\pdfoutput=1

\documentclass{article}

\usepackage[main, final]{neurips_2026}

\usepackage[utf8]{inputenc}
\usepackage[T1]{fontenc}
\usepackage{hyperref}
\hypersetup{
  pdftitle={3D Point Splatting for mmWave Radar Novel View Synthesis},
  pdfauthor={Adnan Armouti, Yixuan Gao, Rajalakshmi Nandakumar}
}
\usepackage{url}
\usepackage{booktabs}
\usepackage{amsfonts}
\usepackage{amsmath}
\usepackage{amssymb}
\usepackage{nicefrac}
\usepackage{microtype}
\usepackage{xcolor}
\usepackage{graphicx}
\usepackage{orcidlink}         %
\usepackage{multirow}
\usepackage{array}                  %
\usepackage{enumitem}
\usepackage{etoc}
\usepackage{placeins}               %

\definecolor{lightgreen}{HTML}{D5F5E3}
\definecolor{lightyellow}{HTML}{FEF9E7}
\definecolor{lightred}{HTML}{FADBD8}
\definecolor{lightergray}{HTML}{F2F3F4}

\usepackage{colortbl}
\definecolor{magbg}{HTML}{FFF1A8}     %
\definecolor{vrbg}{HTML}{B5DCEC}      %
\definecolor{mediumgray}{HTML}{D5D8DC}

\title{3D Point Splatting for mmWave Radar Novel View Synthesis}

\author{%
  Adnan Armouti\,\orcidlink{0000-0002-4448-3089} \\
  Cornell Tech \\
  New York, NY, USA \\
  \texttt{aa2546@cornell.edu} \\
  \And
  Yixuan Gao\,\orcidlink{0000-0003-1778-3104} \\
  Cornell Tech \\
  New York, NY, USA \\
  \texttt{yg478@cornell.edu} \\
  \And
  Rajalakshmi Nandakumar\,\orcidlink{0000-0002-1601-148X} \\
  Cornell Tech \\
  New York, NY, USA \\
  \texttt{rn283@cornell.edu} \\
}

\begin{document}
\etocsettocdepth.toc{none}

\maketitle

\begin{abstract}
Solving novel view synthesis (NVS) for millimeter-wave (mmWave) radar requires a renderer that is physically faithful, complex-valued, and multi-viewpoint-tractable. No prior method achieves these three properties simultaneously. Differentiable Monte Carlo (MC) ray tracers implement the radar forward model directly with explicit material modeling and complex outputs, but a joint fit across views takes over an hour per scene. Optical-NVS ports of NeRF, hash grids, and 3D Gaussians train fast but discard phase and replace explicit material modeling with opaque learned features, restricting them to power-only range--azimuth (RA) magnitudes. We propose 3D Point Splatting (3DPS), the first differentiable point splatting renderer for radar, derived directly from the standard solid-angle form of the radar equation. Each oriented 3D point carries a material model built on ITU-R~P.2040 permittivity and slab reflection, evaluated in closed form, with the resulting complex phasor splatted into range bins through a precomputed point spread function (PSF). The complex-valued output makes the renderer product-agnostic. The same optimized scene yields analog-to-digital converter (ADC), complex range profile (CRP), and RA outputs through standard fast Fourier transform (FFT) pipelines without retraining for each format. On six outdoor ColoRadar scenes, 3DPS reaches \textbf{0.587} mean Pearson correlation on held-out RA images. This is between $1.7\times$ and $5.2\times$ the three optical-NVS baselines (RadarSplat, Radar Fields, DART) and $1.3\times$ the mmIR ray tracer at $1/25$ of its wall-clock time. Training takes approximately 3 minutes per scene on a single RTX~4090.
Project page: \url{https://3d-point-splatting.github.io}.
\end{abstract}

\section{Introduction}
\label{sec:intro}

\begin{figure}[t]
  \centering
  \includegraphics[width=\linewidth]{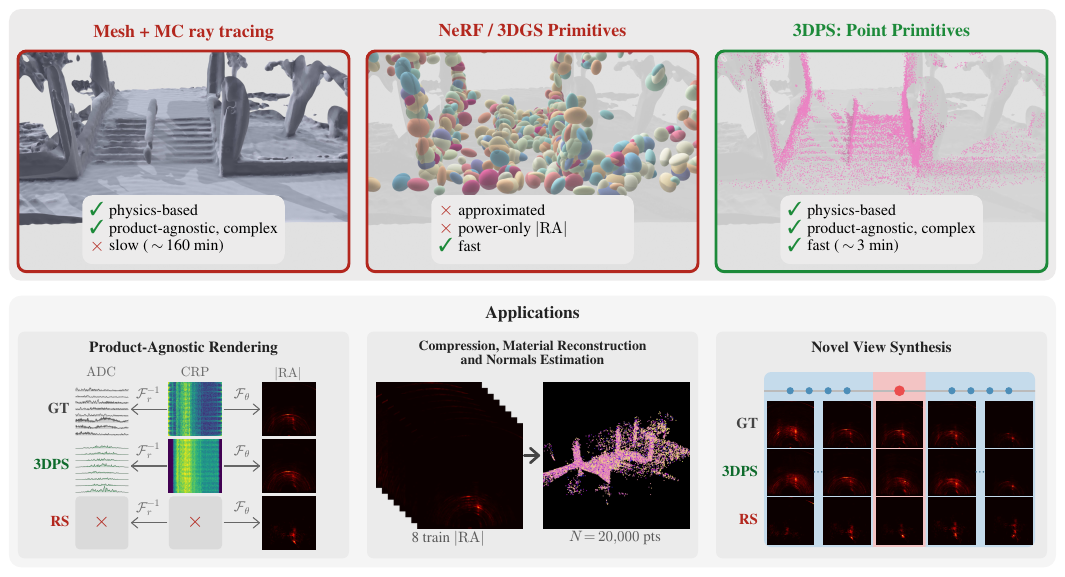}
  \caption{\textbf{The right primitive for radar NVS.} Top: three method families on the three required properties. MC ray-tracing renderers (e.g., mmIR, Sionna RT) are physics-faithful but slow. Optical-NVS adaptations (RadarSplat, Radar Fields, DART) are fast but discard phase. 3DPS satisfies all three. Bottom: 3DPS outputs on S1\,F185 ($|\mathrm{RA}|$, raw ADC magnitude, point cloud colored by $\varepsilon_r'$).}
  \label{fig:teaser}
\end{figure}

Millimeter-wave (mmWave) radar penetrates fog, rain, and airborne dust~\cite{guan2020through,bijelic2020seeing}, returns Doppler velocity directly, and is an essential sensor for robust perception in autonomous driving and robotics~\cite{gao2019experiments,harlow2024mmwave} and for contactless health sensing~\cite{vilesov2022blending,delregno2024thermalimagingradarremote}. However, radar datasets are substantially smaller than their optical or LiDAR counterparts. Radar sensors are not ubiquitous, the captured data structure depends heavily on array geometry and application-specific waveform configuration, and most large-scale datasets remain proprietary. NVS for radar can address this data scarcity by enabling sparse training sets to generalize across novel poses, but this problem is structurally hard. The supervision signal is typically i) sparse with returns dominated by quasi-mirror specular reflections at 60--77\,GHz, ii) constrained to 2D as commercial off-the-shelf (COTS) radar array geometries lack a useful elevation aperture and integrate out the elevation dimension, and iii) limited in view count as NVS aggregates only a handful of training views per scene.

Achieving high-fidelity radar NVS requires a renderer with three properties. First, physically faithful, so that predictions at unseen poses match the underlying sensor physics and the optimized scene parameters remain interpretable, editable, and transferable across sensor views. Second, complex-valued, since the radar forward model emits a complex phasor that is product-agnostic. The same optimized scene yields ADC, CRP, and RA outputs at novel viewpoints without retraining (Fig.~\ref{fig:teaser}). This also allows the renderer to reproduce antenna patterns, sidelobes, and FFT windowing from the signal model, sharpening RA correctness. Phase-discarding renderers forfeit both. Third, multi-viewpoint-tractable, since NVS by definition demands joint optimization across many training poses.

No prior method has all three properties. Differentiable Monte Carlo (MC) ray tracers~\cite{sionna, hofmann2025inverse, chen2025invertwin} implement the radar forward model with explicit bidirectional scattering distribution function (BSDF) material modeling, real antenna beam patterns, and complex outputs. The cost is structural, since each surface hit must be evaluated against every transmitter--receiver (TX, RX) pair to maintain multiple-input multiple-output (MIMO) coherence. A joint eight-viewpoint fit with the mmIR ray tracer~\cite{mmIR2026} takes 77.6 minutes per scene at our 500-iteration budget (Sec.~\ref{app:runtime}). Optical-NVS adaptations~\cite{huang2024dart, 10.1145/3641519.3657510, takawale2025spinr, kung2025radarsplat} of neural radiance fields, hash grids, and 3D Gaussians~\cite{mildenhall2021nerf, 3dgs} train fast on real radar captures, because they trade explicit modeling of the radar signal chain for runtime. On real captures they supervise on post-FFT magnitudes, so phase is dropped by construction, and scattering is carried by learned features rather than explicit materials.

3DPS realizes all three properties because we derive our closed-form, deterministic, and differentiable pipeline directly from the standard solid-angle form of the radar equation~\cite{skolnik2001introduction}. We initialize an oriented 3D point set from a co-registered LiDAR point cloud (Sec.~\ref{sec:methods:impl}, see Scope below). For physical fidelity, the antenna geometry, BSDF (per ITU-R~P.2040), and antenna gains are evaluated in closed form with no learned features, and phase is integrated from the geometric path length rather than learned. Each oriented 3D point contributes exactly once per (TX, RX) pair, with no Monte Carlo sampling. For complex-valued output, the per-(TX, RX) range profile is preserved end-to-end through PSF splatting, enabling product-agnostic rendering to ADC, CRP, and RA without retraining via standard FFT pipelines. For multi-viewpoint tractability, we use a Hann-windowed PSF splat at constant cost per (point, range bin) and fuse BSDF evaluation, splatting, and the analytical backward into three CUDA kernels. Across six outdoor ColoRadar~\cite{kramer2022coloradar} scenes, 3DPS reaches \textbf{0.587} mean Pearson correlation on held-out novel-view $|\mathrm{RA}|$, between $1.7\times$ and $5.2\times$ the three optical-NVS baselines and $1.3\times$ mmIR, in approximately 3 minutes per scene on a single RTX~4090.

\textbf{Contributions.} \textbf{(i)} The first differentiable point splatting renderer for radar, with a radar-native primitive (oriented 3D points carrying ITU-R~P.2040 material parameters) and multi-viewpoint-tractable rendering operation (PSF splatting of complex phasors into range bins), derived from the solid-angle form of the radar equation. \textbf{(ii)} Product-agnostic rendering, since the same optimized scene yields CRP, ADC, and RA outputs through standard FFT pipelines without retraining. \textbf{(iii)} Sparse-view radar NVS at the state of the art held-out correlation on $|\mathrm{RA}|$ and $|\mathrm{CRP}|$ across six ColoRadar scenes, in approximately 3 minutes per scene.

\textbf{Scope.} \textbf{(i) LiDAR initialization.} 3DPS initializes the oriented 3D point set from a co-registered LiDAR point cloud, following the LiDAR-derived geometric prior described in Sec.~\ref{sec:methods:impl}. LiDAR initialization is common in 3DGS for driving scenes~\cite{yan2024streetgaussians}. A 3D prior is necessary because the supervision signal suffers from cascading sparsity: a sparse set of training views, sparse radar returns per view, and elevation integrated out by 2D RA. Together these compound to leave 3D scene structure unrecoverable from RA-only supervision in the sparse-view setting NVS demands. LiDAR-free 3D recovery from radar (shape-from-radar) is an active open problem outside our scope. \textbf{(ii) Single-bounce path tracing.} 3DPS evaluates a single TX$\to$scatterer$\to$RX path per oriented point, matching the single-bounce assumption in RadarSplat~\cite{kung2025radarsplat}.

\section{Related Works}
\label{sec:related}

\begin{table}[t]
\centering
\caption{\textbf{Comparison of radar rendering methods.} Signal types: analog-to-digital converter samples (ADC), channel impulse response (CIR), intermediate frequency (IF), range profile (RP), range--azimuth (RA), range--Doppler--antenna (RDA). Columns mark scene representation (Repr.), explicit material model (Mat.), per-point or per-vertex storage (Per-pt), real-data validation (Real), and wall-clock on one RTX~4090 over 8 views (Time). mmIR is the multi-view extension in Sec.~\ref{app:mmir_baseline}.}
\label{tab:comparison}
\small
\setlength{\tabcolsep}{3pt}
\renewcommand{\arraystretch}{1.1}
\resizebox{\linewidth}{!}{
\begin{tabular}{lcccccccccc}
\toprule
\textbf{Method} & \textbf{Repr.} & \textbf{Signal} & \textbf{Complex} & \textbf{Diff.} & \textbf{Mat.} & \textbf{Per-pt} & \textbf{BSDF} & \textbf{NVS} & \textbf{Real} & \textbf{Time} \\
\midrule
Sionna RT~\cite{sionna}              & Mesh     & CIR & \checkmark & Partial    & \checkmark & --         & \checkmark & --         & --         & --       \\
SFCW Inv.~\cite{hofmann2025inverse}  & Mesh     & IF  & \checkmark & Partial    & \checkmark & \checkmark & \checkmark & --         & \checkmark & --       \\
InverTwin~\cite{chen2025invertwin}   & Mesh     & CIR, RP & \checkmark & \checkmark & \checkmark & --         & --         & --         & \checkmark & --       \\
mmIR~\cite{mmIR2026}                & Mesh     & ADC, CRP, RA & \checkmark & \checkmark & \checkmark & \checkmark & \checkmark & \checkmark & \checkmark & 78 min   \\
\midrule
DART~\cite{huang2024dart}            & Implicit & RDA & --         & \checkmark & --         & --         & --         & \checkmark & \checkmark & 0.4 min  \\
Radar Fields~\cite{10.1145/3641519.3657510} & Implicit & RA & --   & \checkmark & --         & --         & --         & \checkmark & \checkmark & 1.5 min  \\
SpINR~\cite{takawale2025spinr}       & Implicit & Beat  & \checkmark & \checkmark & --         & --         & --         & \checkmark & --         & --       \\
RadarSplat~\cite{kung2025radarsplat} & 3DGS     & RA  & --         & \checkmark & --         & --         & --         & \checkmark & \checkmark & 20 min   \\
\midrule
\textbf{Ours}                        & Points   & ADC, CRP, RA & \checkmark & \checkmark & \checkmark & \checkmark & \checkmark & \checkmark & \checkmark & 3 min    \\
\bottomrule
\end{tabular}
}
\end{table}

\textbf{Differentiable Monte Carlo (MC) ray tracers.}
Building on the differentiable-rendering toolkit pioneered for graphics --- edge-sampling gradients~\cite{li2018differentiable}, reparameterized boundary integrals~\cite{loubet2019reparameterizing}, unbiased warped-area sampling~\cite{bangaru2020unbiased}, and the broader differentiable-rendering survey by Kato et al.~\cite{kato2020differentiable} --- recent radar-domain ray tracers extend the same machinery to electromagnetic propagation: Sionna RT~\cite{sionna}, Hofmann et al.~\cite{hofmann2025inverse}, InverTwin~\cite{chen2025invertwin}, and the deterministic MIMO ray tracer of Sch{\"u}{\ss}ler et al.~\cite{9533181} variously suspend gradients through their path solvers, do not model wideband frequency-modulated continuous-wave (FMCW) radar, assume known geometry, or validate only on synthetic data. Earlier mmWave FMCW ray-tracing simulators~\cite{dudek2010millimeter} establish the methodology but lack differentiability. SAR-domain inverse rendering~\cite{wei2024csvbsdf} pursues a similar materials-fitting agenda but on synthetic-aperture geometry rather than real-aperture FMCW. This family is physically faithful and complex-valued but slow at the multi-viewpoint optimization NVS demands: mmIR~\cite{mmIR2026} takes 77.6 minutes per scene for the 8-viewpoint fit (Sec.~\ref{app:runtime}), against 3 minutes for 3DPS. 3DPS preserves the physical fidelity and complex output of MC ray tracers while replacing per-iteration MC sampling with deterministic PSF splatting at constant cost per (point, range bin).

\textbf{Optical-NVS adaptations.}
Following neural radiance fields~\cite{mildenhall2021nerf} and 3D Gaussian Splatting~\cite{3dgs} (with hash-grid acceleration~\cite{muller2022instantngp}, point-based light-field variants~\cite{ost2022neuralpoint}, and subsequent surfel-style refinements~\cite{jiang2025surfels}), DART~\cite{huang2024dart} trains a NeRF on range--Doppler magnitude, Radar Fields~\cite{10.1145/3641519.3657510} synthesizes power-only RA via a hash-grid and multilayer perceptron (MLP), and RadarSplat~\cite{kung2025radarsplat} ports 3DGS to RA rendering with opaque per-Gaussian features. Other neural-field adaptations~\cite{takawale2025spinr,zhang2025rf4d,rafidashti2025neuradar,lu2024newrf,farrell2023coir} extend to dynamic scenes or fuse radar with other modalities, but render no complex output. The same neural-field machinery has been ported to other coherent sensors: SAR~\cite{lei2024sarnerf,ehret2024radarfieldssar} and ISAR~\cite{deng2024isarnerf} for radar-like apertures, LiDAR for point-cloud novel views~\cite{tao2024lidarnerf,huang2023neurallidar}, sonar via implicit fields~\cite{reed2023sas} and Gaussian splatting~\cite{qu2024zsplat,sethuraman2025sonarsplat}, and physically-based scattering NeRFs for adverse-weather media~\cite{ramazzina2023scatternerf}. This family is multi-viewpoint-tractable but neither physically faithful nor complex-valued. The three methods in our benchmark supervise on post-FFT magnitudes and carry scattering in learned features without explicit materials. 3DPS preserves multi-view tractability while adding explicit ITU-R~P.2040 BSDF material modeling, real antenna beam patterns, and per-(TX, RX) complex range profile output.

\section{Method}
\label{sec:methods}

3DPS renders complex-valued FMCW range profiles from a learned set of \emph{oriented 3D points}, anchored to a LiDAR-derived geometric scaffold. Each point carries a quaternion-encoded surface normal and a six-parameter ITU-R~P.2040 material vector, evaluated by a closed-form bidirectional scattering distribution function (BSDF) and splatted into range-profile bins through a precomputed sensor point spread function (PSF). Sec.~\ref{sec:methods:fwd} states the forward model from the radar equation through the hemisphere integral to a deterministic discrete sum and its splatted form, plus the per-point BSDF. Sec.~\ref{sec:methods:pipeline} walks through the rendering pipeline, justifies the choice of primitive, and ties the design back to the three requirements in Sec.~\ref{sec:intro}. Sec.~\ref{sec:methods:impl} details optimization and implementation.

\begin{figure}[t]
  \centering
  \includegraphics[width=\linewidth]{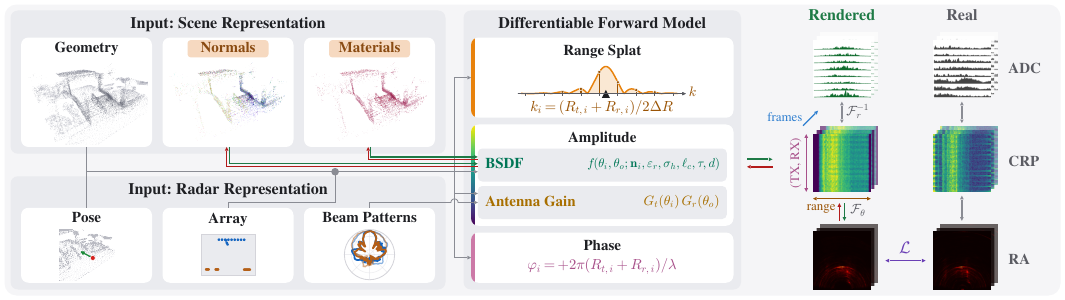}
  \caption{\textbf{3DPS overview.} A scene is a set of $N{=}20{,}000$ oriented points seeded from LiDAR, each carrying a normal and a material. For each (TX, RX) pair, closed-form gain, BSDF and phase give a complex amplitude (Eqs.~\ref{eq:radar_eqn},~\ref{eq:bsdf}), which a Hann-windowed PSF splats into the $L{=}15$ nearest range bins (Eq.~\ref{eq:crp_splat}). The loss compares $|\mathrm{RA}|$ with the measurement; ADC and CRP come from the same render.}
  \label{fig:pipeline}
\end{figure}

\subsection{Radar forward model}
\label{sec:methods:fwd}

Consider an FMCW radar with a MIMO array of $N_{\text{TX}}$ transmitters and $N_{\text{RX}}$ receivers, chirp sweep slope $\mu$, carrier wavenumber $k=2\pi/\lambda$, and per-chirp ADC sampling $\{t_n = n/f_s\}_{n=0}^{N_s-1}$. A scene is illuminated by transmitter $t$ at $\mathbf{p}_t$ and observed by receiver $r$ at $\mathbf{p}_r$. We let $R_t(\mathbf{x})=\|\mathbf{x}-\mathbf{p}_t\|$, $R_r(\mathbf{x})=\|\mathbf{x}-\mathbf{p}_r\|$, $\hat{\boldsymbol{\omega}}_i$ the incident direction (TX$\to\mathbf{x}$), and $\hat{\boldsymbol{\omega}}_o$ the scattered direction ($\mathbf{x}\to$RX).

\textbf{Rendering equation.} Following prior radar inverse rendering~\cite{mmIR2026}, we take the bistatic radar equation~\cite{skolnik2001introduction}, replace the radar cross-section by a spatially varying BSDF $f$ and integrate over the receiver-centered hemisphere $\Omega^{+}(\mathbf{p}_r)$, giving the received power (full derivation in Sec.~\ref{app:fwd_derivation})
\begin{equation}
P_r \;=\; \frac{P_t\,\lambda^2}{(4\pi)^2}
\int_{\Omega^{+}(\mathbf{p}_r)}
\frac{G_t(\hat{\boldsymbol{\omega}}_i)\,G_r(\hat{\boldsymbol{\omega}}_o)\,
f(\hat{\boldsymbol{\omega}}_i,\mathbf{x},\hat{\boldsymbol{\omega}}_o)\,\cos\theta_i\,V(\mathbf{x})}
{R_t^2(\mathbf{x})}\;\mathrm{d}\omega,
\label{eq:rendering_eqn}
\end{equation}
where $G_t,G_r$ are the antenna power gain patterns, $\cos\theta_i=|\mathbf{n}\!\cdot\!\hat{\boldsymbol{\omega}}_i|$, $V\in\{0,1\}$ is visibility and $\mathbf{x}(\hat{\boldsymbol{\omega}})$ is the first scene intersection along $\hat{\boldsymbol{\omega}}$ from $\mathbf{p}_r$.

\textbf{Discretization on an oriented point set.} We discretize $\Omega^{+}(\mathbf{p}_r)$ on $N$ oriented points $\{(\mathbf{x}_i,\mathbf{n}_i,\boldsymbol{\theta}_i)\}_{i=1}^{N}$, where $\boldsymbol{\theta}_i$ is the per-point material vector. Because Eq.~\eqref{eq:rendering_eqn} integrates over solid angle, the surface Jacobian has already cancelled and the points need no per-point area. A Riemann sum with uniform solid-angle weight $\Delta\omega$ leaves a closed-form, per-point complex amplitude with the carrier phase attached,
\begin{equation}
S_{tr}(\mathbf{x}_i) \;=\; C\,
\frac{\sqrt{G_t(\hat{\boldsymbol{\omega}}_{i,i})\,G_r(\hat{\boldsymbol{\omega}}_{o,i})}}{R_{t,i}}\,
\sqrt{f_i\cos\theta_{i}}\;
e^{\,jk\,(R_{t,i}+R_{r,i})},
\label{eq:radar_eqn}
\end{equation}
with $C=\frac{\lambda}{4\pi}\sqrt{P_t\,\Delta\omega}$, which in practice we treat as a constant calibrated to the measured level. The dechirped ADC sample is the closed-form, deterministic single-bounce sum
\begin{equation}
\hat{a}_{tr}[n] \;=\; \sum_{i=1}^{N}
v_i\, S_{tr}(\mathbf{x}_i)\,
e^{\,j 2\pi\, \mu\, \tau_i\, t_n},
\qquad \tau_i \;=\; \tfrac{R_{t,i}+R_{r,i}}{c},
\label{eq:adc_discrete}
\end{equation}
where $v_i\in\{0,1\}$ is the visibility of point $i$ (Sec.~\ref{app:fwd_derivation}). Each point contributes exactly once per (TX, RX) pair, with no Monte Carlo sampling or path enumeration.

\textbf{CRP via Hann-windowed PSF splat.} The complex range profile is the windowed range FFT of the ADC, $\mathrm{CRP}_{tr}[k] = \mathcal{F}_n\{w[n]\,a_{tr}[n]\}$. Linearity of Eq.~\eqref{eq:adc_discrete} lets us push the FFT inside the sum, so each point becomes a known kernel centered at its (fractional) range bin
\begin{equation}
\mathrm{CRP}_{tr}[k] \;=\; \sum_{i=1}^{N} v_i\, S_{tr}(\mathbf{x}_i)\,\Phi\!\bigl(k - k_i;\, \delta_i\bigr),
\qquad k_i \;=\; \tfrac{R_{t,i}+R_{r,i}}{2\Delta R},\;\; \delta_i \;=\; k_i - \lfloor k_i\rfloor,
\label{eq:crp_splat}
\end{equation}
where $\Delta R$ is the range bin width and $\Phi(\cdot;\delta)$ is the precomputed Hann-window PSF (length $L\!=\!15$ taps). Splatting replaces the per-sample chirp synthesis of Eq.~\eqref{eq:adc_discrete} with a single scatter-add into $L$ adjacent range bins per point, dropping the per-pair cost from $\mathcal{O}(N\!\cdot\!N_s)$ to $\mathcal{O}(N\!\cdot\!L)$. Validity bounds (Hann-tail truncation, fractional-bin handling, numerical equivalence to the standard ADC-then-FFT pipeline) are quantified in Sec.~\ref{app:splat_validity}.

\textbf{Per-point BSDF.} Each point's BSDF blends a specular and a scattering lobe, each weighted by reflection from an ITU-R~P.2040~\cite{itu_p2040} slab:
\begin{equation}
f_i(\hat{\boldsymbol{\omega}}_i,\hat{\boldsymbol{\omega}}_o) \;=\;
\tau_{\text{eff}}\, R_{h}\, f_{\text{KA}}(\hat{\boldsymbol{\omega}}_i,\hat{\boldsymbol{\omega}}_o;\mathbf{n}_i,\sigma_h)
+
(1-\tau_{\text{eff}})\, R_{n}\, f_{\text{SPM}}(\hat{\boldsymbol{\omega}}_i,\hat{\boldsymbol{\omega}}_o;\mathbf{n}_i,\sigma_h,\ell_c),
\label{eq:bsdf}
\end{equation}
with all material parameters drawn from the per-point vector $\boldsymbol{\theta}_i=(\varepsilon'_r,\varepsilon''_r,\sigma_h,\ell_c,\tau,d)_i$ (subscript $i$ suppressed inside the equation for compactness). $f_{\text{KA}}$ is a Kirchhoff-approximation (KA) GGX microfacet lobe and $f_{\text{SPM}}$ a small-perturbation-method (SPM) von Mises--Fisher lobe about the mirror direction. $R_{h}$ and $R_{n}$ are slab reflection powers at the half vector and at the mean normal, parameterized by complex permittivity $\varepsilon_r=\varepsilon'_r-j\varepsilon''_r$ and slab thickness $d$. $\tau_{\text{eff}}$ blends the learned $\tau$ with lobe validity at the current geometry. The full closed form and parameter ranges are in Sec.~\ref{app:bsdf_full}.

\subsection{Rendering pipeline}
\label{sec:methods:pipeline}

Figure~\ref{fig:pipeline} traces a single forward pass. Input: an oriented 3D point set $\{(\mathbf{x}_i,\mathbf{n}_i,\boldsymbol{\theta}_i)\}_{i=1}^{N}$ with $N{=}20{,}000$ (Sec.~\ref{sec:methods:impl}). For each (TX, RX) pair the renderer (i)~computes per-point geometry $R_{t,i}, R_{r,i}, \hat{\boldsymbol{\omega}}_{i,i}, \hat{\boldsymbol{\omega}}_{o,i}$ and the visibility $v_i$; (ii)~evaluates the per-point BSDF $f_i$ from $\boldsymbol{\theta}_i$ via Eq.~\eqref{eq:bsdf}; (iii)~assembles the per-(TX, RX) complex amplitude $S_{tr}(\mathbf{x}_i)$ of Eq.~\eqref{eq:radar_eqn}; and (iv)~scatter-adds each point's contribution into the $L{=}15$ nearest range bins of $\mathrm{CRP}_{tr}[k]$ via the Hann PSF kernel $\Phi$.

The output of the splat is a set of $N_{\text{TX}}\!\times\!N_{\text{RX}}$ complex range profiles. Stacking these into the virtual-array dimension and applying a Hann window and a 128-point azimuth discrete Fourier transform (DFT) across the 86 elements of the azimuth row, then dropping the DC bin, yields the range--azimuth (RA) map of shape $(N_{\text{az}},K) = (127, 256)$, with arcsin-spaced azimuth bins $\sin\theta_a = (2a-N_{\text{az}}+1)/(N_{\text{az}}+1)$ and forward-cone half-angle $\arcsin(126/128) \approx 79.86^\circ$. ADC samples are recovered by inverse range FFT of $\mathrm{CRP}$. The same optimized scene therefore yields all three downstream products (ADC, CRP, and RA) through standard FFT pipelines without retraining.

\textbf{Choice of primitive.} The rendering equation (Eq.~\ref{eq:rendering_eqn}) requires hit positions, surface normals, and material parameters at every scattering site, and it operates on the receiver-centered solid angle, removing any need for attached surface area on the primitives. Oriented 3D points are therefore the radar-native primitive. MC ray tracers (Sionna~RT~\cite{sionna}) recompute these hit points every iteration. We make them persistent and learnable, removing the per-iteration ray-tracing cost. 3DGS~\cite{3dgs} notes that point sample rendering ``suffers from holes, causes aliasing, and is strictly discontinuous,'' and that splatting primitives with spatial extent addresses this. Radar resolution is an order of magnitude coarser at $\Delta R{=}5.93$~cm range and ${\sim}1.4^\circ$ azimuth, and each point splats a continuous $L{=}15$-tap Hann PSF over ${\sim}89$~cm of range. Those arguments do not apply.

\textbf{Choosing $N$ and $L$.} $L$ follows from the waveform and window, not the scene: beyond 15 taps the Hann kernel holds under $-59$\,dB of its energy. $N$ follows from the sensor's resolvable cells, not scene volume, because initialization keeps only points inside the $[1.5, 15.2]$\,m shell and the azimuth cone. $N{=}20{,}000$ gives about 0.6 points per cell over all $127\times256$ cells and 1.6 over range bins 15--110. A larger scene at the same waveform therefore needs no more points, while more range bins or a denser virtual array would. Held-out correlation is 0.586 at 10k, 0.587 at 20k and 0.538 at 50k (Sec.~\ref{app:ablations}), and the indoor scenes use the same $N$ and $L$ (Sec.~\ref{app:per_scene_ra}).

\textbf{Realizing the three properties (Sec.~\ref{sec:intro}).} \textbf{Physically faithful:} antenna gains, the ITU-R~P.2040 BSDF, and the carrier phase $e^{jk(R_{t,i}+R_{r,i})}$ are all evaluated in closed form from geometric path lengths, so the optimized $\boldsymbol{\theta}_i$ and $\mathbf{n}_i$ remain physically interpretable, editable, and transferable across sensor views. \textbf{Complex-valued:} Eq.~\eqref{eq:crp_splat} preserves the per-(TX, RX) complex range profile as a first-class output. The same optimized scene synthesizes ADC, CRP, and RA at novel viewpoints without retraining. \textbf{Multi-viewpoint-tractable:} each oriented point contributes exactly once per (TX, RX) pair with no Monte Carlo sampling or path enumeration. The PSF splat has fixed cost $\mathcal{O}(N\,L)$ per pair, and the fused CUDA implementation (Sec.~\ref{sec:methods:impl}) trains across 8 viewpoints jointly in approximately 3 minutes per scene on a single RTX~4090.

\subsection{Optimization and implementation}
\label{sec:methods:impl}

\textbf{LiDAR-derived geometric prior.} A 3D geometric prior is necessary for the cascading-sparsity reasons in Sec.~\ref{sec:intro}. We initialize the oriented point set from a co-registered LiDAR cloud of between 1.8 and 3.0 million raw points across our scenes, reducing to $N{=}20{,}000$ via four stages: frustum and range culling, occlusion ray-casting against the scene with Mitsuba~3~\cite{jakob2022dr}, cosine-weighted resampling biased toward array boresight, and farthest-point sampling. Each surviving point inherits a quaternion-encoded LiDAR surface normal and a uniform ITU-concrete material prior. Perturbing seed normals by up to $20^\circ$ leaves held-out correlation within 0.006 of the default (Sec.~\ref{app:ablations}). Full reduction-stage parameters are in Sec.~\ref{app:lidar_reduction}.

\textbf{Learned parameters and carrier-phase detach.} For each point we learn the 6-parameter ITU material vector $\boldsymbol{\theta}_i$, the surface-normal quaternion $\mathbf{q}_i$ (initialized from the LiDAR normal), and the 3D position $\mathbf{p}_i$ through the \emph{amplitude path only}. The carrier-phase term $e^{jk(R_{t,i}+R_{r,i})}$ in $S_{tr}(\mathbf{x}_i)$ is detached from position gradients because a sub-millimeter step rotates the phase by $\mathcal{O}(1)$ at this wavelength, which makes the loss locally periodic in position. The forward-pass carrier phase is still computed from the unrounded $R_{t,i}+R_{r,i}$. An L2 anchor $\lambda_{\text{pos}}\|\mathbf{p}_i - \mathbf{p}_i^{(0)}\|^2$ ($\lambda_{\text{pos}}{=}100$) keeps the median point within about 1 mm of the LiDAR cloud, at carrier wavelength $\lambda \approx 3.9$ mm at 77 GHz.

\textbf{Optimization and CUDA implementation.} The loss is per-pose MSE between $|\mathrm{RA}_F^{\text{pred}}|$ and $|\mathrm{RA}_F^{\text{GT}}|$, divided by the squared mean of $|\mathrm{RA}_F^{\text{GT}}|$, optimized with Adam ($\beta_1{=}0.9$, $\beta_2{=}0.999$) at per-parameter learning rates $\eta_{\text{mat}}{=}10^{-2}$, $\eta_{\text{rot}}{=}5{\cdot}10^{-3}$, $\eta_{\text{pos}}{=}10^{-5}$. Learning rates follow a cosine decay from iteration 200, and gradients are RMS-clipped per parameter. Following 3DGS~\cite{3dgs} but adapted to radar physics, we split the top 5\% and prune the bottom 5\% of points by accumulated amplitude-path position-gradient magnitude $g_i = \sum_F \|\nabla_{\mathbf{p}_i} \mathcal{L}_F\|$ at iterations $\{100,200,300,400\}$, preserving the budget $N{=}20{,}000$. Each split child inherits its parent's parameters with Gaussian noise (2 cm on position), so 16--20\% of final points lie over 5 mm from the LiDAR cloud. Eqs.~\eqref{eq:crp_splat}--\eqref{eq:bsdf} plus the virtual-array DFT run as three fused CUDA kernels (BSDF, scatter-splat, and analytical backward). The fused forward pass renders all $N{=}20{,}000$ points across 192 (TX, RX) pairs in approximately 5 ms per viewpoint on an RTX~4090, and a full training step takes approximately 30 ms. A 500-iteration 8-viewpoint training pass completes in approximately 3 minutes per scene.

\section{Experiments}
\label{sec:experiments}

We evaluate 3DPS on six outdoor ColoRadar~\cite{kramer2022coloradar} scenes captured with a TI MMWCAS cascaded mmWave radar (12\,TX $\times$ 16\,RX, 77\,GHz carrier, 5.93\,cm range resolution) and a co-registered Ouster OS1-64 LiDAR.

\textbf{Why ColoRadar.} 3DPS requires four inputs: complete radar array geometry, raw I/Q ADC samples, co-registered LiDAR, and dense pose annotations. ColoRadar provides all four. Other automotive radar datasets~\cite{caesar2020nuscenes,schumann2021radarscenes,barnes2020oxford,burnett2023boreas,sheeny2021radiate,paek2022kradar,rebut2022rawhd} either ship pre-processed radar tensors without per-element array geometry, use mechanically-rotated single-channel sensors that do not expose a virtual array, or lack the co-registered LiDAR and dense poses 3DPS requires.

\subsection{Experimental setup}
\label{sec:experiments:setup}

\textbf{Scene splits.} We follow the periodic held-out-frame protocol of RadarSplat~\cite{kung2025radarsplat}, which reserves every fifth frame as the novel-view test frame. Adapted to the cascade's 5\,Hz frame rate and our short-window (9-frame) experimental setup, this becomes a centred 1-of-9 variant: for each scene we select 9 adjacent cascaded radar frames at 5\,Hz (200\,ms inter-frame spacing) and use only the first chirp loop of each frame. The middle frame $F$ is the held-out test frame and the eight bracketing frames $\{F\!\pm\!1,\ldots,\!\pm\!4\}$ form the training set, giving 8 training and 1 held-out test RA map per scene.

\textbf{Baselines.} Three published radar NVS methods from the optical-NVS family (full descriptions in Sec.~\ref{sec:related}): \textbf{DART}~\cite{huang2024dart}, \textbf{Radar Fields}~\cite{10.1145/3641519.3657510}, and \textbf{RadarSplat}~\cite{kung2025radarsplat}. We also compare against the \textbf{mmIR} ray tracer~\cite{mmIR2026}, extended to joint multi-view fitting. Only 3DPS and mmIR render complex range profiles, so CRP and ADC columns are reported for those two; the three optical-NVS baselines emit power-only RA by construction.

\textbf{Metrics.} We report Pearson correlation (Corr), peak signal-to-noise ratio (PSNR), structural similarity (SSIM), and root-mean-square error (RMSE), all computed on min-max-normalized $399\!\times\!399$ Cartesian $|\mathrm{RA}|$ images (Sec.~\ref{app:fmcw_primer:cartesian}).

\textbf{Implementation.} 3DPS training takes approximately 3 minutes per scene for joint 8-viewpoint optimization on a single NVIDIA RTX~4090, or 30 to 45 seconds per single-viewpoint fit at the same 500-iteration budget (full details in Sec.~\ref{sec:methods:impl}). Measured baseline wall-clocks on identical hardware are 0.4, 1.5, 19.7, and 77.6 min per scene for DART, Radar Fields, RadarSplat, and mmIR respectively. Per-scene wall-clock and peak GPU memory are in Sec.~\ref{app:runtime}. \textbf{Sample efficiency.} 3DPS is robust at the view axis: dropping from the 8-view default to just 2 bracketing training frames yields \textbf{0.573} mean test Pearson correlation, within $0.014$ of the 8-view default ($0.587$); per-view-count breakdown in Sec.~\ref{app:ablations}. Two adjacent frames are therefore enough when the test view lies between them; wider spacing costs correlation (Sec.~\ref{app:per_scene_ra}).

\subsection{Quantitative results}
\label{sec:experiments:results}

Table~\ref{tab:results} reports the cross-method, cross-product results, averaged over the six ColoRadar scenes for training views (8 per scene) and the held-out novel-view test frame. The three optical-NVS baselines emit power-only $|\mathrm{RA}|$, so their CRP and ADC columns are marked `--'. Per-scene breakdowns and CRP/ADC evaluation methodology are in supplement Sec.~\ref{app:per_scene_ra} and~\ref{app:rdadc}.

\begin{table*}[t]
\centering
\caption{\textbf{Cross-method, cross-product results on six ColoRadar scenes (mean across scenes).} \colorbox{magbg}{Magnitude}: Pearson correlation, PSNR, SSIM, RMSE on min-max-normalized $|\cdot|$. \colorbox{vrbg}{Rel.\ phase}: magnitude-weighted phase coherence $\langle\cos(\Delta\varphi)\rangle$ across range ($\Delta\varphi_r$) and across virtual antennas ($\Delta\varphi_v$); on ADC, $\Delta\varphi_t$ is the differential across fast-time samples. All in $[-1,1]$, random reference 0. These metrics are absolute-phase-invariant by construction so are unaffected by per-VA / per-range calibration drift the magnitude-only training loss does not constrain. \textbf{Bold}: best per column (ties jointly). \underline{Underline}: second-best. Per-scene breakdowns and standard deviations in Tables~\ref{tab:test_ra}--\ref{tab:crp_adc_test}.}
\label{tab:results}
\setlength{\tabcolsep}{2.5pt}
\resizebox{\textwidth}{!}{
\begin{tabular}{l | cccc | ccc | cc | c | c}
\toprule
& \multicolumn{4}{c|}{\textbf{RA cart}} & \multicolumn{5}{c|}{\textbf{CRP}} & \multicolumn{2}{c}{\textbf{ADC}} \\
\cmidrule(lr){2-5}\cmidrule(lr){6-10}\cmidrule(lr){11-12}
& \multicolumn{4}{c|}{\colorbox{magbg}{\textbf{Magnitude}}} & \multicolumn{3}{c|}{\colorbox{magbg}{\textbf{Magnitude}}} & \multicolumn{2}{c|}{\colorbox{vrbg}{\textbf{Rel.\ phase}}} & \colorbox{magbg}{\textbf{Env.}} & \colorbox{vrbg}{\textbf{Rel.\ phase}} \\
Method & Corr & PSNR & SSIM & RMSE & Corr & PSNR & SSIM & $\Delta\varphi_r$ & $\Delta\varphi_v$ & $\rho_{|\cdot|}$ & $\Delta\varphi_t$ \\
\midrule
\multicolumn{12}{l}{\textbf{Train Mean}} \\
DART~\cite{huang2024dart} & 0.033 & 23.9 & 0.540 & 0.0673 & -- & -- & -- & -- & -- & -- & -- \\
Radar Fields~\cite{10.1145/3641519.3657510} & 0.218 & 24.9 & 0.480 & 0.0581 & -- & -- & -- & -- & -- & -- & -- \\
RadarSplat~\cite{kung2025radarsplat} & 0.365 & 24.9 & 0.526 & 0.0595 & -- & -- & -- & -- & -- & -- & -- \\
mmIR~\cite{mmIR2026} & \underline{0.570} & \underline{27.9} & \underline{0.752} & \underline{0.0440} & \underline{0.566} & \underline{24.5} & \underline{0.611} & \underline{0.500} & \underline{0.410} & \underline{0.587} & \underline{0.395} \\
\textbf{3DPS (Ours)} & \textbf{0.812} & \textbf{31.9} & \textbf{0.819} & \textbf{0.0272} & \textbf{0.701} & \textbf{27.9} & \textbf{0.721} & \textbf{0.518} & \textbf{0.494} & \textbf{0.689} & \textbf{0.444} \\
\midrule
\multicolumn{12}{l}{\textbf{Test Mean}} \\
DART~\cite{huang2024dart} & 0.112 & 22.5 & 0.408 & 0.0817 & -- & -- & -- & -- & -- & -- & -- \\
Radar Fields~\cite{10.1145/3641519.3657510} & 0.133 & 28.0 & 0.496 & 0.0402 & -- & -- & -- & -- & -- & -- & -- \\
RadarSplat~\cite{kung2025radarsplat} & 0.339 & 24.2 & 0.511 & 0.0643 & -- & -- & -- & -- & -- & -- & -- \\
mmIR~\cite{mmIR2026} & \underline{0.454} & \textbf{28.8} & \underline{0.685} & \textbf{0.0364} & \underline{0.470} & \underline{23.5} & \underline{0.568} & \textbf{0.401} & \underline{0.425} & \textbf{0.627} & \textbf{0.434} \\
\textbf{3DPS (Ours)} & \textbf{0.587} & \textbf{28.8} & \textbf{0.734} & \underline{0.0369} & \textbf{0.603} & \textbf{24.7} & \textbf{0.650} & \underline{0.371} & \textbf{0.434} & \underline{0.613} & \underline{0.414} \\
\bottomrule
\end{tabular}
}
\end{table*}

\textbf{Magnitude on $|\mathrm{RA}|$.} 3DPS achieves \textbf{0.812} mean Pearson correlation on training views and \textbf{0.587} on the held-out test view, on Cartesian $|\mathrm{RA}|$. This is $2.2\!\times$ the next-best optical-NVS baseline on training views and $1.7\!\times$ on the test view. Relative to mmIR, the next-best method on both splits at 0.570 train and 0.454 test, 3DPS reaches $1.4\!\times$ and $1.3\!\times$. Among the three optical-NVS baselines, RadarSplat is best (0.365 train, 0.339 test), then Radar Fields, then DART. On held-out images 3DPS ranks first on Corr and SSIM, and mmIR matches it on PSNR and RMSE (28.8\,dB; 0.0364 vs.\ 0.0369). On training views 3DPS ranks first on all four metrics in every scene (Tables~\ref{tab:test_ra} and~\ref{tab:train_ra}). The ranking matches the structural mismatches identified in Sec.~\ref{sec:related}. RadarSplat opaque per-Gaussian features cannot share information with the radar antenna pattern, capping at 0.37 train and 0.34 test. Radar Fields combines a hash-grid encoder with a multilayer perceptron decoder, supervised on post-FFT magnitudes. Without a virtual-array model it fails to recover scene support from sparse RA alone, reaching only 0.22 train and 0.13 test. DART is designed for $N_d{=}256$ chirps but operates here at the cascade's $N_d{=}16$, leaving insufficient Doppler resolution at 0.03 train and 0.11 test.

\textbf{Comparison with mmIR.} mmIR marginally leads on held-out ADC envelope (0.627 vs.\ 0.613), $\Delta\varphi_r$ (0.401 vs.\ 0.371) and $\Delta\varphi_t$ (0.434 vs.\ 0.414), while 3DPS leads CRP Corr (0.603 vs.\ 0.470) and $\Delta\varphi_v$ (0.434 vs.\ 0.425). An eight-view mmIR fit takes 77.6 minutes against 3 minutes for 3DPS.

\textbf{Train-test gap.} 3DPS exhibits a 0.225 absolute gap between train and held-out test Pearson correlation. This reflects the cascading sparsity defined in Sec.~\ref{sec:intro}. Outdoor surfaces at 60 to 77 GHz act as quasi-mirrors, so views show little cross-view consistency to regularize the held-out pose. The binding constraint is viewpoint spacing. Training on 2 / 4 / 6 / 8 views gives 0.573 / 0.575 / 0.586 / 0.587 (Table~\ref{tab:ablations}), but all four are subsets of the same $F\pm1\ldots\pm4$ bracket at 5\,Hz and sample identical angular support. The saturation means the bracket is exhausted, not that view count is irrelevant. The cascade's large physical aperture also limits frame rate, leaving inter-viewpoint baselines at 5\,Hz that are not ideal for radar NVS. That is a capture constraint rather than model capacity. Our train correlation of 0.812 marks the ceiling. Held-out correlation falls to 0.442 and 0.419 at two and three times the frame spacing, and 3DPS stays ahead of the optical-NVS baselines at both (Sec.~\ref{app:per_scene_ra}).

\textbf{Per-scene breakdown.} Test correlation ranges from \textbf{0.508} (S2\,F300) to \textbf{0.657} (S2\,F160) across the six scenes (per-scene values: 0.644, 0.519, 0.642, 0.553, 0.657, 0.508 for S0\,F135, S1\,F185, S1\,F438, S2\,F105, S2\,F160, S2\,F300). The strongest scene (S2\,F160) is dominated by extended planar walls that present consistent specular returns across the 8 training views, regularizing the held-out pose well. The weakest scene (S2\,F300) is a tighter staging area with dense pedestrian-scale support structure: many small-feature targets contribute high-frequency RA support that the magnitude-only loss does not constrain at the held-out pose, increasing the train-test gap.

\textbf{Magnitude on CRP and ADC.} The 3DPS native complex range profile gives mean training and test CRP Pearson correlation of \textbf{0.701} and \textbf{0.603}. These numbers are directly comparable to the $|\mathrm{RA}|$ Pearson correlation in the same table. CRP correlation falls below $|\mathrm{RA}|$ on training views (0.701 vs.\ 0.812) and sits level on the held-out view (0.603 vs.\ 0.587). The azimuth FFT provides approximately $\sqrt{86}$ of coherent gain at target directions of arrival, which the per-(v,r) $|\mathrm{CRP}|$ lacks. The ADC envelope $\rho_{|\cdot|}$ reaches \textbf{0.689} train and \textbf{0.613} test. PSNR and SSIM are not reported for ADC because image-domain metrics are inappropriate for raw I/Q time series.

\textbf{Relative-phase fidelity.} The renderer's per-(VA, range) absolute phase is unconstrained by the magnitude-only $|\mathrm{RA}|$ loss --- approximately $22{,}000$ per-bin phase degrees of freedom are free up to per-VA RF calibration drift and per-range ADC sample-zero offsets. Strict complex correlation $|\rho|$ and per-bin phase RMSE penalize that nuisance phase directly, so we instead report the magnitude-weighted phase coherence $\langle\cos(\Delta\varphi)\rangle$ of the \emph{differential} phase across each axis (denoted $\Delta\varphi_r$, $\Delta\varphi_v$, $\Delta\varphi_t$ in Table~\ref{tab:results}). Differential phase across an axis is invariant to any per-bin rotation along the orthogonal axis by construction, so it isolates the relative-phase content that drives downstream products (range-bin localization, beamforming) while ignoring the absolute-phase nuisance subspace the magnitude loss does not cover. The metric lies in $[-1, 1]$ with random reference 0; perfect agreement is $1$. 3DPS reaches mean training and test CRP $\Delta\varphi_r$ of \textbf{0.518} and \textbf{0.371}, $\Delta\varphi_v$ of \textbf{0.494} and \textbf{0.434}, and ADC $\Delta\varphi_t$ of \textbf{0.444} and \textbf{0.414}, which is 0.37--0.43 of perfect on the held-out view with no phase-domain supervision. Subspace DOA on the render is $2.8\times$ to $3.6\times$ more accurate than a zero-phase control (Sec.~\ref{app:doa}). The three optical-NVS baselines emit power-only RA, so the rel.\ phase columns apply only to 3DPS and mmIR.

\begin{figure}[t]
  \centering
  \includegraphics[width=\linewidth]{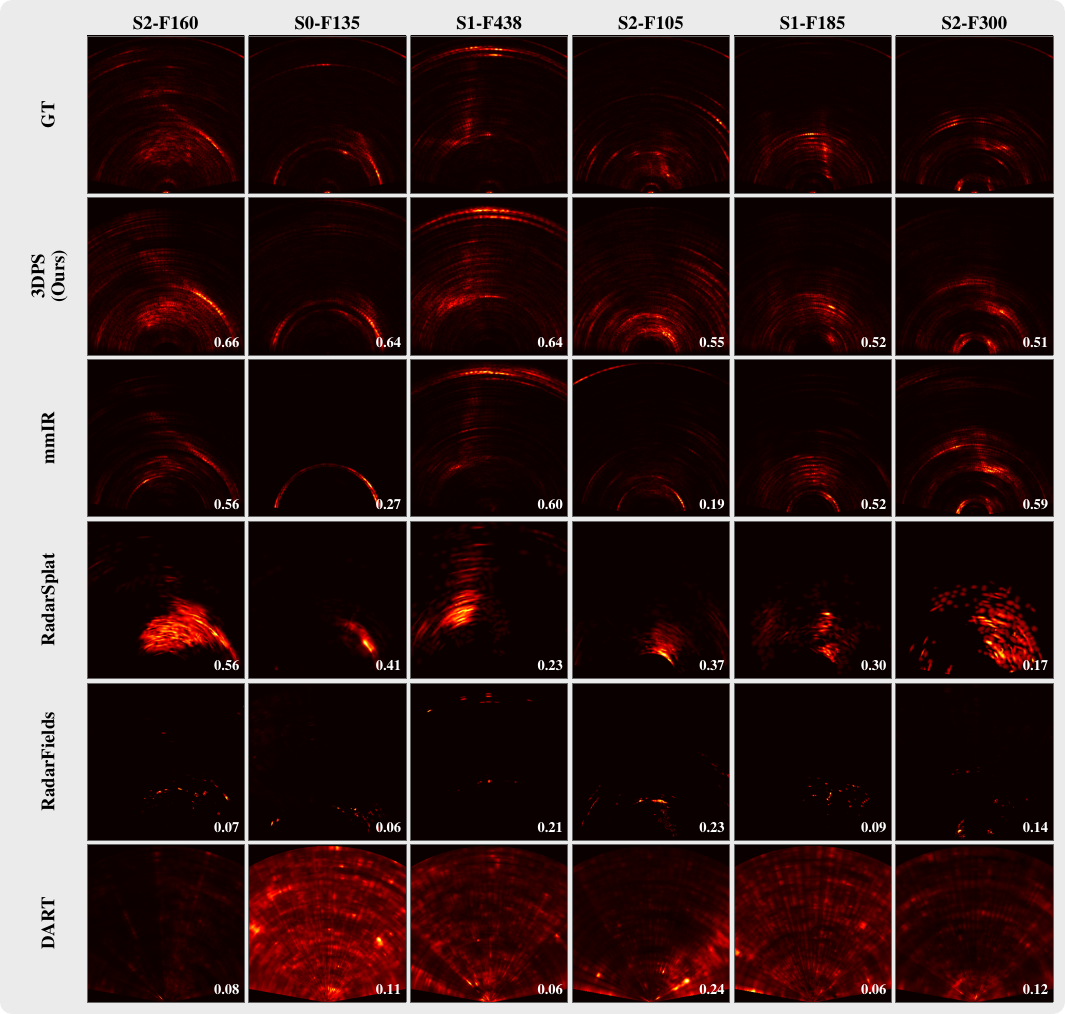}
  \caption{\textbf{Held-out novel-view $|\mathrm{RA}|$ comparison.} Columns: six ColoRadar scenes sorted by 3DPS test Pearson correlation, from 0.657 on S2\,F160 to 0.508 on S2\,F300 (mean 0.587). Rows: GT, 3DPS (Ours), mmIR, RadarSplat, Radar Fields, DART. Test Pearson correlation in white.}
  \label{fig:training_ra}
\end{figure}

\textbf{Qualitative comparison.} Figure~\ref{fig:training_ra} shows held-out test $|\mathrm{RA}|$ renders for 3DPS and the four baselines on the six scenes, sorted left to right by descending 3DPS test Pearson correlation. 3DPS preserves dominant scene structure (walls and support columns) with consistent range and azimuth localization. RadarSplat produces blurred opaque blobs from its per-Gaussian features. Radar Fields under-fits with a near-constant background dotted by sparse hot-spots, the failure mode of post-FFT supervision without a virtual-array model. DART produces noisy, low-amplitude renders dominated by Doppler ambiguity at the cascade's $N_d{=}16$ chirps per frame. The visual ranking matches the quantitative order. Beyond magnitude, 3DPS preserves $\sim$40\% of held-out relative-phase coherence on CRP ($\Delta\varphi_r{=}0.371$, $\Delta\varphi_v{=}0.434$) and on ADC ($\Delta\varphi_t{=}0.414$) without any phase supervision --- meaningful structure for downstream beamforming and range-localization. Per-scene comparisons across all training frames, CRP and ADC heatmaps, runtime, and limitations are in Sec.~\ref{app:per_scene_train},~\ref{app:rdadc:figures},~\ref{app:runtime}, and~\ref{app:limitations}.

\section{Conclusion}
\label{sec:conclusion}

We presented 3DPS, a differentiable physically based point renderer for FMCW mmWave radar that synthesizes complex range profiles from novel sensor poses. Oriented 3D points carry ITU-R~P.2040 material parameters and are evaluated by closed-form BSDFs and splatted through precomputed sensor PSFs, removing the Monte Carlo sampling that constrains prior physically based radar renderers. On six outdoor ColoRadar scenes, 3DPS reaches \textbf{0.587} mean Pearson correlation on held-out $|\mathrm{RA}|$, between $1.7\times$ and $5.2\times$ the three optical-NVS baselines and $1.3\times$ mmIR. Training takes approximately 3 minutes per scene over 8 viewpoints. \textbf{Limitations and future work.} The loss supervises $|\mathrm{RA}|$, which constrains relative phase across antennas and leaves absolute per-bin phase free. We therefore do not claim phase accuracy sufficient for deployment: subspace DOA on our render beats a zero-phase control by $2.8\times$ to $3.6\times$, yet its median error remains 2.4 to 2.8 beamwidths (Sec.~\ref{app:doa}). We evaluate six static outdoor and six static indoor scenes on one radar. Held-out correlation falls at wider view gaps (0.587 to 0.419) and indoors (0.541), where single-bounce evaluation is a limit, and moving objects and other radar hardware remain untested. LiDAR-free radar NVS, a phase-aware loss exploiting the renderer complex output, dynamic scenes, downstream perception tasks, and broader radar geometries are natural next steps.

\textbf{Reproducibility.} Code, configuration files, and the analysis scripts that generated every table and figure in this paper are available at \url{https://3d-point-splatting.github.io}.

\textbf{Ethics statement.} Radar can sense people through darkness and fog without their knowledge, so radar simulation carries privacy risks. Our experiments use public data of static scenes and involve no human subjects. Encrypted radar processing~\cite{ahmed2026mmfhe} and privacy-aware sharing of raw sensor data~\cite{10.1145/3789514.3792039} can reduce these risks. Sec.~\ref{app:broader_impact} covers broader impact and dual use.

\begin{ack}This work received no external funding. The authors declare no competing interests.\end{ack}

\clearpage
\bibliographystyle{plainnat_titlecase}
\bibliography{main}

\etocsettocdepth.toc{subsection}
\clearpage
\appendix

\etocsettocstyle{\section*{Table of Contents}\vspace{-0.5em}}{}
\begingroup
\hypersetup{linkcolor=black,pdfborder={0 0 0}}
\small
\tableofcontents
\endgroup
\clearpage

The supplement opens with a self-contained primer on FMCW mmWave radar
signal processing for readers more familiar with vision/graphics than
radar hardware (Sec.~\ref{app:fmcw_primer}). It then provides forward-model
details deferred from Sec.~\ref{sec:methods:fwd} (full surface-to-hemisphere
derivation, PSF-splat validity bounds, full ITU-R~P.2040 BSDF closed form,
LiDAR reduction parameters), the per-scene quantitative evidence and
qualitative comparisons that anchor the main paper's claims, the methodology
details for the CRP and ADC evaluation introduced in
Sec.~\ref{sec:experiments:results}, implementation details (wall-clock,
hardware), and a more detailed discussion of limitations and future work.

\section{FMCW mmWave radar primer}
\label{app:fmcw_primer}

This section summarizes the FMCW mmWave radar signal-processing chain that 3DPS targets, for readers more familiar with vision, graphics, or general machine learning than with radar hardware. Notation is kept consistent with the main paper. We follow the standard treatments in Skolnik~\cite{skolnik2001introduction} and Richards~\cite{richards2014fundamentals}, with cascade-radar parameters as in the TI MMWCAS reference~\cite{iovescu2017fundamentals}.

\subsection{FMCW chirp and dechirped ADC samples}
\label{app:fmcw_primer:chirp}

FMCW radar~\cite{jankiraman2018fmcw} transmits a linear-frequency-sweep waveform (a ``chirp'') of duration $T_c$ that rises from carrier frequency $f_c$ at slope $\mu = B/T_c$ over total swept bandwidth $B$:
\begin{equation}
s_{\text{TX}}(t) \;=\; \cos\!\bigl(2\pi(f_c\,t + \tfrac{1}{2}\mu\,t^2)\bigr), \qquad t \in [0, T_c].
\end{equation}
A reflection from a scatterer at total round-trip path length $R = R_t + R_r$ arrives at the receiver after delay $\tau = R/c$. Multiplying the transmitted chirp by the conjugate of the received signal (``dechirping'') and low-pass filtering produces a beat signal whose instantaneous frequency is constant over the chirp at $\mu \tau$. After analog-to-digital conversion at sample rate $f_s$, the per-chirp dechirped baseband samples are
\begin{equation}
a[n] \;=\; A_r\, e^{\,j 2\pi\, \mu\, \tau\, t_n}, \qquad t_n = n / f_s,\quad n = 0, \ldots, N_s-1,
\label{eq:fmcw_baseband}
\end{equation}
where the complex amplitude $A_r$ aggregates antenna gains, scattering cross section, path-length attenuation, and the carrier-phase term $e^{jk(R_t+R_r)}$ with $k = 2\pi/\lambda$. Eq.~\eqref{eq:fmcw_baseband} is the single-scatterer complex baseband response, summed over the point set in our Eq.~\eqref{eq:adc_discrete}.

\subsection{Complex baseband (I/Q) and the range profile}
\label{app:fmcw_primer:rangefft}

The dechirp mixer is implemented in I/Q form so that each ADC sample is complex, $a[n] = a_I[n] + j\,a_Q[n]$. The complex form is what makes radar a phase-coherent imaging modality: the magnitude $|a[n]|$ encodes scatterer reflectivity, while the phase argument retains fractional-wavelength path information. At a 77\,GHz carrier the wavelength is $\lambda \approx 3.9$\,mm, an order of magnitude finer than the range-bin width $\Delta R = c/(2B) \approx 5.93$\,cm produced by the range FFT below. Optical-NVS adaptations (Sec.~\ref{sec:related}) train on $|\cdot|$ alone and discard this phase content; 3DPS retains it through the splat in Eq.~\eqref{eq:crp_splat}.

The complex range profile (CRP) is the windowed range FFT of the chirp's ADC samples,
\begin{equation}
\mathrm{CRP}[k] \;=\; \sum_{n=0}^{N_s-1} w[n]\, a[n]\, e^{-j 2\pi n k / N_s}, \qquad k = 0, \ldots, N_s-1,
\label{eq:rangefft}
\end{equation}
with $w[\cdot]$ a windowing function (we use Hann, peak sidelobe $-31$\,dB). A scatterer at delay $\tau$ appears as a peak at fractional bin $k^\star = \mu\tau\,N_s/f_s$, smeared by the discrete-Fourier image of $w$ across neighboring bins --- the ``point spread function'' that our PSF splat (Eq.~\ref{eq:crp_splat}) applies.

\subsection{TDM-MIMO and the virtual array}
\label{app:fmcw_primer:mimo}

A MIMO radar array transmits and receives on multiple antenna elements. Time-division multiplexing (TDM) cycles through TX elements one at a time within a chirp loop, so each (TX, RX) pair gets its own dedicated chirp and resulting CRP. Under far-field assumptions and electrically small array baselines, every (TX, RX) pair behaves like a single ``virtual element'' located at the position
\begin{equation}
\mathbf{p}_v(t, r) \;=\; \mathbf{p}_t + \mathbf{p}_r,
\label{eq:virtual_array}
\end{equation}
the convolution of the TX and RX position sets~\cite{iovescu2017fundamentals}. Thus $N_{\text{TX}}\!\cdot\!N_{\text{RX}}$ physical channels yield up to $N_{\text{TX}}\,N_{\text{RX}}$ virtual elements at the locations of $\{\mathbf{p}_t + \mathbf{p}_r\}$.

For the TI MMWCAS cascade ($N_{\text{TX}}{=}12$, $N_{\text{RX}}{=}16$), the 192 channels form 134 unique virtual positions on four elevation rows. Only the elevation-0 row is a filled uniform linear array, with 86 elements. The other three rows are sparse, so the cascade has no useful elevation aperture. We use only the 86-element row, which fixes the 2D-only supervision constraint discussed below.

\subsection{Range--azimuth (RA) image and the 2D-only constraint}
\label{app:fmcw_primer:ra}

Stacking per-(TX, RX) CRPs along the virtual-element axis gives a virtual-array CRP of shape $(N_{\text{VA}}, N_s)$. Applying a windowed FFT across the virtual-element axis converts inter-element phase progression (which encodes the angle of arrival of each scatterer) into an azimuth-bin index, yielding the range--azimuth image $\mathrm{RA}[a, k]$. Bin widths are $\Delta R = c/(2B) \approx 5.93$\,cm in range and approximately $\theta_{\text{res}} \approx \lambda / D_{\text{VA}} \approx 1.4^\circ$ in azimuth, where $D_{\text{VA}}$ is the virtual-array aperture (${\sim}16$\,cm for the cascade's 86-element extent).

Crucially, a single azimuth row with zero elevation aperture cannot resolve elevation: scatterers at any height in the same (range, azimuth) cell project to the same RA pixel. RA supervision therefore integrates out the elevation dimension by construction, which is the underlying reason for the cascading-sparsity argument in Sec.~\ref{sec:intro}: 2D RA + sparse views + sparse returns make 3D scene structure information-theoretically unrecoverable from RA alone.

\subsection{Polar-to-Cartesian conversion for evaluation}
\label{app:fmcw_primer:cartesian}

The native RA image is intrinsically polar --- rows index range bins $k$, columns index azimuth bins $a$. For human inspection and to keep image-domain metrics meaningful in physical units, we resample the polar RA onto a Cartesian grid $(x, y)$ in the radar frame:
\begin{equation}
\mathrm{RA}_{\text{cart}}[x, y] \;=\; \mathrm{RA}_{\text{polar}}\!\bigl[k(x,y),\, a(x,y)\bigr],\qquad k = \tfrac{\sqrt{x^2 + y^2}}{\Delta R},\;\; a = \tfrac{N_{\text{az}}-1}{2} + \tfrac{N_{\text{az}}+1}{2}\sin\theta,
\label{eq:polar_to_cart}
\end{equation}
with $\theta = \arctan(x/y)$ and bilinear interpolation in $(k, a)$. The arcsin-spaced inverse for $a(x,y)$ matches the azimuth-bin grid in Sec.~\ref{sec:methods:pipeline}. We use a $399\!\times\!399$ Cartesian grid over all 256 range bins and apply the same interpolation to all methods.

\subsection{ITU-R~P.2040 mmWave material model}
\label{app:fmcw_primer:itu}

ITU-R Recommendation P.2040~\cite{itu_p2040} gives reflection and transmission of building materials through complex relative permittivity $\varepsilon_r = \varepsilon'_r - j\varepsilon''_r$ and slab thickness $d$. We add surface roughness $\sigma_h$ and correlation length $\ell_c$ for the scattering lobes, following the CSVBSDF formulation of Wei et al.~\cite{wei2024csvbsdf}. 3DPS attaches a per-point 6-vector $\boldsymbol{\theta}_i = (\varepsilon'_r, \varepsilon''_r, \sigma_h, \ell_c, \tau, d)_i$ to each oriented point, where $\tau$ is the Kirchhoff/SPM blend introduced in Eq.~\eqref{eq:bsdf}; the full closed form is in Sec.~\ref{app:bsdf_full}.

\FloatBarrier
\section{Forward model: surface-to-hemisphere derivation}
\label{app:fwd_derivation}

Building on the FMCW signal model in Sec.~\ref{app:fmcw_primer}, this section derives Eq.~\eqref{eq:rendering_eqn} from the bistatic radar equation. The per-point amplitude in Eq.~\eqref{eq:radar_eqn} then follows from a Riemann sum.

\textbf{Radar equation.} The power received over the single-bounce path TX$\,\to\,\mathbf{x}\,\to\,$RX from a point target is~\cite{skolnik2001introduction}
\begin{equation}
P_r \;=\; \frac{P_t\,G_t(\hat{\boldsymbol{\omega}}_i)\,G_r(\hat{\boldsymbol{\omega}}_o)\,\lambda^2\,\sigma}{(4\pi)^3\,R_t^2\,R_r^2},
\label{eq:radar_bistatic}
\end{equation}
where $P_t$ is transmit power, $G_t,G_r$ are the antenna power gains and $\sigma$ is the bistatic radar cross section.

\textbf{Area-domain form.} Following mmIR~\cite{mmIR2026}, we replace the point target's cross section with a distributed BSDF $f$:
\begin{equation}
\sigma \;=\; 4\pi \int f(\hat{\boldsymbol{\omega}}_i,\mathbf{x},\hat{\boldsymbol{\omega}}_o)\,\cos\theta_i\,\cos\theta_o\;\mathrm{d}A,
\label{eq:sigma_bsdf}
\end{equation}
with $\cos\theta_i=|\mathbf{n}\!\cdot\!\hat{\boldsymbol{\omega}}_i|$ and $\cos\theta_o=|\mathbf{n}\!\cdot\!\hat{\boldsymbol{\omega}}_o|$. Substituting Eq.~\eqref{eq:sigma_bsdf} into Eq.~\eqref{eq:radar_bistatic}, the differential power received from a surface element $\mathrm{d}A$ at $\mathbf{x}$ is
\begin{equation}
\mathrm{d}P_r \;=\; V\,\frac{P_t\,\lambda^2}{(4\pi)^2}\,
\frac{G_t(\hat{\boldsymbol{\omega}}_i)\,G_r(\hat{\boldsymbol{\omega}}_o)\,f(\hat{\boldsymbol{\omega}}_i,\mathbf{x},\hat{\boldsymbol{\omega}}_o)\,\cos\theta_i\,\cos\theta_o}{R_t^2\,R_r^2}\;\mathrm{d}A,
\label{eq:radar_area}
\end{equation}
where $V\in\{0,1\}$ is visibility.

\textbf{Change of variables to receiver solid angle.} A surface element subtends solid angle $\mathrm{d}\omega$ at the receiver:
\begin{equation}
\mathrm{d}\omega \;=\; \frac{\cos\theta_o}{R_r^2}\,\mathrm{d}A
\qquad\Longleftrightarrow\qquad
\cos\theta_o\,\mathrm{d}A \;=\; R_r^2\,\mathrm{d}\omega.
\label{eq:omega_area}
\end{equation}
Substituting Eq.~\eqref{eq:omega_area} into Eq.~\eqref{eq:radar_area}, the factor $\cos\theta_o\,\mathrm{d}A/R_r^2$ becomes $\mathrm{d}\omega$, so both $R_r^2$ and $\cos\theta_o$ are absorbed. Integrating over the receiver hemisphere gives Eq.~\eqref{eq:rendering_eqn}:
\begin{equation}
P_r \;=\; \frac{P_t\,\lambda^2}{(4\pi)^2}
\int_{\Omega^{+}(\mathbf{p}_r)}
\frac{G_t(\hat{\boldsymbol{\omega}}_i)\,G_r(\hat{\boldsymbol{\omega}}_o)\,f(\hat{\boldsymbol{\omega}}_i,\mathbf{x},\hat{\boldsymbol{\omega}}_o)\,\cos\theta_i\,V}{R_t^2}\;\mathrm{d}\omega.
\end{equation}

\textbf{Deterministic discretization.} Where mmIR estimates this integral by Monte Carlo sampling, we evaluate the integrand once at each of $N$ oriented points $\{(\mathbf{x}_i,\mathbf{n}_i,\boldsymbol{\theta}_i)\}_{i=1}^{N}$ with uniform solid-angle weight $\Delta\omega$:
\begin{equation}
\hat{P}_r \;=\; \frac{P_t\,\lambda^2}{(4\pi)^2}\,\Delta\omega
\sum_{i=1}^{N}
\frac{G_t(\hat{\boldsymbol{\omega}}_{i,i})\,G_r(\hat{\boldsymbol{\omega}}_{o,i})\,f_i\,\cos\theta_i\,v_i}{R_{t,i}^2}.
\label{eq:power_sum}
\end{equation}

\textbf{Per-point complex amplitude.} The square root of each term in Eq.~\eqref{eq:power_sum}, with the carrier phase attached, is the amplitude of Eq.~\eqref{eq:radar_eqn}:
\begin{equation}
S_{tr}(\mathbf{x}_i) \;=\; \underbrace{\frac{\lambda}{4\pi}\sqrt{P_t\,\Delta\omega}}_{C}\;
\frac{\sqrt{G_t(\hat{\boldsymbol{\omega}}_{i,i})\,G_r(\hat{\boldsymbol{\omega}}_{o,i})}}{R_{t,i}}\,
\sqrt{f_i\cos\theta_{i}}\;
e^{\,jk\,(R_{t,i}+R_{r,i})}.
\end{equation}
The rendered ADC sums these amplitudes coherently, which gives Eq.~\eqref{eq:adc_discrete}.

\textbf{Visibility at discretization.} A LiDAR-derived point set is not a hemisphere-conformal partition of $\Omega^{+}$, so occluded points can enter the sum. We remove occluded points up-front via Mitsuba~3 ray-casting in the LiDAR reduction pipeline (Sec.~\ref{app:lidar_reduction}). During training, $v_i$ in Eq.~\eqref{eq:adc_discrete} keeps a point when it lies at range $(1.5\,\text{m},K\Delta R)$, which excludes near-field returns and ranges past the last bin, lies in front of the array, and has $|\mathbf{n}_i\!\cdot\!\hat{\mathbf{d}}_i|>0.05$, with $\hat{\mathbf{d}}_i$ the direction to the array, which excludes grazing incidence. The BSDF is double-sided, so each point's normal flips toward each transmitter.

\FloatBarrier
\section{Validity of the PSF splat}
\label{app:splat_validity}

Eq.~\eqref{eq:crp_splat} replaces the per-sample chirp synthesis of Eq.~\eqref{eq:adc_discrete} with an $L$-tap scatter-add into range bins. We bound the two approximation sources here.

\textbf{Hann-tail truncation.} The Hann-windowed PSF $\Phi(\cdot;\delta)$ is the discrete Fourier transform of a length-$N_s$ Hann window evaluated at fractional bin offset $\delta\in[0,1)$. Its sidelobe envelope falls below $-62$~dB beyond $L{=}15$ taps relative to peak, well under the noise floor of real ColoRadar captures. Truncation to 15 taps therefore introduces no measurable energy loss into the rendered CRP.

\textbf{Fractional-bin handling.} The splat support index $\lfloor k_i \rfloor$ rounds the path's bistatic range to the nearest bin, but the carrier phase $e^{jk(R_{t,i}+R_{r,i})}$ in $S_{tr}(\mathbf{x}_i)$ (Eq.~\ref{eq:crp_splat}) is computed from the unrounded $R_{t,i}+R_{r,i}$, and the kernel $\Phi(\cdot;\delta_i)$ encodes the fractional offset $\delta_i = k_i - \lfloor k_i \rfloor$ exactly. The bin rounding therefore affects only the splat support (which $L$ adjacent bins receive contributions), not the per-bin amplitude or phase.

\textbf{Numerical equivalence to the standard ADC-then-FFT pipeline.} We evaluate Eq.~\eqref{eq:adc_discrete} directly in double precision, apply a length-$N_s$ Hann-windowed range FFT, and compare against the splat of Eq.~\eqref{eq:crp_splat} on identical inputs: the optimized points of all six scenes across 192 (TX, RX) pairs. The largest error in any scene is $8.0\!\times\!10^{-4}$ of the peak CRP magnitude, and the normalized RMS error averages $7.8\!\times\!10^{-4}$ ($-62$\,dB), set by the 15-tap truncation and the interpolated PSF table.

\FloatBarrier
\section{Full BSDF closed form}
\label{app:bsdf_full}

Following the ITU-R~P.2040 introduction in Sec.~\ref{app:fmcw_primer:itu}, the BSDF $f_i$ in Eq.~\eqref{eq:bsdf} expands as follows.

\textbf{Slab Fresnel.} The reflection coefficient for a slab of thickness $d$ and complex permittivity $\varepsilon_r=\varepsilon'_r-j\varepsilon''_r$ tracks the coherent superposition of the air--material reflection and the internal round-trip through the slab, evaluated independently on the $s$ and $p$ polarizations of the local Jones basis. mmWave polarimetric scattering studies~\cite{vahidpour2012polarimetric,moallem2014polarimetric} support the per-polarization treatment. We use the single-layer slab form in ITU-R~P.2040~\cite{itu_p2040}, which captures internal reflections in thin materials and is differentiable in $(\varepsilon'_r,\varepsilon''_r,d)$ in closed form. $R_h$ is the resulting power, projected onto the TX and RX polarizations, at the half vector, and $R_n$ is the same power at the mean normal.

\textbf{KA lobe.} $f_{\text{KA}} = D\,G/(4\cos\theta_i\cos\theta_o)$ is a GGX microfacet lobe~\cite{walter2007microfacet}. $D$ is the GGX distribution with $\alpha=\sqrt{4\pi\sigma_h/\lambda}$, and $G=1/(1+\Lambda_i+\Lambda_o)$ is the joint Smith term. We clamp $\alpha$ to $[0.05,0.95]$.

\textbf{SPM lobe.} $f_{\text{SPM}} = \frac{\kappa}{4\pi\sinh\kappa}\exp\bigl(\kappa(\hat{\boldsymbol{\omega}}_o\!\cdot\!\hat{\boldsymbol{\omega}}_r-1)\bigr)\,\epsilon_c$ is a von~Mises--Fisher lobe about the mirror direction $\hat{\boldsymbol{\omega}}_r$. Its concentration is $\kappa=\sqrt{\ell_c/\lambda}/(1+3\sigma_h/\ell_c)$, and $\epsilon_c=(|\varepsilon'_r-1|+\varepsilon''_r)/5$ scales it by permittivity contrast. We clamp $\kappa$ to $[0.5,10]$ and $\epsilon_c$ to $[0.2,1]$.

\textbf{Lobe blend.} $\tau_{\text{eff}}=w_{\text{KA}}/(w_{\text{KA}}+w_{\text{SPM}})$, with $w_{\text{KA}}=s_\theta\,\tau\,s(2(k\ell_c-6))$ and $w_{\text{SPM}}=(1-s_\theta)(1-\tau)\,s(10(0.3-k\sigma_h))$. Here $s$ is the sigmoid and $s_\theta=s(20(\cos\theta_i-0.94))$ favors KA near normal incidence. We clamp $\tau_{\text{eff}}$ to $[0.01,0.99]$.

\textbf{Reparameterizations and bounds.} We reparameterize $\varepsilon'_r=1+\mathrm{softplus}(x)$ and $\tau=0.05+0.9\,\mathrm{sigmoid}(x)$, and $\varepsilon''_r,\sigma_h,\ell_c,d$ as exponentials of clamped values, with ranges $[10^{-3},9{\cdot}10^{6}]$, $[10^{-7},10^{-3}]$\,m, $[5{\cdot}10^{-5},7.4]$\,m and $[10^{-3},7.4]$\,m. Every point starts from a concrete-like prior ($\varepsilon'_r{=}5.31$, $\varepsilon''_r{=}0.0326$, $\sigma_h{=}50\,\mu$m, $\ell_c{=}5$\,mm, $\tau{=}0.5$, $d{=}15$\,cm), with per-point freedom afterward.

\FloatBarrier
\section{LiDAR reduction parameters}
\label{app:lidar_reduction}

The LiDAR-to-oriented-point-set reduction in Sec.~\ref{sec:methods:impl} runs in four stages on each scene's co-registered LiDAR cloud (positions, normals, intensity; ${\sim}1.8$--$3.0$M points across our scenes), reducing it to exactly $N{=}20{,}000$ oriented points. Points come directly from the LiDAR cloud, which keeps its per-point density. A surface mesh~\cite{kazhdan2006poisson} serves only as the occluder in Stage 2, because the radar's beamwidth and range resolution are both coarser than typical reconstruction error.

\textbf{Stage 1 --- frustum and range cull.} Points with $\cos(\angle(\hat{\mathbf{d}},\hat{\mathbf{b}})) \le 0.1761$ (boresight half-angle ${\approx}79.9^\circ$ matching the radar's azimuth FOV) or radial range outside $[1.5\,\text{m}, K\Delta R{=}15.2\,\text{m}]$ are dropped. This typically removes 80--95\% of the raw LiDAR points.

\textbf{Stage 2 --- occlusion cull.} Surviving points are ray-cast toward the array centroid via Mitsuba~3~\cite{jakob2022dr}; any point whose ray intersects another scene element before reaching the centroid is dropped as self-occluded. Stage 2 runs once at initialization, and the renderer does not ray-cast during training.

\textbf{Stage 3 --- cosine-weighted resampling.} We draw $3N$ candidates with replacement at probability $\propto \max(\cos(\angle(\hat{\mathbf{d}},\hat{\mathbf{b}})), 0.01)$, biasing selection toward boresight-aligned regions where the antenna pattern is strongest while preserving a long tail of off-boresight returns.

\textbf{Stage 4 --- farthest-point sampling.} The $3N$ candidates are reduced to exactly $N{=}20{,}000$ via greedy farthest-point sampling, yielding a near-uniform spatial distribution across the boresight-biased candidate pool.

Each surviving point inherits a quaternion encoding its LiDAR surface normal and a uniform ITU-concrete material prior $\boldsymbol{\theta}_i^{(0)}$~\cite{itu_p2040}. Per-point material drift afterward is unconstrained (Sec.~\ref{sec:methods:impl}).

\textbf{LiDAR--radar alignment.} ColoRadar's shipped calibration is not accurate enough for coherent rendering, so we align each frame's radar pose to the LiDAR scene in two passes. Pass 1 aligns each frame independently and keeps the candidate with the highest Cartesian correlation against the measured RA. Pass 2 fits a smoothed trajectory across the scene's frames, flags frames far from it, and re-aligns those with a search centered on the trajectory, rejecting candidates beyond 2.5 median absolute deviations.

\FloatBarrier
\section{Per-scene $|\mathrm{RA}|$ breakdowns}
\label{app:per_scene_ra}

The main paper Table~\ref{tab:results} reports the 6-scene mean for each method on training views (8 views per scene) and the held-out test view. Tables~\ref{tab:test_ra} and~\ref{tab:train_ra} below give the per-scene breakdown that anchors those means, using the $|\mathrm{RA}|$ metric set (Corr, PSNR, SSIM, RMSE on min-max-normalized $399\!\times\!399$ Cartesian images, computed identically for all methods). The Mean row of each table matches the corresponding 3DPS row in Table~\ref{tab:results} by construction.

\begin{table*}[t]
\centering
\caption{\textbf{Per-scene held-out test $|\mathrm{RA}|$ metrics on six ColoRadar scenes.} \textbf{Bold}: best per scene per metric (ties bolded jointly). \underline{Underline}: second-best. The Mean row anchors the corresponding test row of main paper Table~\ref{tab:results}, and the Std row reports the per-method standard deviation across the six scenes.}
\label{tab:test_ra}
\small
\setlength{\tabcolsep}{1.6pt}
\resizebox{\linewidth}{!}{
\begin{tabular}{l|cccccc|cccccc|cccccc|cccccc}
\toprule
 & \multicolumn{6}{c}{RA Corr $\uparrow$} & \multicolumn{6}{c}{RA PSNR $\uparrow$} & \multicolumn{6}{c}{RA SSIM $\uparrow$} & \multicolumn{6}{c}{RA RMSE $\downarrow$} \\
\cmidrule(lr){2-7}\cmidrule(lr){8-13}\cmidrule(lr){14-19}\cmidrule(lr){20-25}
Scene & Ours & mmIR & Sionna & RSplat & RFields & DART & Ours & mmIR & Sionna & RSplat & RFields & DART & Ours & mmIR & Sionna & RSplat & RFields & DART & Ours & mmIR & Sionna & RSplat & RFields & DART \\
\midrule
S0\,F135 & \textbf{0.644} & 0.268 & 0.394 & \underline{0.407} & 0.058 & 0.114 & \textbf{31.3} & 28.1 & 27.9 & 26.4 & \underline{28.8} & 16.6 & \textbf{0.839} & 0.543 & \underline{0.652} & 0.545 & 0.616 & 0.177 & \textbf{0.0273} & 0.0395 & 0.0403 & 0.0478 & \underline{0.0363} & 0.1481 \\
S1\,F185 & 0.519 & \underline{0.523} & \textbf{0.548} & 0.305 & 0.087 & 0.057 & \underline{29.5} & \textbf{30.1} & 29.1 & 26.9 & 28.2 & 21.1 & \textbf{0.780} & \underline{0.770} & 0.761 & 0.618 & 0.583 & 0.276 & \underline{0.0333} & \textbf{0.0311} & 0.0351 & 0.0452 & 0.0389 & 0.0880 \\
S1\,F438 & \textbf{0.642} & \underline{0.599} & 0.410 & 0.226 & 0.210 & 0.060 & 27.1 & \textbf{28.7} & \underline{28.3} & 22.7 & 27.4 & 21.6 & 0.663 & \textbf{0.779} & \underline{0.736} & 0.502 & 0.435 & 0.349 & 0.0442 & \textbf{0.0367} & \underline{0.0385} & 0.0732 & 0.0426 & 0.0827 \\
S2\,F105 & \textbf{0.553} & 0.193 & 0.088 & \underline{0.369} & 0.226 & 0.241 & 27.6 & \textbf{28.3} & 25.1 & 27.0 & \underline{28.2} & 22.4 & \underline{0.683} & \textbf{0.692} & 0.548 & 0.518 & 0.439 & 0.516 & 0.0415 & \textbf{0.0385} & 0.0556 & 0.0446 & \underline{0.0388} & 0.0760 \\
S2\,F160 & \textbf{0.657} & 0.557 & 0.495 & \underline{0.559} & 0.075 & 0.085 & 27.7 & \textbf{29.0} & 27.2 & 20.8 & 26.2 & \underline{27.8} & \underline{0.653} & 0.567 & 0.626 & 0.372 & 0.319 & \textbf{0.687} & 0.0413 & \textbf{0.0355} & 0.0435 & 0.0908 & 0.0488 & \underline{0.0407} \\
S2\,F300 & \underline{0.508} & \textbf{0.585} & 0.341 & 0.168 & 0.141 & 0.117 & \textbf{29.4} & 28.6 & 27.3 & 21.5 & \underline{28.9} & 25.3 & \textbf{0.785} & \underline{0.755} & 0.711 & 0.512 & 0.585 & 0.446 & \textbf{0.0340} & 0.0371 & 0.0429 & 0.0841 & \underline{0.0360} & 0.0545 \\
\midrule
\textbf{Mean} & \textbf{0.587} & \underline{0.454} & 0.380 & 0.339 & 0.133 & 0.112 & \textbf{28.8} & \textbf{28.8} & 27.5 & 24.2 & \underline{28.0} & 22.5 & \textbf{0.734} & \underline{0.685} & 0.672 & 0.511 & 0.496 & 0.408 & \underline{0.0369} & \textbf{0.0364} & 0.0426 & 0.0643 & 0.0402 & 0.0817 \\
\textbf{Std} & 0.068 & 0.177 & 0.161 & 0.140 & 0.072 & 0.068 & 1.58 & 0.73 & 1.35 & 2.86 & 1.00 & 3.83 & 0.077 & 0.105 & 0.079 & 0.080 & 0.117 & 0.182 & 0.0064 & 0.0030 & 0.0070 & 0.0210 & 0.0048 & 0.0372 \\
\bottomrule
\end{tabular}
}
\end{table*}

\begin{table*}[t]
\centering
\caption{\textbf{Per-scene training-view $|\mathrm{RA}|$ metrics on six ColoRadar scenes.} Each cell is the mean over the 8 training frames per scene. Same metrics as Table~\ref{tab:test_ra}. \textbf{Bold}: best per scene per metric (ties bolded jointly). \underline{Underline}: second-best. The Mean row anchors the corresponding train row of main paper Table~\ref{tab:results}, and the Std row reports the per-method standard deviation across the six scenes.}
\label{tab:train_ra}
\small
\setlength{\tabcolsep}{1.6pt}
\resizebox{\linewidth}{!}{
\begin{tabular}{l|cccccc|cccccc|cccccc|cccccc}
\toprule
 & \multicolumn{6}{c}{RA Corr $\uparrow$} & \multicolumn{6}{c}{RA PSNR $\uparrow$} & \multicolumn{6}{c}{RA SSIM $\uparrow$} & \multicolumn{6}{c}{RA RMSE $\downarrow$} \\
\cmidrule(lr){2-7}\cmidrule(lr){8-13}\cmidrule(lr){14-19}\cmidrule(lr){20-25}
Scene & Ours & mmIR & Sionna & RSplat & RFields & DART & Ours & mmIR & Sionna & RSplat & RFields & DART & Ours & mmIR & Sionna & RSplat & RFields & DART & Ours & mmIR & Sionna & RSplat & RFields & DART \\
\midrule
S0\,F135 & \textbf{0.822} & \underline{0.545} & 0.382 & 0.356 & 0.308 & -0.028 & \textbf{32.0} & \underline{29.5} & 27.3 & 25.5 & 25.6 & 22.5 & \textbf{0.837} & \underline{0.801} & 0.671 & 0.540 & 0.522 & 0.468 & \textbf{0.0261} & \underline{0.0346} & 0.0434 & 0.0540 & 0.0527 & 0.0763 \\
S1\,F185 & \textbf{0.836} & 0.506 & \underline{0.598} & 0.354 & 0.196 & 0.016 & \textbf{32.9} & 25.0 & 26.8 & \underline{28.0} & 26.3 & 26.0 & \textbf{0.819} & 0.703 & \underline{0.708} & 0.639 & 0.594 & 0.484 & \textbf{0.0266} & 0.0577 & 0.0472 & \underline{0.0406} & 0.0495 & 0.0504 \\
S1\,F438 & \textbf{0.848} & \underline{0.660} & 0.641 & 0.268 & 0.091 & 0.033 & \textbf{32.6} & \underline{29.0} & 28.1 & 23.2 & 23.4 & 25.3 & \textbf{0.852} & \underline{0.820} & 0.701 & 0.512 & 0.419 & 0.578 & \textbf{0.0241} & \underline{0.0360} & 0.0425 & 0.0707 & 0.0684 & 0.0578 \\
S2\,F105 & \textbf{0.804} & \underline{0.578} & 0.099 & 0.427 & 0.229 & 0.153 & \textbf{31.3} & \underline{29.2} & 25.0 & 26.5 & 24.0 & 23.6 & \textbf{0.815} & \underline{0.778} & 0.541 & 0.494 & 0.409 & 0.656 & \textbf{0.0289} & \underline{0.0354} & 0.0578 & 0.0477 & 0.0633 & 0.0678 \\
S2\,F160 & \textbf{0.784} & 0.532 & 0.561 & \underline{0.575} & 0.306 & -0.038 & \textbf{29.7} & 23.0 & \underline{26.8} & 22.2 & 23.6 & 20.4 & \textbf{0.743} & 0.579 & \underline{0.605} & 0.377 & 0.296 & 0.541 & \textbf{0.0338} & 0.0717 & \underline{0.0469} & 0.0790 & 0.0661 & 0.0968 \\
S2\,F300 & \textbf{0.774} & \underline{0.599} & 0.296 & 0.211 & 0.174 & 0.062 & \textbf{32.9} & \underline{31.5} & 27.9 & 24.0 & 26.7 & 25.8 & \textbf{0.847} & \underline{0.833} & 0.733 & 0.596 & 0.639 & 0.511 & \textbf{0.0236} & \underline{0.0288} & 0.0415 & 0.0649 & 0.0487 & 0.0546 \\
\midrule
\textbf{Mean} & \textbf{0.812} & \underline{0.570} & 0.430 & 0.365 & 0.218 & 0.033 & \textbf{31.9} & \underline{27.9} & 27.0 & 24.9 & 24.9 & 23.9 & \textbf{0.819} & \underline{0.752} & 0.660 & 0.526 & 0.480 & 0.540 & \textbf{0.0272} & \underline{0.0440} & 0.0466 & 0.0595 & 0.0581 & 0.0673 \\
\textbf{Std} & 0.029 & 0.055 & 0.210 & 0.128 & 0.083 & 0.070 & 1.25 & 3.19 & 1.11 & 2.16 & 1.43 & 2.20 & 0.040 & 0.097 & 0.073 & 0.091 & 0.129 & 0.069 & 0.0038 & 0.0168 & 0.0060 & 0.0146 & 0.0088 & 0.0173 \\
\bottomrule
\end{tabular}
}
\end{table*}

\textbf{Indoor scenes.} We run six indoor ColoRadar scenes under the same protocol and hyperparameters. Held-out Corr: 3DPS 0.541, RadarSplat 0.267, Radar Fields 0.132 ($n{=}5$; one run diverged), DART 0.099. Per scene (3DPS): 0.319, 0.455, 0.476, 0.752, 0.687, 0.559. mmIR and Sionna RT were not run on these splits.

\textbf{View-gap sweep.} We hold out $F$ and train on every second ($F\!\pm\!2,4,6,8$) or third ($F\!\pm\!3,6,9,12$) frame. Held-out Corr at stride 1, 2, 3: 3DPS 0.587, 0.442, 0.419; RadarSplat 0.339, 0.198, 0.203 ($n{=}5$ at stride 3, because one run has no result); Radar Fields 0.133, 0.078, 0.077; DART 0.112, 0.073, 0.078. mmIR and Sionna RT were not run on these splits.

\FloatBarrier
\section{CRP and ADC evaluation}
\label{app:rdadc}

Because 3DPS produces complex range profiles, the same optimized scene yields range--azimuth and raw ADC outputs by applying the standard azimuth FFT and inverse range FFT to the per-(TX,~RX) range-profile output of Eq.~\eqref{eq:crp_splat} --- with no retraining. The aggregate fidelity numbers were reported in Table~\ref{tab:results} of the main paper. This section provides the evaluation conventions and the per-scene CRP/ADC tables that anchor the 3DPS row of Table~\ref{tab:results}.

\subsection{Pipeline conventions}
\label{app:rdadc:pipeline}

CRP and ADC are evaluated with the windows used for the training loss, so they compare directly with the $|\mathrm{RA}|$ results. The GT side applies range Hann + range FFT to the raw ADC; both GT and predicted CRPs then receive the azimuth Hann window on the virtual-array axis, matching the conventions used during training. Range bins $[0{:}15]$ are zeroed in both signals to reject TX--RX coupling. ADC is the inverse range FFT of the resulting CRP. Magnitude metrics ($\rho$, PSNR, SSIM, $\rho_{|\cdot|}$) are computed on independently min-max-normalized $|\cdot|$, identical to the $|\mathrm{RA}|$ convention. Relative-phase metrics ($\Delta\varphi_r$, $\Delta\varphi_v$, $\Delta\varphi_t$) are computed on the raw (uncorrected) signals and are invariant to per-bin absolute phase, so they require no calibration removal.

\subsection{Per-scene CRP and ADC fidelity}
\label{app:rdadc:per_scene}

Tables~\ref{tab:crp_adc_train} and~\ref{tab:crp_adc_test} give the per-scene CRP/ADC fidelity that anchors the 3DPS row of Table~\ref{tab:results}. For each scene, the value is the 8-train-frame mean (train table) or the held-out test frame value (test table); the Mean row at the bottom of each is the average across the six scenes and matches the corresponding 3DPS Train Mean / Test Mean row of Table~\ref{tab:results} by construction. Same metric set and column layout as Table~\ref{tab:results}.

\begin{table}[t]
\centering
\small
\caption{\textbf{Per-scene CRP and ADC fidelity (training-view), 3DPS (Ours), mmIR and Sionna RT.} Mean row anchors the 3DPS (Ours) train row of main paper Table~\ref{tab:results}. Std is across the six scenes. \colorbox{magbg}{\textbf{Magnitude}} columns: Pearson correlation, PSNR, SSIM, RMSE on min-max-normalized $|\cdot|$. \colorbox{vrbg}{\textbf{Rel.\ phase}} columns: magnitude-weighted $\langle\cos(\Delta\varphi)\rangle$ across range ($\Delta\varphi_r$), across virtual antennas ($\Delta\varphi_v$), and across fast-time ($\Delta\varphi_t$); all in $[-1,1]$, random reference 0.}
\label{tab:crp_adc_train}
\setlength{\tabcolsep}{4pt}
\begin{tabular}{l | cccc | cc | c | c}
\toprule
& \multicolumn{6}{c|}{\textbf{CRP}} & \multicolumn{2}{c}{\textbf{ADC}} \\
\cmidrule(lr){2-7}\cmidrule(lr){8-9}
& \multicolumn{4}{c|}{\colorbox{magbg}{\textbf{Magnitude}}} & \multicolumn{2}{c|}{\colorbox{vrbg}{\textbf{Rel.\ phase}}} & \colorbox{magbg}{\textbf{Env.}} & \colorbox{vrbg}{\textbf{Rel.\ phase}} \\
Scene & Corr & PSNR & SSIM & RMSE & $\Delta\varphi_r$ & $\Delta\varphi_v$ & $\rho_{|\cdot|}$ & $\Delta\varphi_t$ \\
\midrule
\multicolumn{9}{l}{\textit{3DPS (Ours)}} \\
S0\,F135 & 0.677 & 28.1 & 0.697 & 0.0398 & 0.598 & 0.528 & 0.665 & 0.422 \\
S1\,F185 & 0.741 & 26.5 & 0.753 & 0.0483 & 0.533 & 0.507 & 0.674 & 0.757 \\
S1\,F438 & 0.718 & 29.6 & 0.742 & 0.0343 & 0.468 & 0.586 & 0.725 & 0.128 \\
S2\,F105 & 0.690 & 27.3 & 0.707 & 0.0449 & 0.494 & 0.395 & 0.666 & 0.422 \\
S2\,F160 & 0.671 & 26.4 & 0.627 & 0.0505 & 0.435 & 0.376 & 0.681 & 0.419 \\
S2\,F300 & 0.709 & 29.6 & 0.798 & 0.0352 & 0.579 & 0.569 & 0.719 & 0.519 \\
\midrule
\textbf{Mean} & \textbf{0.701} & \textbf{27.9} & \textbf{0.721} & \textbf{0.0422} & \textbf{0.518} & \textbf{0.494} & \textbf{0.689} & \textbf{0.444} \\
\textbf{Std}  & 0.027 & 1.44 & 0.058 & 0.0068 & 0.064 & 0.088 & 0.027 & 0.203 \\
\midrule
\multicolumn{9}{l}{\textit{mmIR~\cite{mmIR2026}}} \\
S0\,F135 & 0.561 & 25.0 & 0.568 & 0.0584 & 0.605 & 0.434 & 0.615 & 0.348 \\
S1\,F185 & 0.572 & 22.7 & 0.649 & 0.0744 & 0.440 & 0.302 & 0.574 & 0.699 \\
S1\,F438 & 0.622 & 26.5 & 0.660 & 0.0478 & 0.514 & 0.578 & 0.613 & 0.129 \\
S2\,F105 & 0.559 & 24.9 & 0.614 & 0.0586 & 0.466 & 0.285 & 0.578 & 0.391 \\
S2\,F160 & 0.520 & 22.1 & 0.492 & 0.0816 & 0.424 & 0.343 & 0.552 & 0.362 \\
S2\,F300 & 0.560 & 25.6 & 0.682 & 0.0536 & 0.552 & 0.518 & 0.587 & 0.438 \\
\midrule
\textbf{Mean} & 0.566 & 24.5 & 0.611 & 0.0624 & 0.500 & 0.410 & 0.587 & 0.395 \\
\textbf{Std} & 0.033 & 1.73 & 0.070 & 0.0129 & 0.070 & 0.120 & 0.024 & 0.183 \\
\midrule
\multicolumn{9}{l}{\textit{Sionna RT~\cite{sionna}}} \\
S0\,F135 & 0.338 & 22.7 & 0.567 & 0.0736 & 0.398 & 0.334 & 0.589 & 0.303 \\
S1\,F185 & 0.617 & 23.6 & 0.677 & 0.0661 & 0.468 & 0.399 & 0.619 & 0.755 \\
S1\,F438 & 0.609 & 25.6 & 0.591 & 0.0563 & 0.570 & 0.523 & 0.662 & 0.164 \\
S2\,F105 & 0.232 & 21.6 & 0.525 & 0.0838 & 0.460 & 0.101 & 0.582 & 0.261 \\
S2\,F160 & 0.524 & 23.9 & 0.540 & 0.0653 & 0.462 & 0.325 & 0.600 & 0.301 \\
S2\,F300 & 0.354 & 21.7 & 0.556 & 0.0826 & 0.454 & 0.349 & 0.594 & 0.262 \\
\midrule
\textbf{Mean} & 0.446 & 23.2 & 0.576 & 0.0713 & 0.469 & 0.338 & 0.608 & 0.341 \\
\textbf{Std} & 0.160 & 1.51 & 0.054 & 0.0107 & 0.056 & 0.137 & 0.030 & 0.209 \\
\bottomrule
\end{tabular}
\end{table}

\begin{table}[t]
\centering
\small
\caption{\textbf{Per-scene CRP and ADC fidelity (held-out test), 3DPS (Ours), mmIR and Sionna RT.} Mean row anchors the 3DPS (Ours) test row of main paper Table~\ref{tab:results}. Std is across the six scenes. \colorbox{magbg}{\textbf{Magnitude}} columns: Pearson correlation, PSNR, SSIM, RMSE on min-max-normalized $|\cdot|$. \colorbox{vrbg}{\textbf{Rel.\ phase}} columns: magnitude-weighted $\langle\cos(\Delta\varphi)\rangle$ across range ($\Delta\varphi_r$), across virtual antennas ($\Delta\varphi_v$), and across fast-time ($\Delta\varphi_t$); all in $[-1,1]$, random reference 0.}
\label{tab:crp_adc_test}
\setlength{\tabcolsep}{4pt}
\begin{tabular}{l | cccc | cc | c | c}
\toprule
& \multicolumn{6}{c|}{\textbf{CRP}} & \multicolumn{2}{c}{\textbf{ADC}} \\
\cmidrule(lr){2-7}\cmidrule(lr){8-9}
& \multicolumn{4}{c|}{\colorbox{magbg}{\textbf{Magnitude}}} & \multicolumn{2}{c|}{\colorbox{vrbg}{\textbf{Rel.\ phase}}} & \colorbox{magbg}{\textbf{Env.}} & \colorbox{vrbg}{\textbf{Rel.\ phase}} \\
Scene & Corr & PSNR & SSIM & RMSE & $\Delta\varphi_r$ & $\Delta\varphi_v$ & $\rho_{|\cdot|}$ & $\Delta\varphi_t$ \\
\midrule
\multicolumn{9}{l}{\textit{3DPS (Ours)}} \\
S0\,F135 & 0.626 & 26.2 & 0.714 & 0.0492 & 0.345 & 0.514 & 0.667 & 0.505 \\
S1\,F185 & 0.594 & 26.0 & 0.705 & 0.0501 & 0.437 & 0.482 & 0.612 & 0.711 \\
S1\,F438 & 0.554 & 25.0 & 0.632 & 0.0565 & 0.352 & 0.590 & 0.586 & -0.019 \\
S2\,F105 & 0.611 & 23.2 & 0.639 & 0.0694 & 0.336 & 0.309 & 0.590 & 0.434 \\
S2\,F160 & 0.615 & 23.6 & 0.536 & 0.0662 & 0.317 & 0.266 & 0.588 & 0.460 \\
S2\,F300 & 0.619 & 24.2 & 0.671 & 0.0617 & 0.439 & 0.445 & 0.636 & 0.395 \\
\midrule
\textbf{Mean} & \textbf{0.603} & \textbf{24.7} & \textbf{0.650} & \textbf{0.0589} & \textbf{0.371} & \textbf{0.434} & \textbf{0.613} & \textbf{0.414} \\
\textbf{Std}  & 0.026 & 1.24 & 0.065 & 0.0083 & 0.053 & 0.124 & 0.033 & 0.239 \\
\midrule
\multicolumn{9}{l}{\textit{mmIR~\cite{mmIR2026}}} \\
S0\,F135 & 0.246 & 22.7 & 0.418 & 0.0731 & 0.182 & 0.346 & 0.685 & 0.663 \\
S1\,F185 & 0.525 & 26.0 & 0.728 & 0.0503 & 0.457 & 0.389 & 0.655 & 0.621 \\
S1\,F438 & 0.557 & 24.6 & 0.644 & 0.0590 & 0.449 & 0.579 & 0.561 & 0.122 \\
S2\,F105 & 0.300 & 21.1 & 0.494 & 0.0883 & 0.390 & 0.276 & 0.614 & 0.342 \\
S2\,F160 & 0.563 & 22.8 & 0.487 & 0.0728 & 0.487 & 0.425 & 0.635 & 0.462 \\
S2\,F300 & 0.626 & 23.9 & 0.638 & 0.0640 & 0.441 & 0.536 & 0.611 & 0.395 \\
\midrule
\textbf{Mean} & 0.470 & 23.5 & 0.568 & 0.0679 & 0.401 & 0.425 & 0.627 & 0.434 \\
\textbf{Std} & 0.156 & 1.70 & 0.119 & 0.0132 & 0.112 & 0.115 & 0.042 & 0.198 \\
\midrule
\multicolumn{9}{l}{\textit{Sionna RT~\cite{sionna}}} \\
S0\,F135 & 0.410 & 22.9 & 0.534 & 0.0713 & 0.357 & 0.231 & 0.654 & 0.438 \\
S1\,F185 & 0.525 & 22.0 & 0.604 & 0.0794 & 0.457 & 0.381 & 0.602 & 0.655 \\
S1\,F438 & 0.417 & 23.6 & 0.538 & 0.0659 & 0.437 & 0.519 & 0.574 & 0.091 \\
S2\,F105 & 0.202 & 18.9 & 0.339 & 0.1131 & 0.418 & 0.146 & 0.557 & 0.149 \\
S2\,F160 & 0.502 & 22.4 & 0.481 & 0.0755 & 0.422 & 0.257 & 0.602 & 0.460 \\
S2\,F300 & 0.462 & 21.0 & 0.546 & 0.0893 & 0.389 & 0.434 & 0.622 & 0.358 \\
\midrule
\textbf{Mean} & 0.420 & 21.8 & 0.507 & 0.0824 & 0.413 & 0.328 & 0.602 & 0.359 \\
\textbf{Std} & 0.116 & 1.67 & 0.091 & 0.0170 & 0.036 & 0.140 & 0.034 & 0.210 \\
\bottomrule
\end{tabular}
\end{table}

\FloatBarrier
\section{Qualitative comparisons}
\label{app:qualitative}

\subsection{Per-scene $|\mathrm{RA}|$ comparison across methods}
\label{app:per_scene_train}

\begin{figure}[p]
\centering
\setlength{\tabcolsep}{1pt}\renewcommand{\arraystretch}{0.6}
\begin{tabular}{ccccc}
 & {\footnotesize $F\!-\!4$} & {\footnotesize $F\!-\!3$} & {\footnotesize $F\!-\!2$} & {\footnotesize $F\!-\!1$} \\
\raisebox{2.0em}{\rotatebox{90}{\scriptsize\textbf{GT}}} & \includegraphics[width=0.165\linewidth]{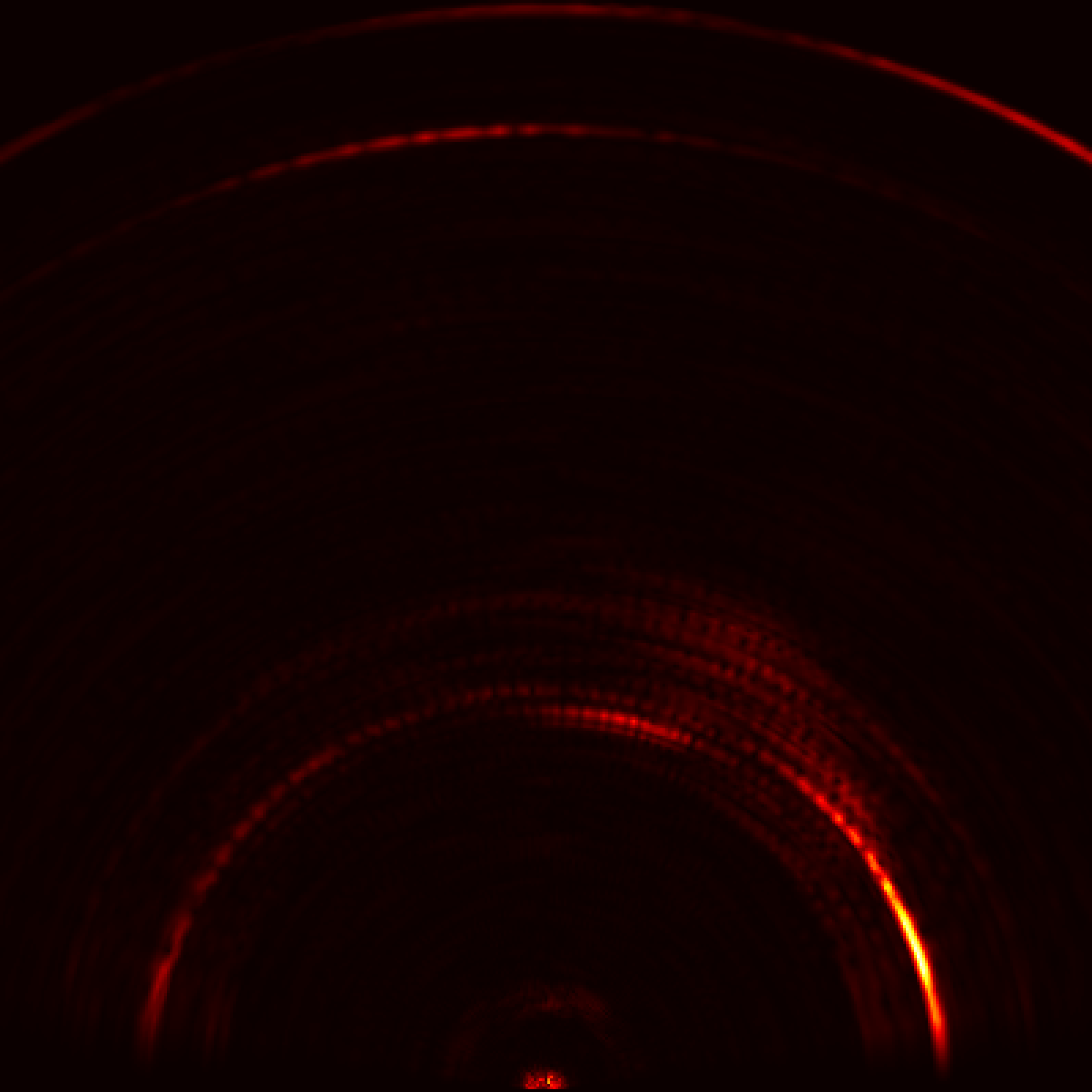} & \includegraphics[width=0.165\linewidth]{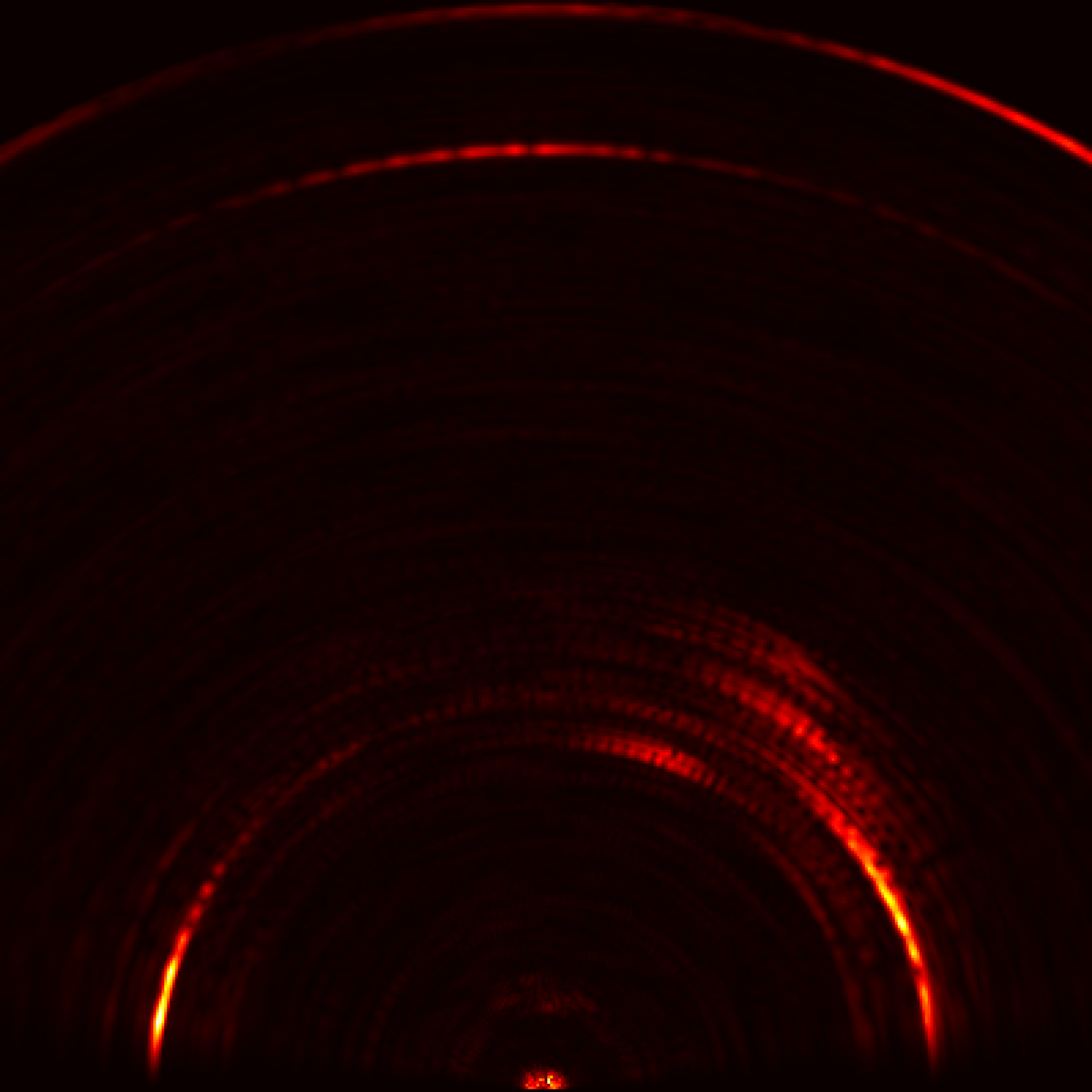} & \includegraphics[width=0.165\linewidth]{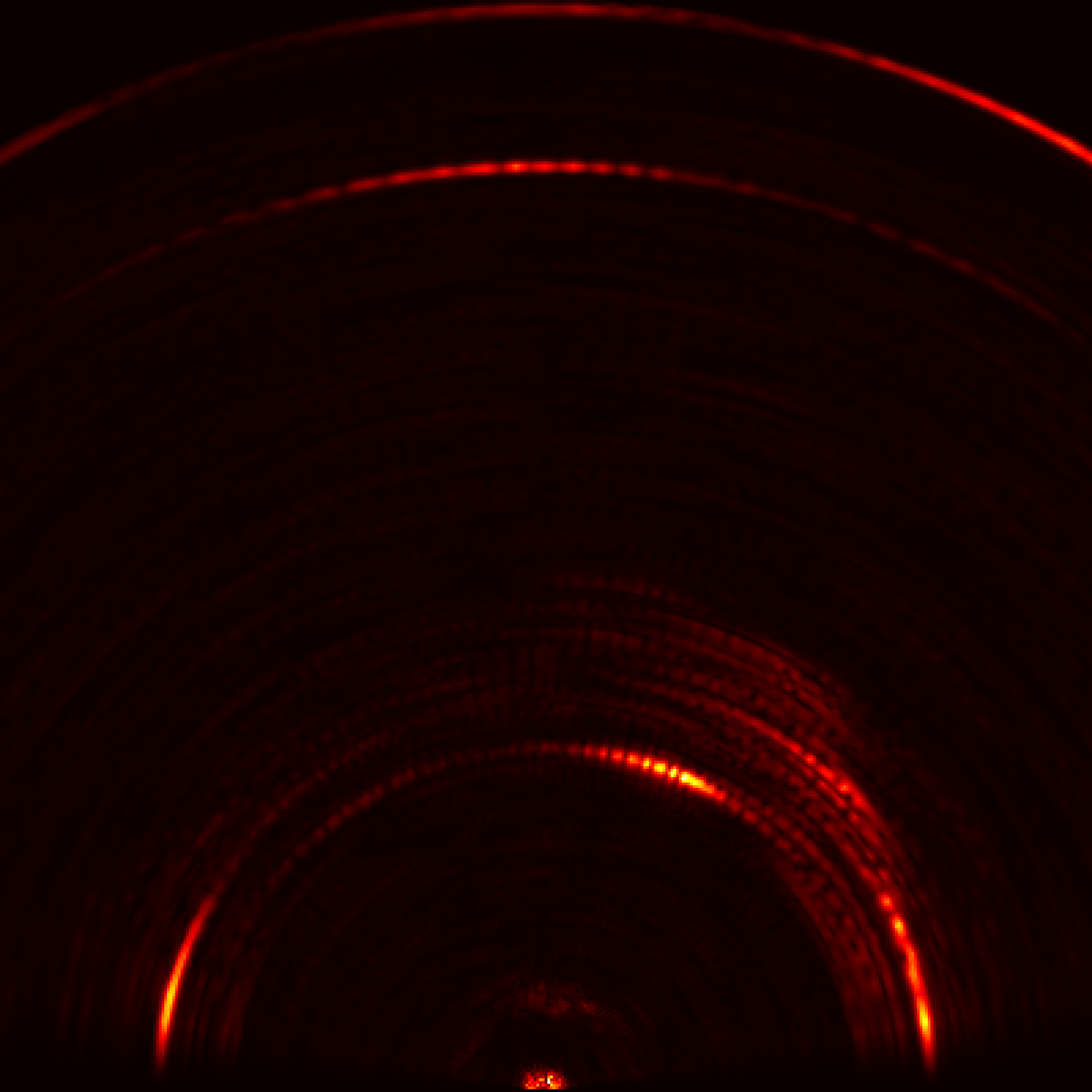} & \includegraphics[width=0.165\linewidth]{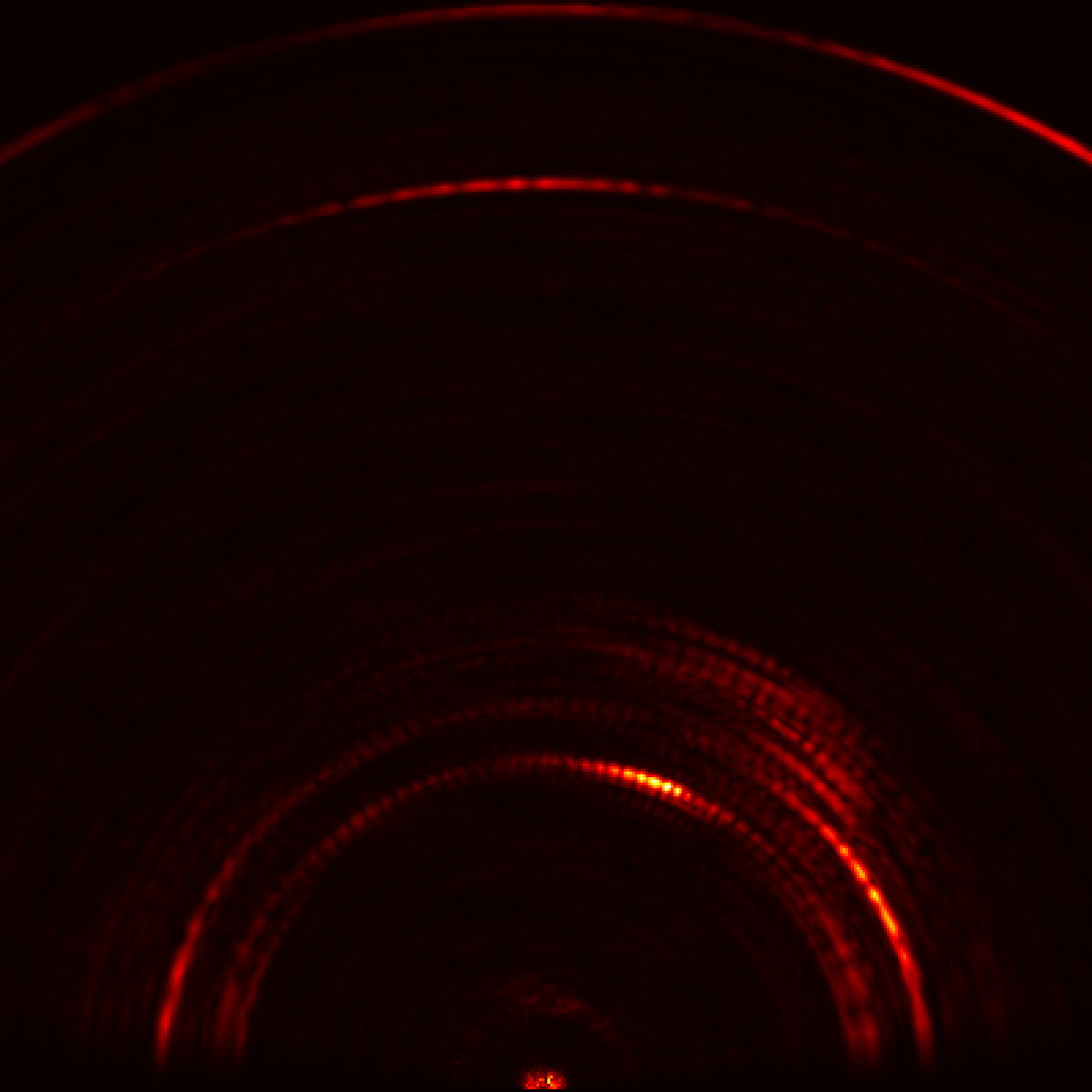} \\
\raisebox{2.0em}{\rotatebox{90}{\scriptsize\textbf{3DPS (Ours)}}} & \includegraphics[width=0.165\linewidth]{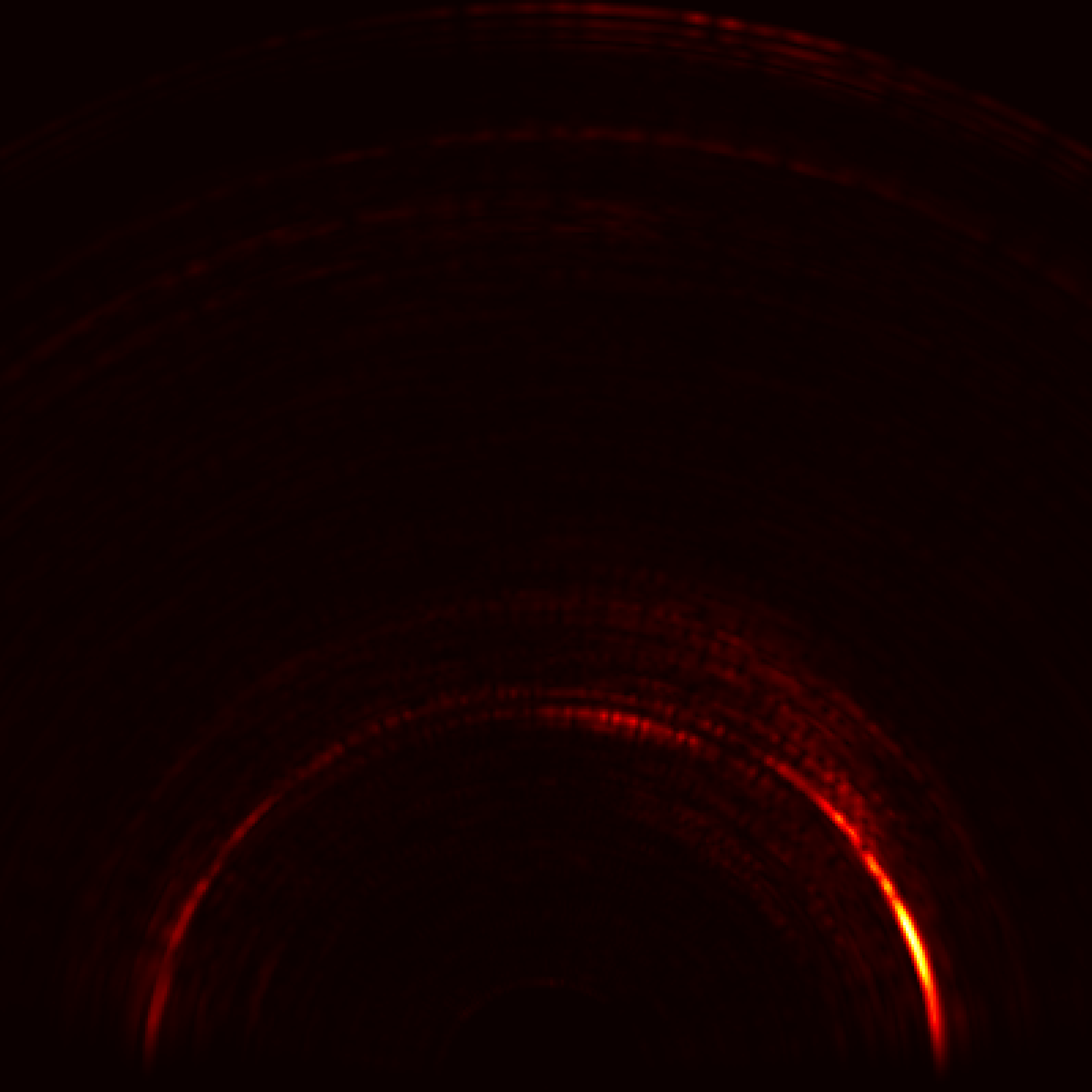} & \includegraphics[width=0.165\linewidth]{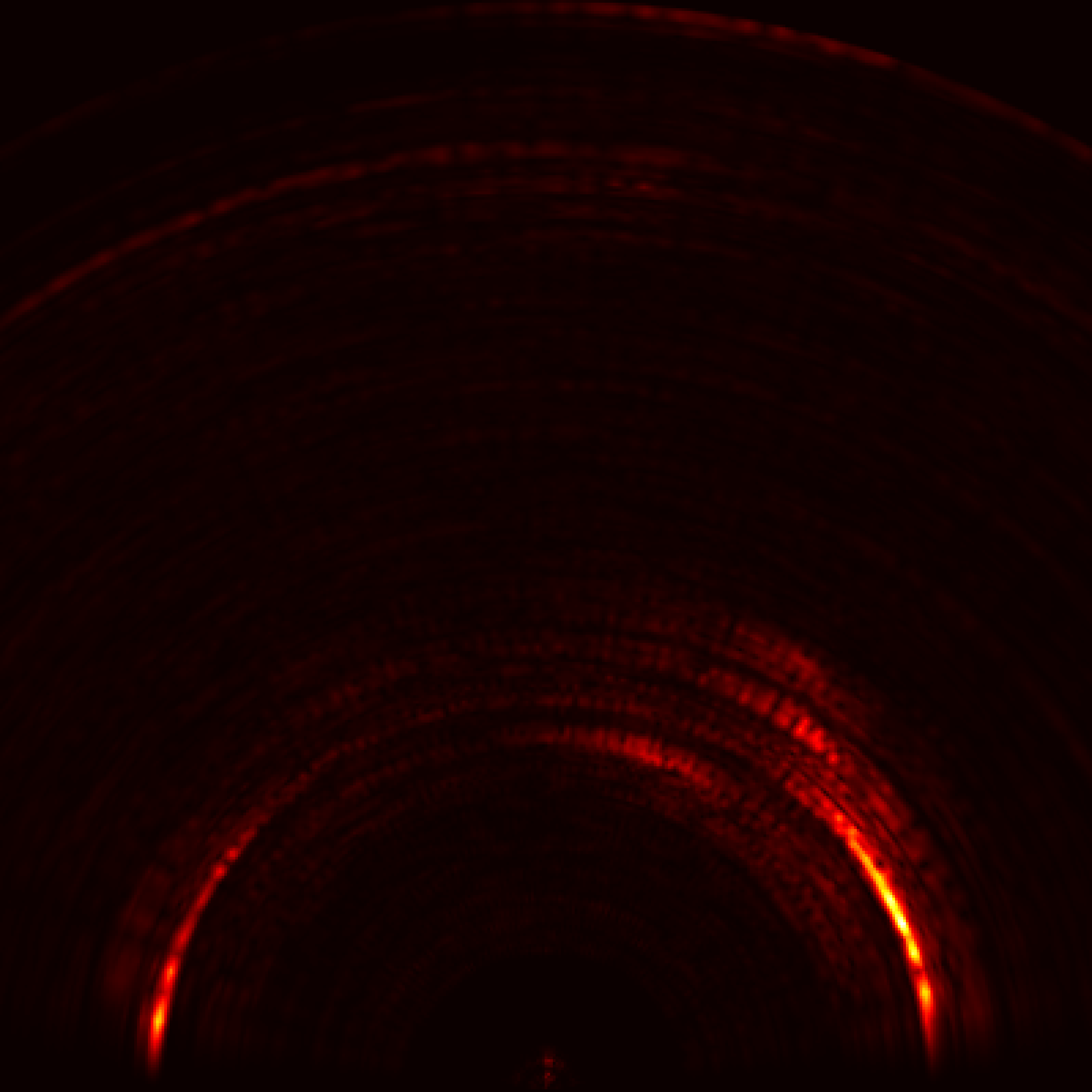} & \includegraphics[width=0.165\linewidth]{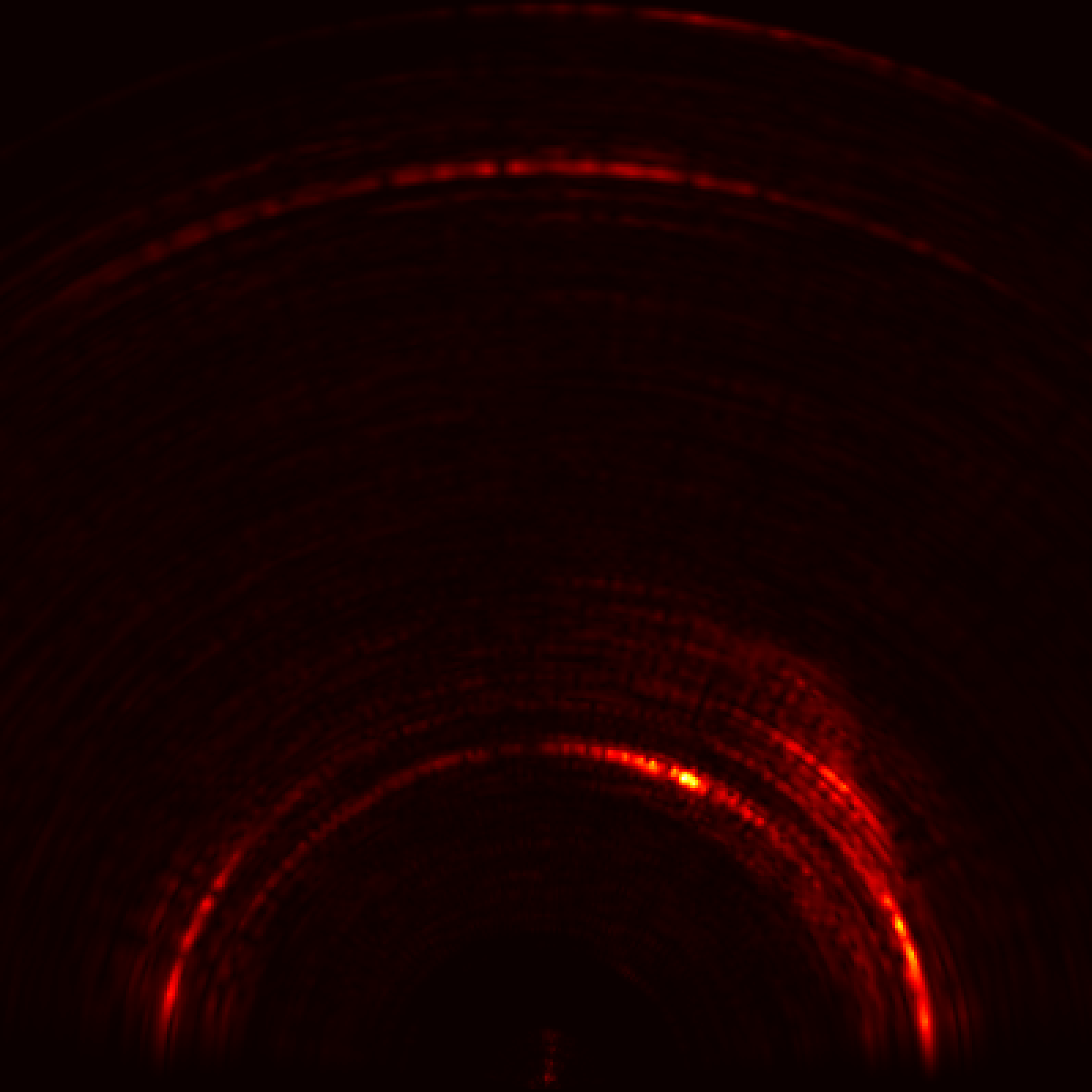} & \includegraphics[width=0.165\linewidth]{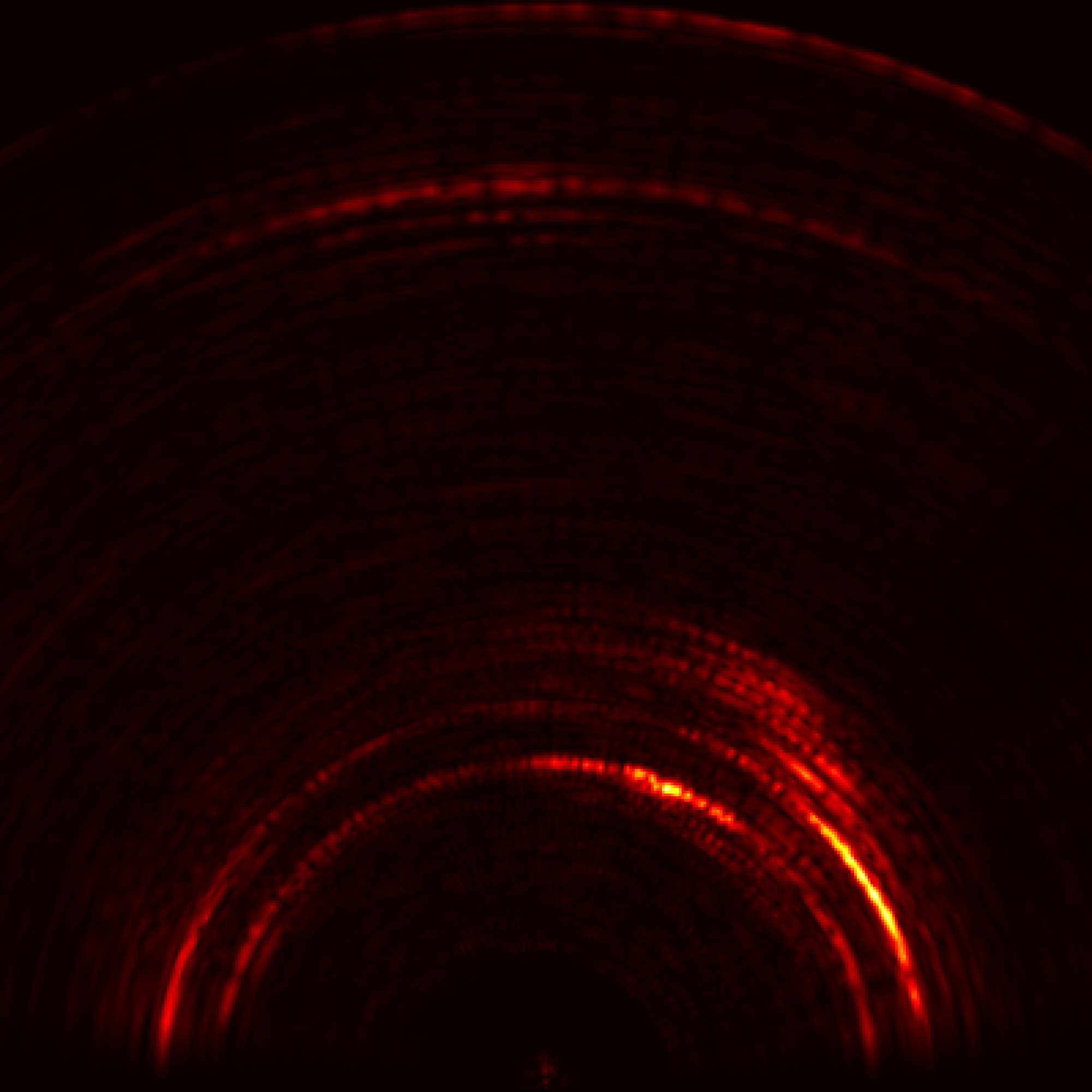} \\
\raisebox{2.0em}{\rotatebox{90}{\scriptsize\textbf{mmIR}}} & \includegraphics[width=0.165\linewidth]{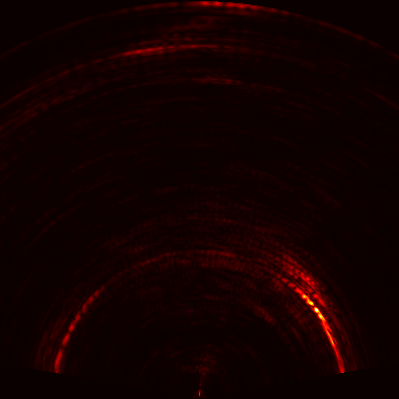} & \includegraphics[width=0.165\linewidth]{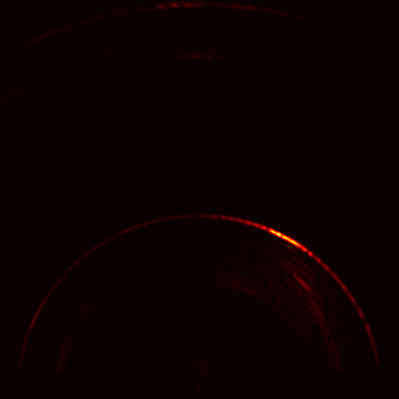} & \includegraphics[width=0.165\linewidth]{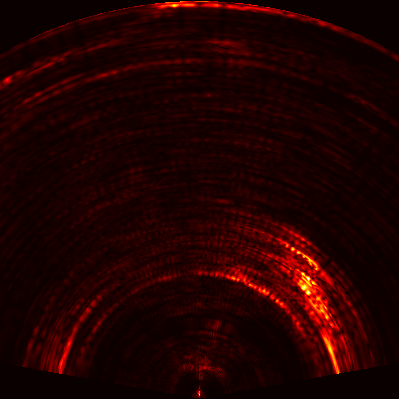} & \includegraphics[width=0.165\linewidth]{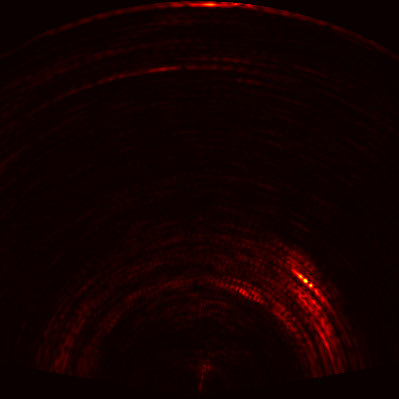} \\
\raisebox{2.0em}{\rotatebox{90}{\scriptsize\textbf{RadarSplat}}} & \includegraphics[width=0.165\linewidth]{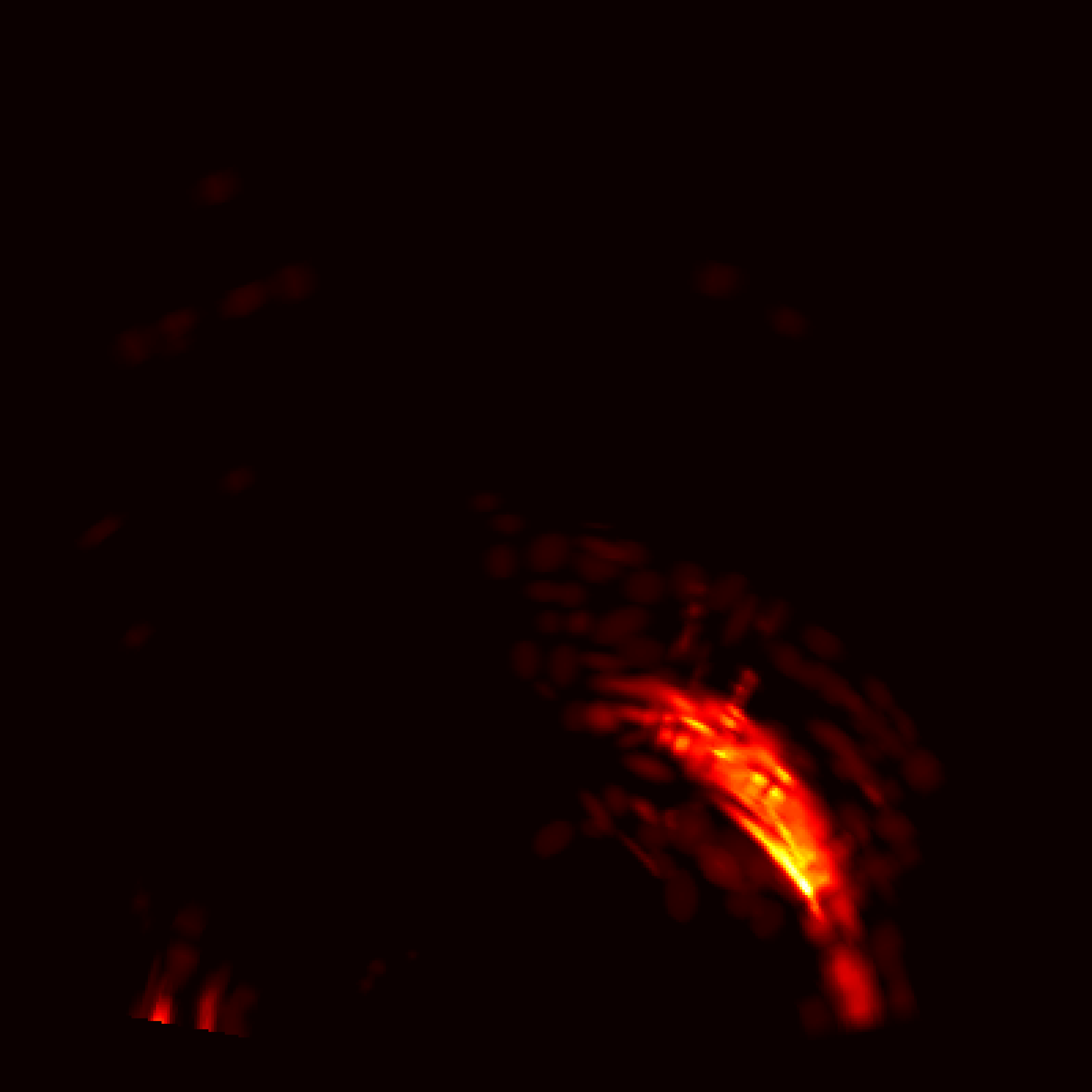} & \includegraphics[width=0.165\linewidth]{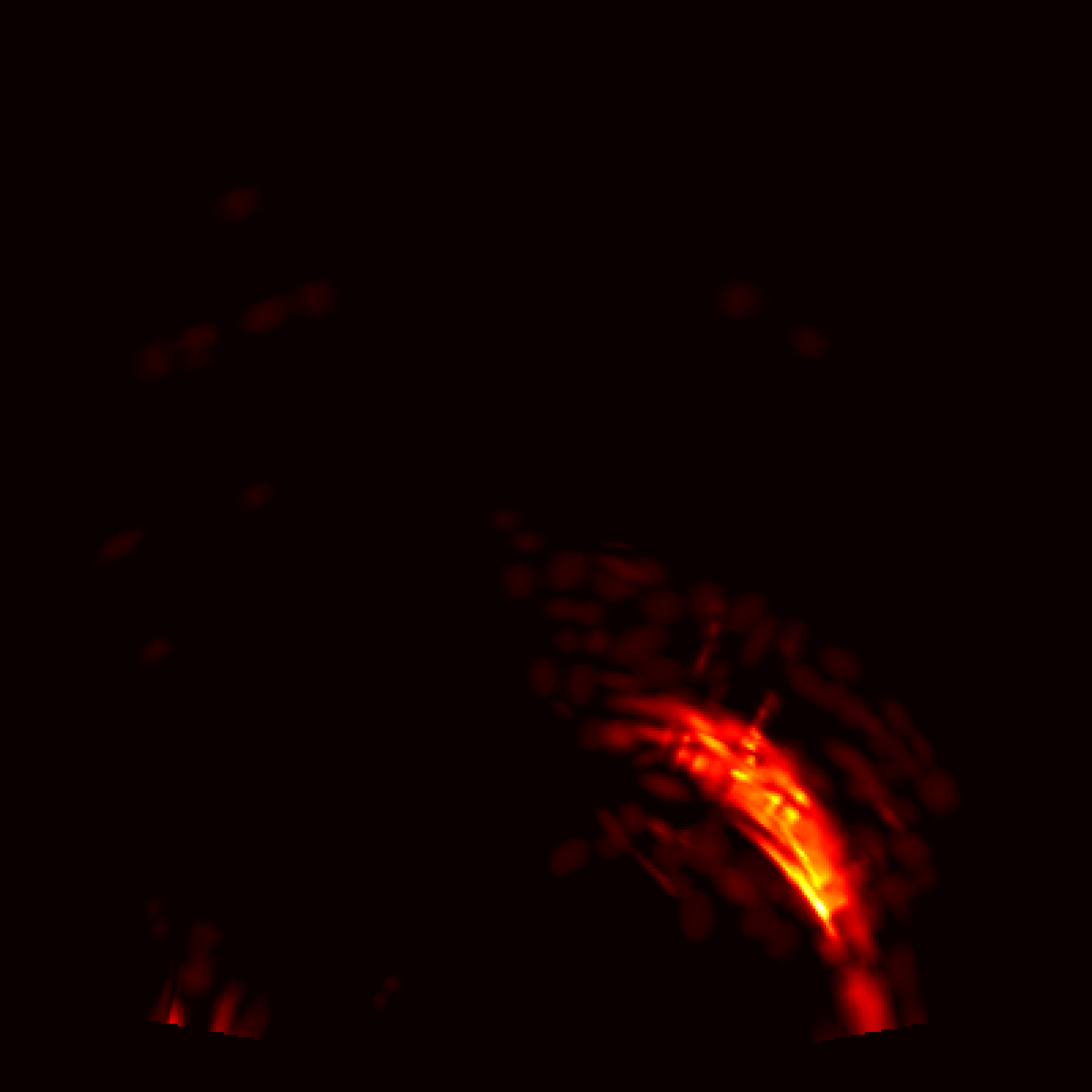} & \includegraphics[width=0.165\linewidth]{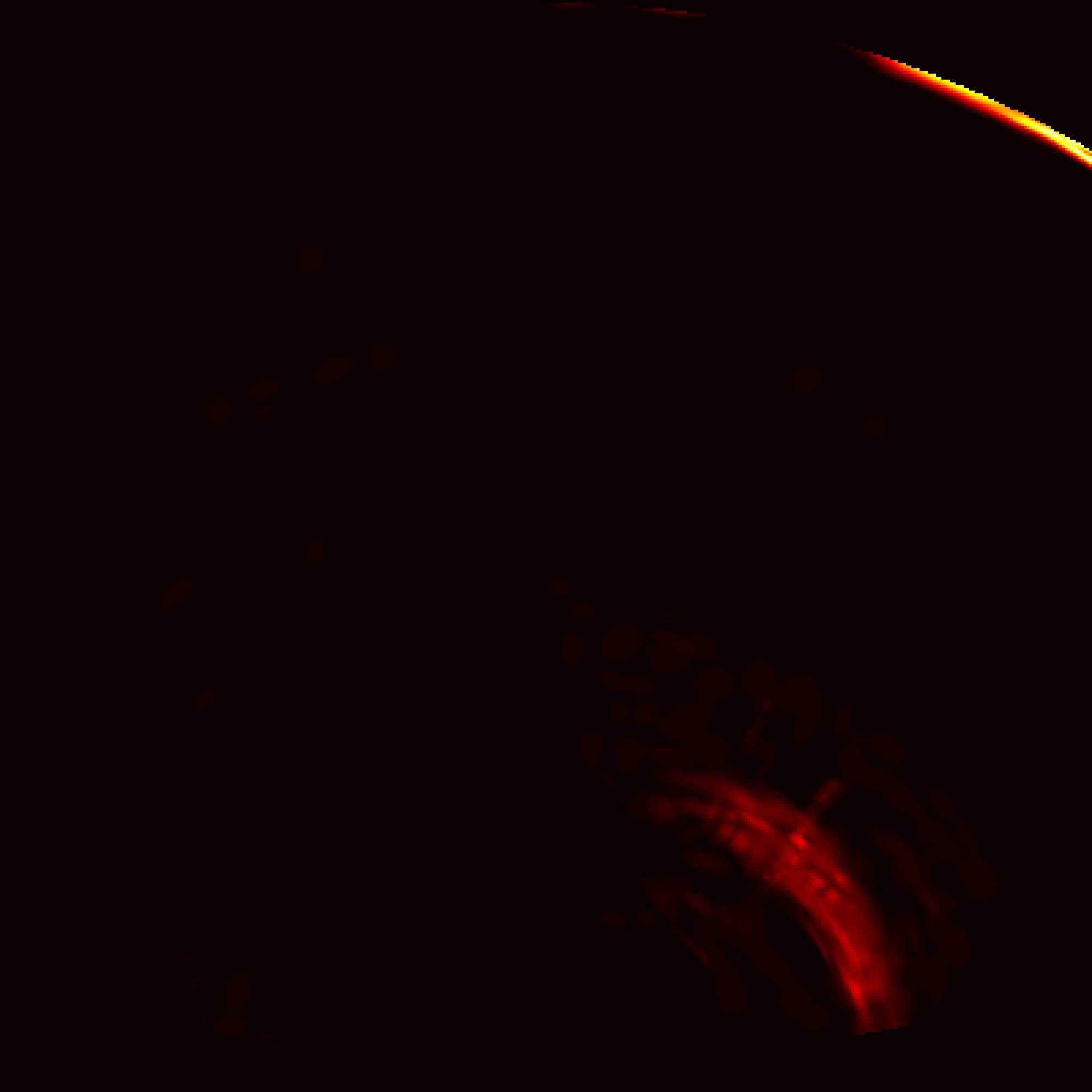} & \includegraphics[width=0.165\linewidth]{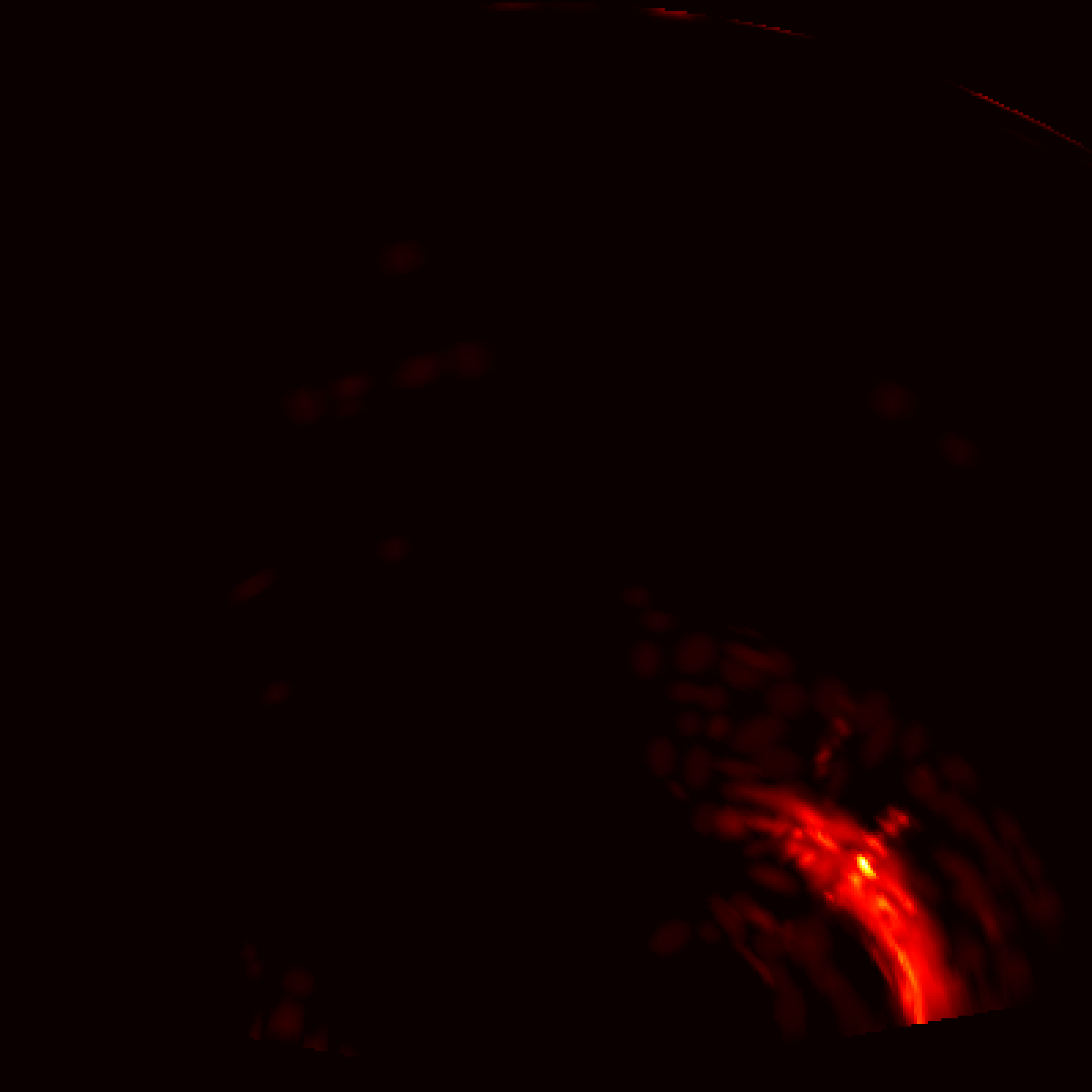} \\
\raisebox{2.0em}{\rotatebox{90}{\scriptsize\textbf{Radar Fields}}} & \includegraphics[width=0.165\linewidth]{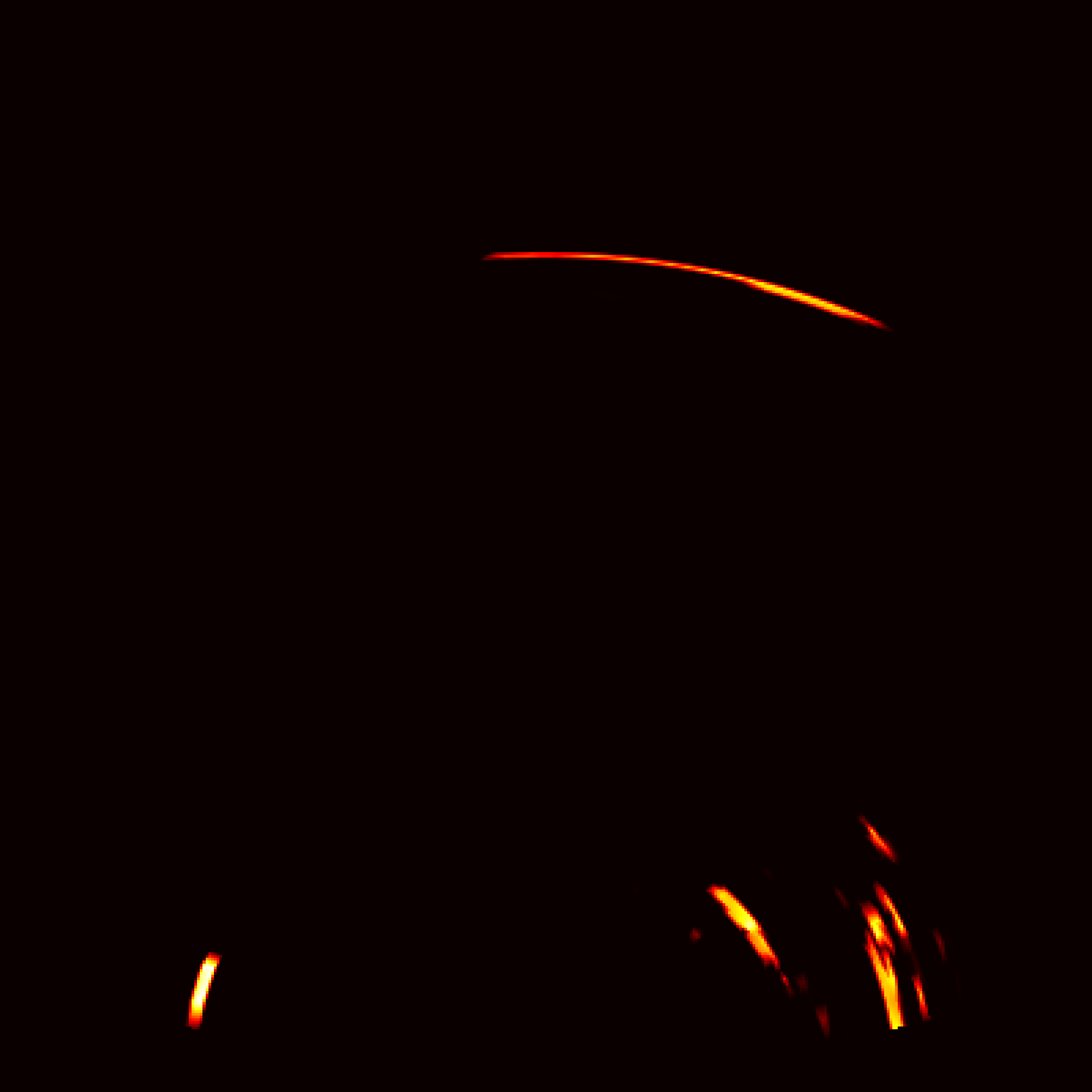} & \includegraphics[width=0.165\linewidth]{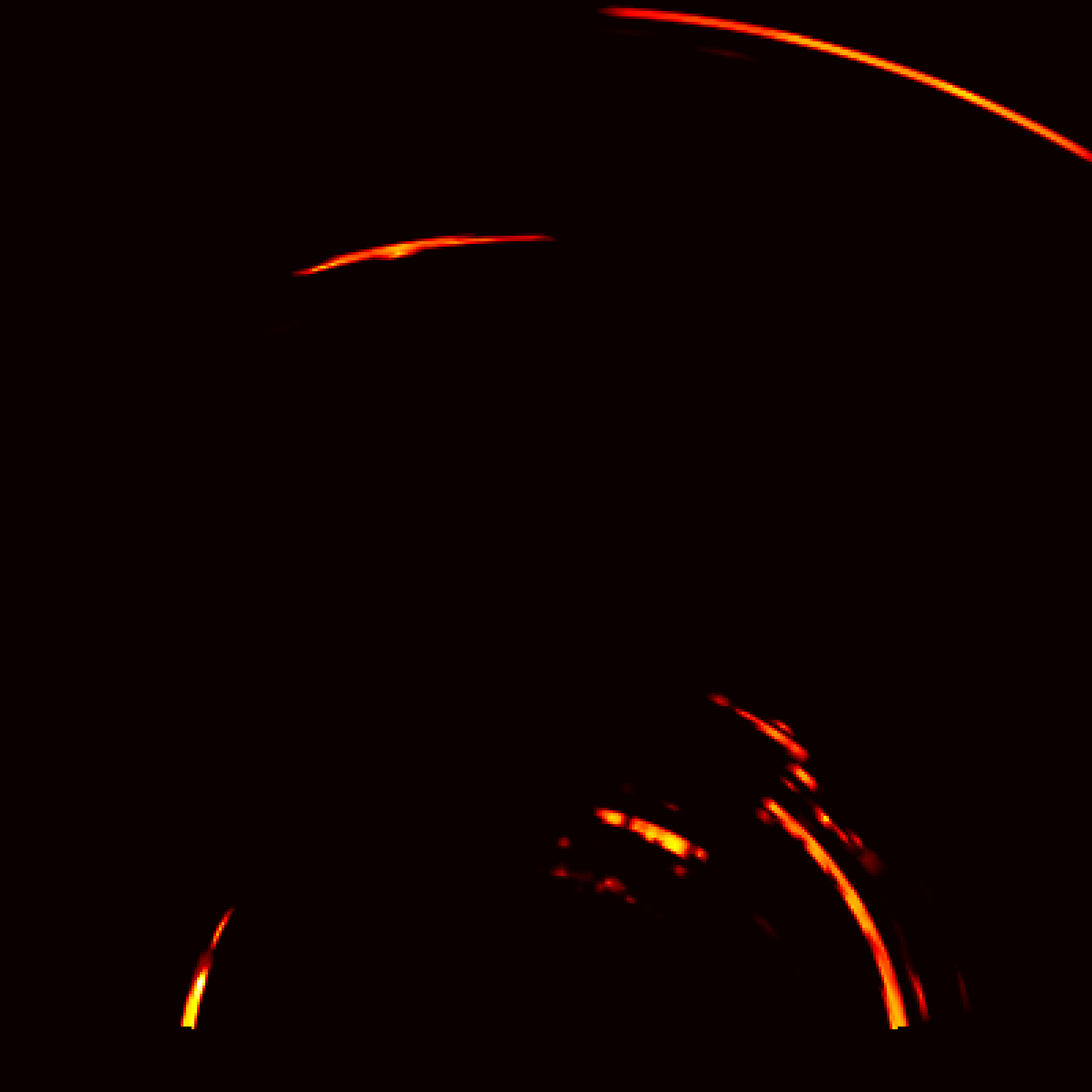} & \includegraphics[width=0.165\linewidth]{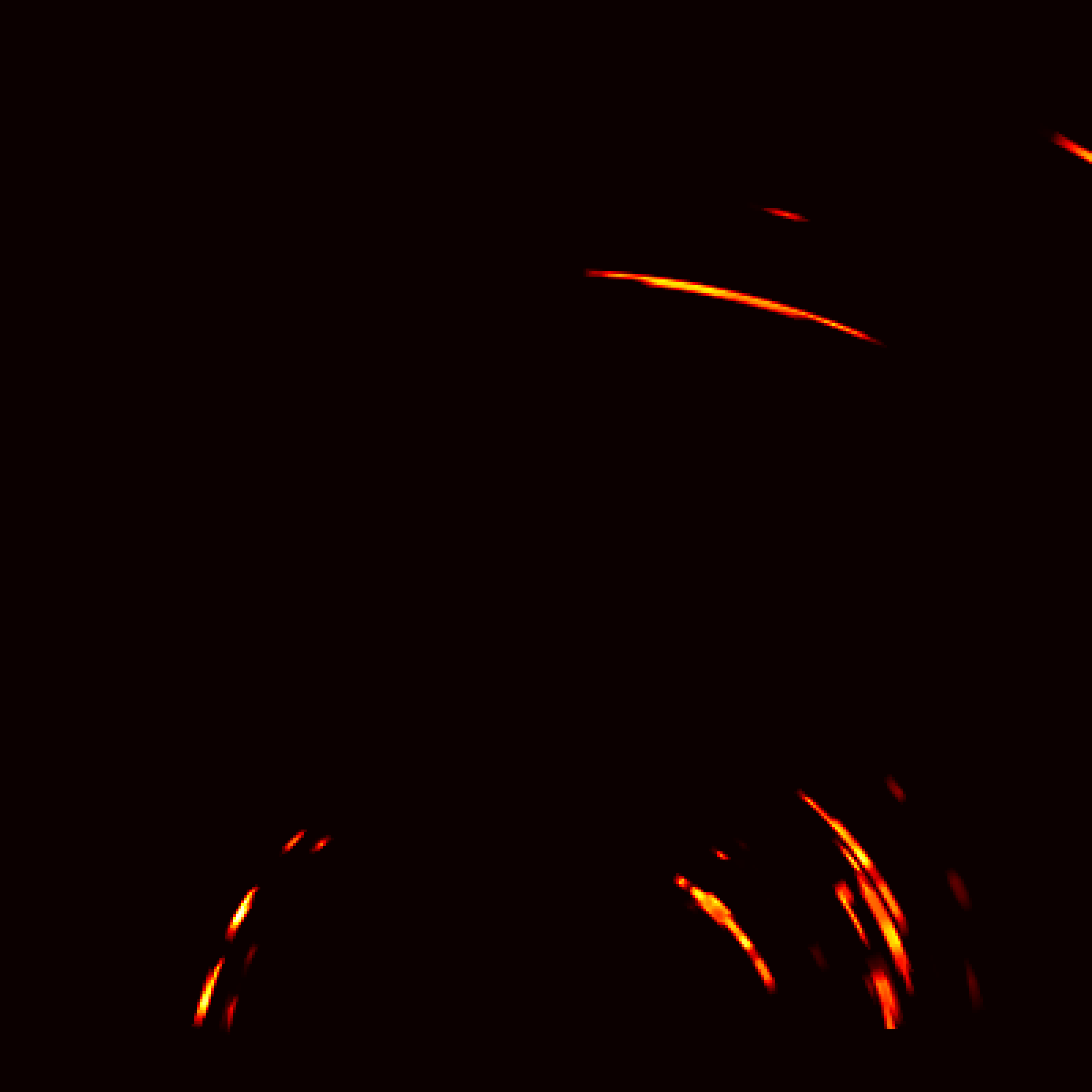} & \includegraphics[width=0.165\linewidth]{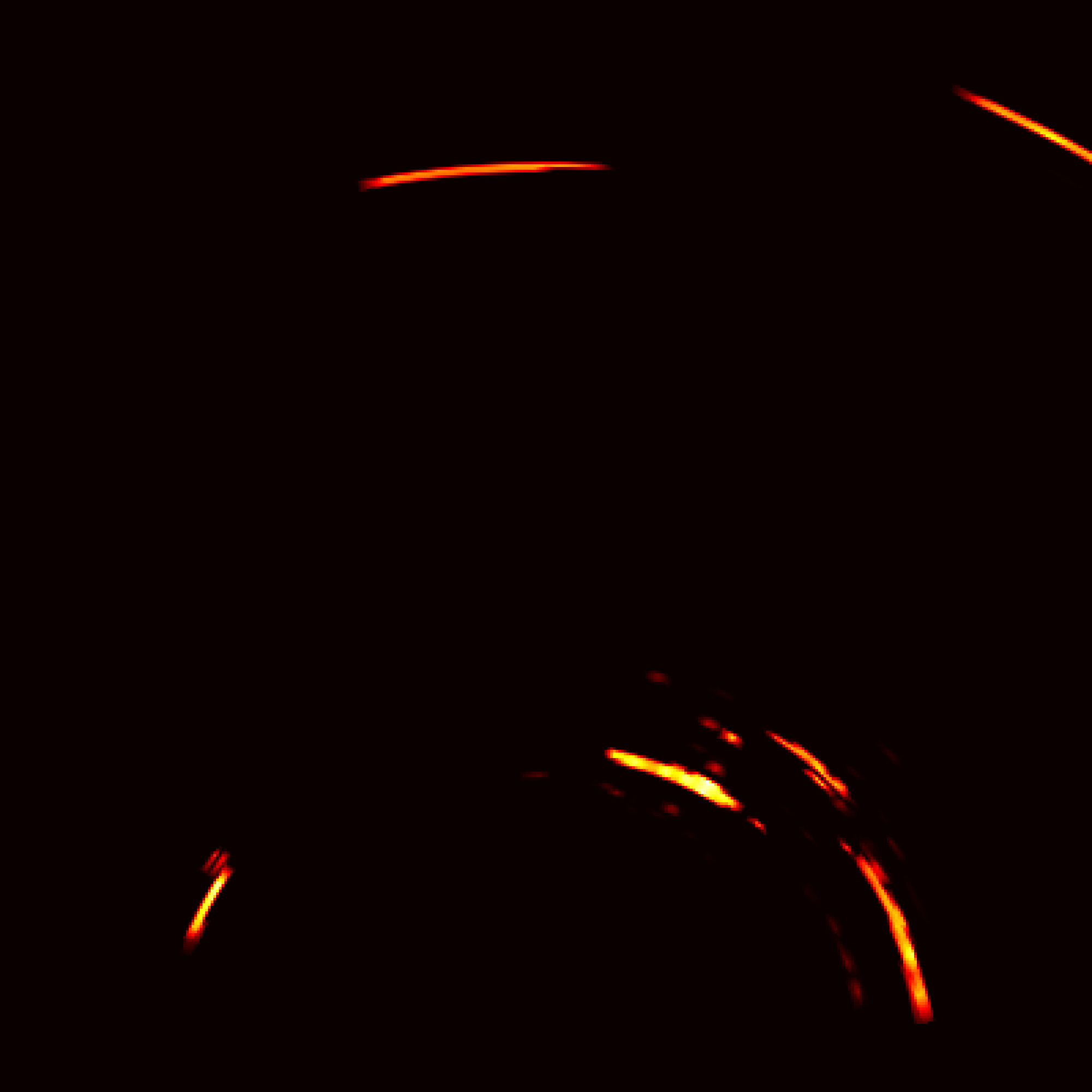} \\
\raisebox{2.0em}{\rotatebox{90}{\scriptsize\textbf{DART}}} & \includegraphics[width=0.165\linewidth]{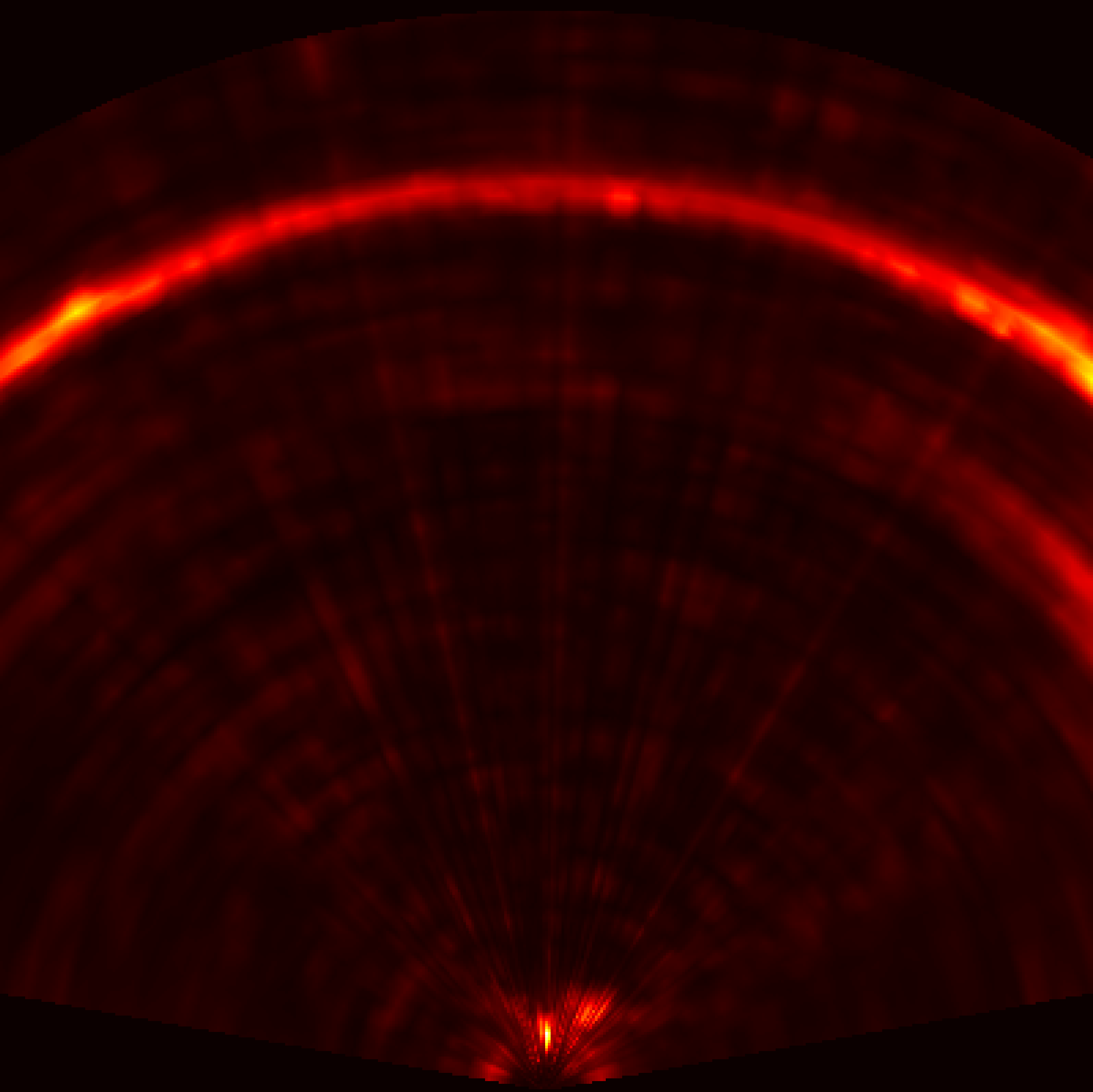} & \includegraphics[width=0.165\linewidth]{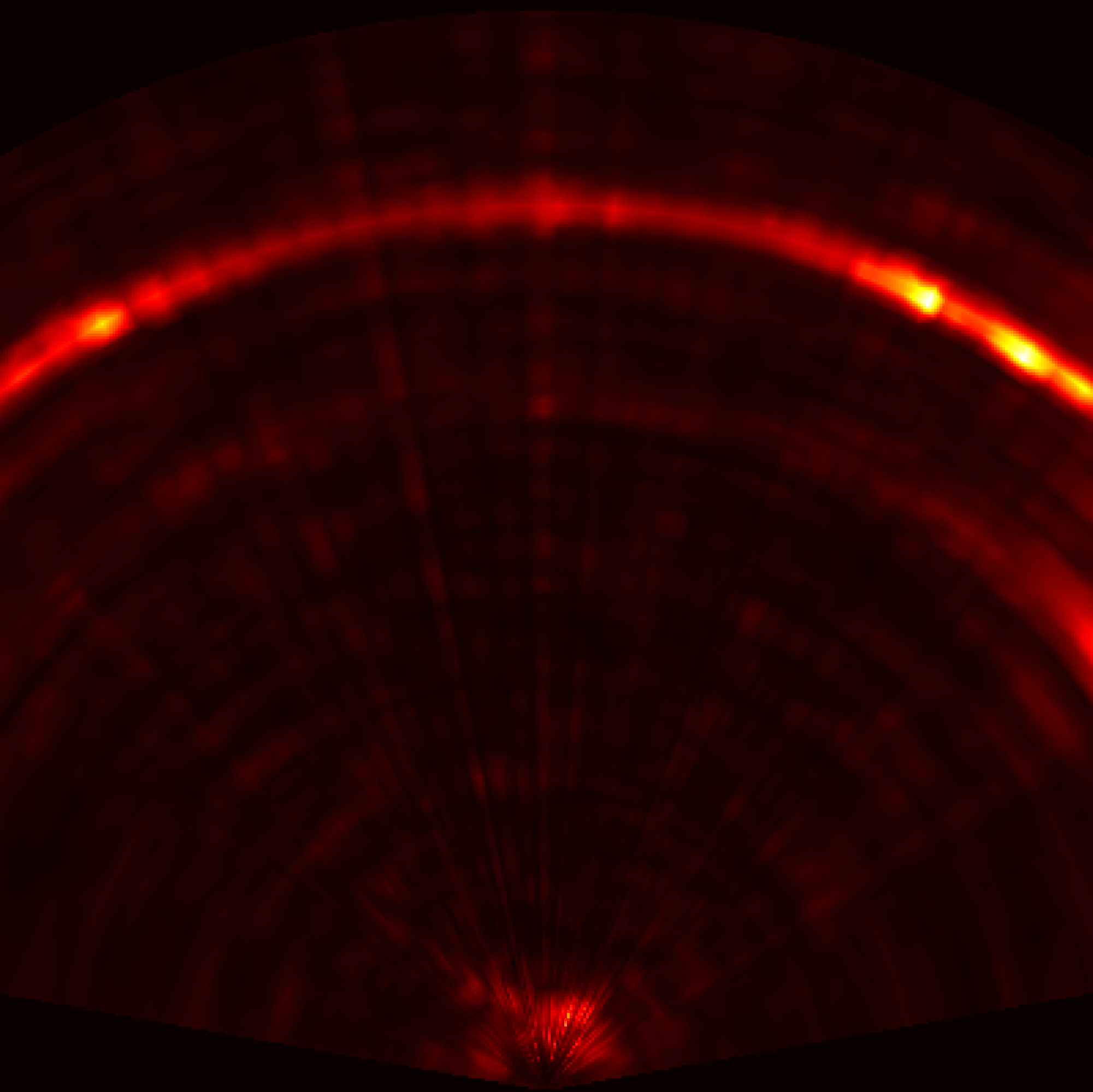} & \includegraphics[width=0.165\linewidth]{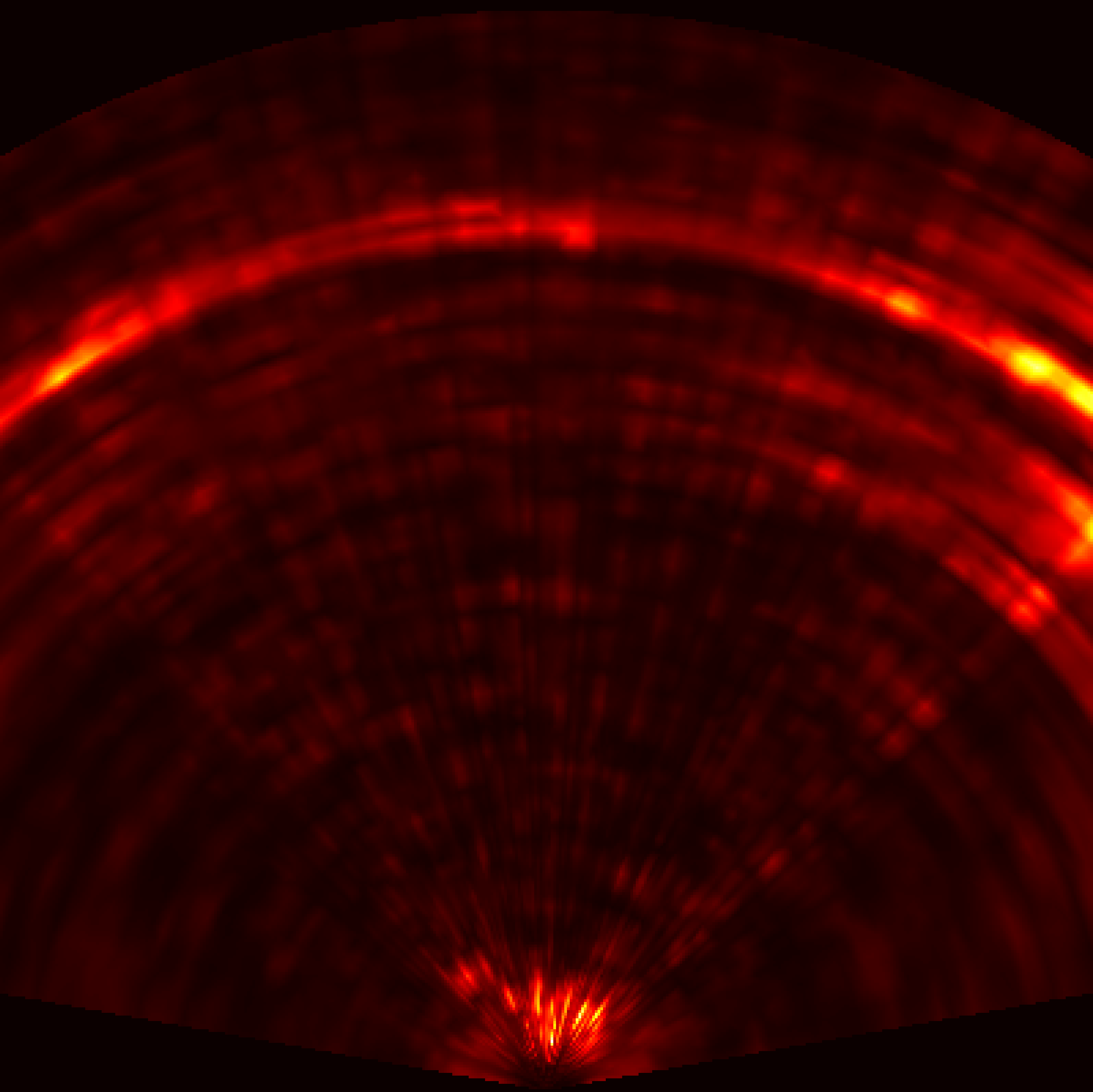} & \includegraphics[width=0.165\linewidth]{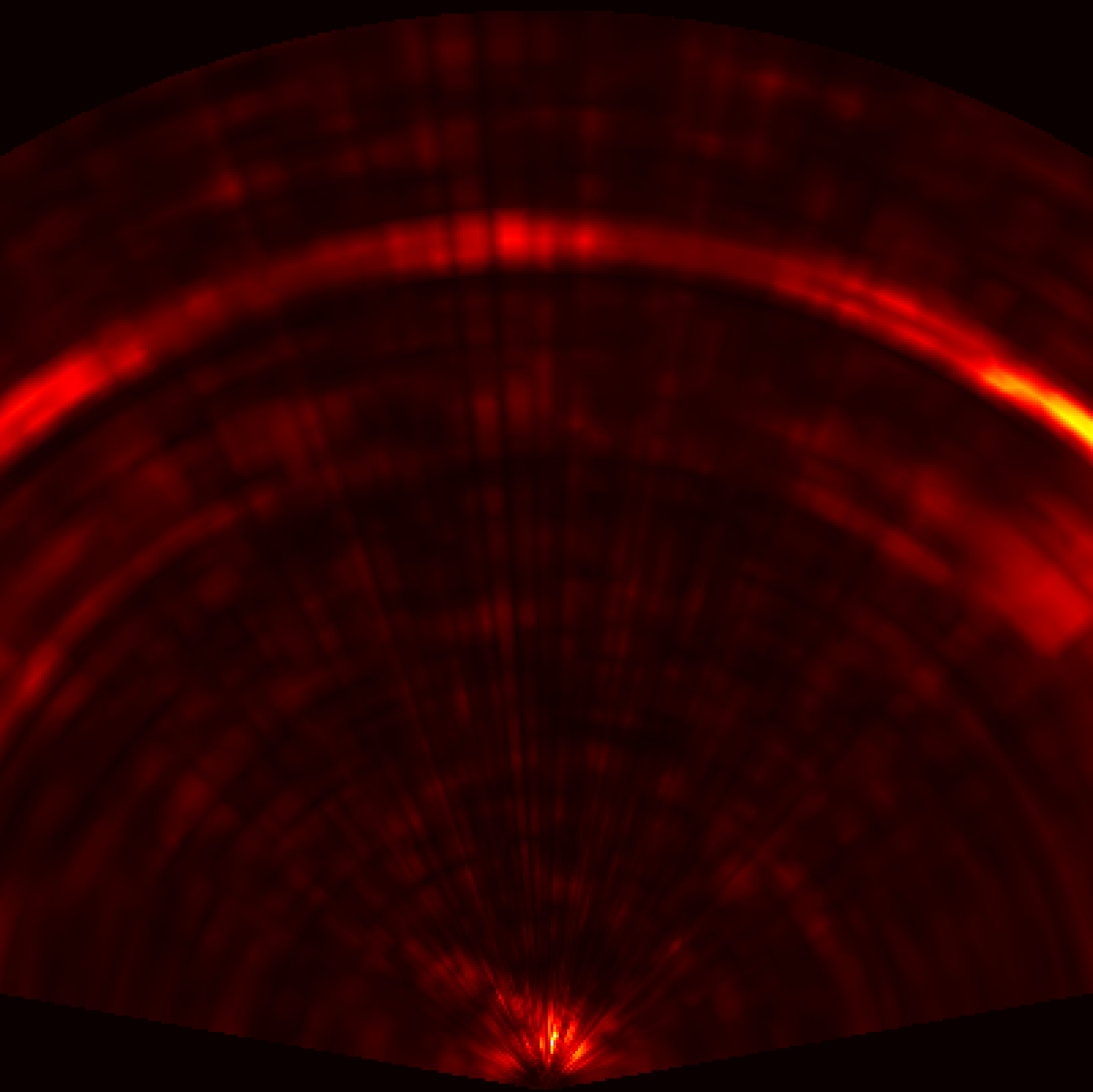} \\
\end{tabular}
\caption{\textbf{Per-frame $|\mathrm{RA}|$ comparison, scene S0\,F135, training frames $F\!-\!4$ to $F\!-\!1$.} Pearson correlation per column, left to right. 3DPS: 0.90, 0.85, 0.81, 0.84; mmIR: 0.69, 0.24, 0.61, 0.55; RadarSplat: 0.19, 0.21, 0.38, 0.41; Radar Fields: 0.05, 0.21, 0.13, 0.28; DART: -0.04, -0.04, -0.04, -0.04.}
\label{fig:supp_s0_f135}
\end{figure}

\begin{figure}[p]
\centering
\setlength{\tabcolsep}{1pt}\renewcommand{\arraystretch}{0.6}
\begin{tabular}{cccccc}
 & {\footnotesize $F\!+\!1$} & {\footnotesize $F\!+\!2$} & {\footnotesize $F\!+\!3$} & {\footnotesize $F\!+\!4$} & \textcolor{red}{\footnotesize\textbf{$F$}} \\
\raisebox{2.0em}{\rotatebox{90}{\scriptsize\textbf{GT}}} & \includegraphics[width=0.165\linewidth]{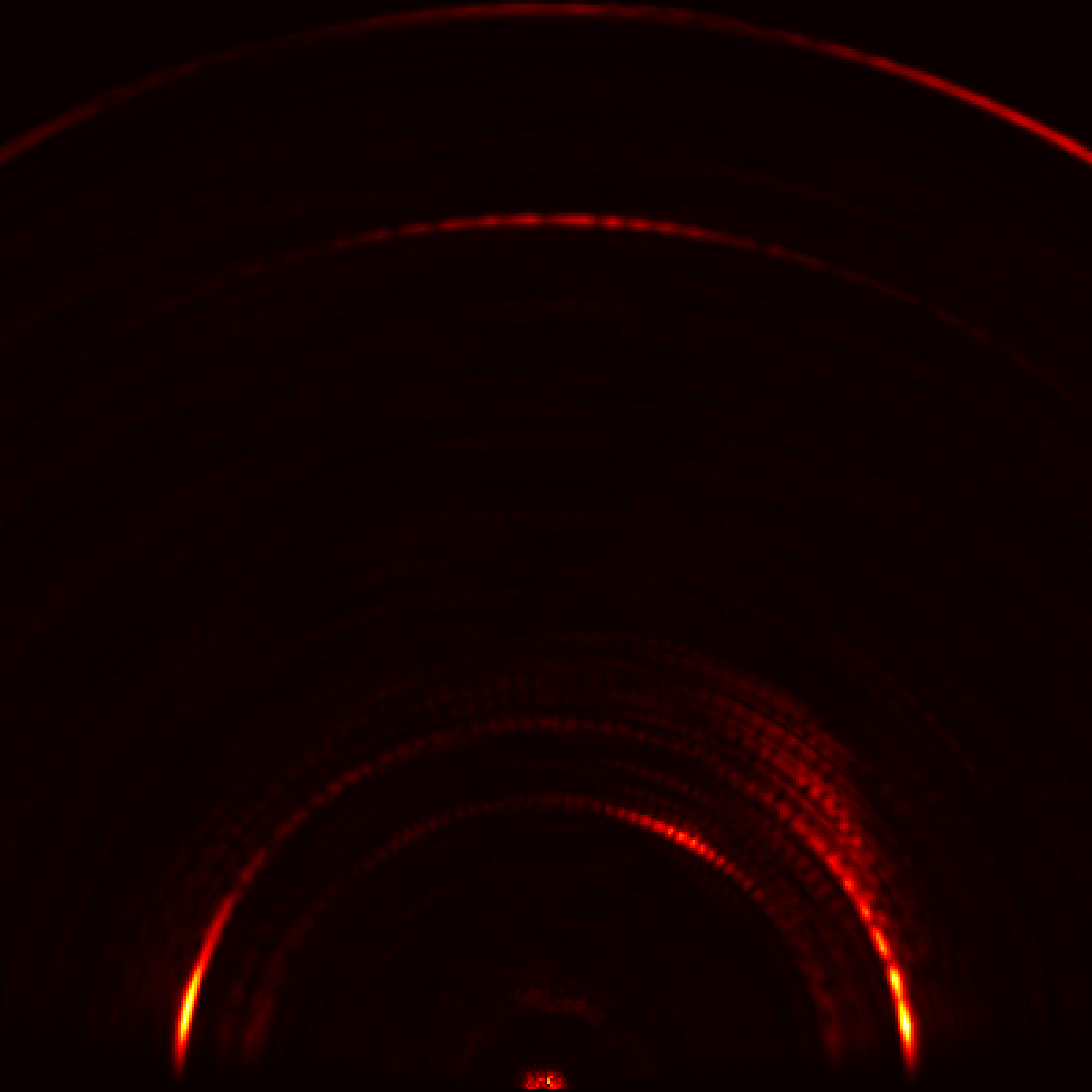} & \includegraphics[width=0.165\linewidth]{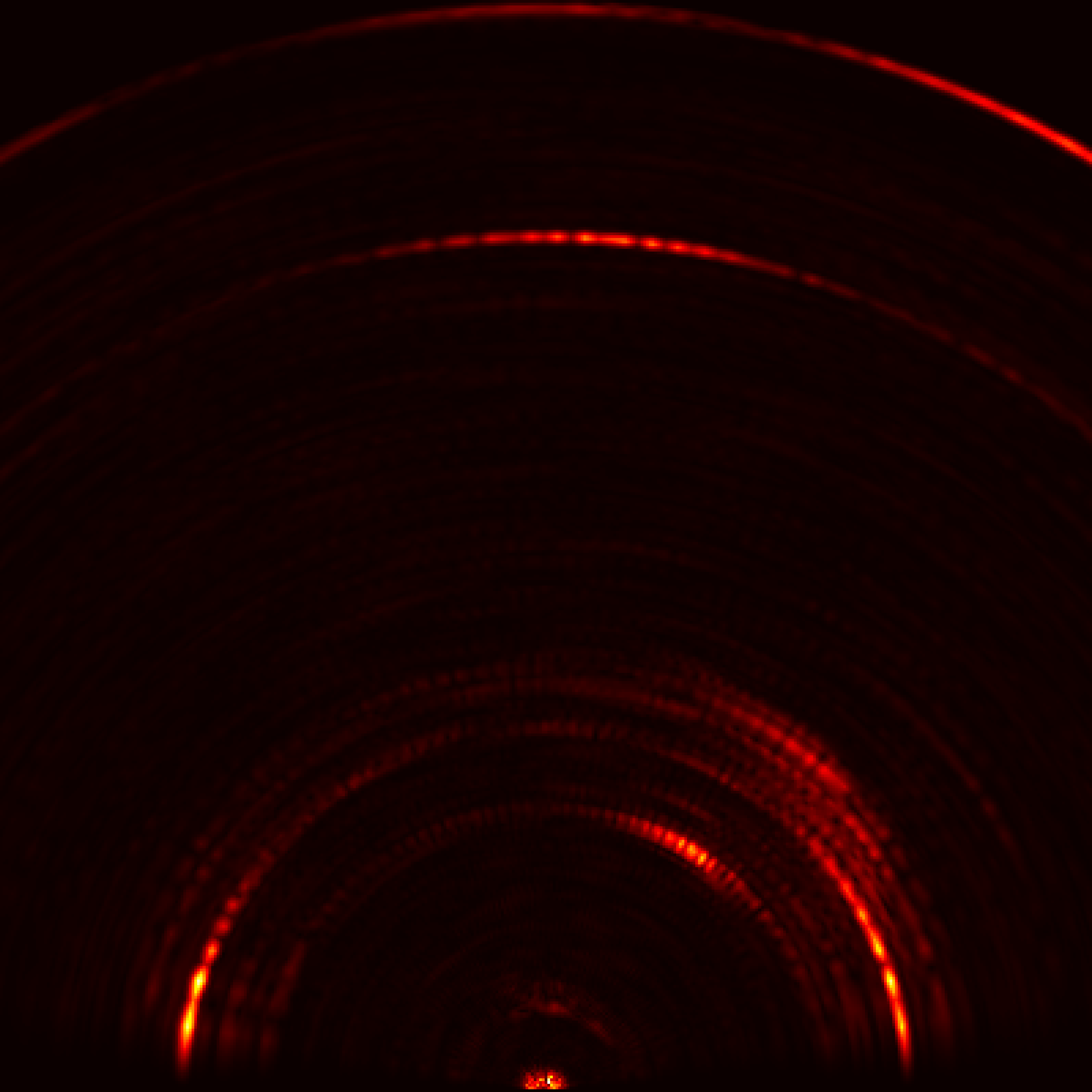} & \includegraphics[width=0.165\linewidth]{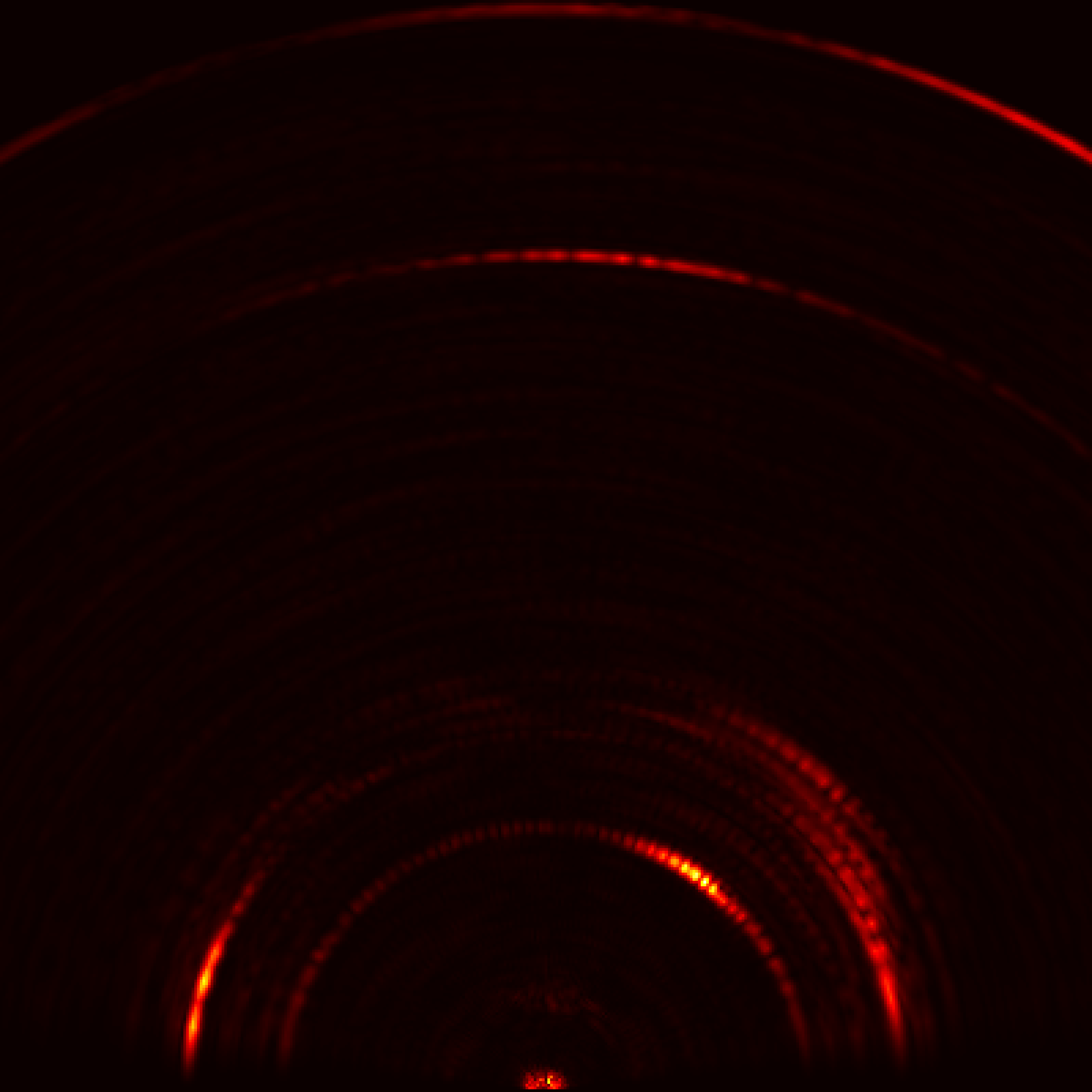} & \includegraphics[width=0.165\linewidth]{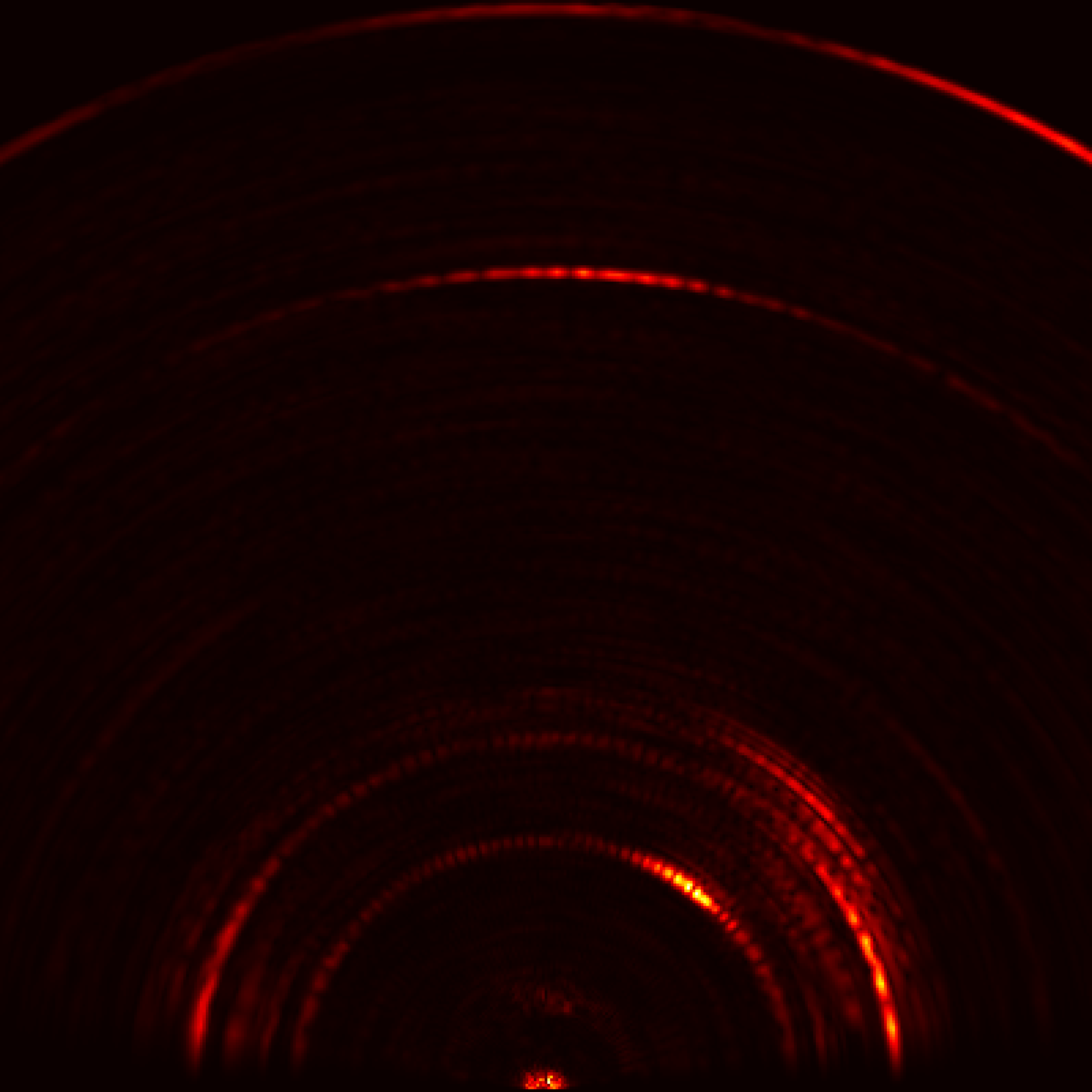} & \includegraphics[width=0.165\linewidth]{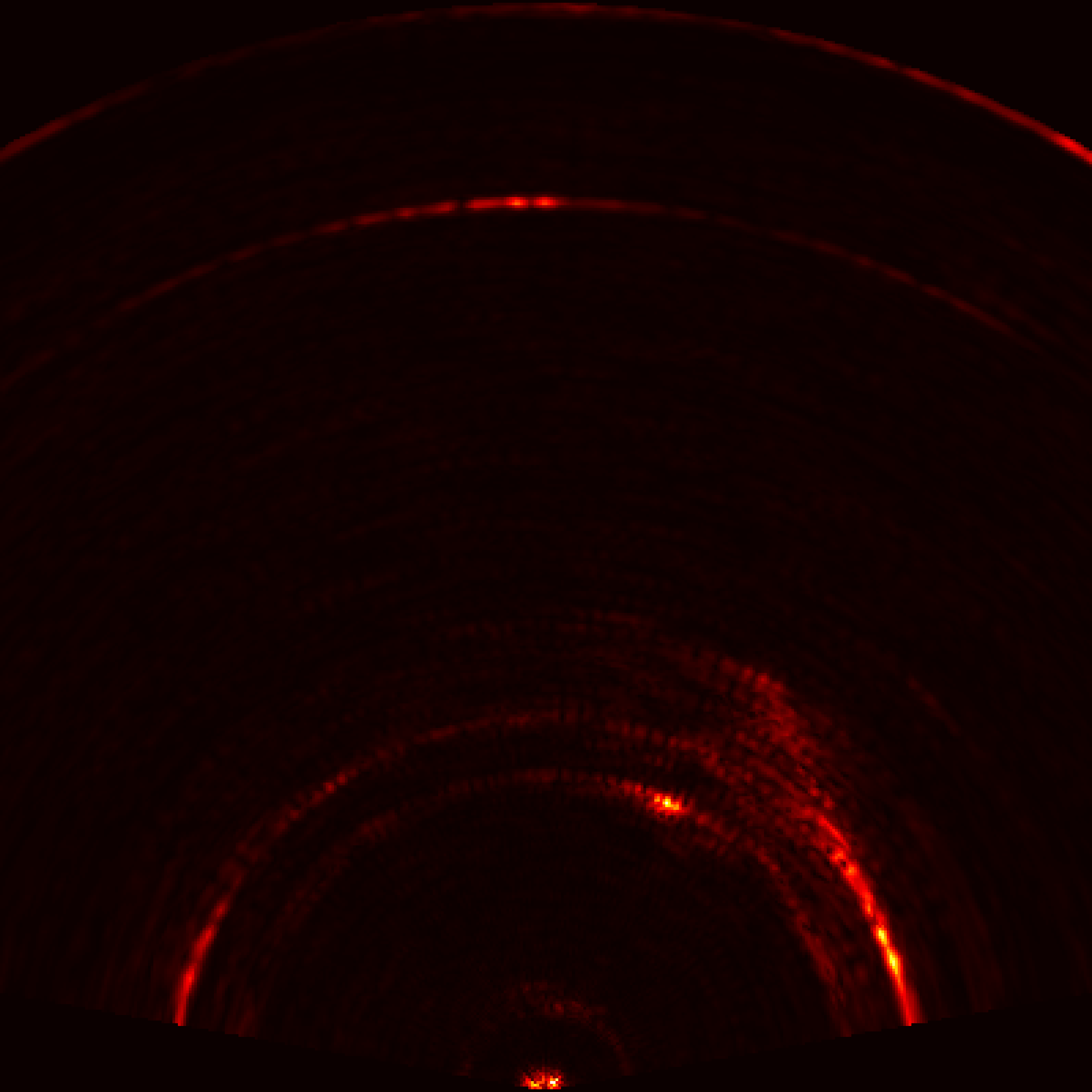} \\
\raisebox{2.0em}{\rotatebox{90}{\scriptsize\textbf{3DPS (Ours)}}} & \includegraphics[width=0.165\linewidth]{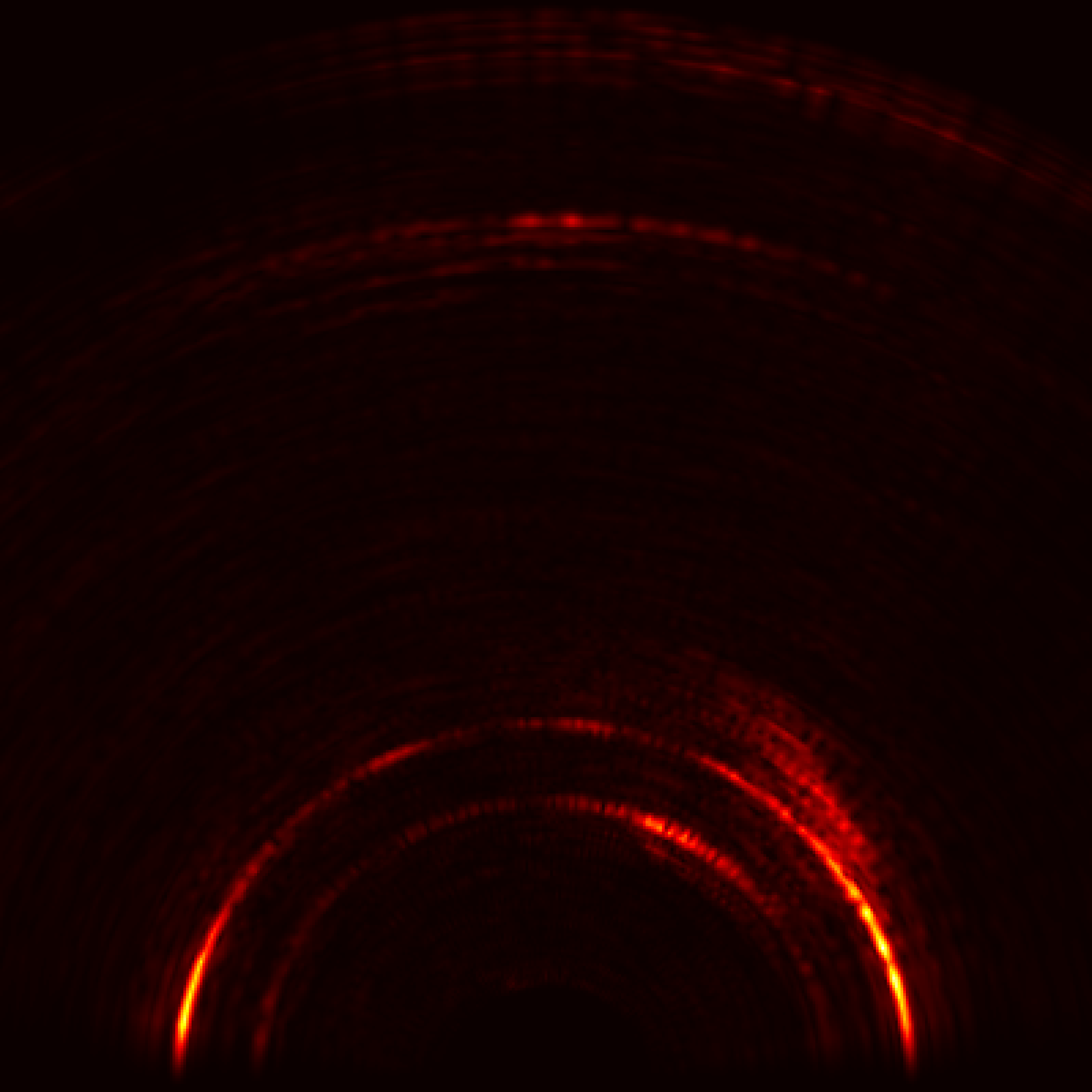} & \includegraphics[width=0.165\linewidth]{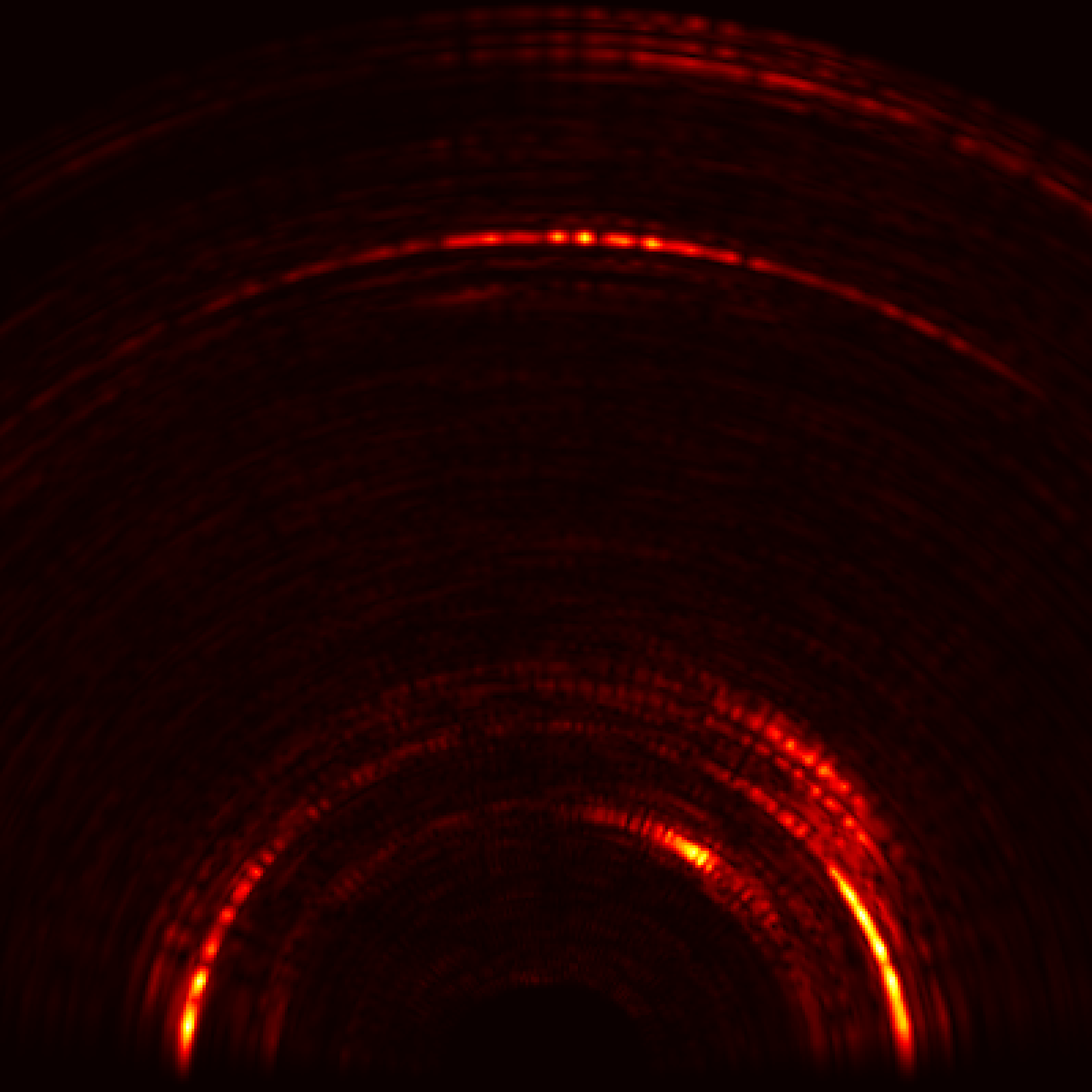} & \includegraphics[width=0.165\linewidth]{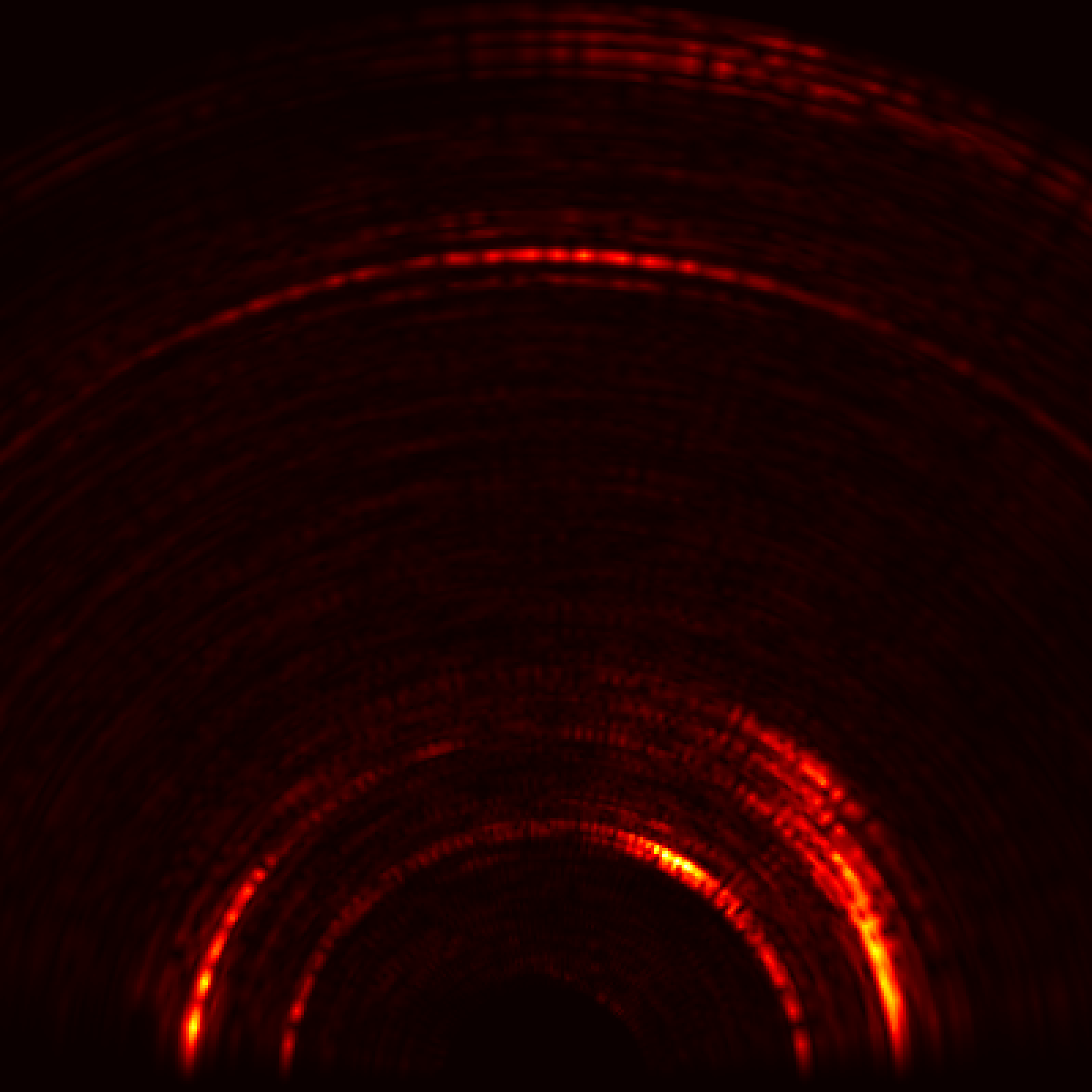} & \includegraphics[width=0.165\linewidth]{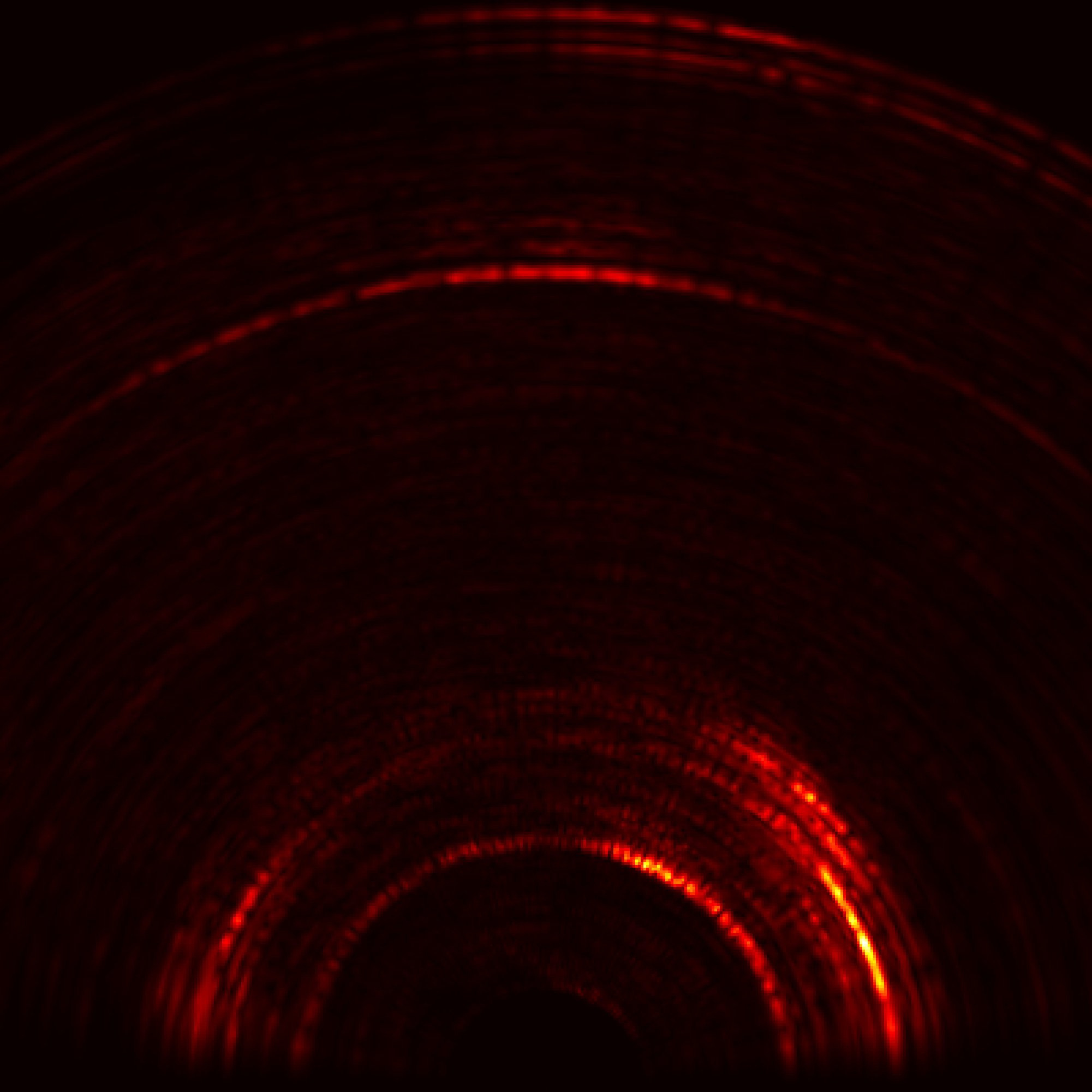} & \includegraphics[width=0.165\linewidth]{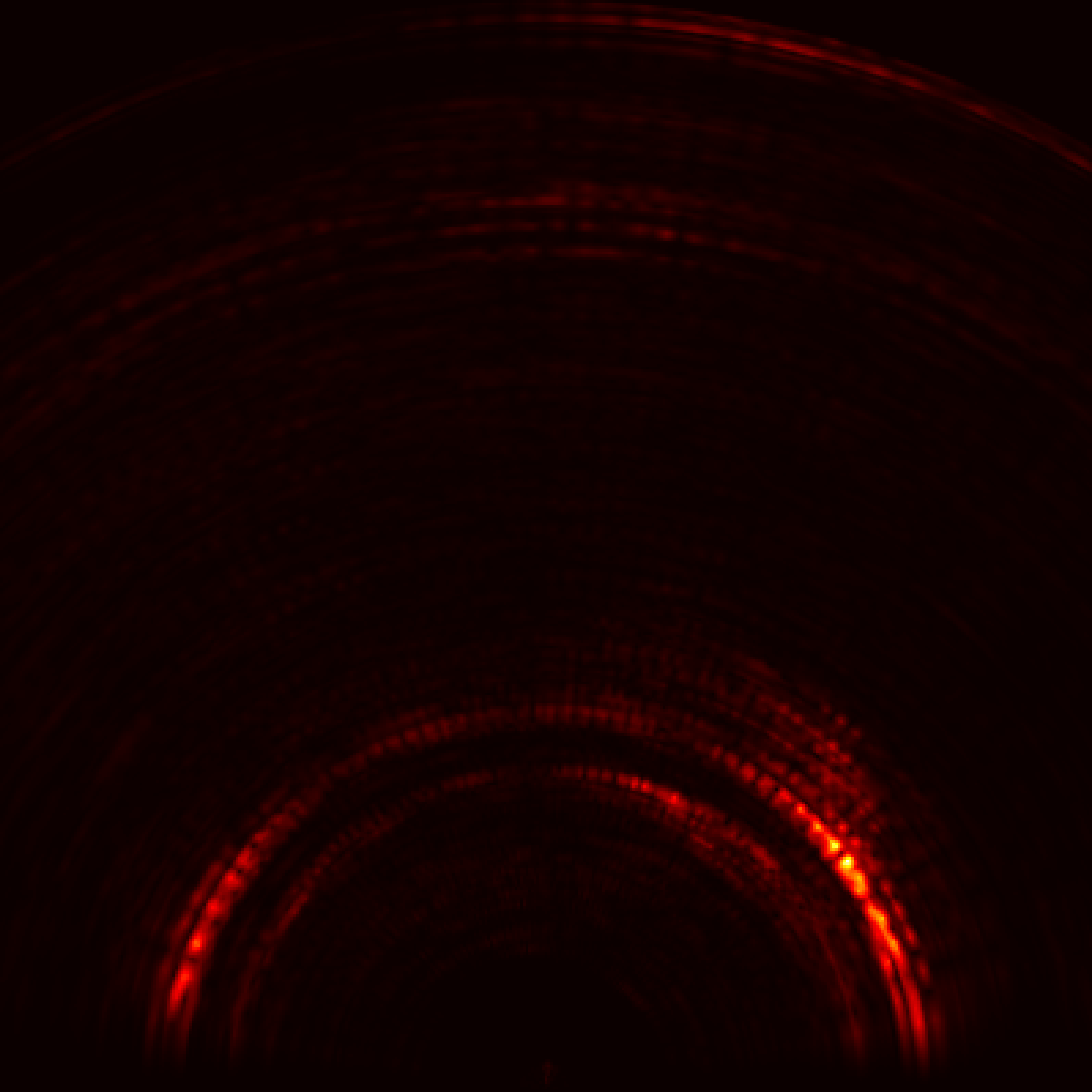} \\
\raisebox{2.0em}{\rotatebox{90}{\scriptsize\textbf{mmIR}}} & \includegraphics[width=0.165\linewidth]{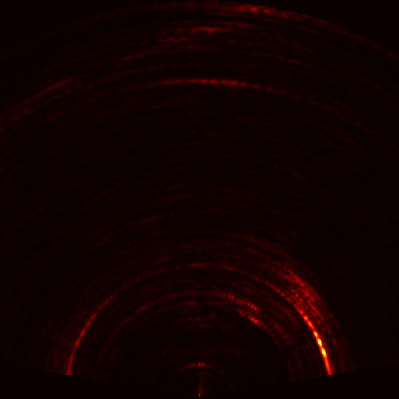} & \includegraphics[width=0.165\linewidth]{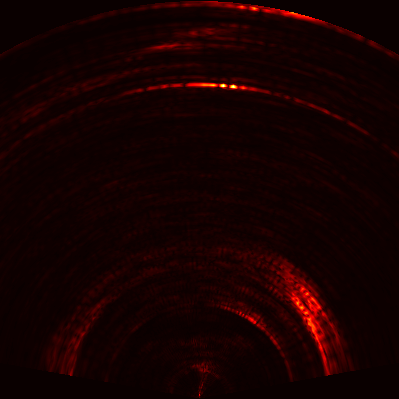} & \includegraphics[width=0.165\linewidth]{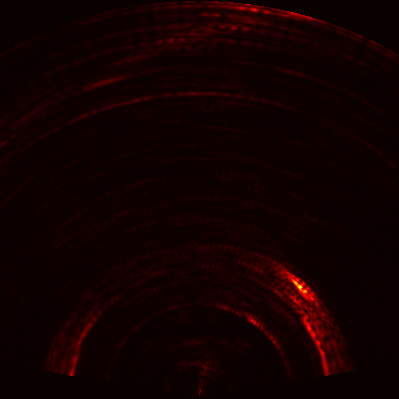} & \includegraphics[width=0.165\linewidth]{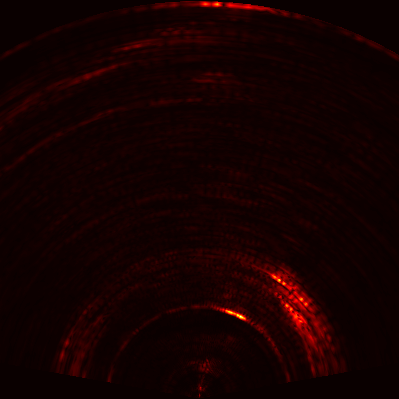} & \includegraphics[width=0.165\linewidth]{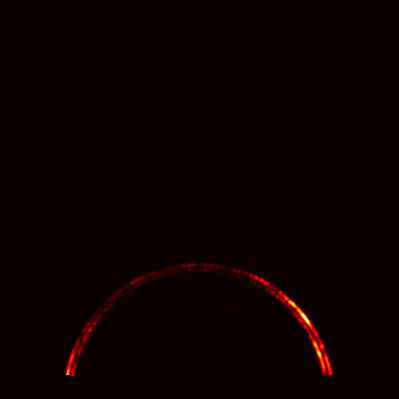} \\
\raisebox{2.0em}{\rotatebox{90}{\scriptsize\textbf{RadarSplat}}} & \includegraphics[width=0.165\linewidth]{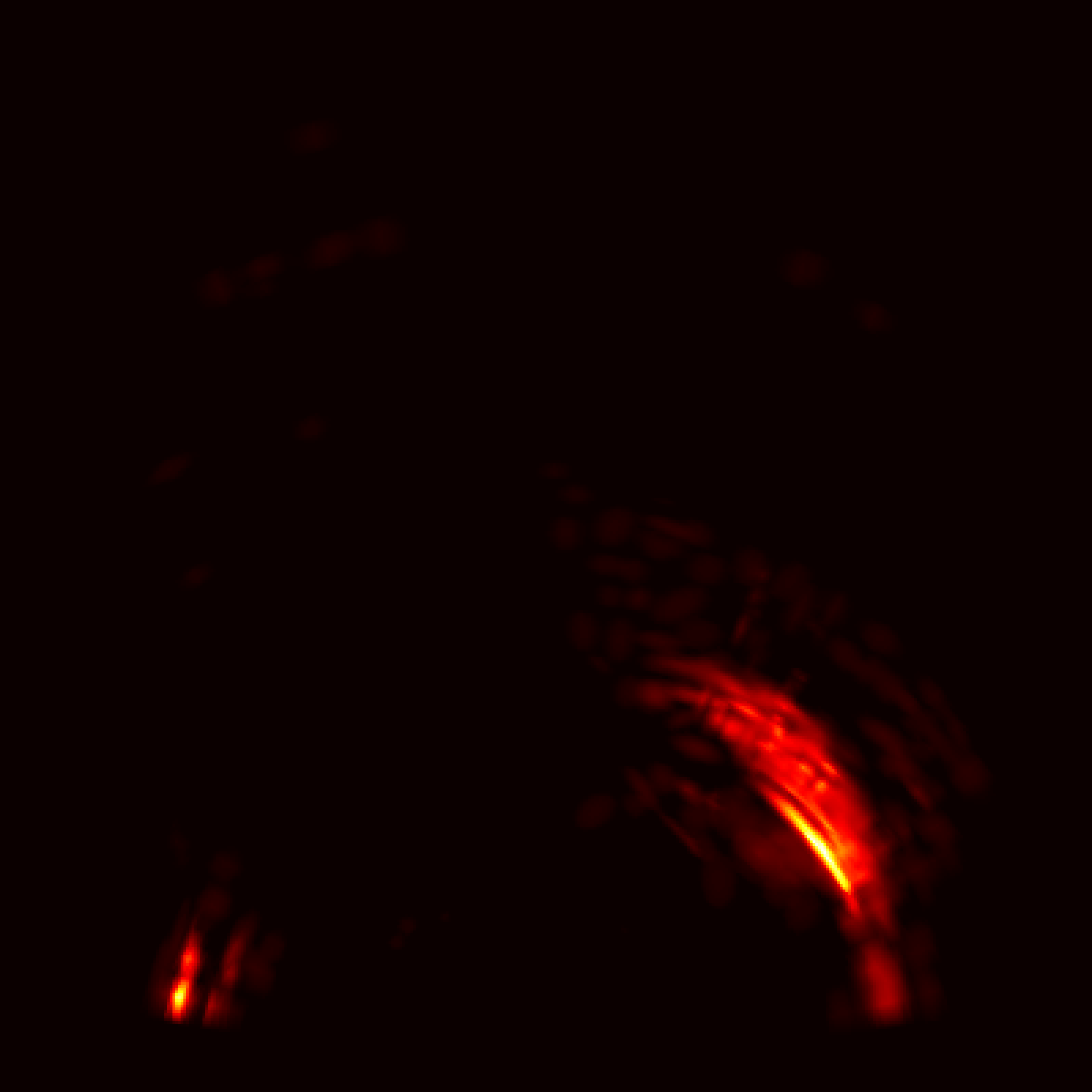} & \includegraphics[width=0.165\linewidth]{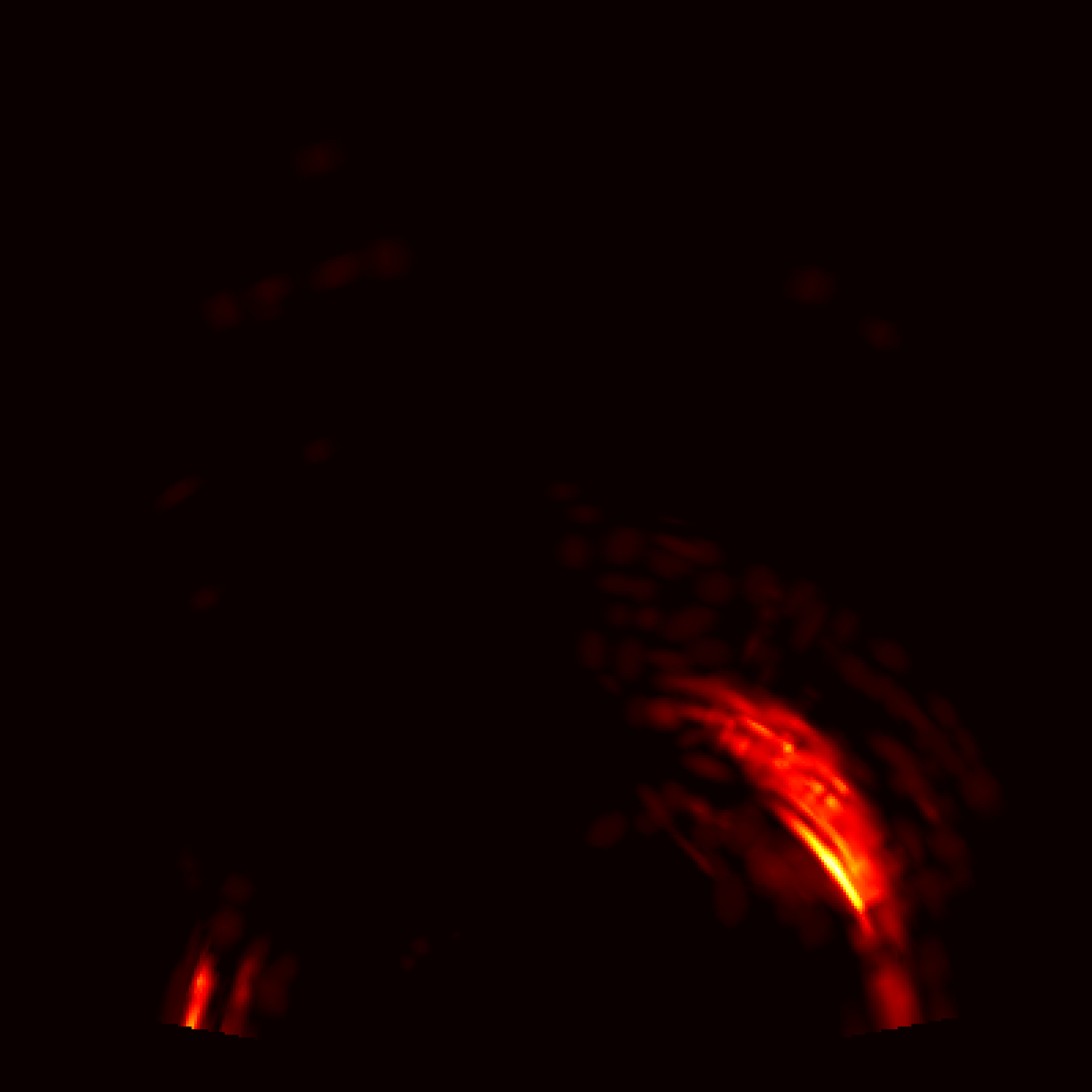} & \includegraphics[width=0.165\linewidth]{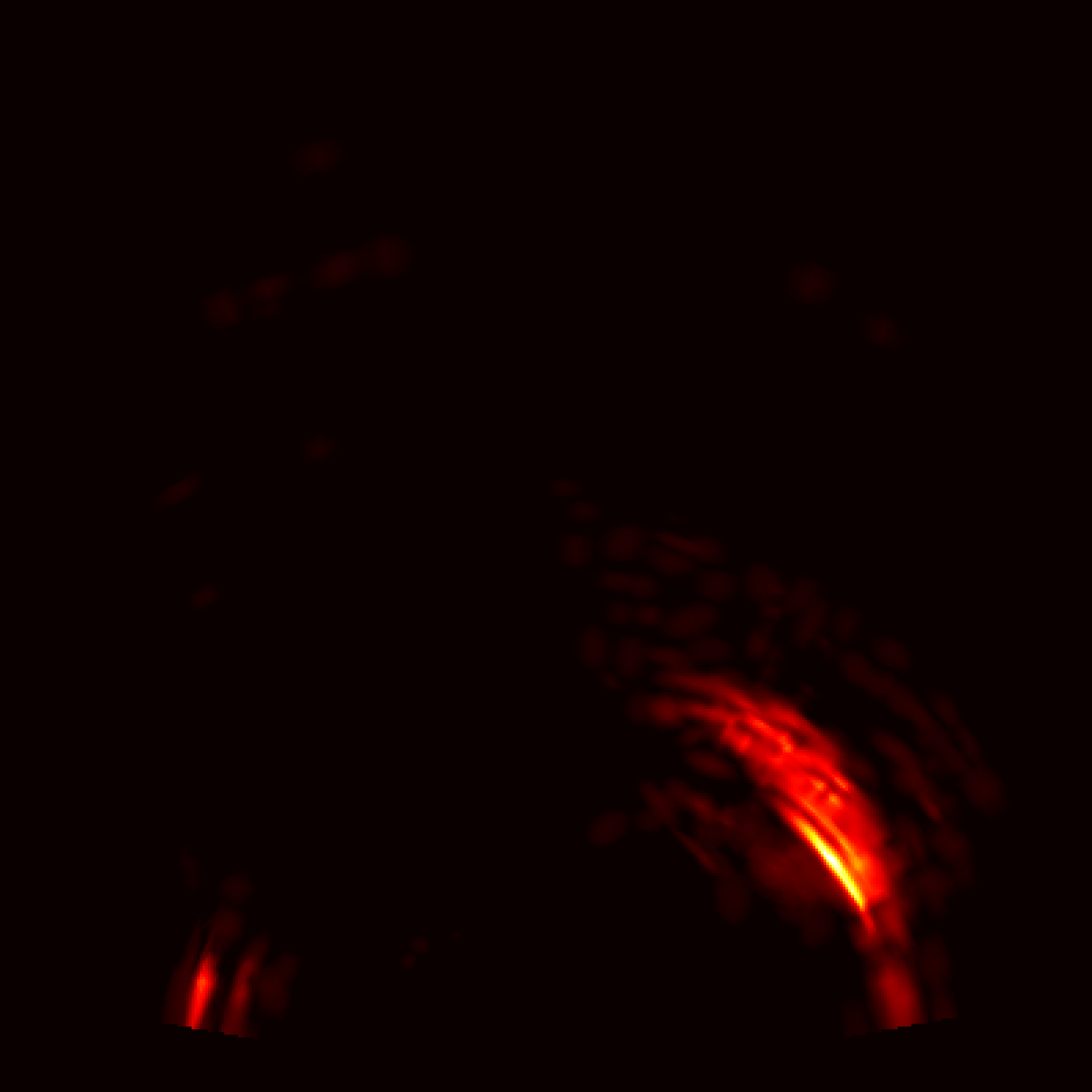} & \includegraphics[width=0.165\linewidth]{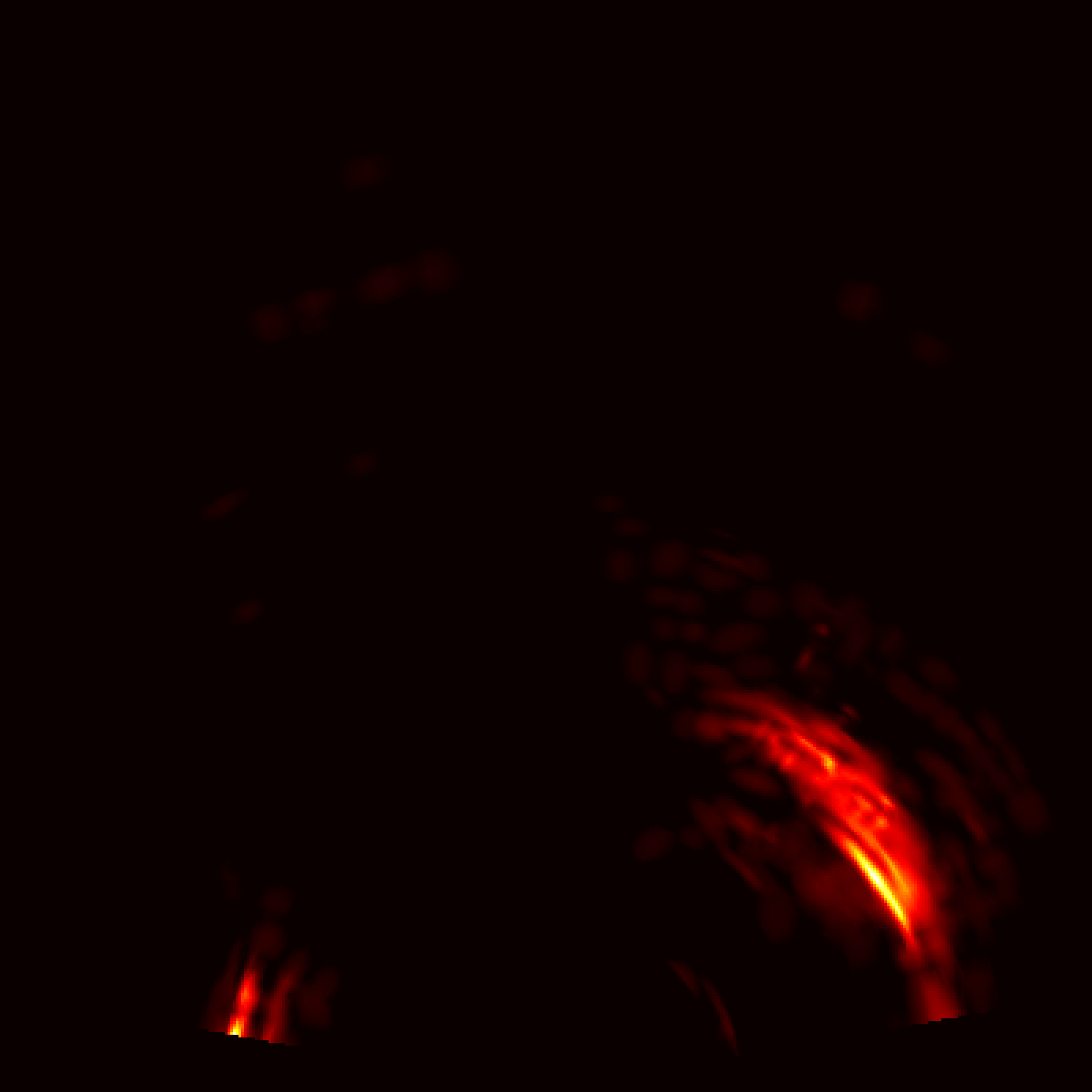} & \includegraphics[width=0.165\linewidth]{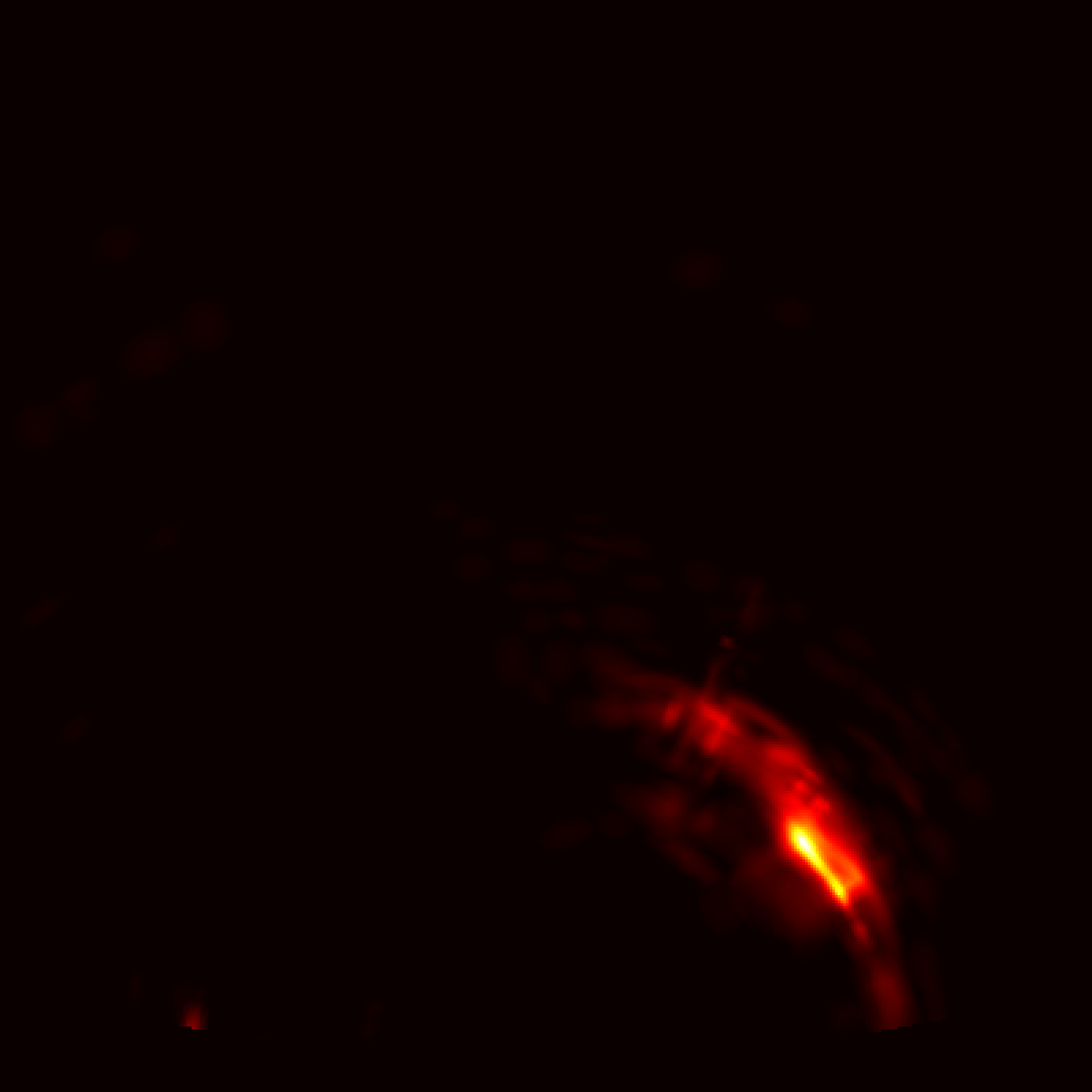} \\
\raisebox{2.0em}{\rotatebox{90}{\scriptsize\textbf{Radar Fields}}} & \includegraphics[width=0.165\linewidth]{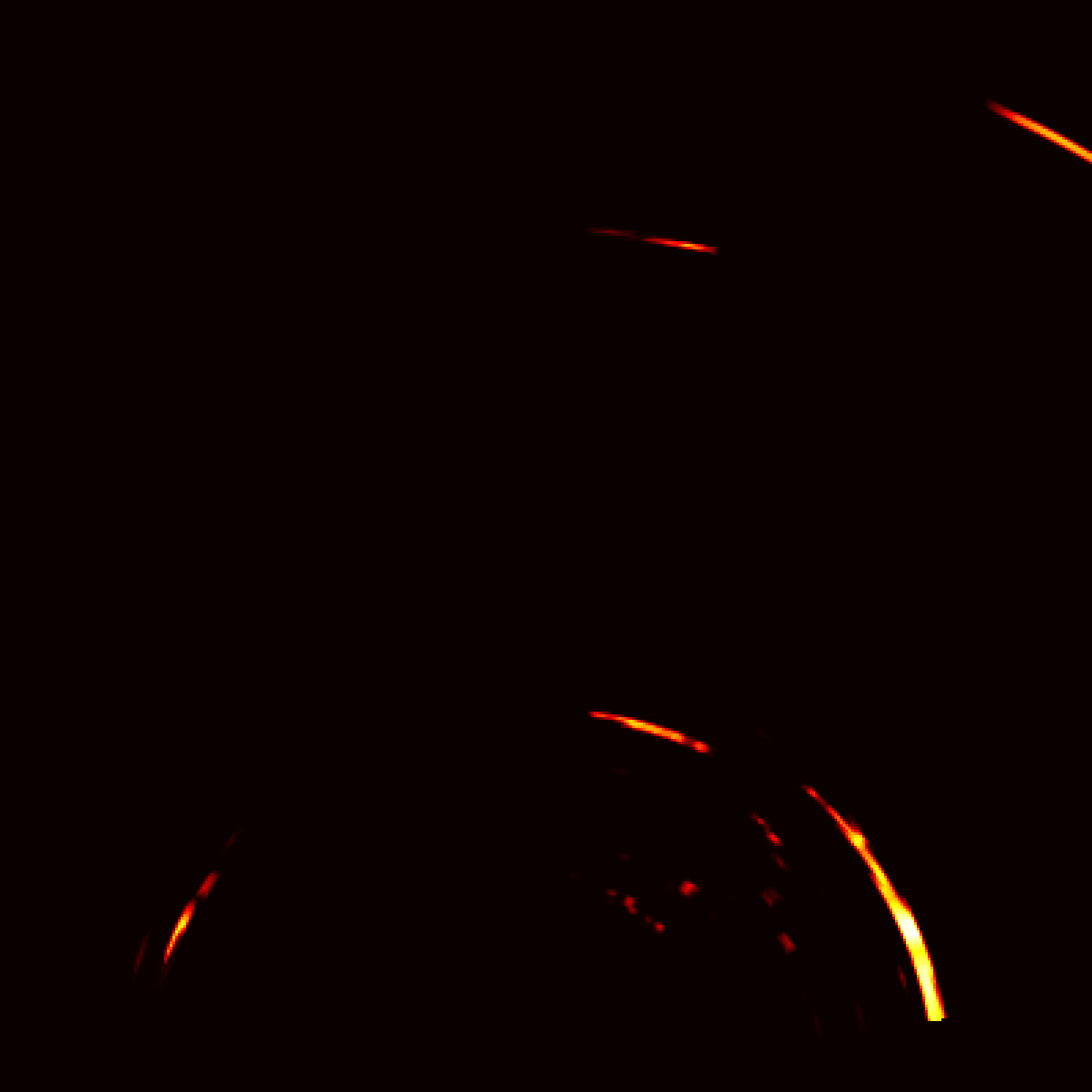} & \includegraphics[width=0.165\linewidth]{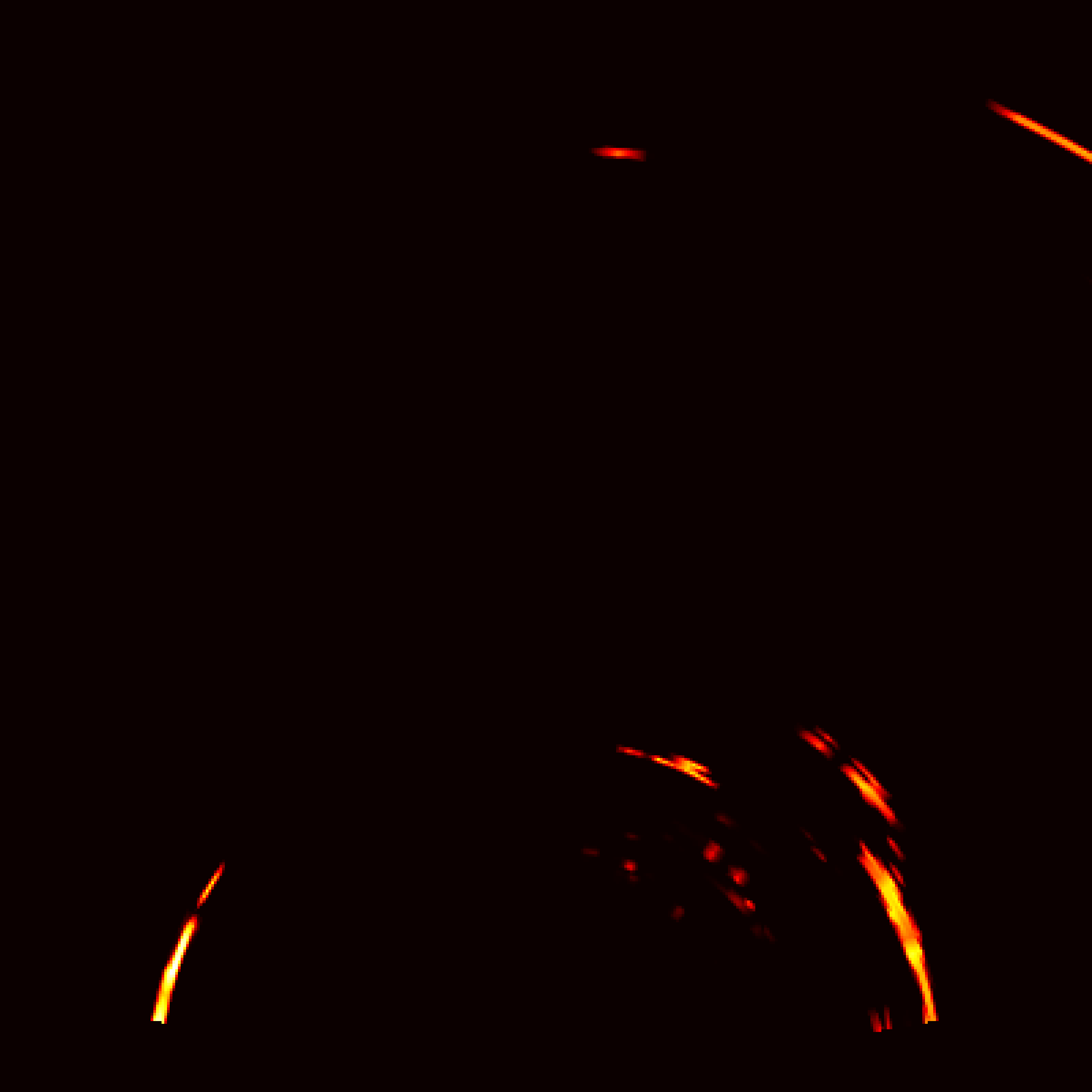} & \includegraphics[width=0.165\linewidth]{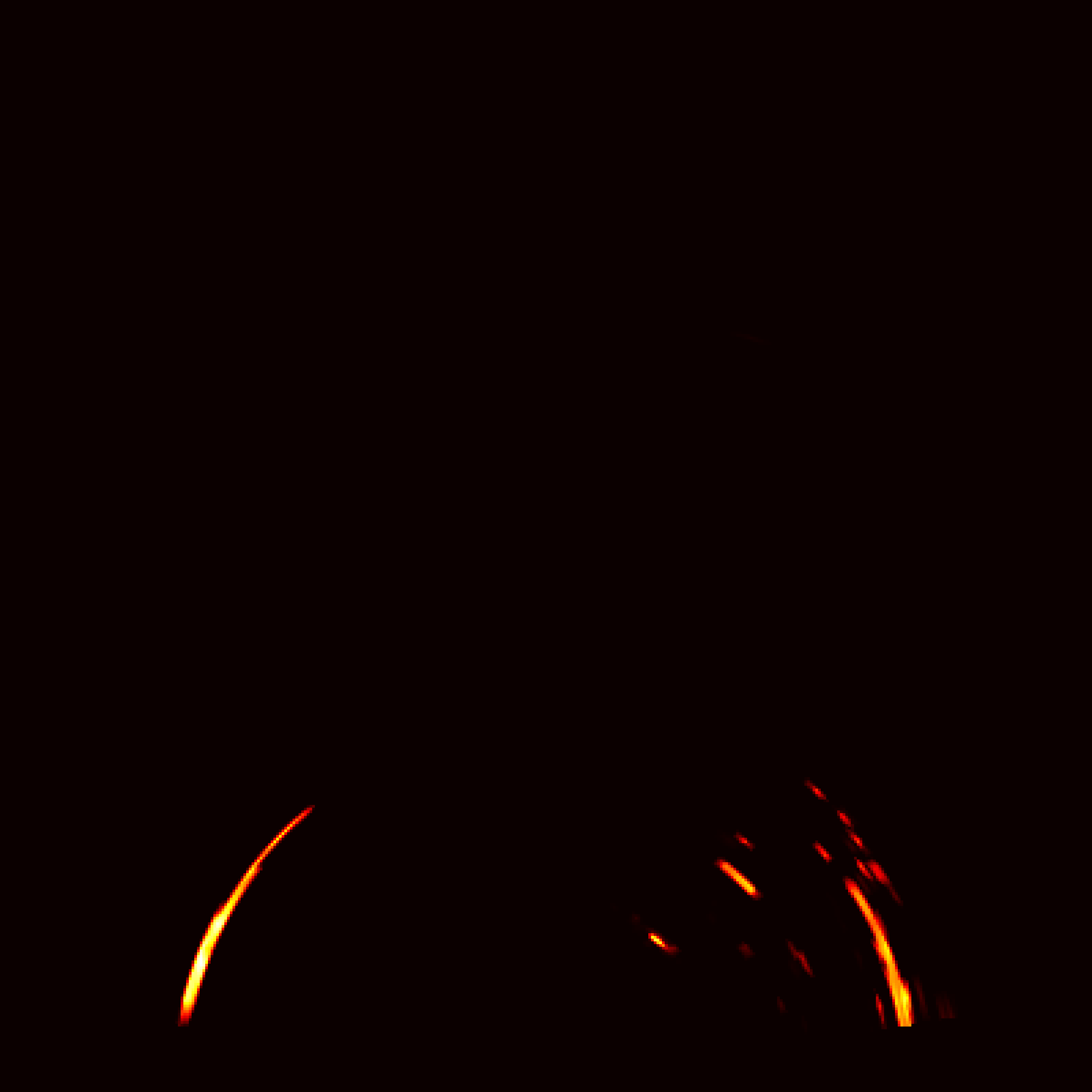} & \includegraphics[width=0.165\linewidth]{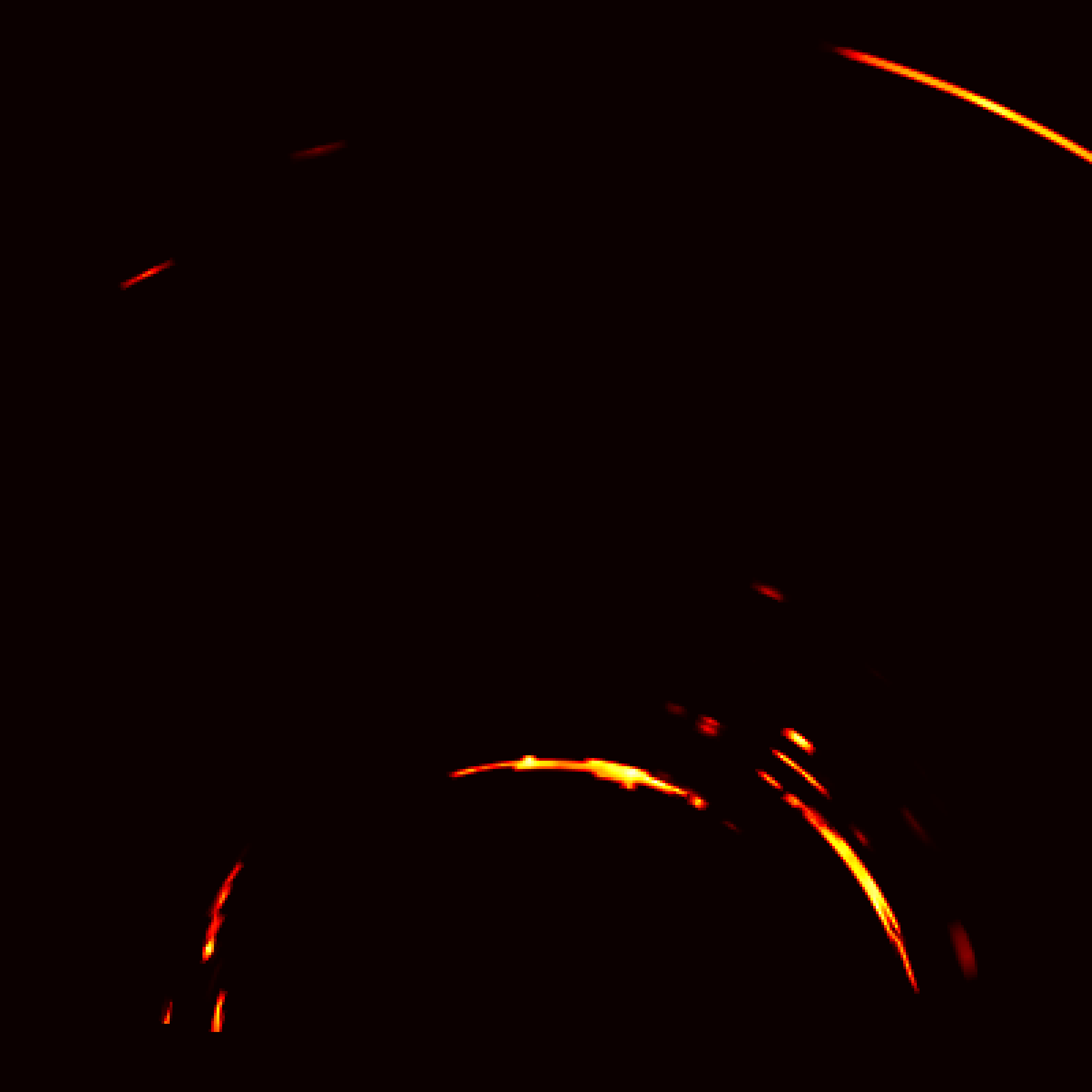} & \includegraphics[width=0.165\linewidth]{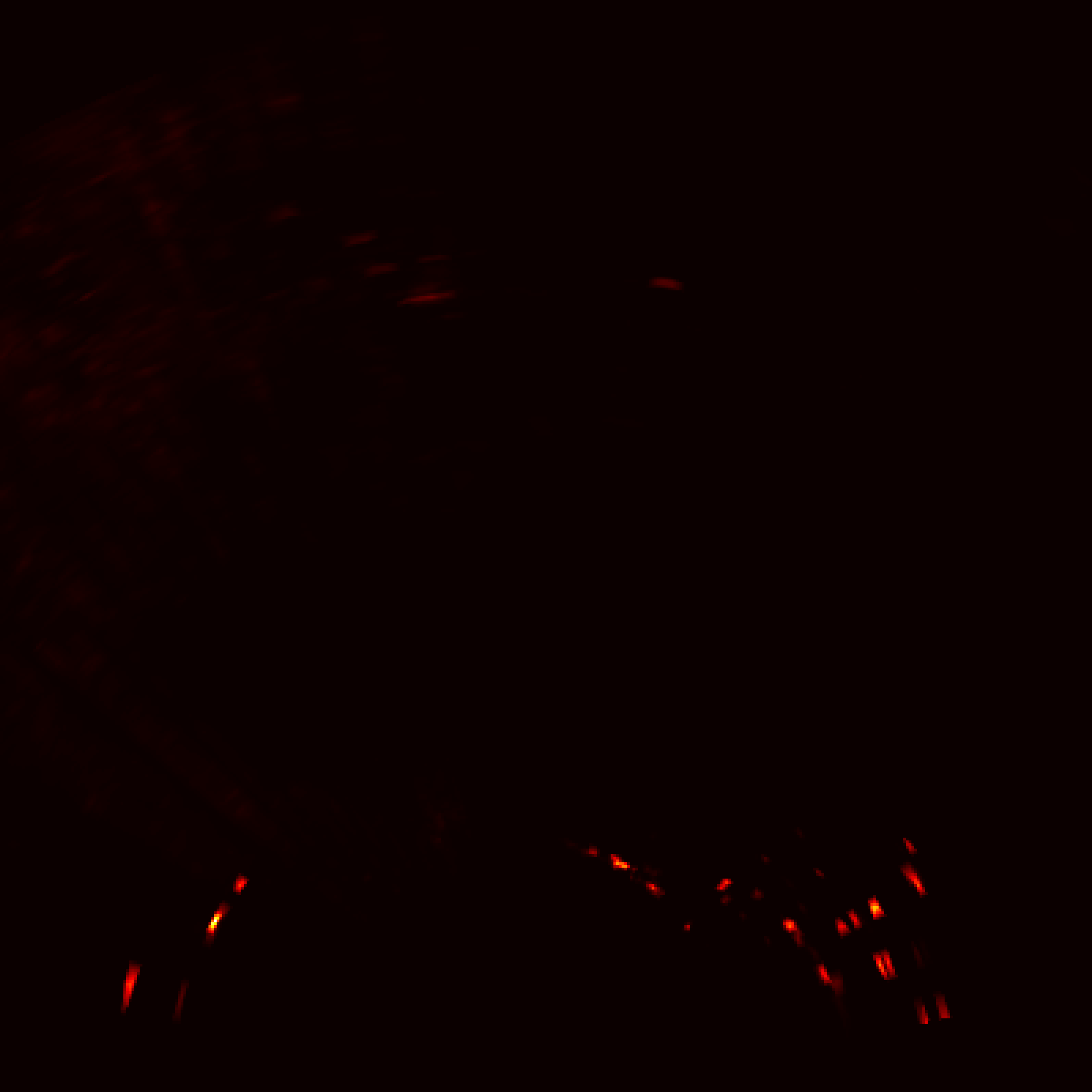} \\
\raisebox{2.0em}{\rotatebox{90}{\scriptsize\textbf{DART}}} & \includegraphics[width=0.165\linewidth]{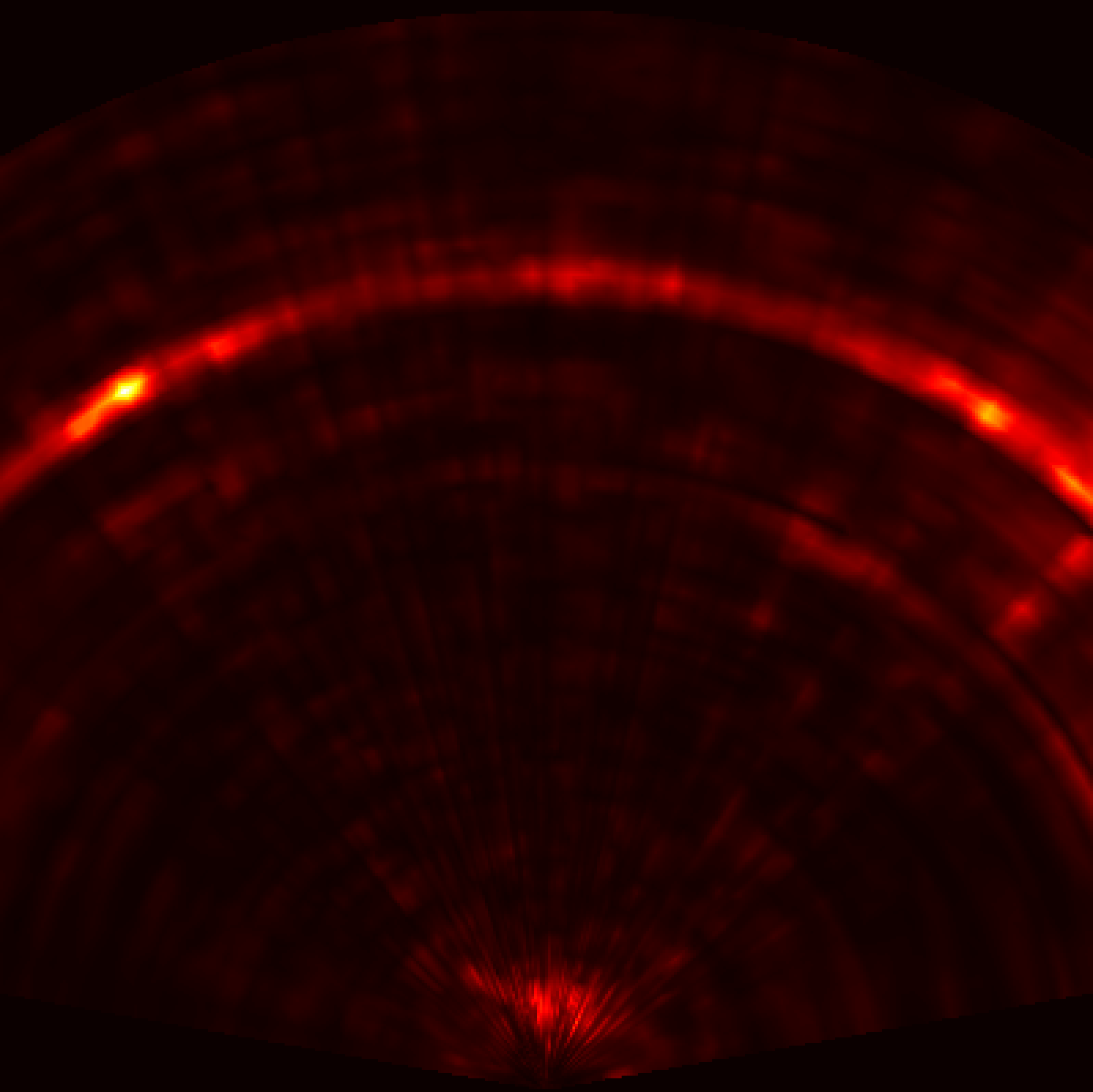} & \includegraphics[width=0.165\linewidth]{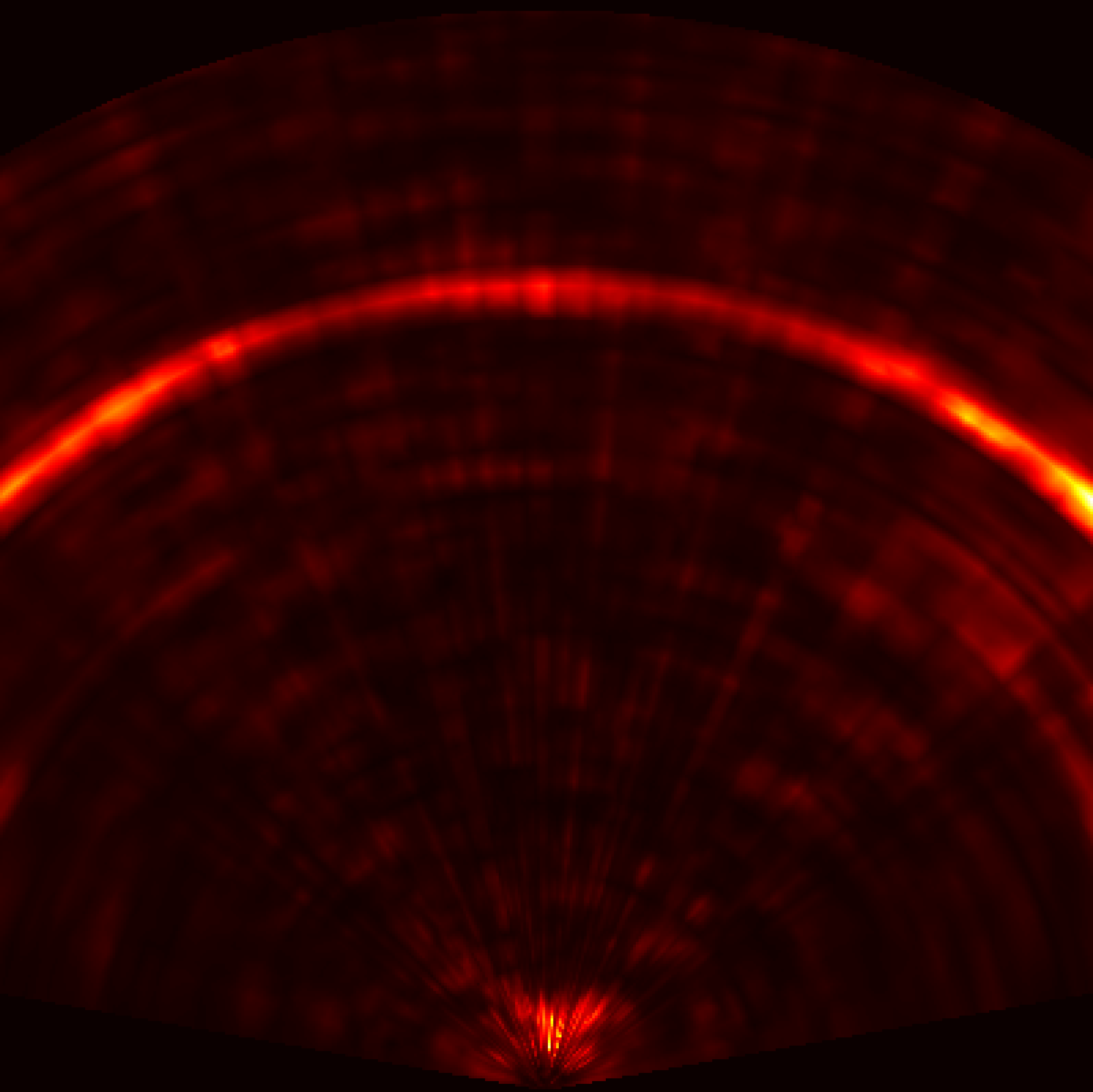} & \includegraphics[width=0.165\linewidth]{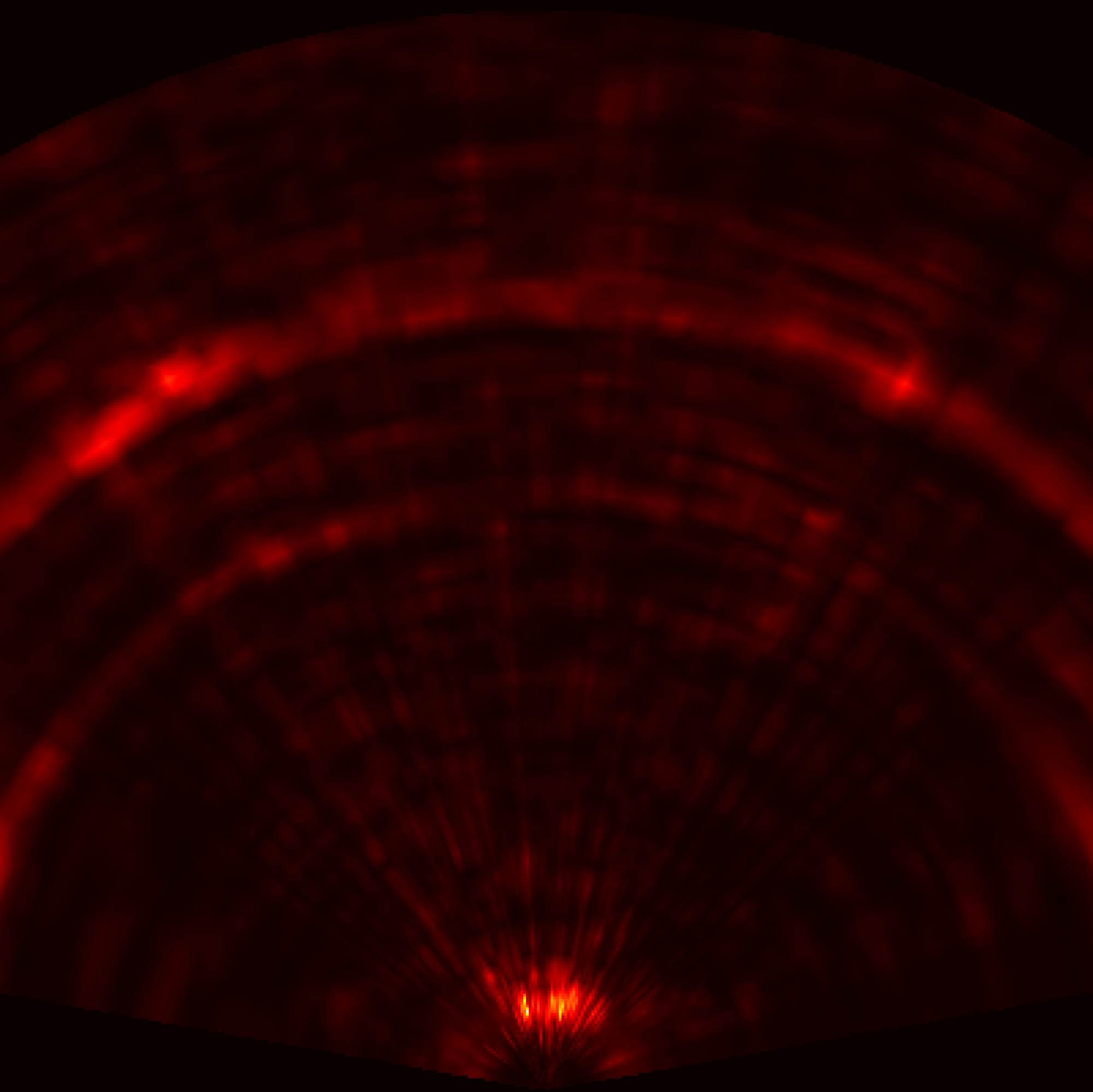} & \includegraphics[width=0.165\linewidth]{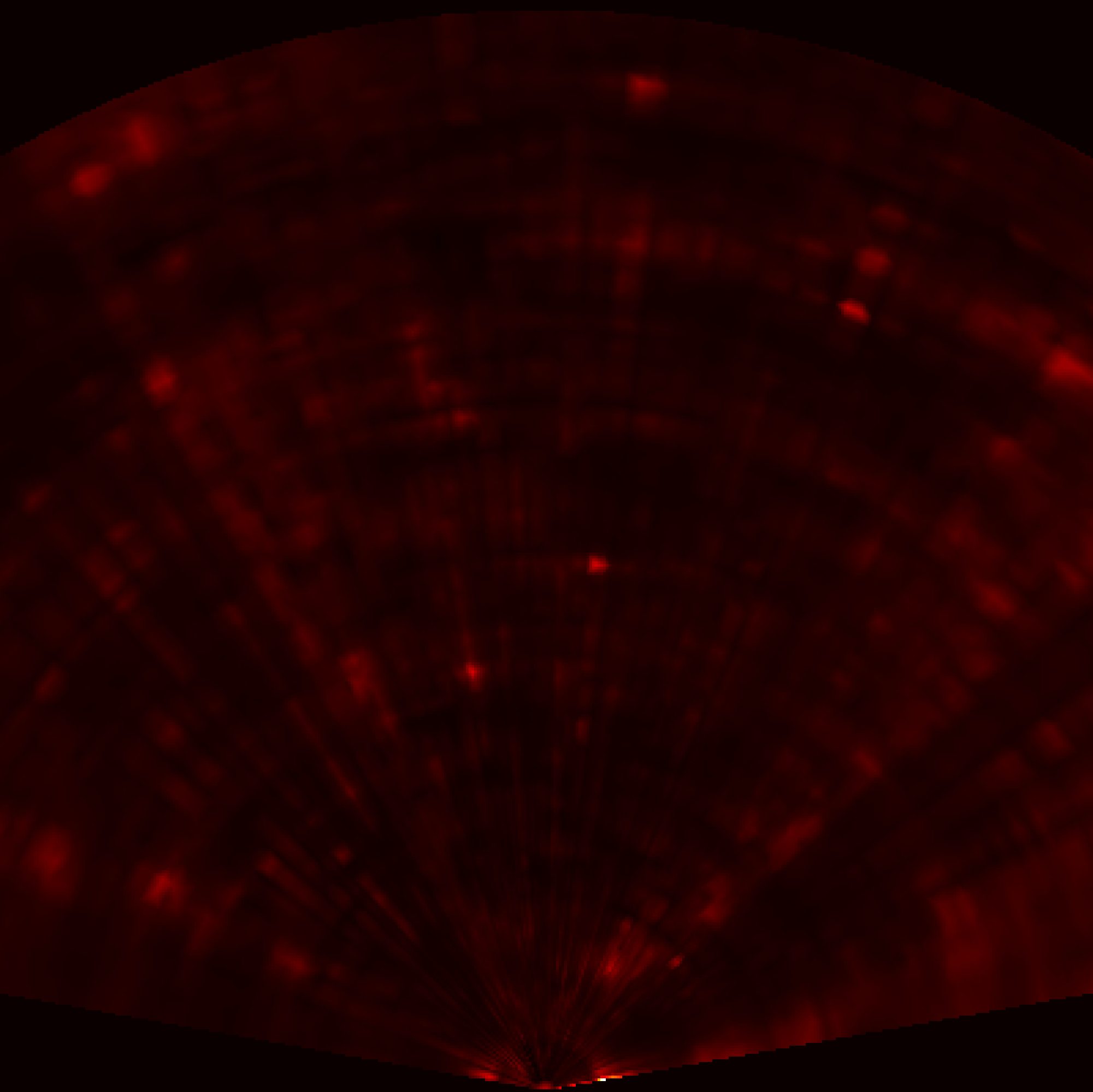} & \includegraphics[width=0.165\linewidth]{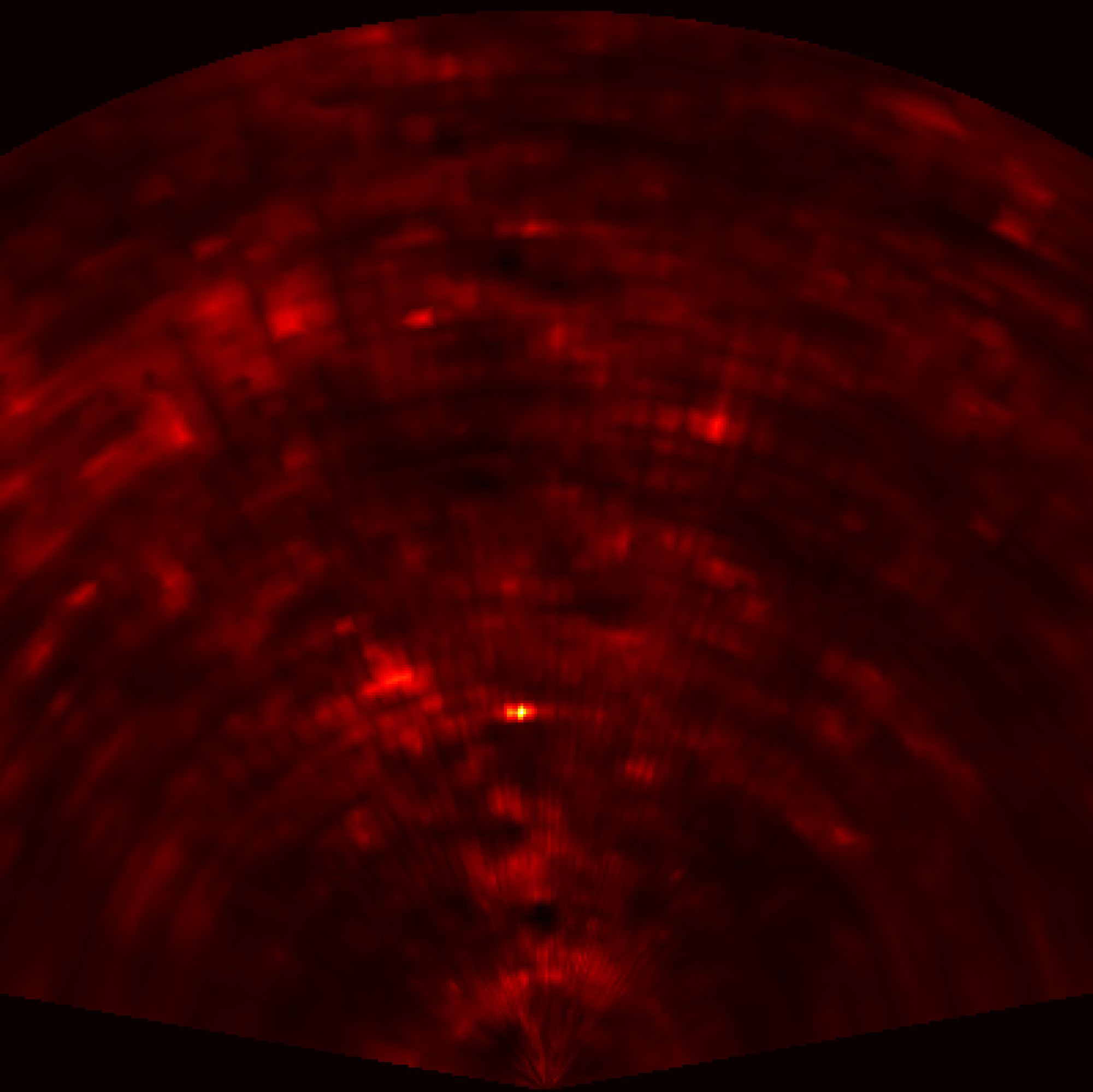} \\
\end{tabular}
\caption{\textbf{Per-frame $|\mathrm{RA}|$ comparison, scene S0\,F135, training frames $F\!+\!1$ to $F\!+\!4$ and held-out test frame $F$ (red header).} Pearson correlation per column, left to right. 3DPS: 0.84, 0.81, 0.77, 0.76, 0.64; mmIR: 0.65, 0.56, 0.54, 0.52, 0.27; RadarSplat: 0.46, 0.52, 0.43, 0.16, 0.41; Radar Fields: 0.13, 0.15, 0.22, 0.25, 0.11; DART: -0.05, -0.04, -0.03, 0.03, 0.03.}
\label{fig:supp_s0_f135_b}
\end{figure}

\begin{figure}[p]
\centering
\setlength{\tabcolsep}{1pt}\renewcommand{\arraystretch}{0.6}
\begin{tabular}{ccccc}
 & {\footnotesize $F\!-\!4$} & {\footnotesize $F\!-\!3$} & {\footnotesize $F\!-\!2$} & {\footnotesize $F\!-\!1$} \\
\raisebox{2.0em}{\rotatebox{90}{\scriptsize\textbf{GT}}} & \includegraphics[width=0.165\linewidth]{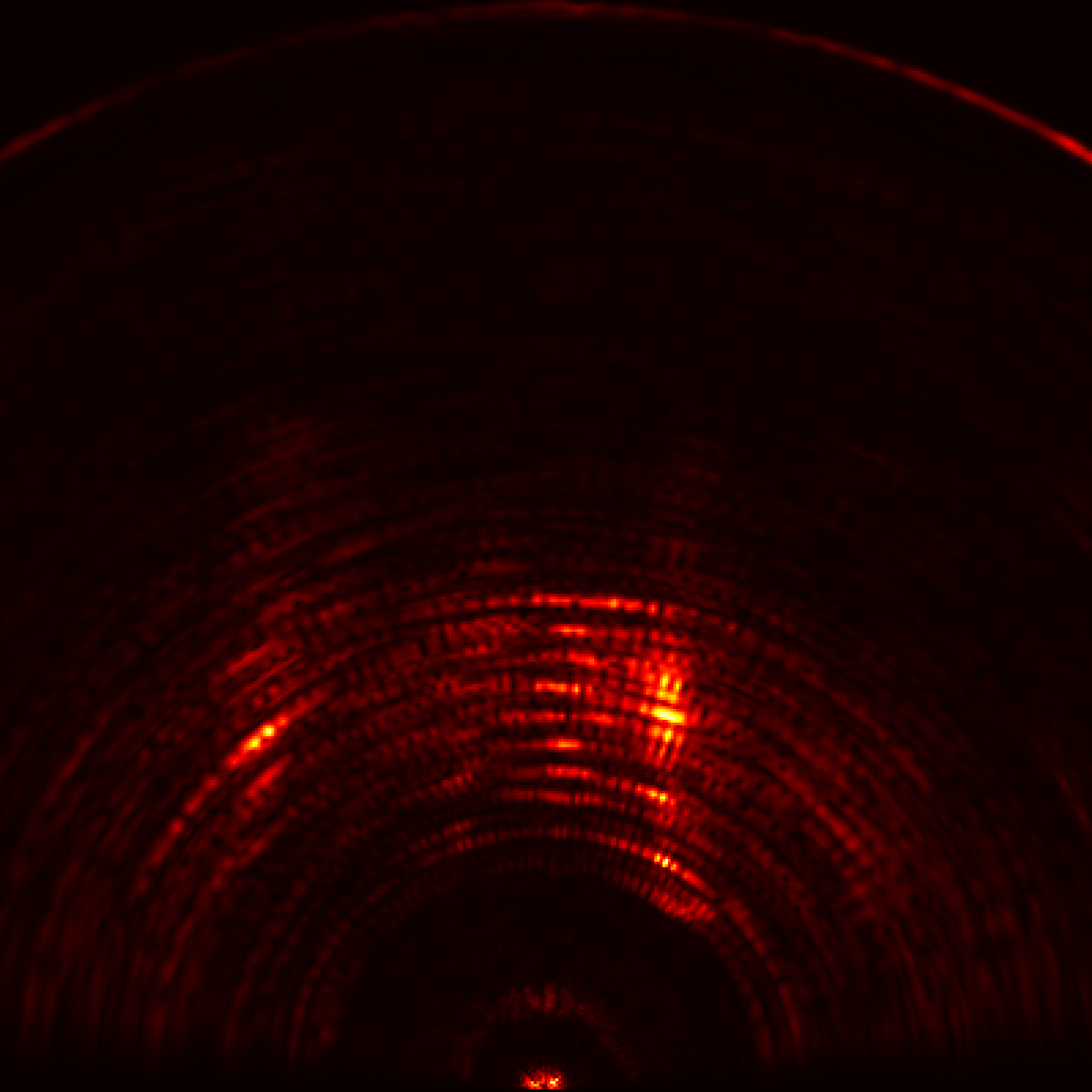} & \includegraphics[width=0.165\linewidth]{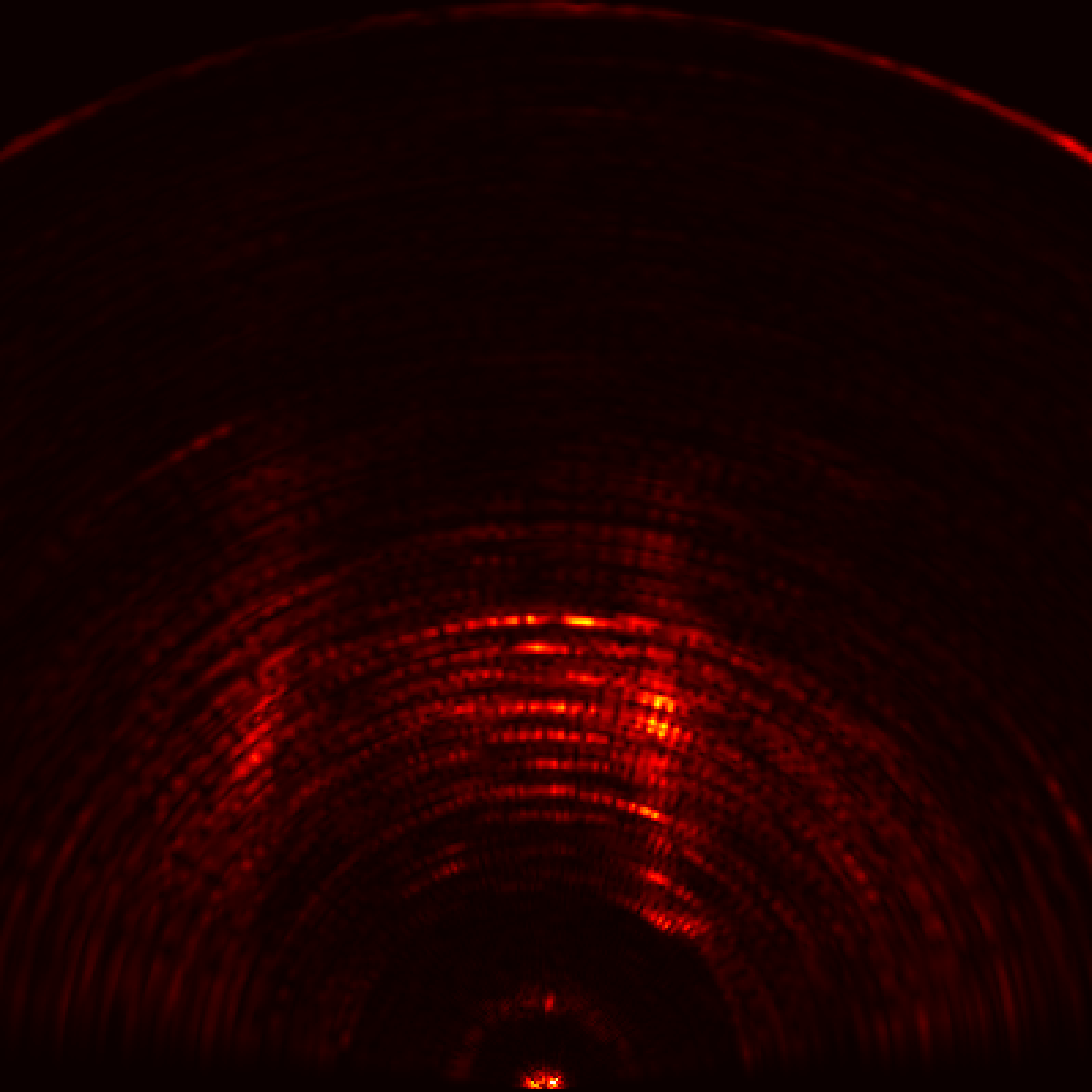} & \includegraphics[width=0.165\linewidth]{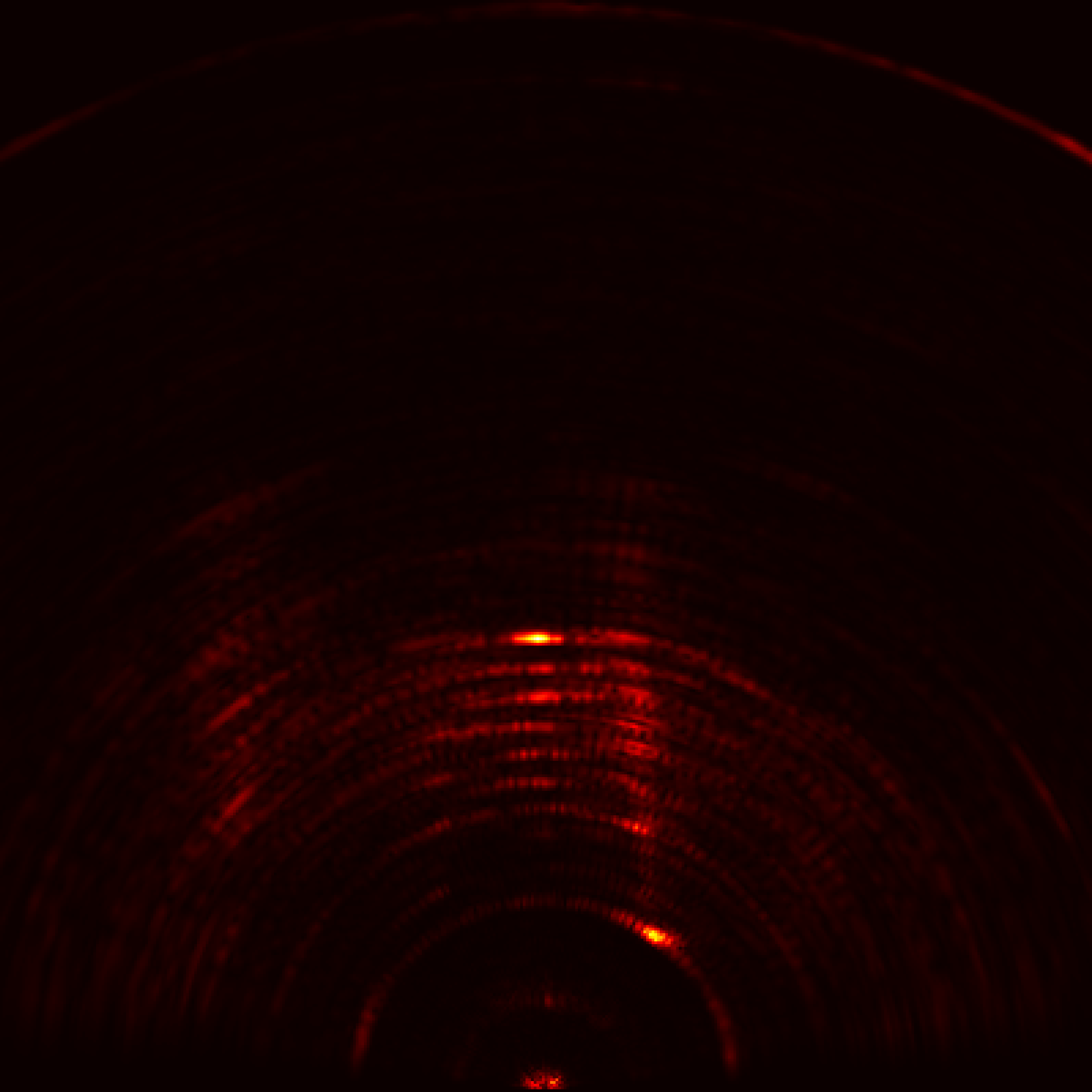} & \includegraphics[width=0.165\linewidth]{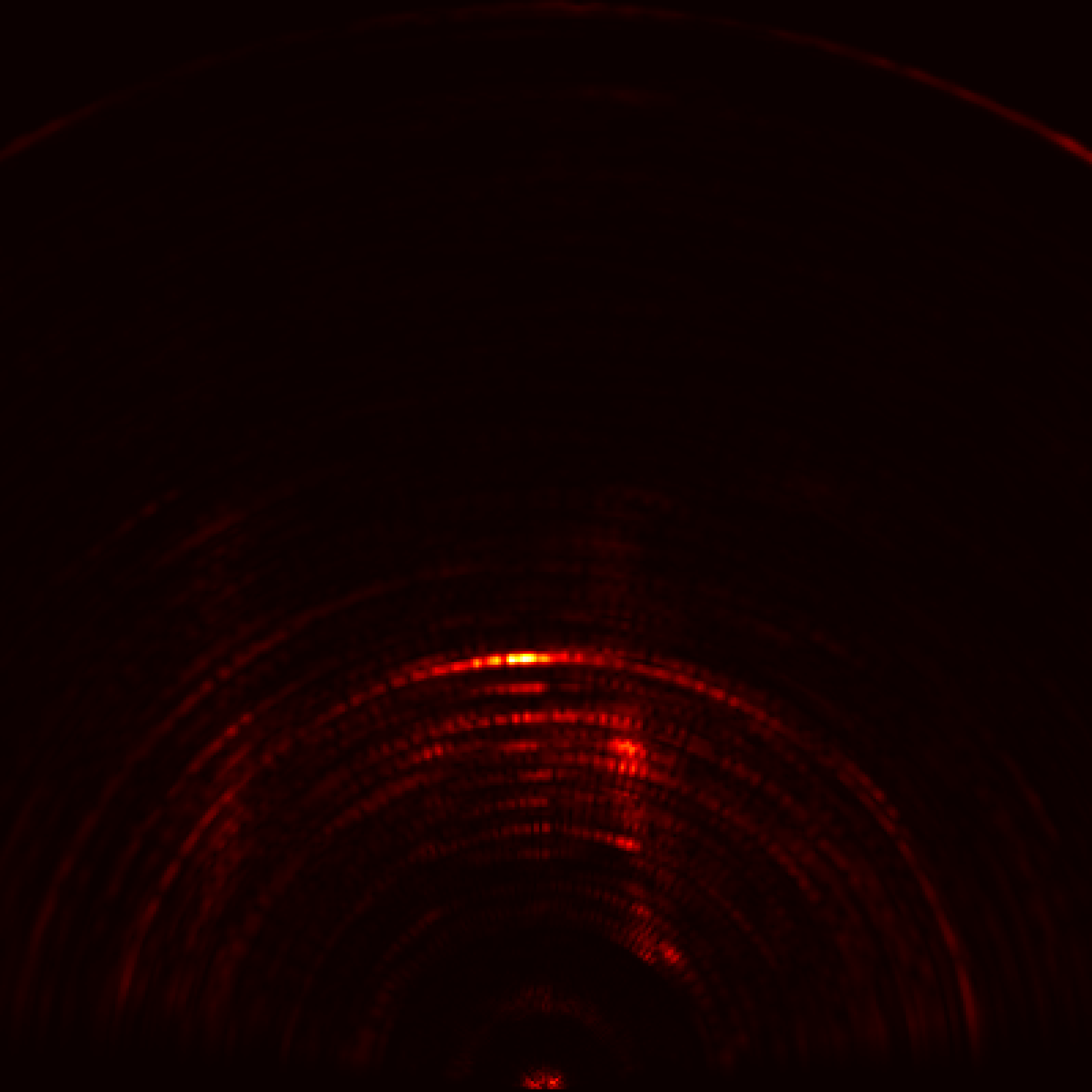} \\
\raisebox{2.0em}{\rotatebox{90}{\scriptsize\textbf{3DPS (Ours)}}} & \includegraphics[width=0.165\linewidth]{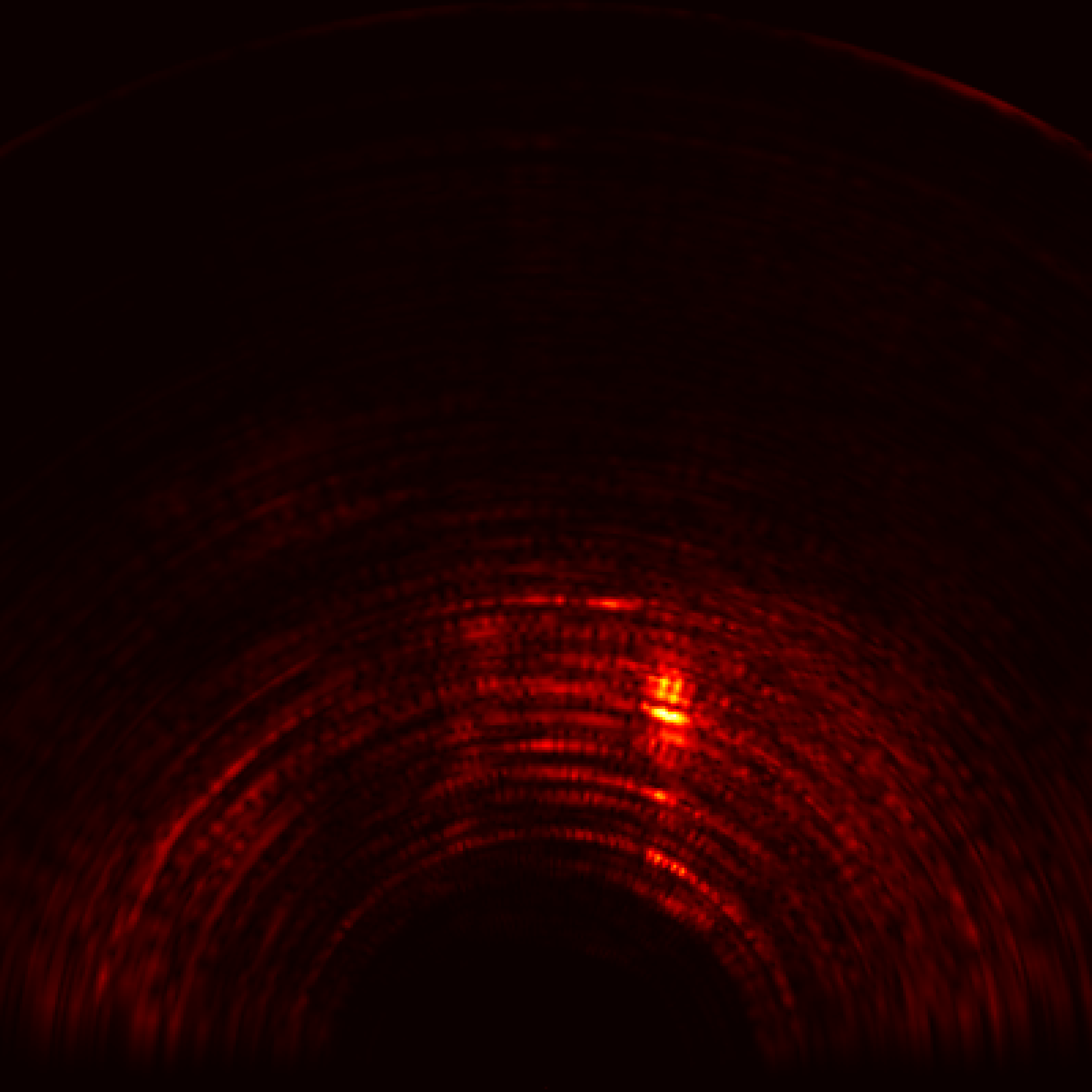} & \includegraphics[width=0.165\linewidth]{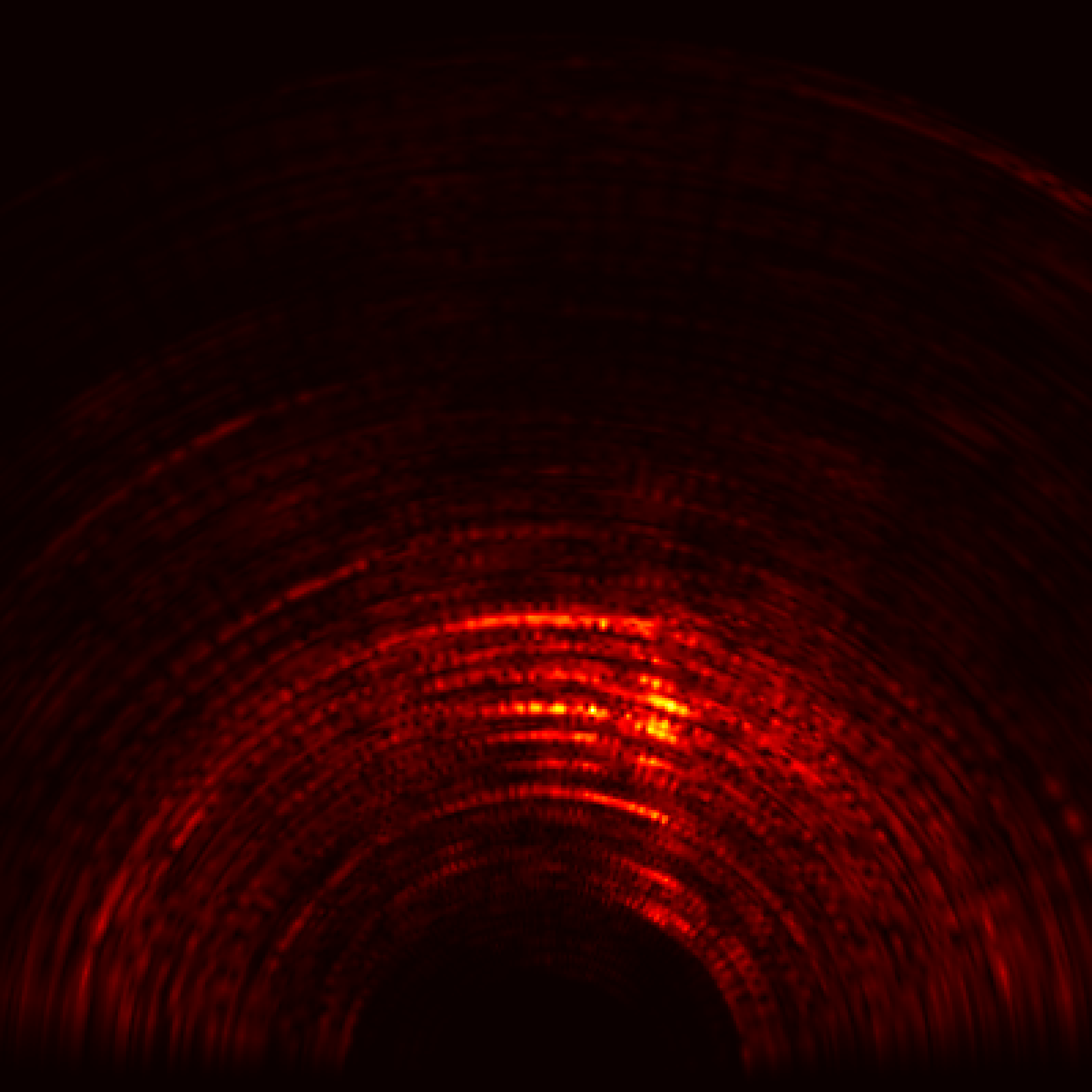} & \includegraphics[width=0.165\linewidth]{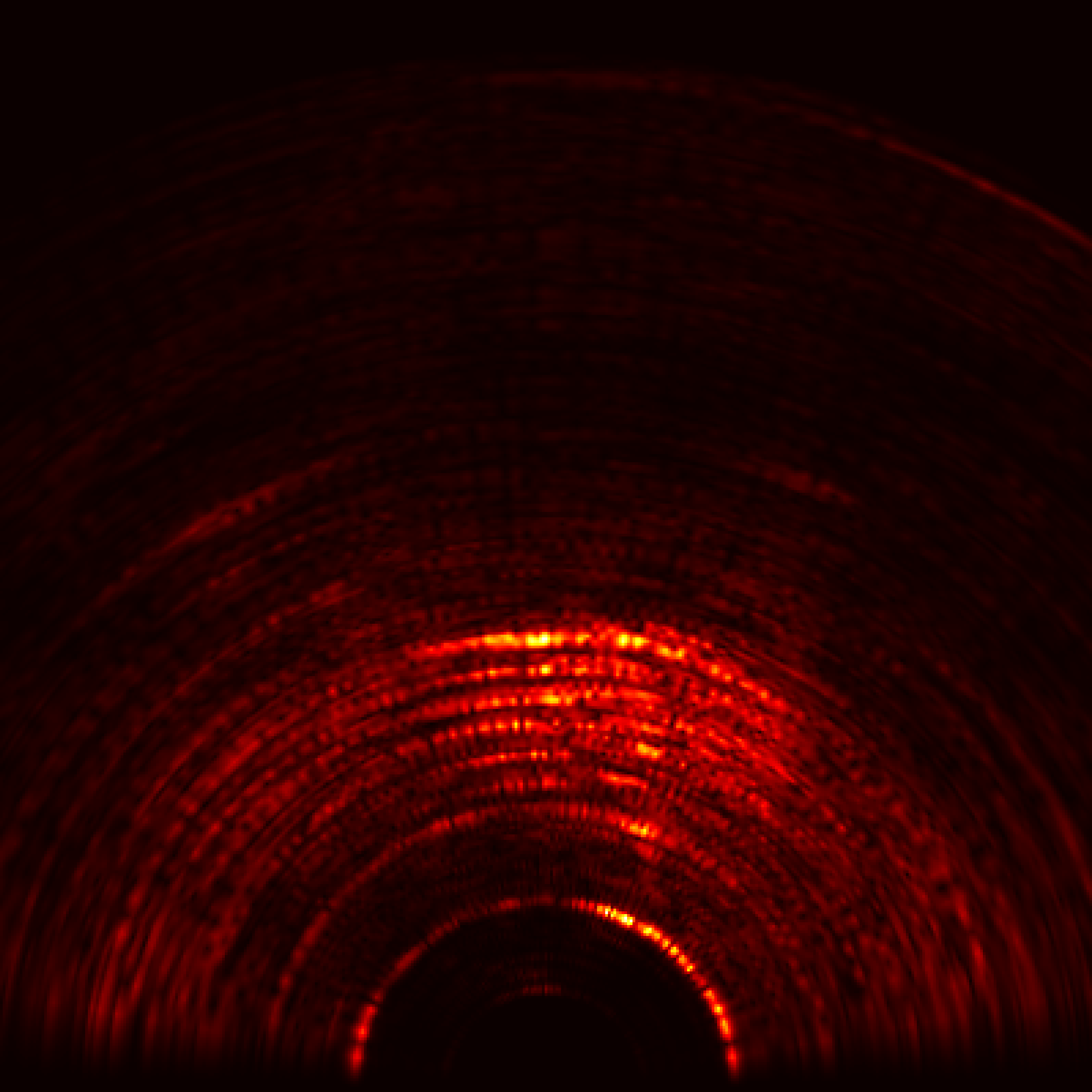} & \includegraphics[width=0.165\linewidth]{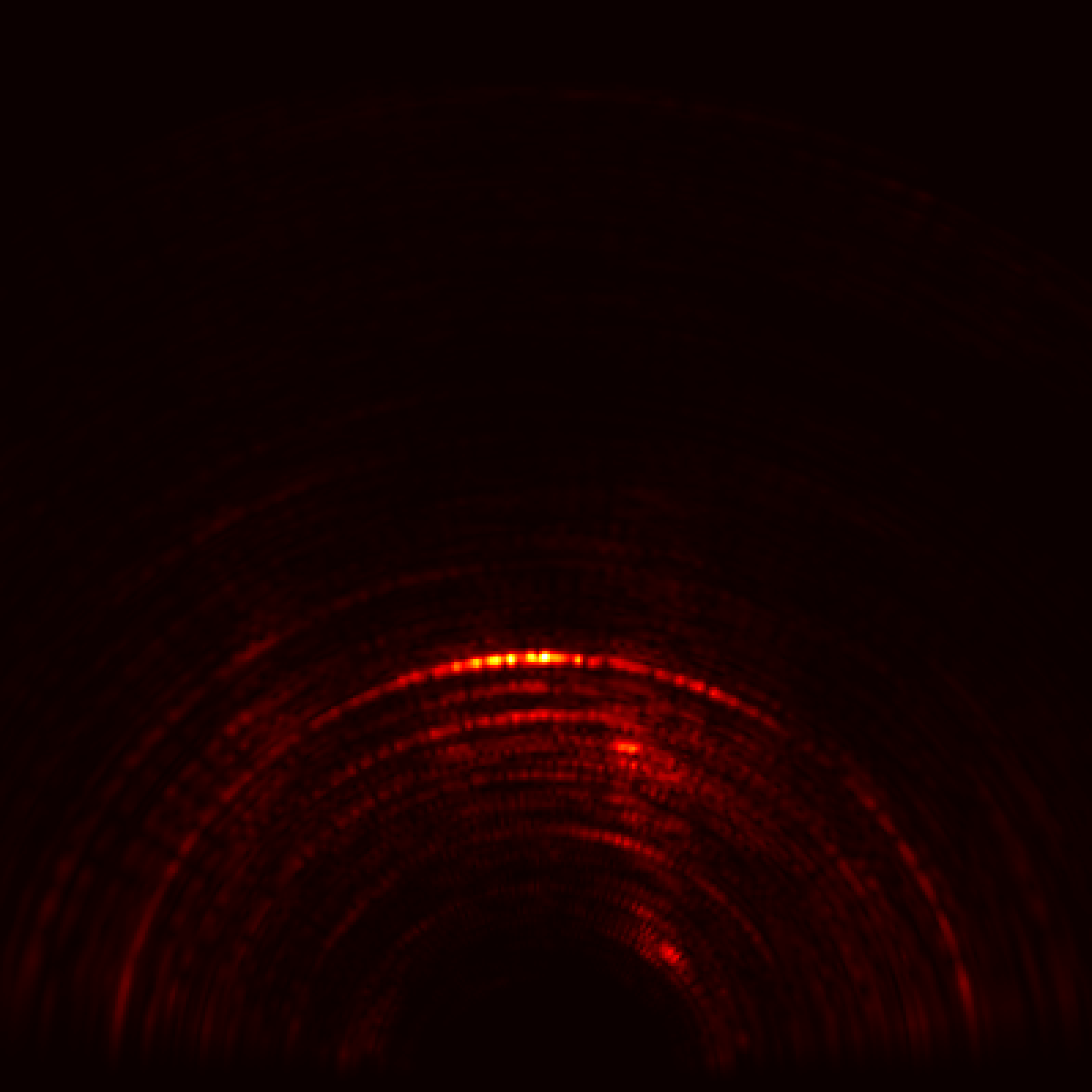} \\
\raisebox{2.0em}{\rotatebox{90}{\scriptsize\textbf{mmIR}}} & \includegraphics[width=0.165\linewidth]{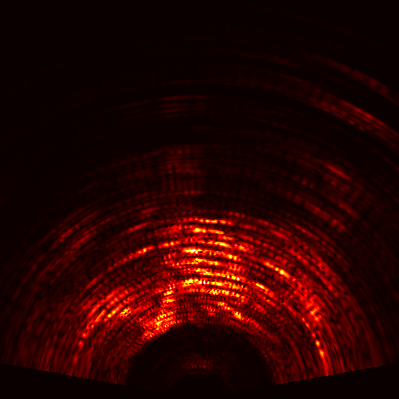} & \includegraphics[width=0.165\linewidth]{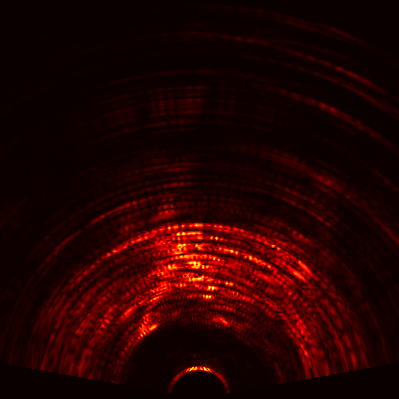} & \includegraphics[width=0.165\linewidth]{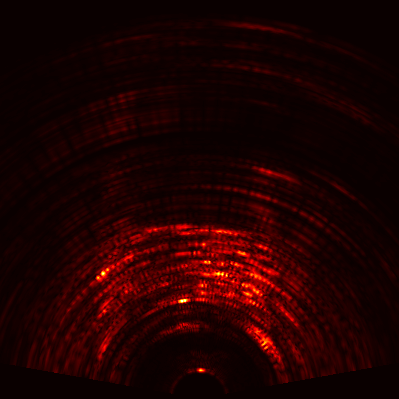} & \includegraphics[width=0.165\linewidth]{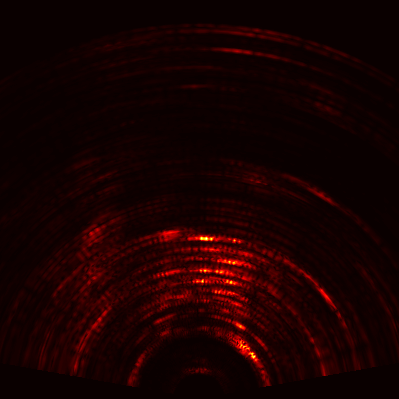} \\
\raisebox{2.0em}{\rotatebox{90}{\scriptsize\textbf{RadarSplat}}} & \includegraphics[width=0.165\linewidth]{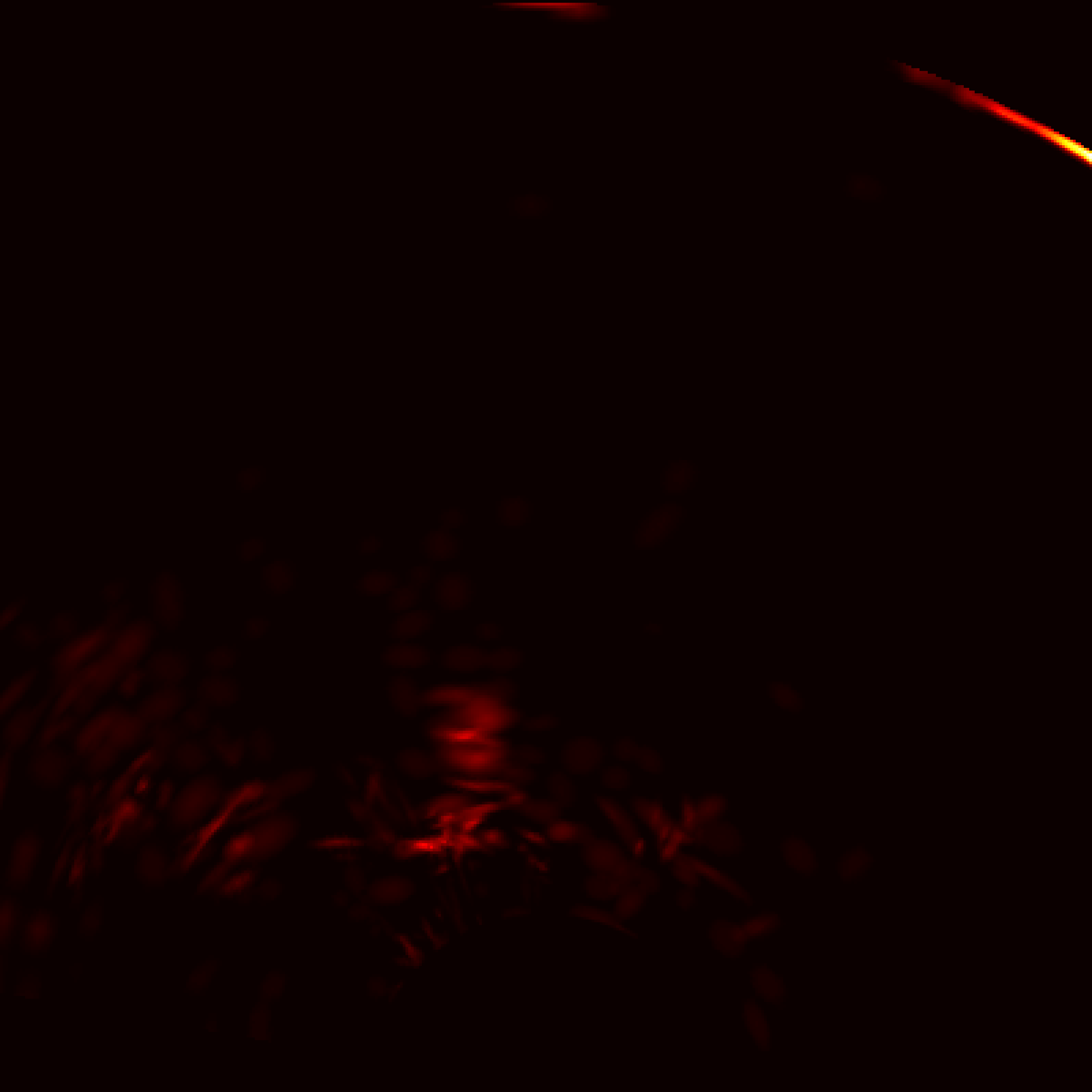} & \includegraphics[width=0.165\linewidth]{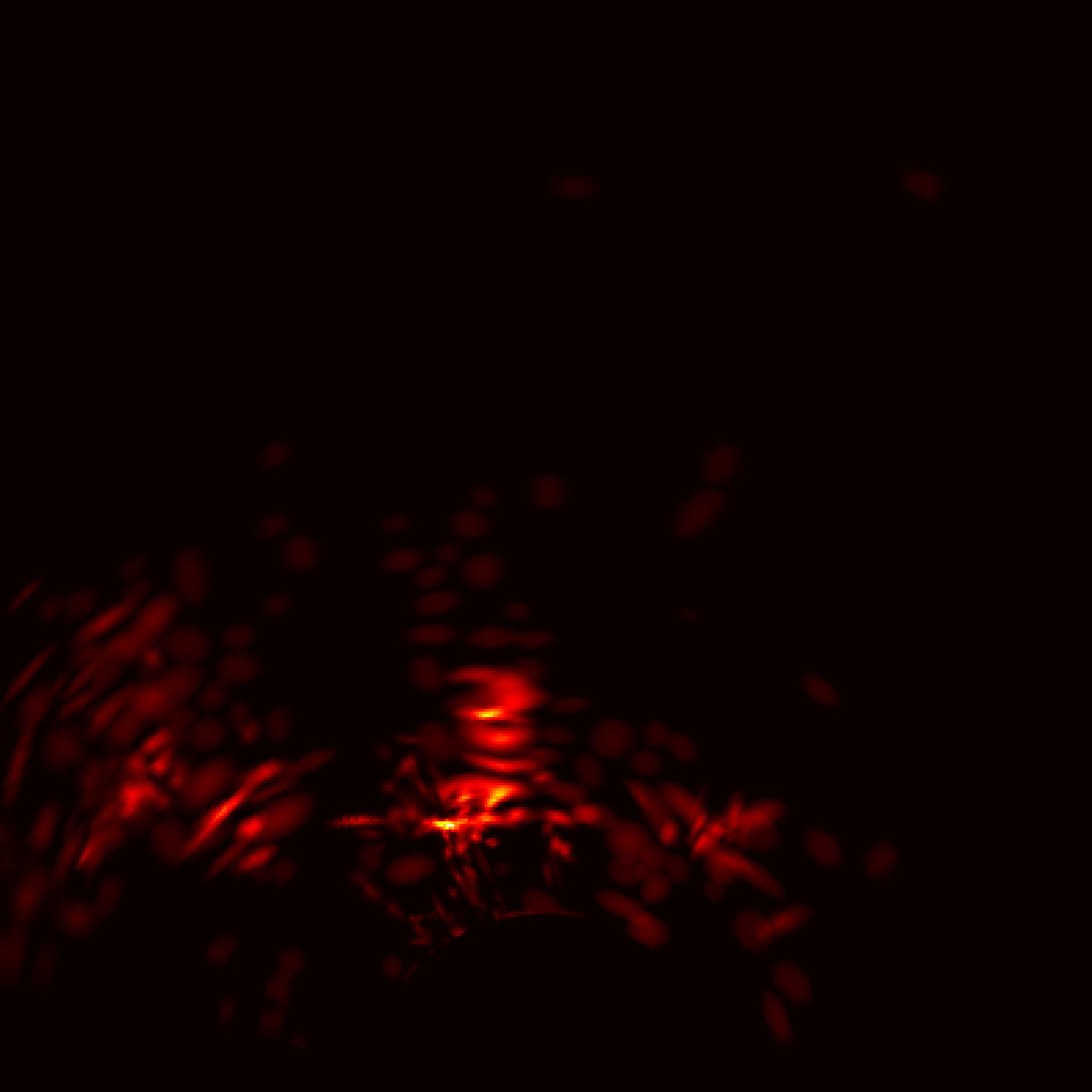} & \includegraphics[width=0.165\linewidth]{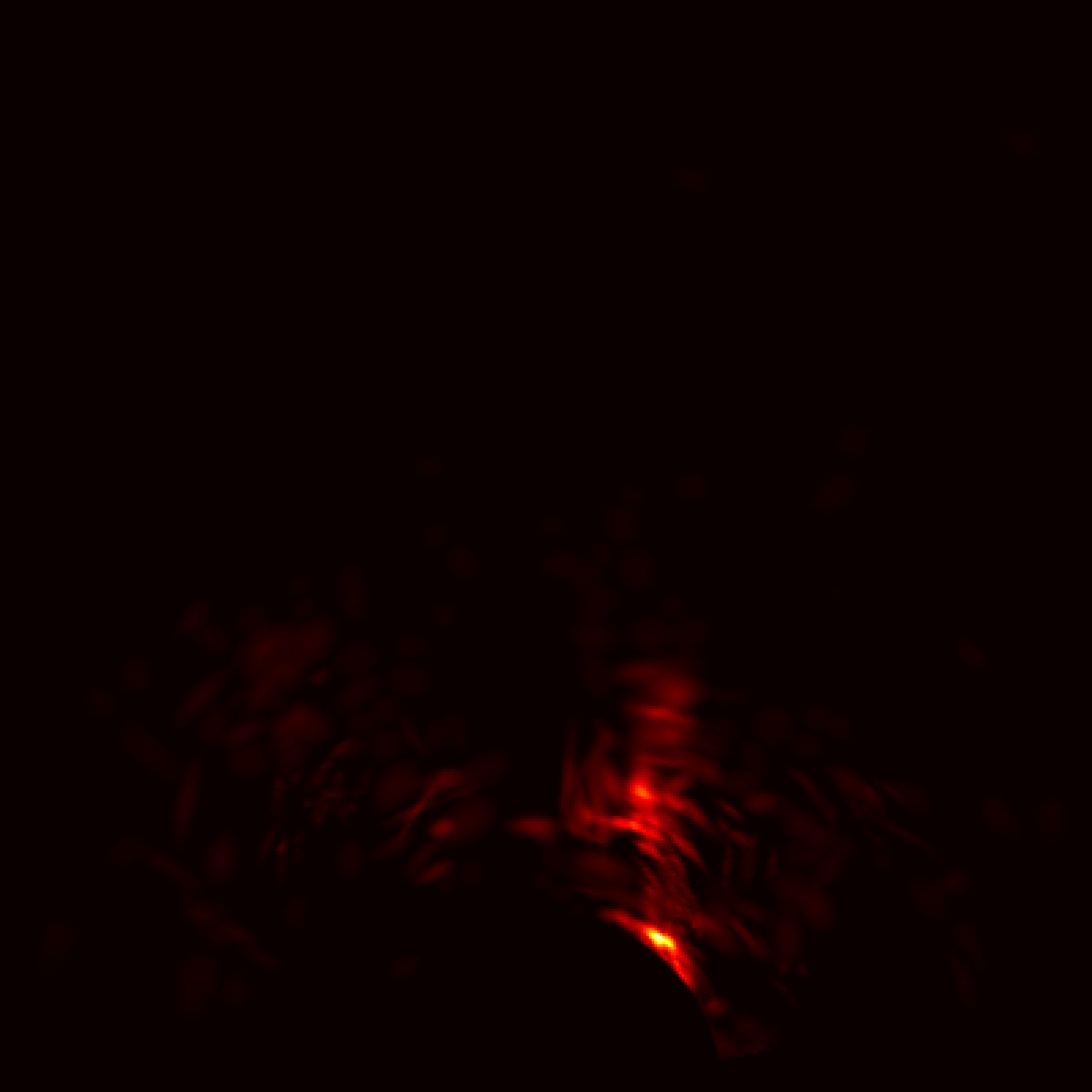} & \includegraphics[width=0.165\linewidth]{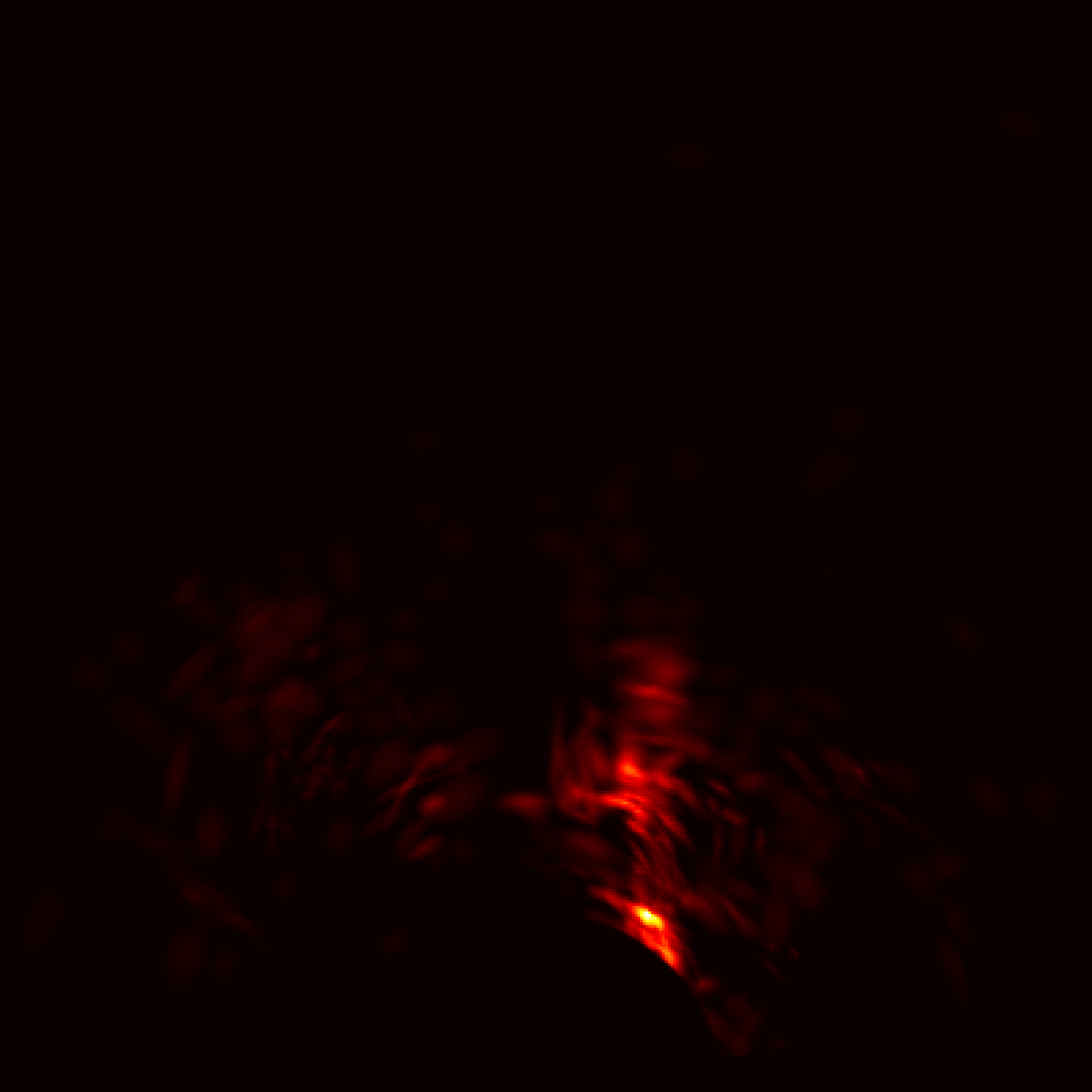} \\
\raisebox{2.0em}{\rotatebox{90}{\scriptsize\textbf{Radar Fields}}} & \includegraphics[width=0.165\linewidth]{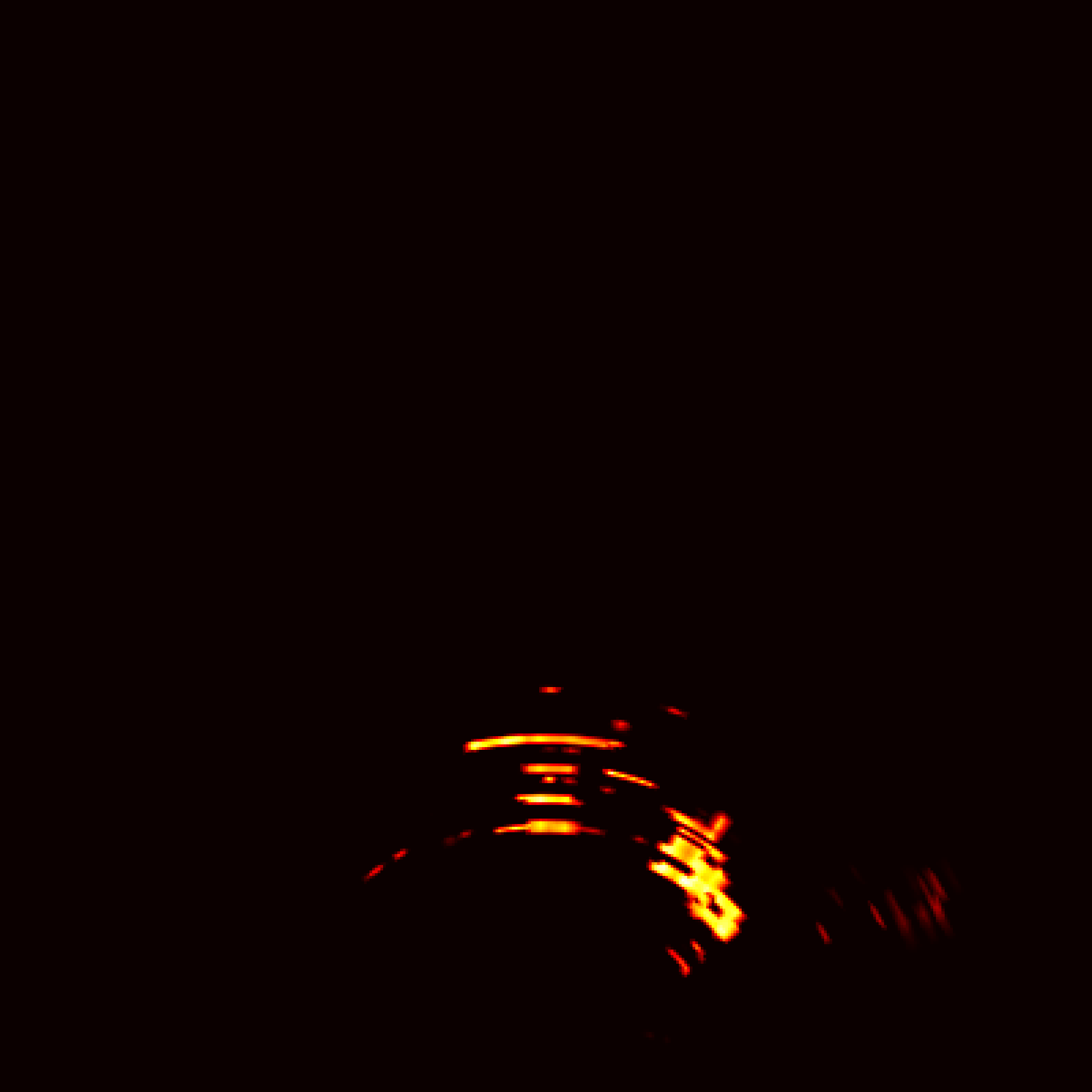} & \includegraphics[width=0.165\linewidth]{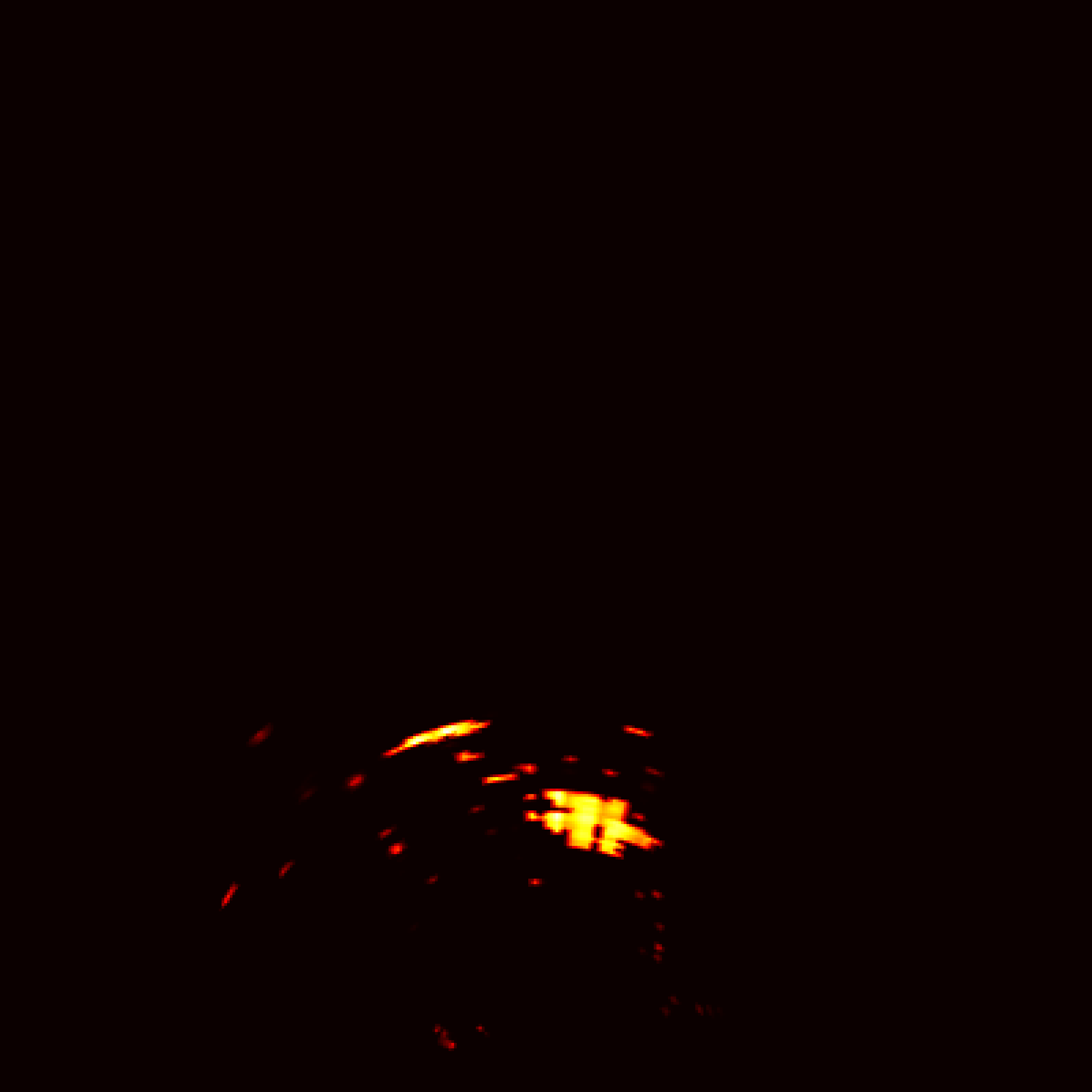} & \includegraphics[width=0.165\linewidth]{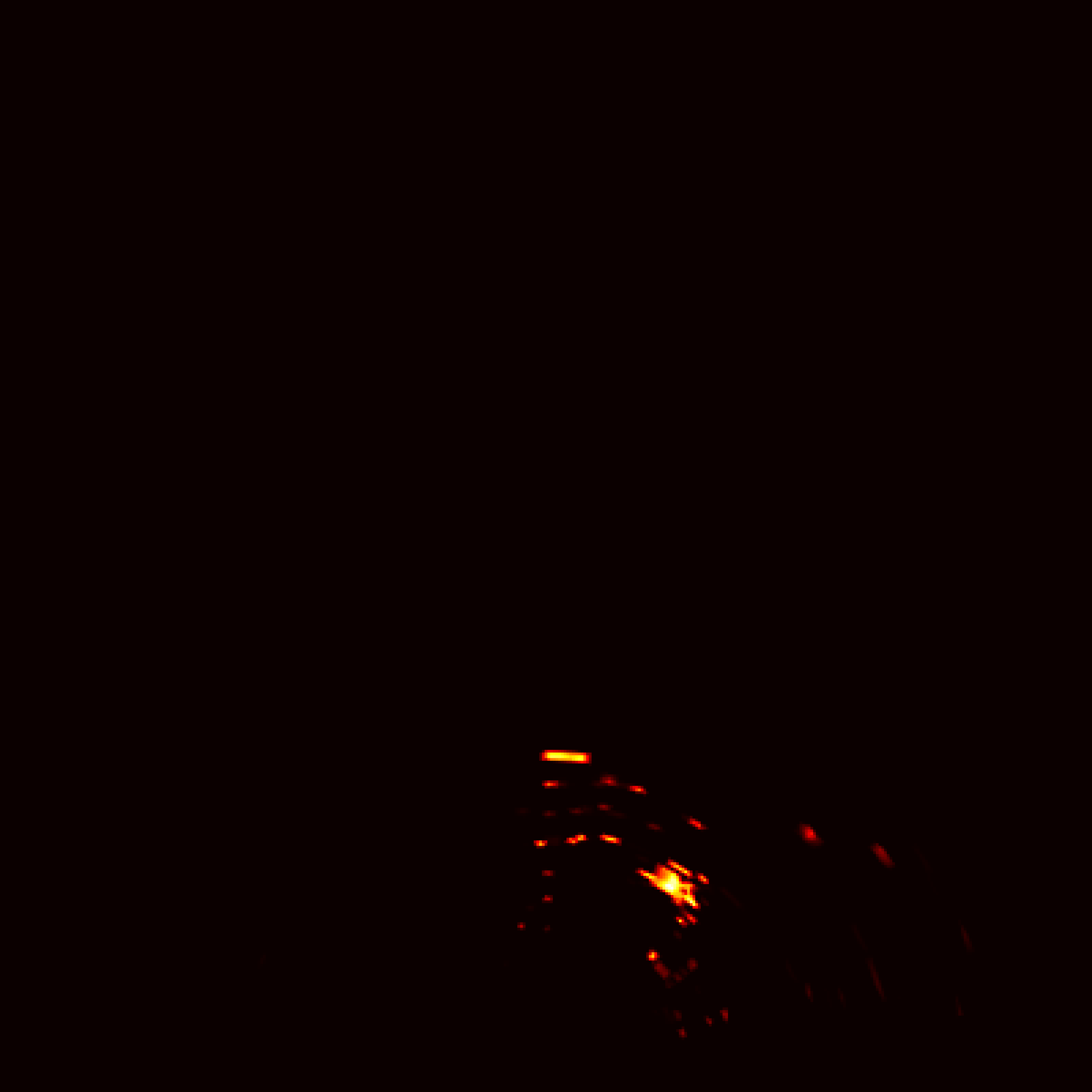} & \includegraphics[width=0.165\linewidth]{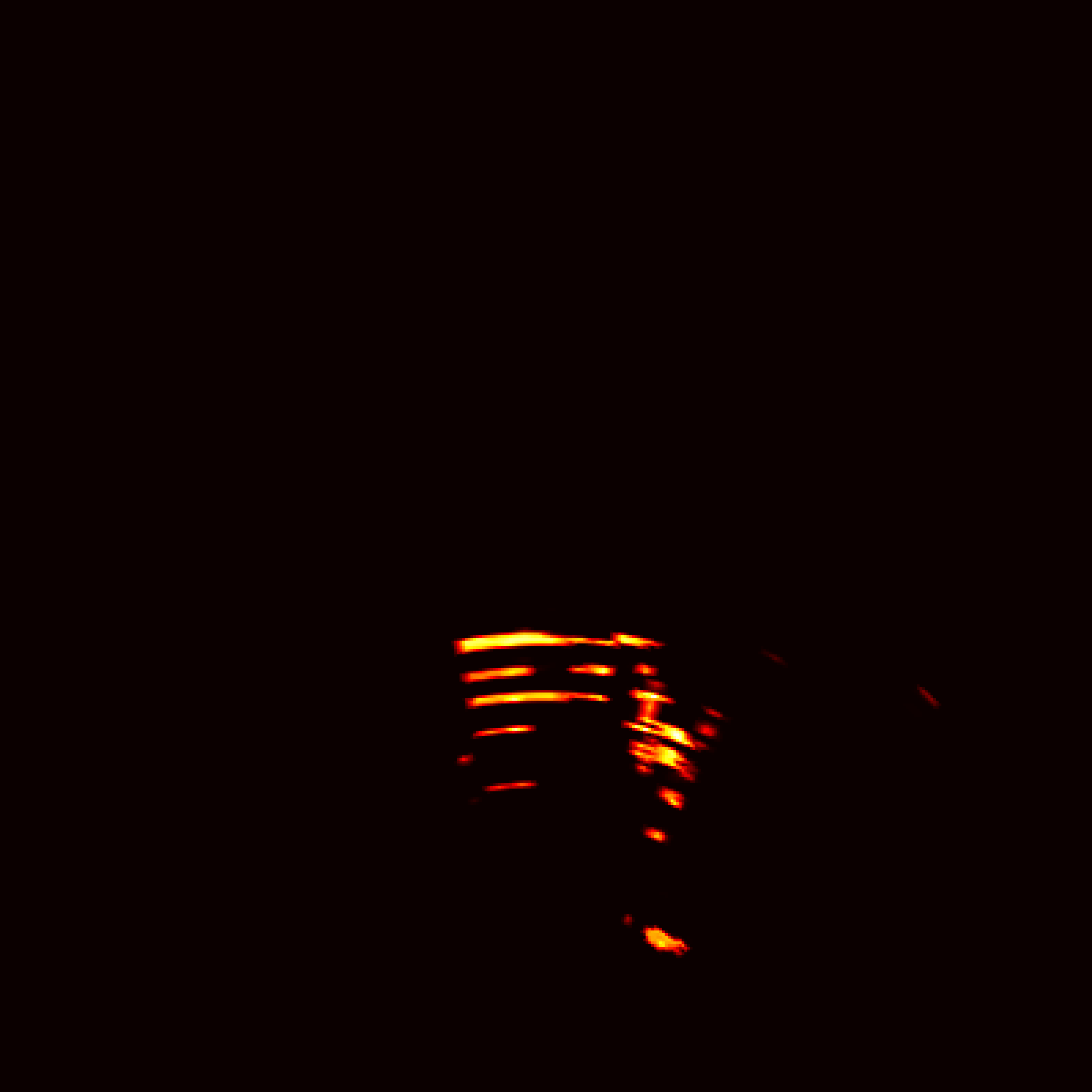} \\
\raisebox{2.0em}{\rotatebox{90}{\scriptsize\textbf{DART}}} & \includegraphics[width=0.165\linewidth]{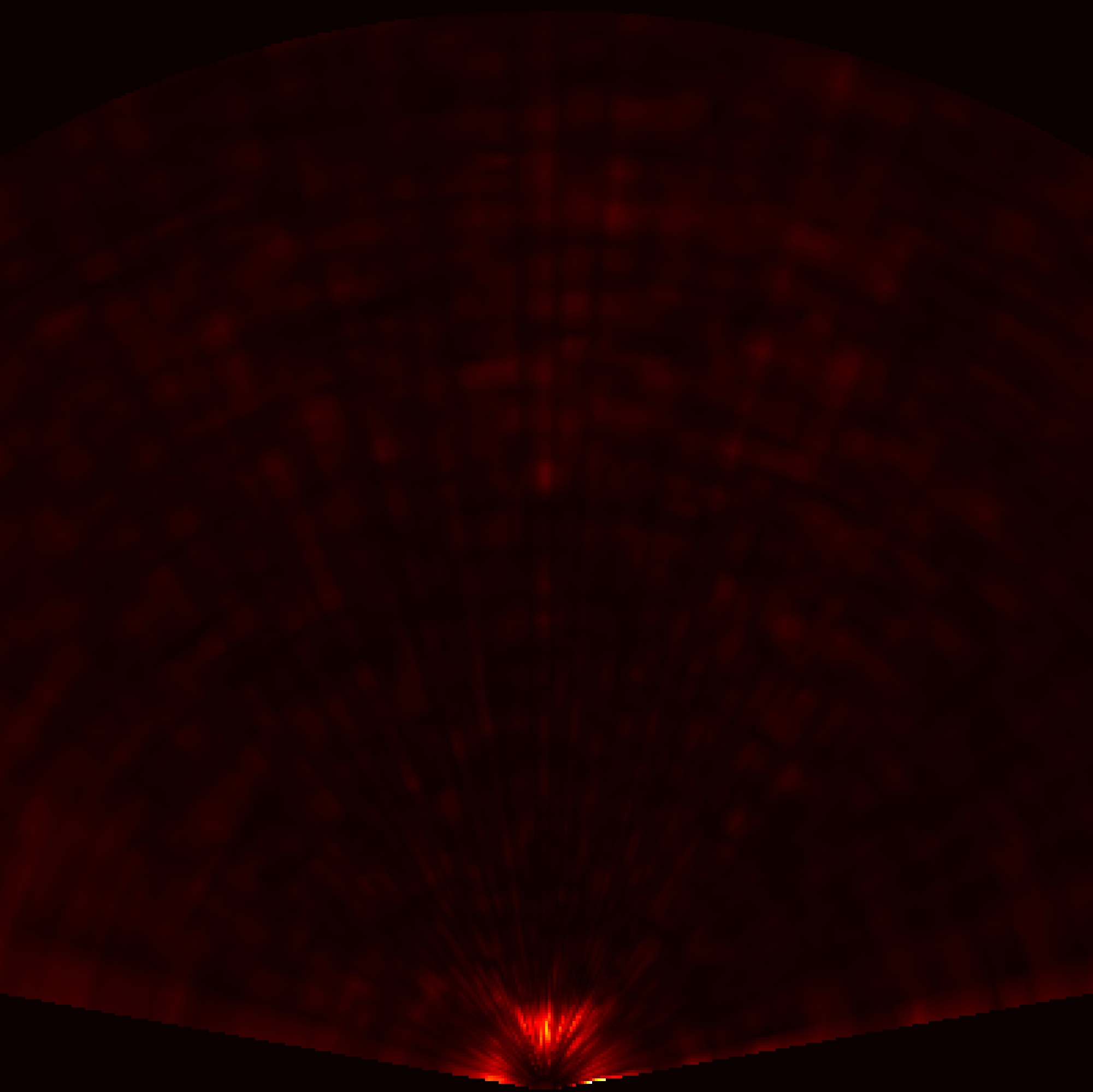} & \includegraphics[width=0.165\linewidth]{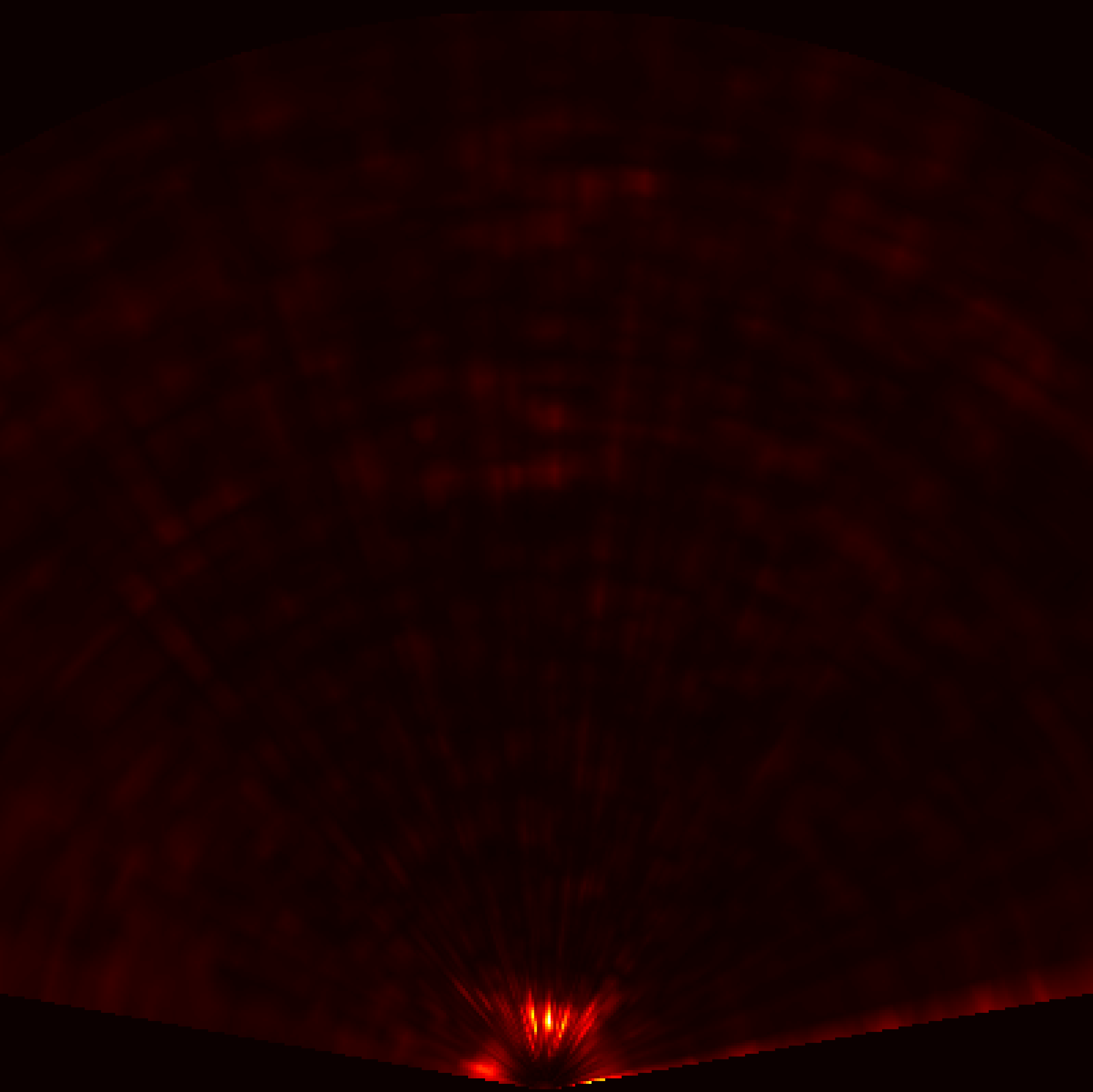} & \includegraphics[width=0.165\linewidth]{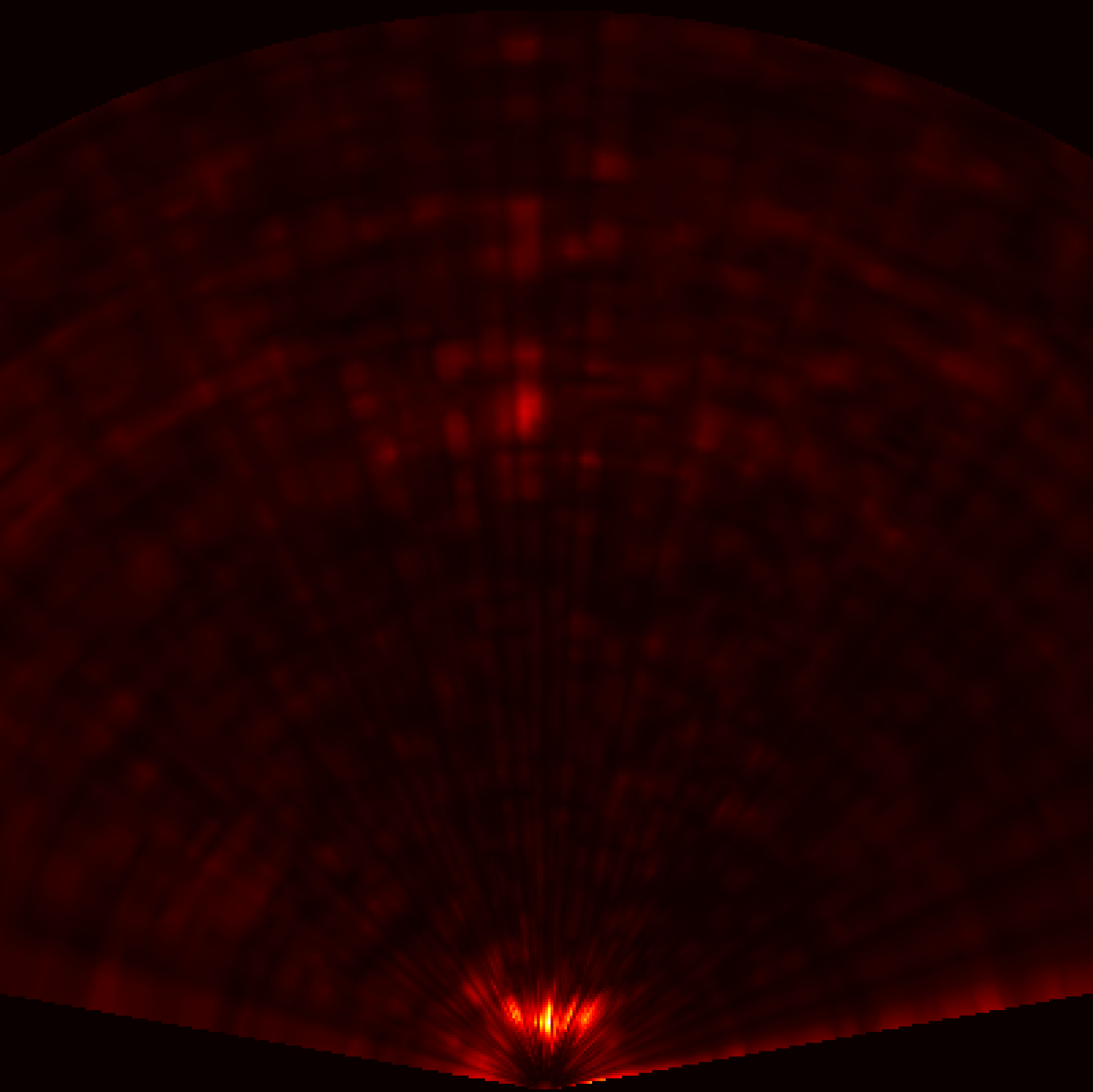} & \includegraphics[width=0.165\linewidth]{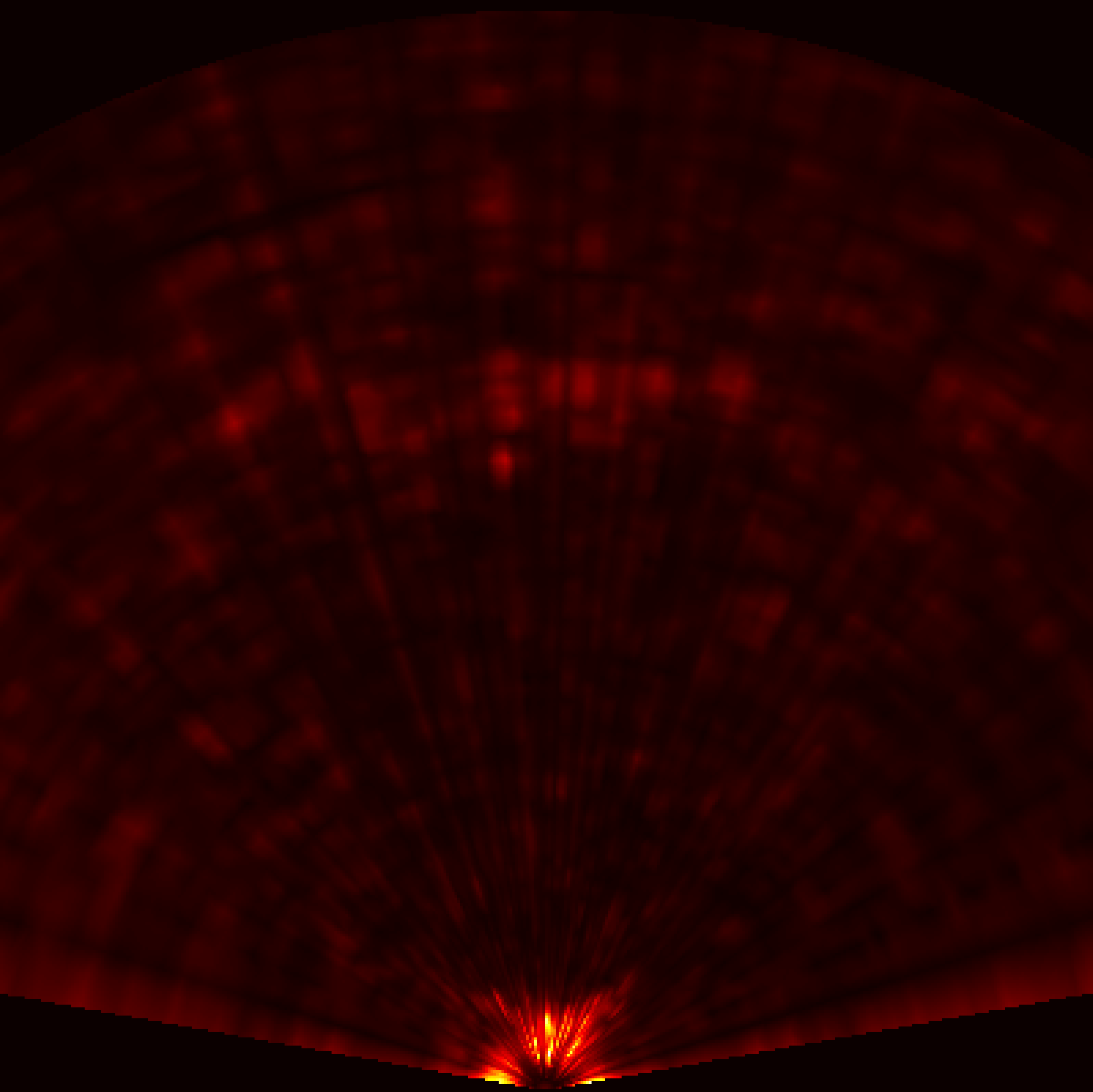} \\
\end{tabular}
\caption{\textbf{Per-frame $|\mathrm{RA}|$ comparison, scene S1\,F185, training frames $F\!-\!4$ to $F\!-\!1$.} Pearson correlation per column, left to right. 3DPS: 0.81, 0.79, 0.74, 0.87; mmIR: 0.51, 0.52, 0.46, 0.50; RadarSplat: 0.19, 0.23, 0.47, 0.39; Radar Fields: 0.17, 0.09, 0.06, 0.14; DART: -0.01, 0.00, -0.02, -0.02.}
\label{fig:supp_s1_f185}
\end{figure}

\begin{figure}[p]
\centering
\setlength{\tabcolsep}{1pt}\renewcommand{\arraystretch}{0.6}
\begin{tabular}{cccccc}
 & {\footnotesize $F\!+\!1$} & {\footnotesize $F\!+\!2$} & {\footnotesize $F\!+\!3$} & {\footnotesize $F\!+\!4$} & \textcolor{red}{\footnotesize\textbf{$F$}} \\
\raisebox{2.0em}{\rotatebox{90}{\scriptsize\textbf{GT}}} & \includegraphics[width=0.165\linewidth]{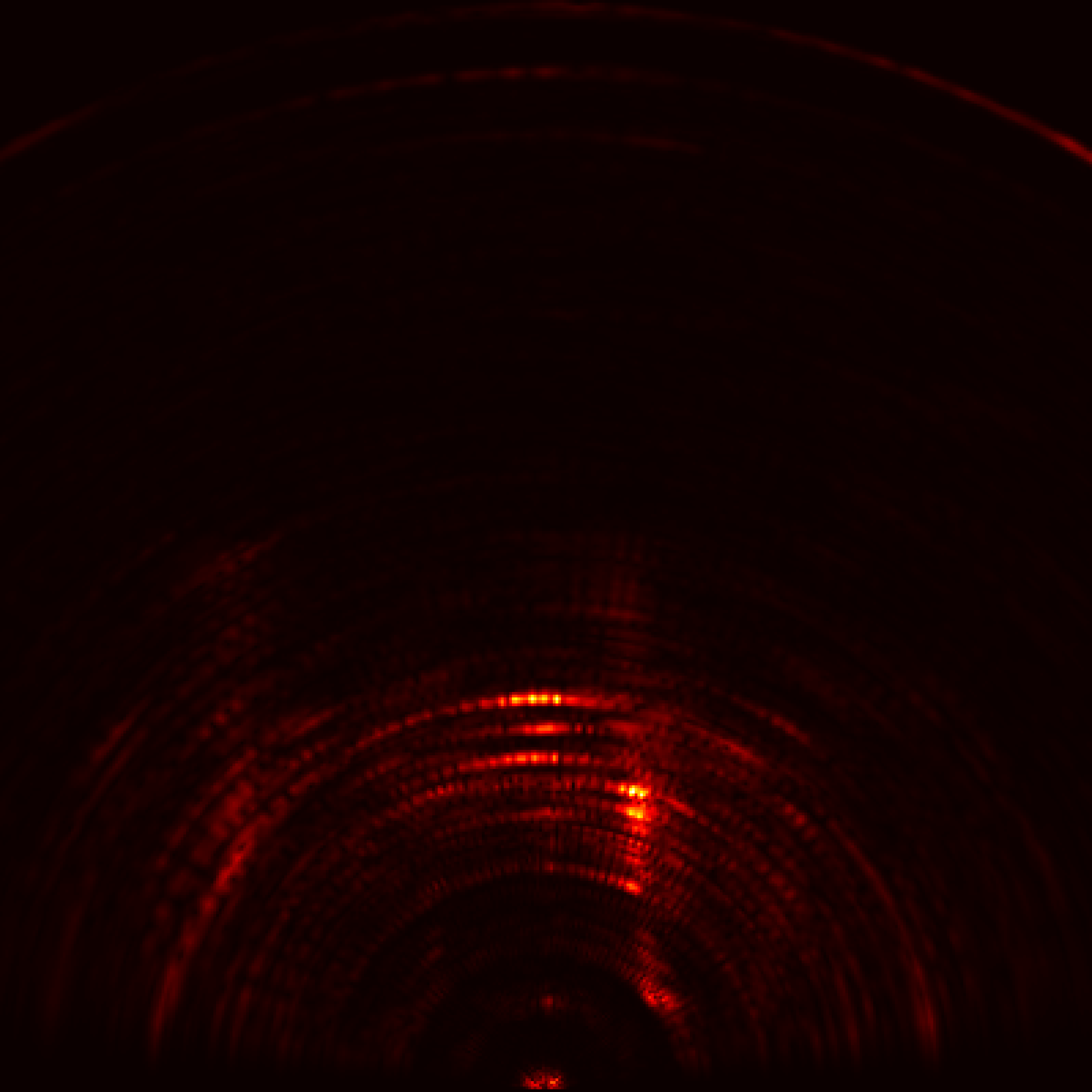} & \includegraphics[width=0.165\linewidth]{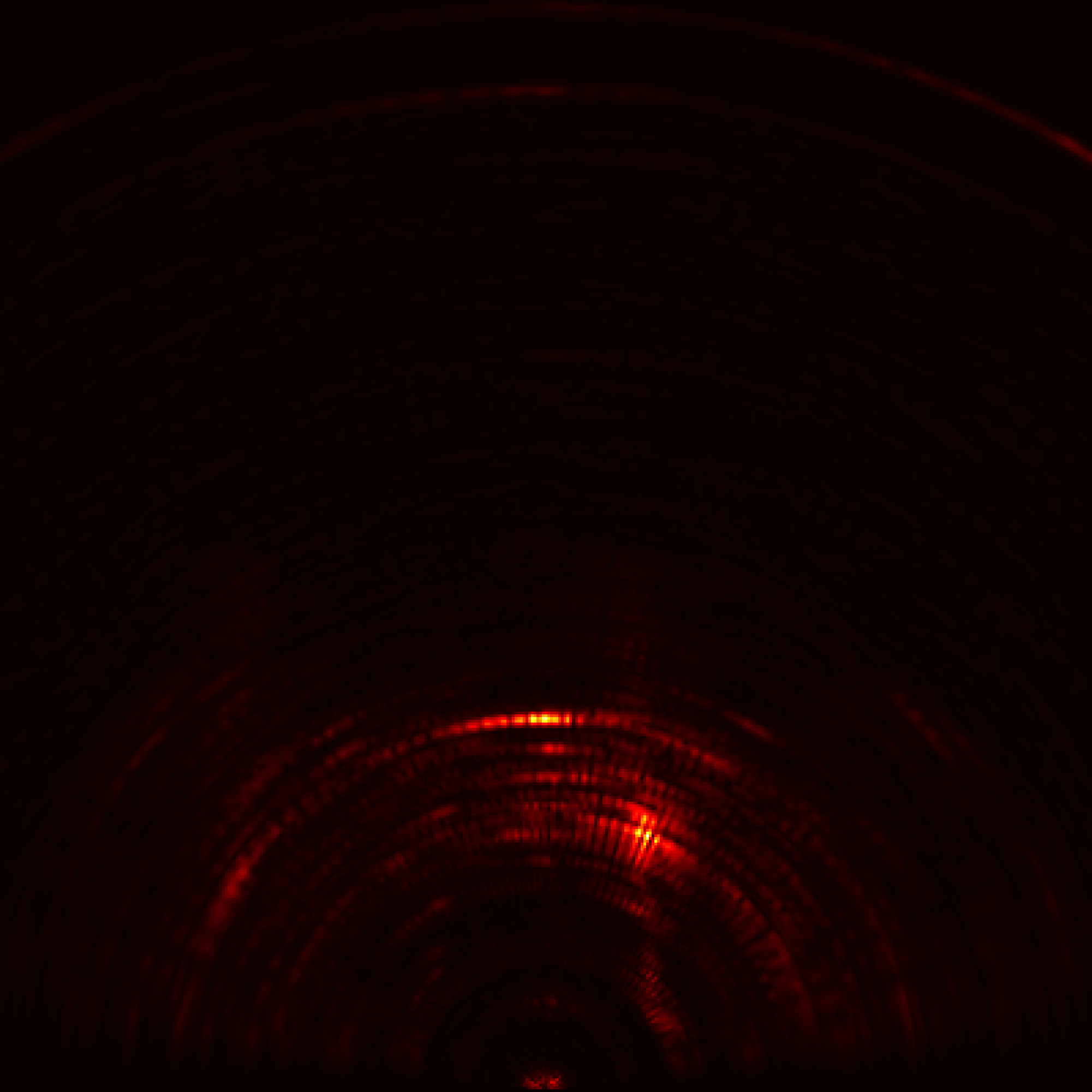} & \includegraphics[width=0.165\linewidth]{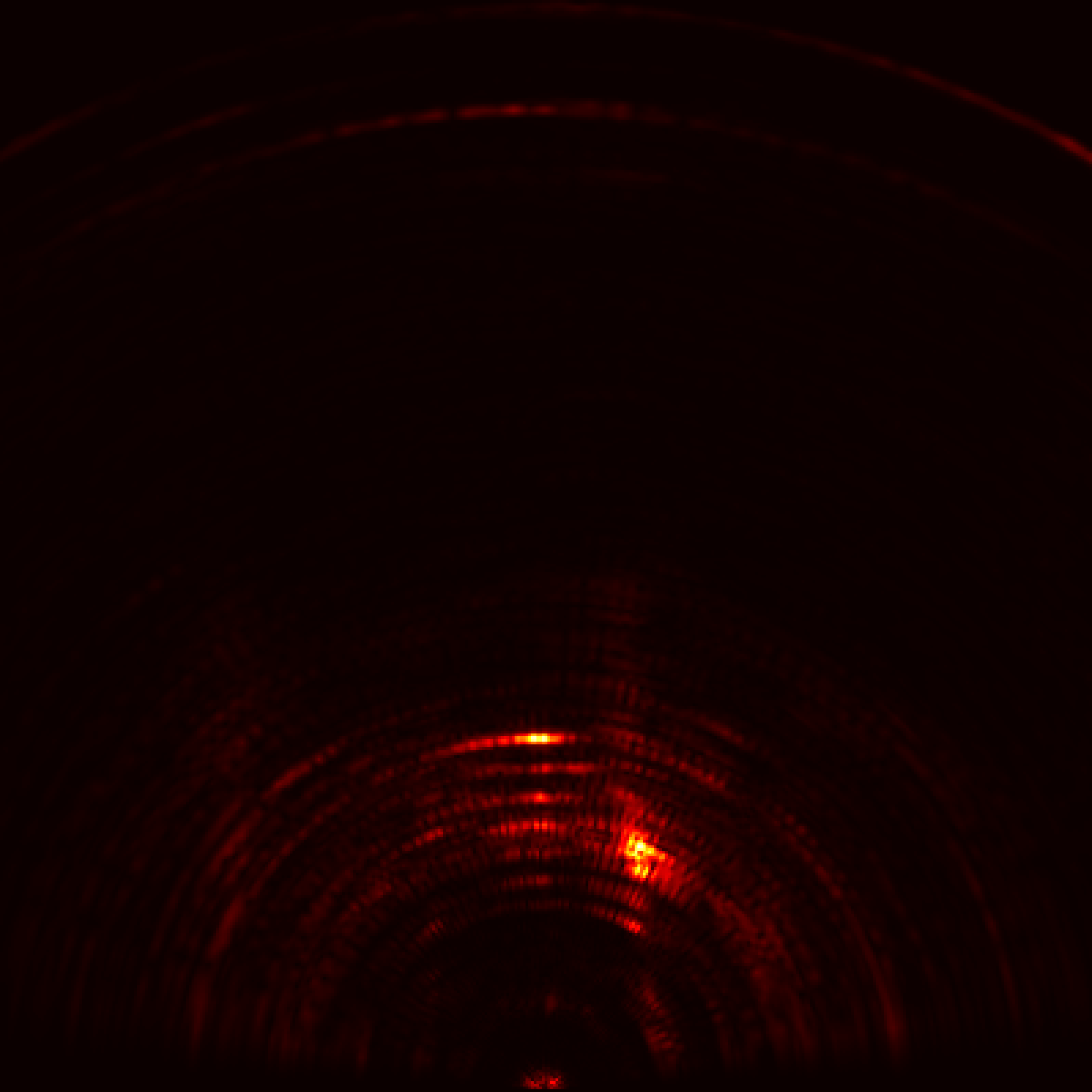} & \includegraphics[width=0.165\linewidth]{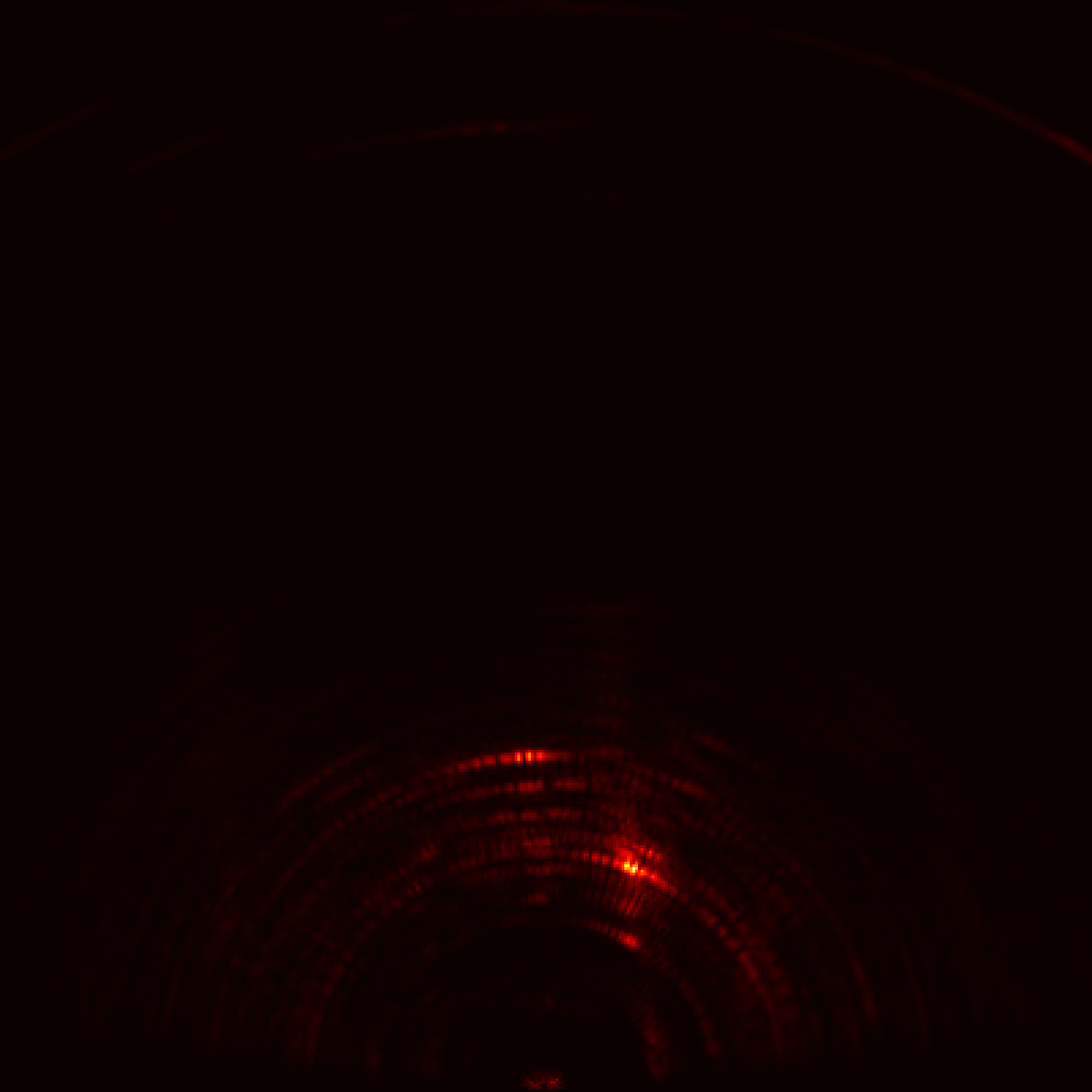} & \includegraphics[width=0.165\linewidth]{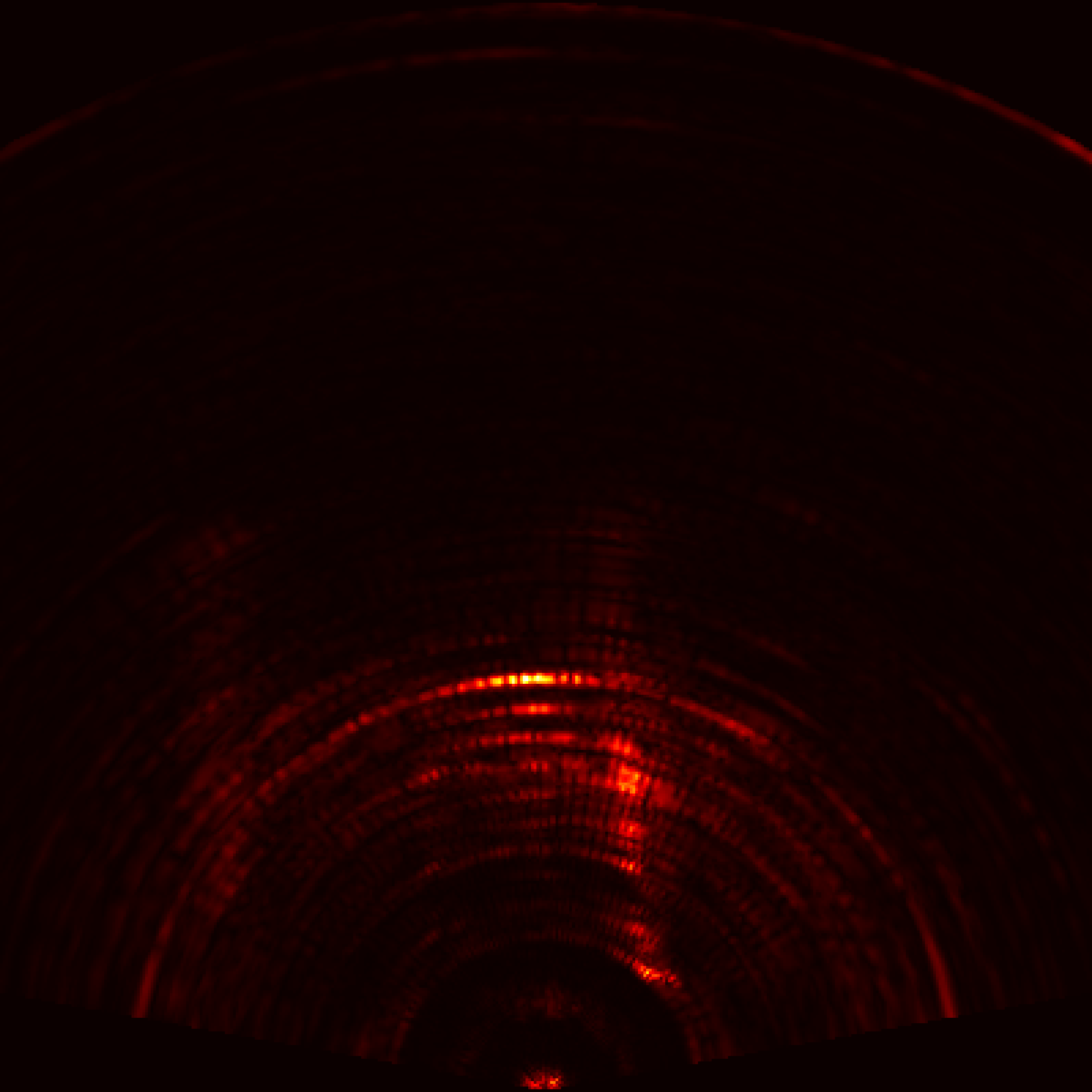} \\
\raisebox{2.0em}{\rotatebox{90}{\scriptsize\textbf{3DPS (Ours)}}} & \includegraphics[width=0.165\linewidth]{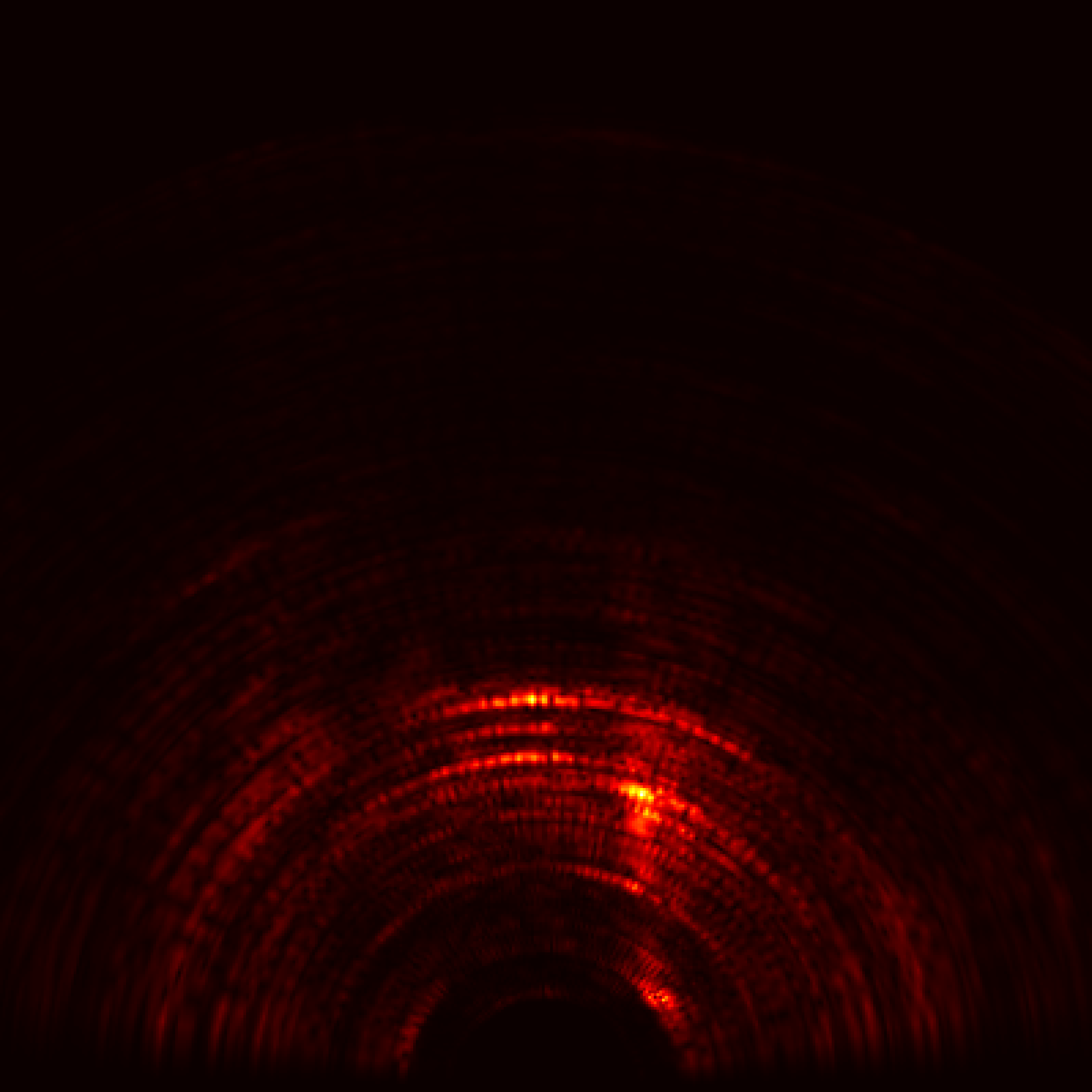} & \includegraphics[width=0.165\linewidth]{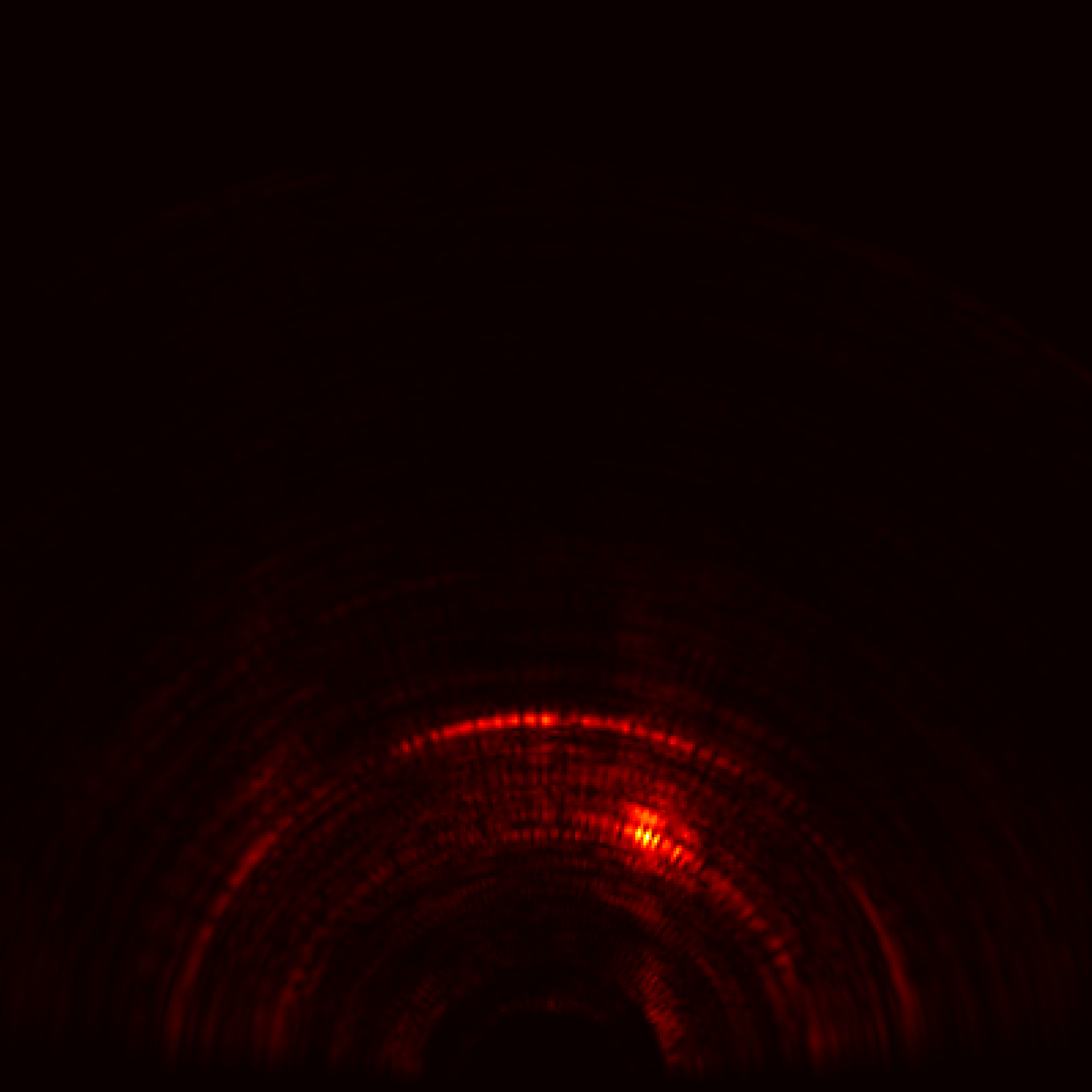} & \includegraphics[width=0.165\linewidth]{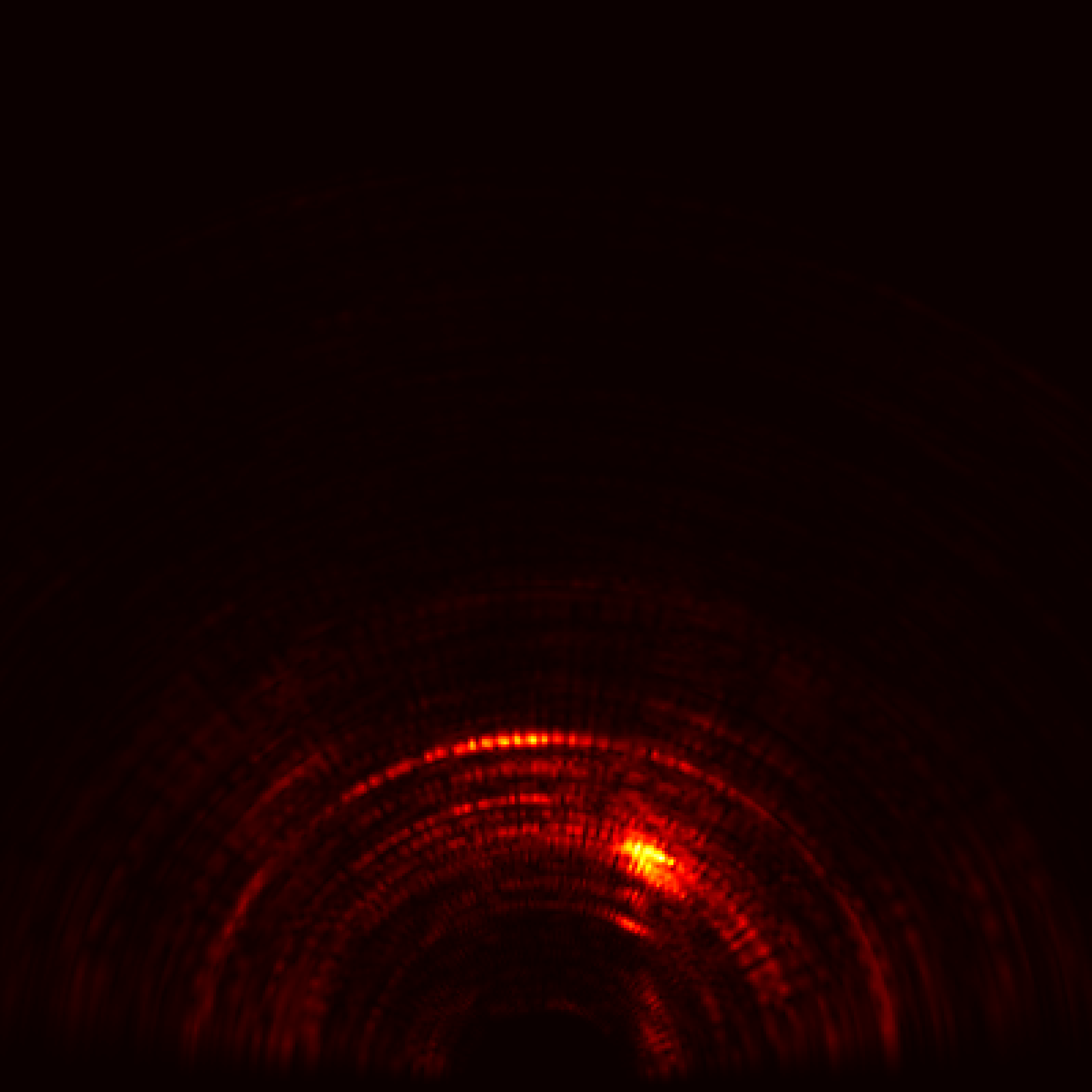} & \includegraphics[width=0.165\linewidth]{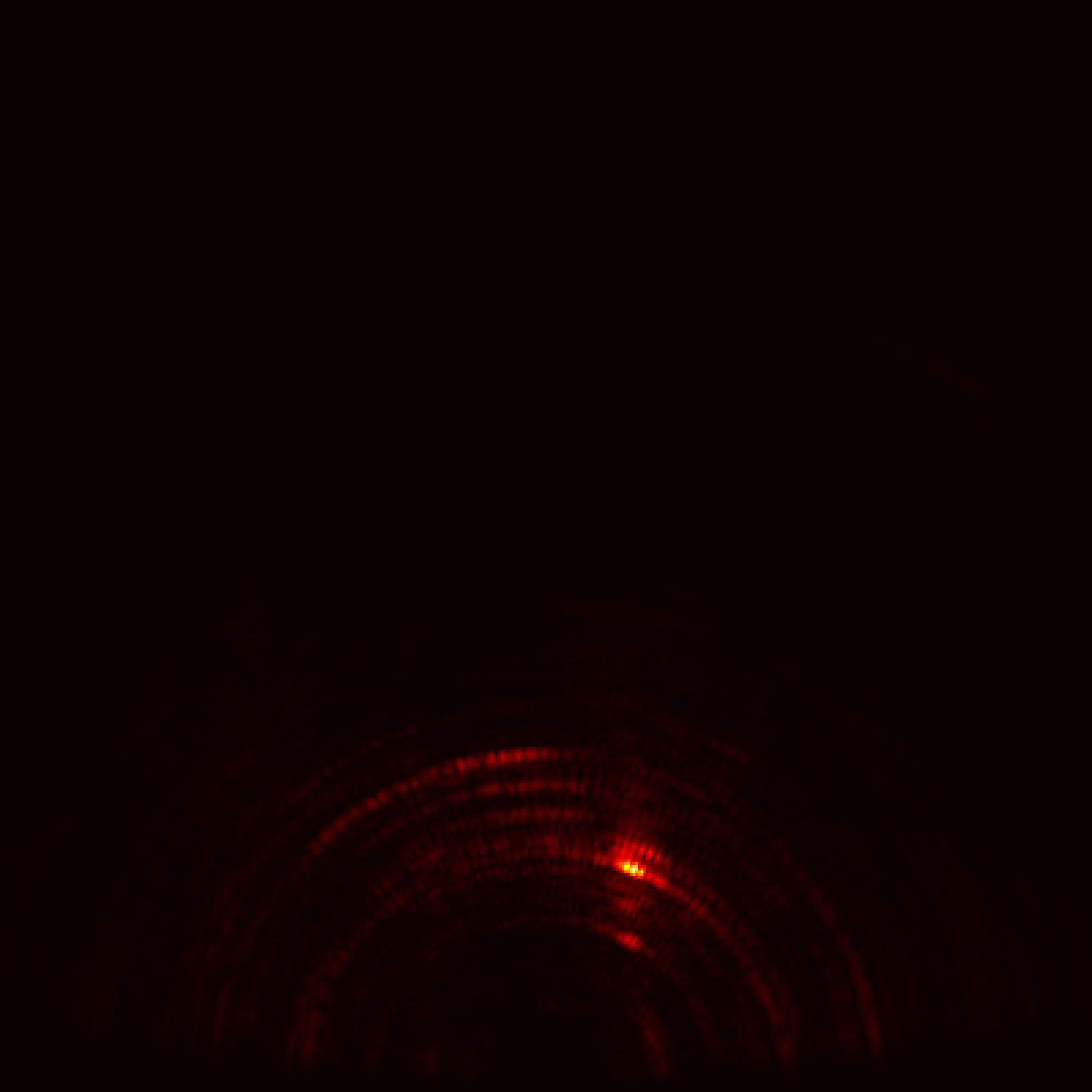} & \includegraphics[width=0.165\linewidth]{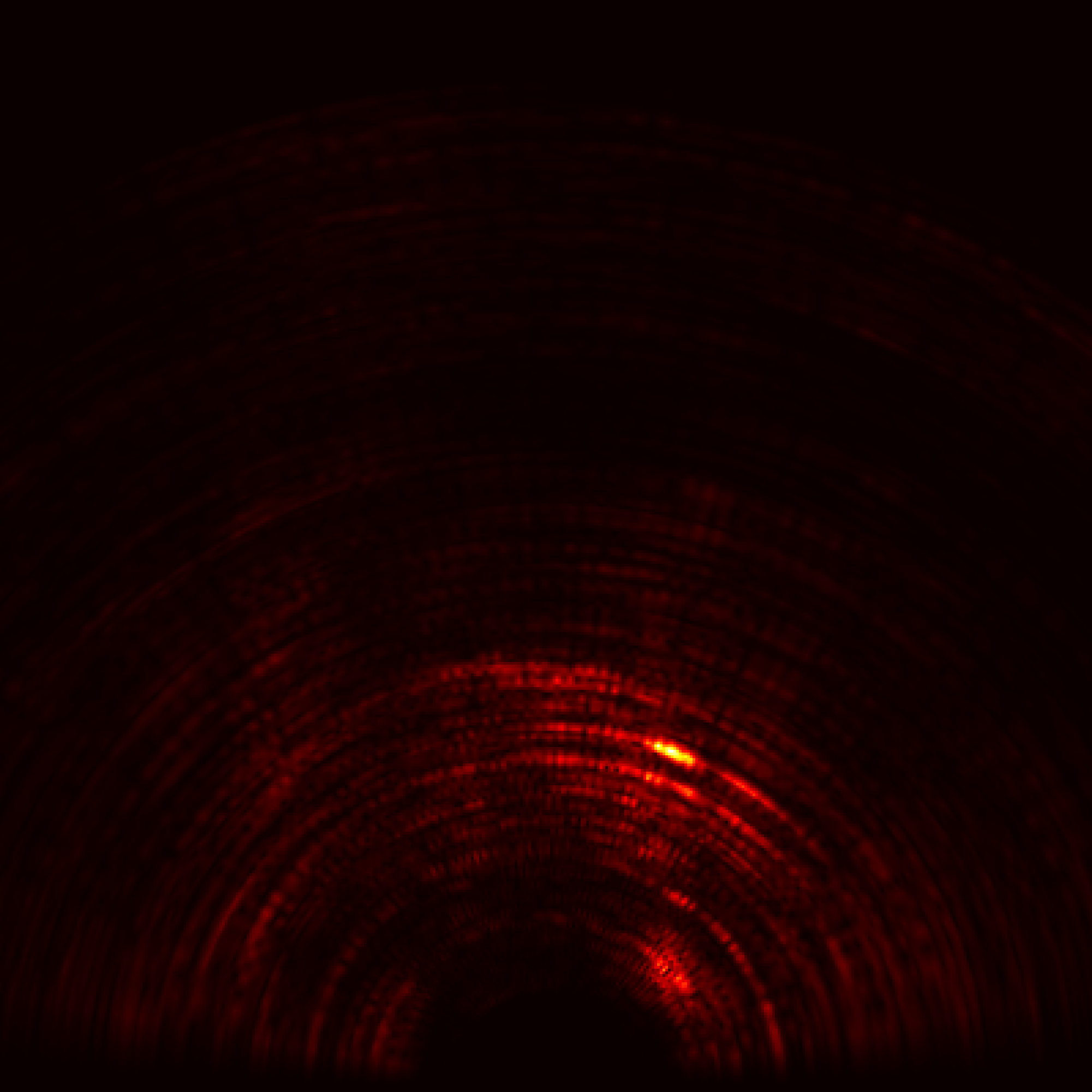} \\
\raisebox{2.0em}{\rotatebox{90}{\scriptsize\textbf{mmIR}}} & \includegraphics[width=0.165\linewidth]{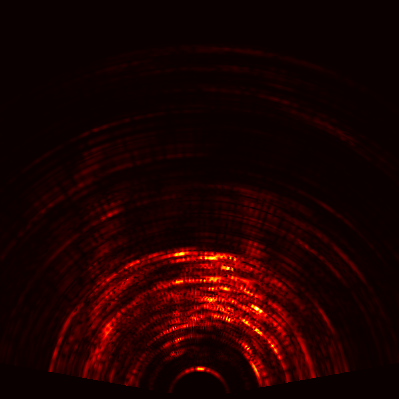} & \includegraphics[width=0.165\linewidth]{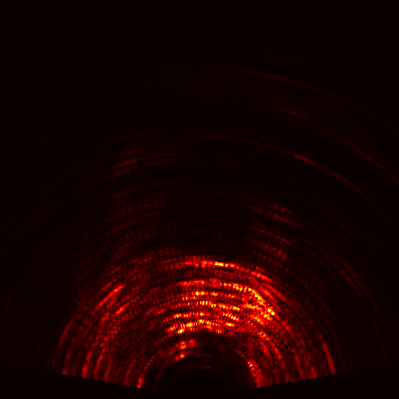} & \includegraphics[width=0.165\linewidth]{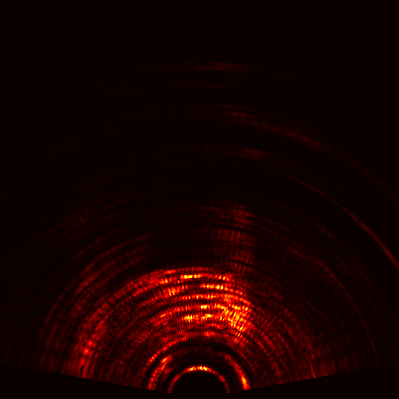} & \includegraphics[width=0.165\linewidth]{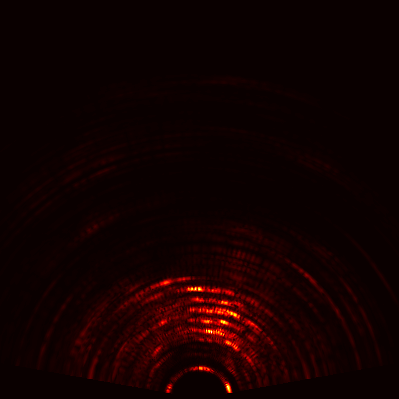} & \includegraphics[width=0.165\linewidth]{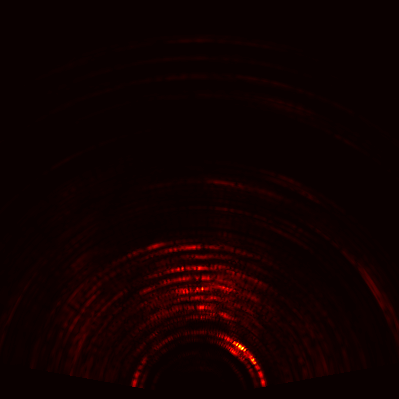} \\
\raisebox{2.0em}{\rotatebox{90}{\scriptsize\textbf{RadarSplat}}} & \includegraphics[width=0.165\linewidth]{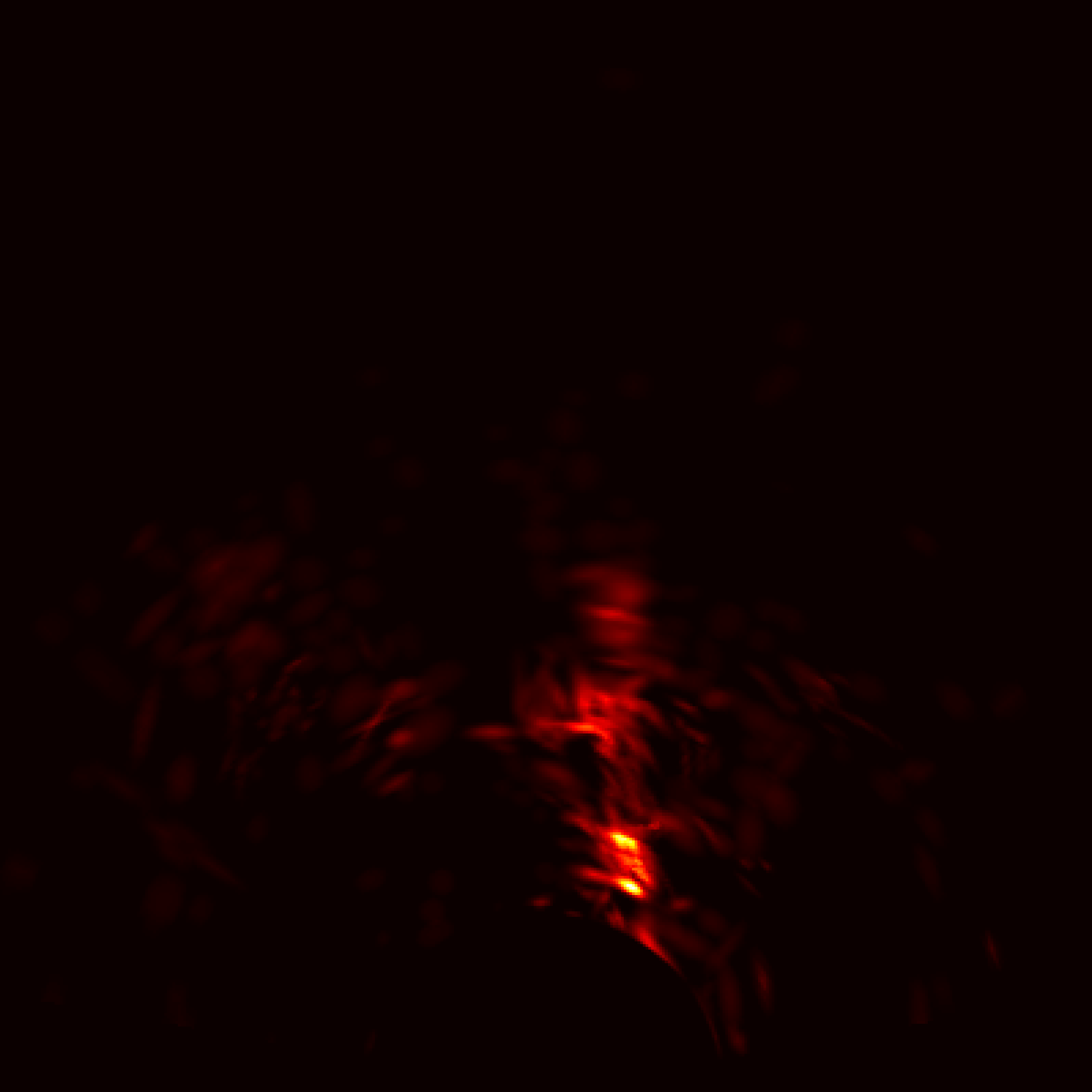} & \includegraphics[width=0.165\linewidth]{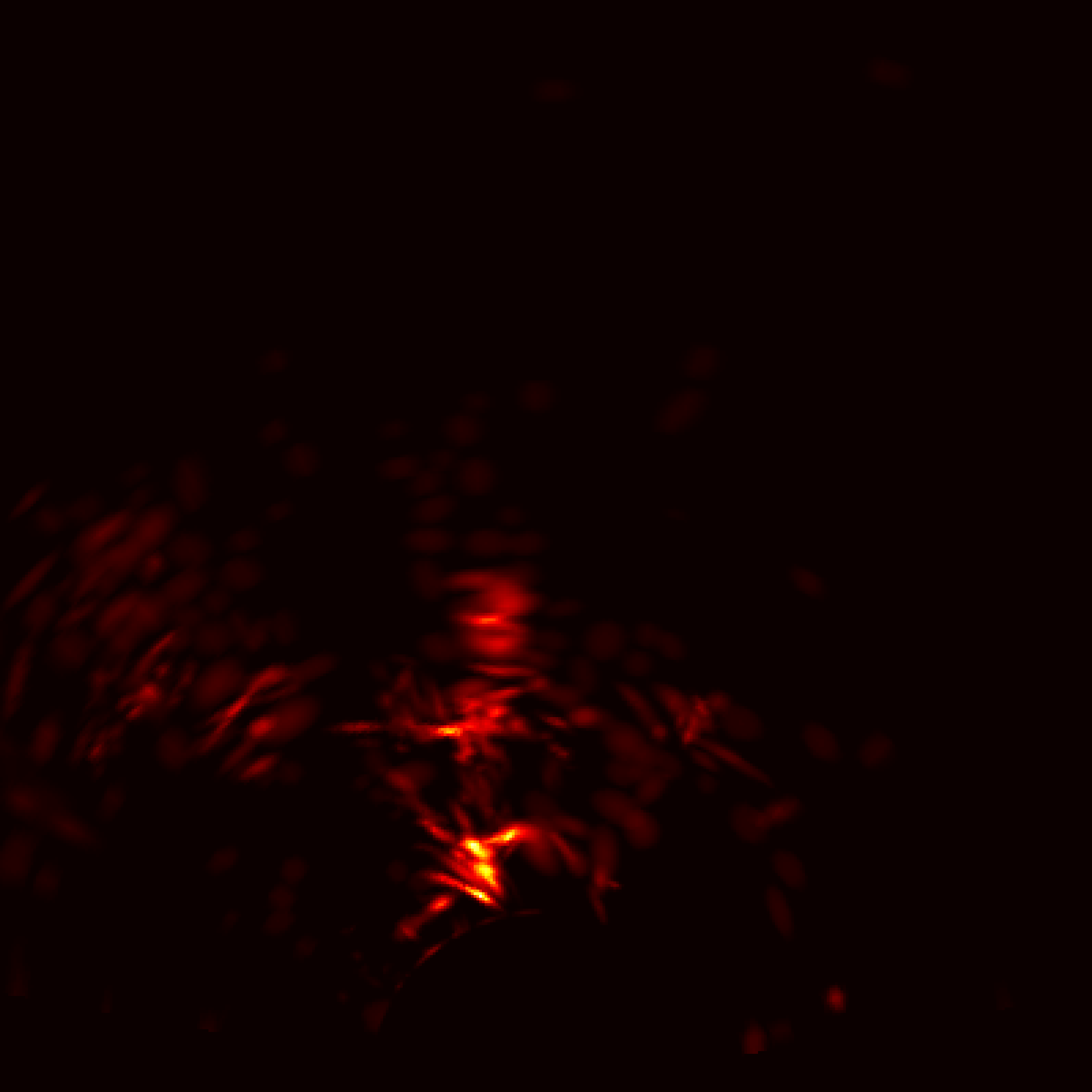} & \includegraphics[width=0.165\linewidth]{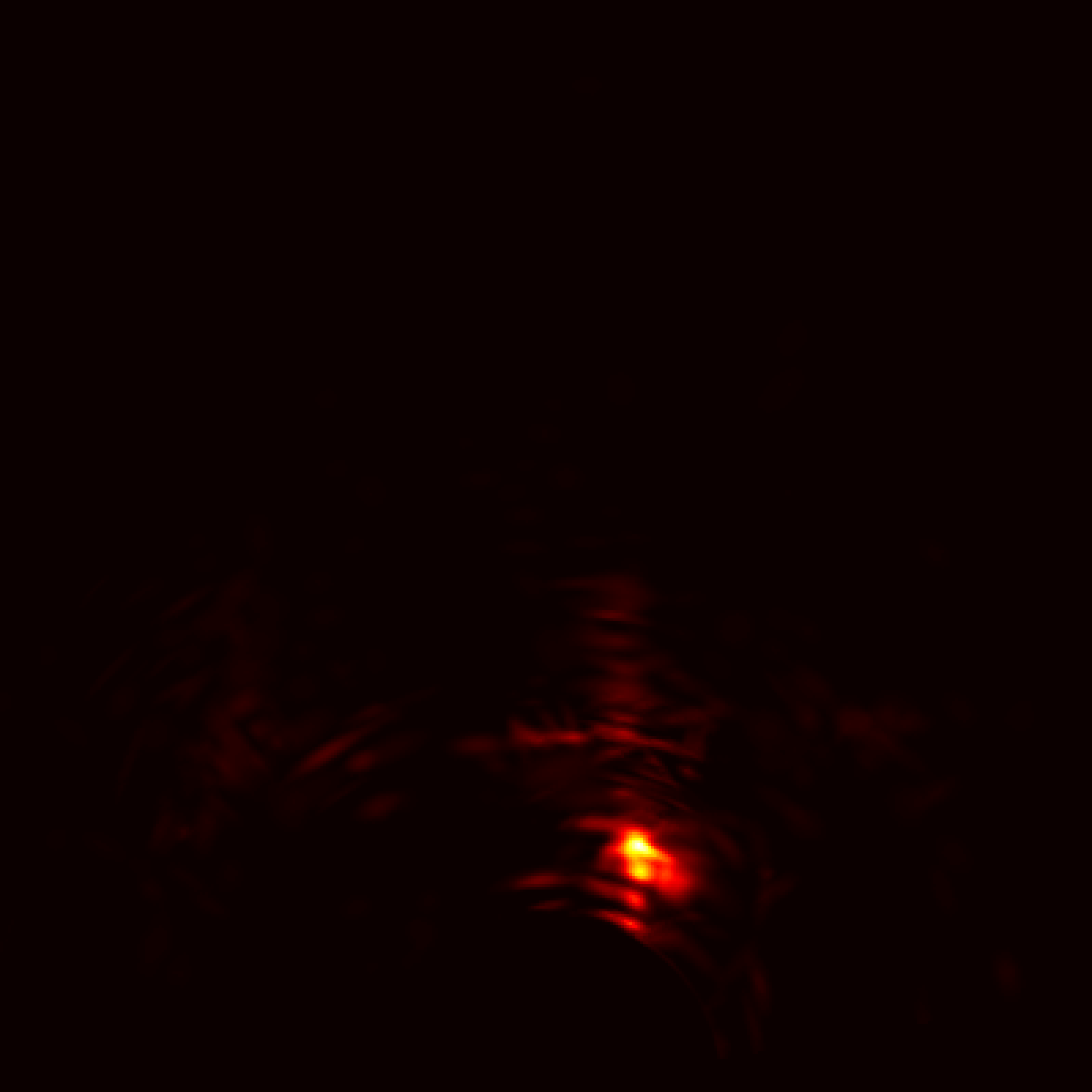} & \includegraphics[width=0.165\linewidth]{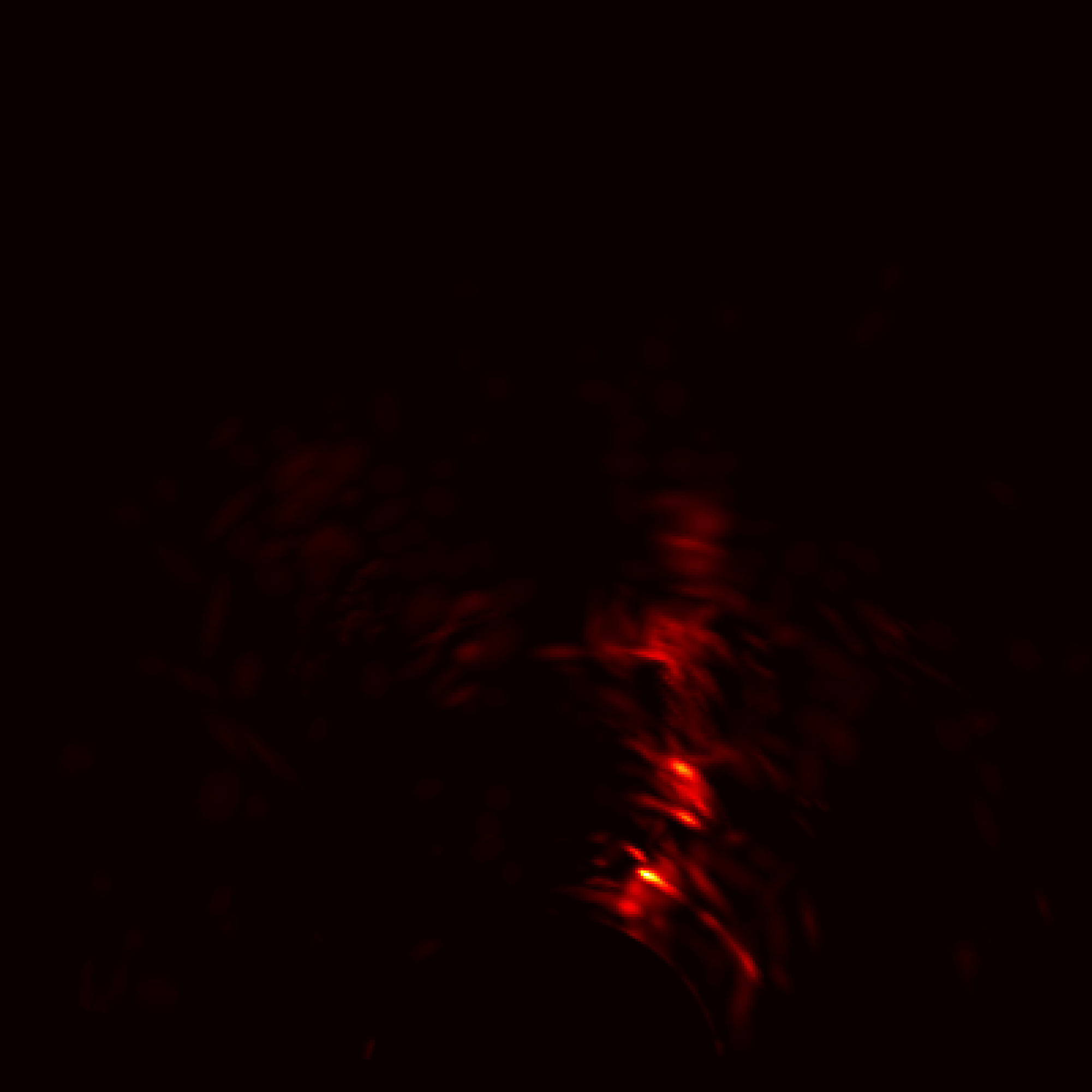} & \includegraphics[width=0.165\linewidth]{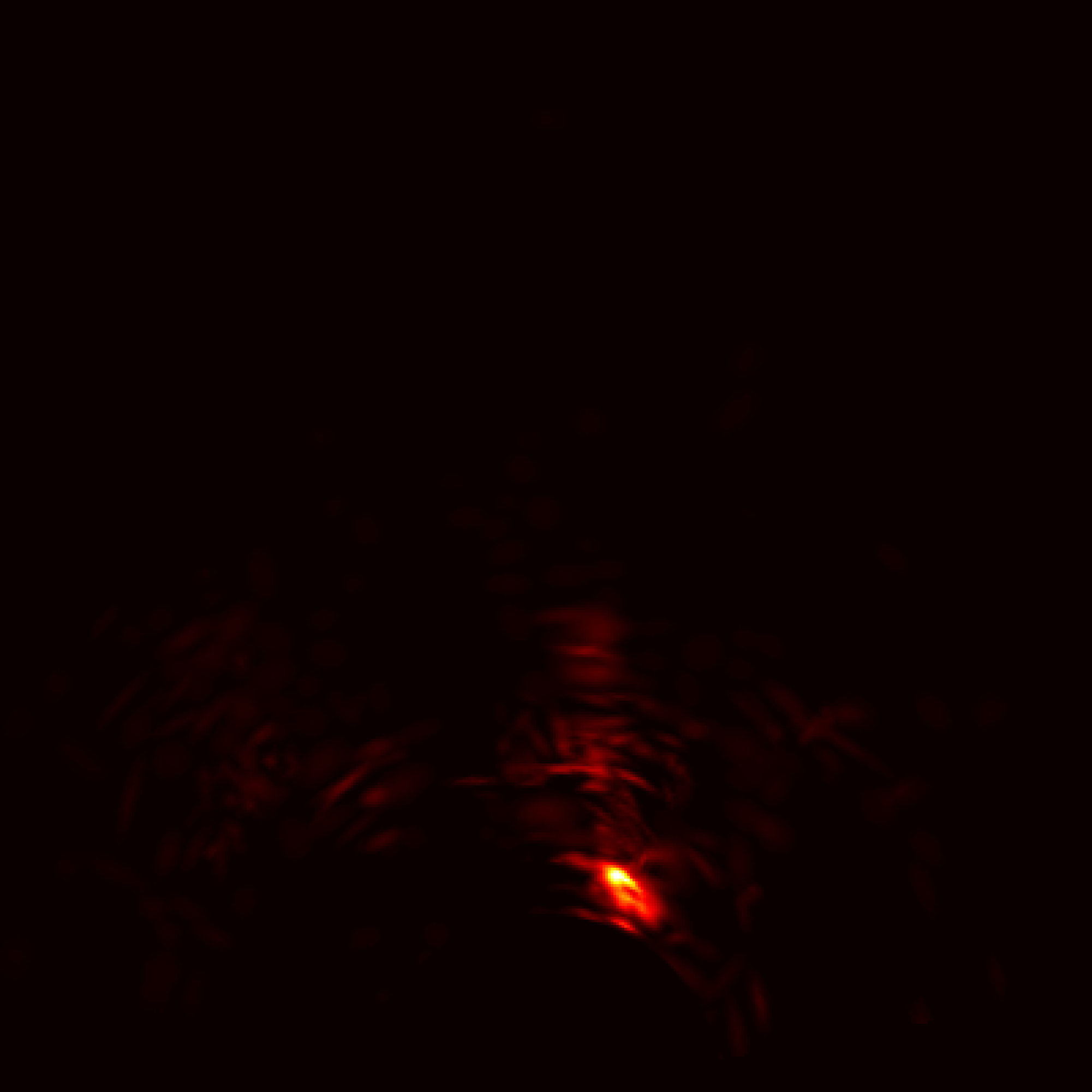} \\
\raisebox{2.0em}{\rotatebox{90}{\scriptsize\textbf{Radar Fields}}} & \includegraphics[width=0.165\linewidth]{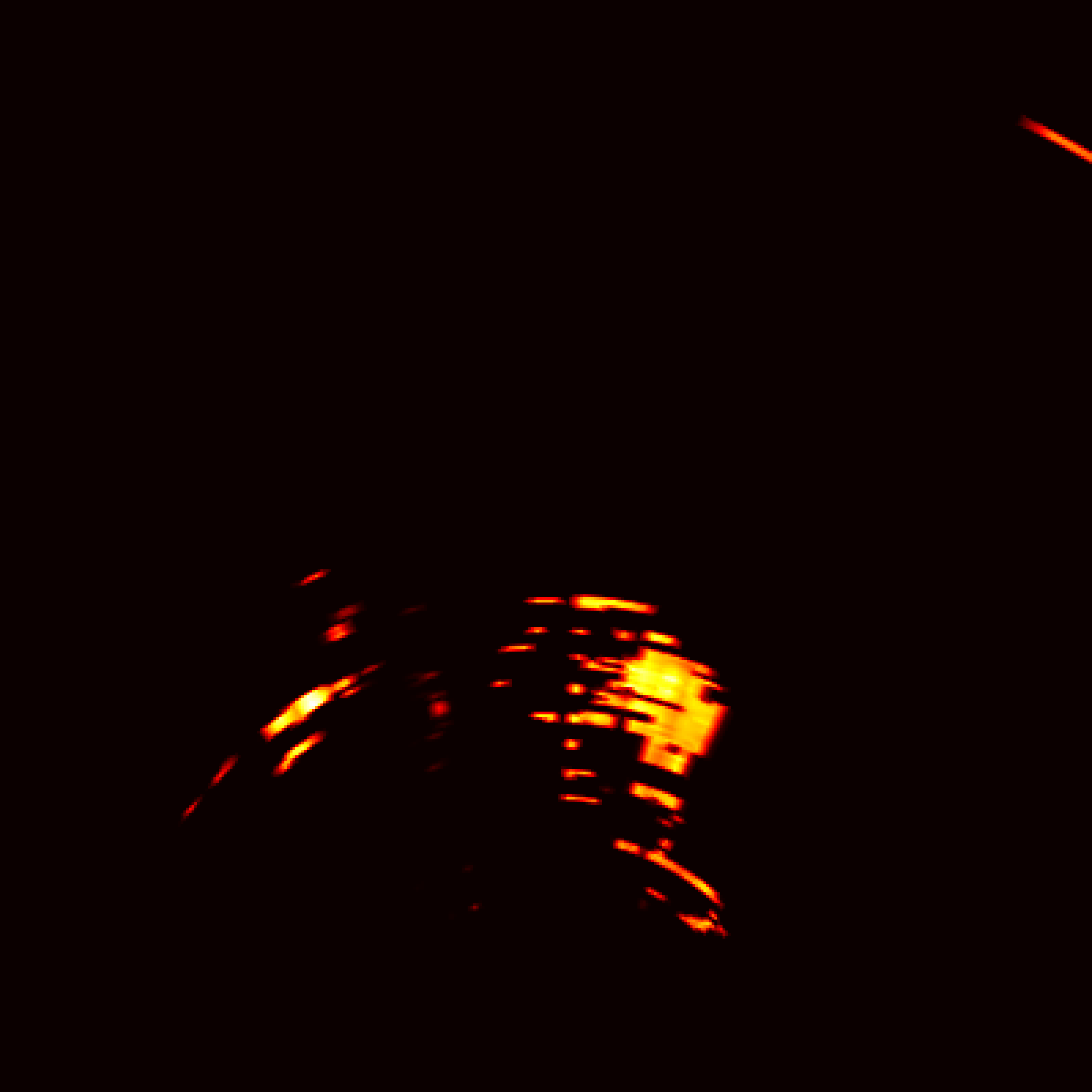} & \includegraphics[width=0.165\linewidth]{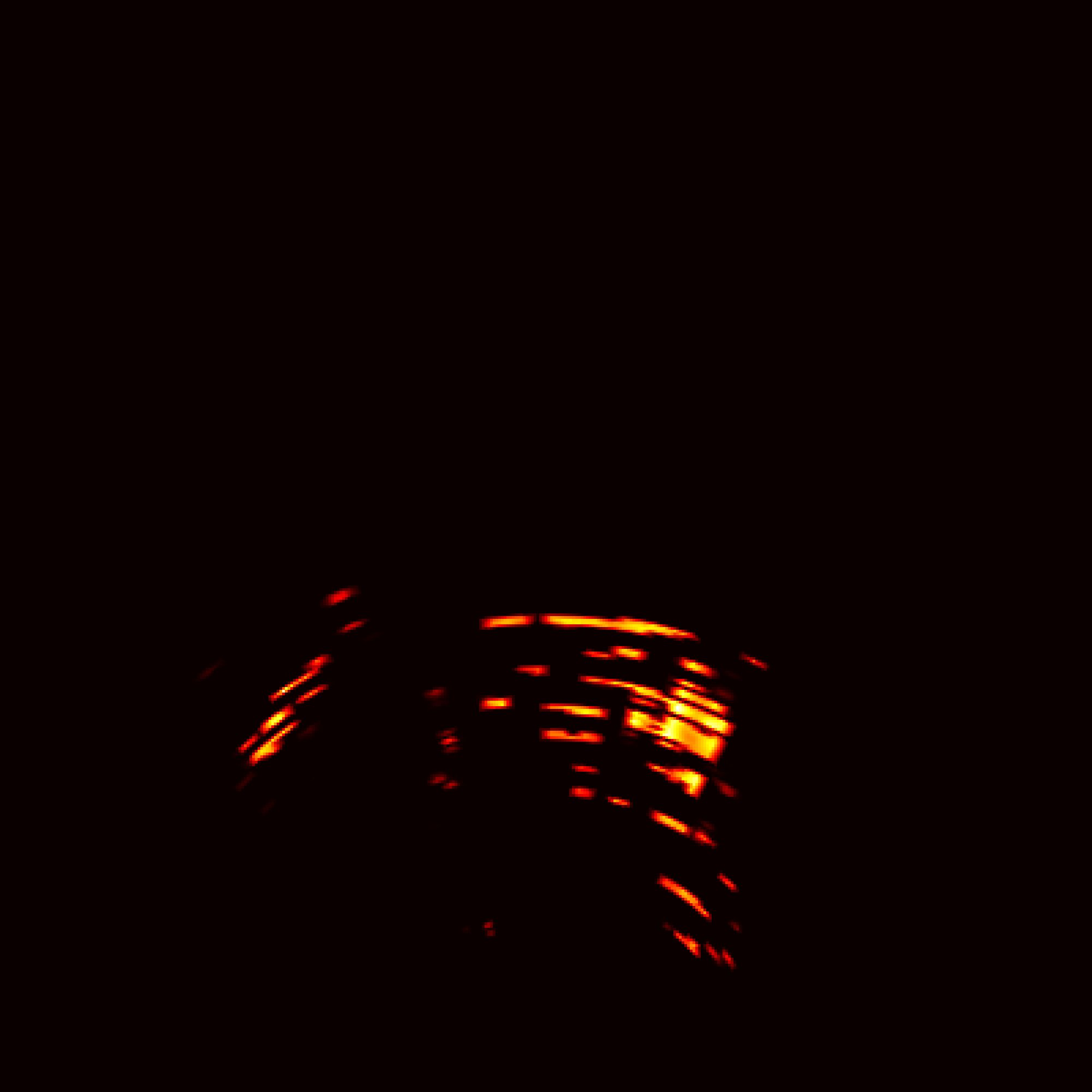} & \includegraphics[width=0.165\linewidth]{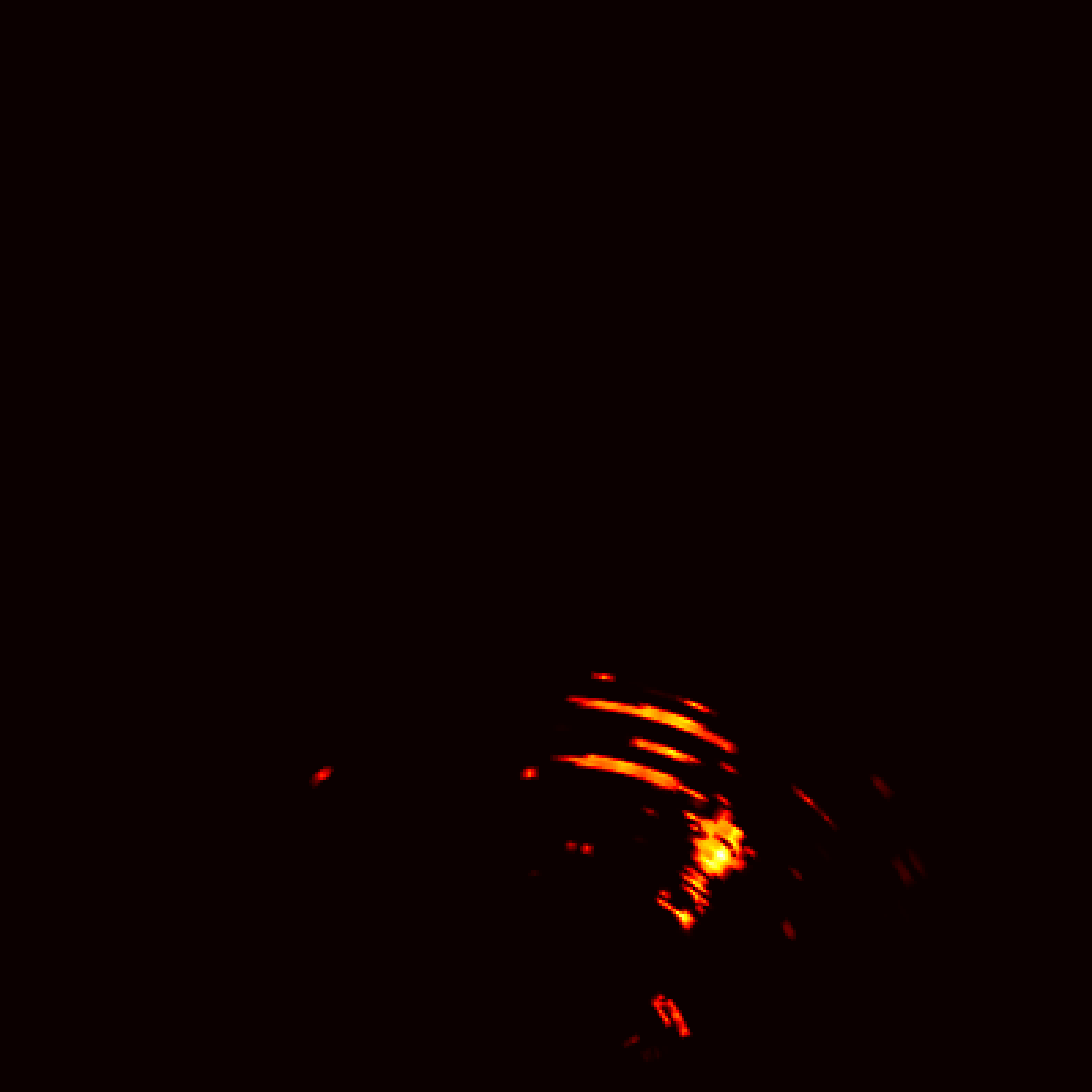} & \includegraphics[width=0.165\linewidth]{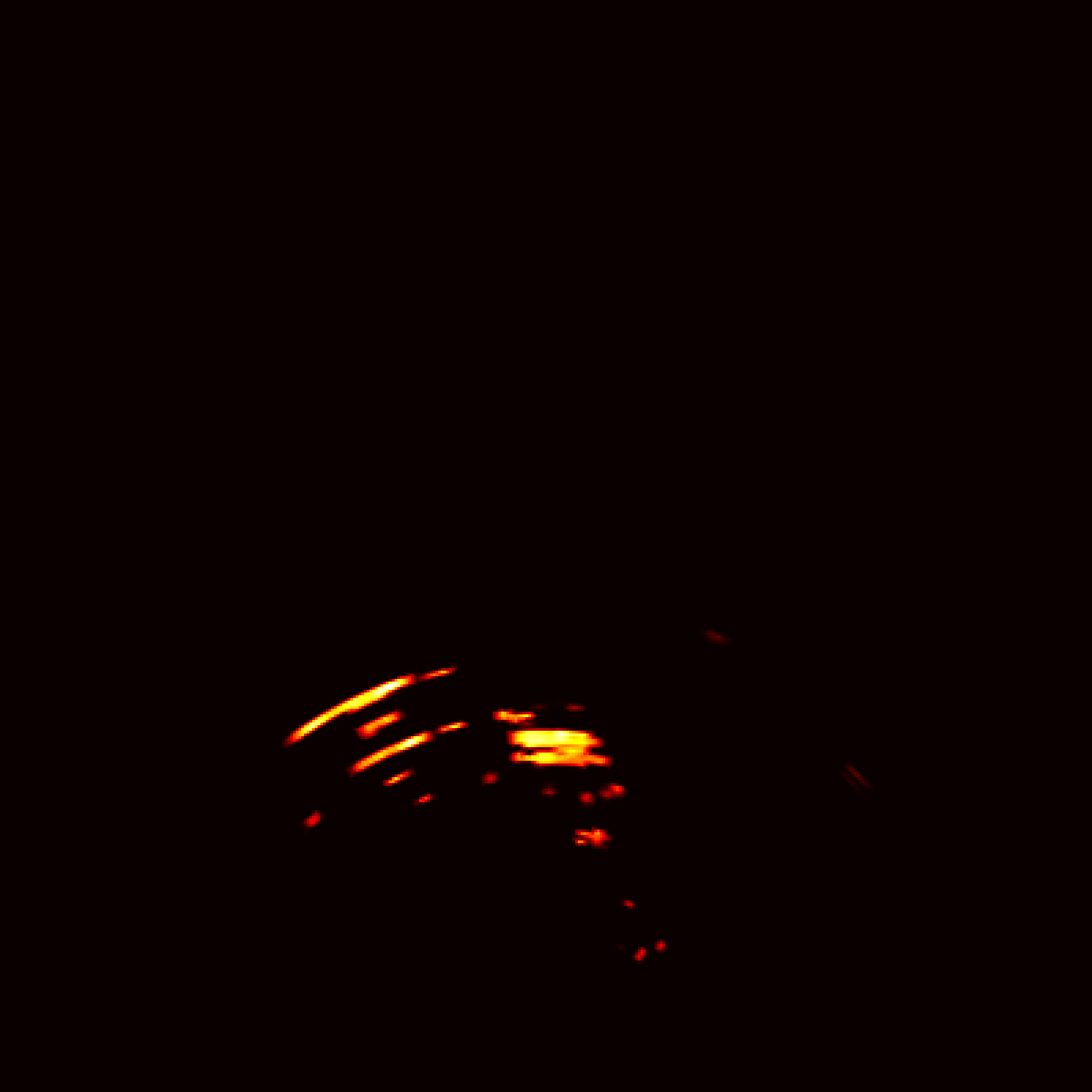} & \includegraphics[width=0.165\linewidth]{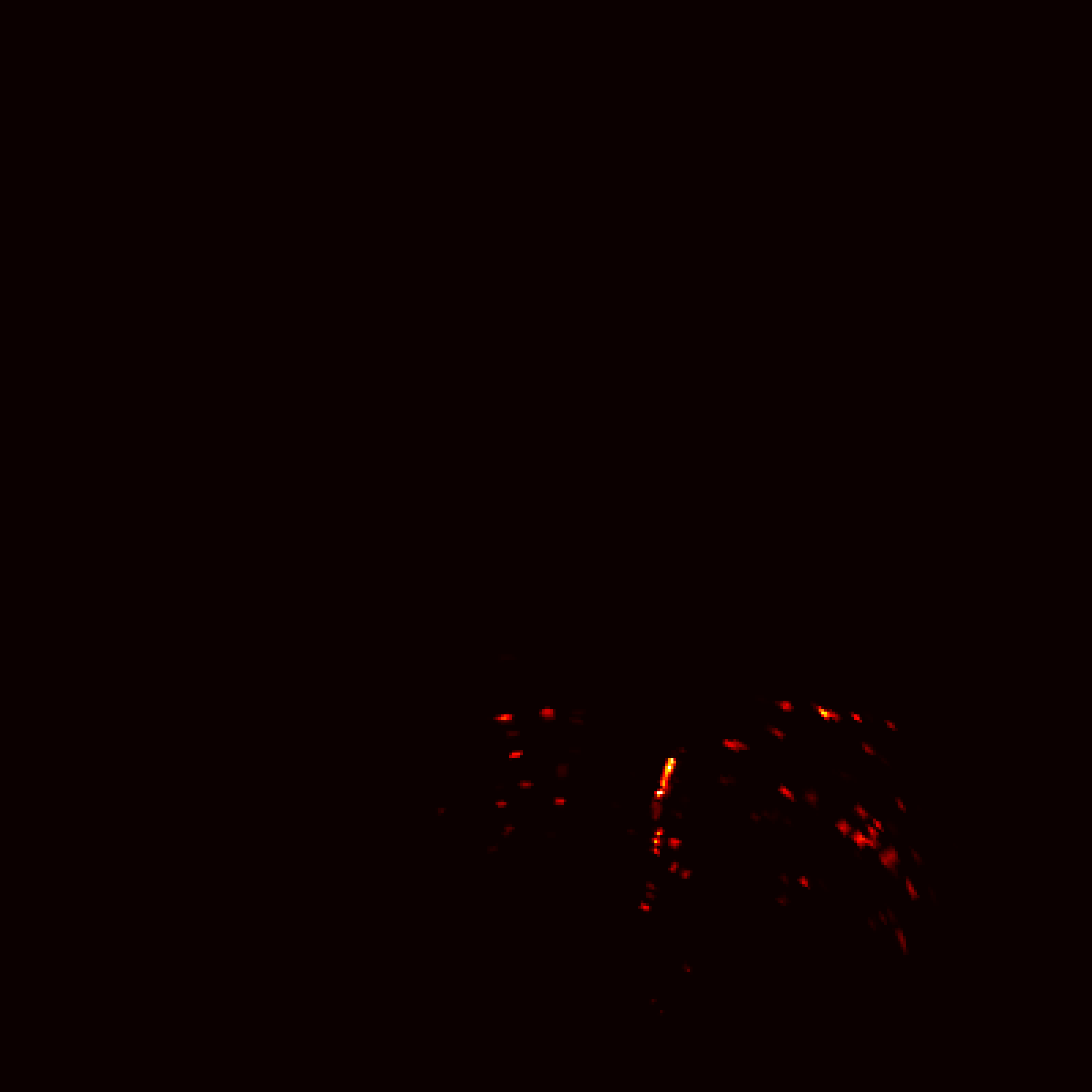} \\
\raisebox{2.0em}{\rotatebox{90}{\scriptsize\textbf{DART}}} & \includegraphics[width=0.165\linewidth]{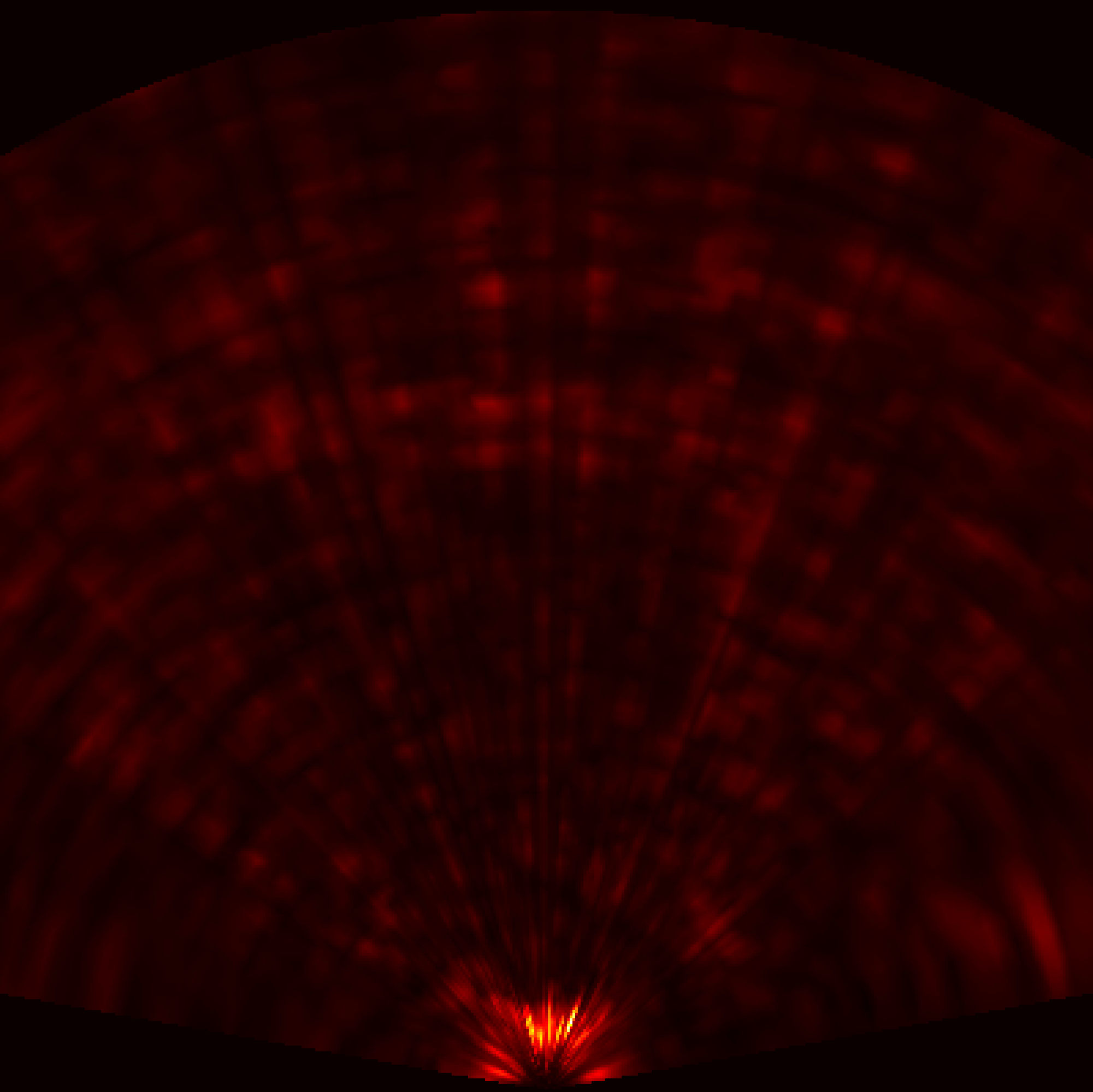} & \includegraphics[width=0.165\linewidth]{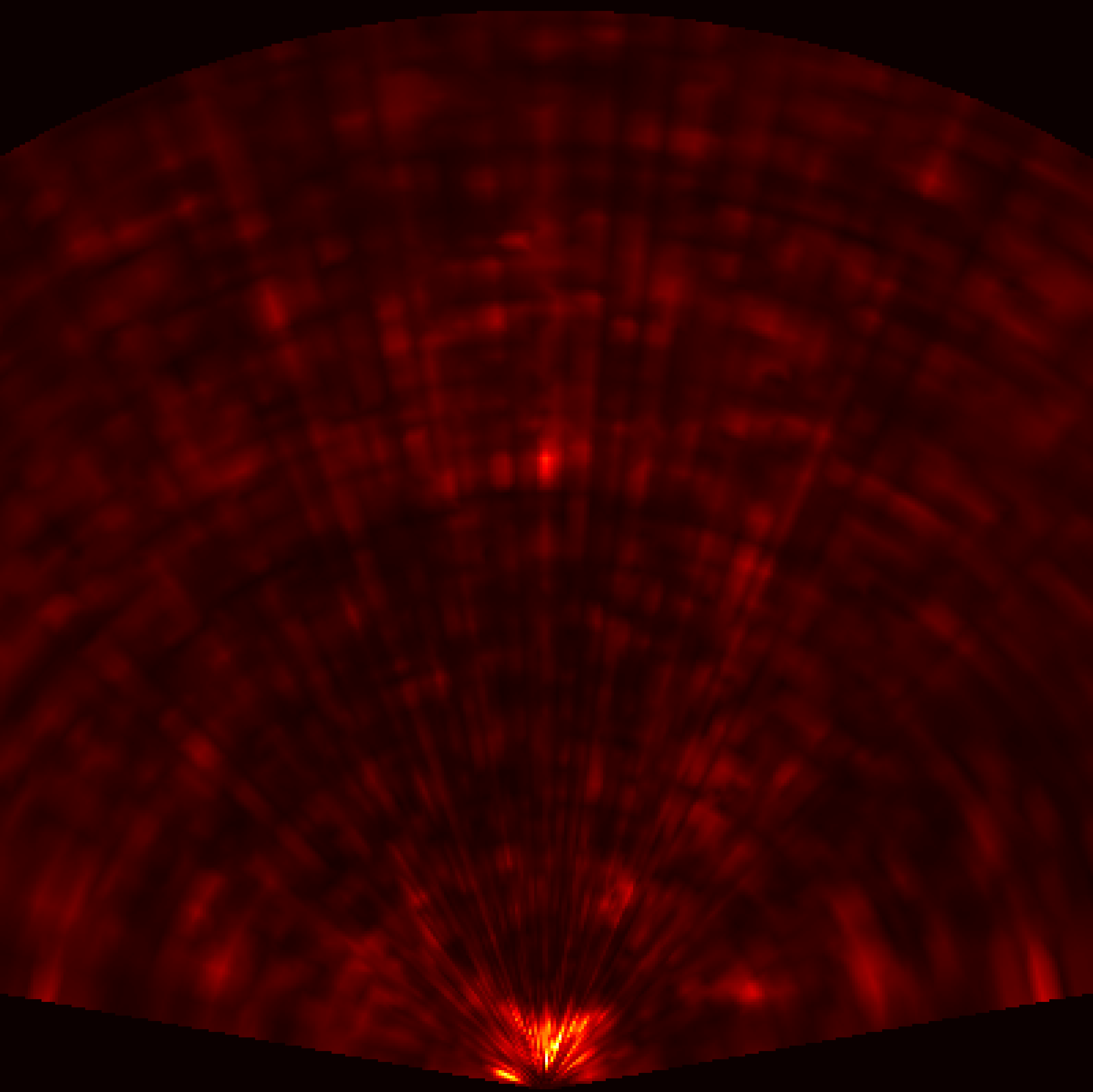} & \includegraphics[width=0.165\linewidth]{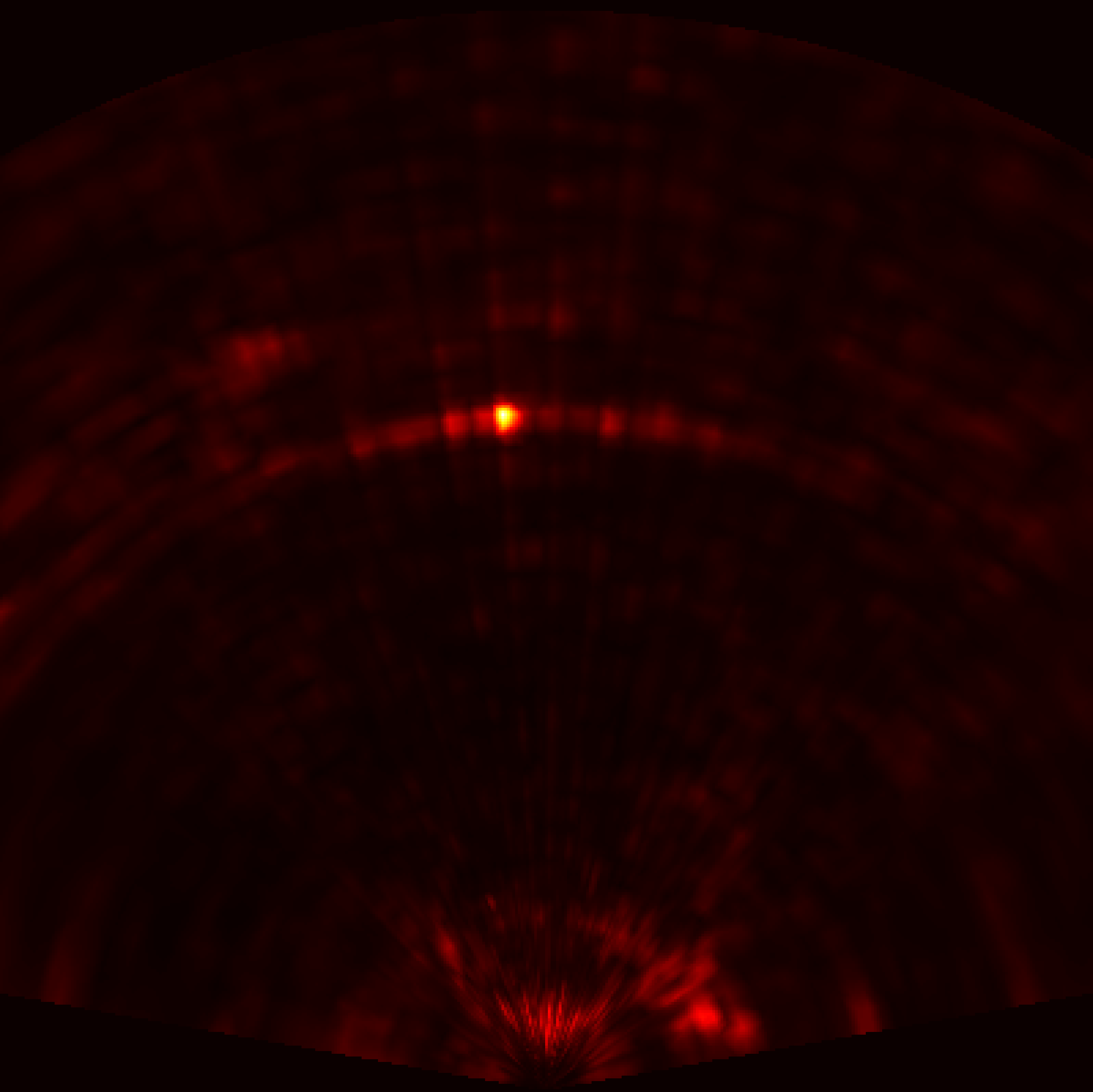} & \includegraphics[width=0.165\linewidth]{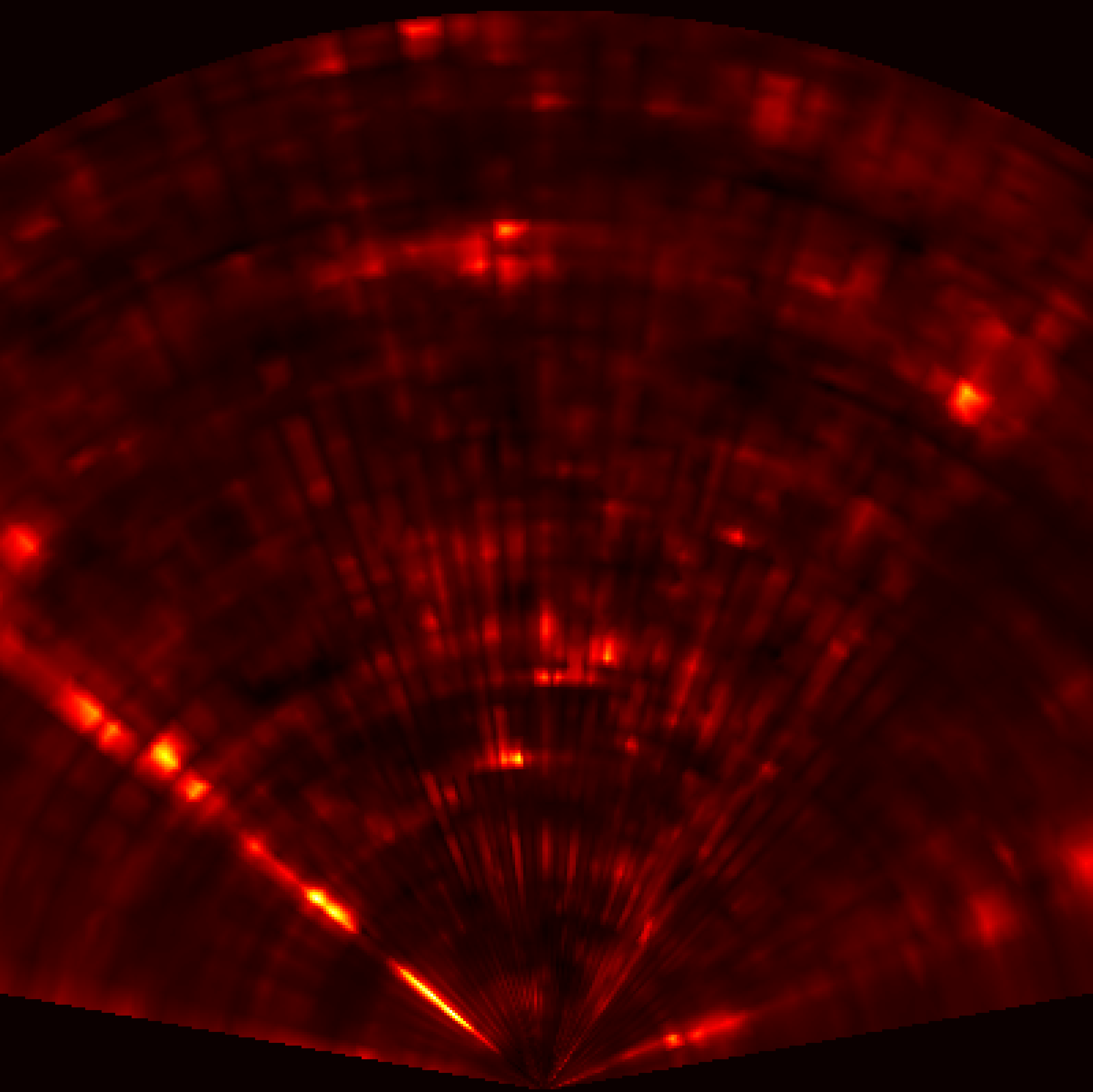} & \includegraphics[width=0.165\linewidth]{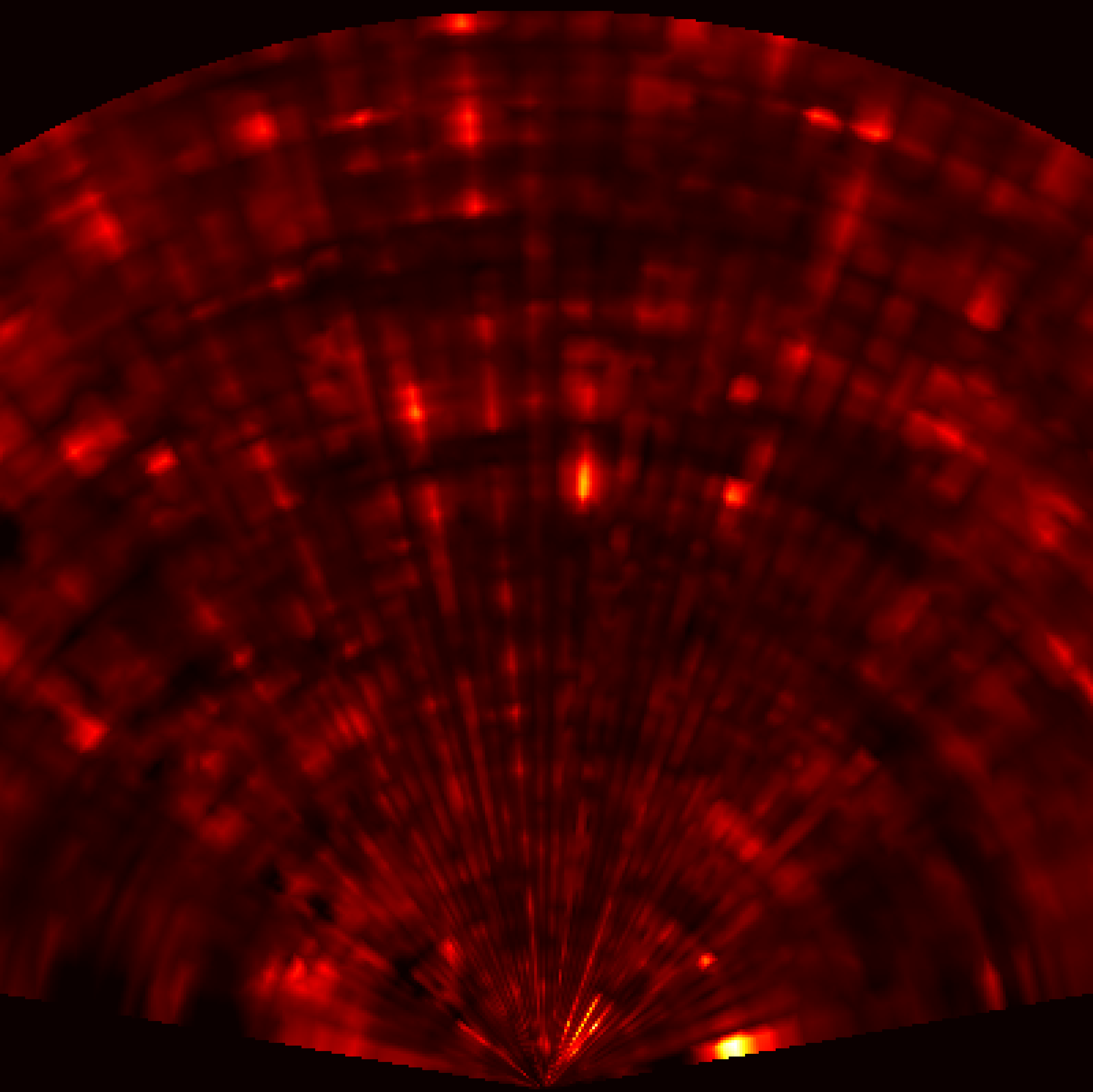} \\
\end{tabular}
\caption{\textbf{Per-frame $|\mathrm{RA}|$ comparison, scene S1\,F185, training frames $F\!+\!1$ to $F\!+\!4$ and held-out test frame $F$ (red header).} Pearson correlation per column, left to right. 3DPS: 0.85, 0.87, 0.85, 0.90, 0.52; mmIR: 0.52, 0.51, 0.54, 0.49, 0.52; RadarSplat: 0.46, 0.22, 0.74, 0.35, 0.25; Radar Fields: 0.21, 0.12, 0.10, 0.12, 0.09; DART: -0.01, 0.05, 0.01, 0.12, 0.04.}
\label{fig:supp_s1_f185_b}
\end{figure}

\begin{figure}[p]
\centering
\setlength{\tabcolsep}{1pt}\renewcommand{\arraystretch}{0.6}
\begin{tabular}{ccccc}
 & {\footnotesize $F\!-\!4$} & {\footnotesize $F\!-\!3$} & {\footnotesize $F\!-\!2$} & {\footnotesize $F\!-\!1$} \\
\raisebox{2.0em}{\rotatebox{90}{\scriptsize\textbf{GT}}} & \includegraphics[width=0.165\linewidth]{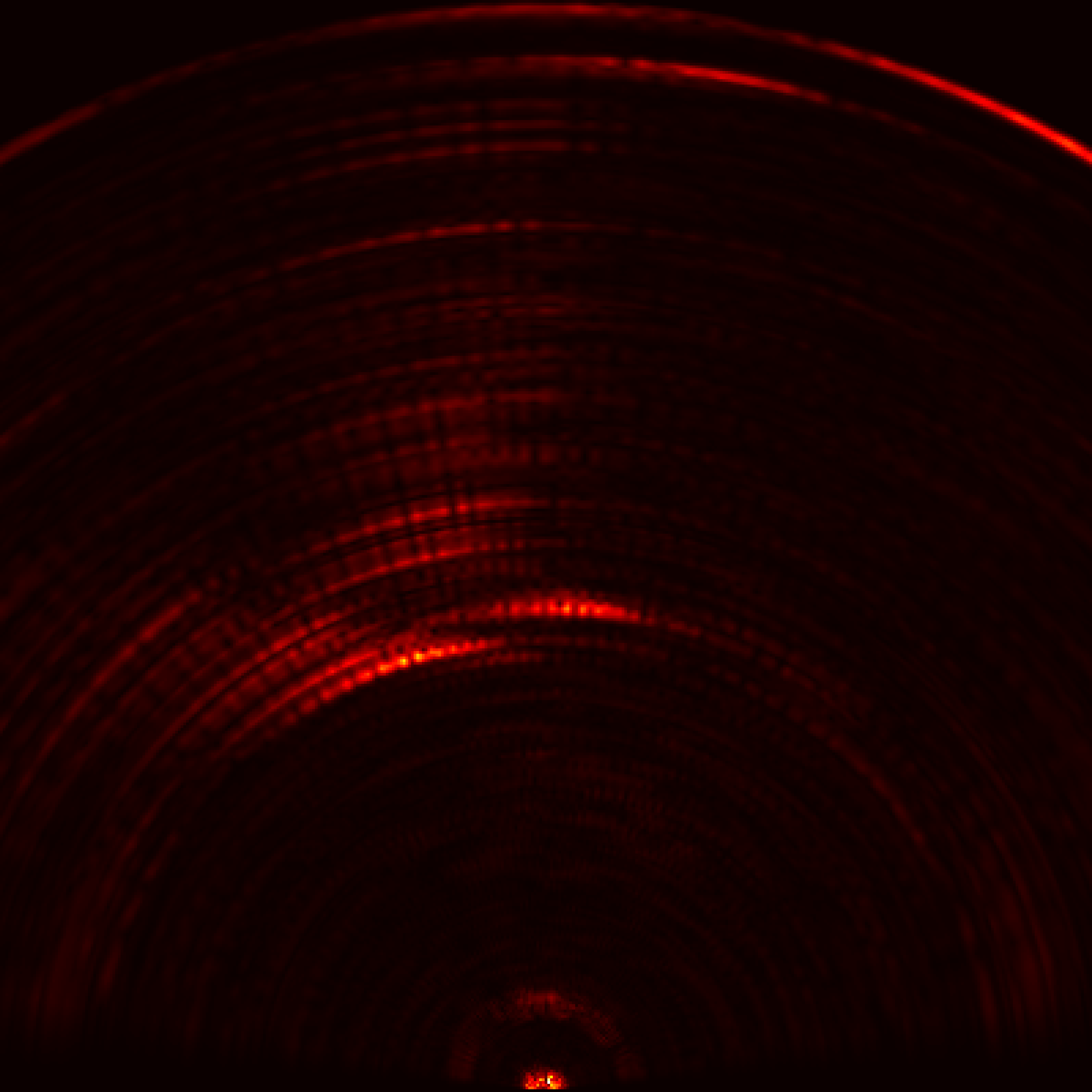} & \includegraphics[width=0.165\linewidth]{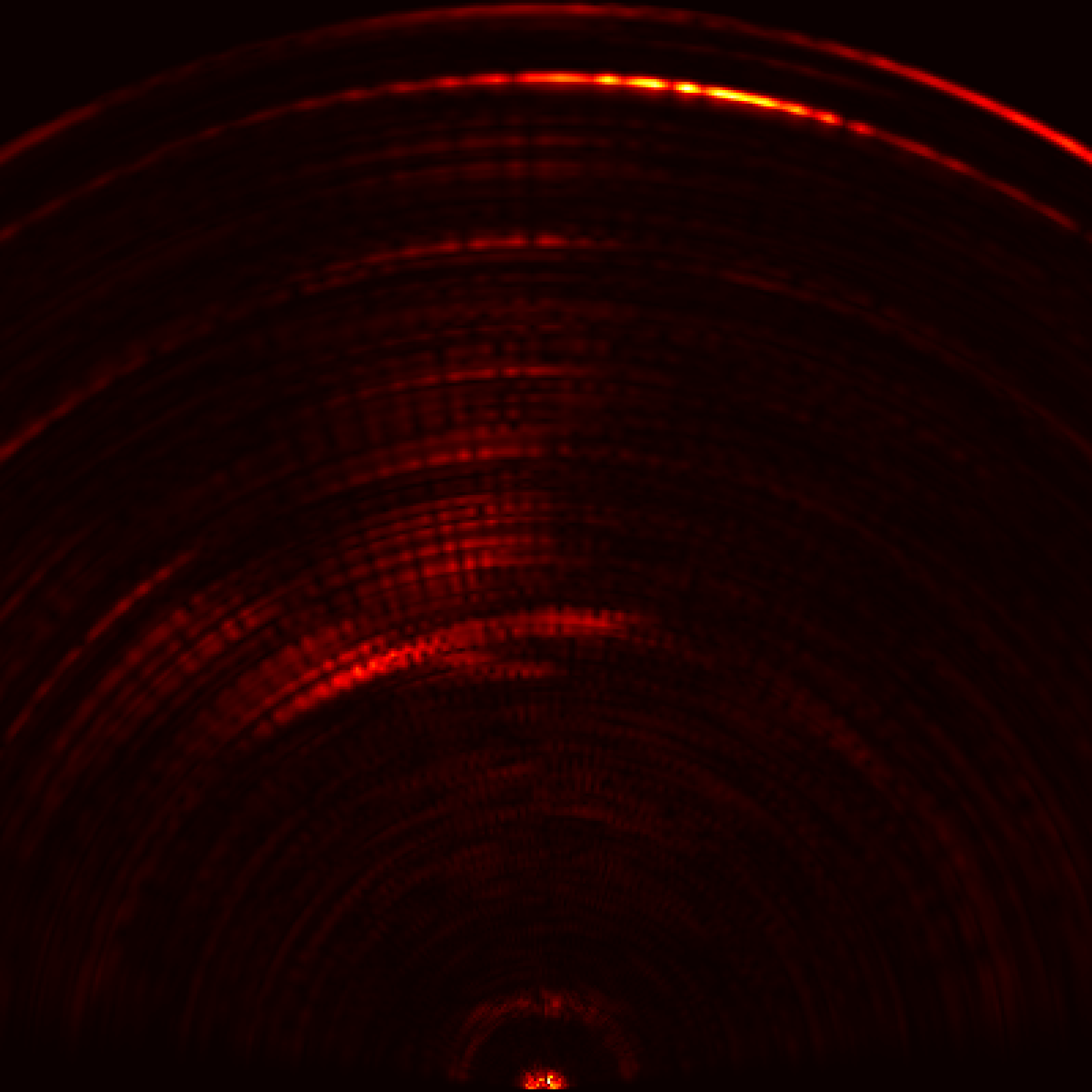} & \includegraphics[width=0.165\linewidth]{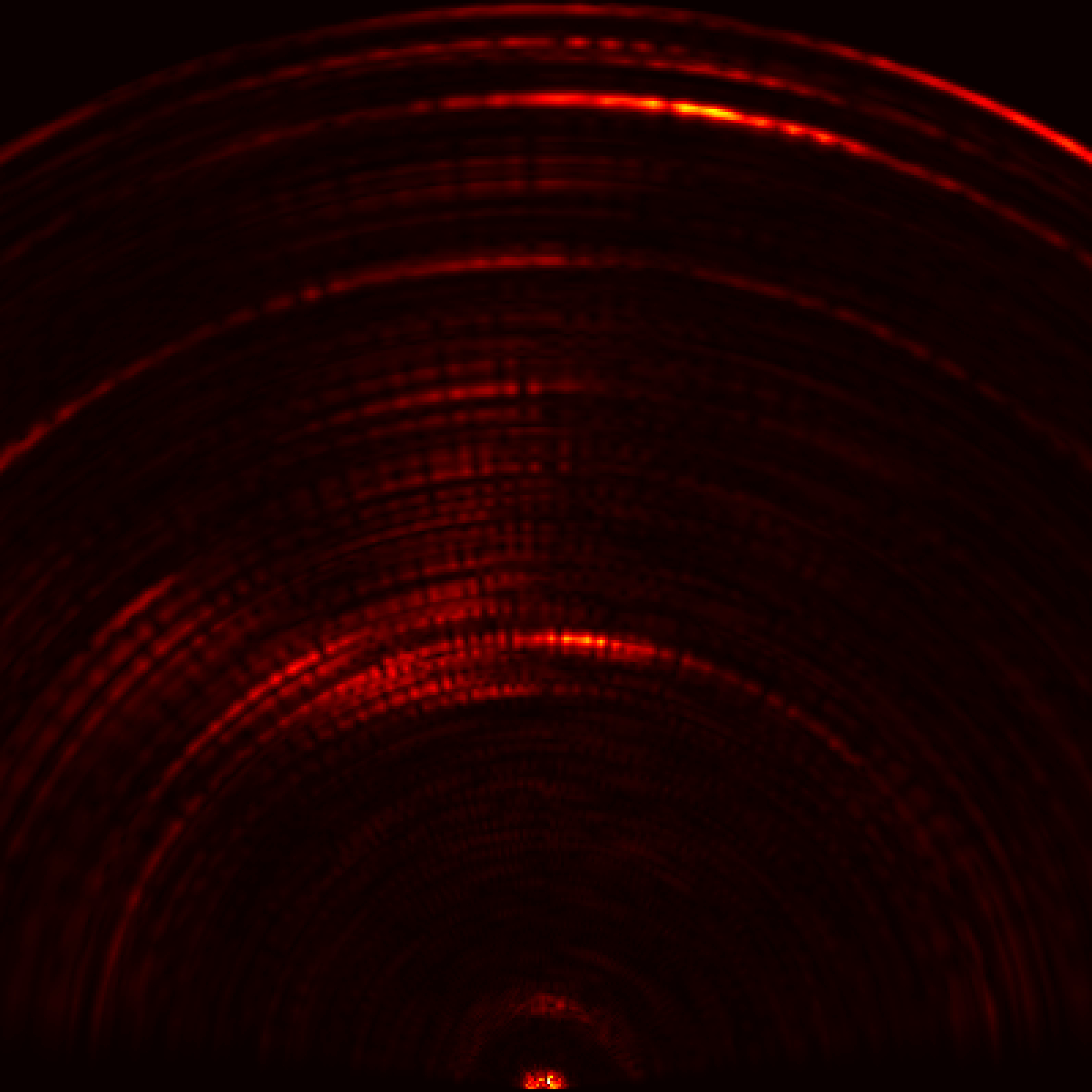} & \includegraphics[width=0.165\linewidth]{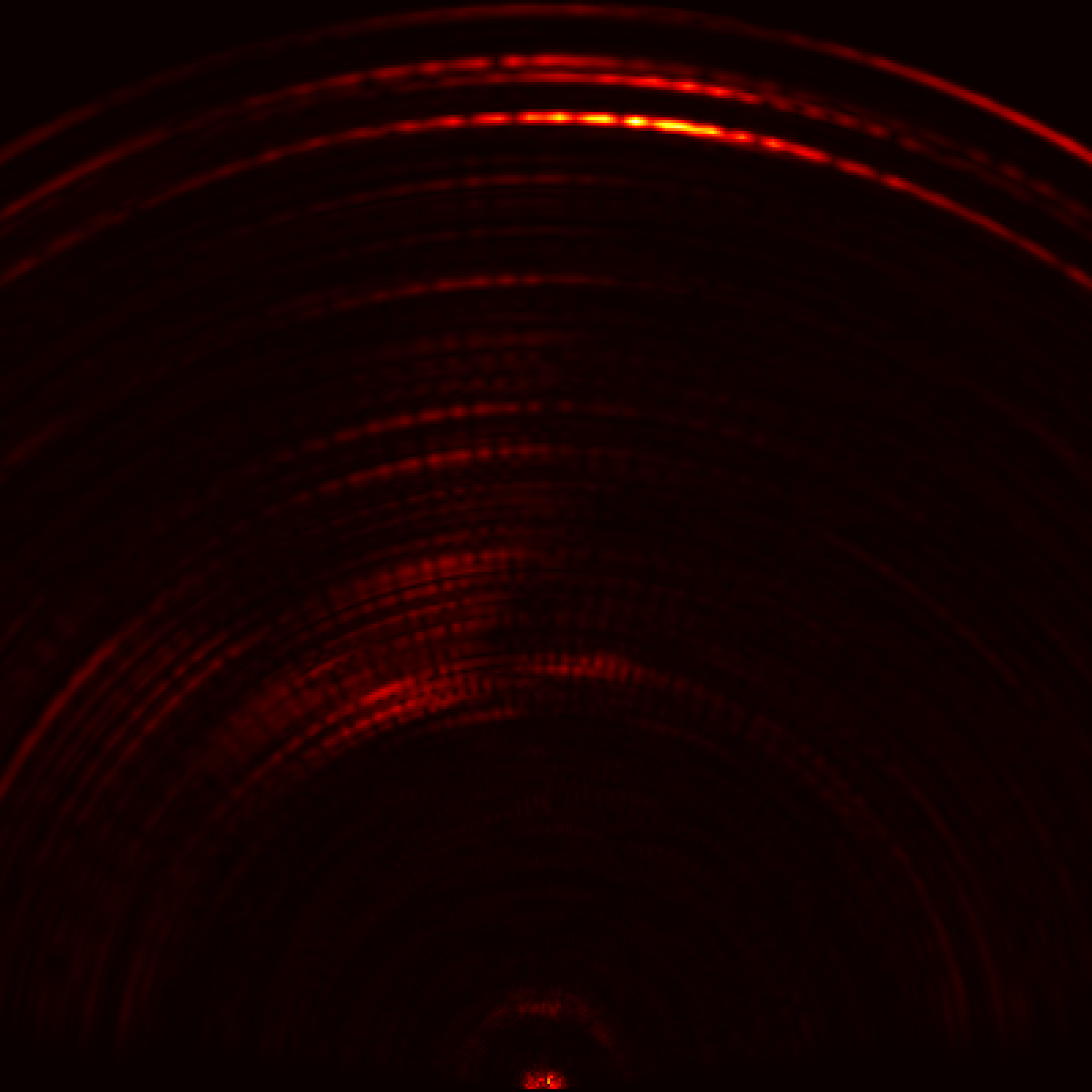} \\
\raisebox{2.0em}{\rotatebox{90}{\scriptsize\textbf{3DPS (Ours)}}} & \includegraphics[width=0.165\linewidth]{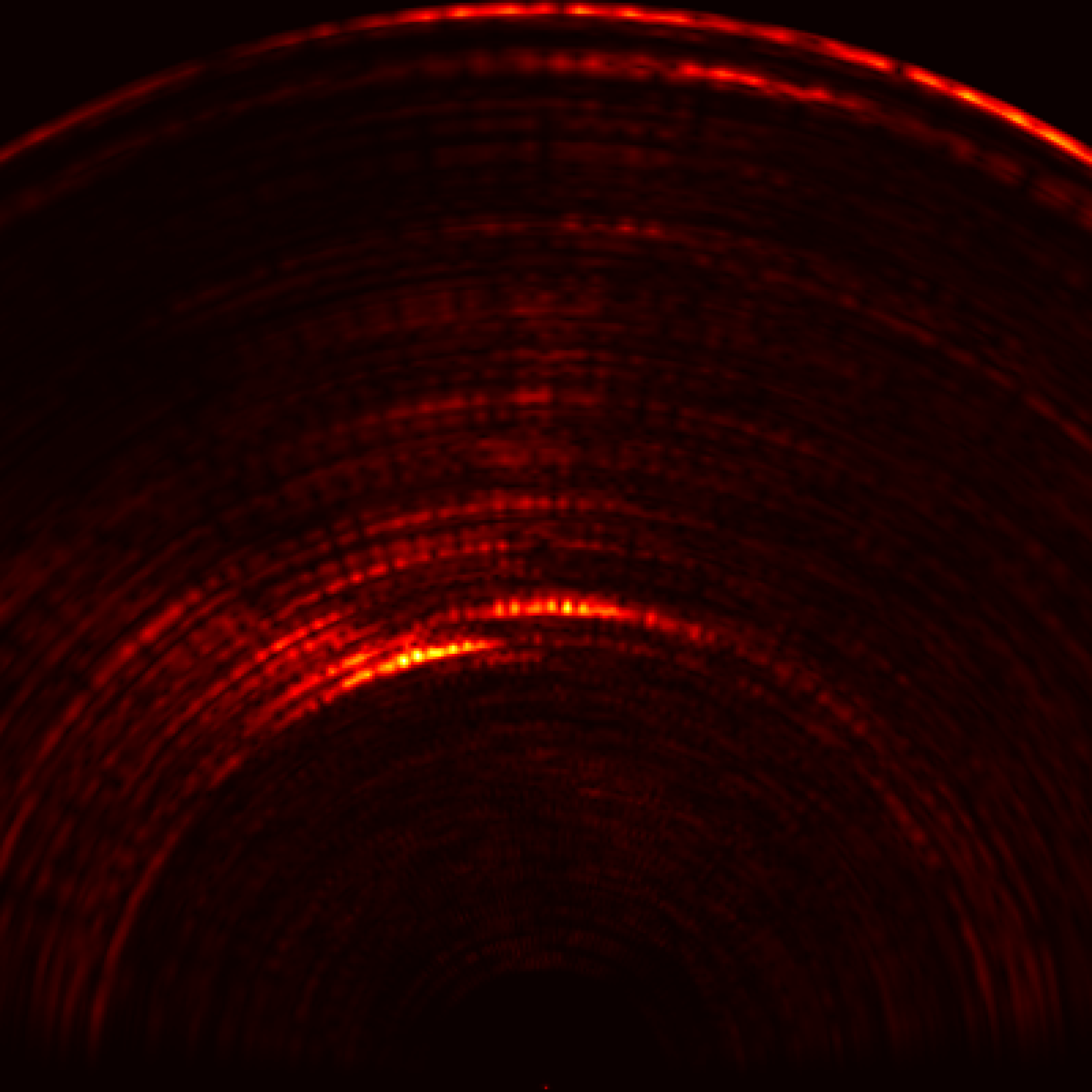} & \includegraphics[width=0.165\linewidth]{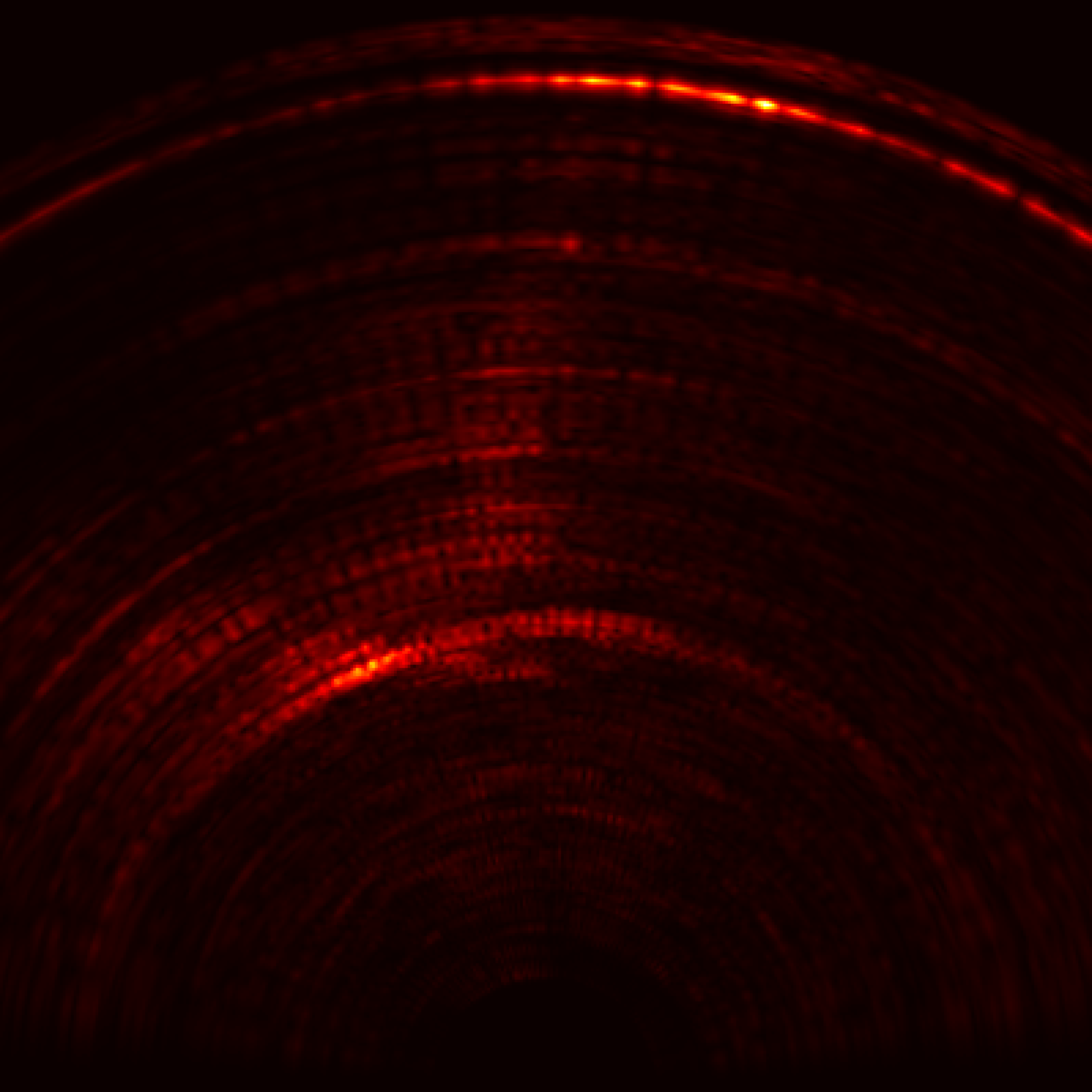} & \includegraphics[width=0.165\linewidth]{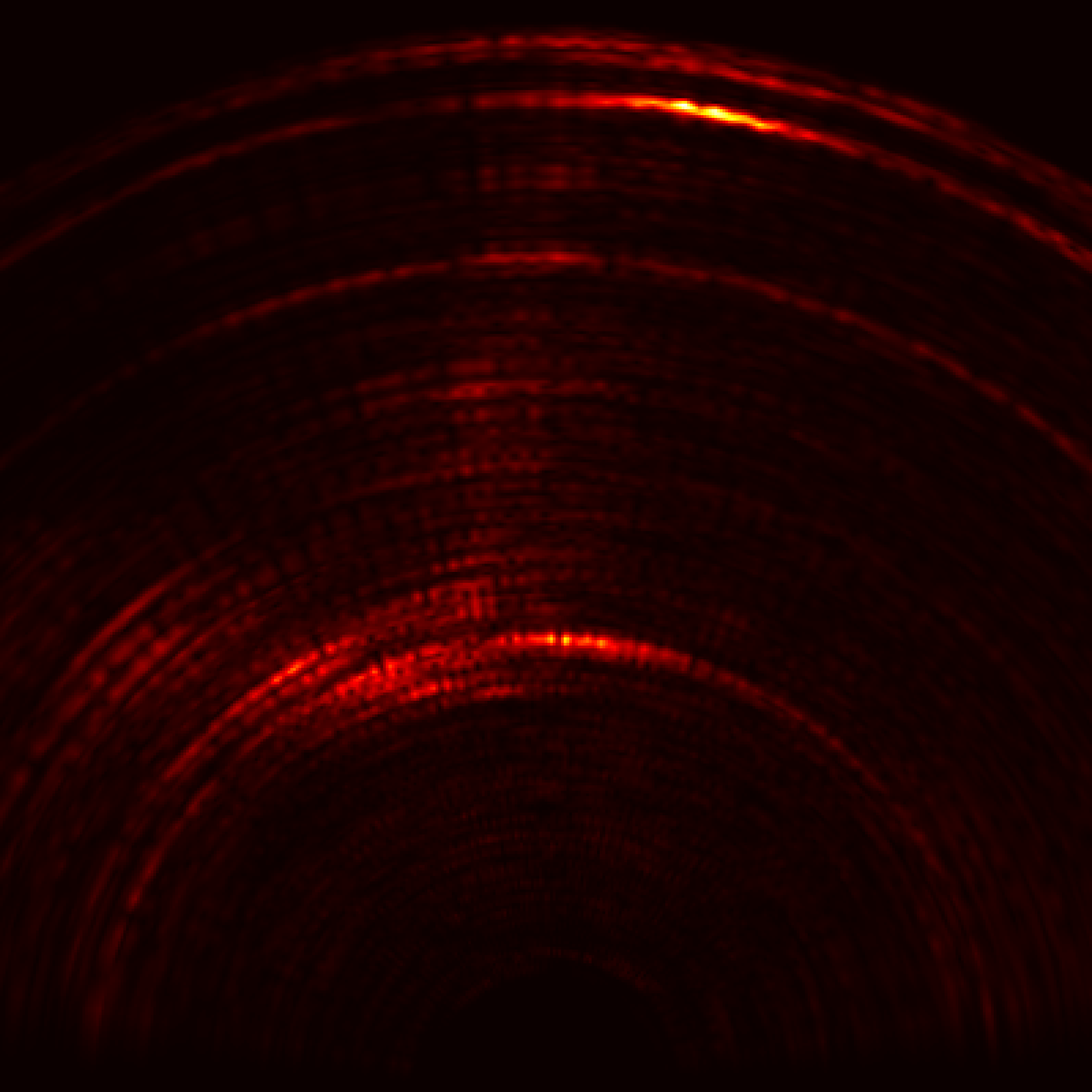} & \includegraphics[width=0.165\linewidth]{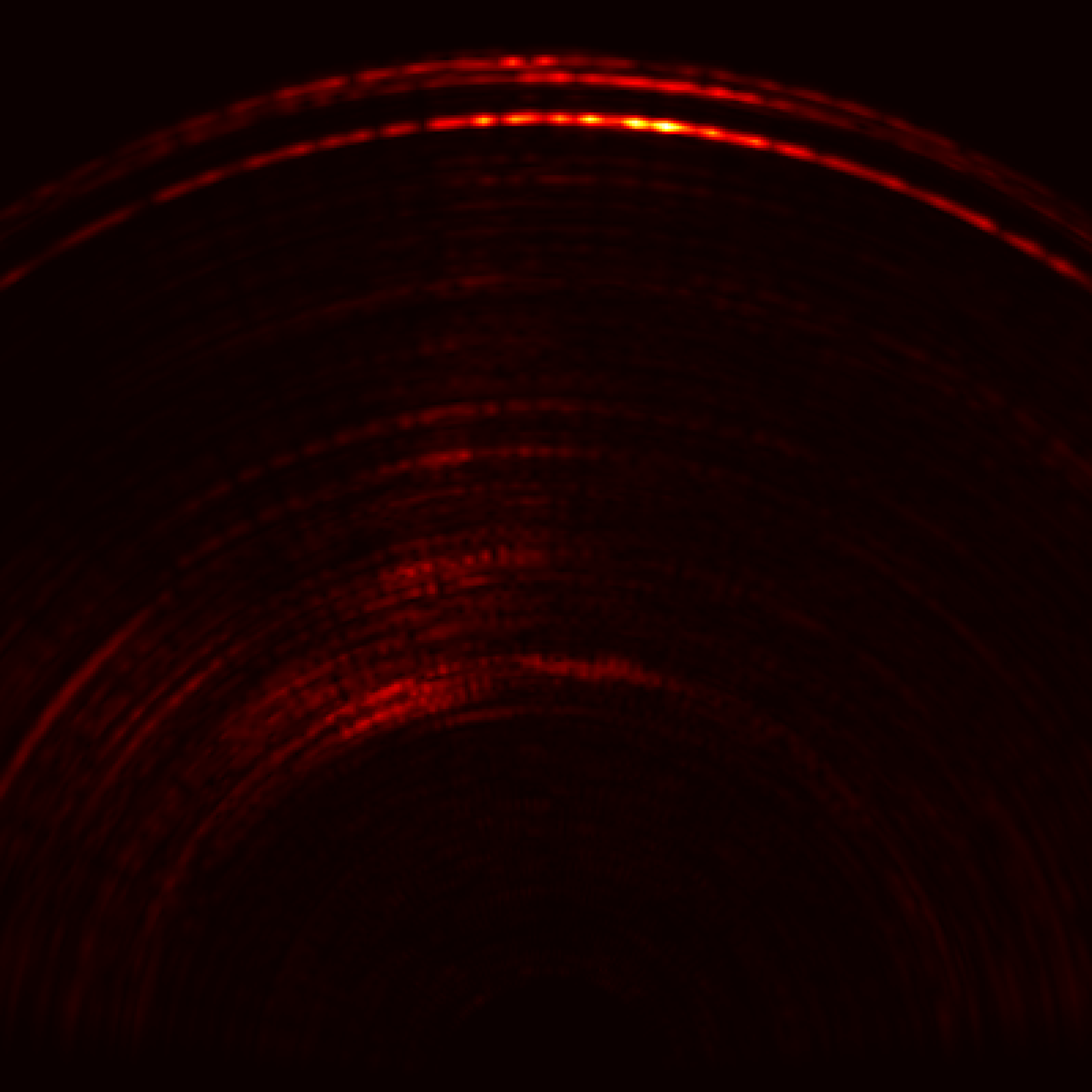} \\
\raisebox{2.0em}{\rotatebox{90}{\scriptsize\textbf{mmIR}}} & \includegraphics[width=0.165\linewidth]{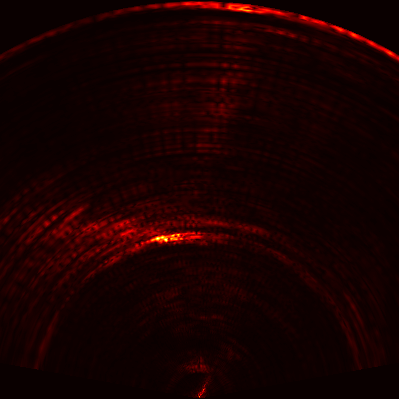} & \includegraphics[width=0.165\linewidth]{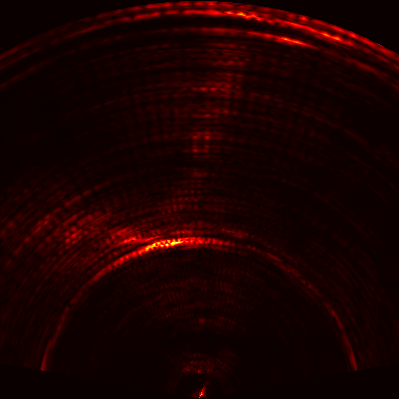} & \includegraphics[width=0.165\linewidth]{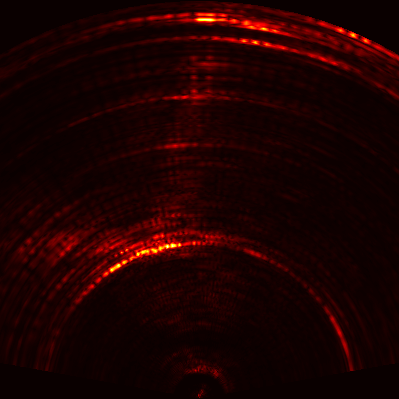} & \includegraphics[width=0.165\linewidth]{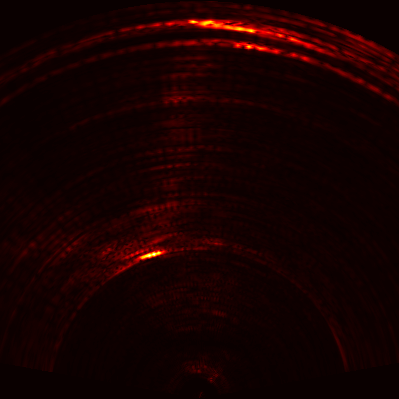} \\
\raisebox{2.0em}{\rotatebox{90}{\scriptsize\textbf{RadarSplat}}} & \includegraphics[width=0.165\linewidth]{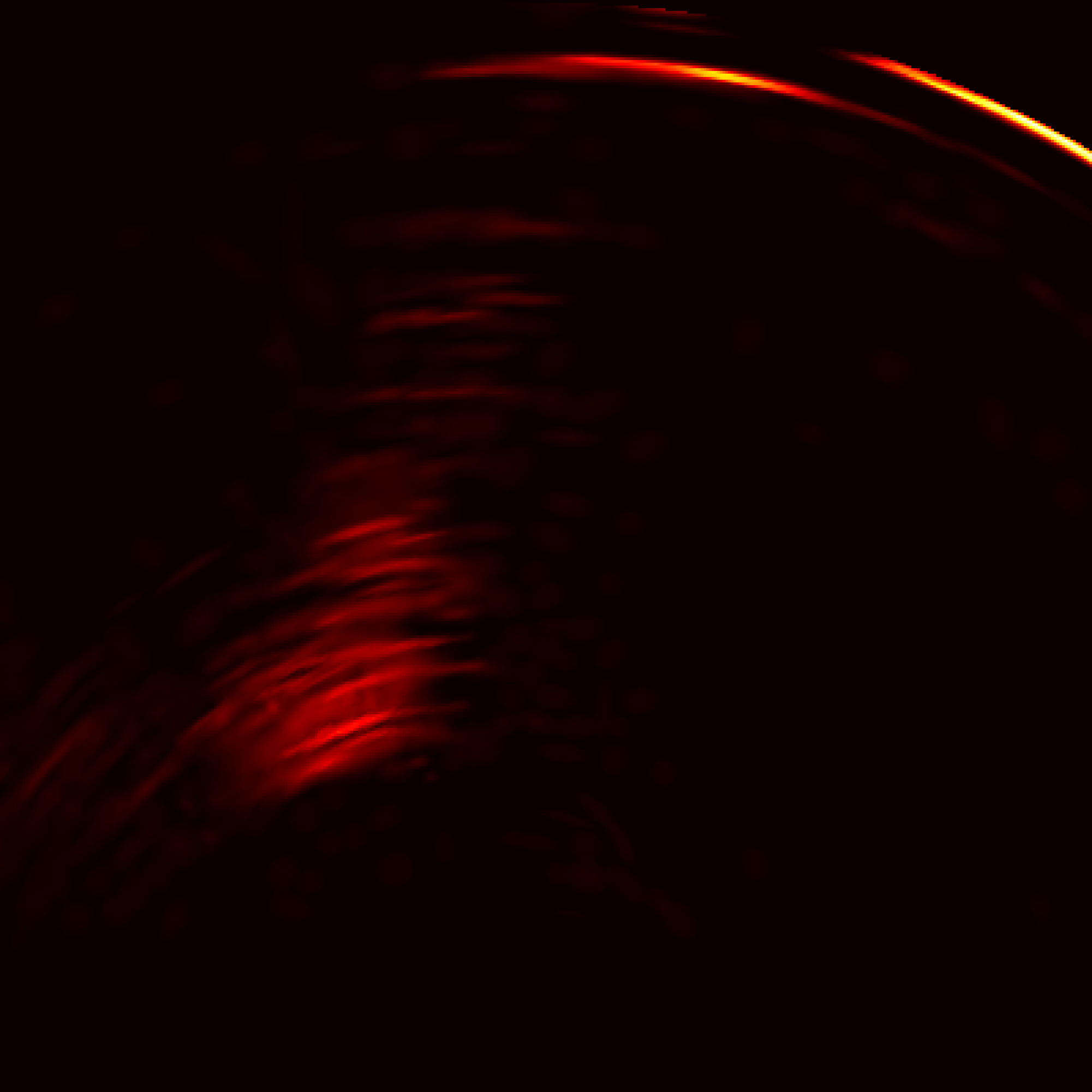} & \includegraphics[width=0.165\linewidth]{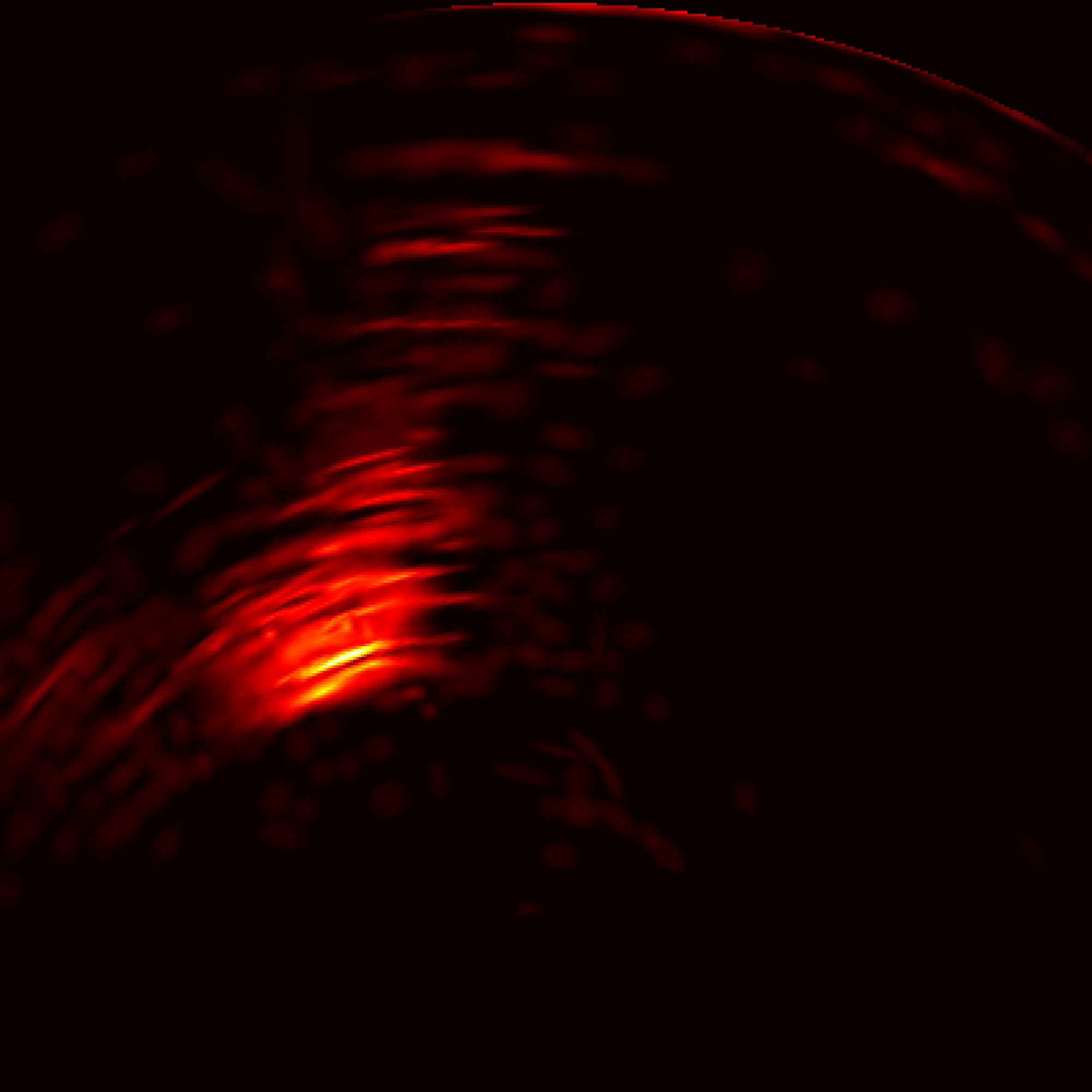} & \includegraphics[width=0.165\linewidth]{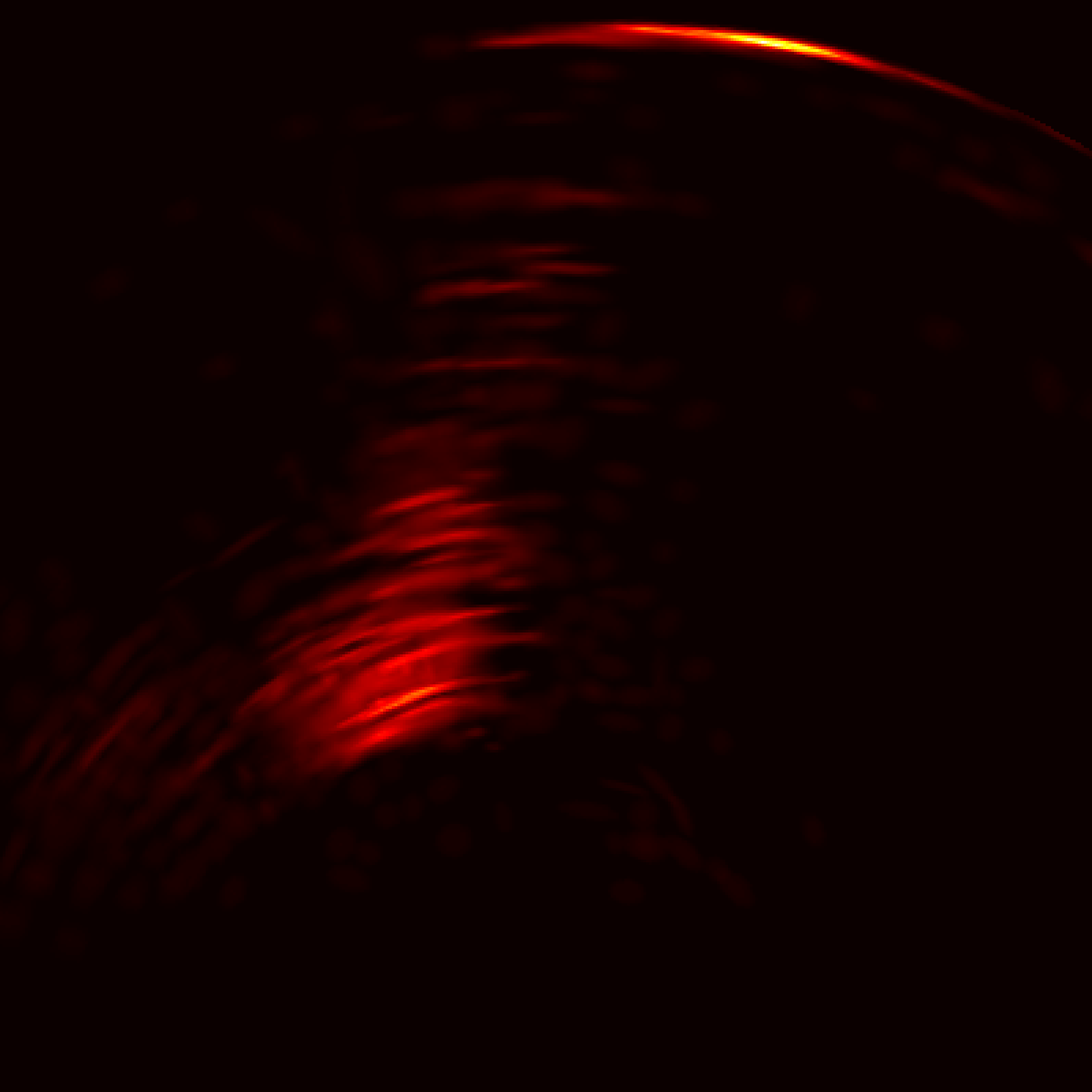} & \includegraphics[width=0.165\linewidth]{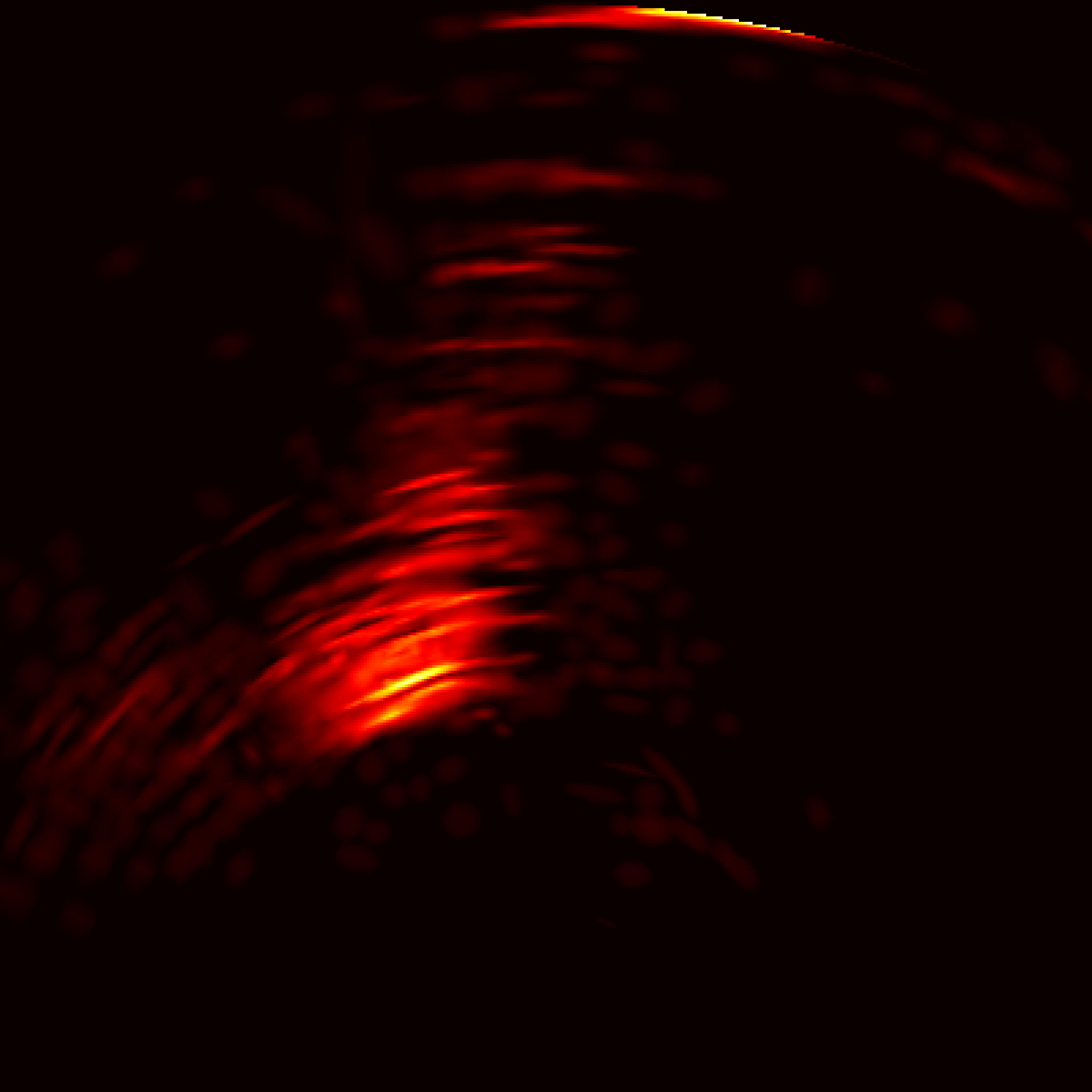} \\
\raisebox{2.0em}{\rotatebox{90}{\scriptsize\textbf{Radar Fields}}} & \includegraphics[width=0.165\linewidth]{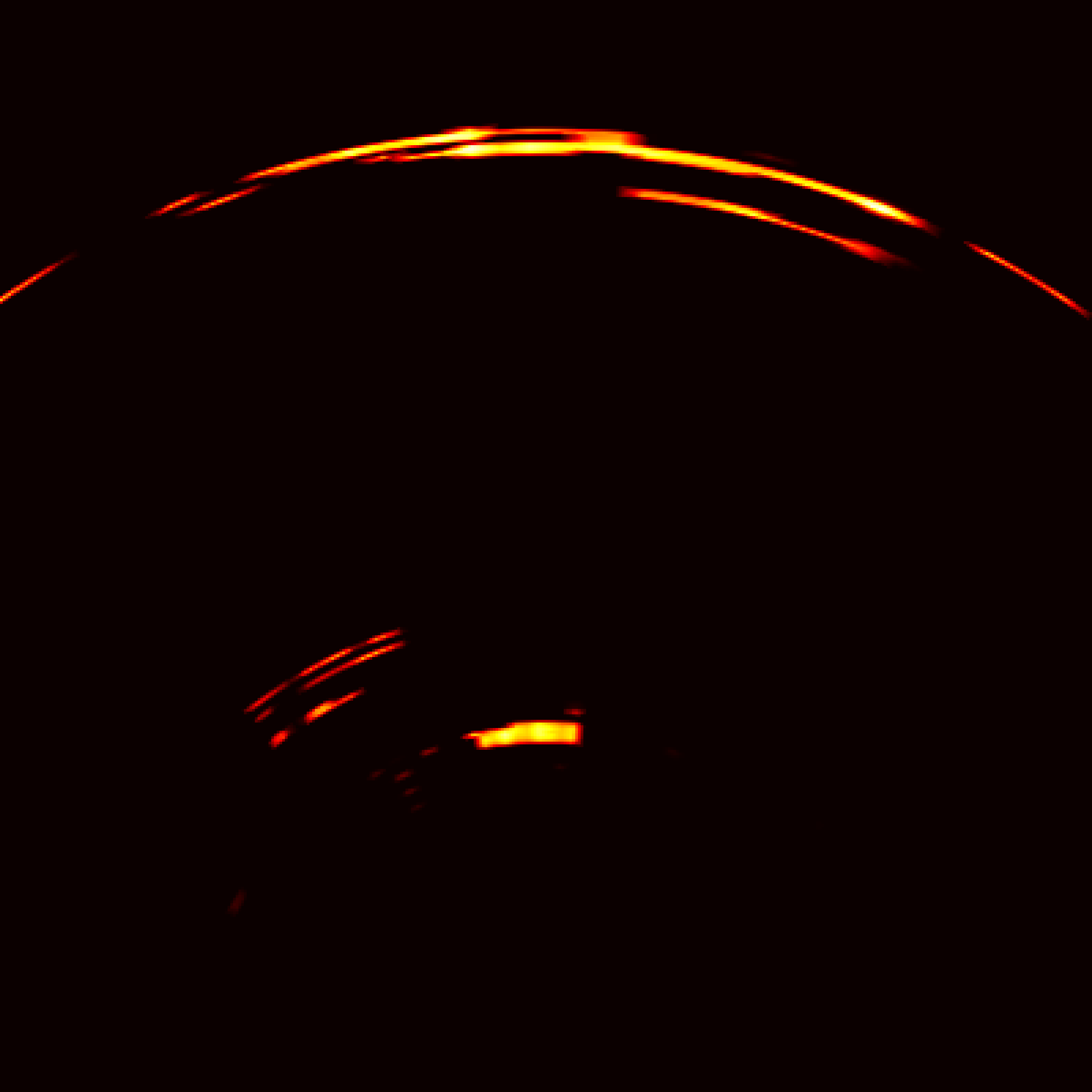} & \includegraphics[width=0.165\linewidth]{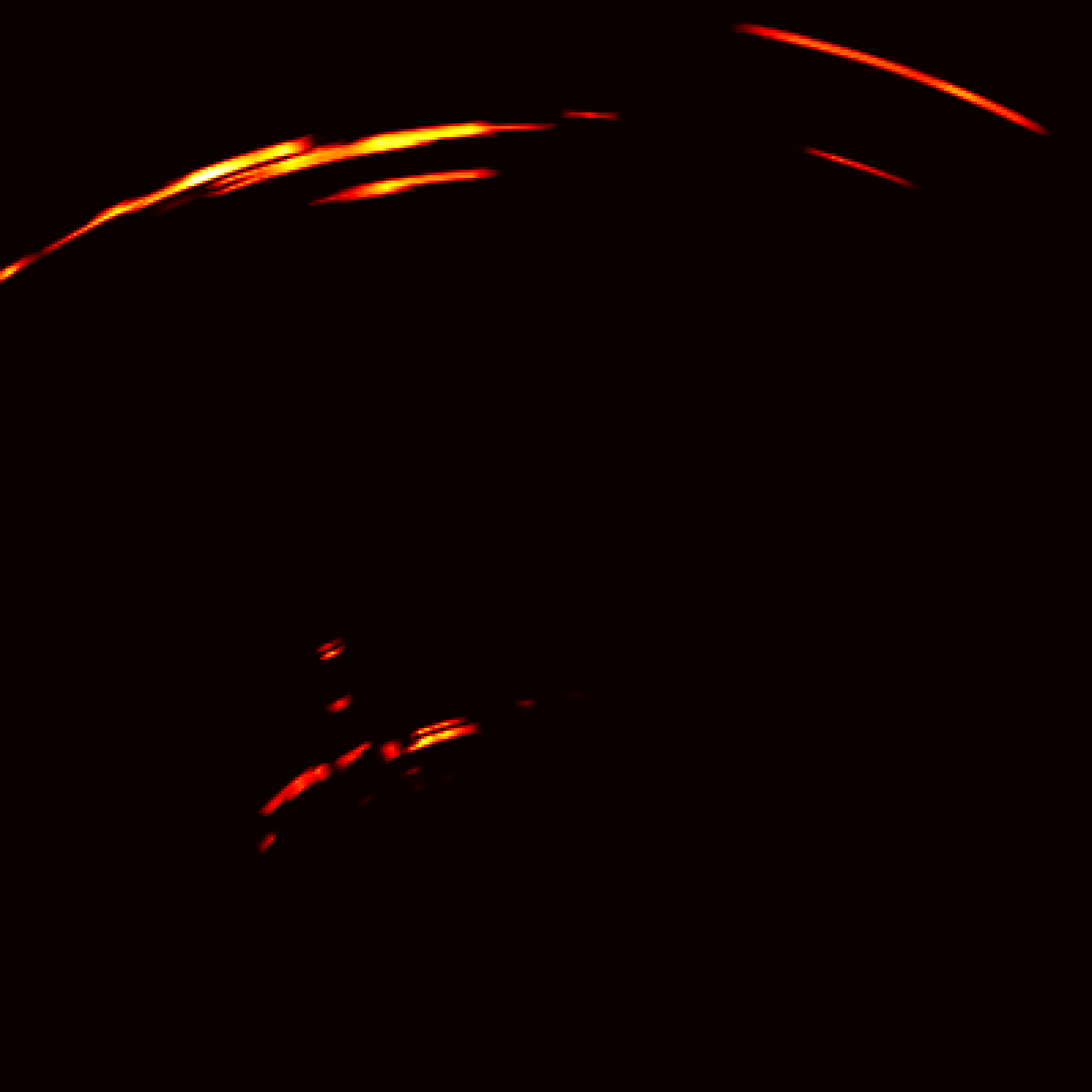} & \includegraphics[width=0.165\linewidth]{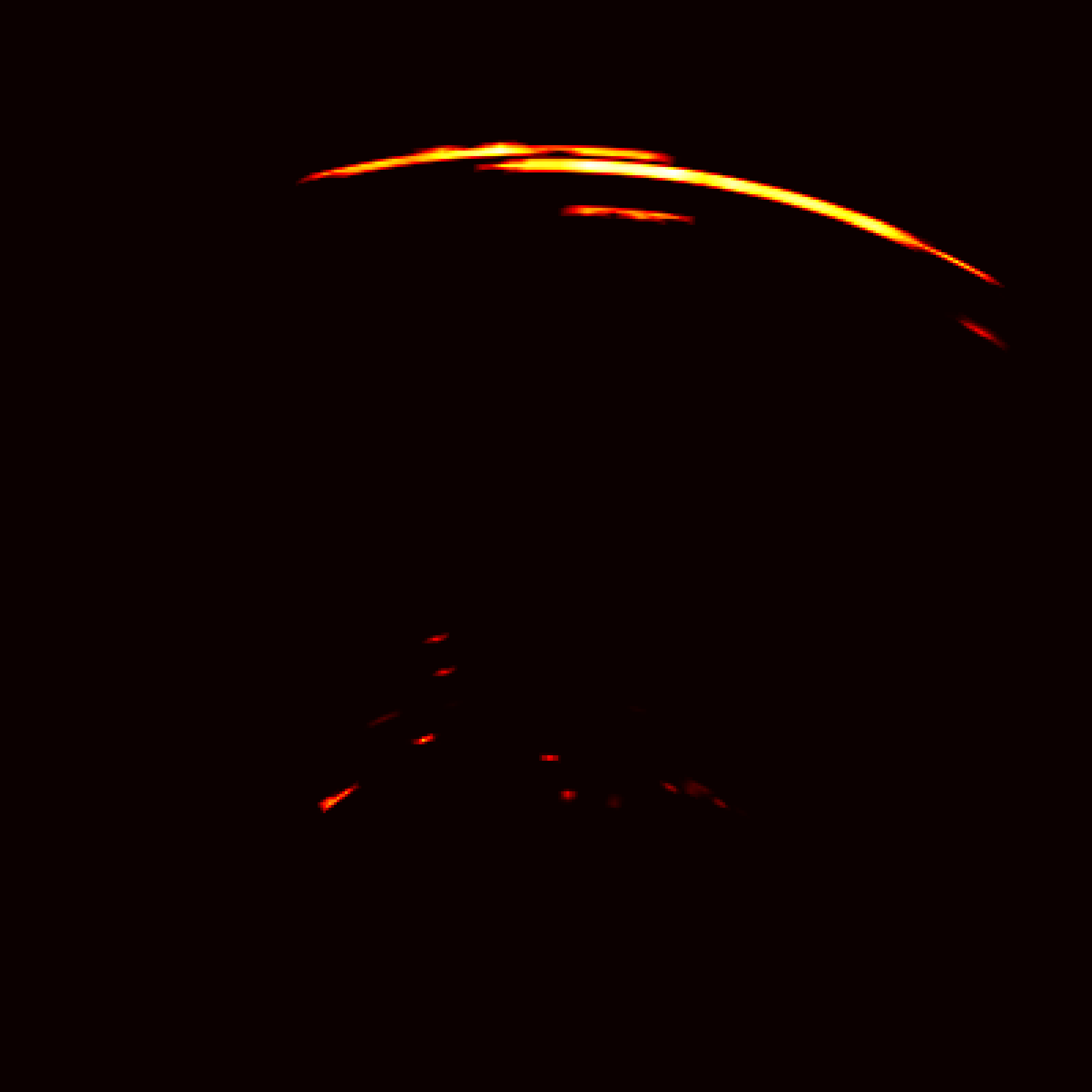} & \includegraphics[width=0.165\linewidth]{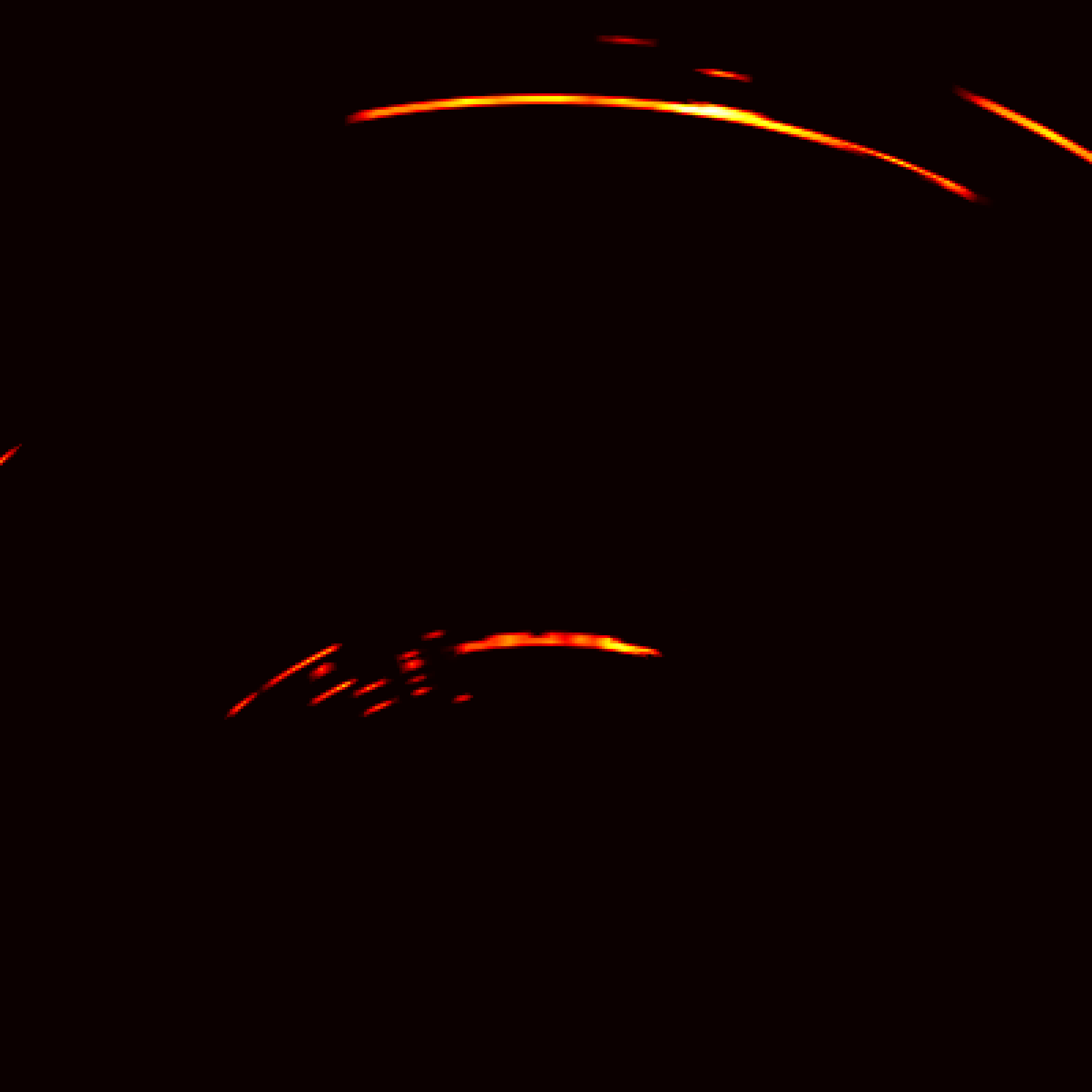} \\
\raisebox{2.0em}{\rotatebox{90}{\scriptsize\textbf{DART}}} & \includegraphics[width=0.165\linewidth]{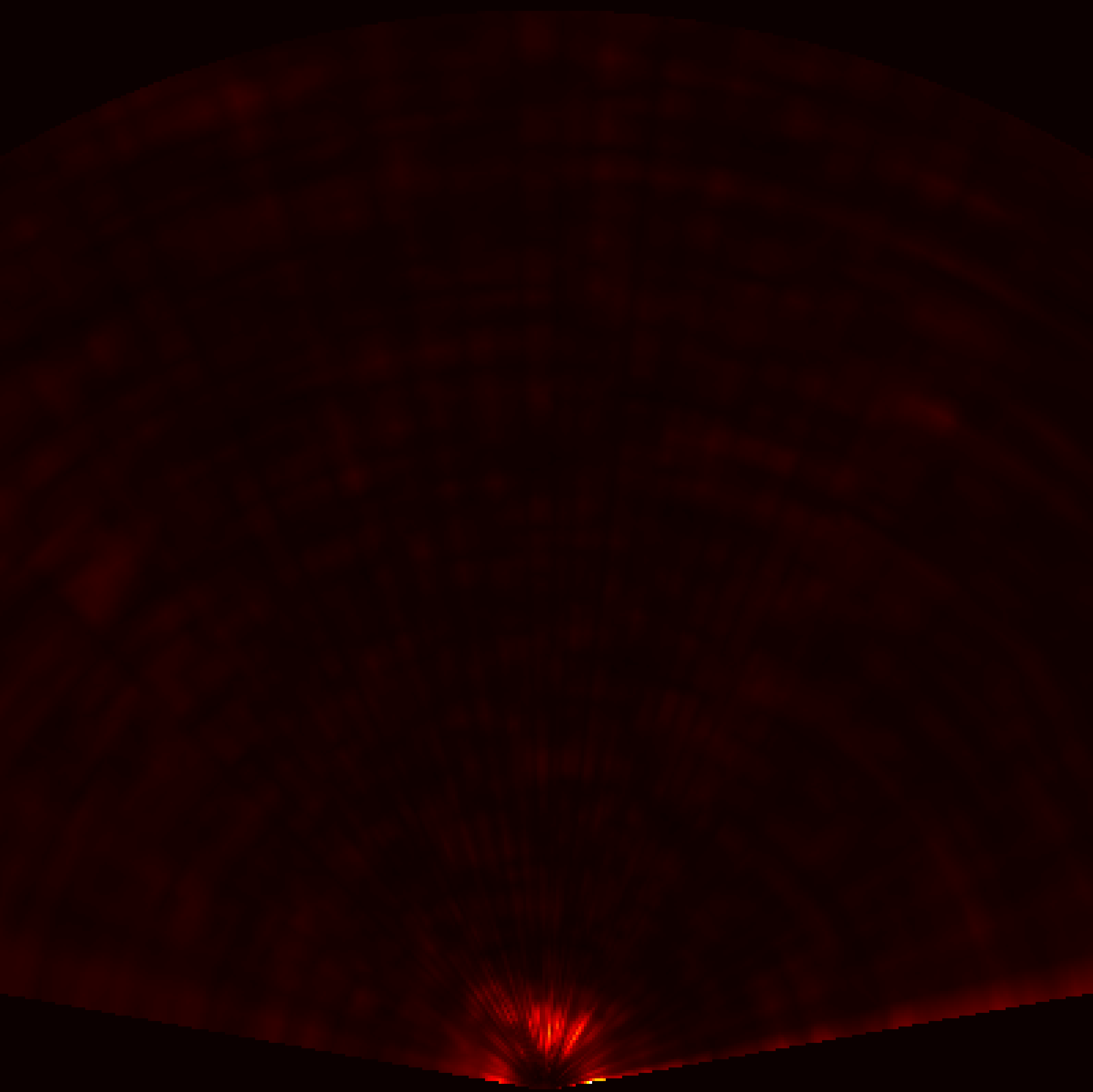} & \includegraphics[width=0.165\linewidth]{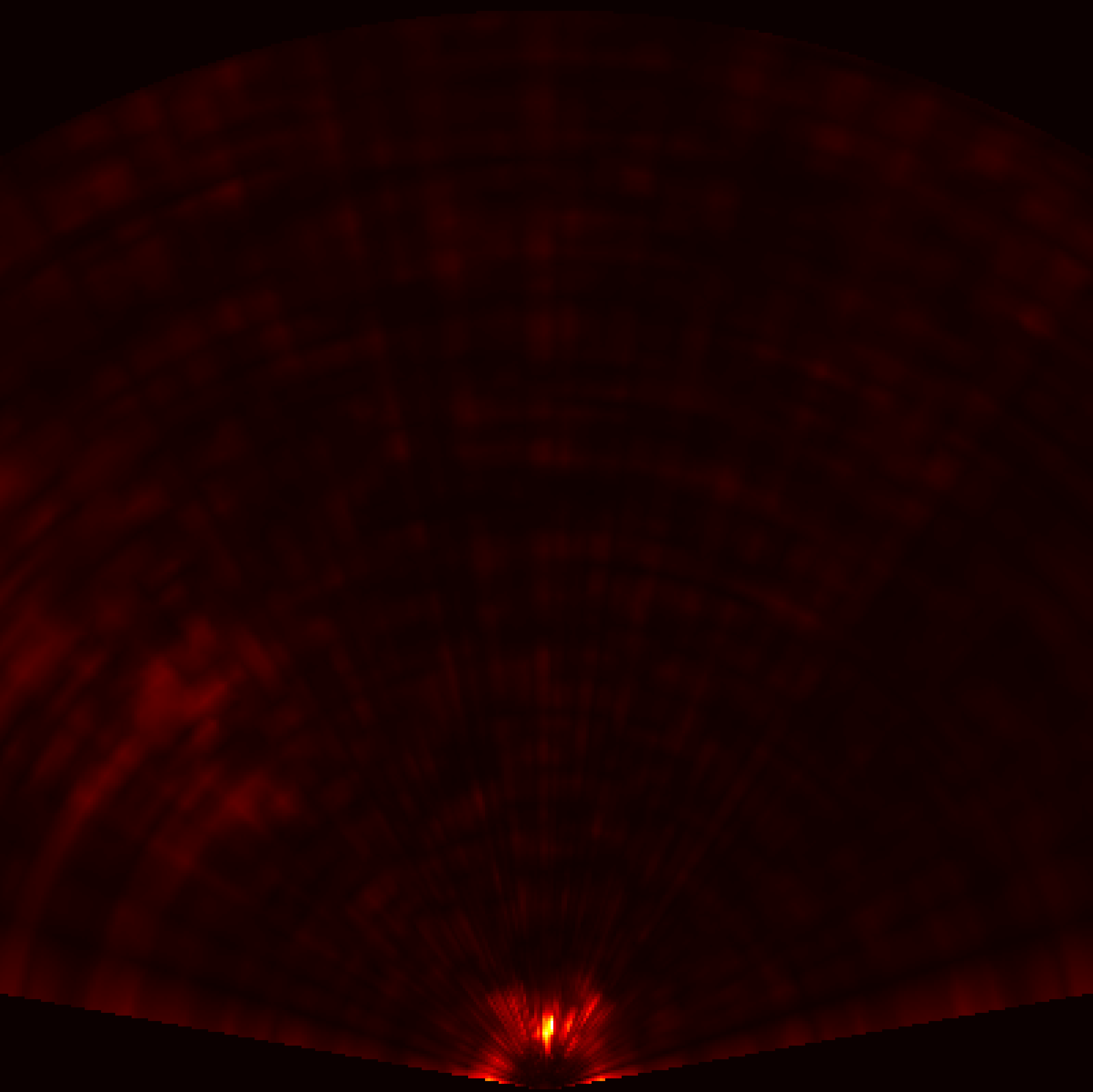} & \includegraphics[width=0.165\linewidth]{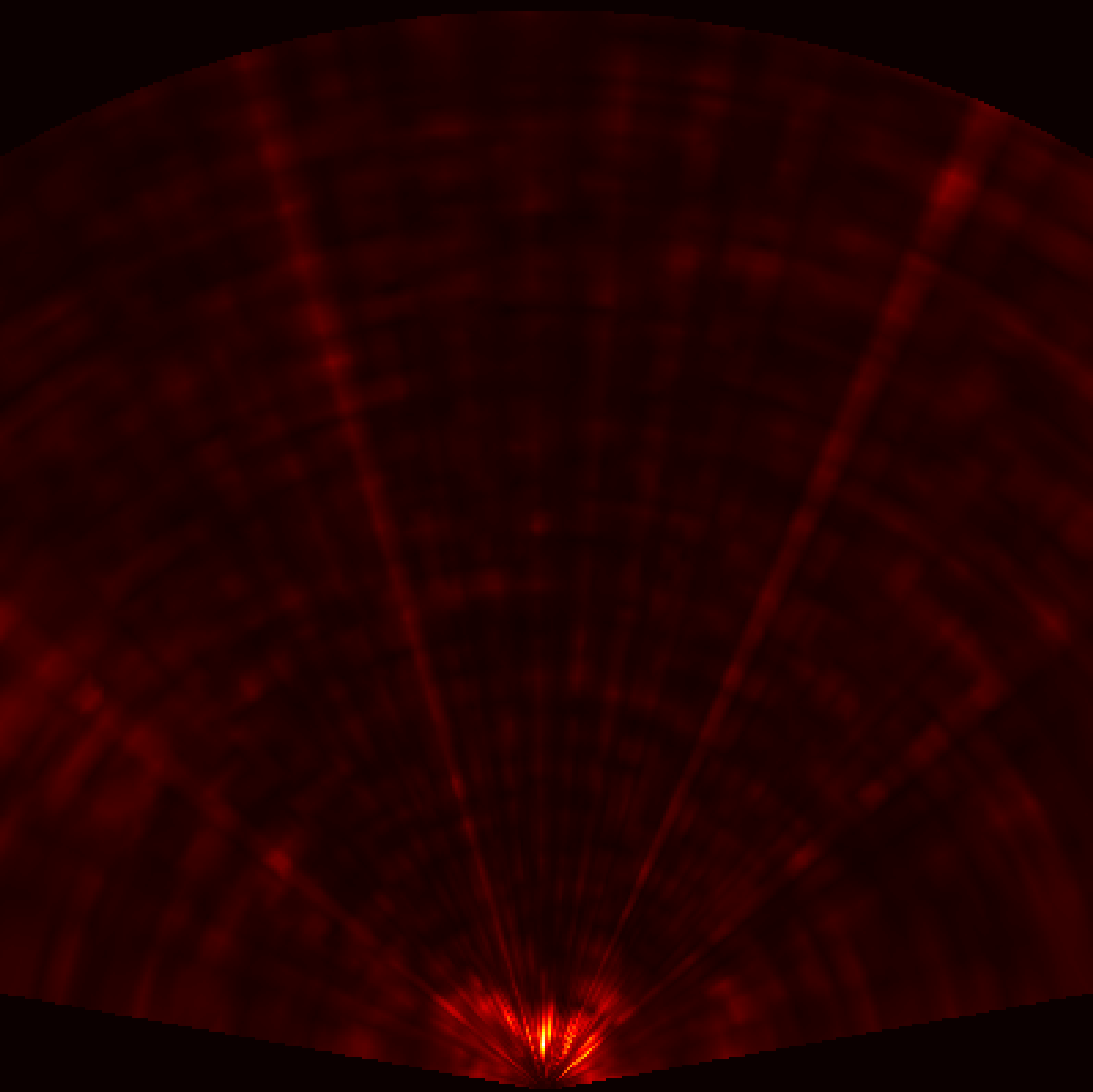} & \includegraphics[width=0.165\linewidth]{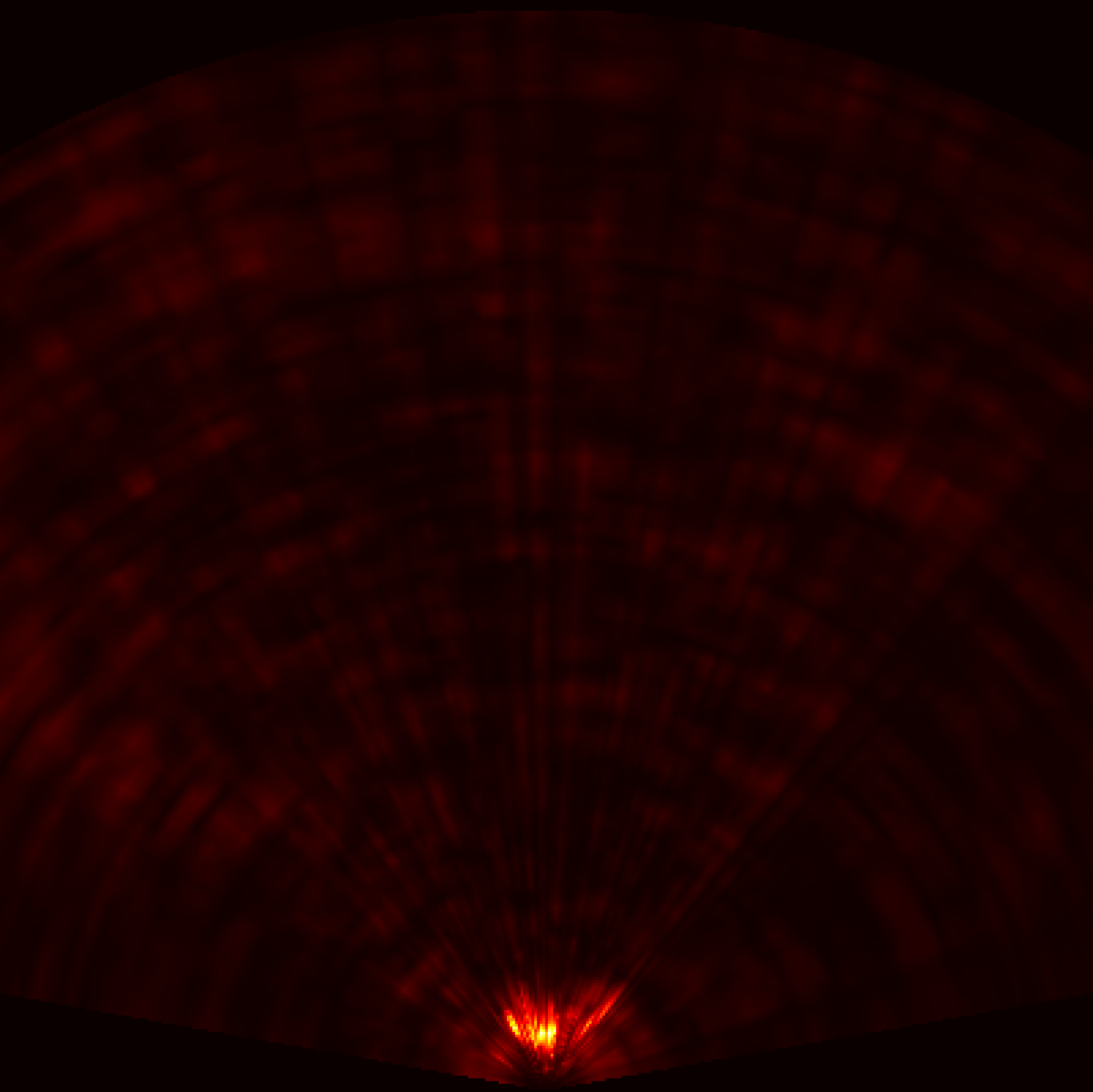} \\
\end{tabular}
\caption{\textbf{Per-frame $|\mathrm{RA}|$ comparison, scene S1\,F438, training frames $F\!-\!4$ to $F\!-\!1$.} Pearson correlation per column, left to right. 3DPS: 0.85, 0.75, 0.80, 0.86; mmIR: 0.62, 0.47, 0.55, 0.64; RadarSplat: 0.57, 0.35, 0.34, 0.27; Radar Fields: 0.02, 0.06, -0.01, 0.07; DART: 0.03, 0.03, 0.04, 0.02.}
\label{fig:supp_s1_f438}
\end{figure}

\begin{figure}[p]
\centering
\setlength{\tabcolsep}{1pt}\renewcommand{\arraystretch}{0.6}
\begin{tabular}{cccccc}
 & {\footnotesize $F\!+\!1$} & {\footnotesize $F\!+\!2$} & {\footnotesize $F\!+\!3$} & {\footnotesize $F\!+\!4$} & \textcolor{red}{\footnotesize\textbf{$F$}} \\
\raisebox{2.0em}{\rotatebox{90}{\scriptsize\textbf{GT}}} & \includegraphics[width=0.165\linewidth]{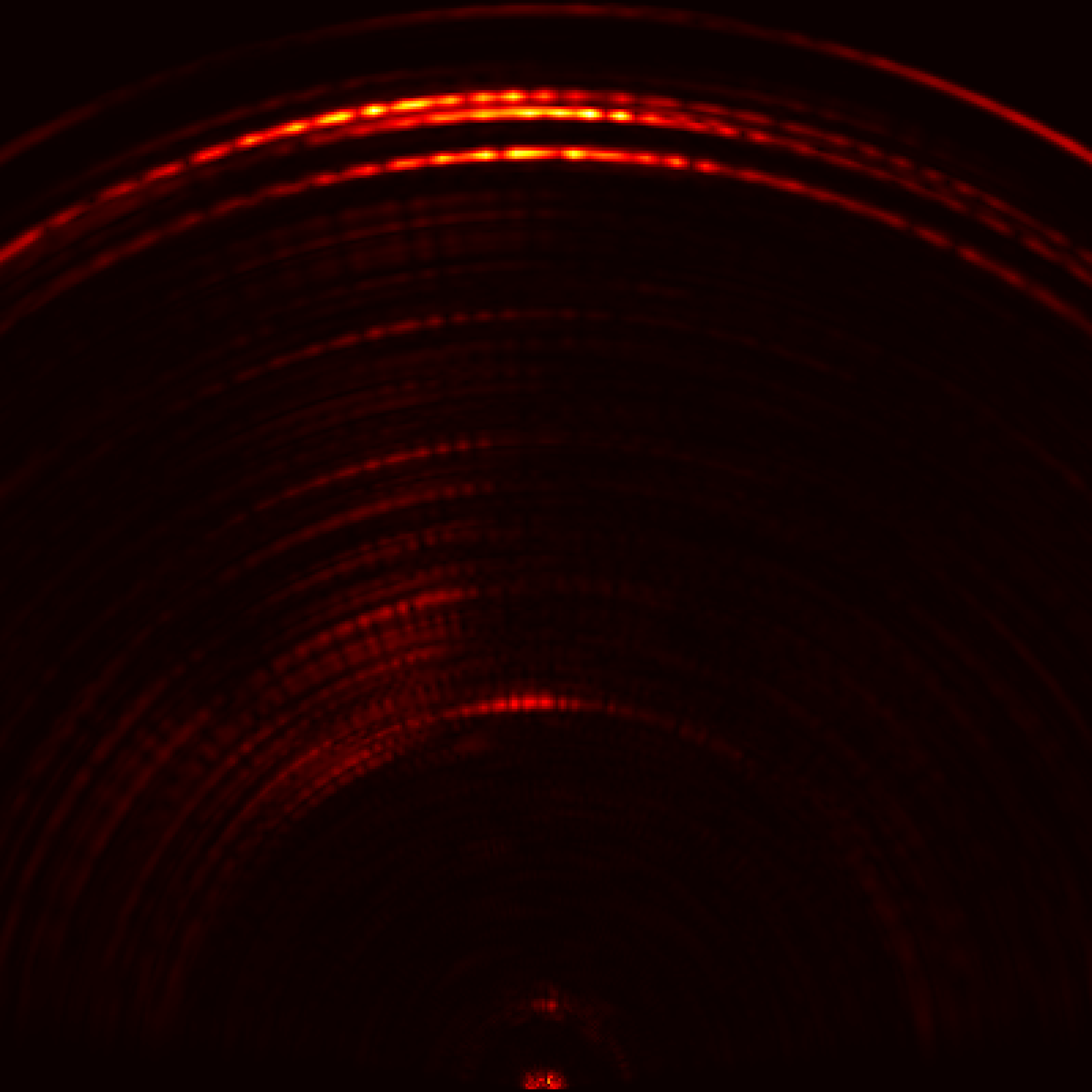} & \includegraphics[width=0.165\linewidth]{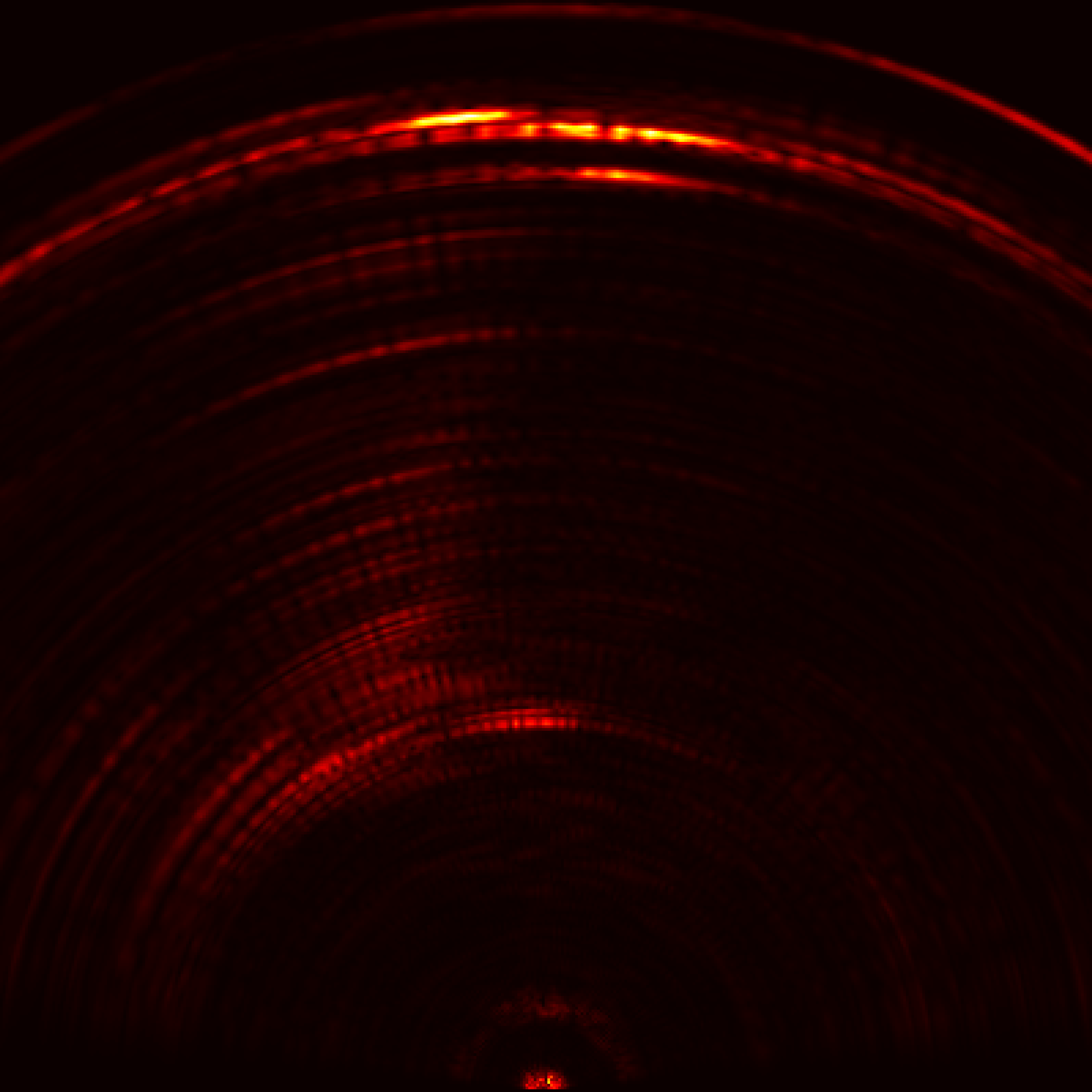} & \includegraphics[width=0.165\linewidth]{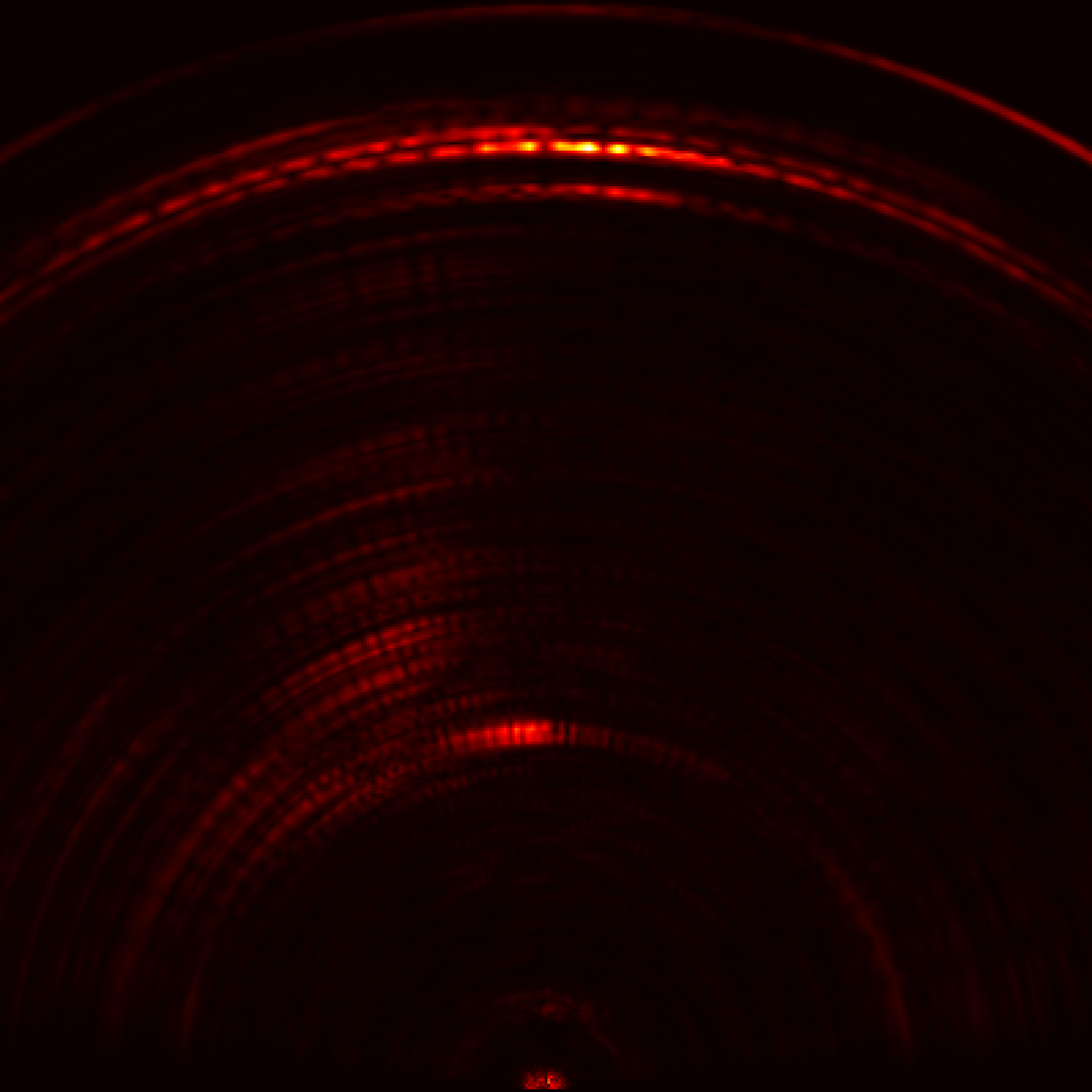} & \includegraphics[width=0.165\linewidth]{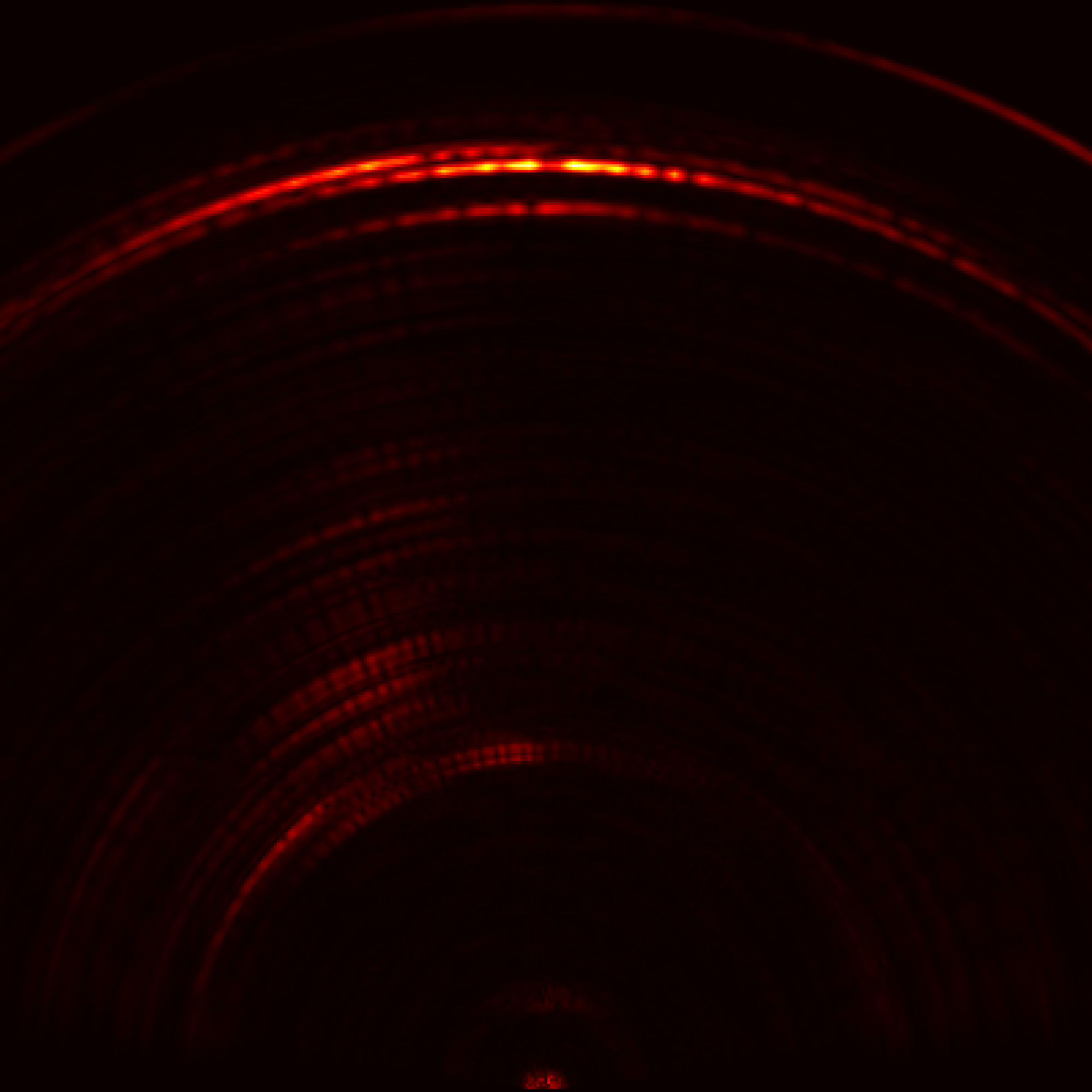} & \includegraphics[width=0.165\linewidth]{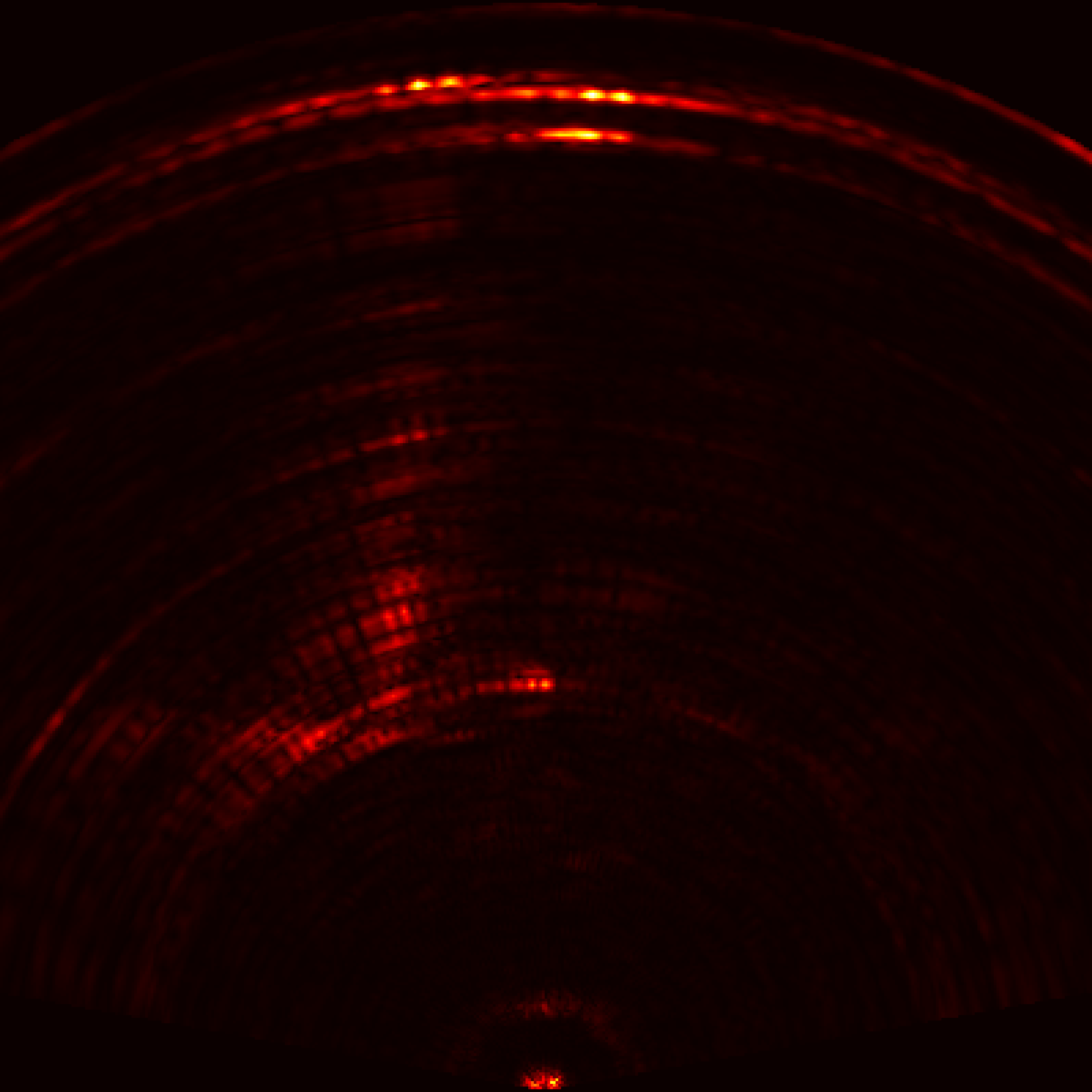} \\
\raisebox{2.0em}{\rotatebox{90}{\scriptsize\textbf{3DPS (Ours)}}} & \includegraphics[width=0.165\linewidth]{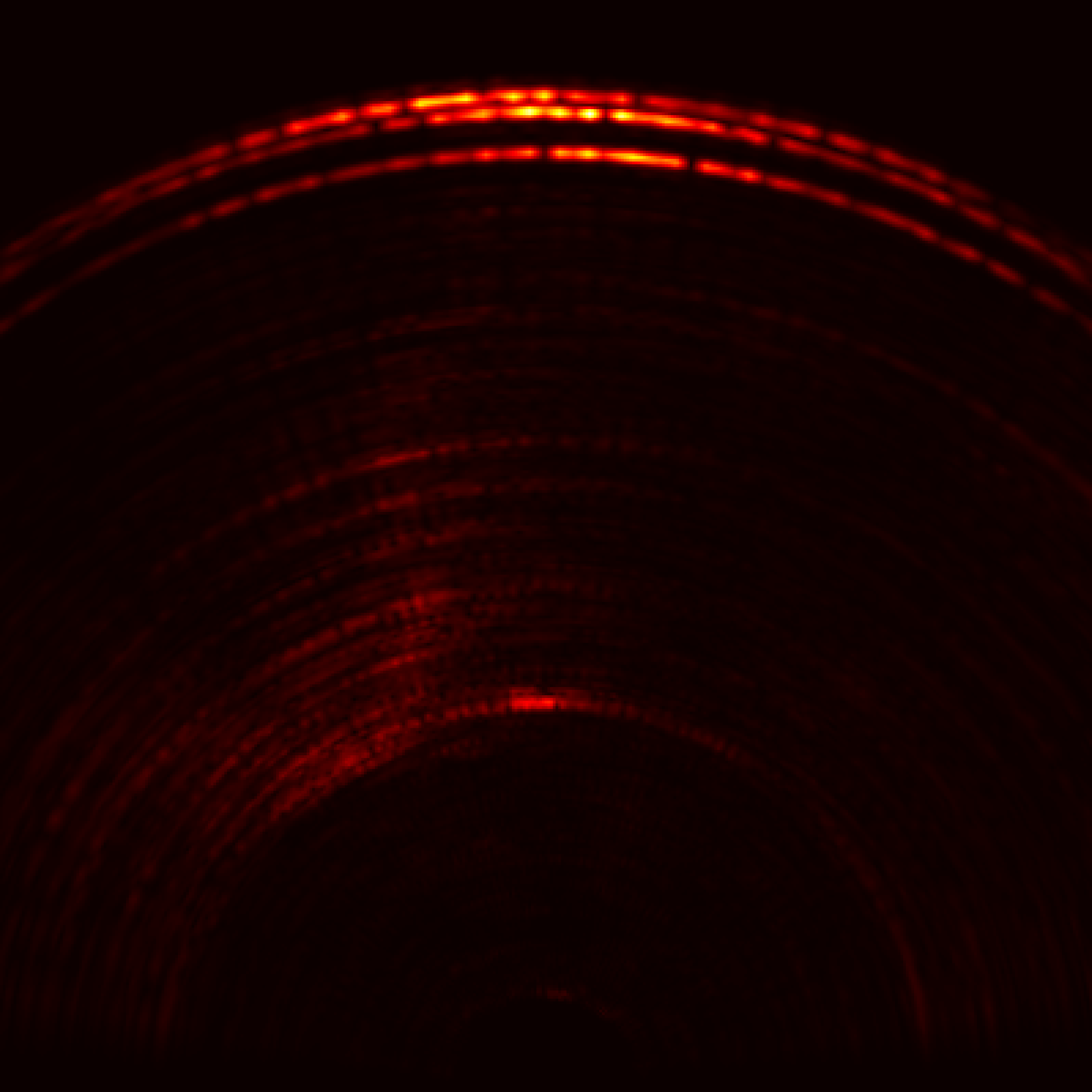} & \includegraphics[width=0.165\linewidth]{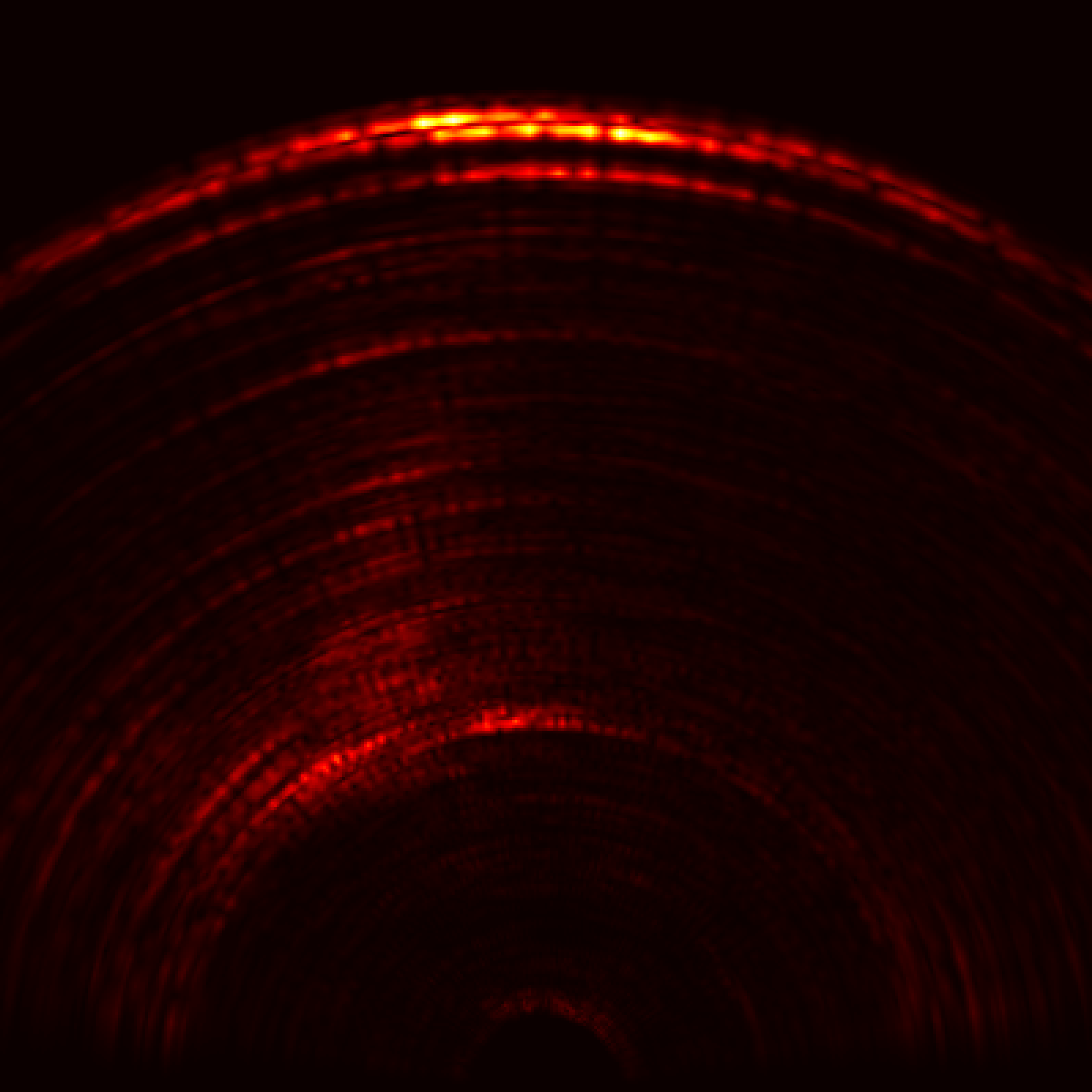} & \includegraphics[width=0.165\linewidth]{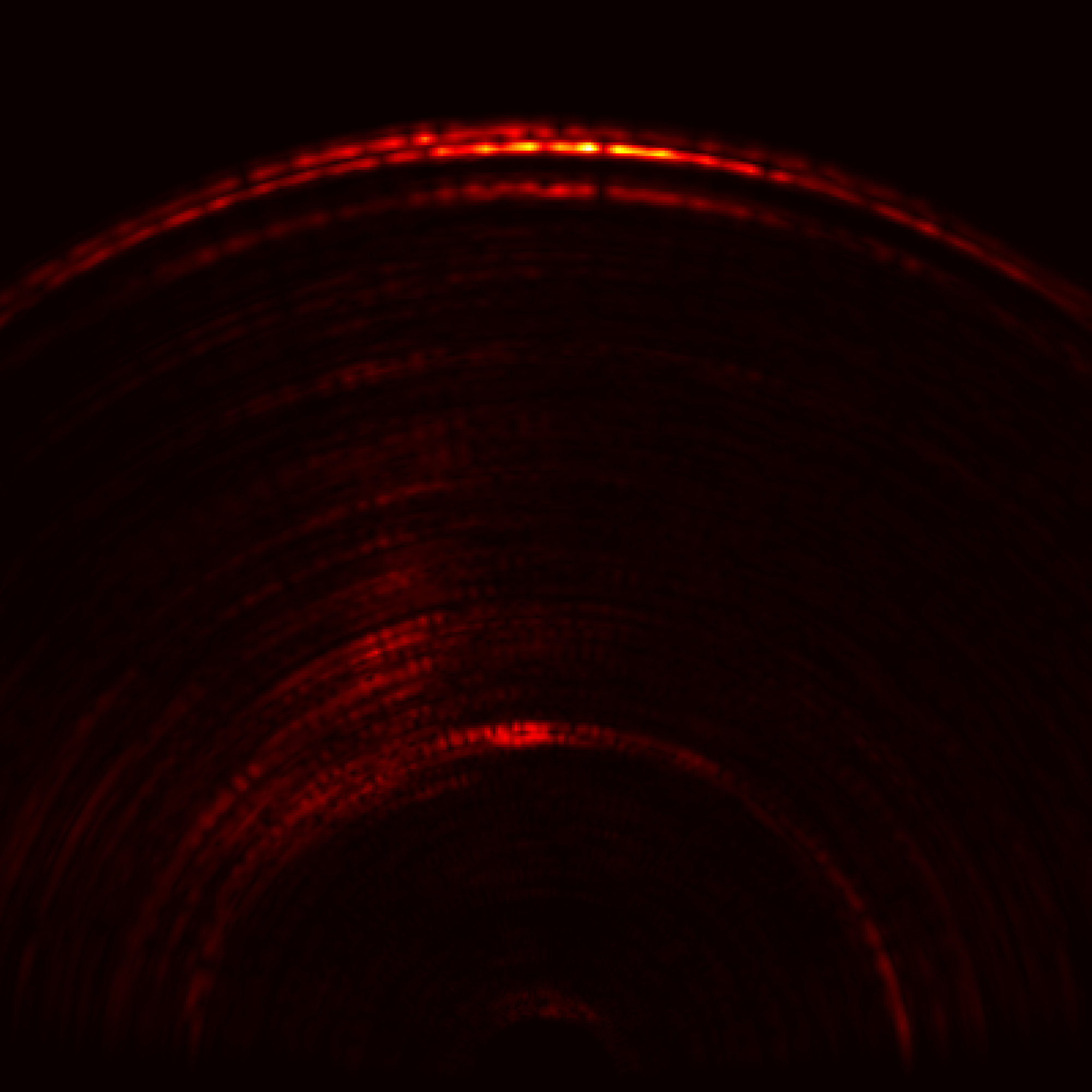} & \includegraphics[width=0.165\linewidth]{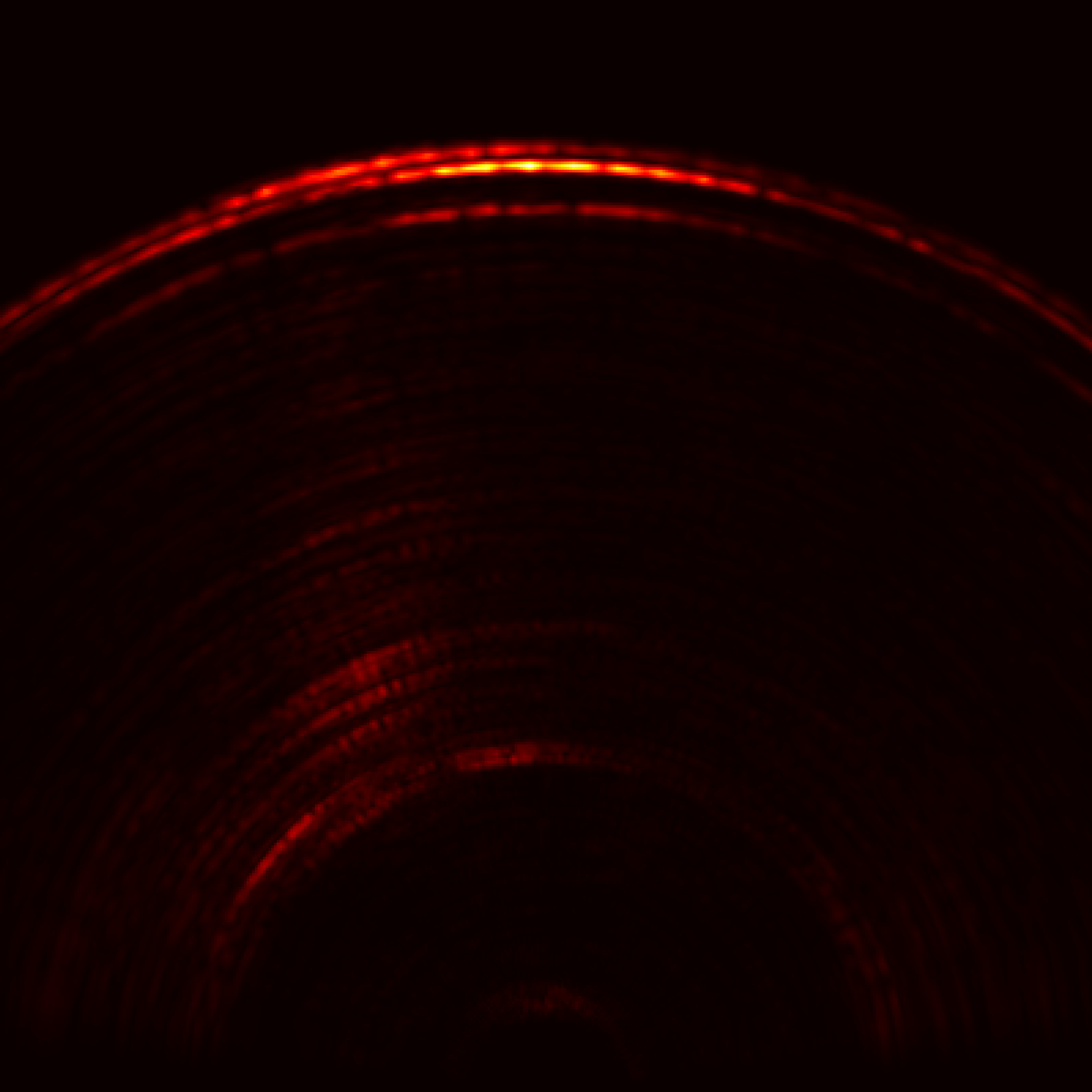} & \includegraphics[width=0.165\linewidth]{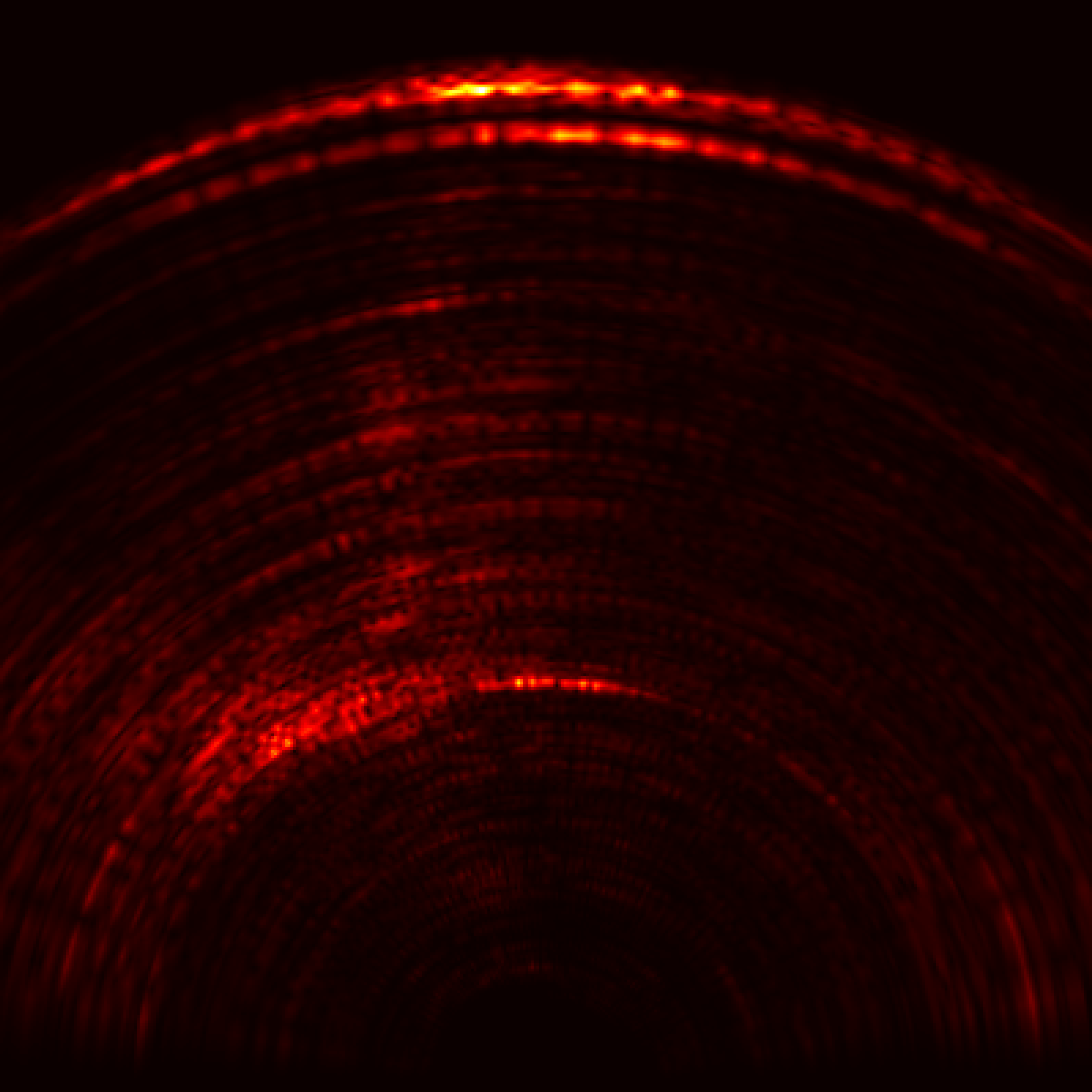} \\
\raisebox{2.0em}{\rotatebox{90}{\scriptsize\textbf{mmIR}}} & \includegraphics[width=0.165\linewidth]{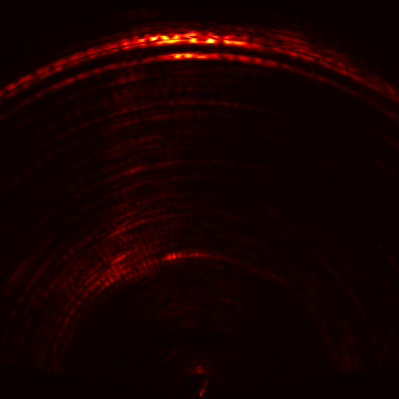} & \includegraphics[width=0.165\linewidth]{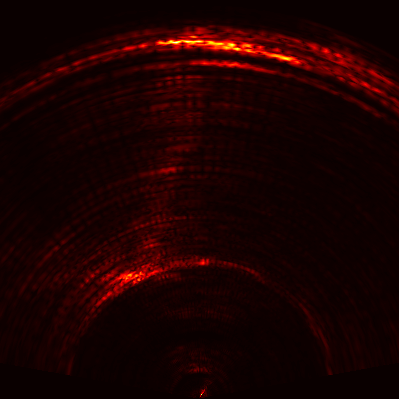} & \includegraphics[width=0.165\linewidth]{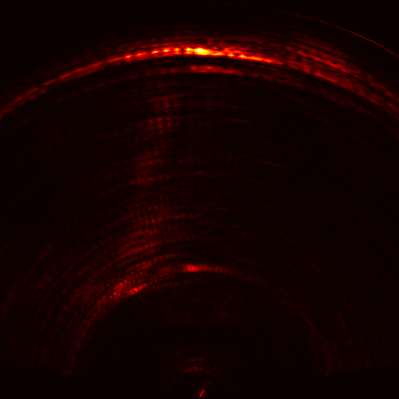} & \includegraphics[width=0.165\linewidth]{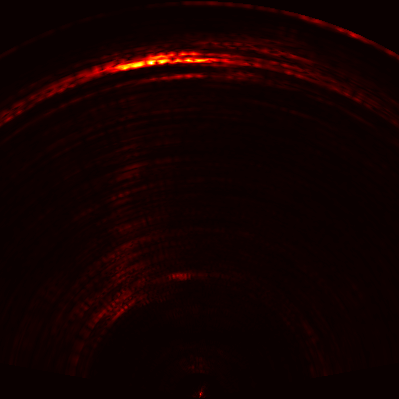} & \includegraphics[width=0.165\linewidth]{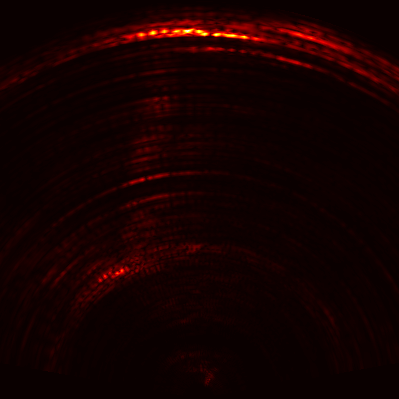} \\
\raisebox{2.0em}{\rotatebox{90}{\scriptsize\textbf{RadarSplat}}} & \includegraphics[width=0.165\linewidth]{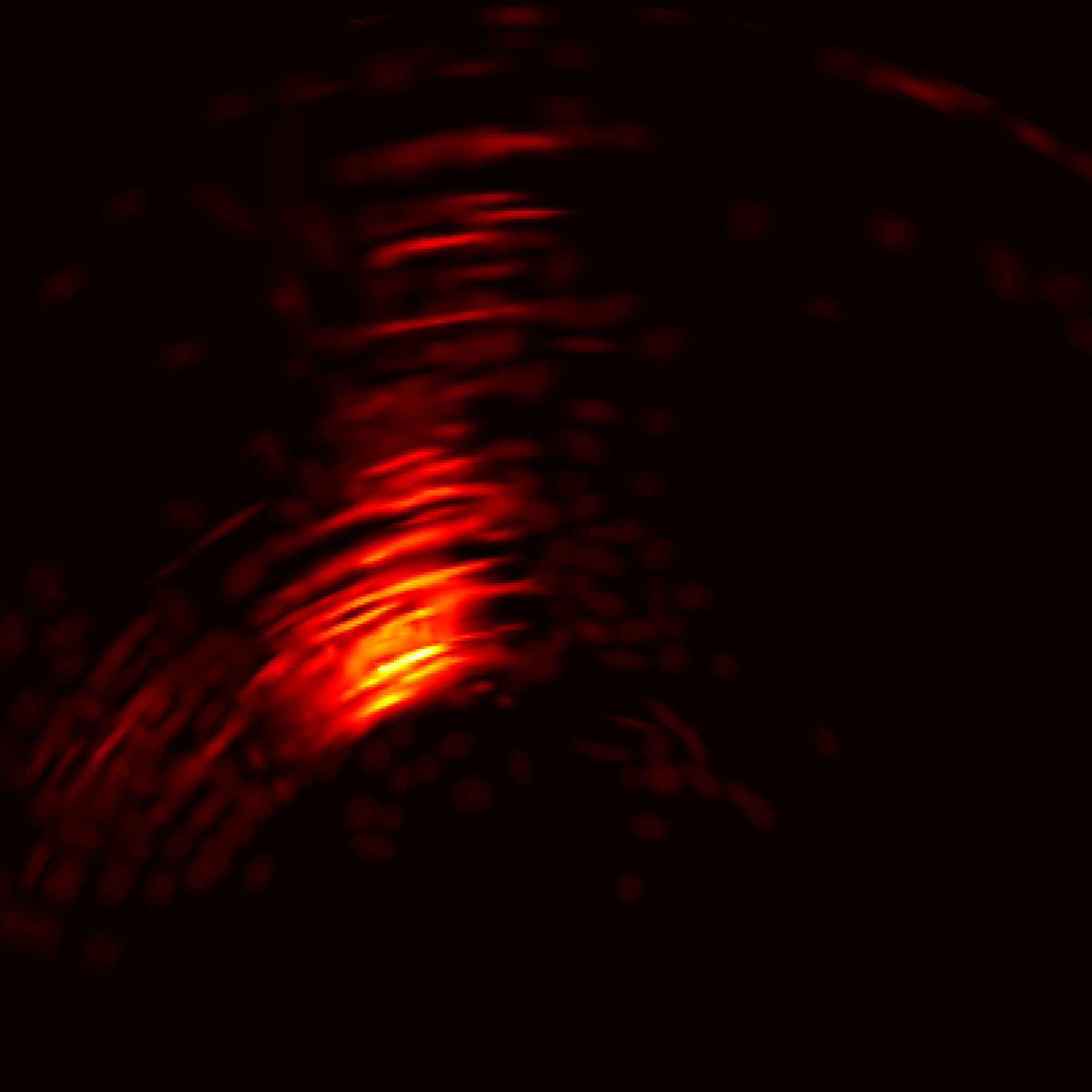} & \includegraphics[width=0.165\linewidth]{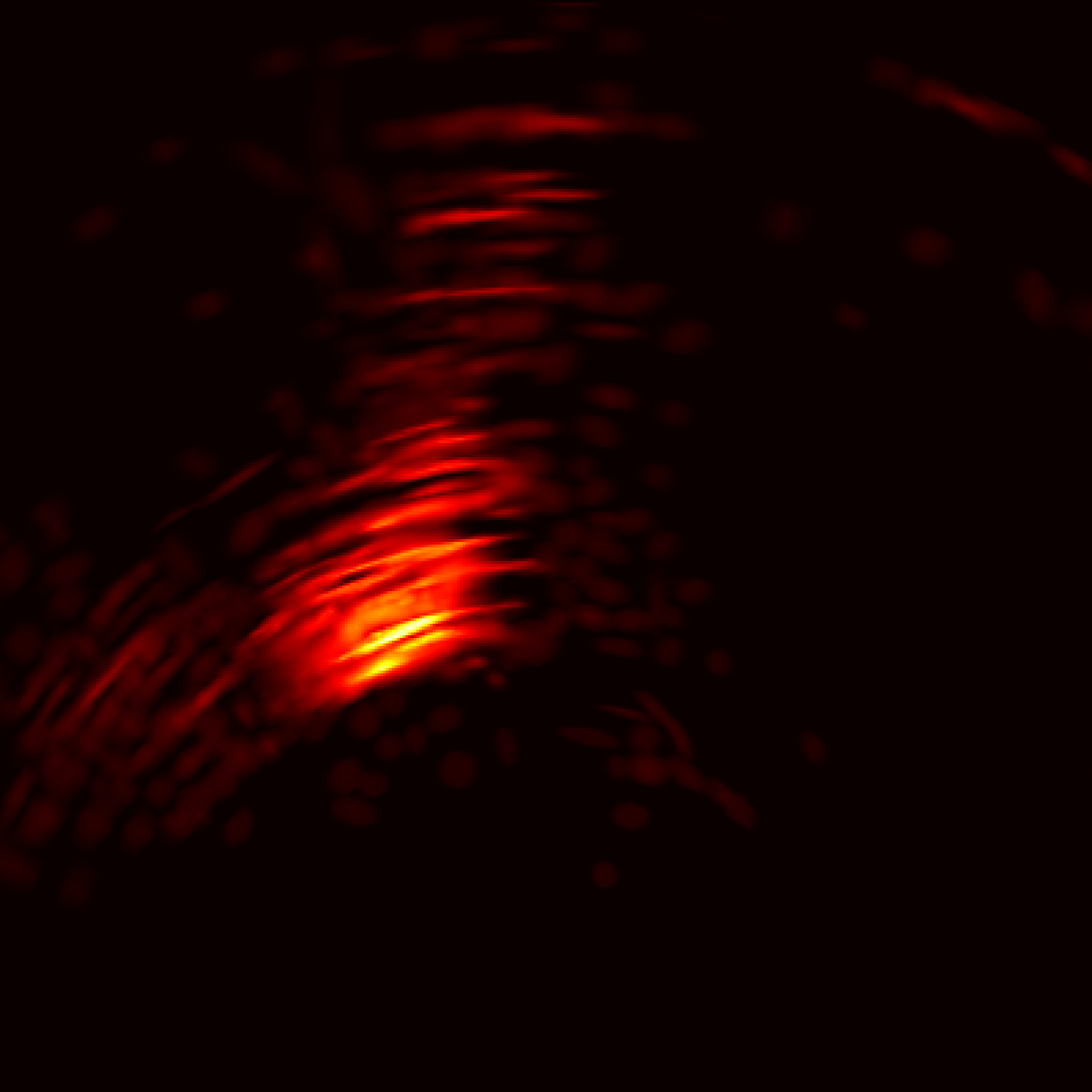} & \includegraphics[width=0.165\linewidth]{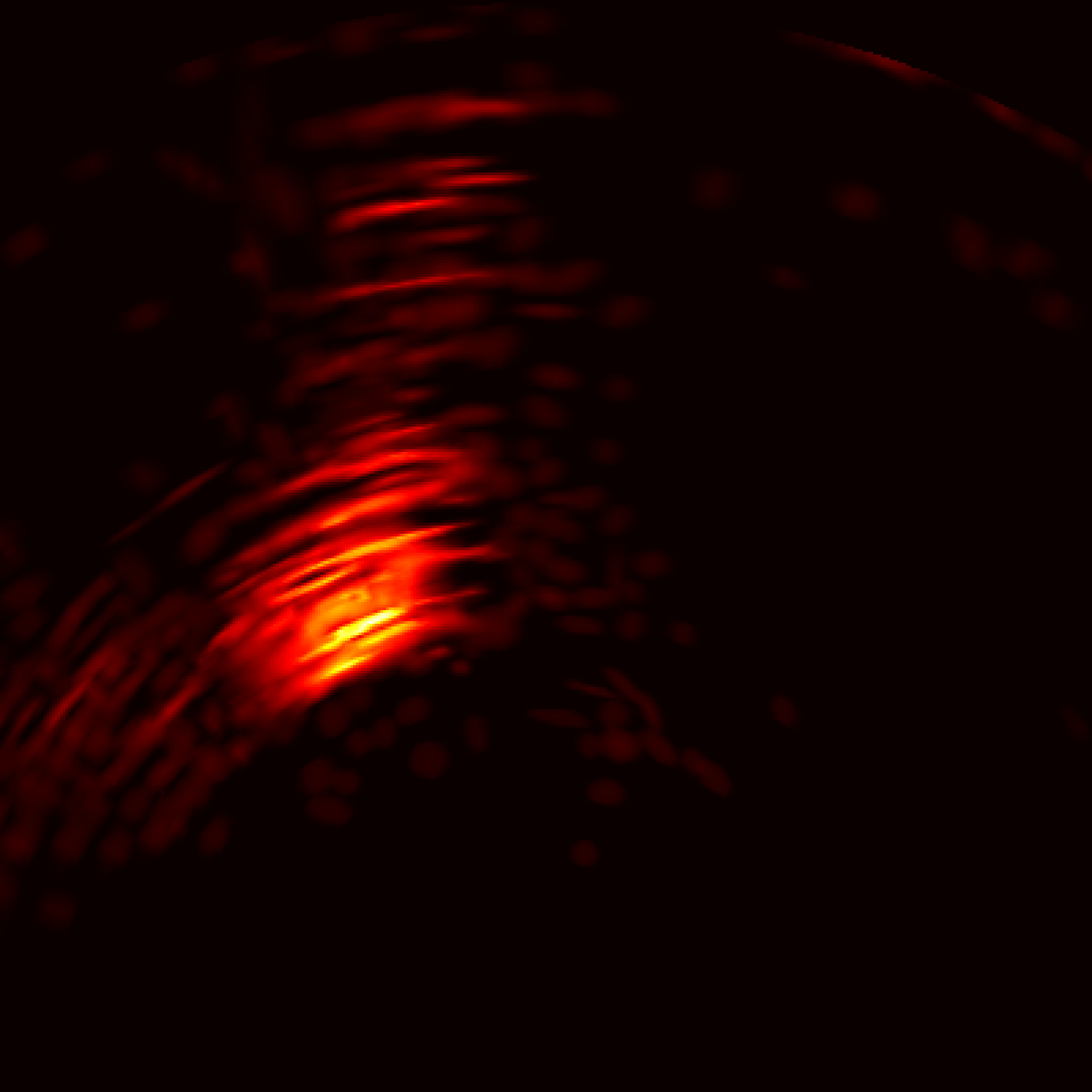} & \includegraphics[width=0.165\linewidth]{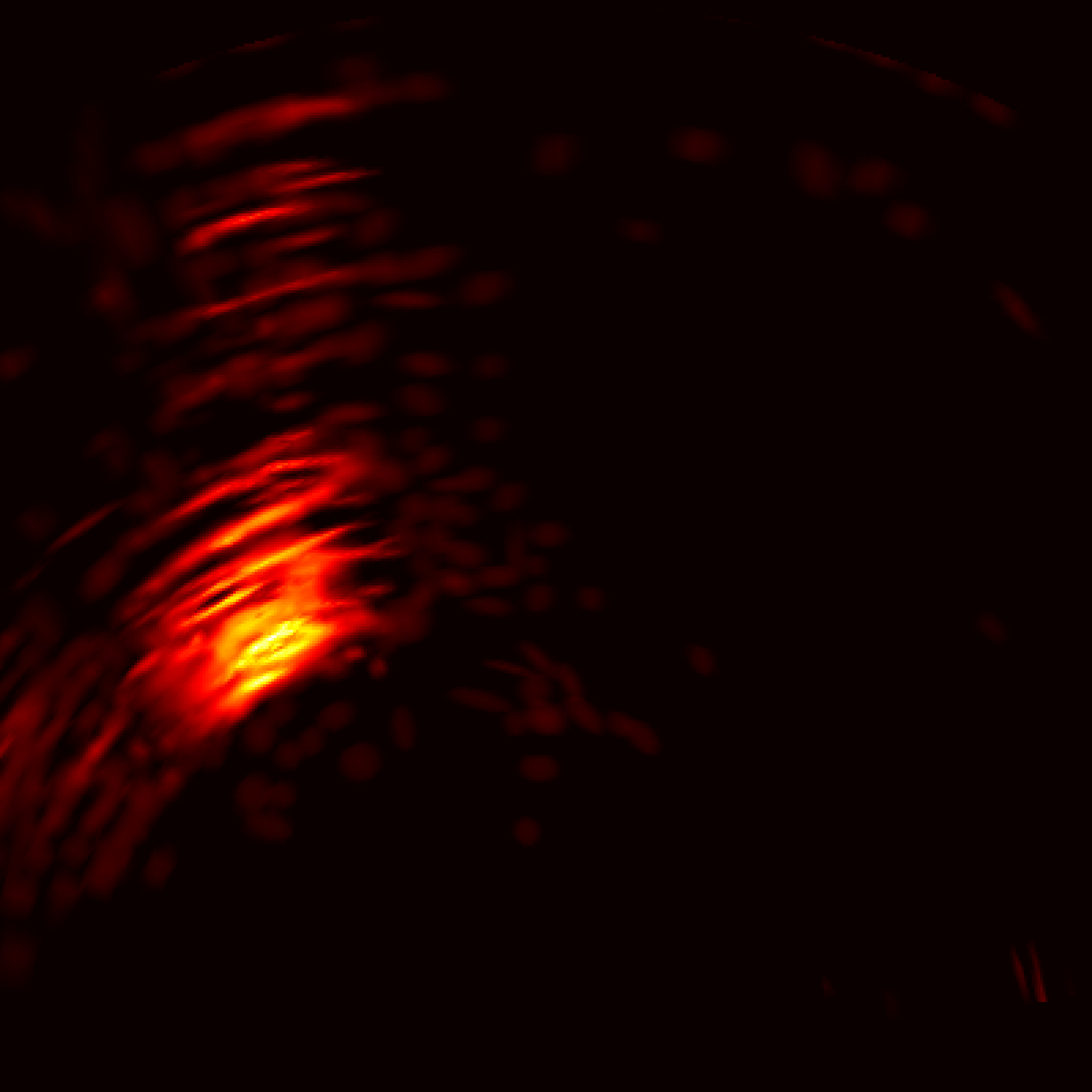} & \includegraphics[width=0.165\linewidth]{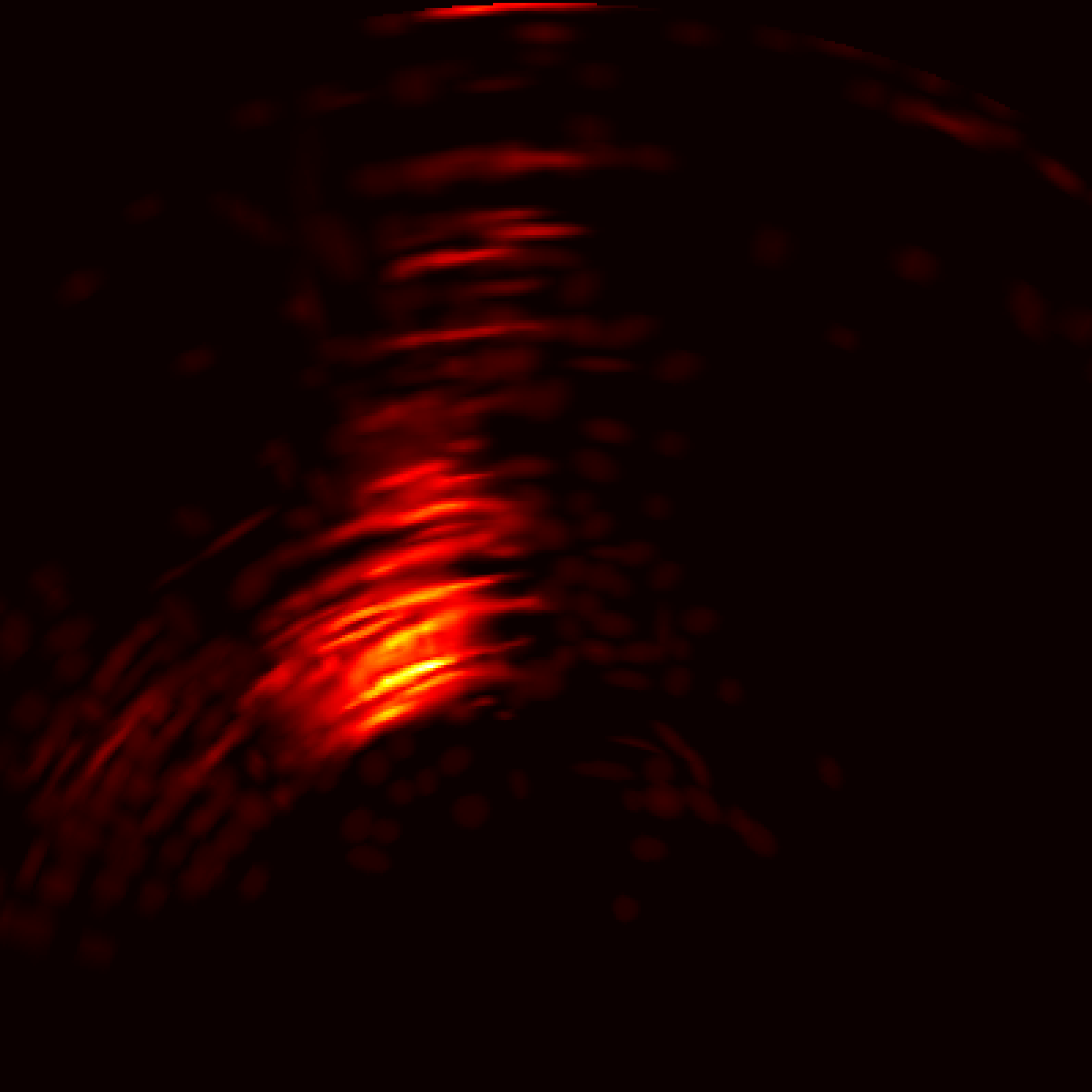} \\
\raisebox{2.0em}{\rotatebox{90}{\scriptsize\textbf{Radar Fields}}} & \includegraphics[width=0.165\linewidth]{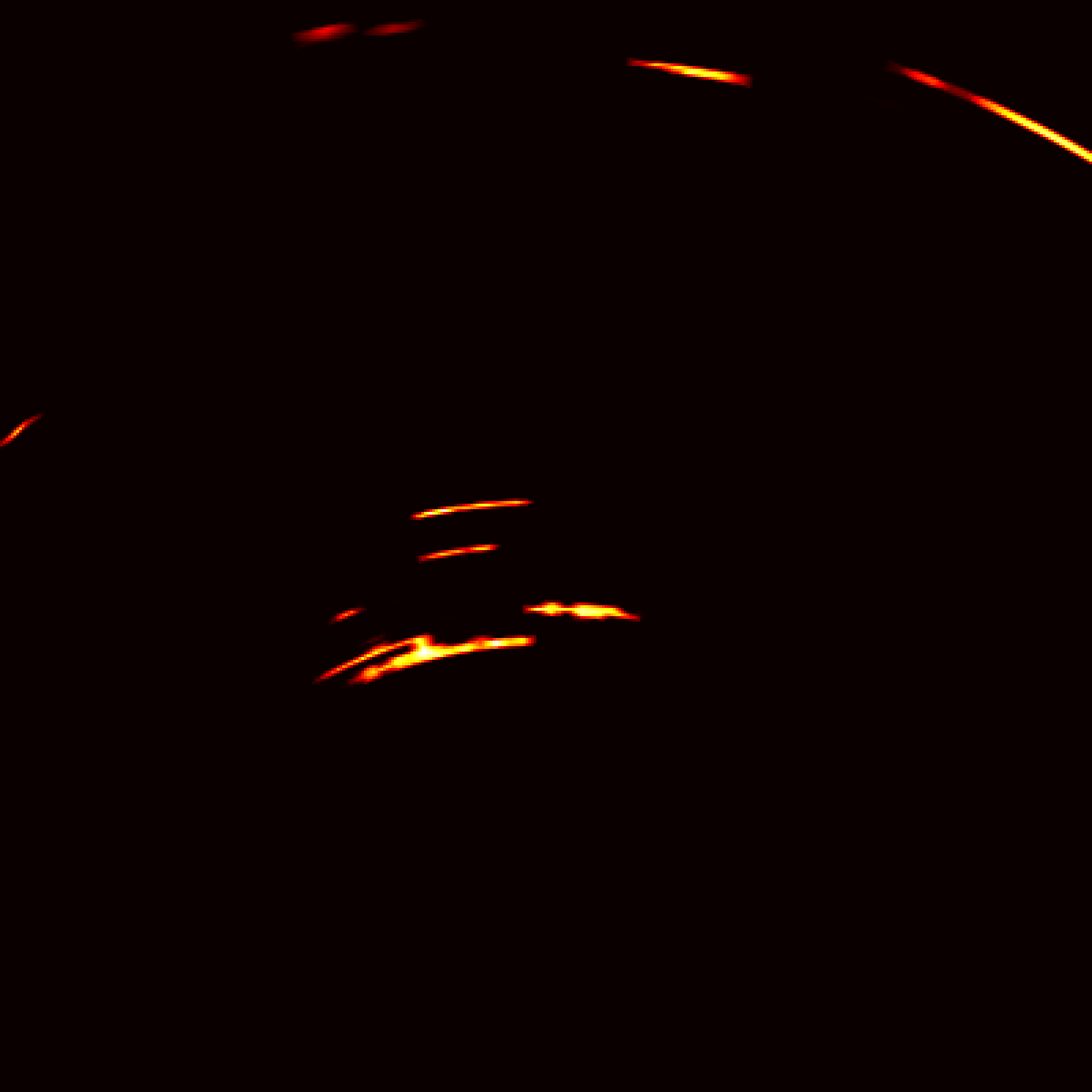} & \includegraphics[width=0.165\linewidth]{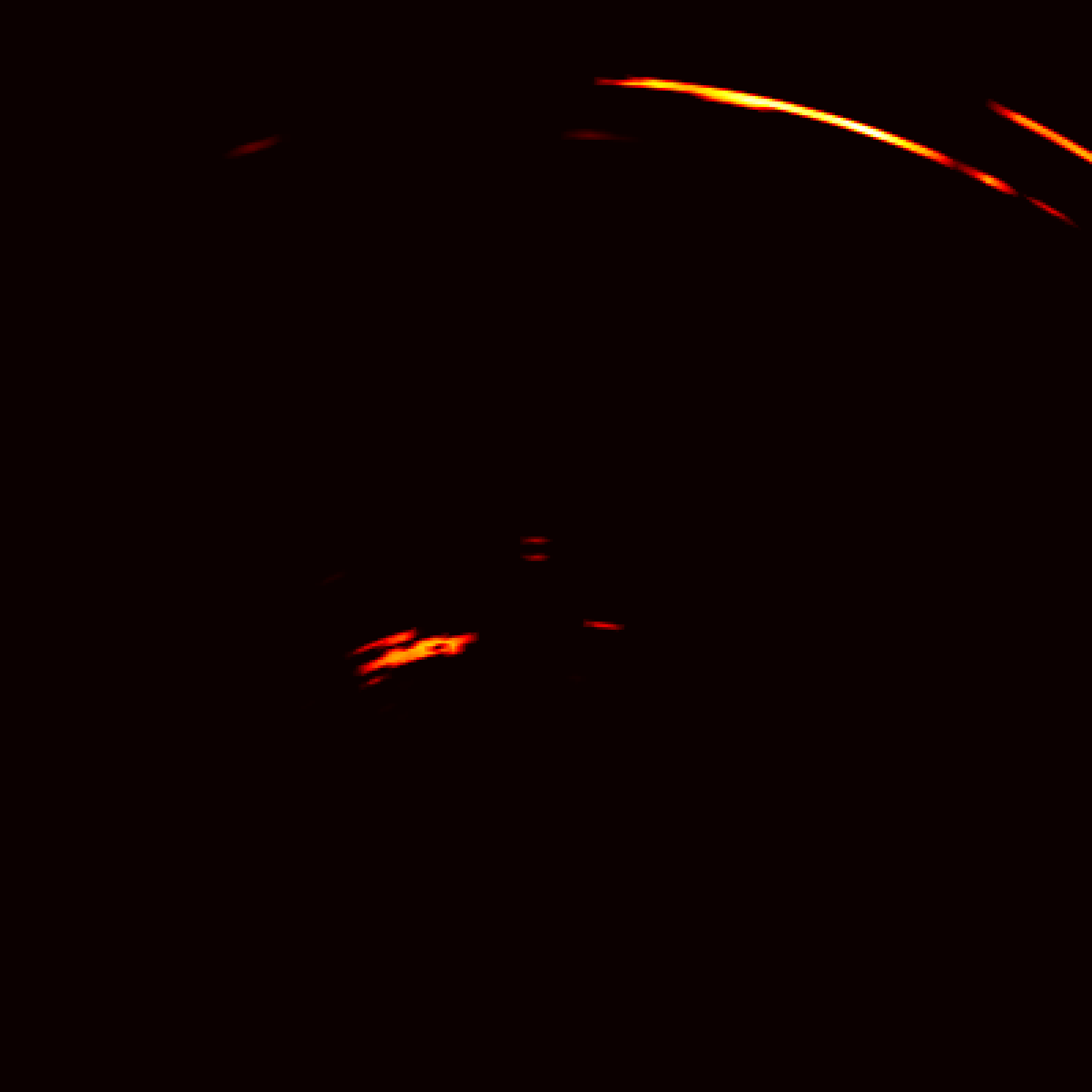} & \includegraphics[width=0.165\linewidth]{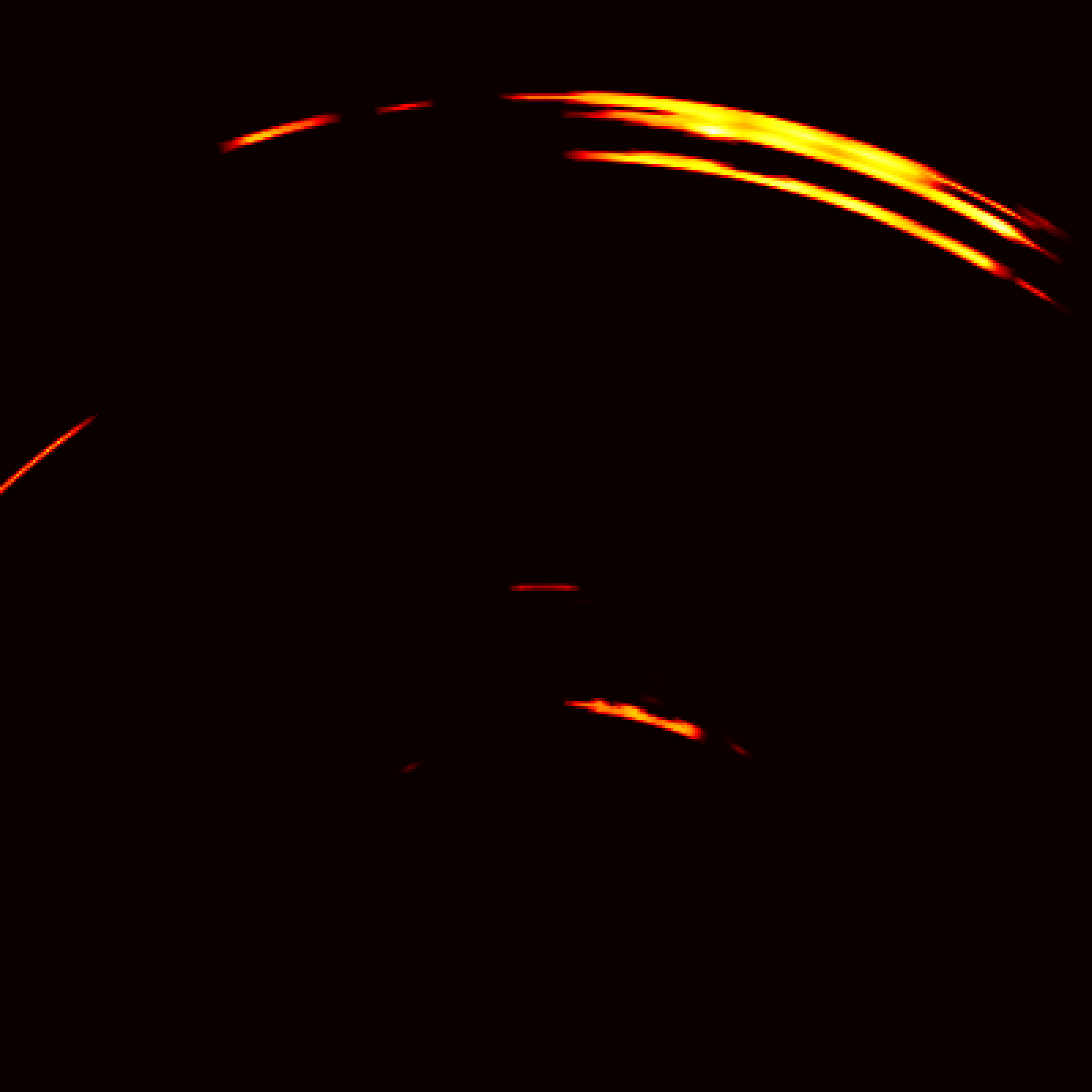} & \includegraphics[width=0.165\linewidth]{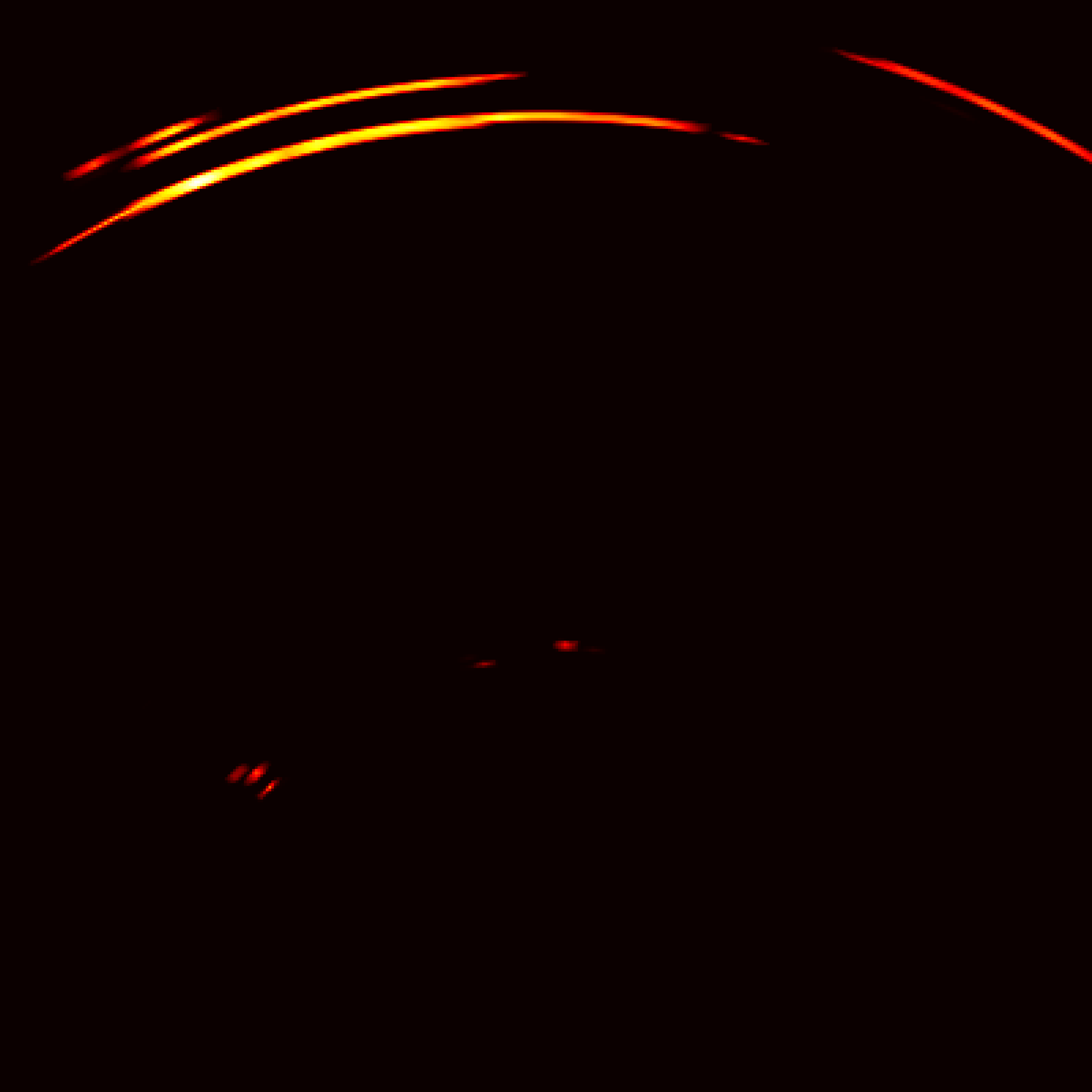} & \includegraphics[width=0.165\linewidth]{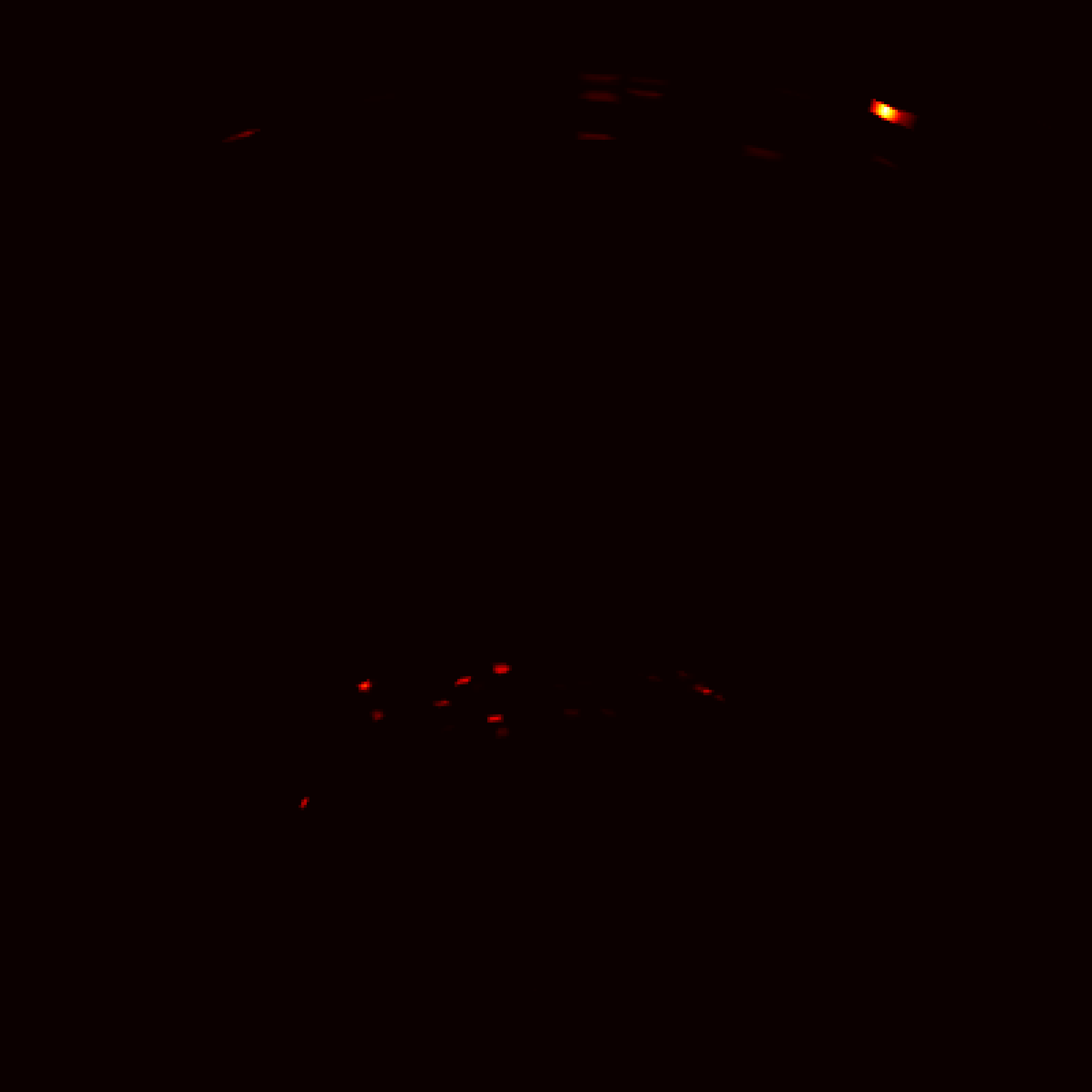} \\
\raisebox{2.0em}{\rotatebox{90}{\scriptsize\textbf{DART}}} & \includegraphics[width=0.165\linewidth]{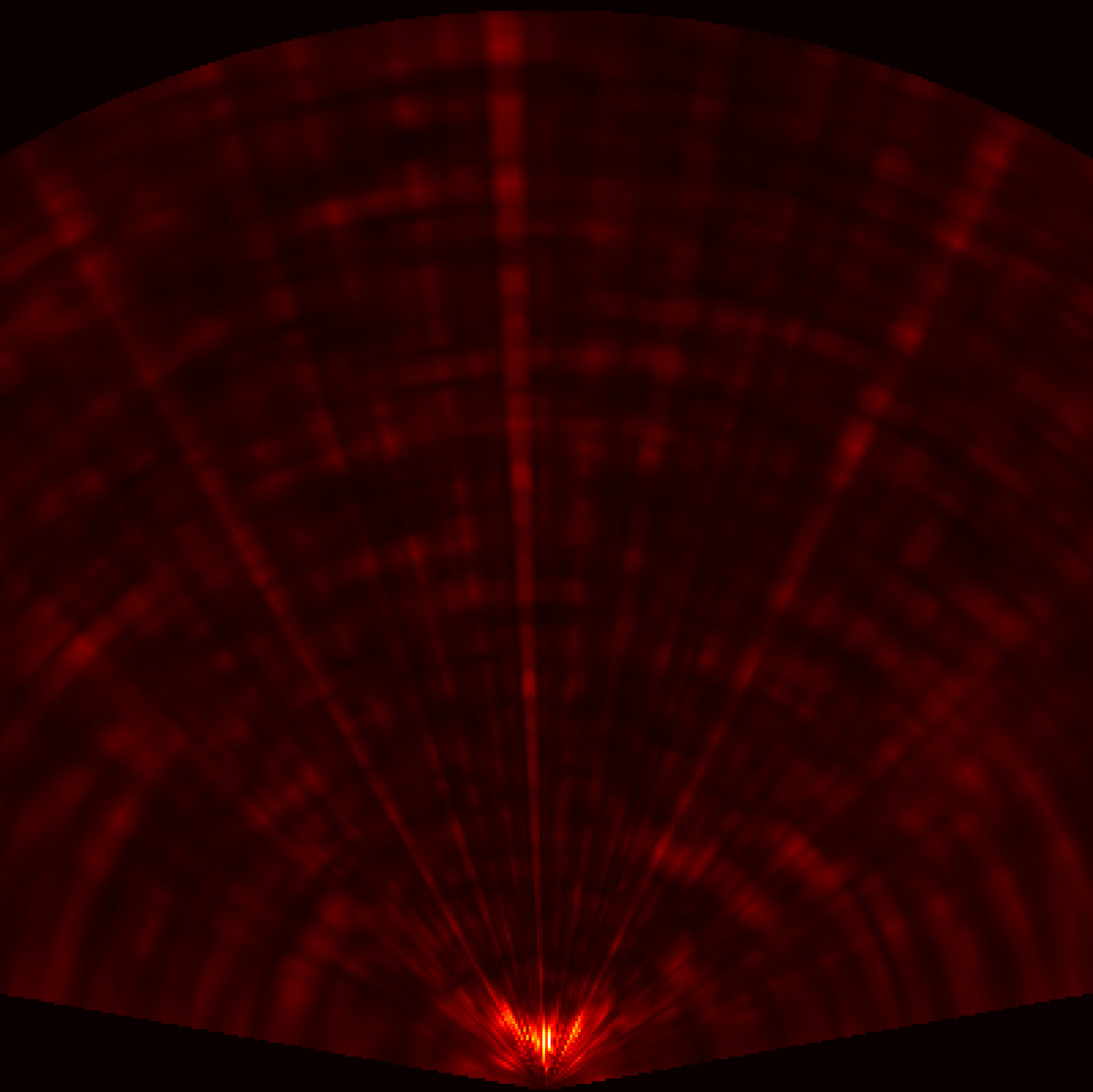} & \includegraphics[width=0.165\linewidth]{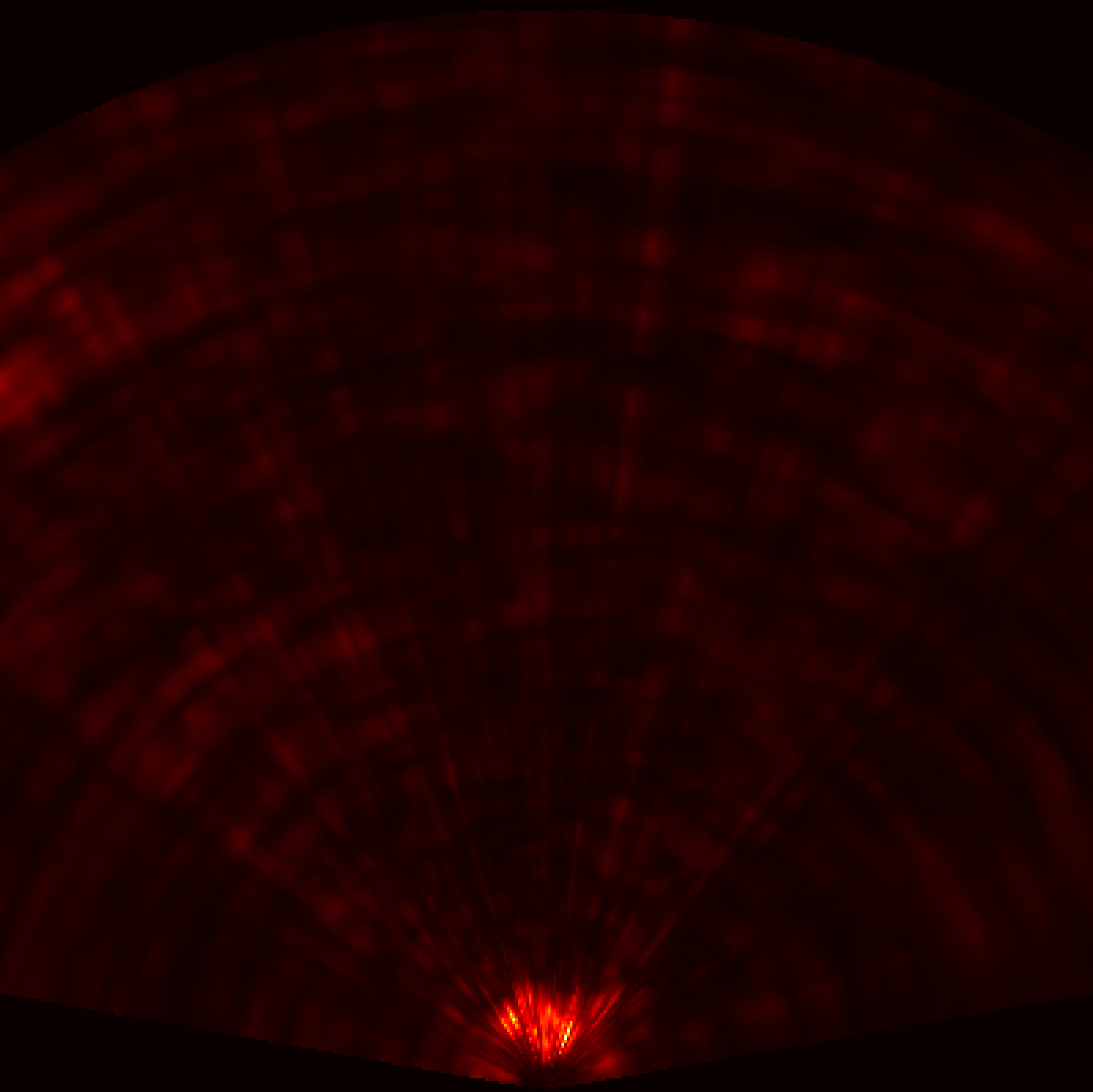} & \includegraphics[width=0.165\linewidth]{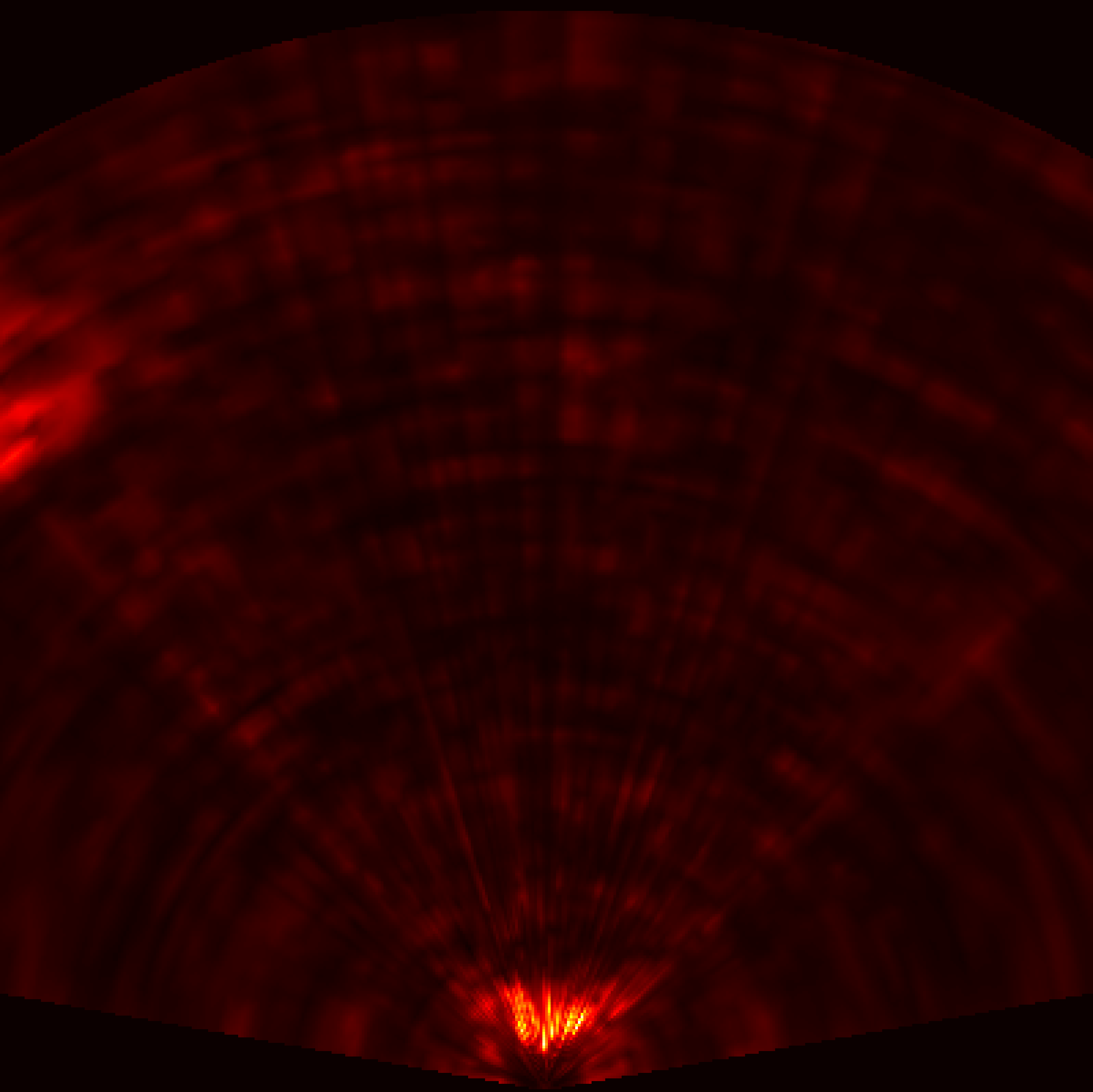} & \includegraphics[width=0.165\linewidth]{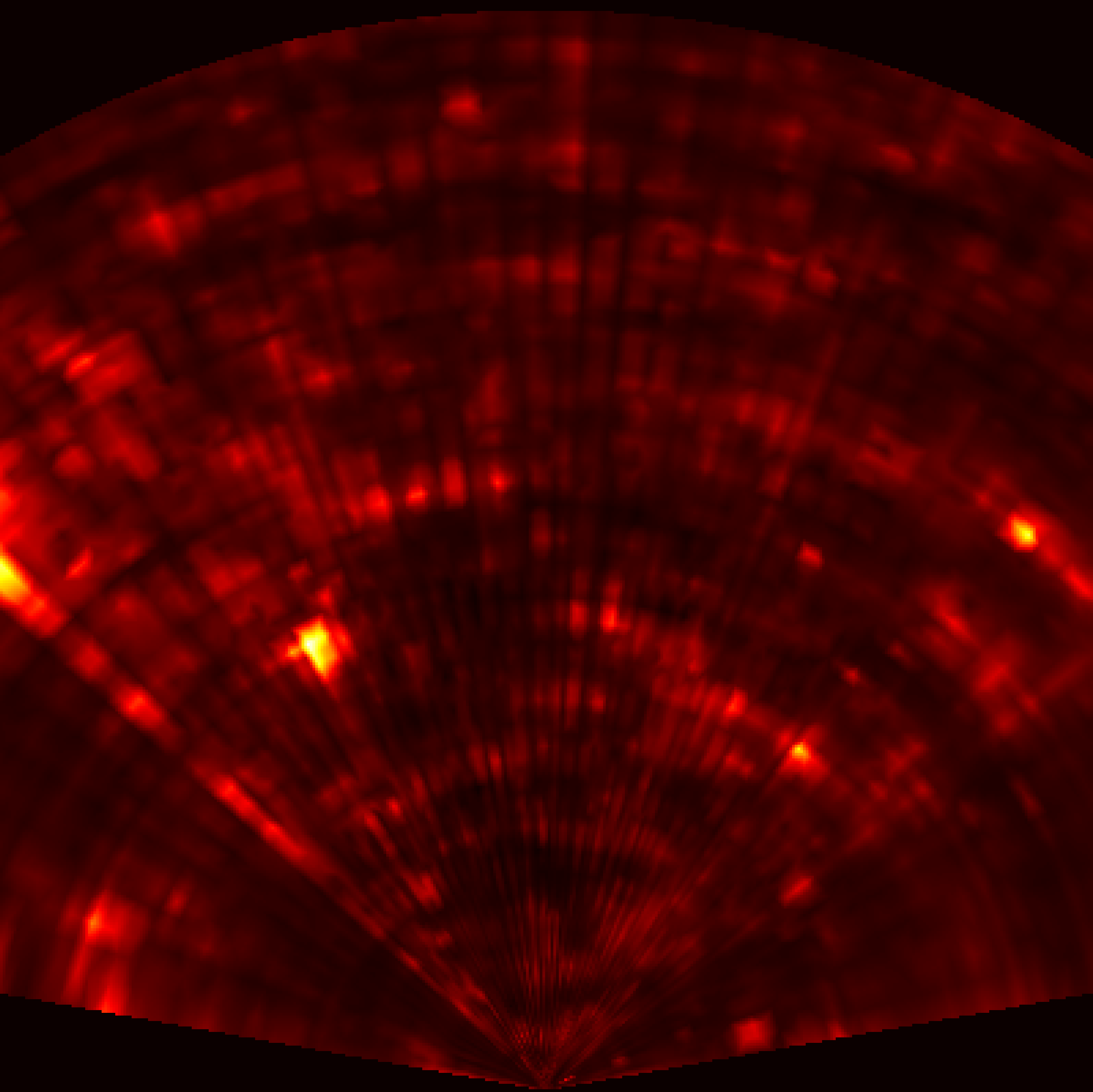} & \includegraphics[width=0.165\linewidth]{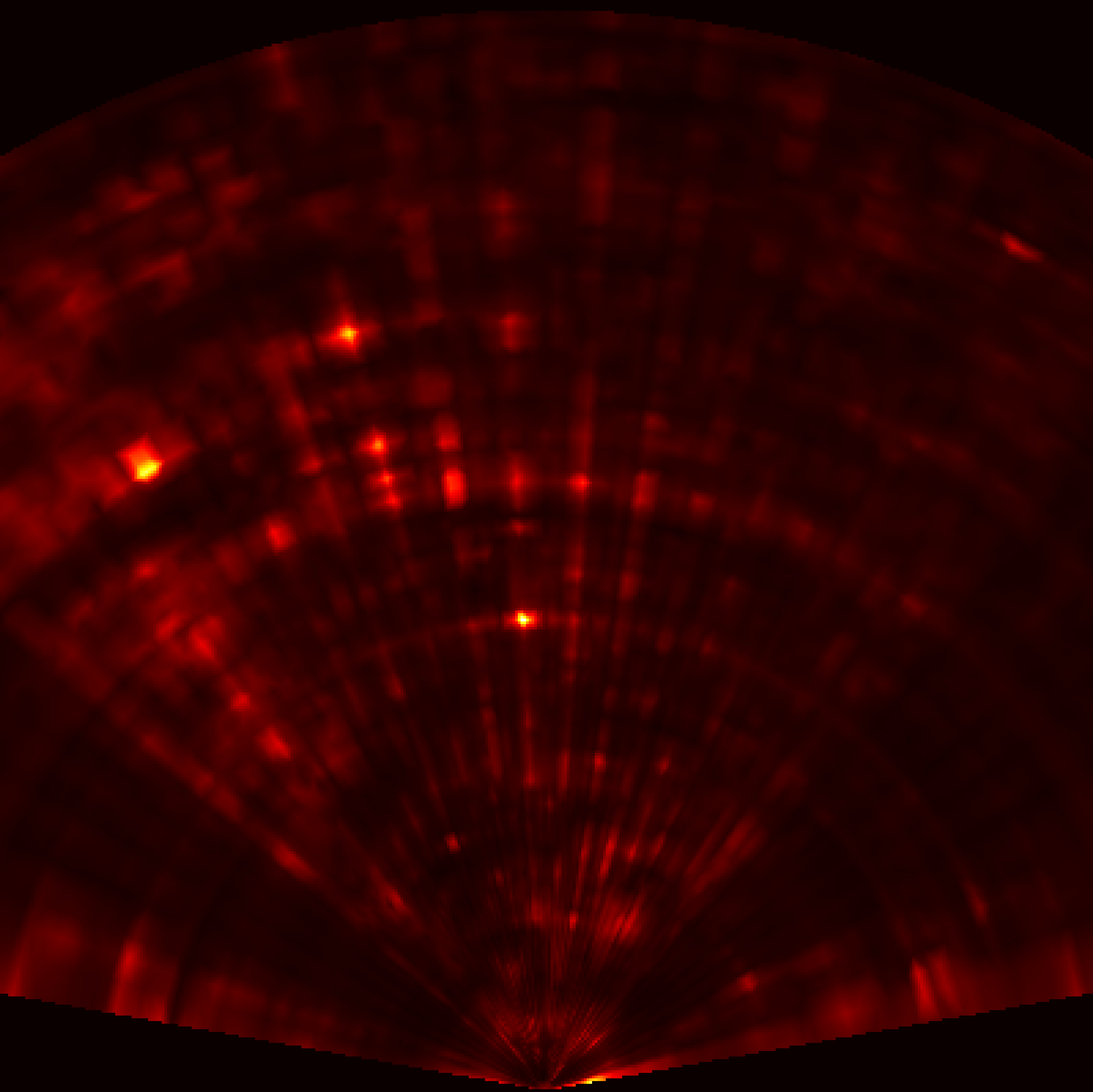} \\
\end{tabular}
\caption{\textbf{Per-frame $|\mathrm{RA}|$ comparison, scene S1\,F438, training frames $F\!+\!1$ to $F\!+\!4$ and held-out test frame $F$ (red header).} Pearson correlation per column, left to right. 3DPS: 0.87, 0.85, 0.88, 0.92, 0.64; mmIR: 0.82, 0.68, 0.74, 0.77, 0.60; RadarSplat: 0.18, 0.21, 0.17, 0.03, 0.26; Radar Fields: 0.09, 0.06, 0.11, 0.05, 0.04; DART: 0.02, 0.05, 0.01, 0.06, 0.07.}
\label{fig:supp_s1_f438_b}
\end{figure}

\begin{figure}[p]
\centering
\setlength{\tabcolsep}{1pt}\renewcommand{\arraystretch}{0.6}
\begin{tabular}{ccccc}
 & {\footnotesize $F\!-\!4$} & {\footnotesize $F\!-\!3$} & {\footnotesize $F\!-\!2$} & {\footnotesize $F\!-\!1$} \\
\raisebox{2.0em}{\rotatebox{90}{\scriptsize\textbf{GT}}} & \includegraphics[width=0.165\linewidth]{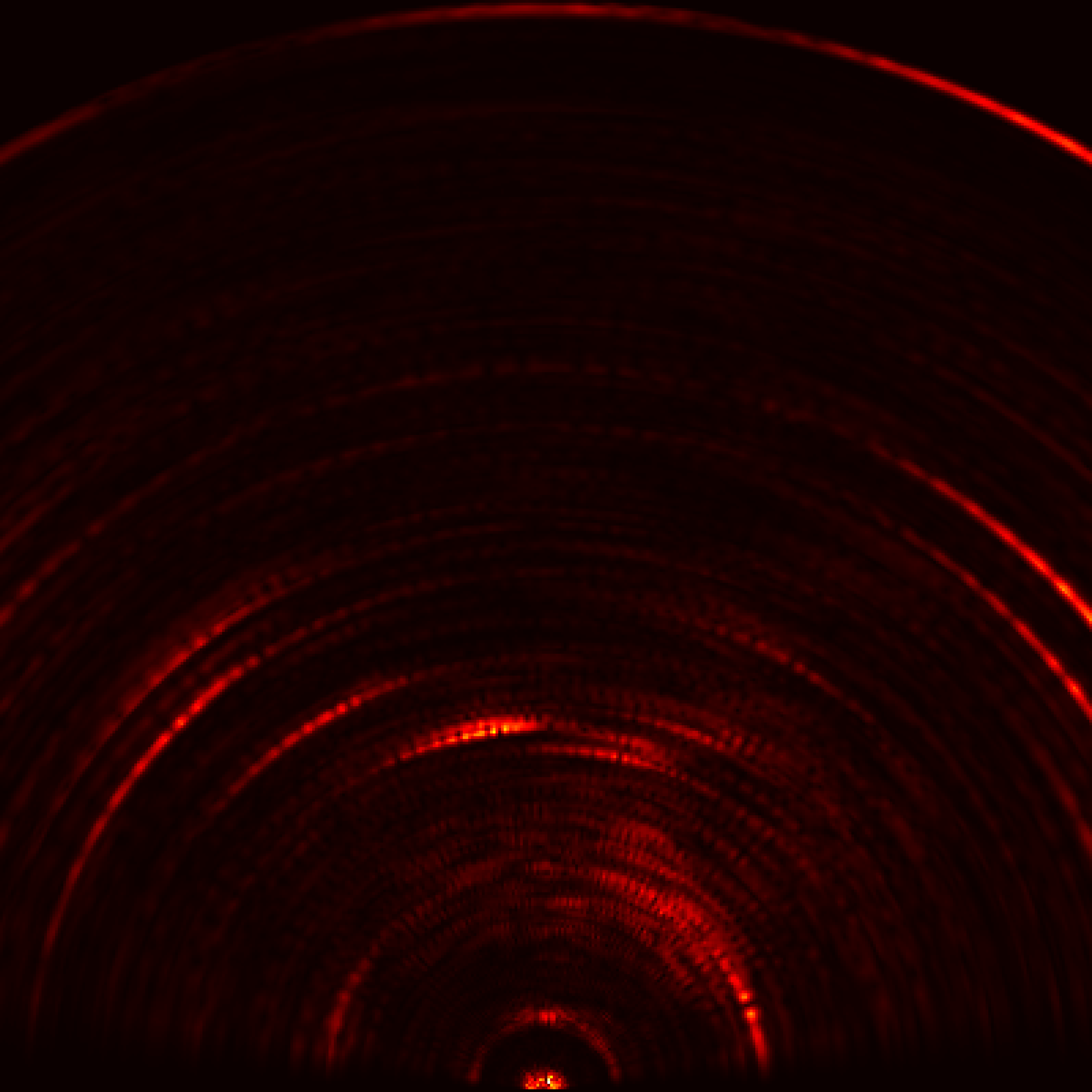} & \includegraphics[width=0.165\linewidth]{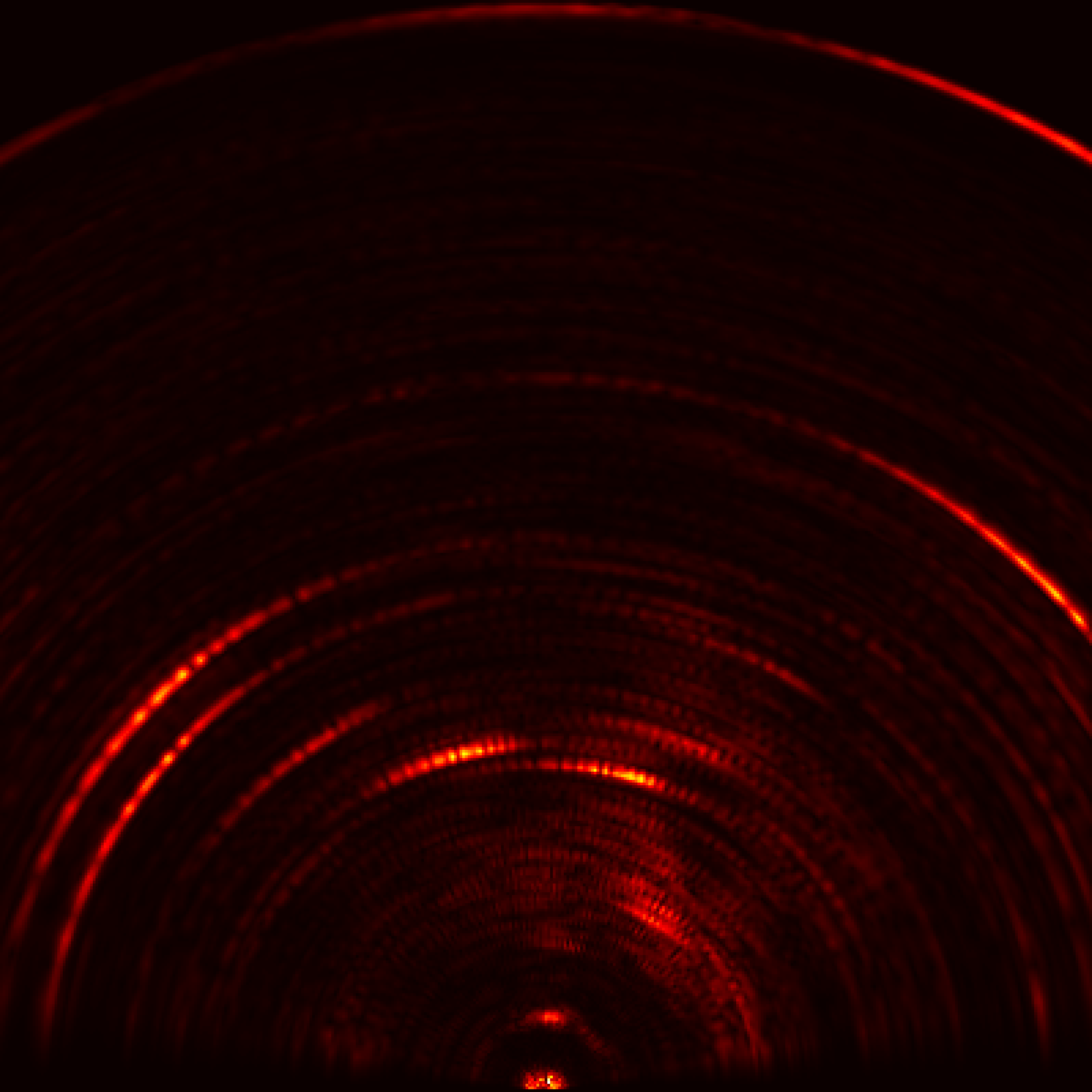} & \includegraphics[width=0.165\linewidth]{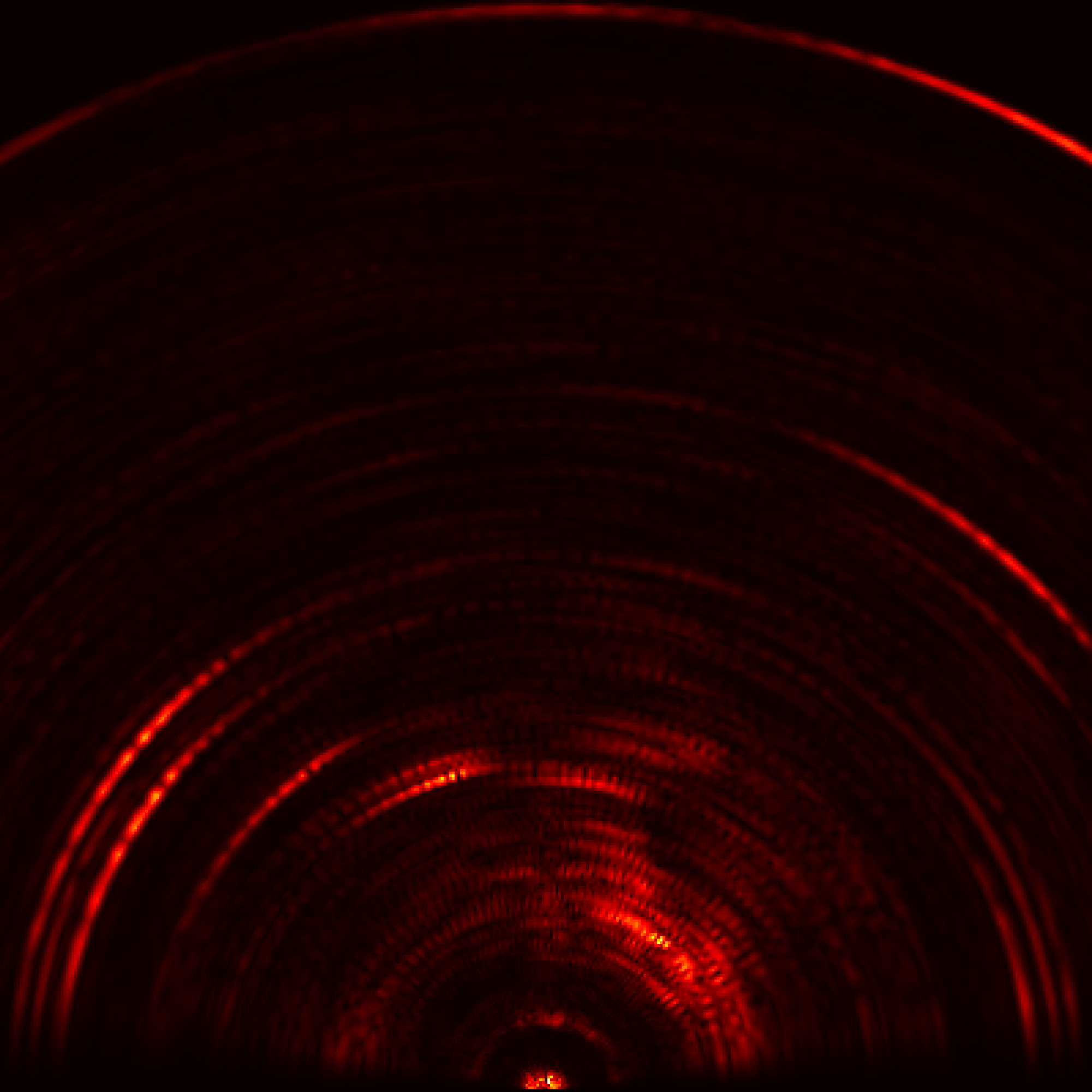} & \includegraphics[width=0.165\linewidth]{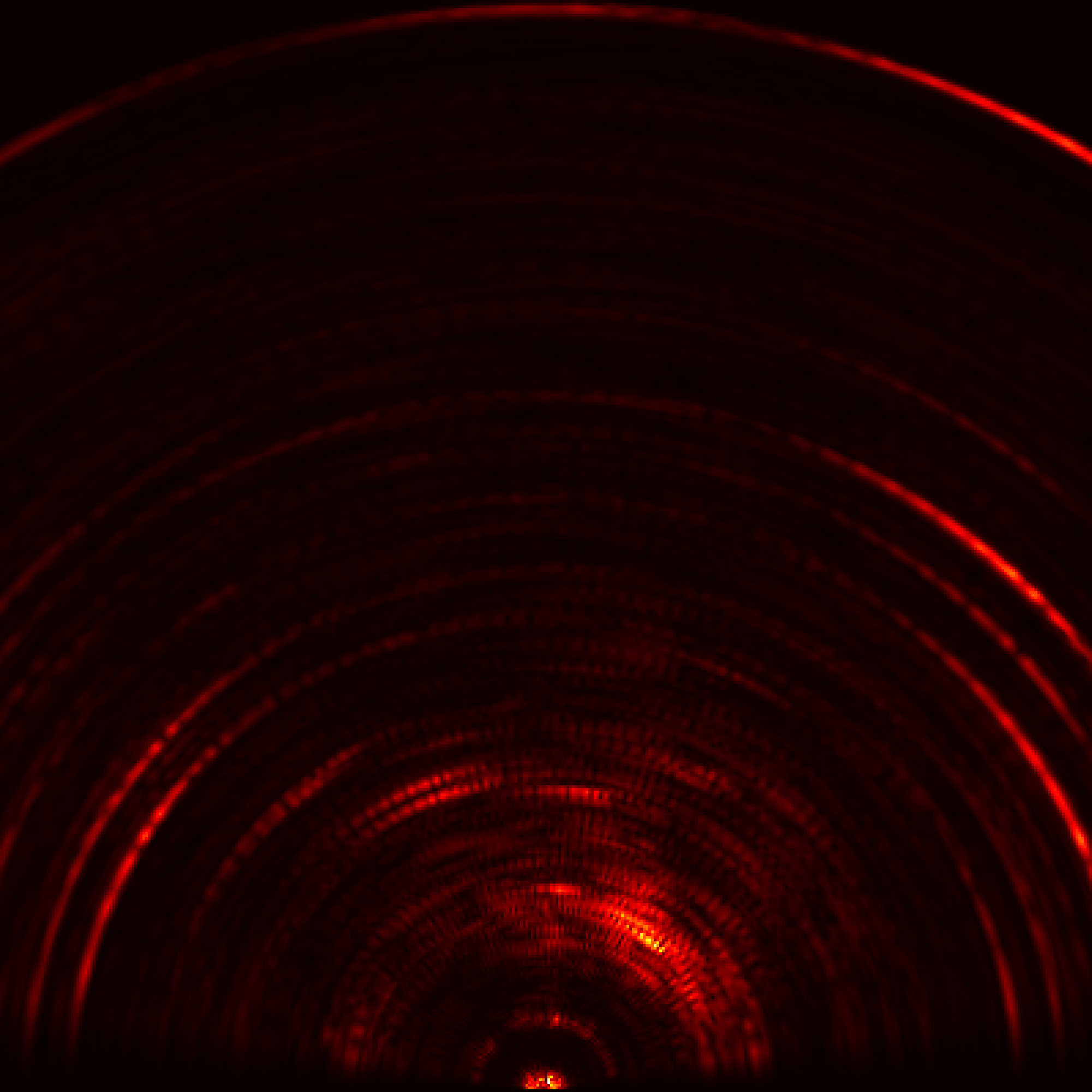} \\
\raisebox{2.0em}{\rotatebox{90}{\scriptsize\textbf{3DPS (Ours)}}} & \includegraphics[width=0.165\linewidth]{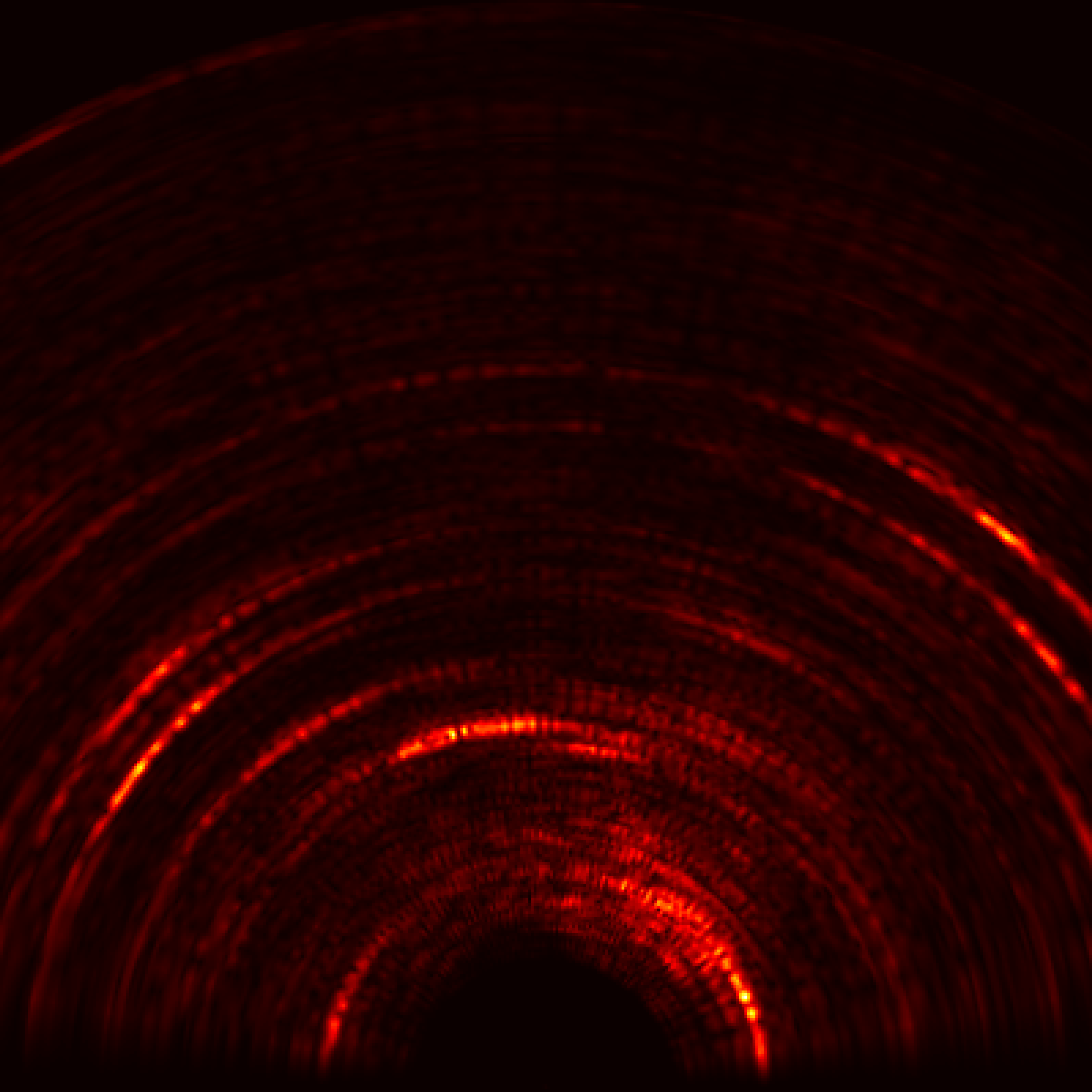} & \includegraphics[width=0.165\linewidth]{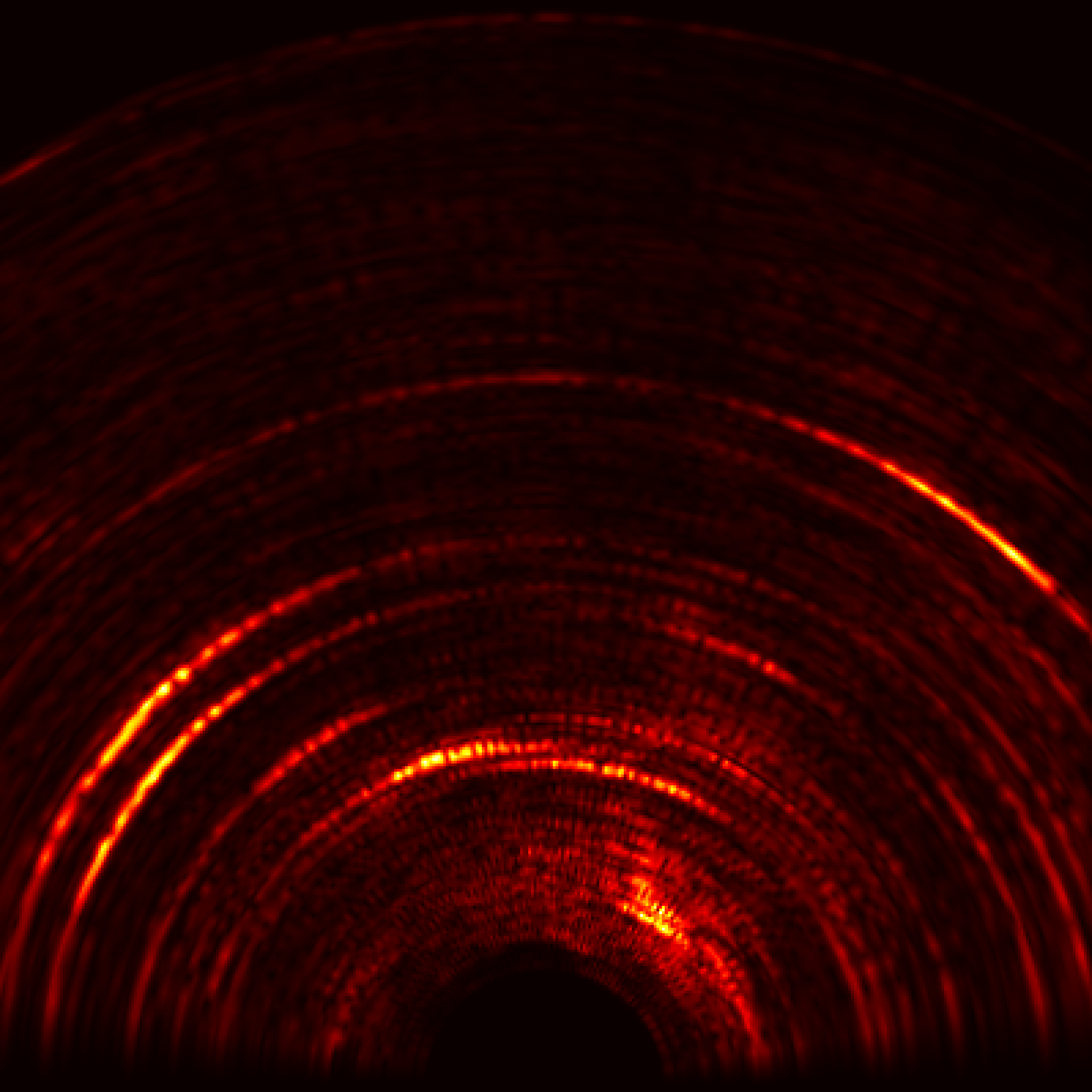} & \includegraphics[width=0.165\linewidth]{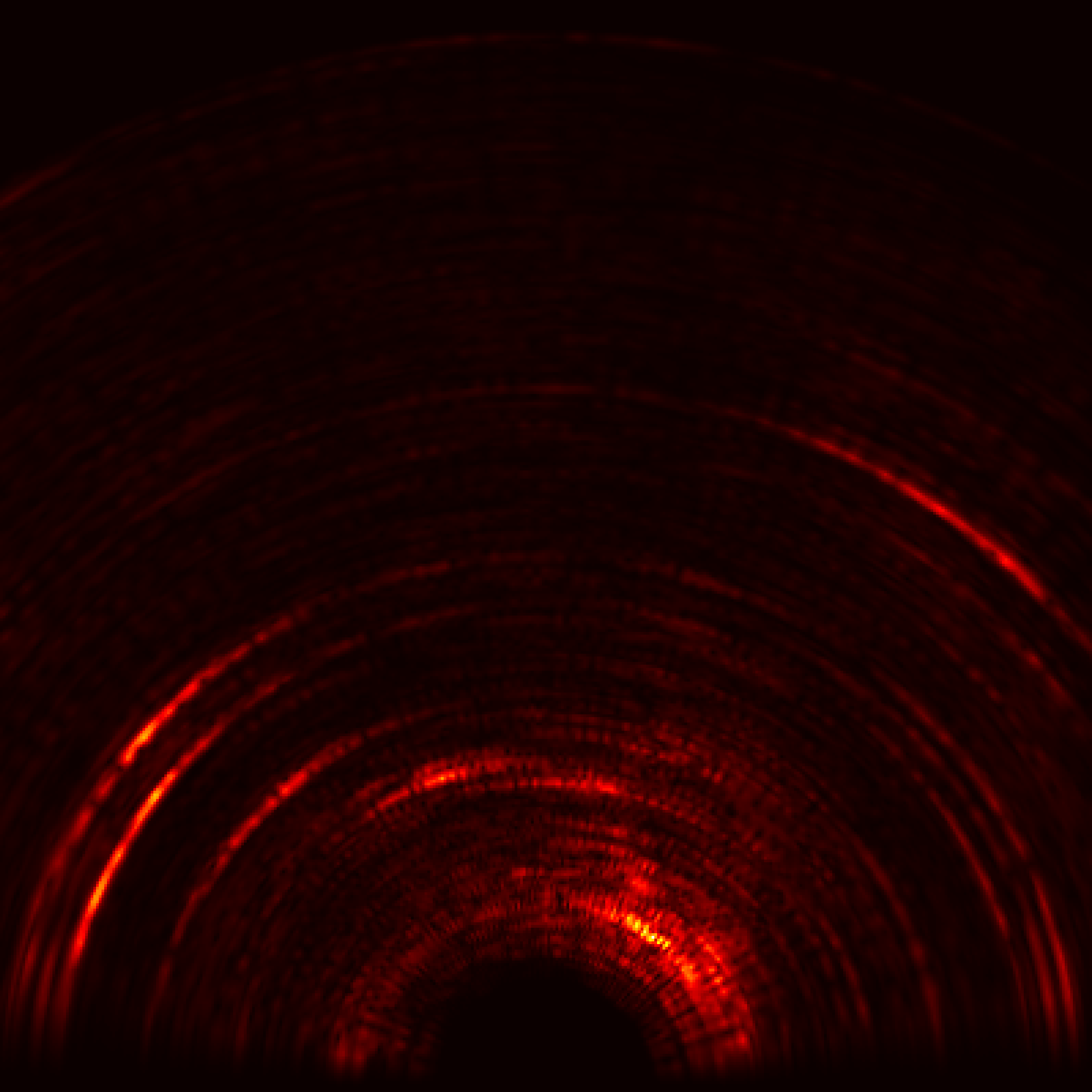} & \includegraphics[width=0.165\linewidth]{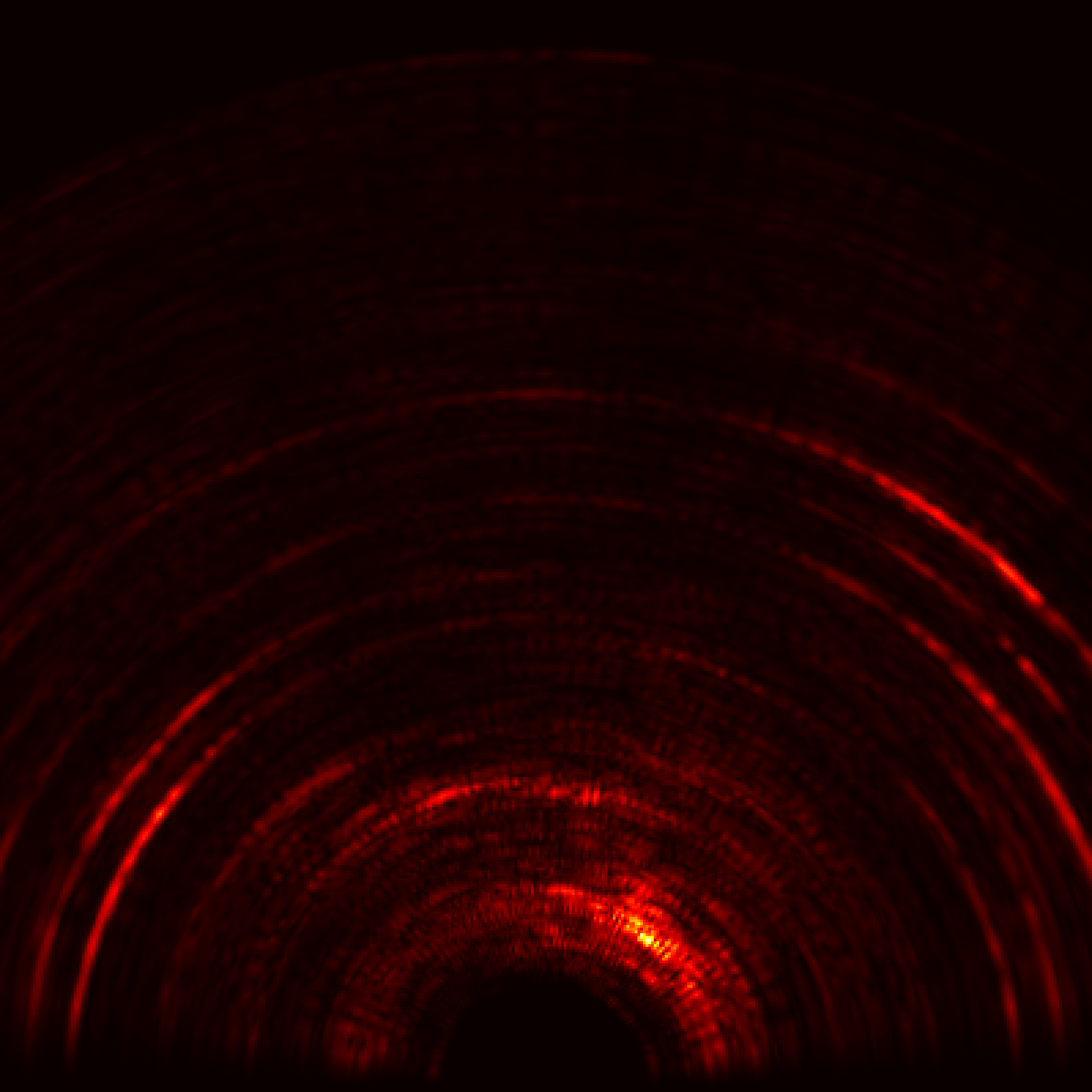} \\
\raisebox{2.0em}{\rotatebox{90}{\scriptsize\textbf{mmIR}}} & \includegraphics[width=0.165\linewidth]{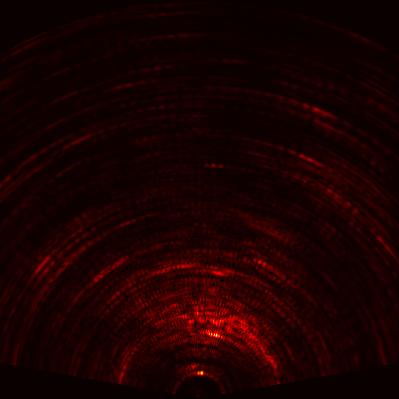} & \includegraphics[width=0.165\linewidth]{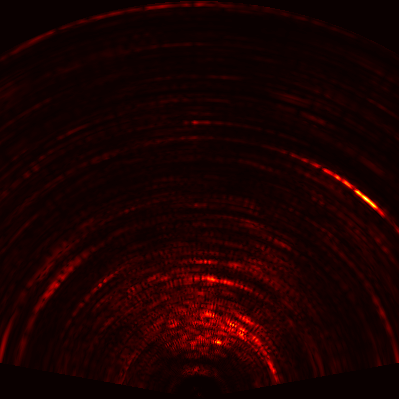} & \includegraphics[width=0.165\linewidth]{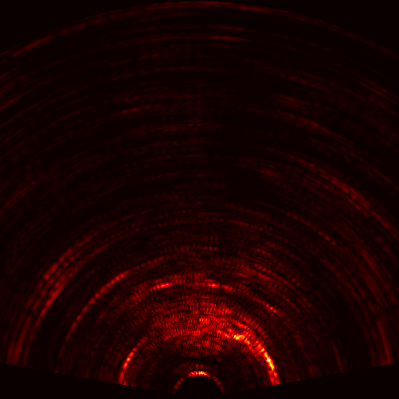} & \includegraphics[width=0.165\linewidth]{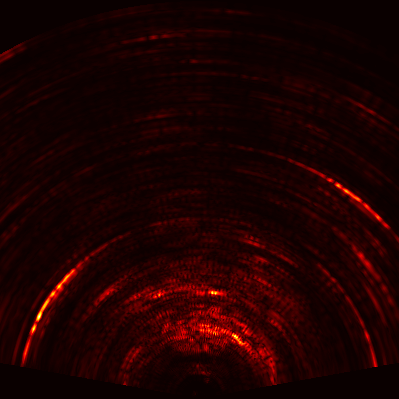} \\
\raisebox{2.0em}{\rotatebox{90}{\scriptsize\textbf{RadarSplat}}} & \includegraphics[width=0.165\linewidth]{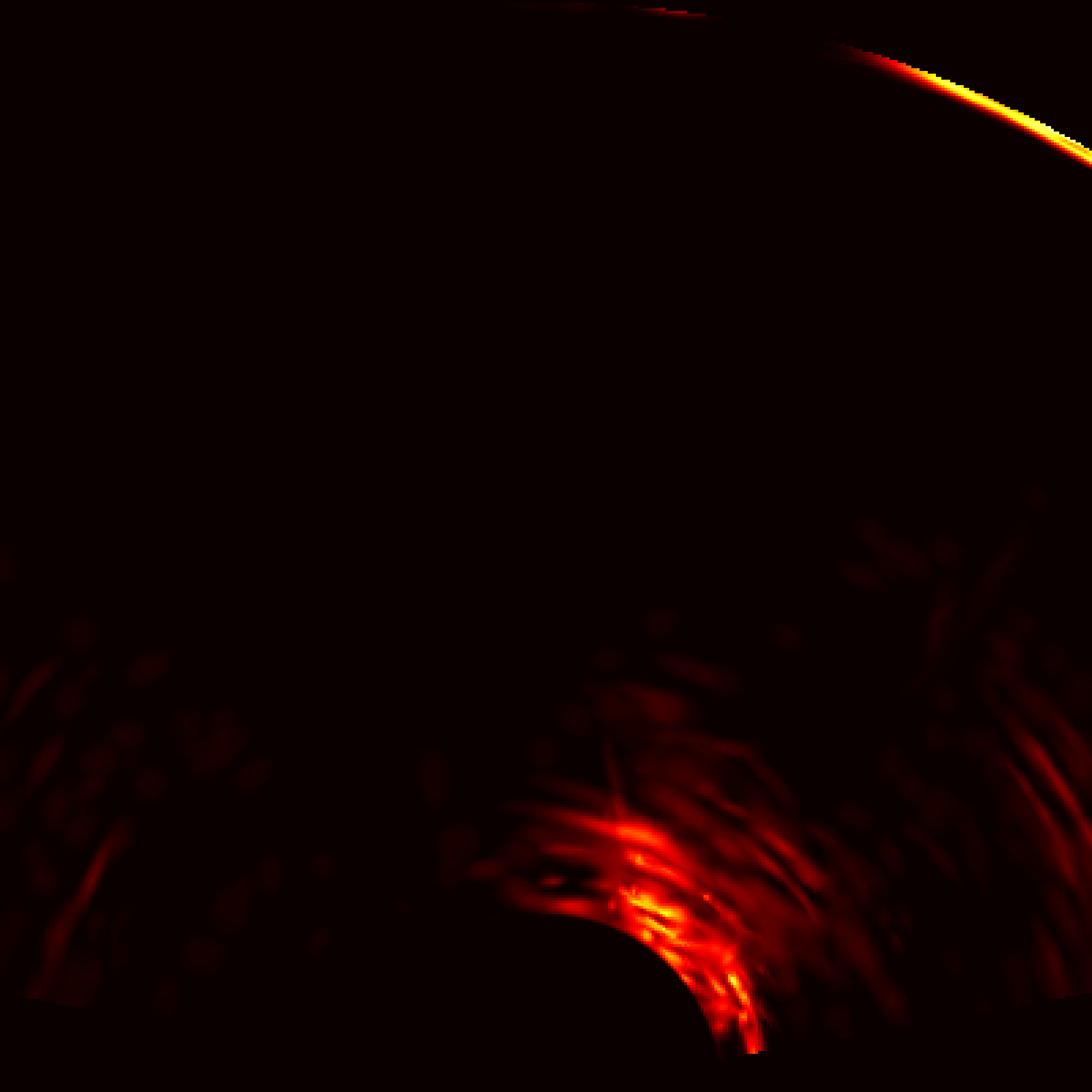} & \includegraphics[width=0.165\linewidth]{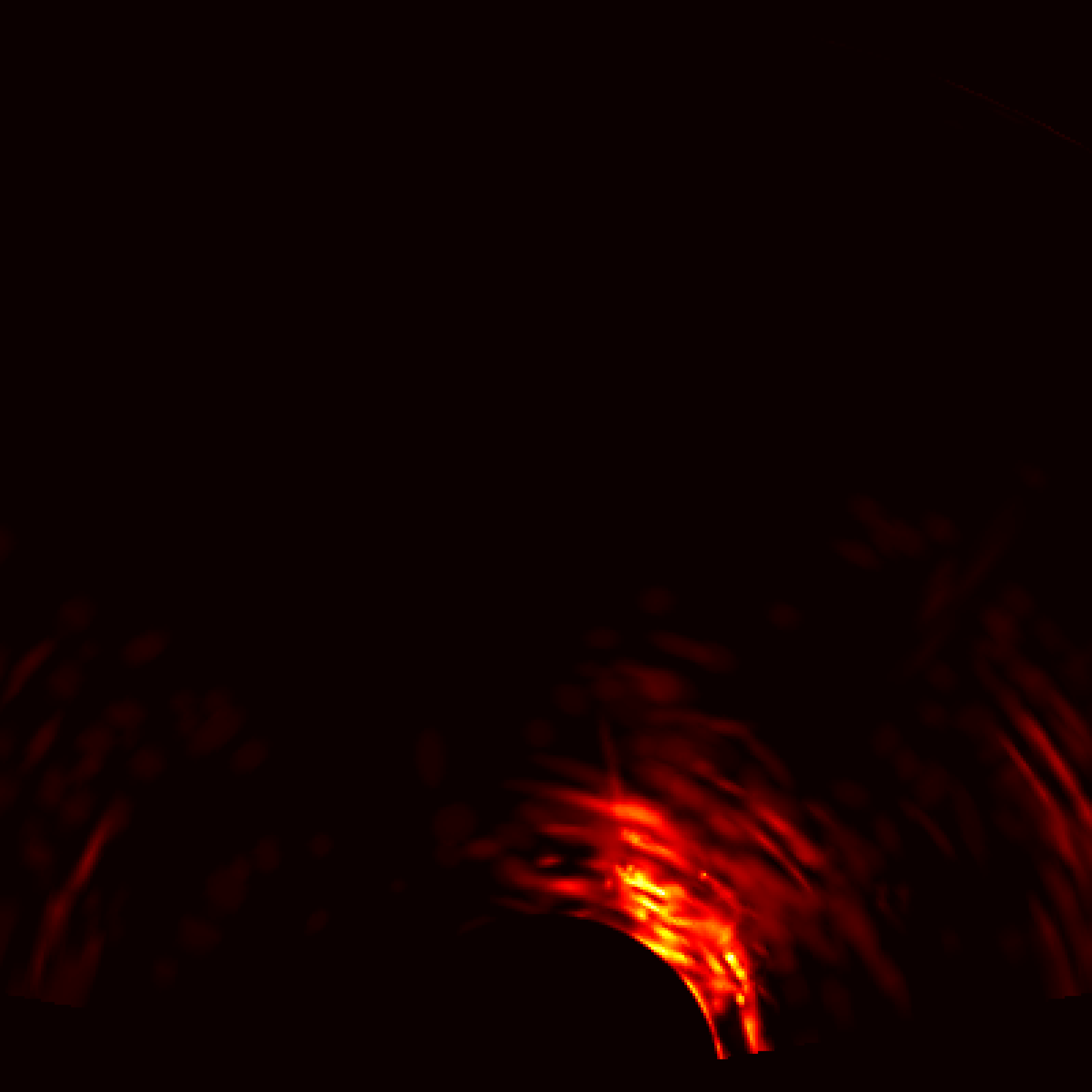} & \includegraphics[width=0.165\linewidth]{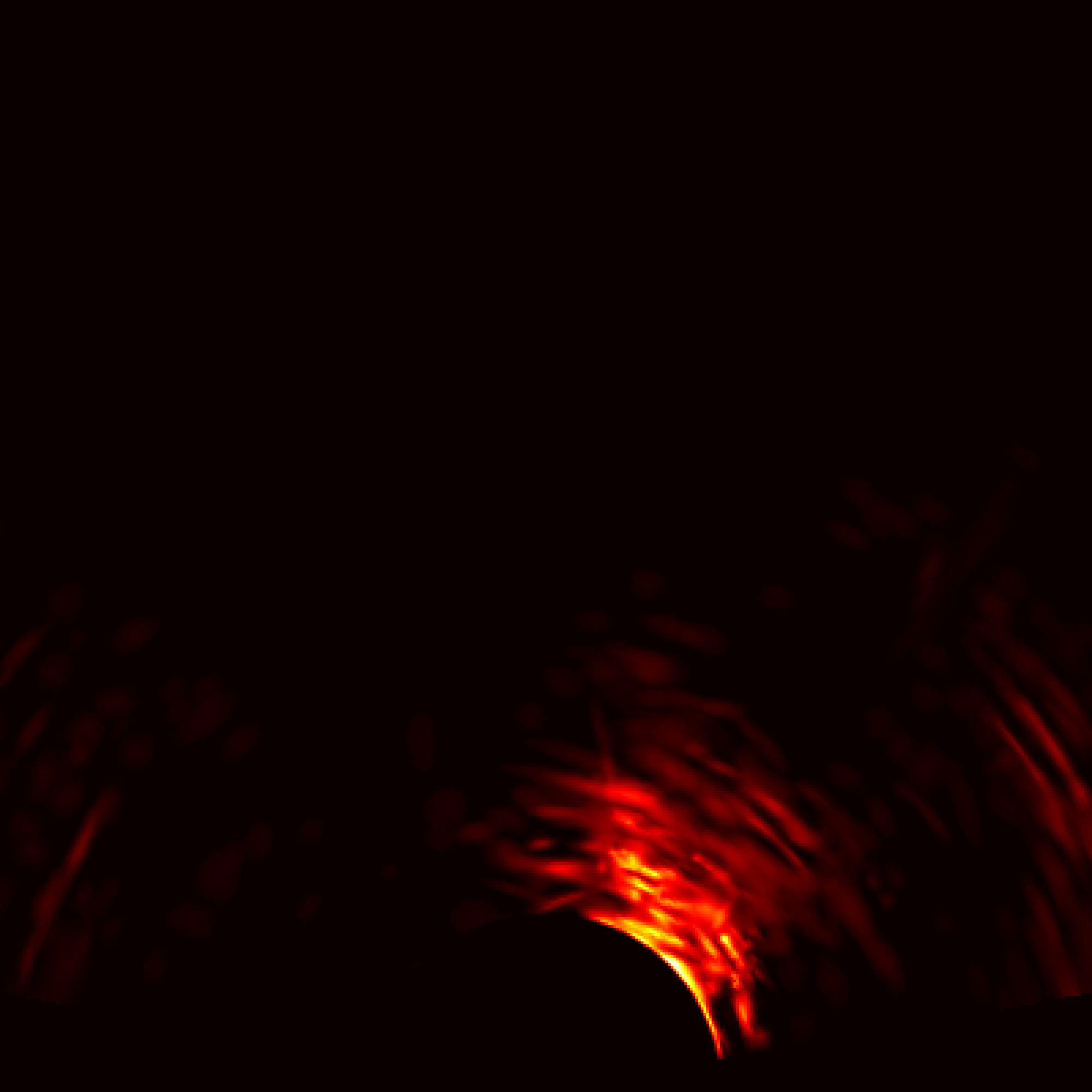} & \includegraphics[width=0.165\linewidth]{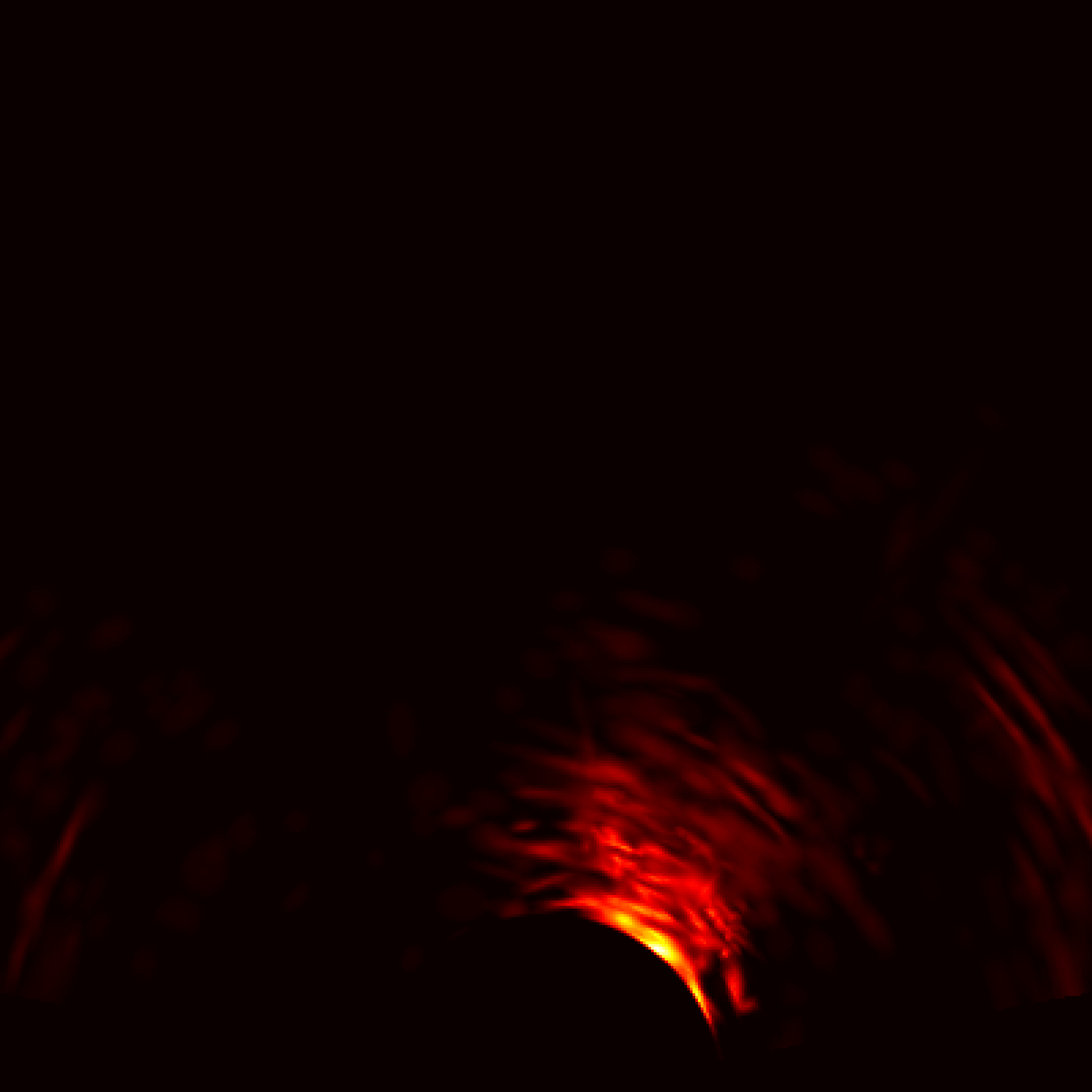} \\
\raisebox{2.0em}{\rotatebox{90}{\scriptsize\textbf{Radar Fields}}} & \includegraphics[width=0.165\linewidth]{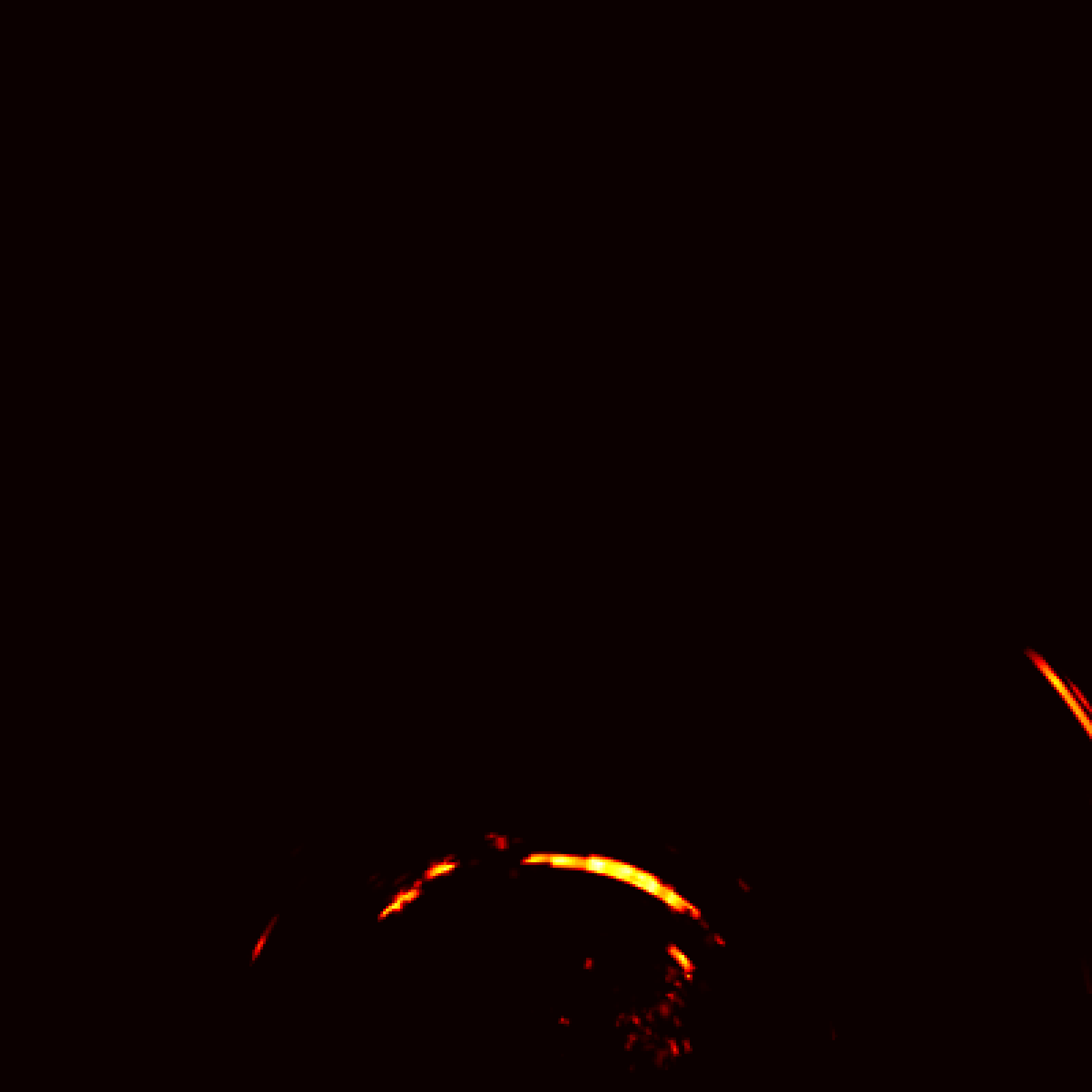} & \includegraphics[width=0.165\linewidth]{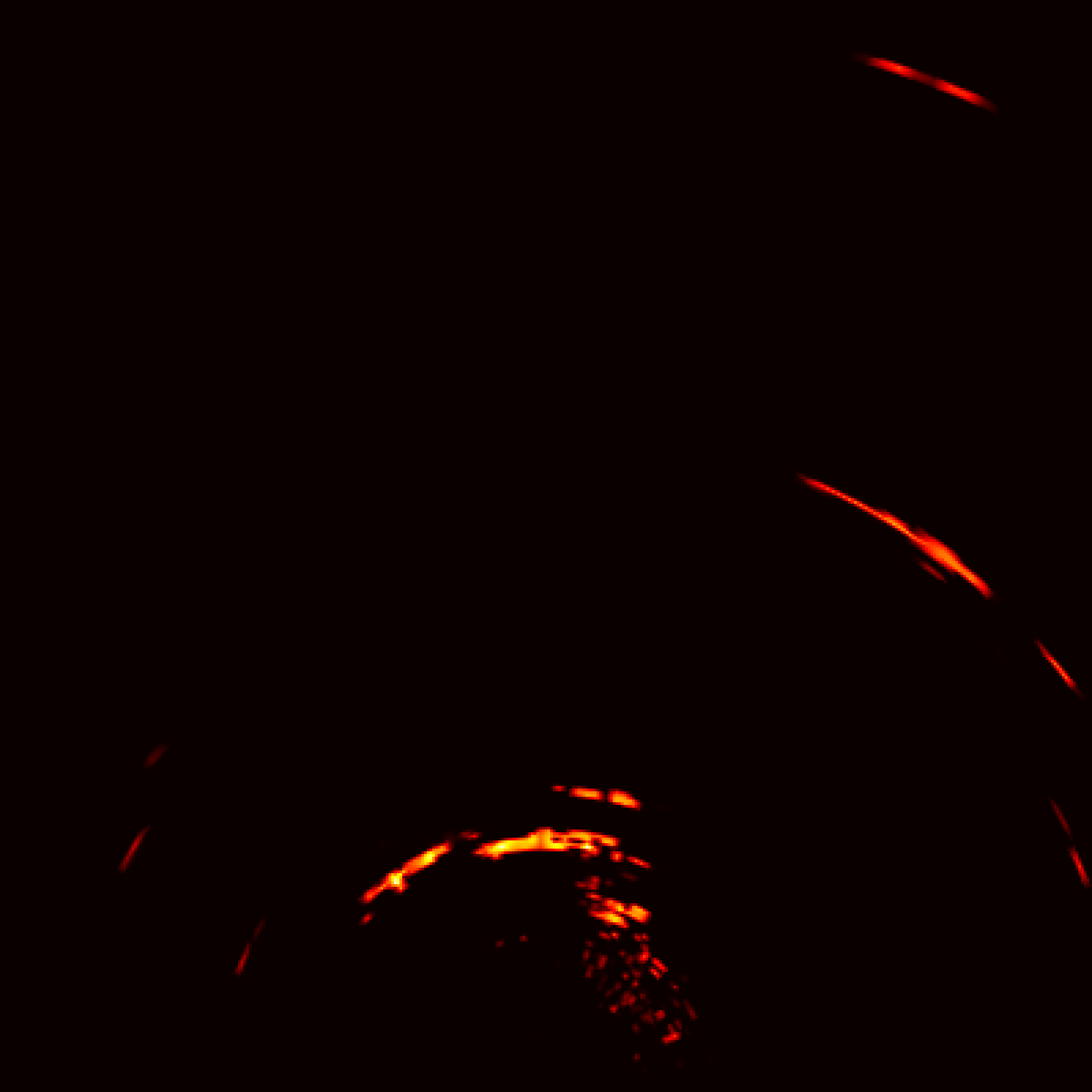} & \includegraphics[width=0.165\linewidth]{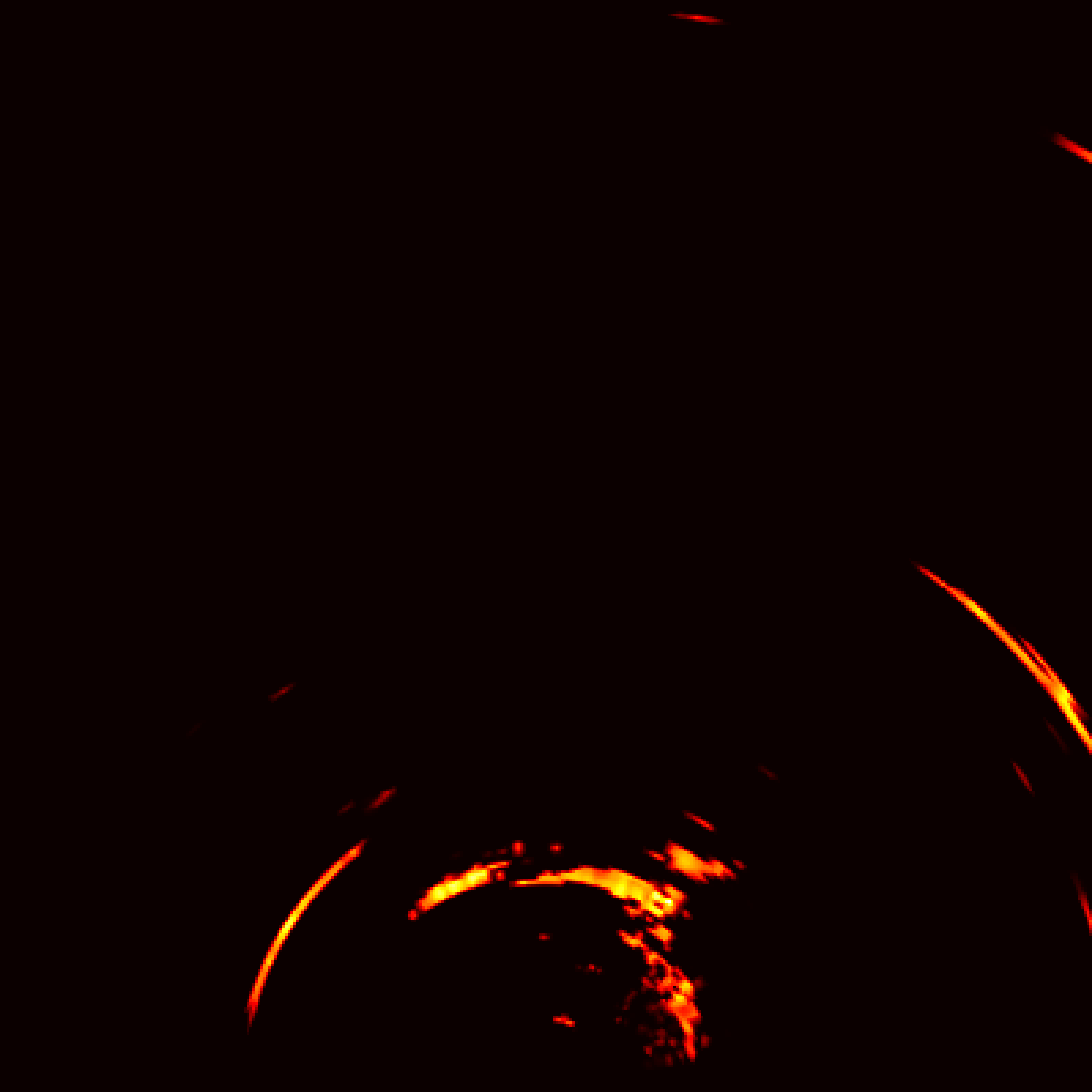} & \includegraphics[width=0.165\linewidth]{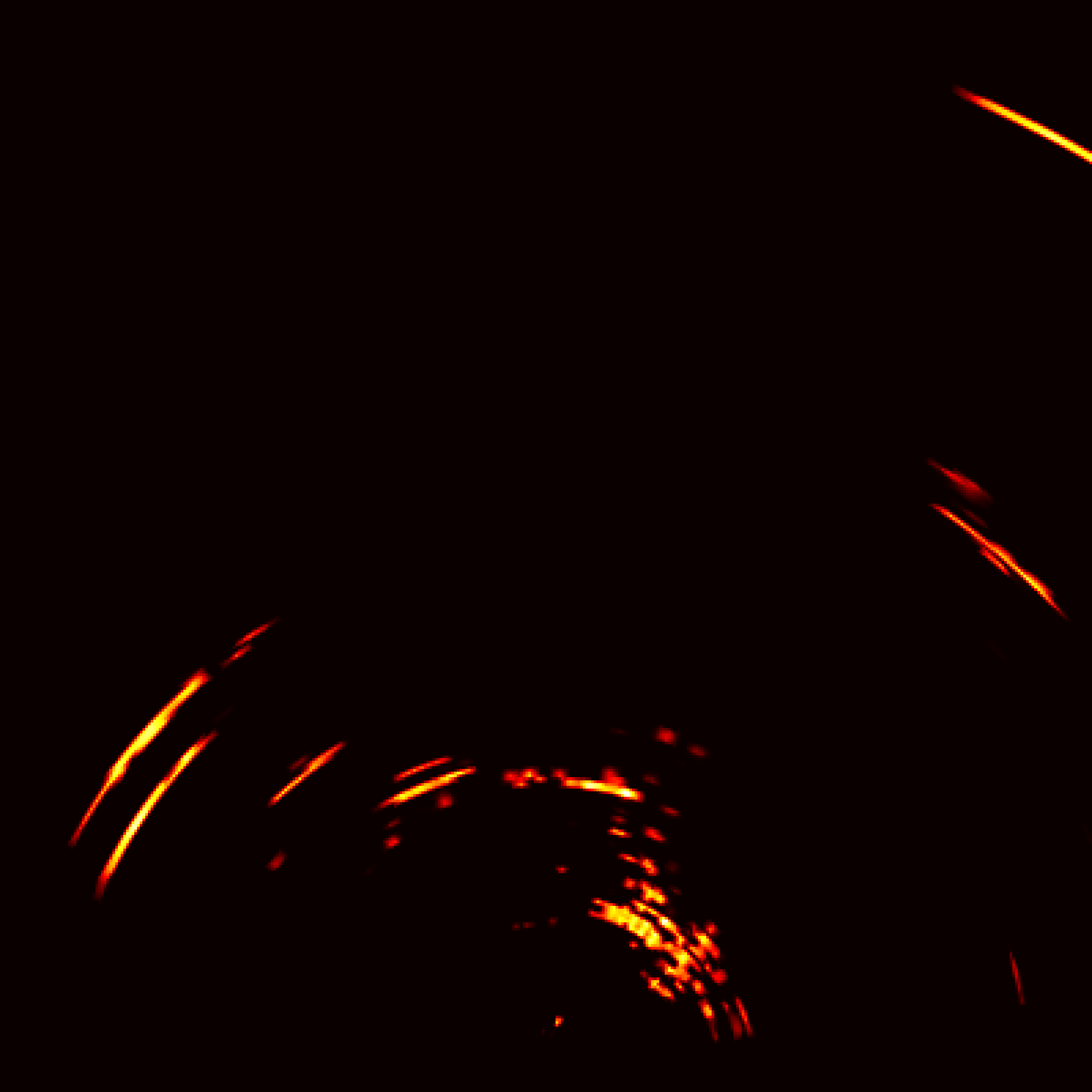} \\
\raisebox{2.0em}{\rotatebox{90}{\scriptsize\textbf{DART}}} & \includegraphics[width=0.165\linewidth]{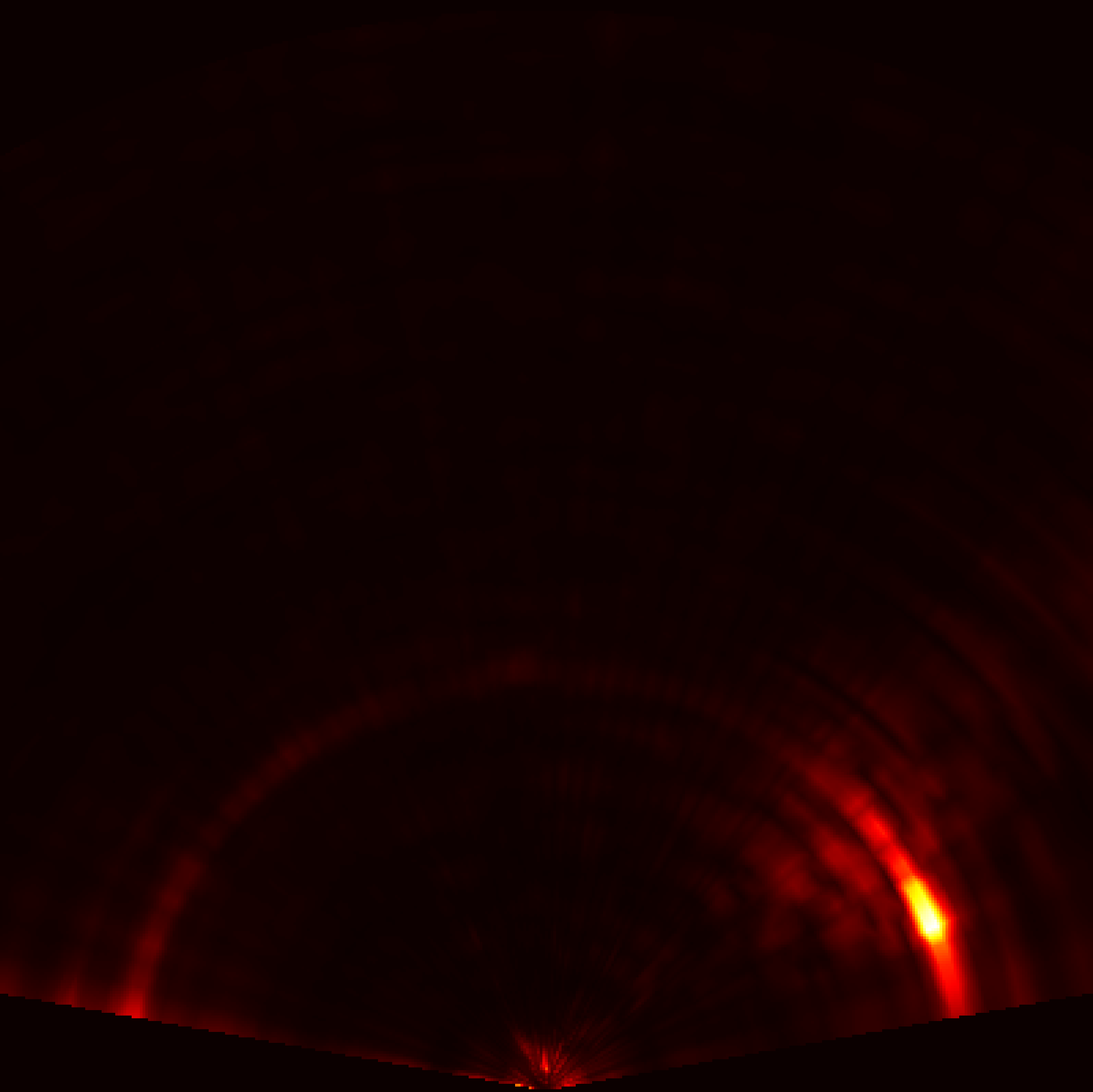} & \includegraphics[width=0.165\linewidth]{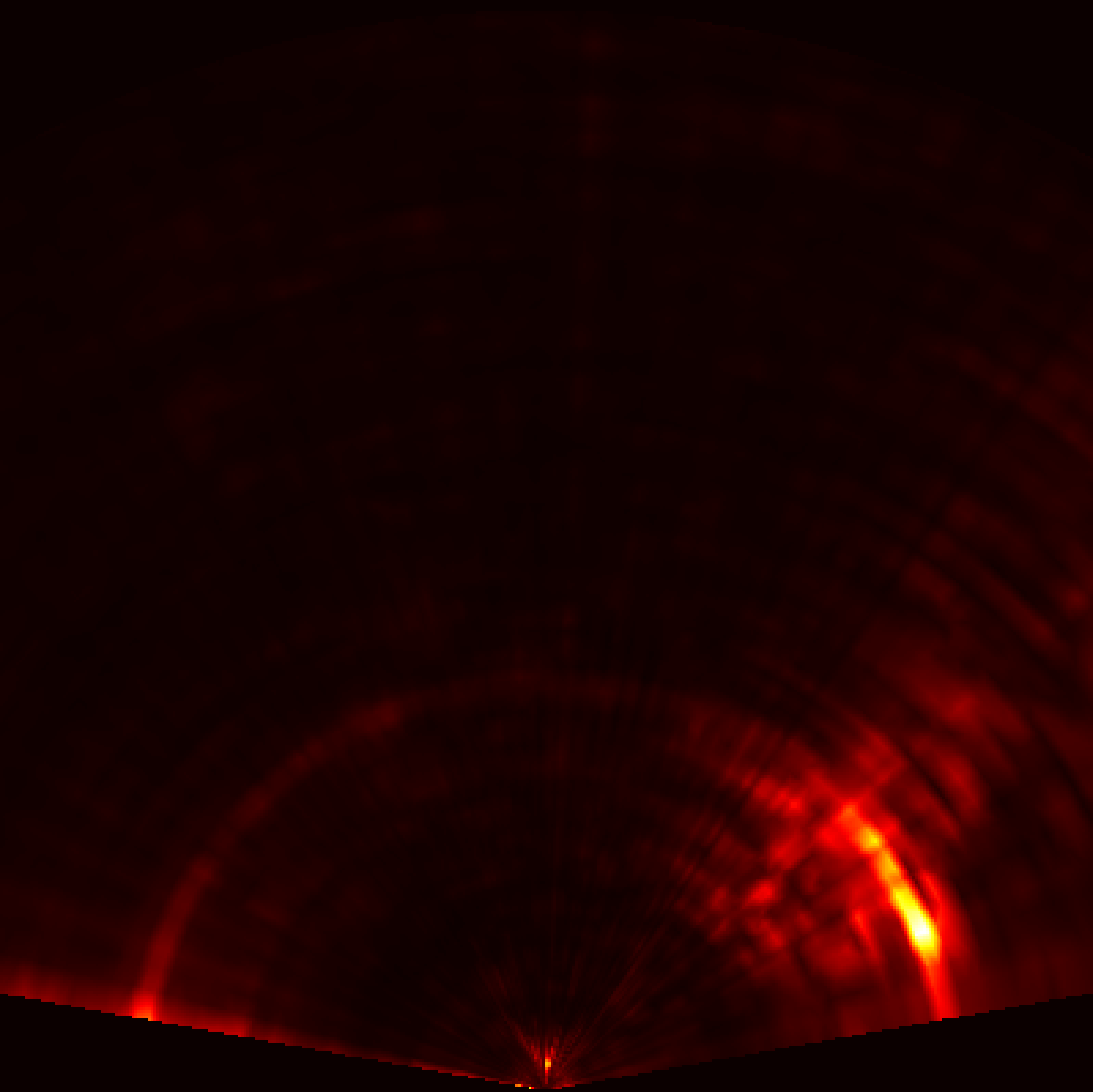} & \includegraphics[width=0.165\linewidth]{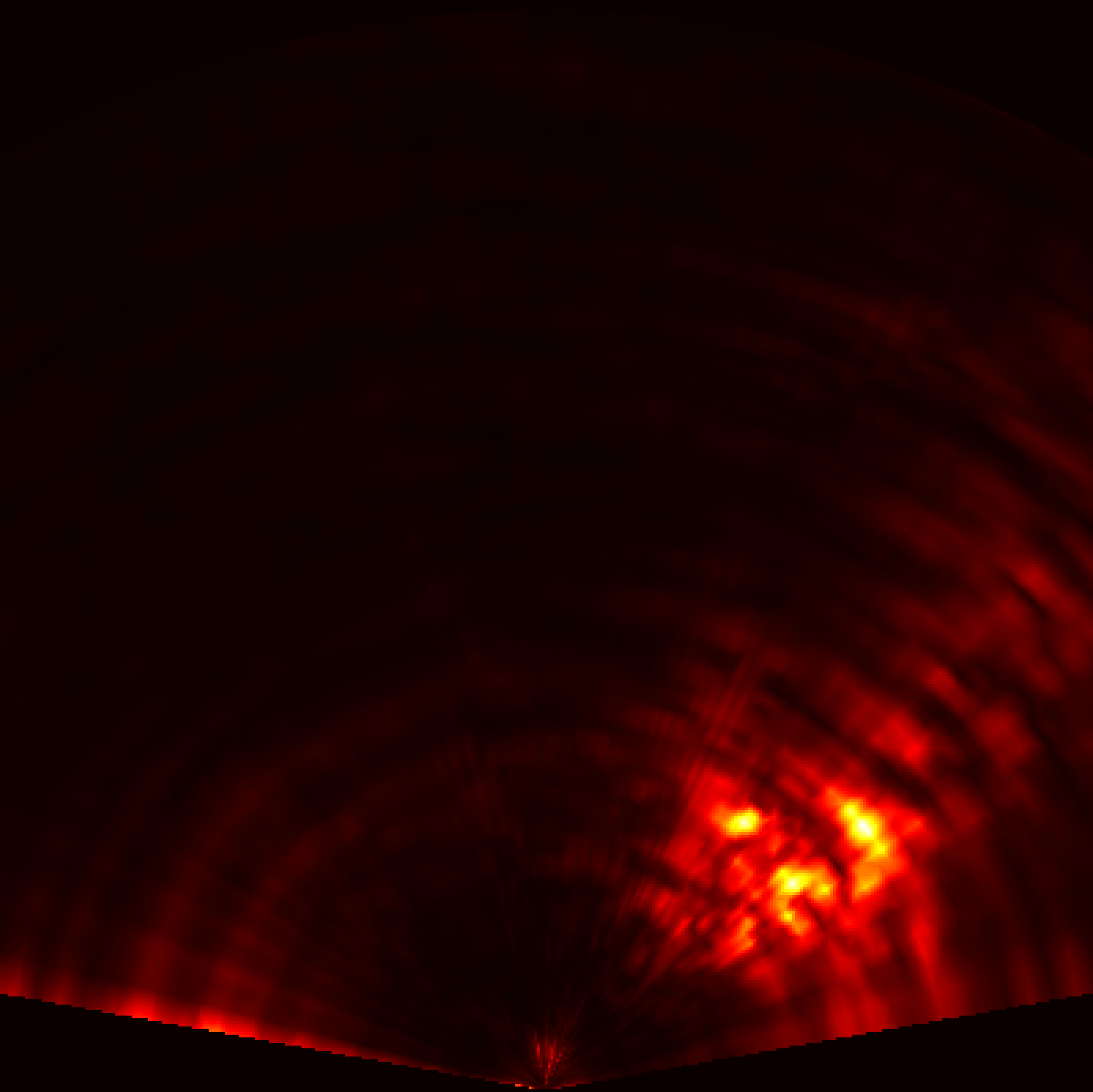} & \includegraphics[width=0.165\linewidth]{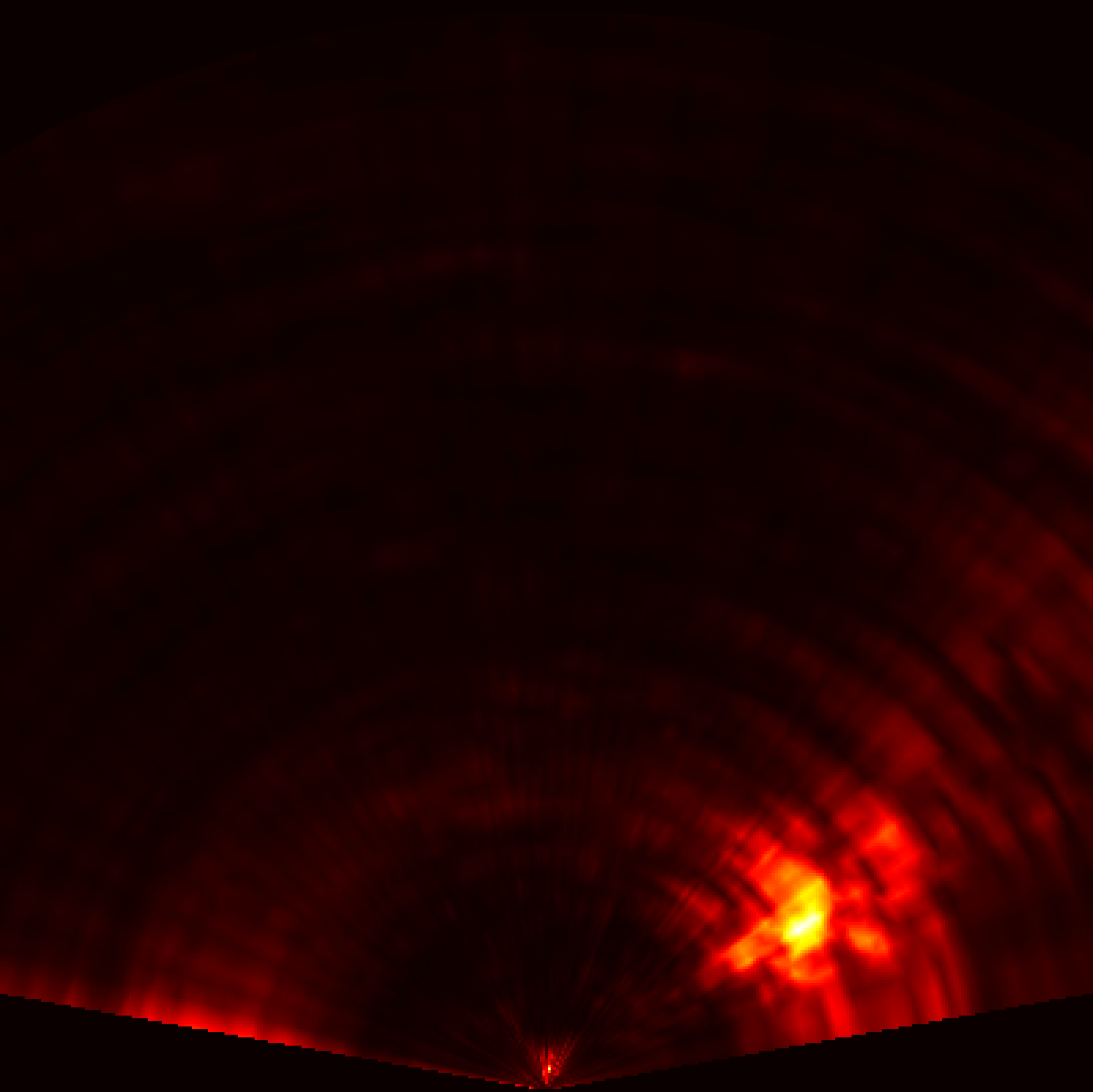} \\
\end{tabular}
\caption{\textbf{Per-frame $|\mathrm{RA}|$ comparison, scene S2\,F105, training frames $F\!-\!4$ to $F\!-\!1$.} Pearson correlation per column, left to right. 3DPS: 0.74, 0.78, 0.80, 0.80; mmIR: 0.51, 0.57, 0.58, 0.54; RadarSplat: 0.51, 0.33, 0.48, 0.50; Radar Fields: 0.12, 0.09, 0.20, 0.37; DART: 0.05, 0.10, 0.18, 0.14.}
\label{fig:supp_s2_f105}
\end{figure}

\begin{figure}[p]
\centering
\setlength{\tabcolsep}{1pt}\renewcommand{\arraystretch}{0.6}
\begin{tabular}{cccccc}
 & {\footnotesize $F\!+\!1$} & {\footnotesize $F\!+\!2$} & {\footnotesize $F\!+\!3$} & {\footnotesize $F\!+\!4$} & \textcolor{red}{\footnotesize\textbf{$F$}} \\
\raisebox{2.0em}{\rotatebox{90}{\scriptsize\textbf{GT}}} & \includegraphics[width=0.165\linewidth]{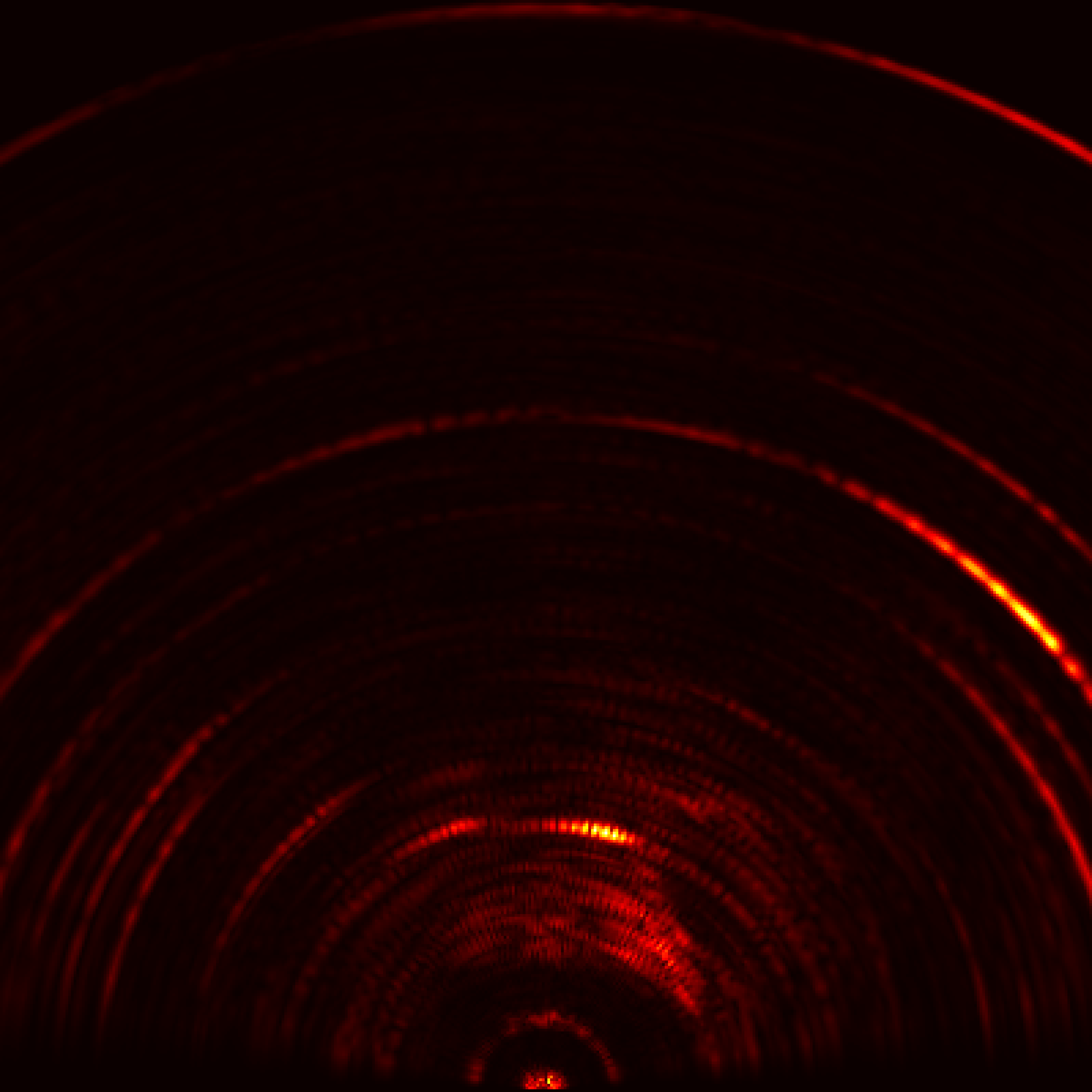} & \includegraphics[width=0.165\linewidth]{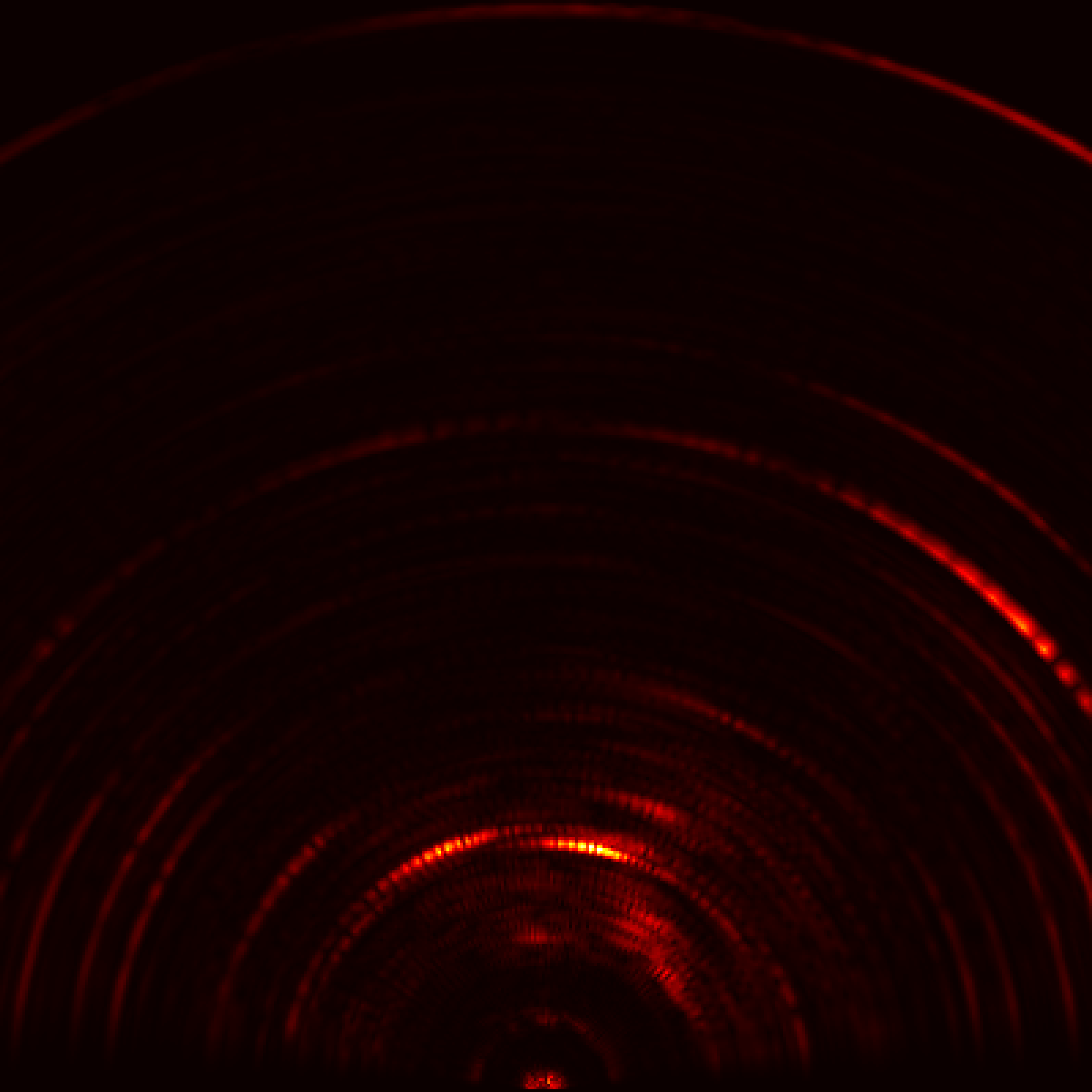} & \includegraphics[width=0.165\linewidth]{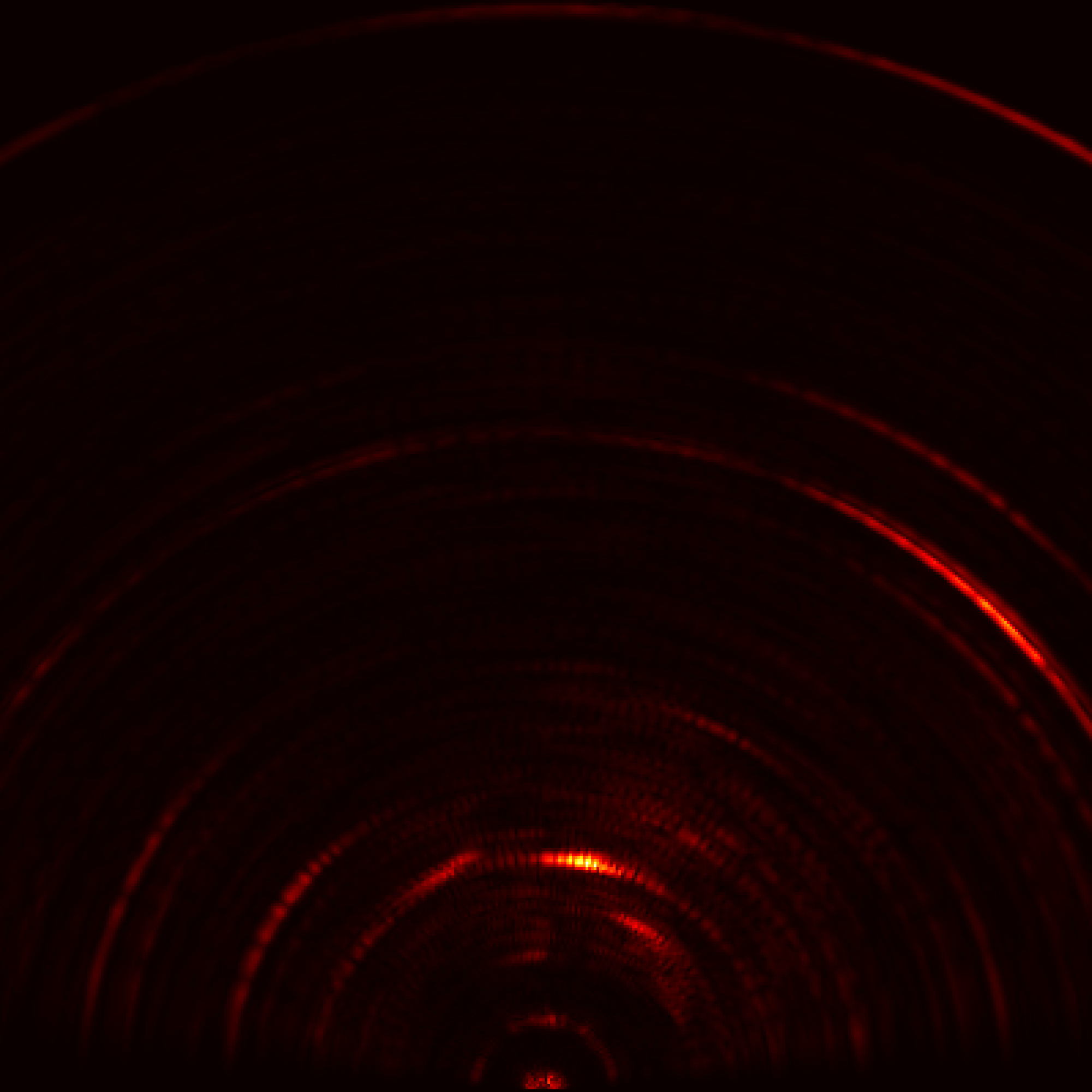} & \includegraphics[width=0.165\linewidth]{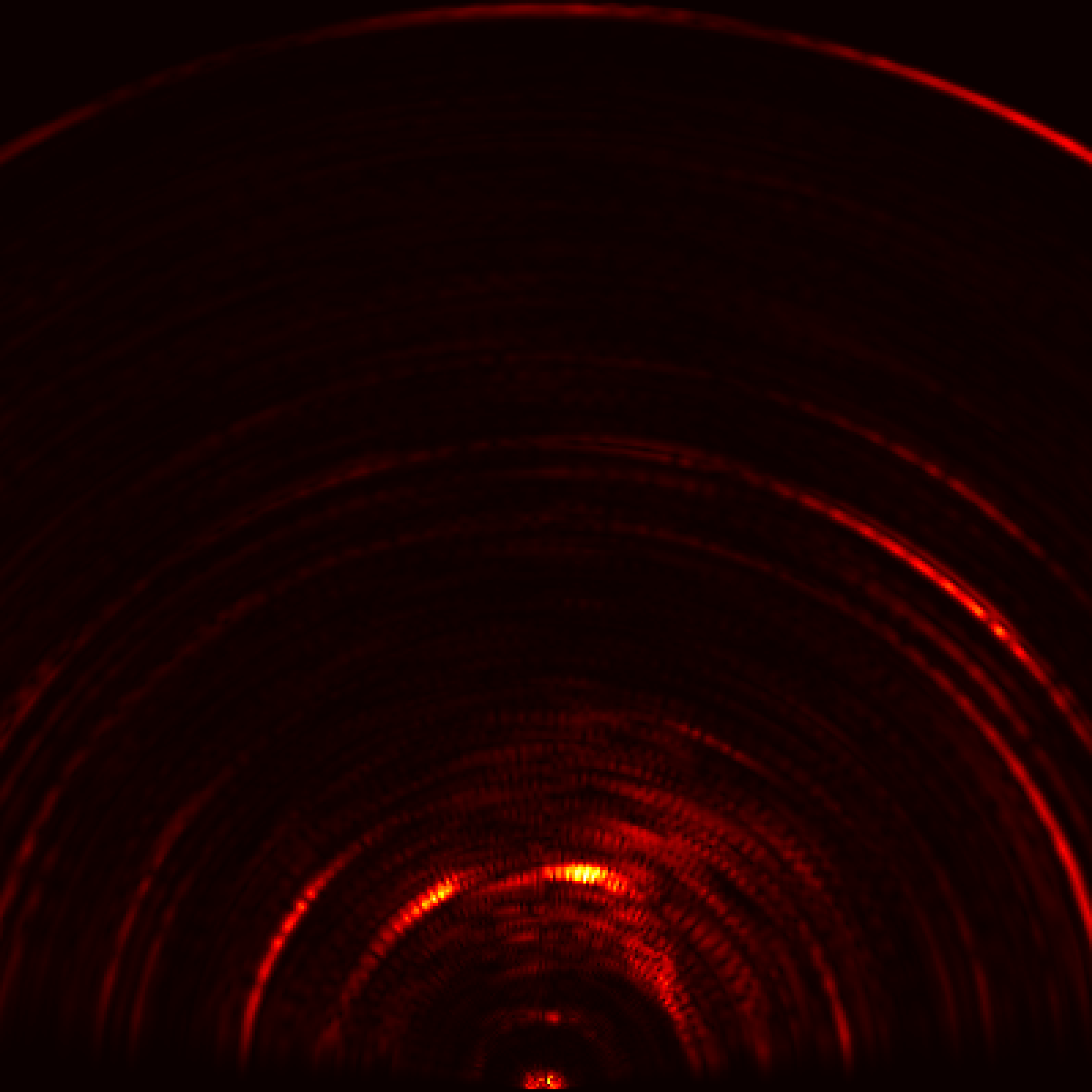} & \includegraphics[width=0.165\linewidth]{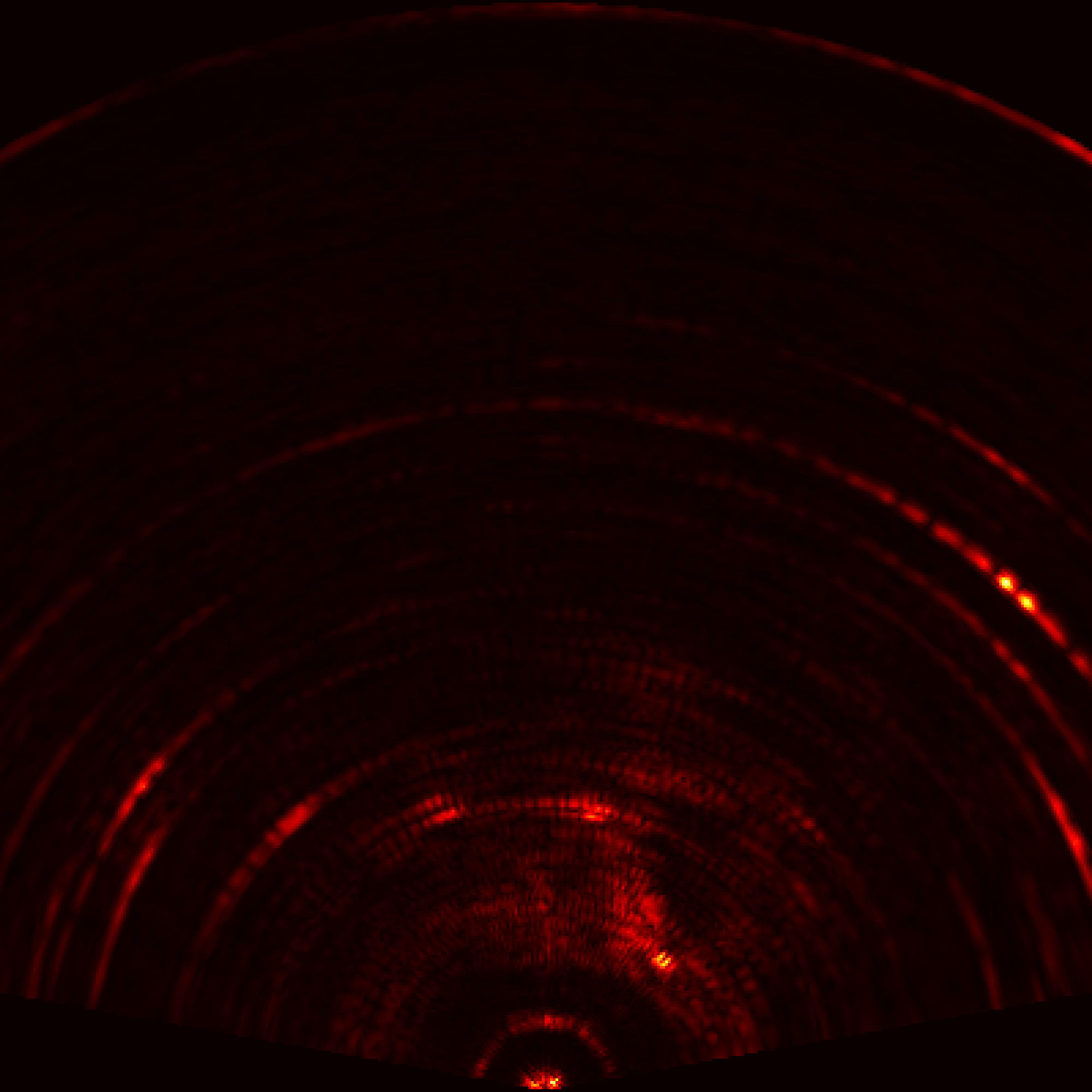} \\
\raisebox{2.0em}{\rotatebox{90}{\scriptsize\textbf{3DPS (Ours)}}} & \includegraphics[width=0.165\linewidth]{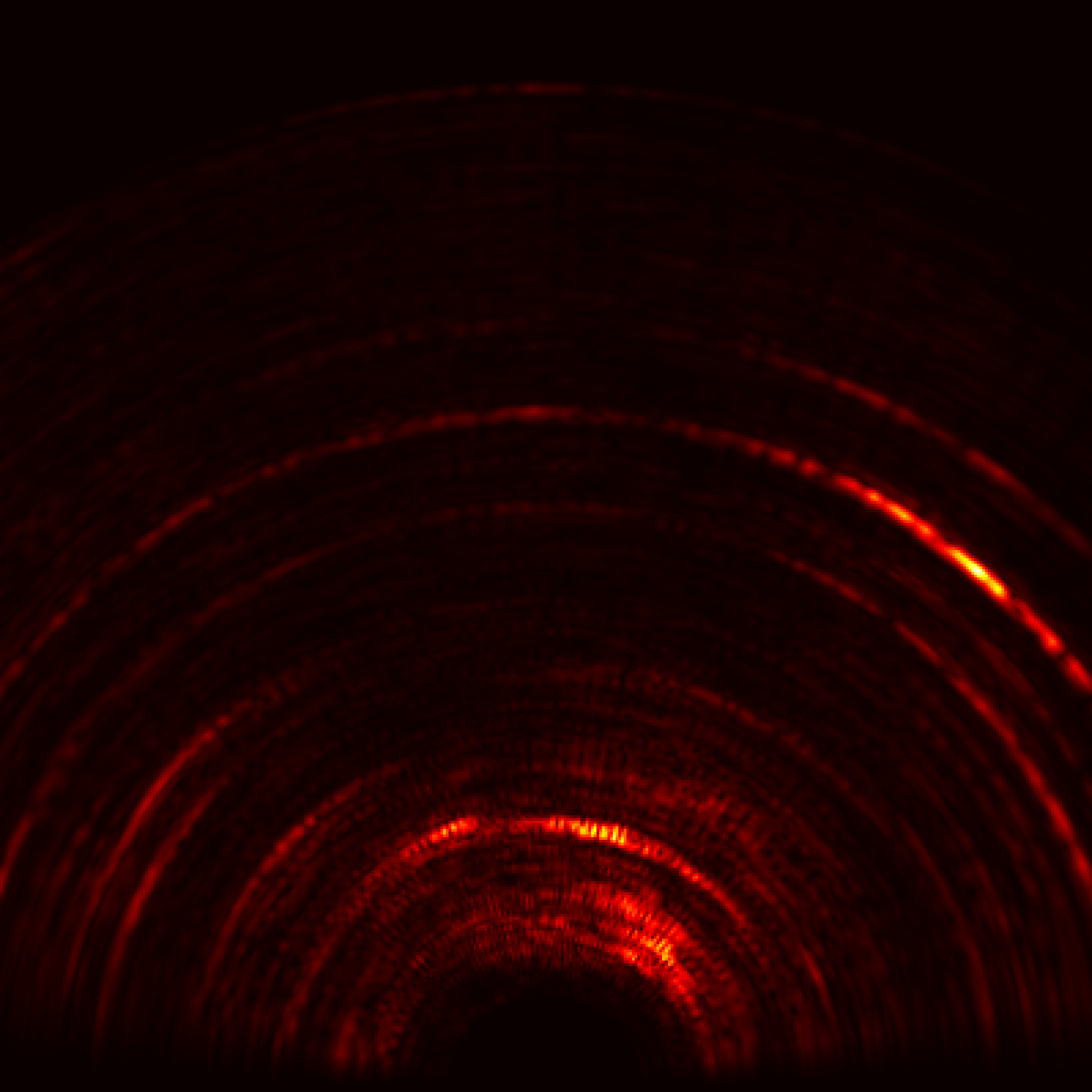} & \includegraphics[width=0.165\linewidth]{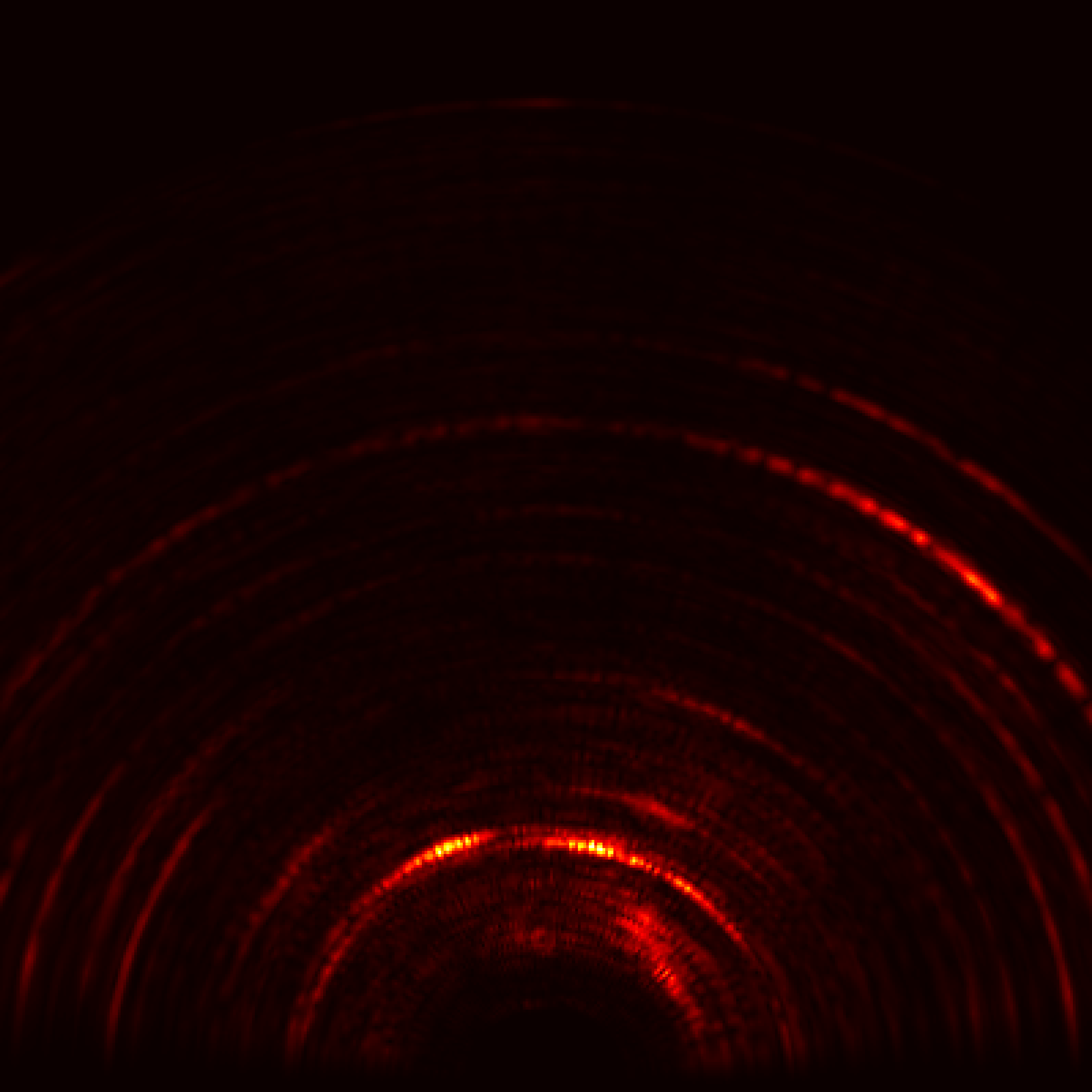} & \includegraphics[width=0.165\linewidth]{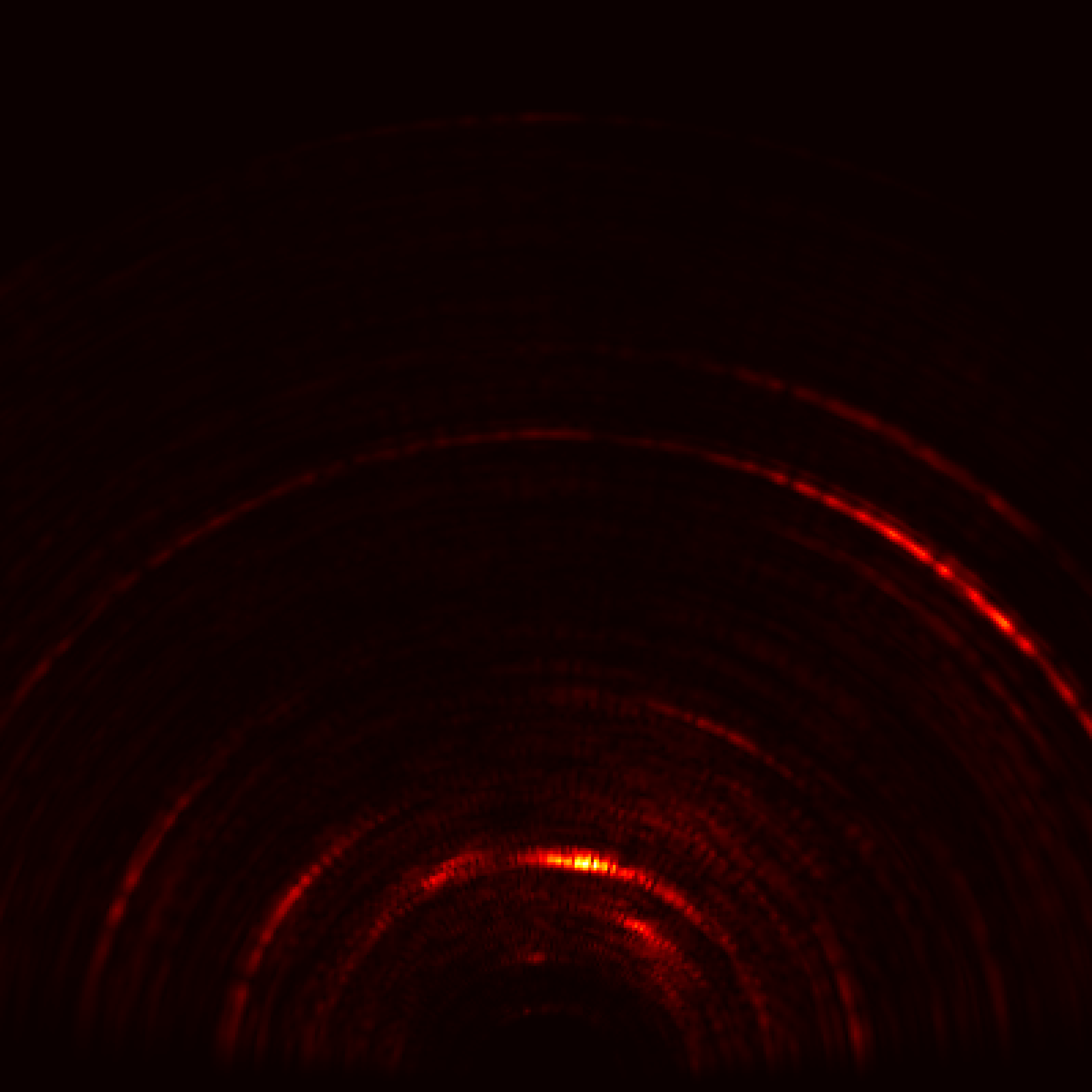} & \includegraphics[width=0.165\linewidth]{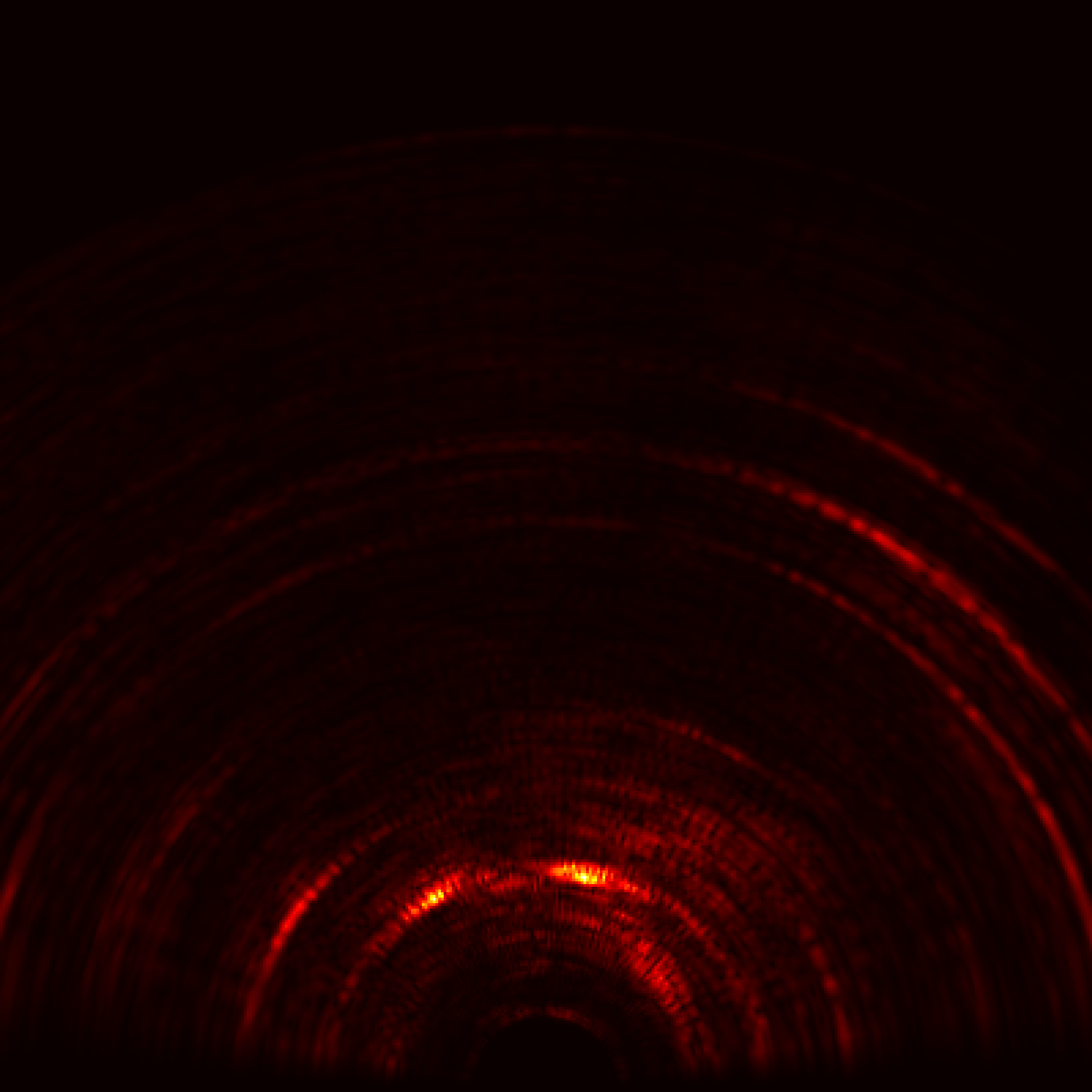} & \includegraphics[width=0.165\linewidth]{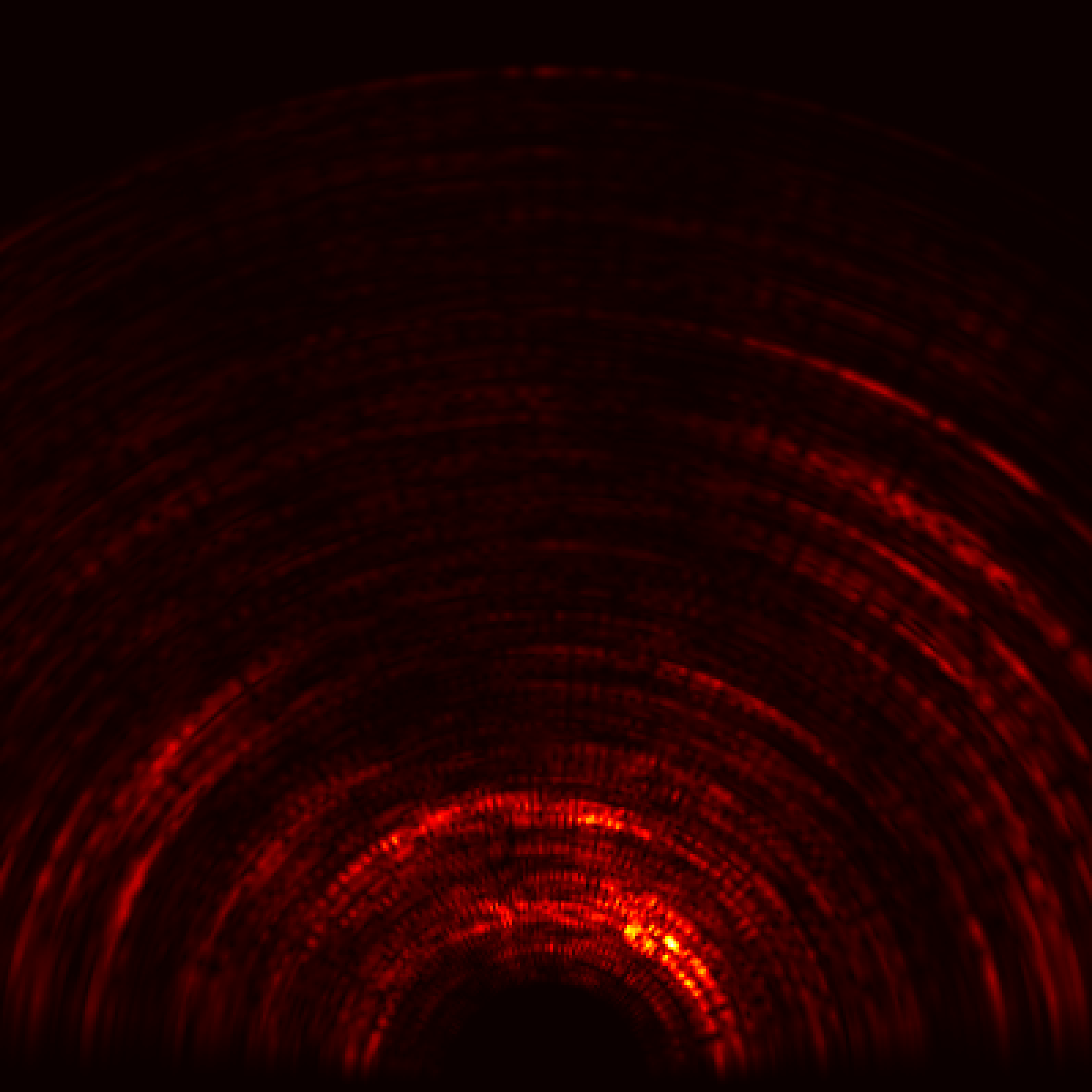} \\
\raisebox{2.0em}{\rotatebox{90}{\scriptsize\textbf{mmIR}}} & \includegraphics[width=0.165\linewidth]{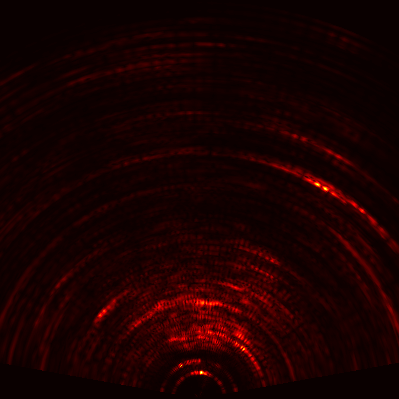} & \includegraphics[width=0.165\linewidth]{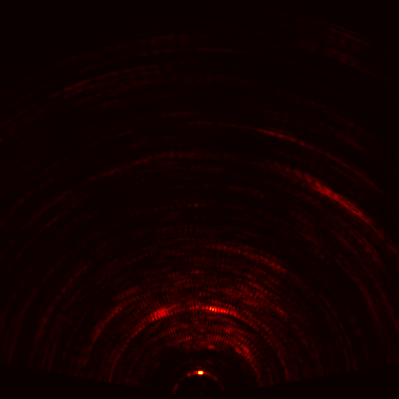} & \includegraphics[width=0.165\linewidth]{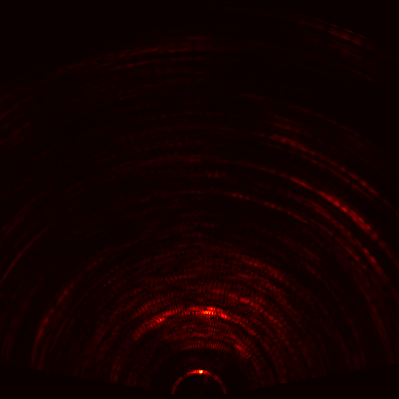} & \includegraphics[width=0.165\linewidth]{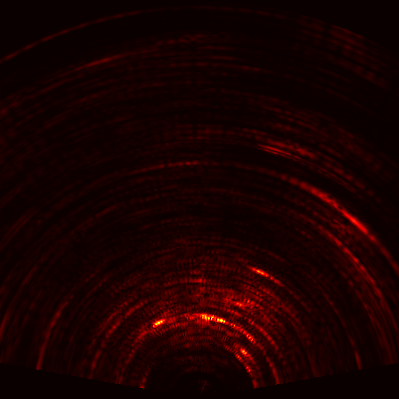} & \includegraphics[width=0.165\linewidth]{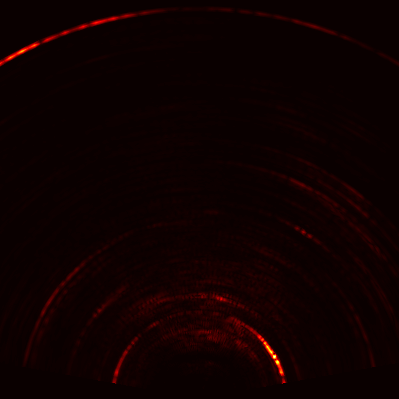} \\
\raisebox{2.0em}{\rotatebox{90}{\scriptsize\textbf{RadarSplat}}} & \includegraphics[width=0.165\linewidth]{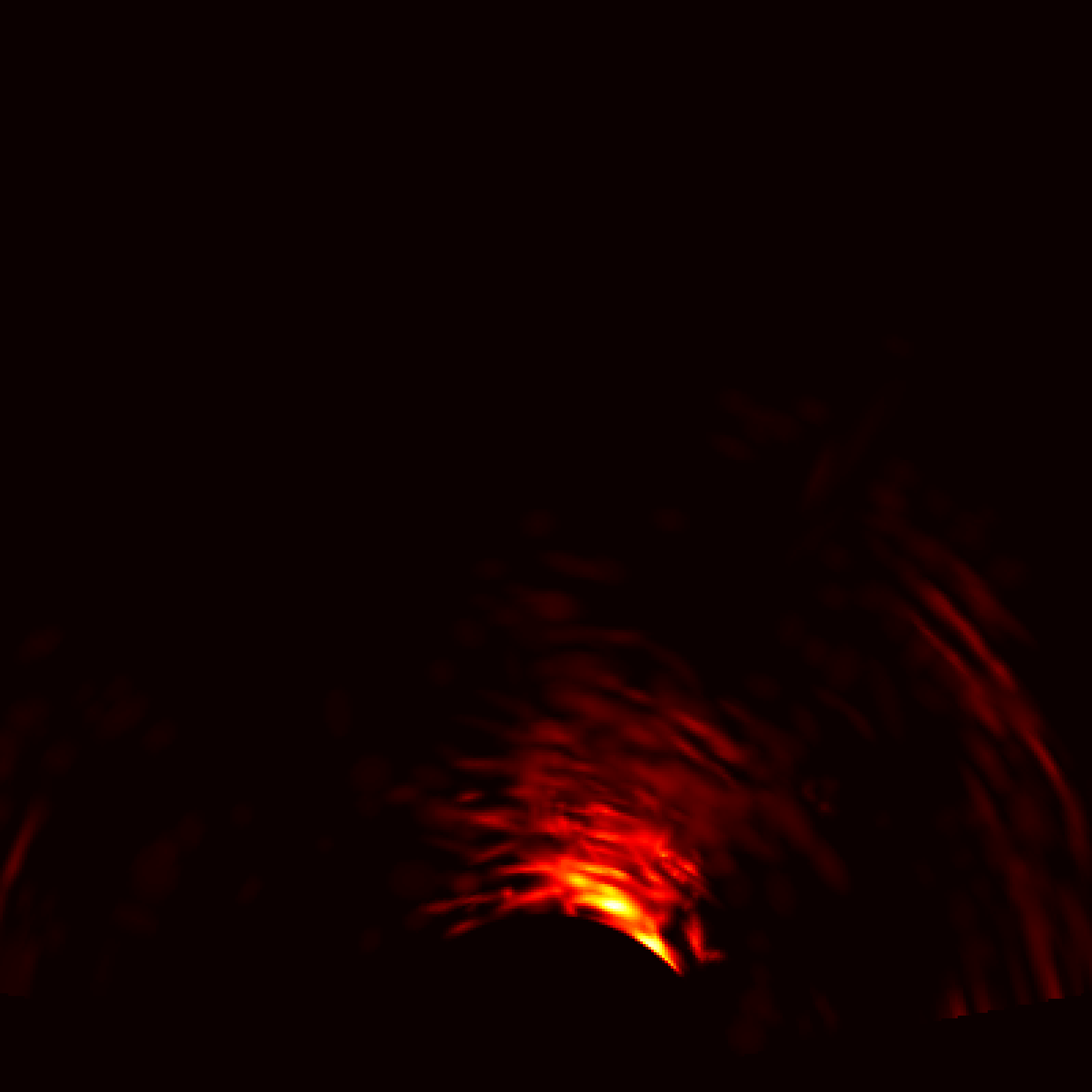} & \includegraphics[width=0.165\linewidth]{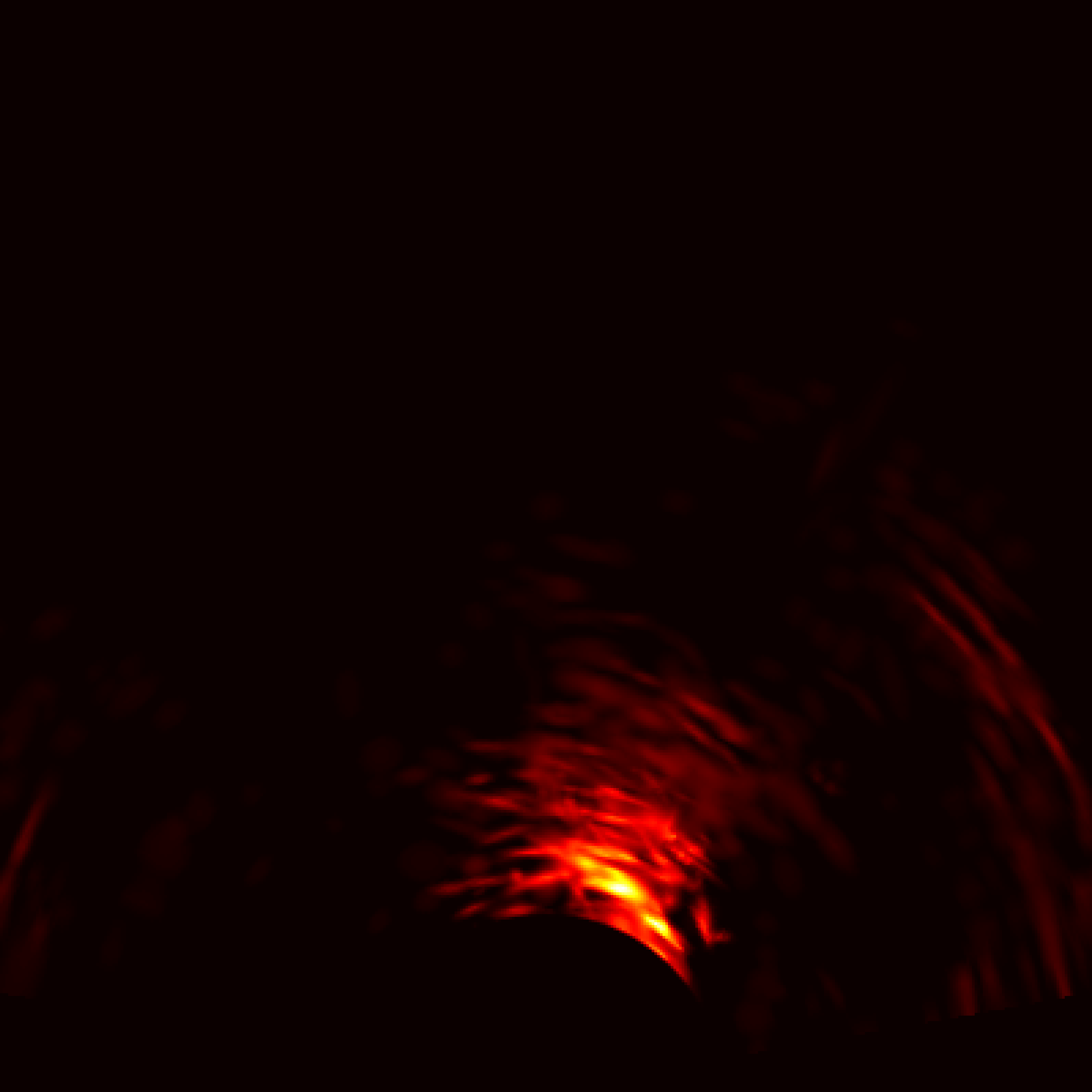} & \includegraphics[width=0.165\linewidth]{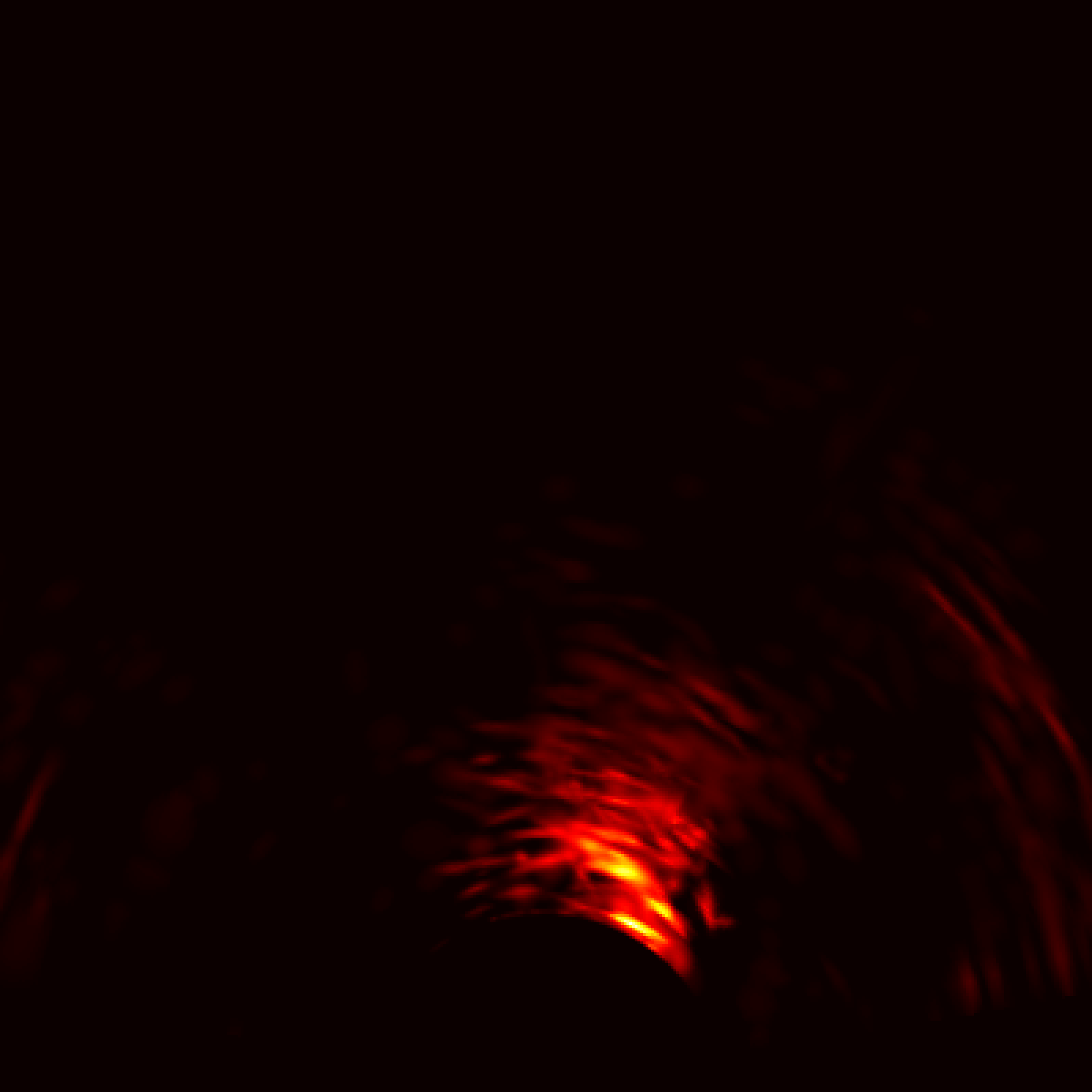} & \includegraphics[width=0.165\linewidth]{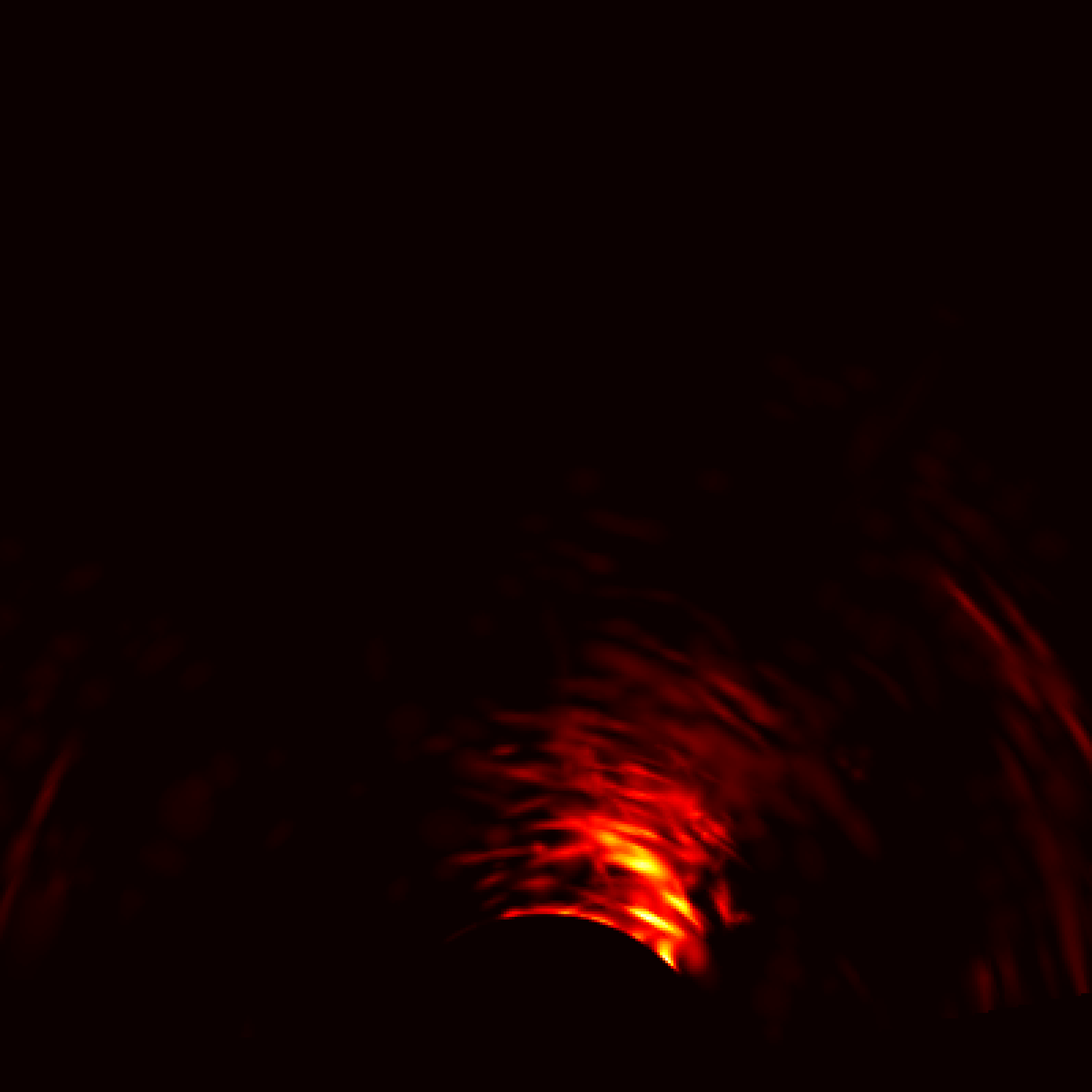} & \includegraphics[width=0.165\linewidth]{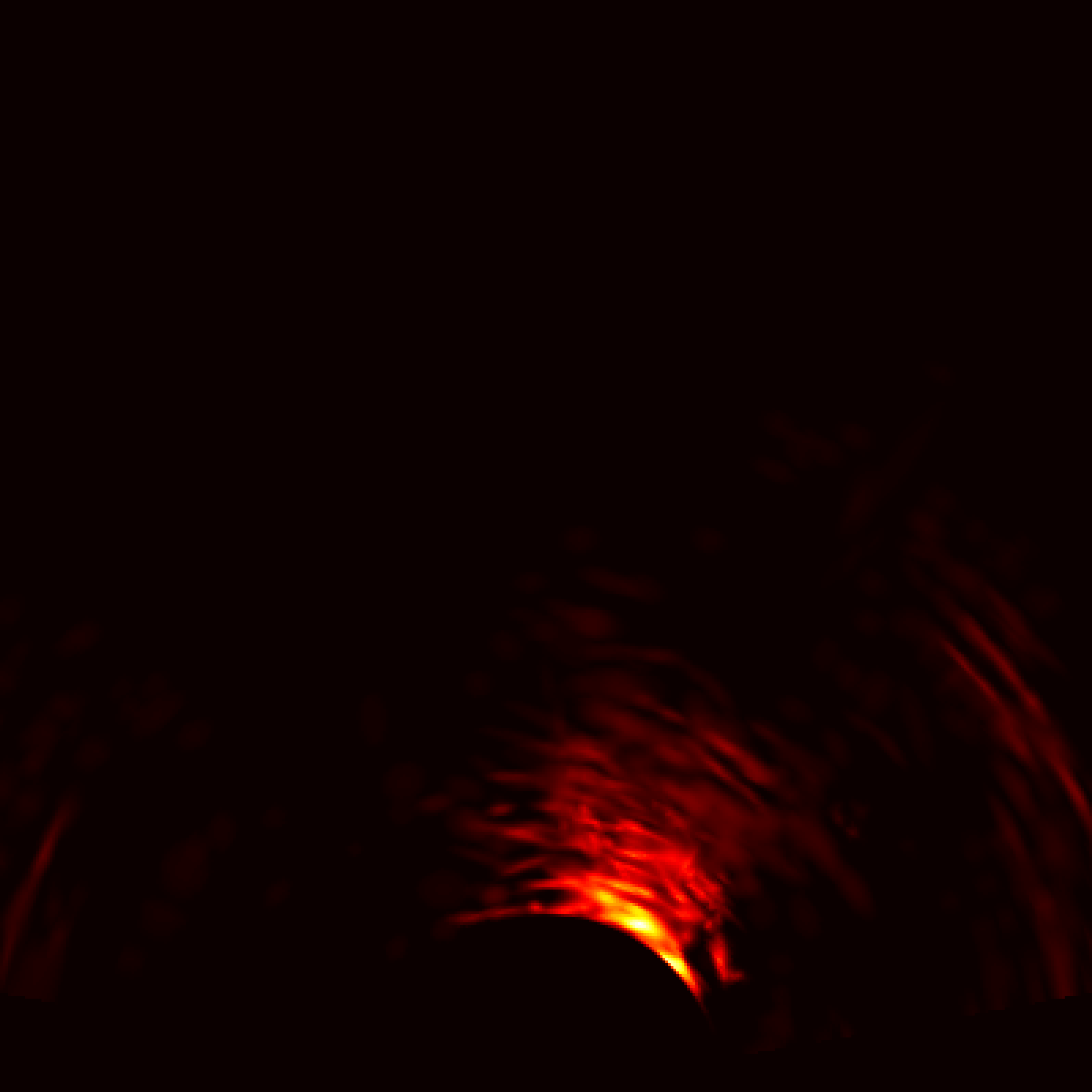} \\
\raisebox{2.0em}{\rotatebox{90}{\scriptsize\textbf{Radar Fields}}} & \includegraphics[width=0.165\linewidth]{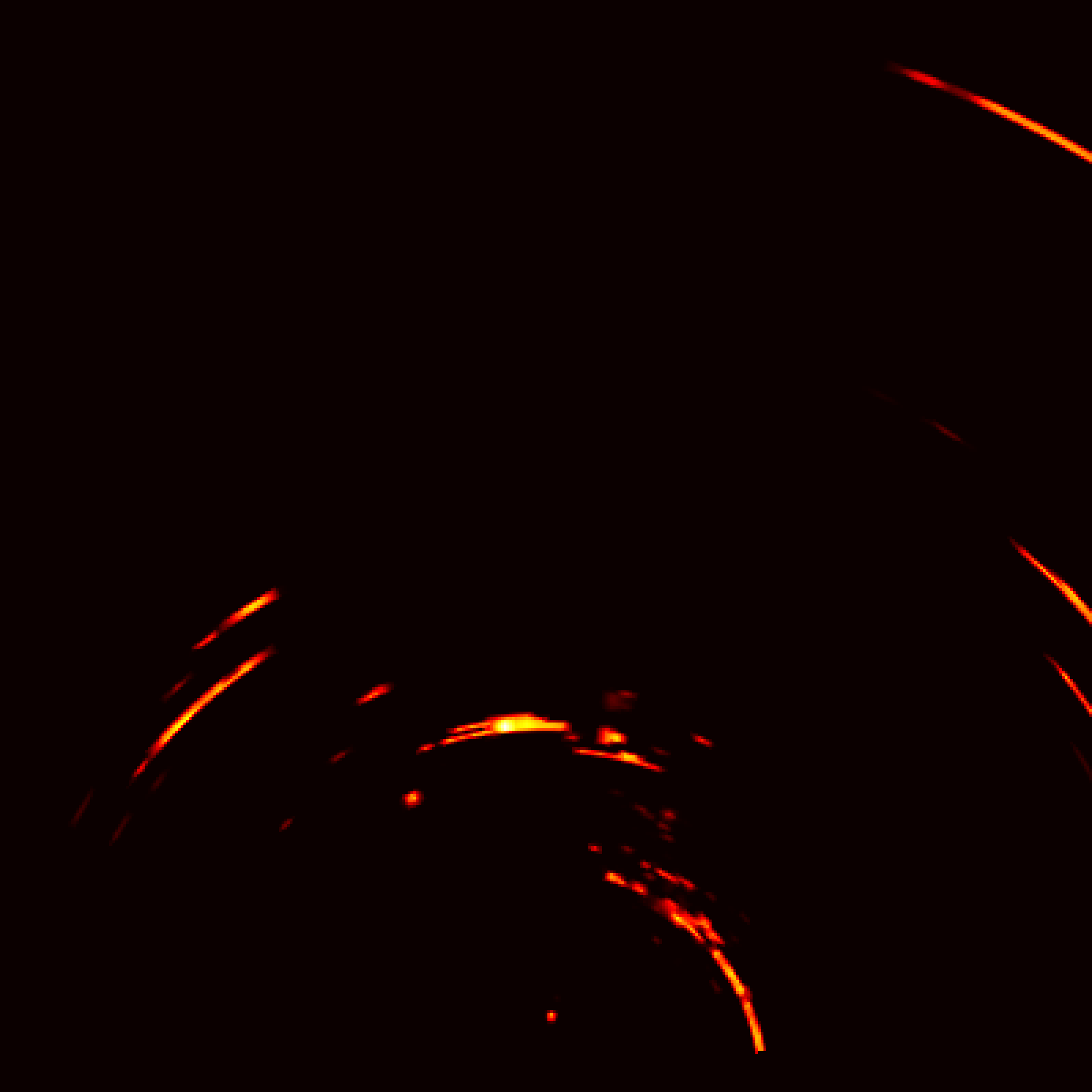} & \includegraphics[width=0.165\linewidth]{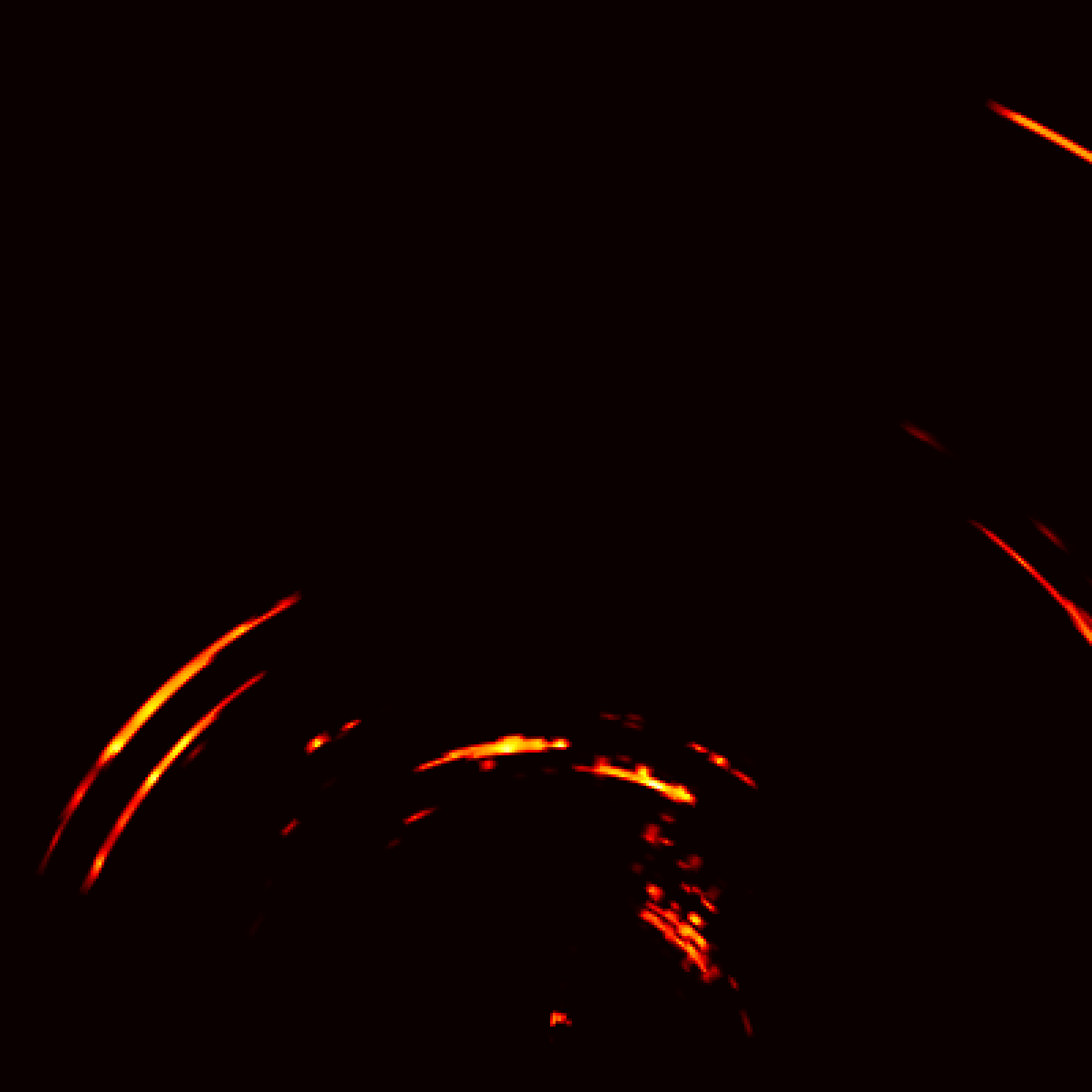} & \includegraphics[width=0.165\linewidth]{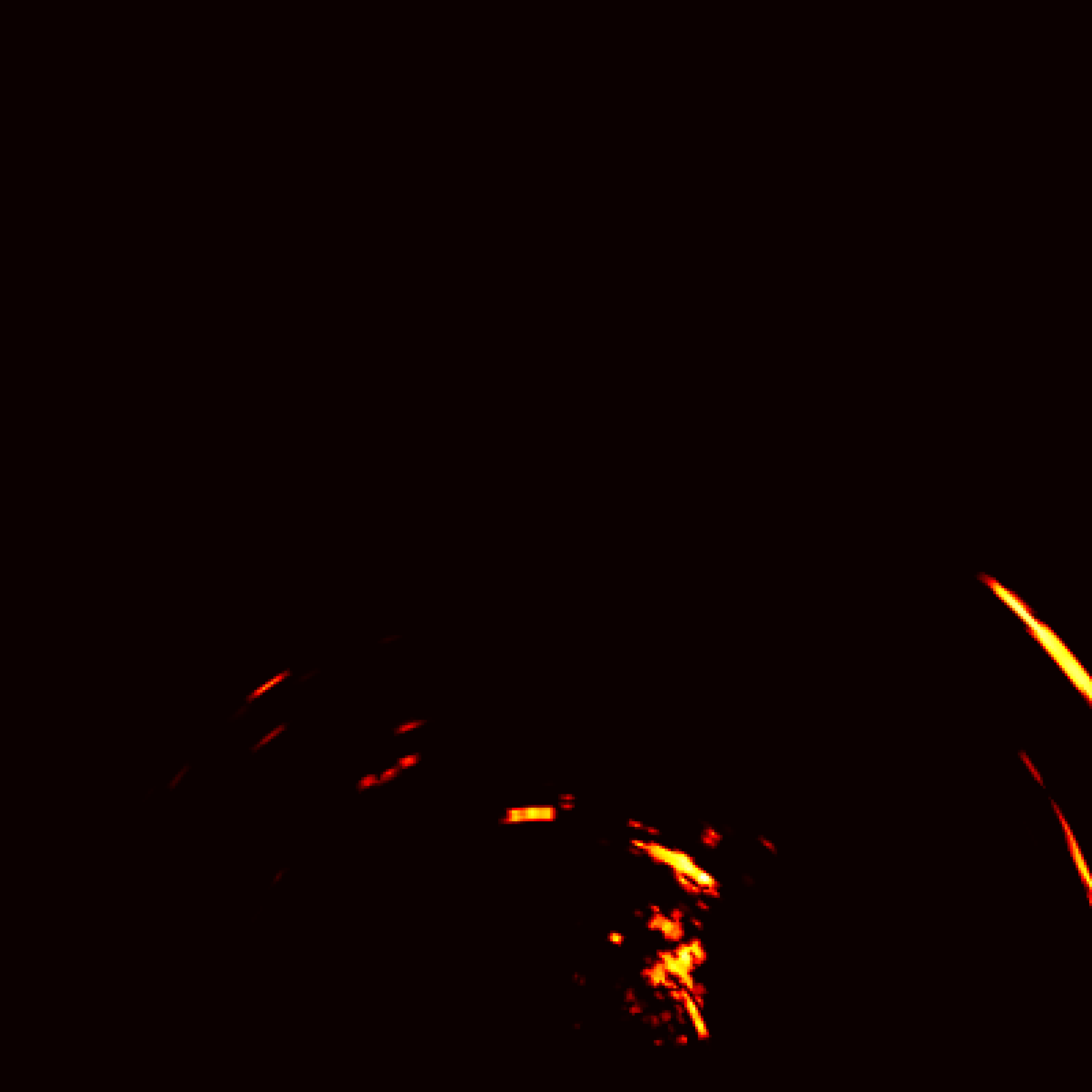} & \includegraphics[width=0.165\linewidth]{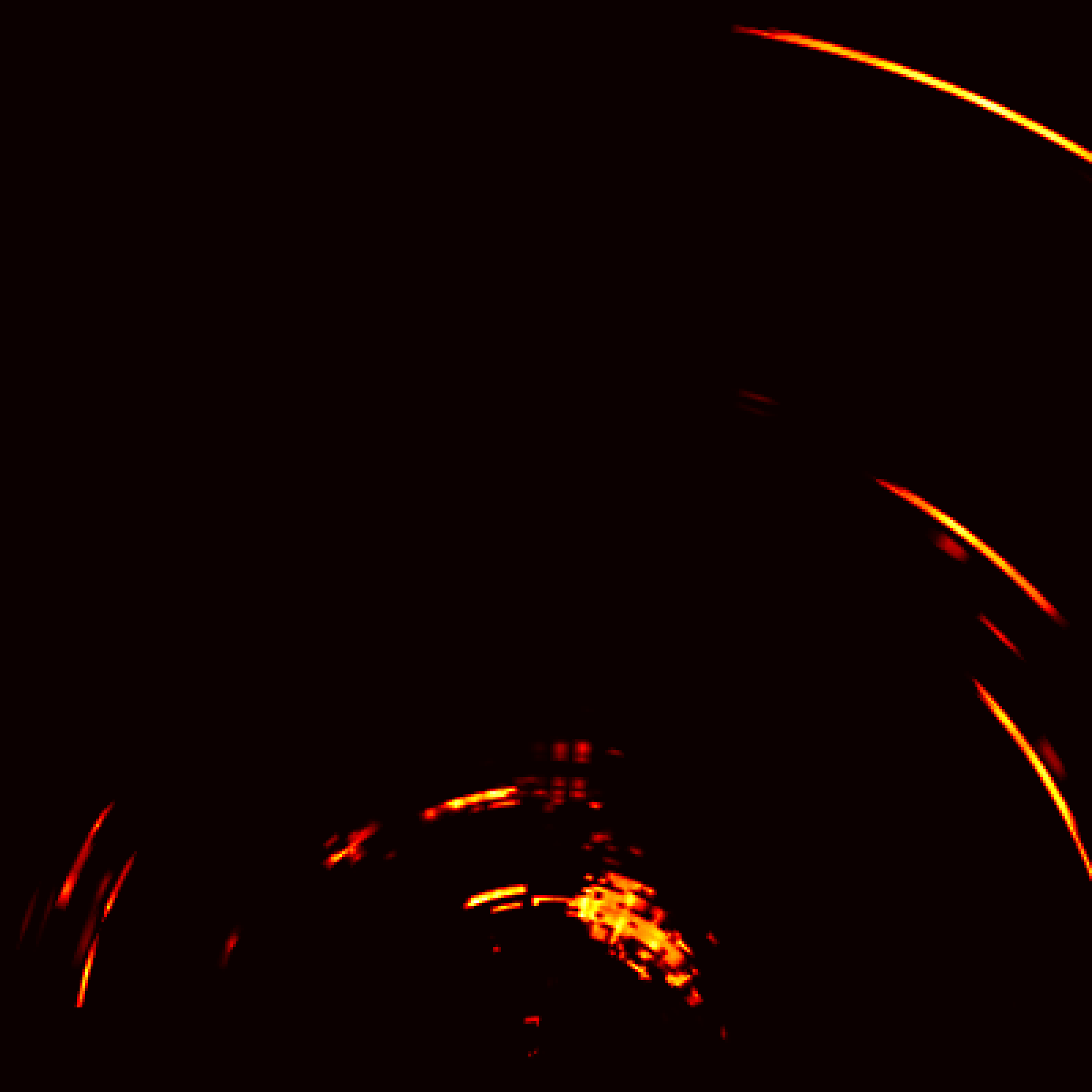} & \includegraphics[width=0.165\linewidth]{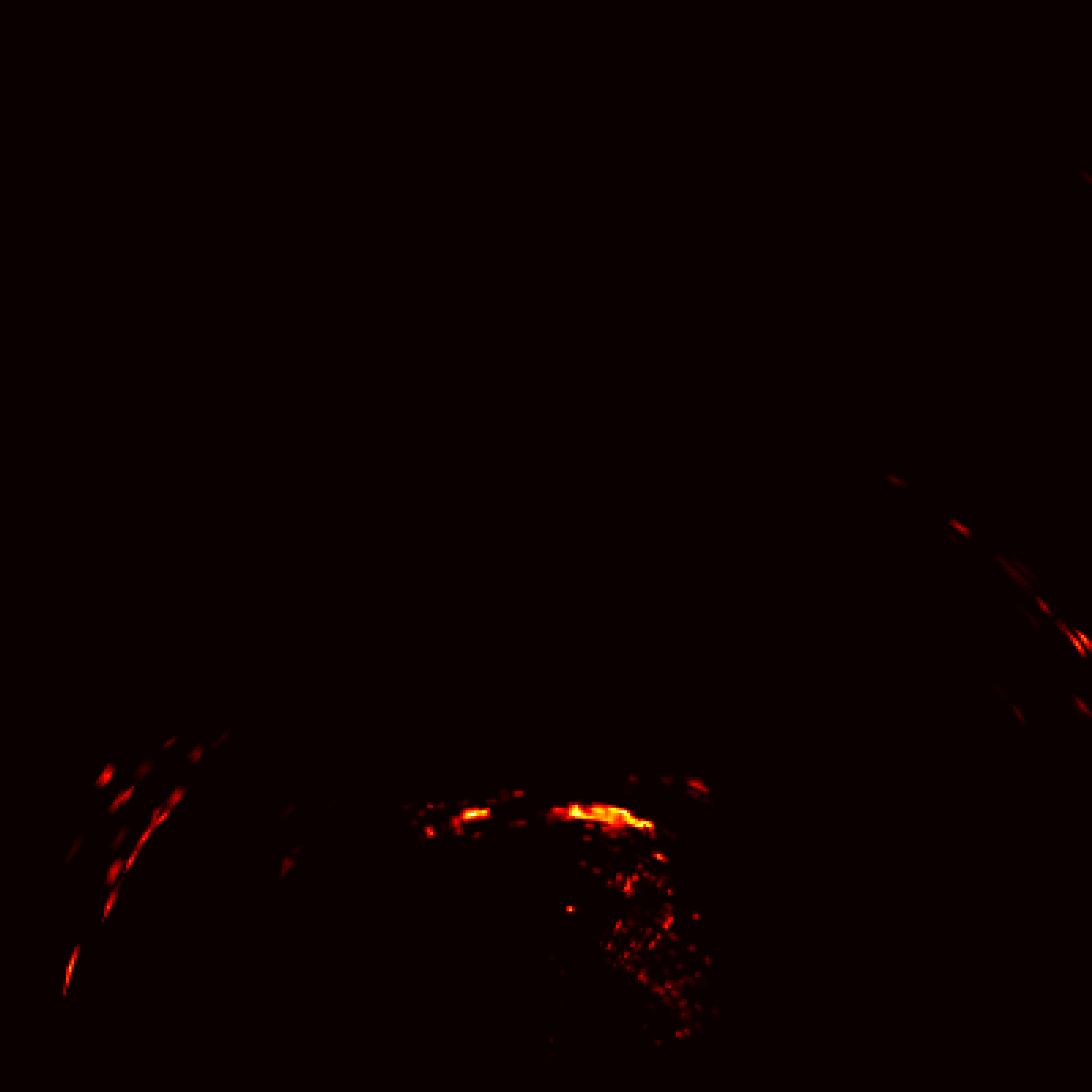} \\
\raisebox{2.0em}{\rotatebox{90}{\scriptsize\textbf{DART}}} & \includegraphics[width=0.165\linewidth]{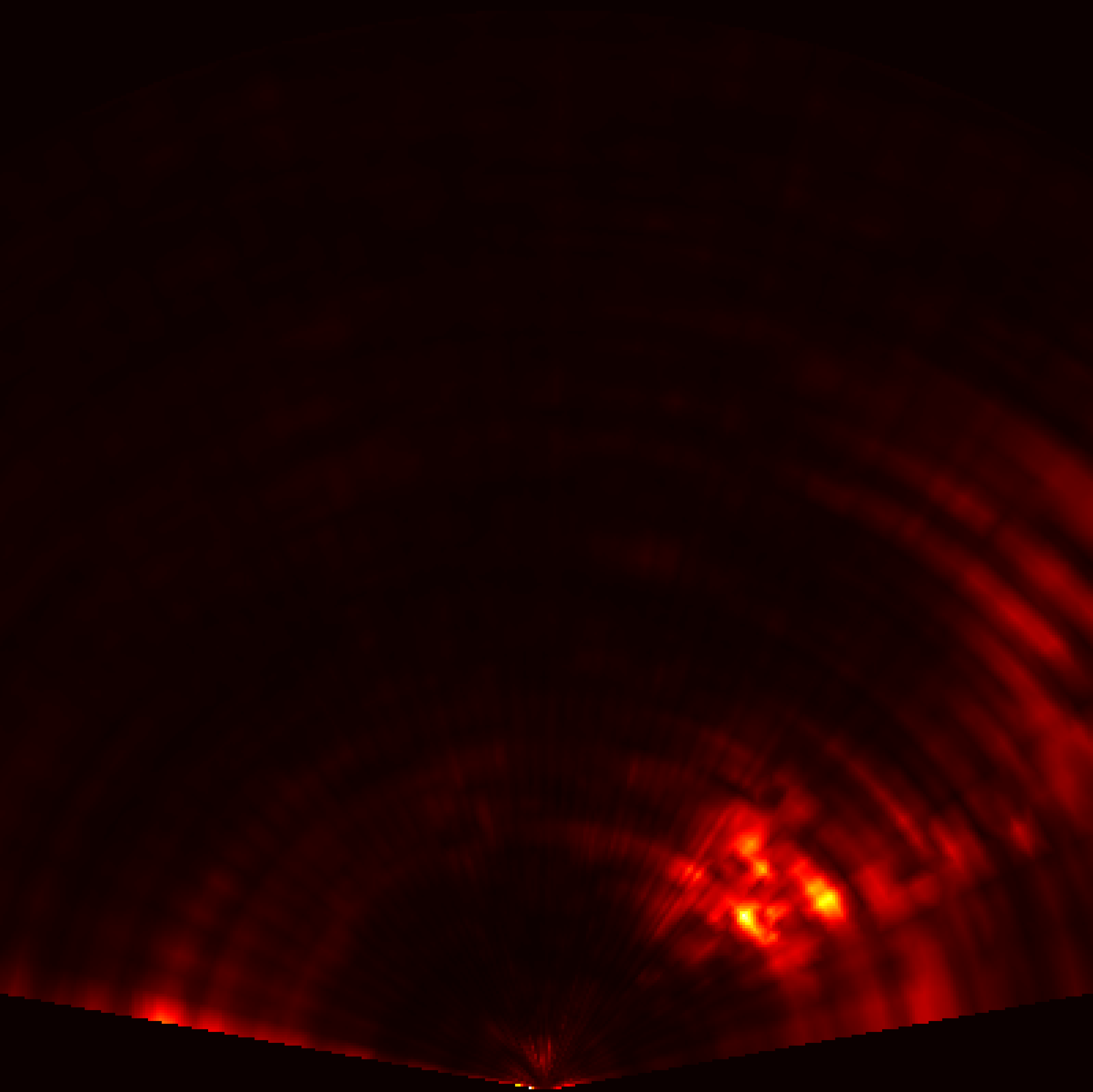} & \includegraphics[width=0.165\linewidth]{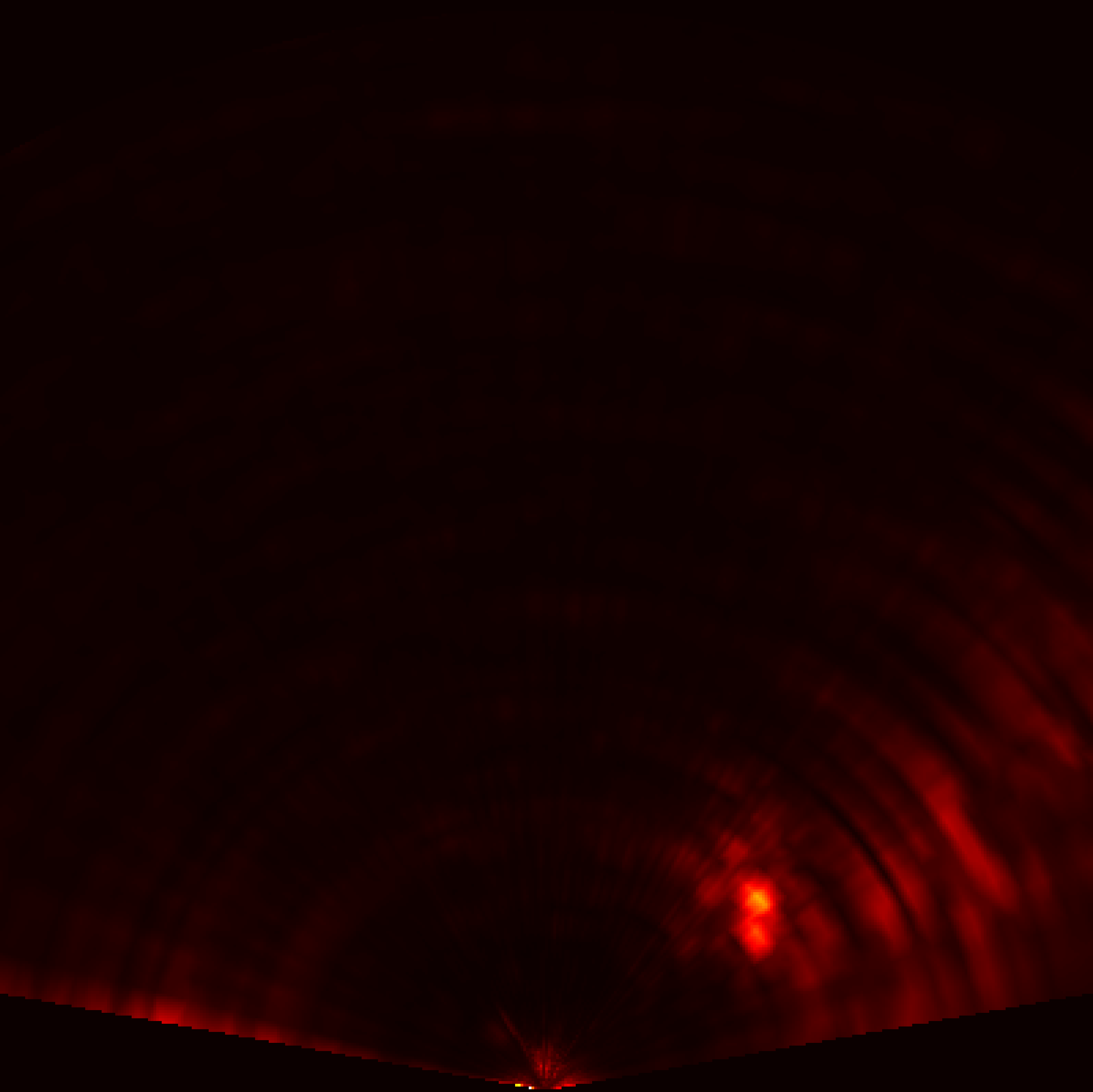} & \includegraphics[width=0.165\linewidth]{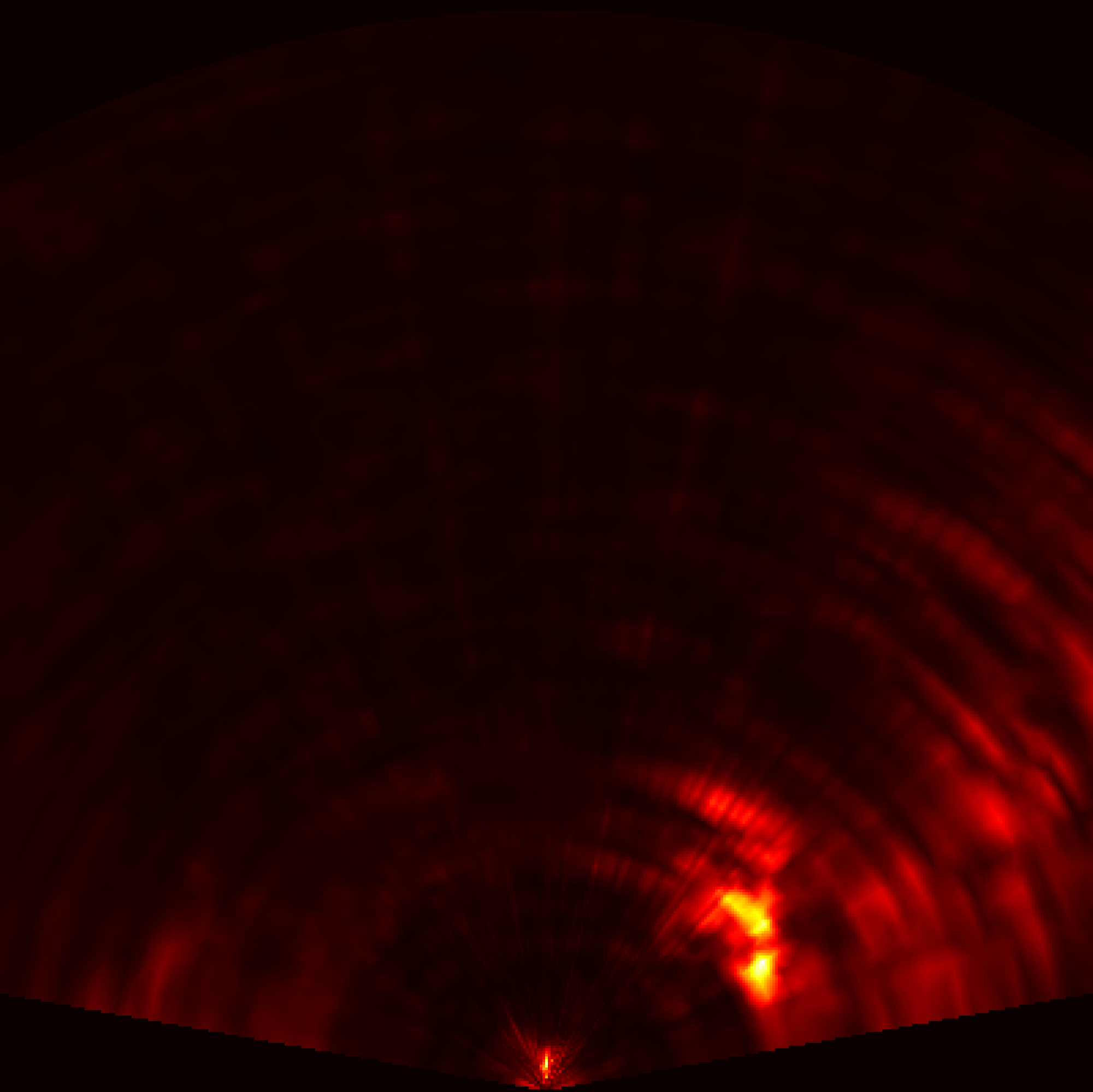} & \includegraphics[width=0.165\linewidth]{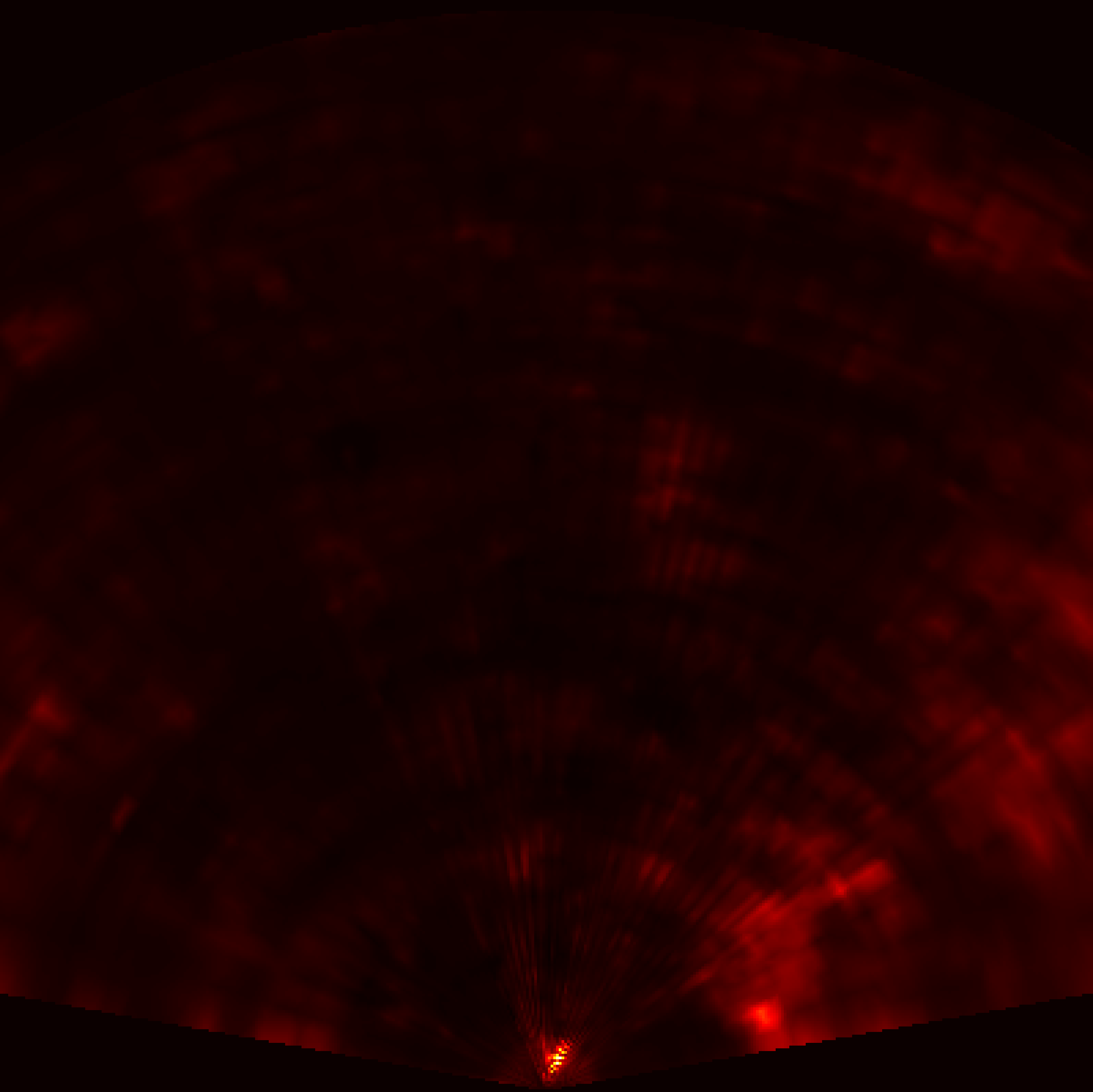} & \includegraphics[width=0.165\linewidth]{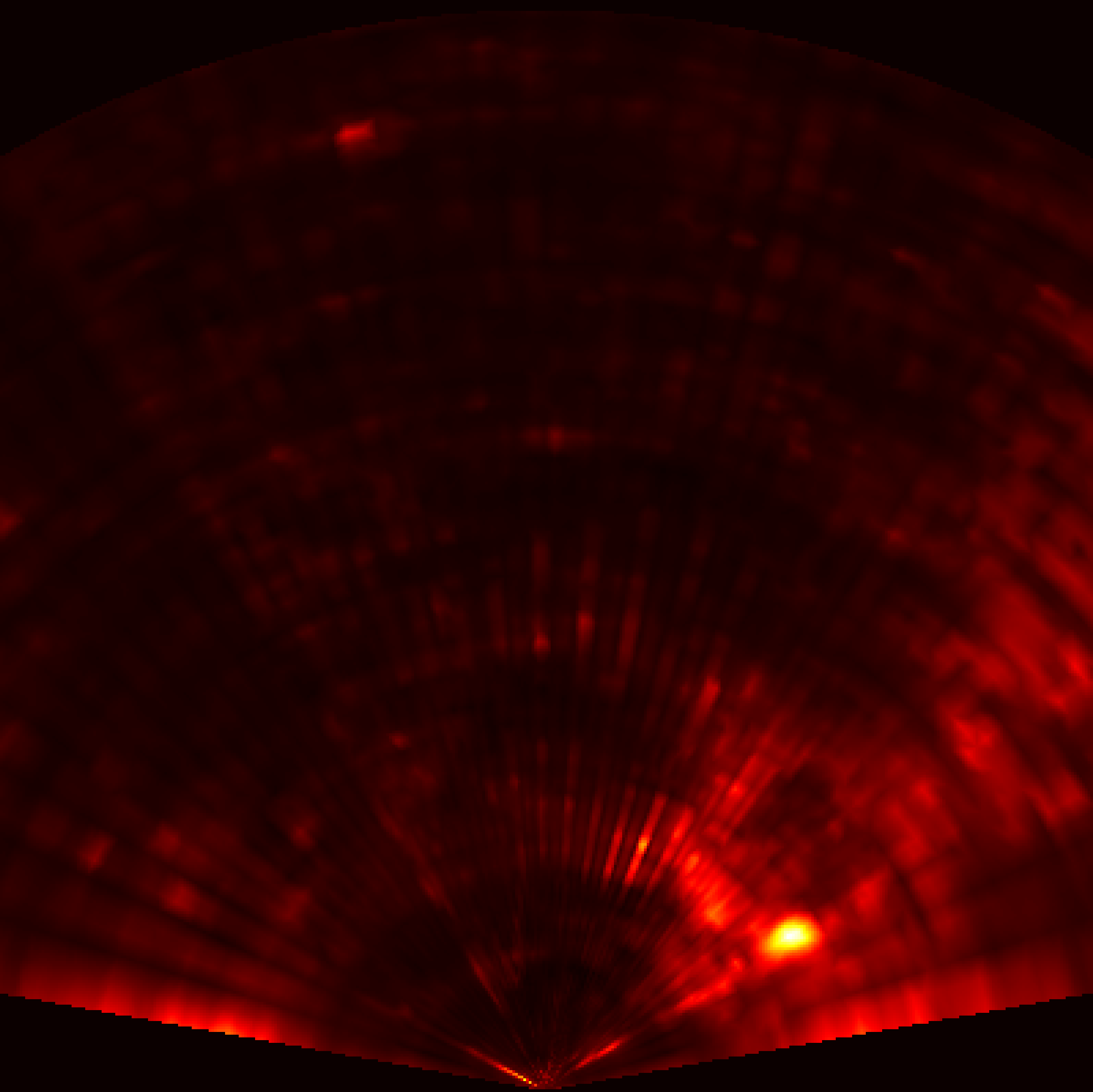} \\
\end{tabular}
\caption{\textbf{Per-frame $|\mathrm{RA}|$ comparison, scene S2\,F105, training frames $F\!+\!1$ to $F\!+\!4$ and held-out test frame $F$ (red header).} Pearson correlation per column, left to right. 3DPS: 0.80, 0.83, 0.85, 0.83, 0.55; mmIR: 0.57, 0.64, 0.57, 0.64, 0.19; RadarSplat: 0.39, 0.44, 0.43, 0.36, 0.37; Radar Fields: 0.10, 0.08, 0.07, 0.28, 0.21; DART: 0.24, 0.17, 0.18, 0.18, 0.24.}
\label{fig:supp_s2_f105_b}
\end{figure}

\begin{figure}[p]
\centering
\setlength{\tabcolsep}{1pt}\renewcommand{\arraystretch}{0.6}
\begin{tabular}{ccccc}
 & {\footnotesize $F\!-\!4$} & {\footnotesize $F\!-\!3$} & {\footnotesize $F\!-\!2$} & {\footnotesize $F\!-\!1$} \\
\raisebox{2.0em}{\rotatebox{90}{\scriptsize\textbf{GT}}} & \includegraphics[width=0.165\linewidth]{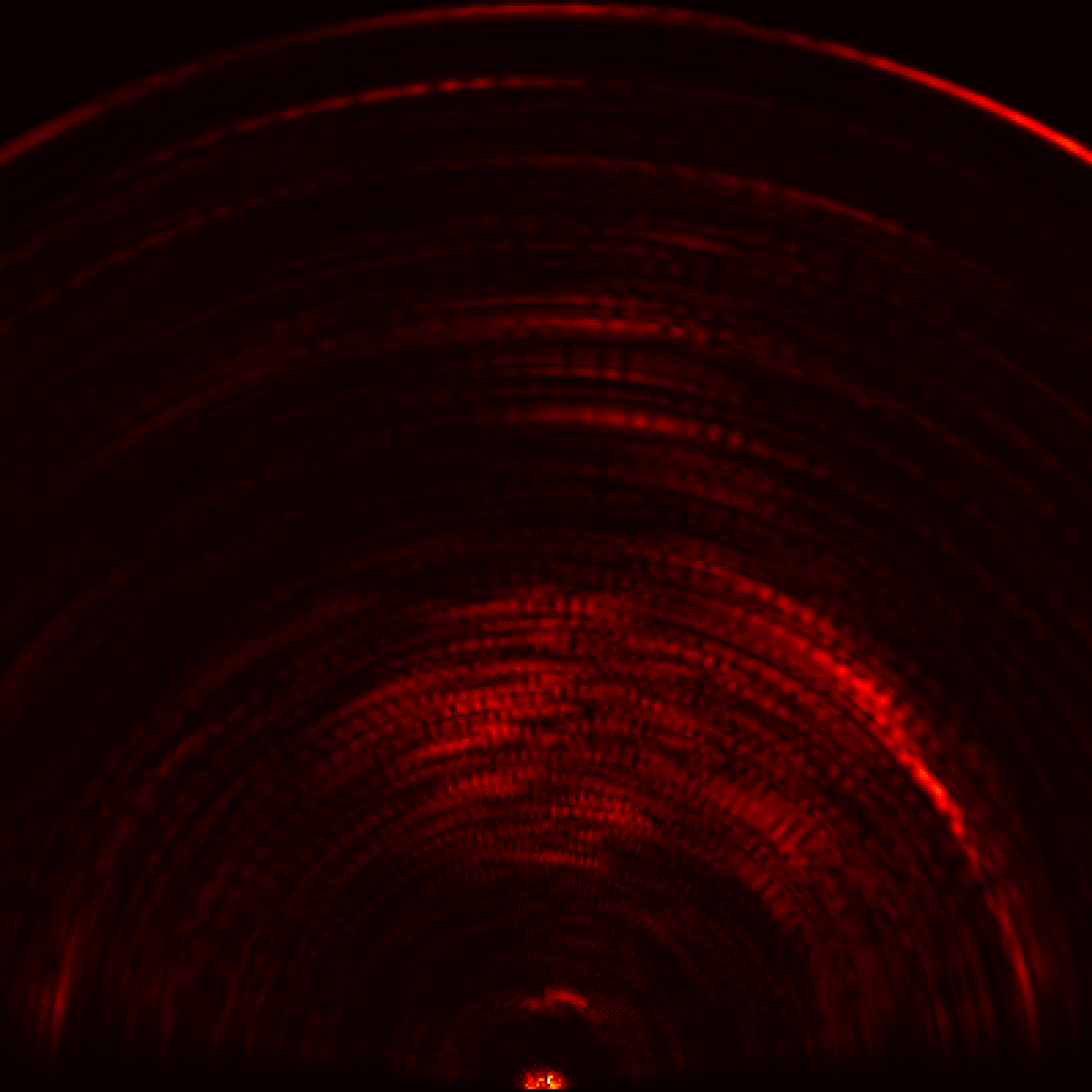} & \includegraphics[width=0.165\linewidth]{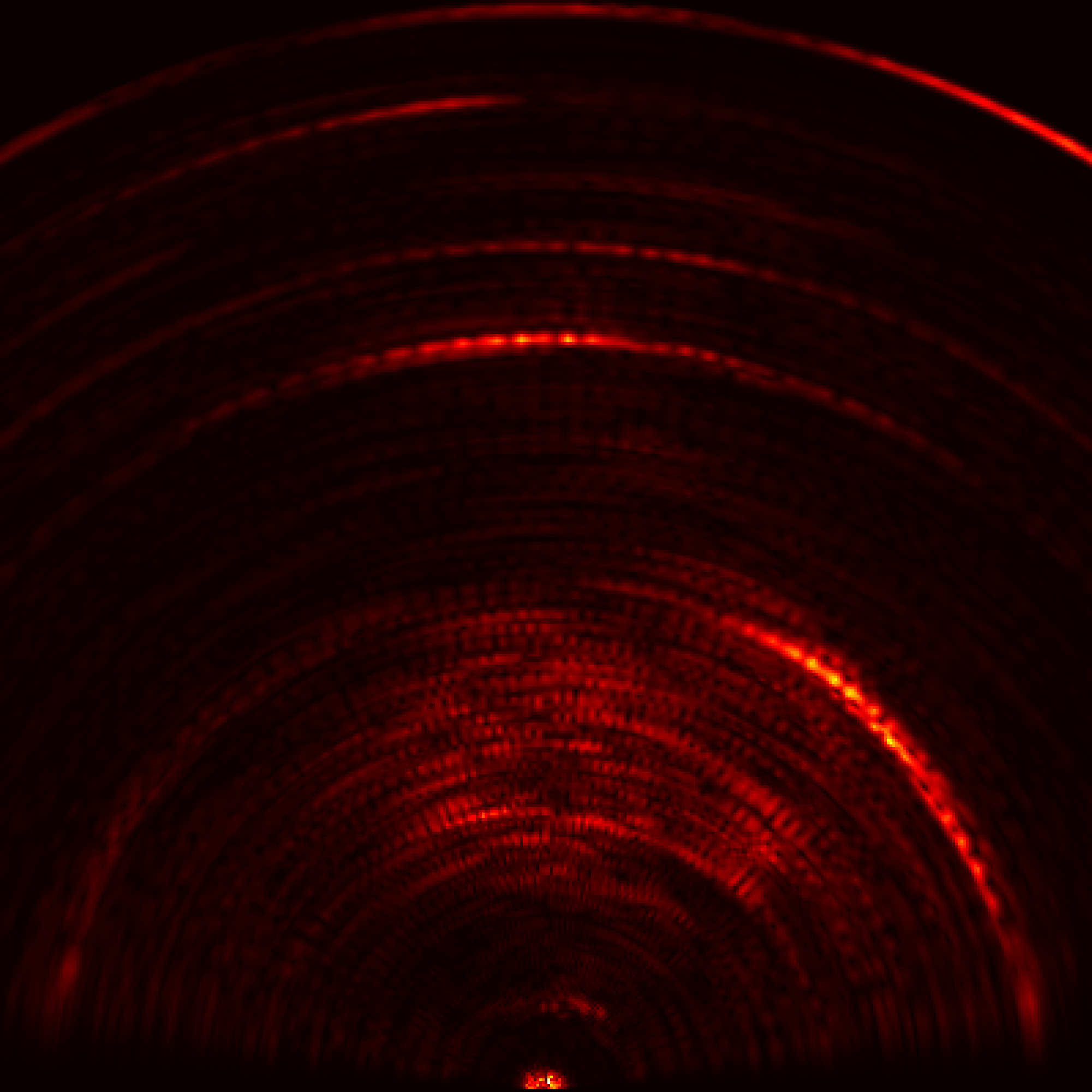} & \includegraphics[width=0.165\linewidth]{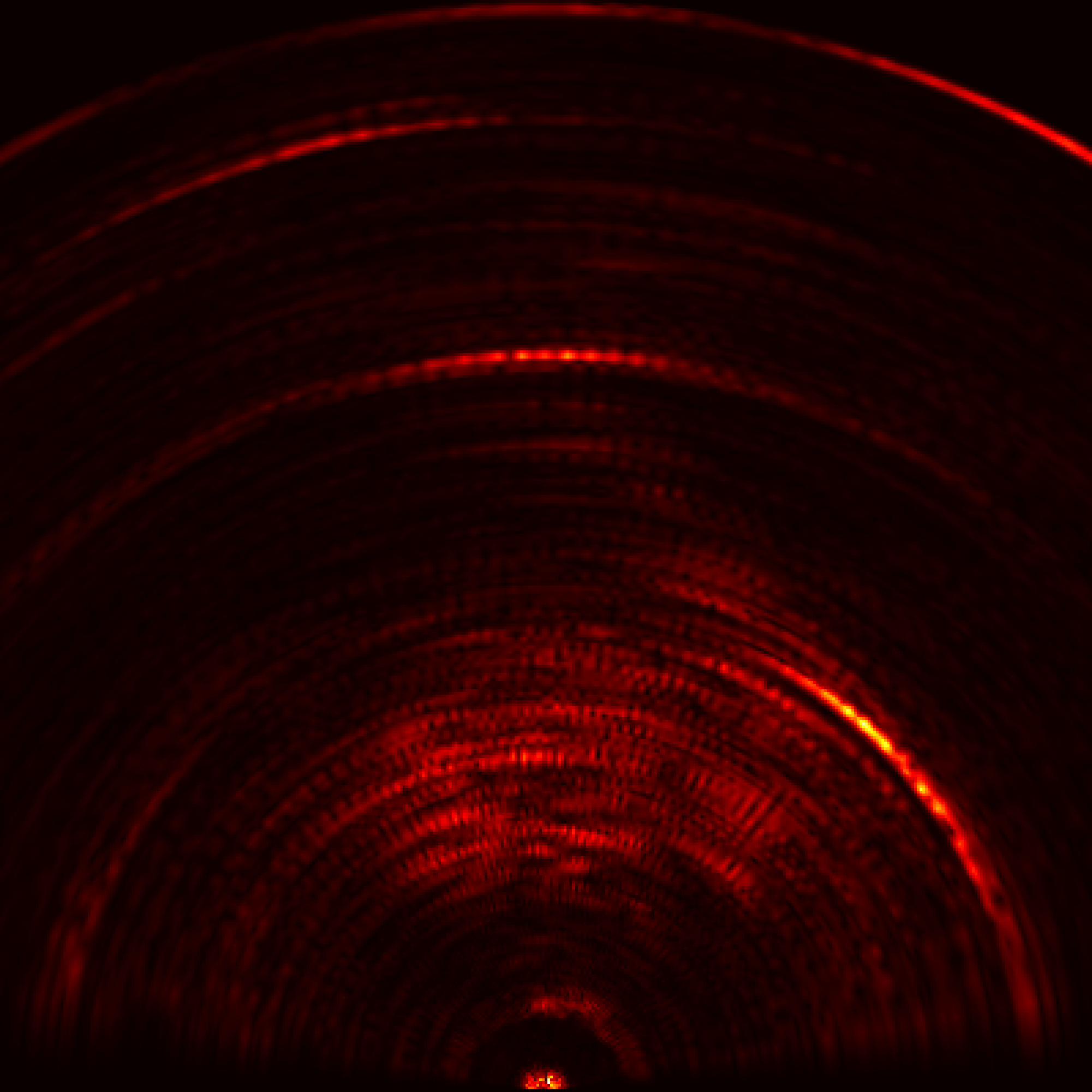} & \includegraphics[width=0.165\linewidth]{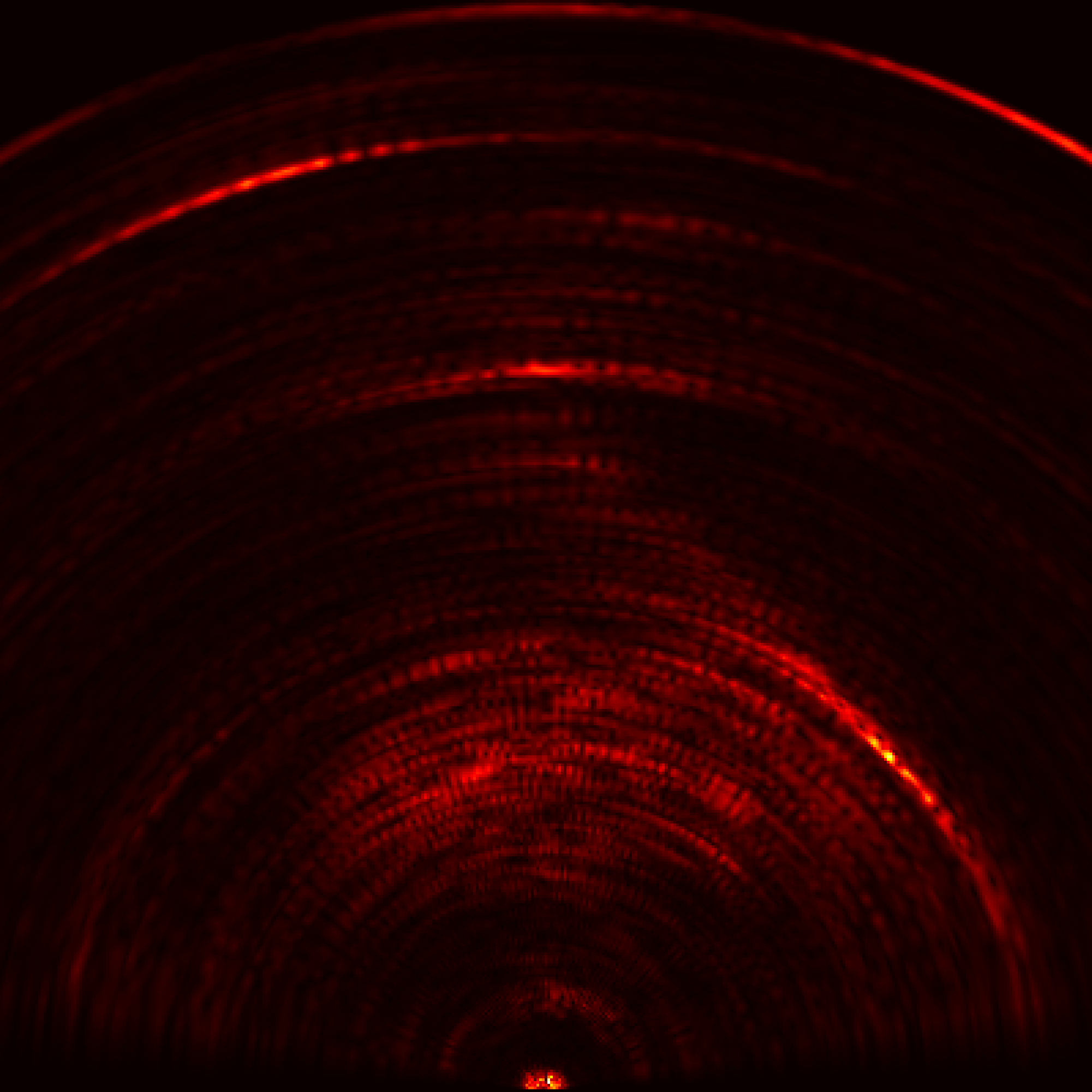} \\
\raisebox{2.0em}{\rotatebox{90}{\scriptsize\textbf{3DPS (Ours)}}} & \includegraphics[width=0.165\linewidth]{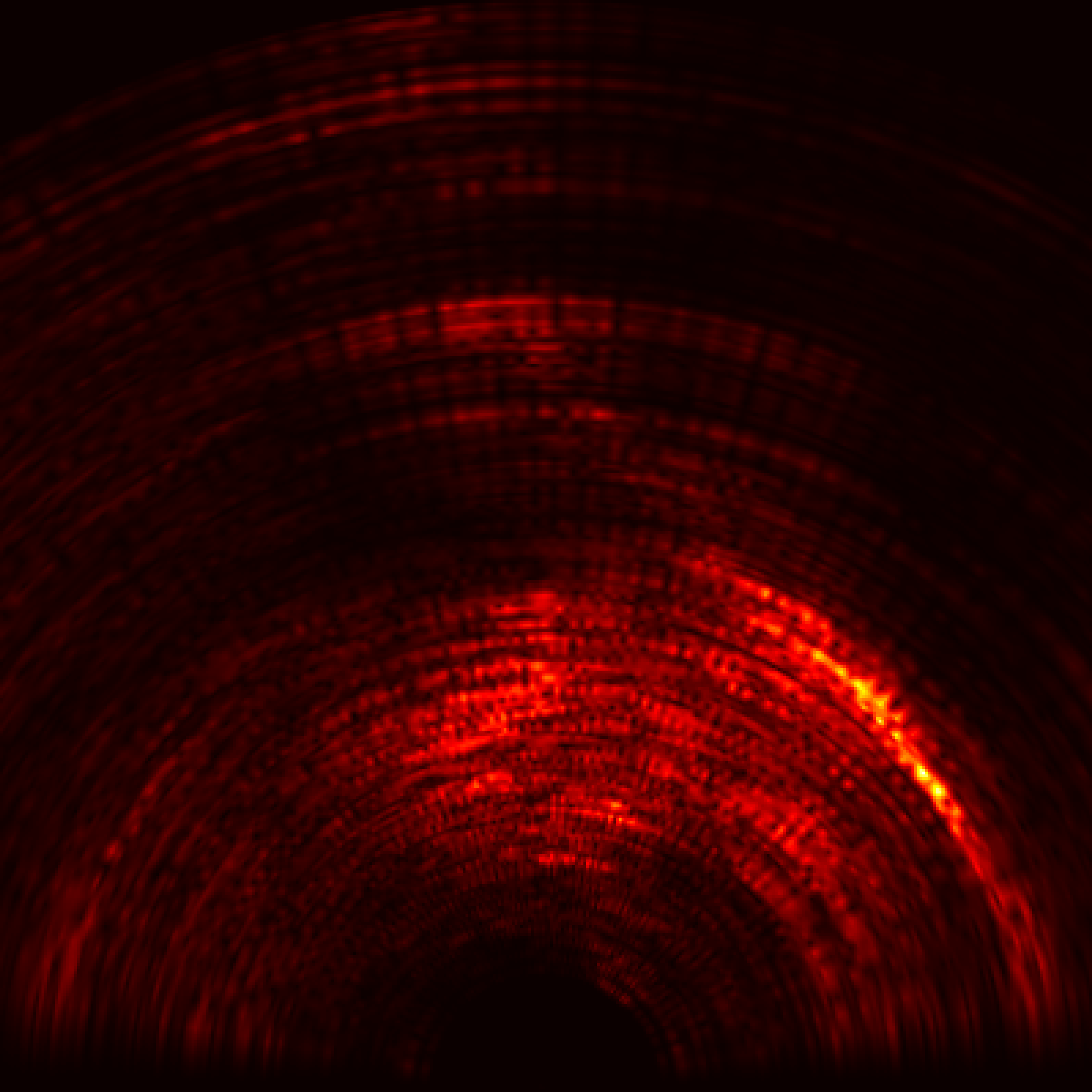} & \includegraphics[width=0.165\linewidth]{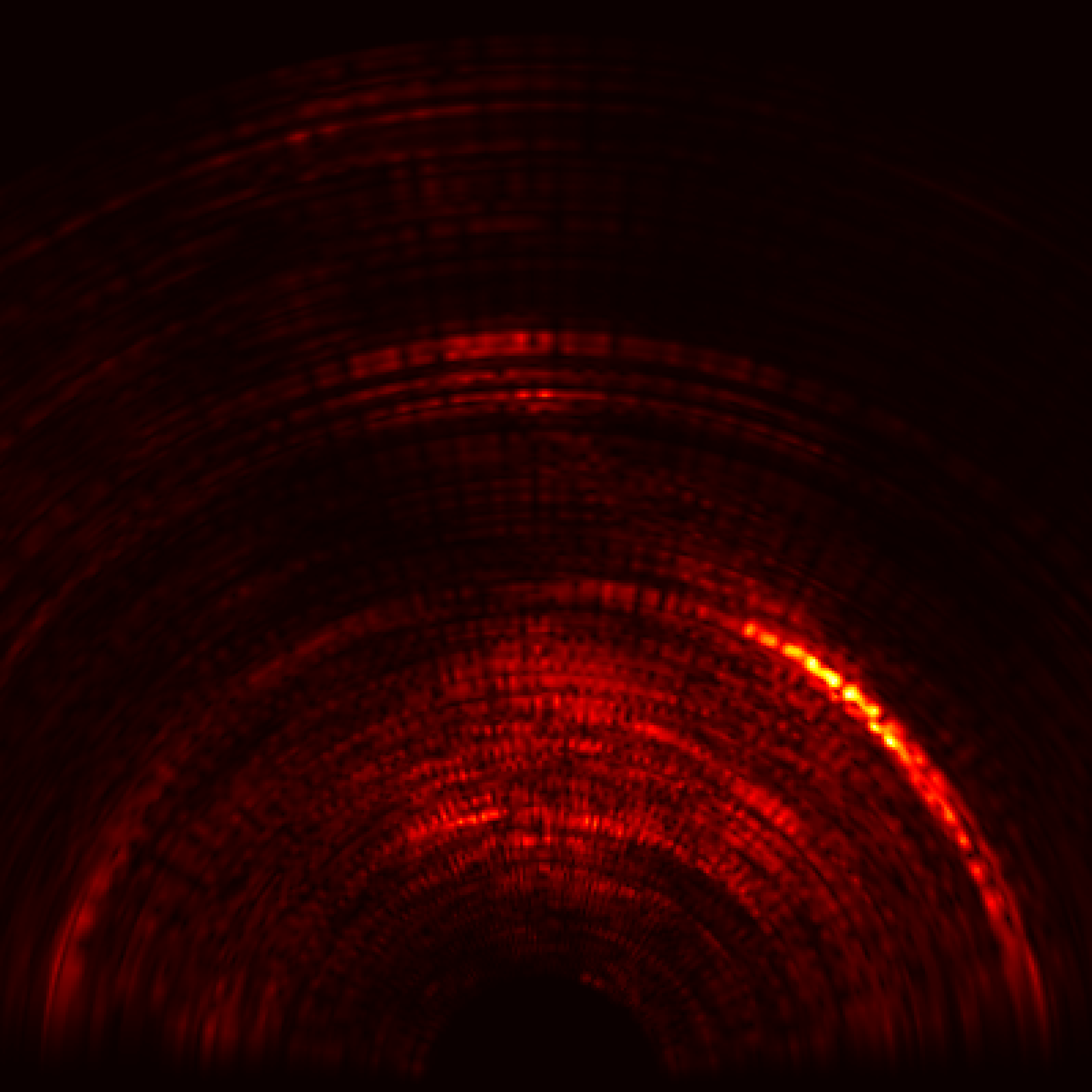} & \includegraphics[width=0.165\linewidth]{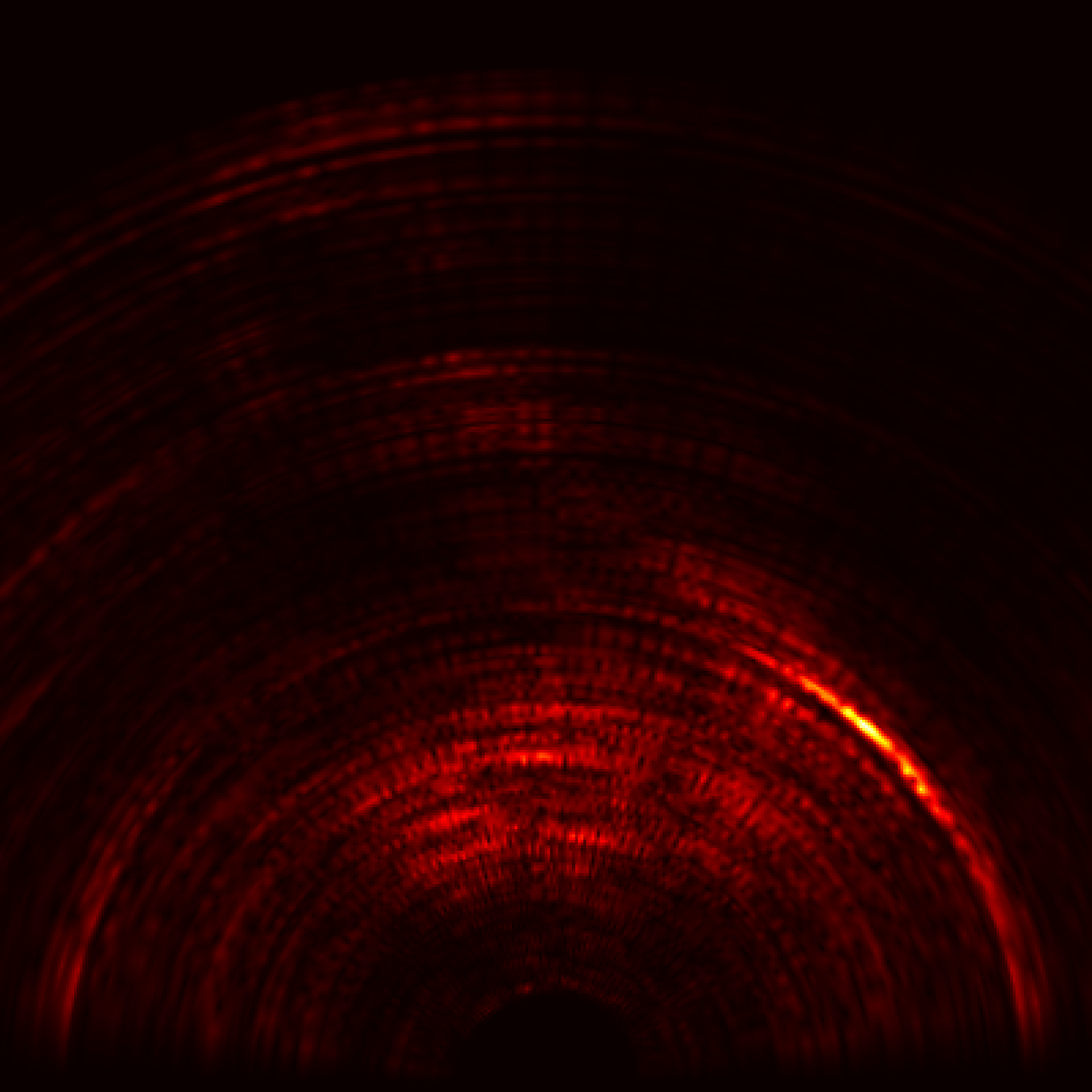} & \includegraphics[width=0.165\linewidth]{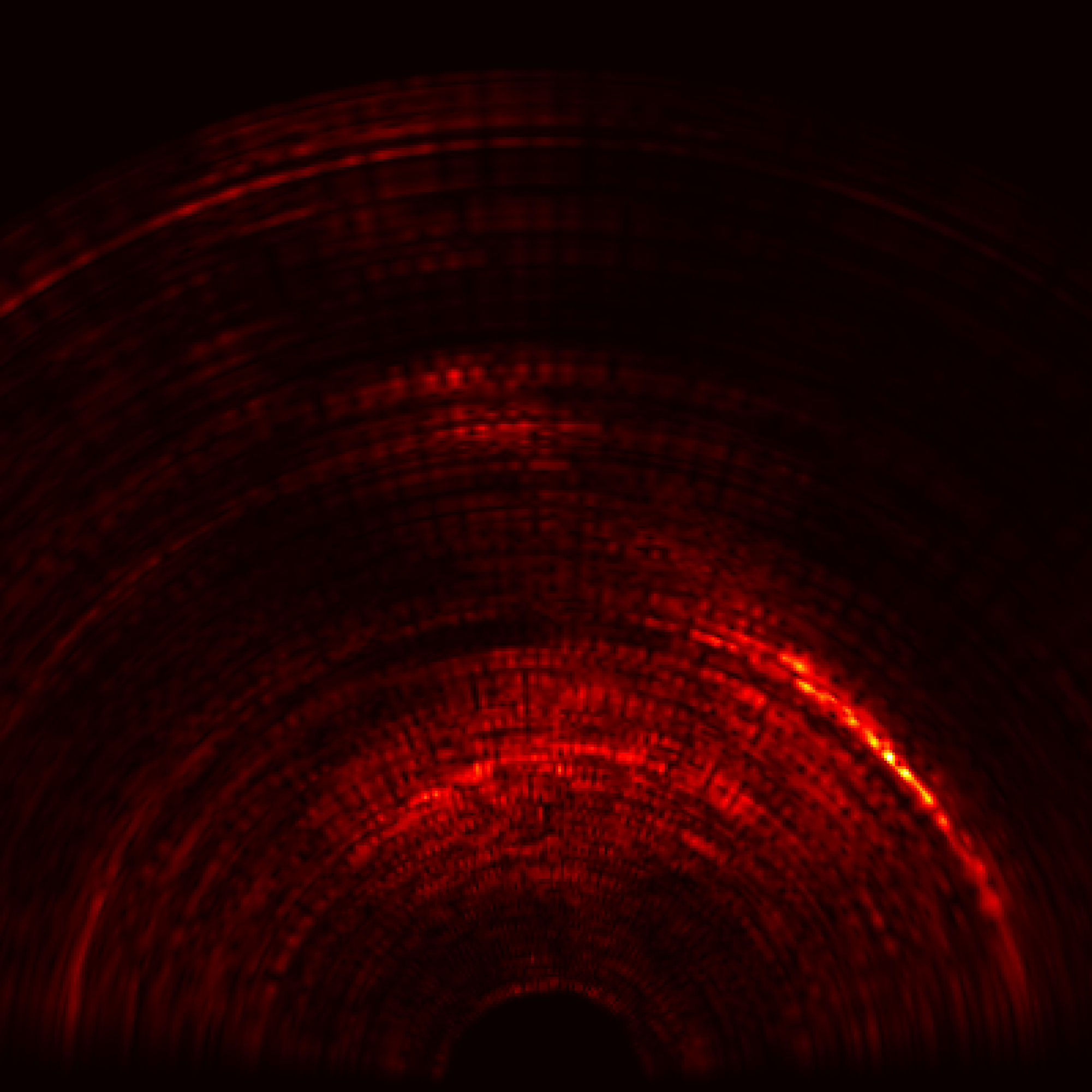} \\
\raisebox{2.0em}{\rotatebox{90}{\scriptsize\textbf{mmIR}}} & \includegraphics[width=0.165\linewidth]{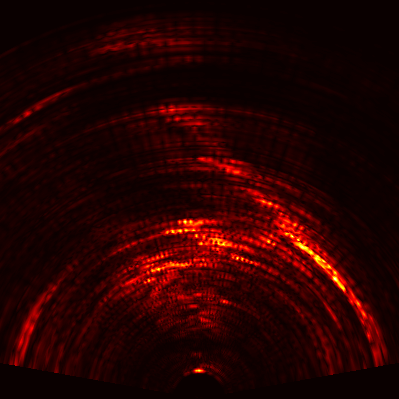} & \includegraphics[width=0.165\linewidth]{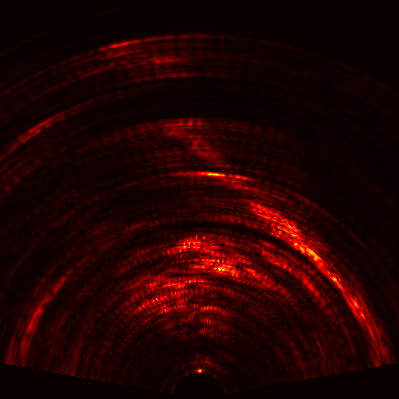} & \includegraphics[width=0.165\linewidth]{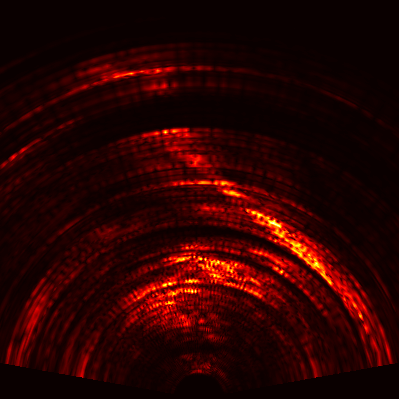} & \includegraphics[width=0.165\linewidth]{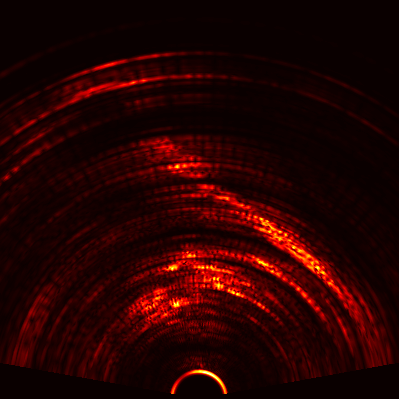} \\
\raisebox{2.0em}{\rotatebox{90}{\scriptsize\textbf{RadarSplat}}} & \includegraphics[width=0.165\linewidth]{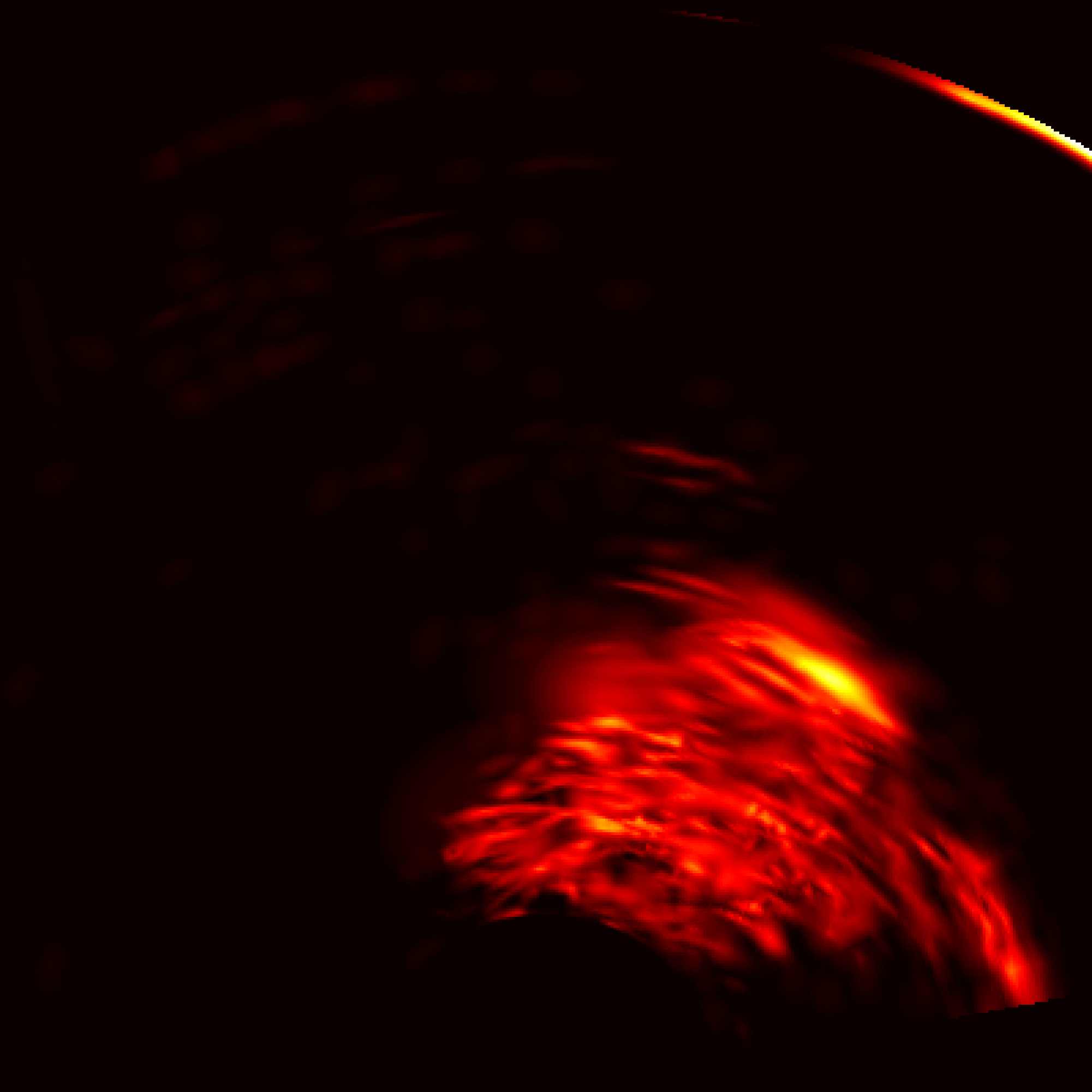} & \includegraphics[width=0.165\linewidth]{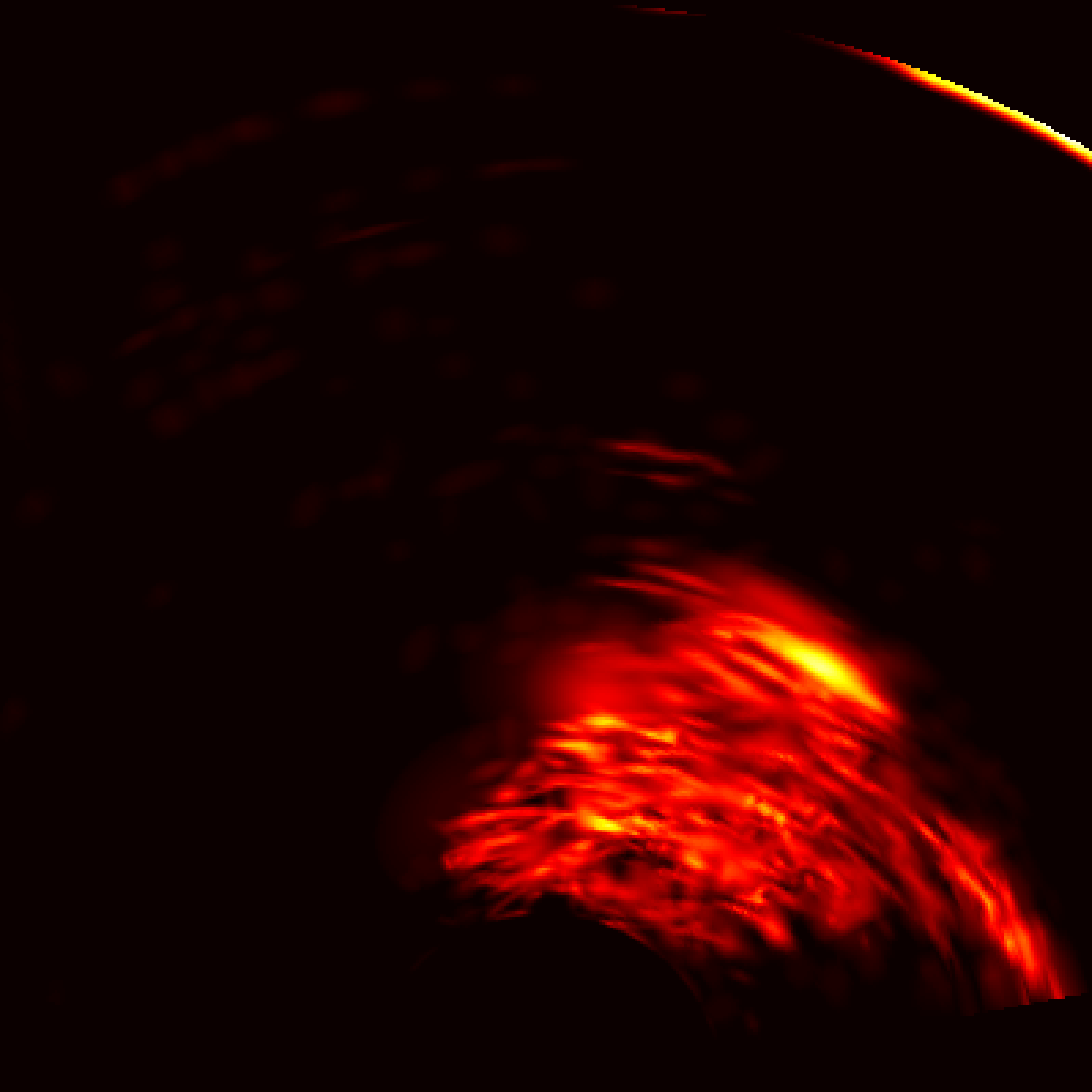} & \includegraphics[width=0.165\linewidth]{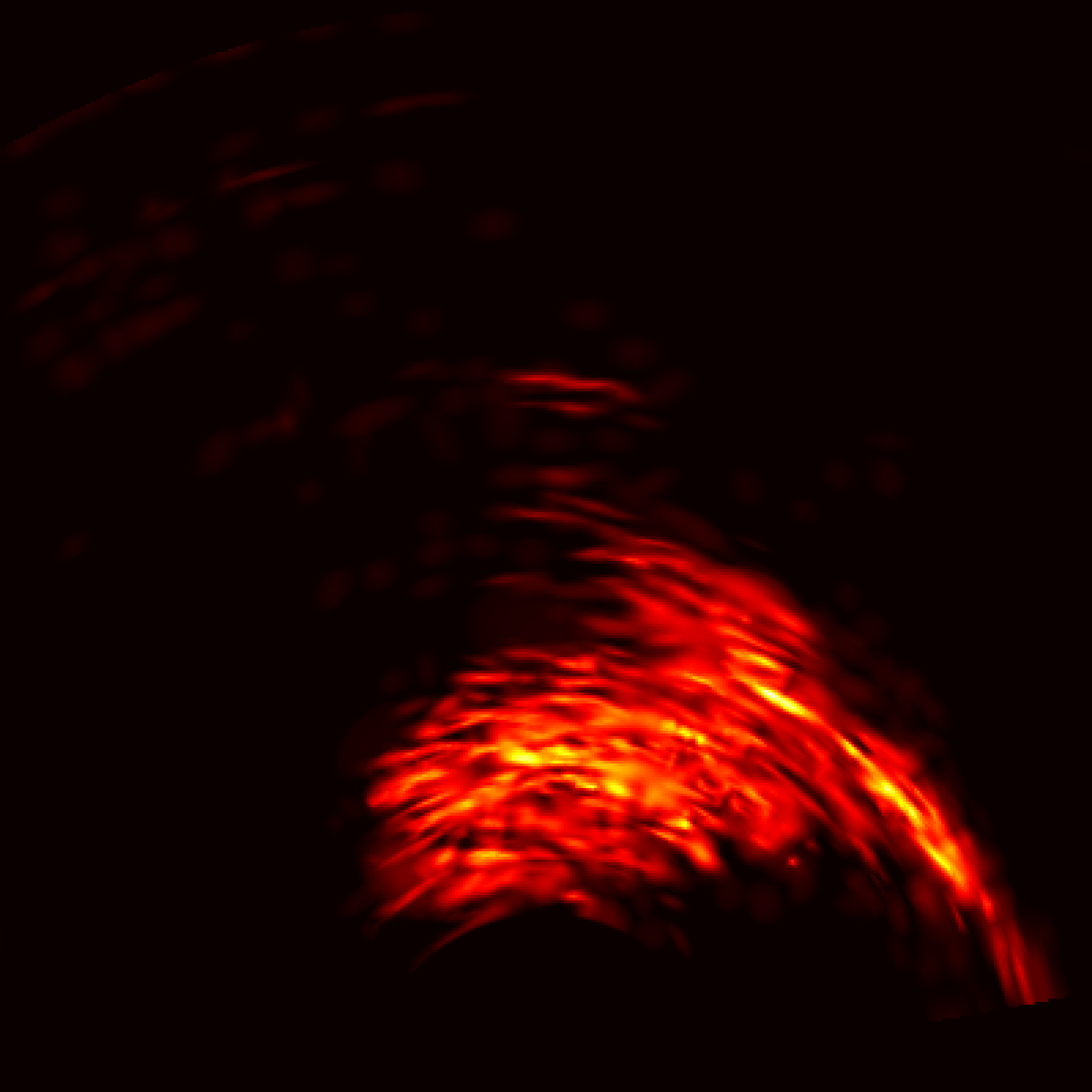} & \includegraphics[width=0.165\linewidth]{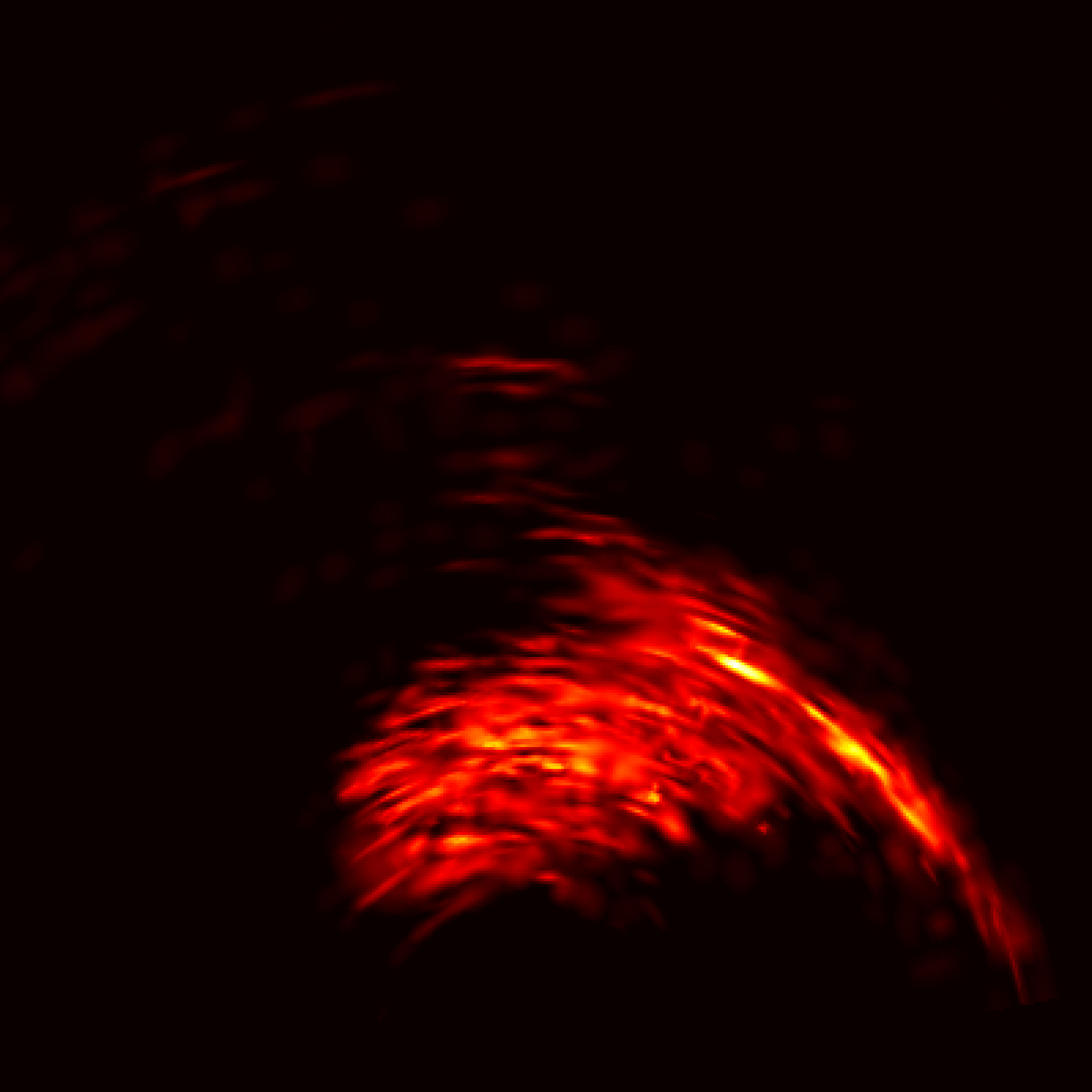} \\
\raisebox{2.0em}{\rotatebox{90}{\scriptsize\textbf{Radar Fields}}} & \includegraphics[width=0.165\linewidth]{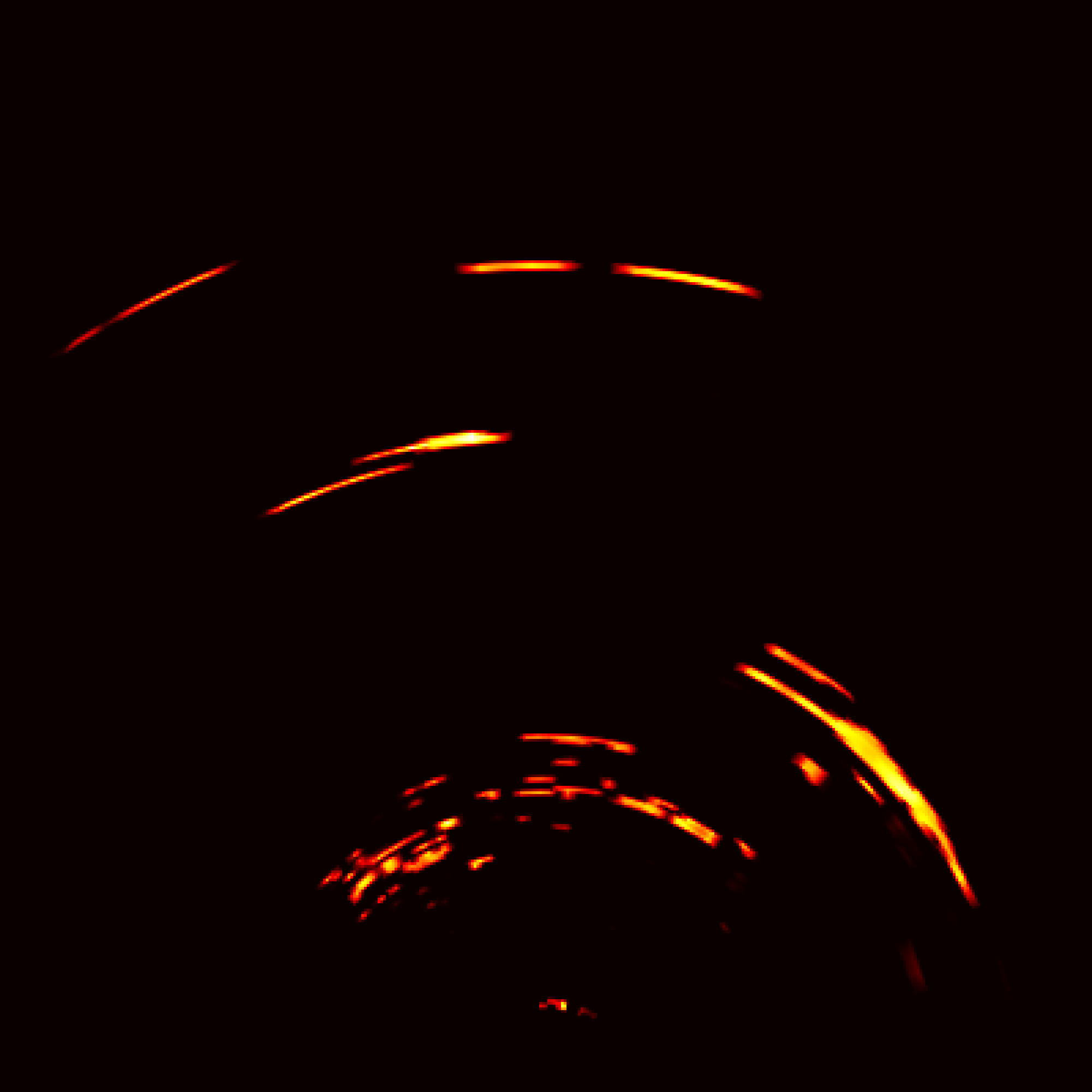} & \includegraphics[width=0.165\linewidth]{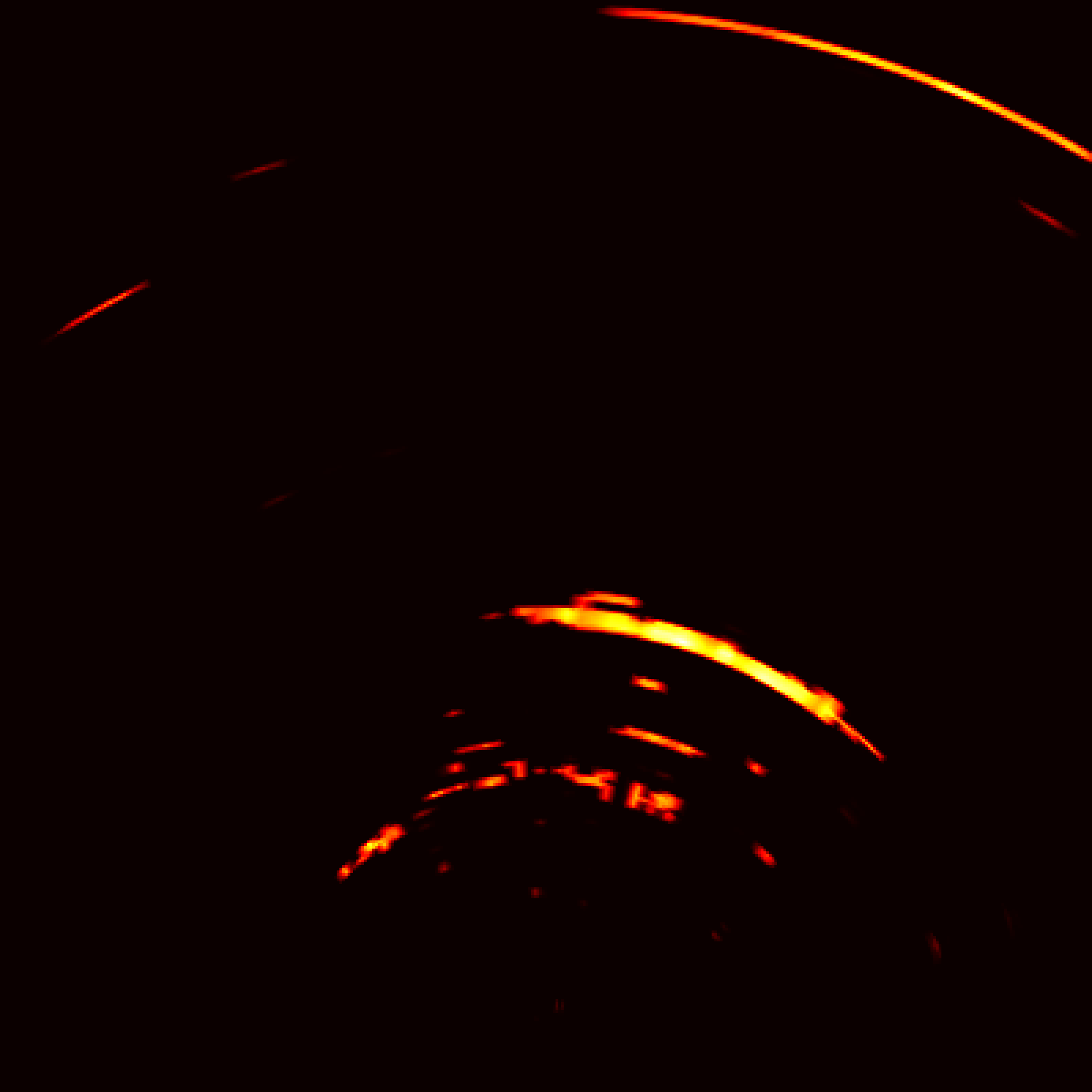} & \includegraphics[width=0.165\linewidth]{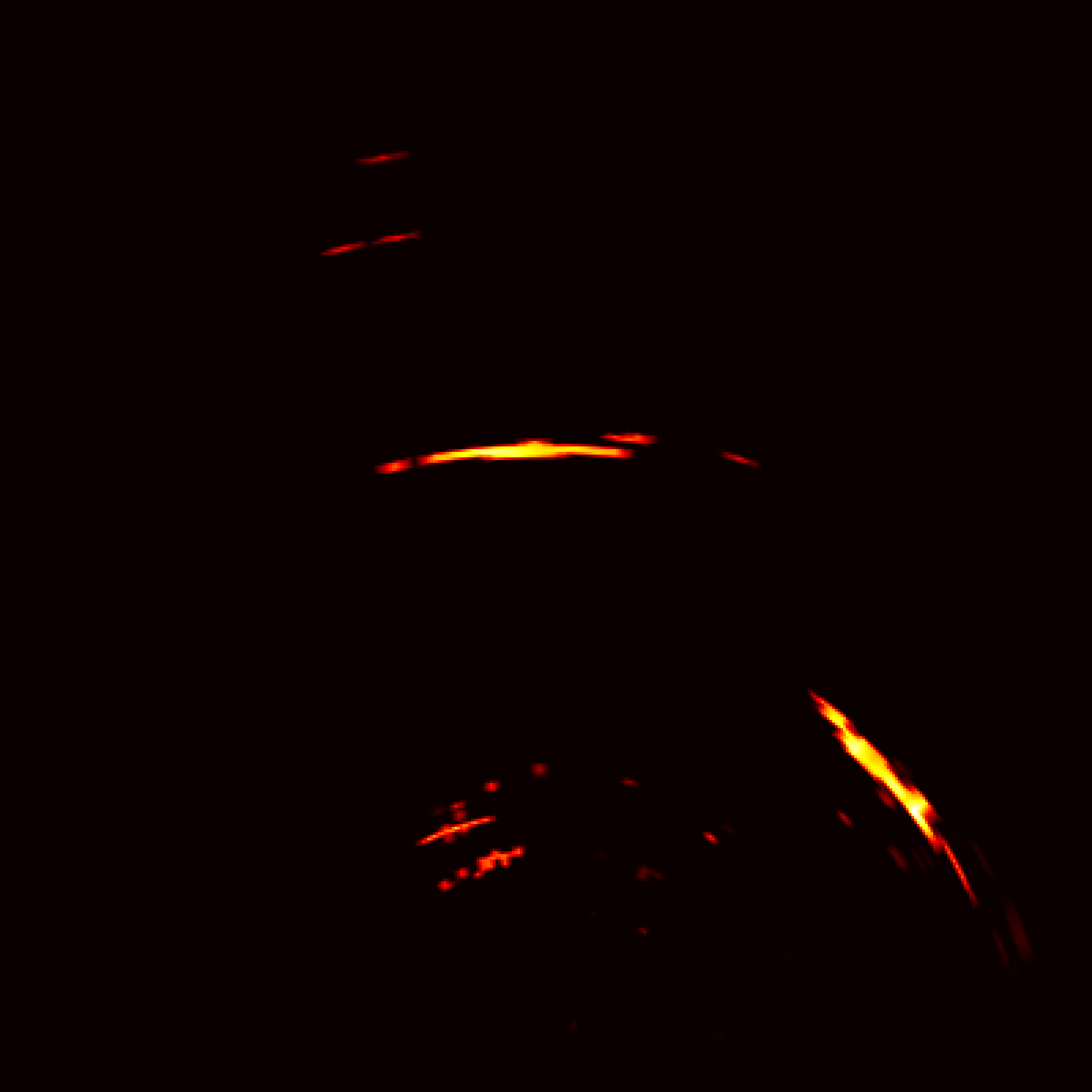} & \includegraphics[width=0.165\linewidth]{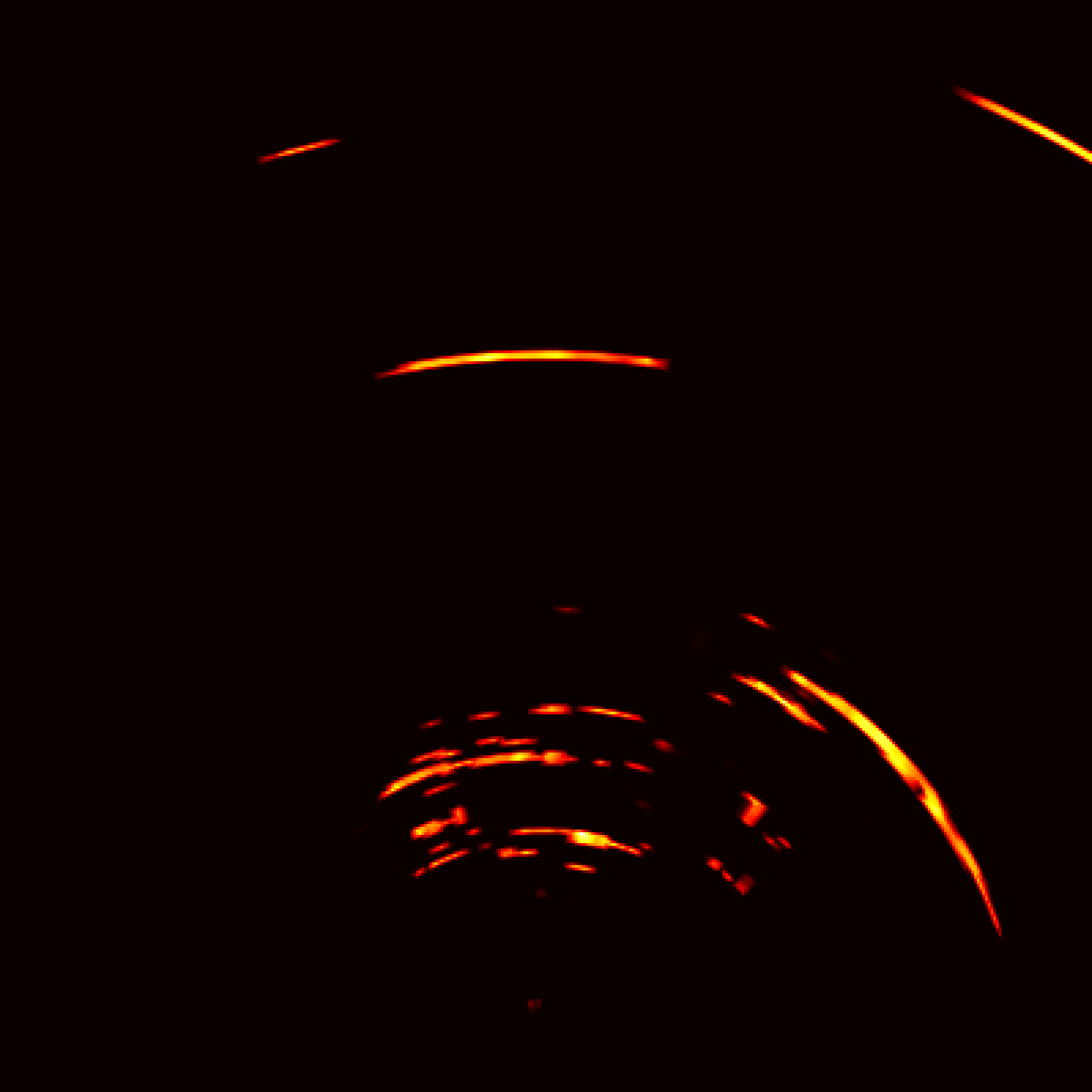} \\
\raisebox{2.0em}{\rotatebox{90}{\scriptsize\textbf{DART}}} & \includegraphics[width=0.165\linewidth]{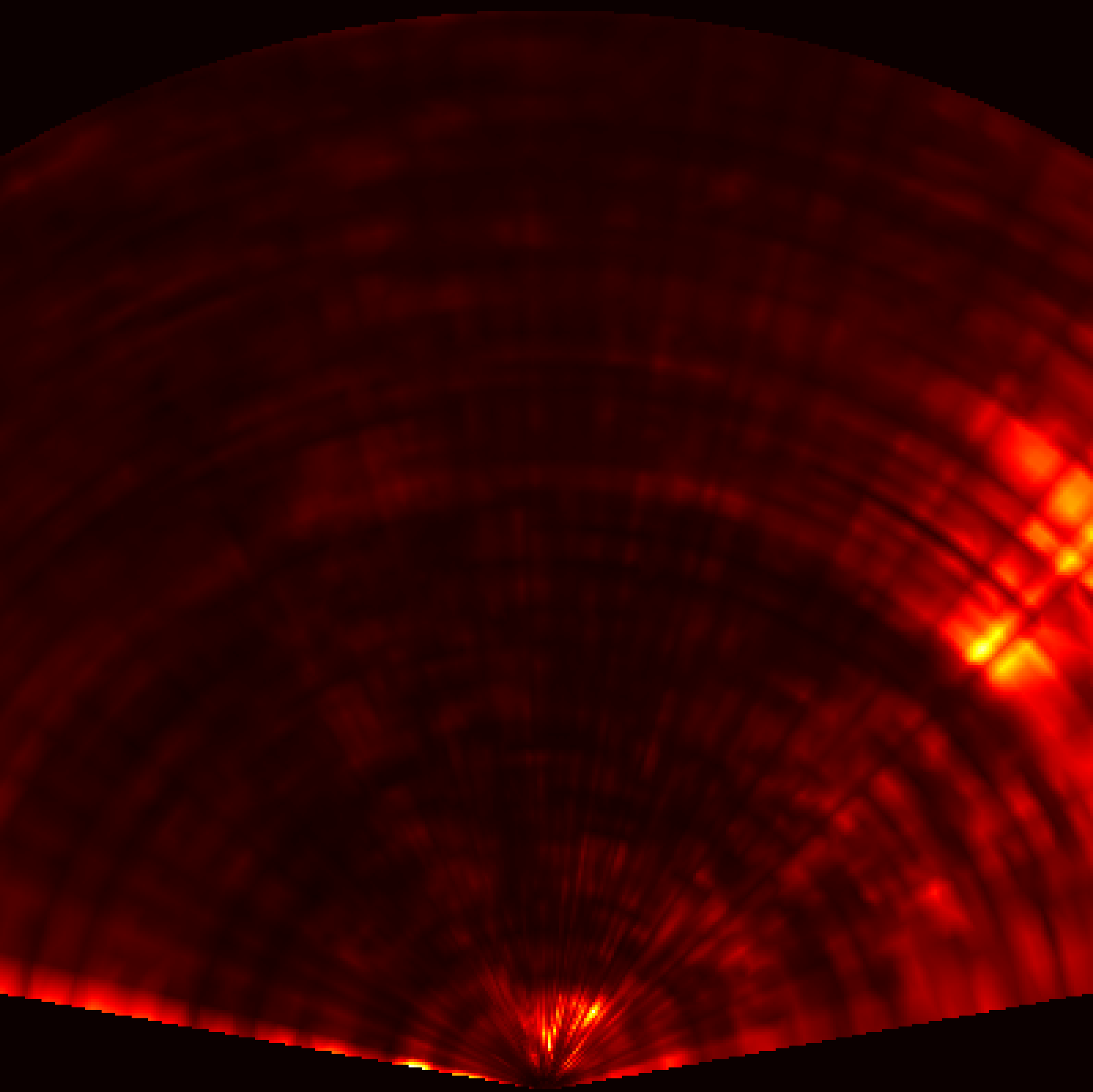} & \includegraphics[width=0.165\linewidth]{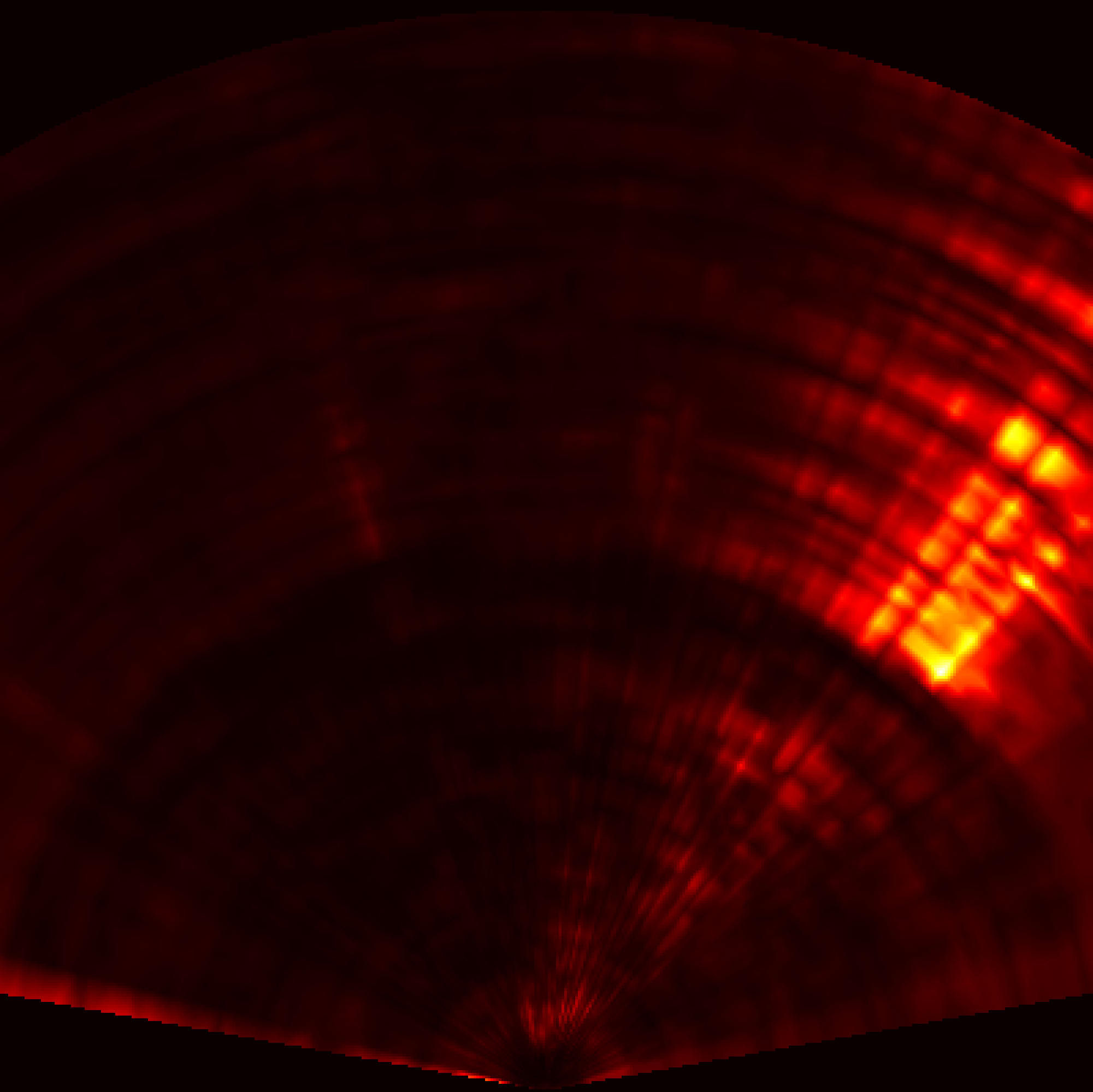} & \includegraphics[width=0.165\linewidth]{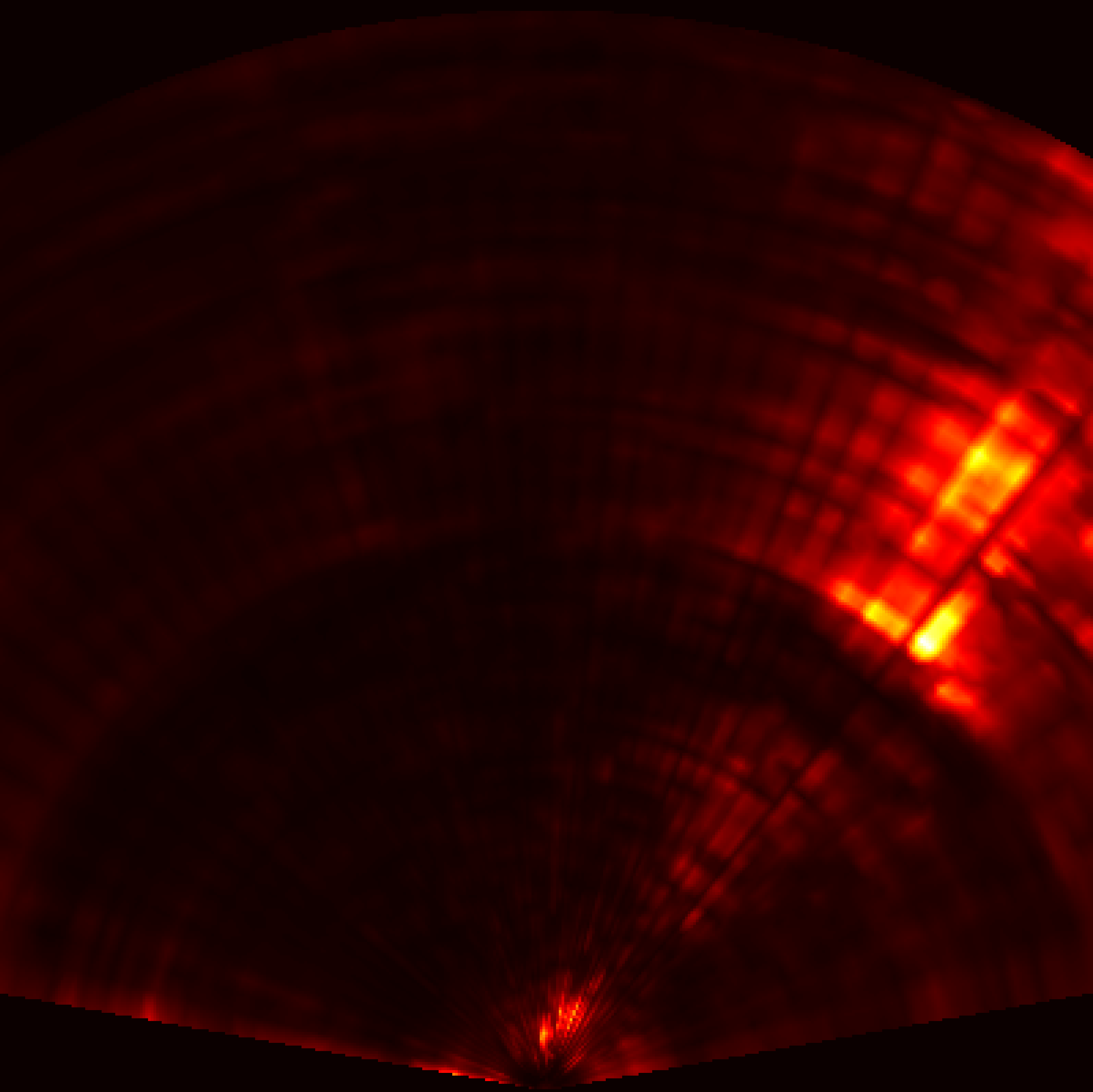} & \includegraphics[width=0.165\linewidth]{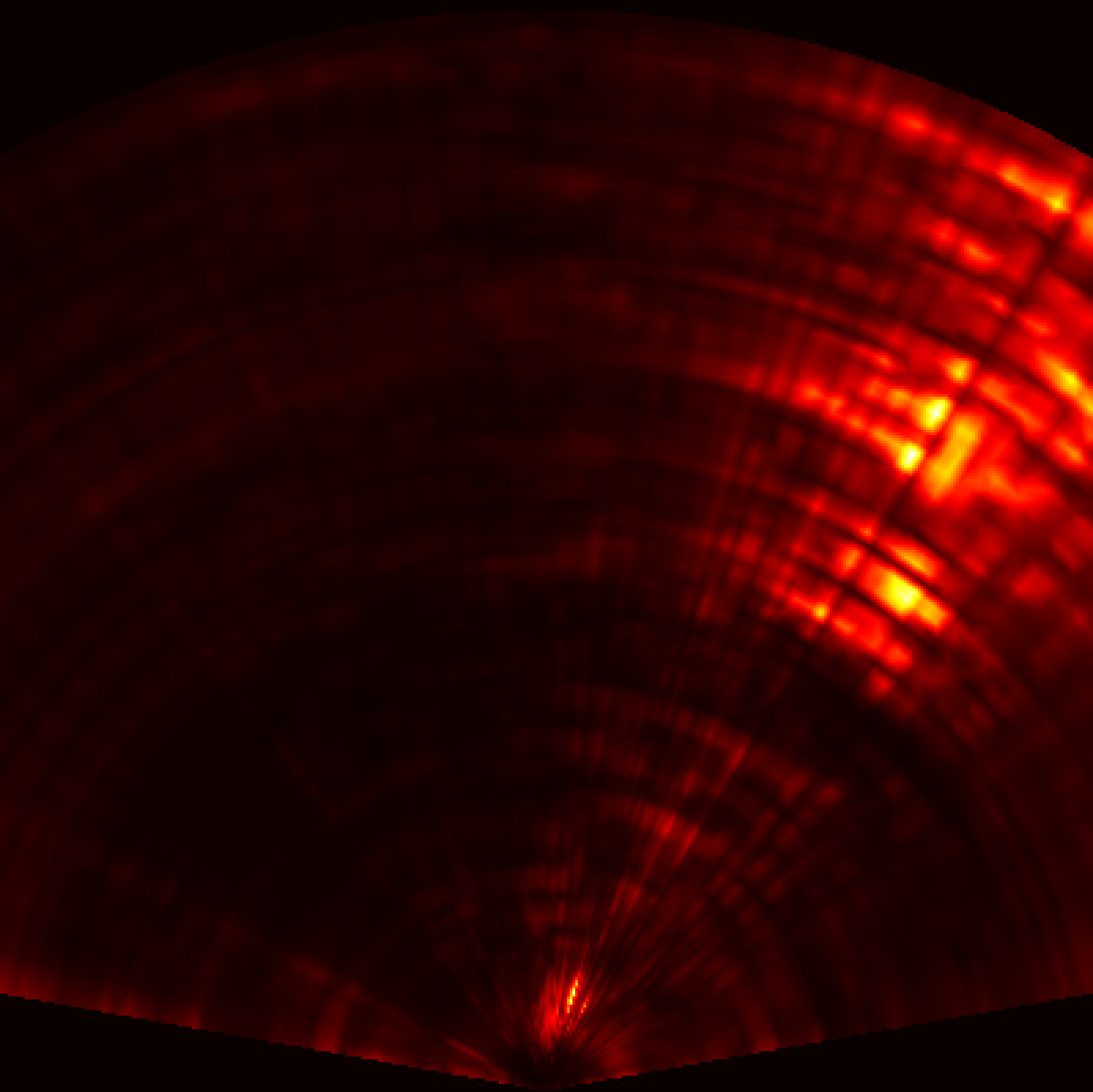} \\
\end{tabular}
\caption{\textbf{Per-frame $|\mathrm{RA}|$ comparison, scene S2\,F160, training frames $F\!-\!4$ to $F\!-\!1$.} Pearson correlation per column, left to right. 3DPS: 0.79, 0.79, 0.80, 0.72; mmIR: 0.55, 0.55, 0.52, 0.45; RadarSplat: 0.61, 0.65, 0.60, 0.57; Radar Fields: 0.08, 0.21, 0.08, 0.35; DART: 0.00, -0.03, -0.05, -0.09.}
\label{fig:supp_s2_f160}
\end{figure}

\begin{figure}[p]
\centering
\setlength{\tabcolsep}{1pt}\renewcommand{\arraystretch}{0.6}
\begin{tabular}{cccccc}
 & {\footnotesize $F\!+\!1$} & {\footnotesize $F\!+\!2$} & {\footnotesize $F\!+\!3$} & {\footnotesize $F\!+\!4$} & \textcolor{red}{\footnotesize\textbf{$F$}} \\
\raisebox{2.0em}{\rotatebox{90}{\scriptsize\textbf{GT}}} & \includegraphics[width=0.165\linewidth]{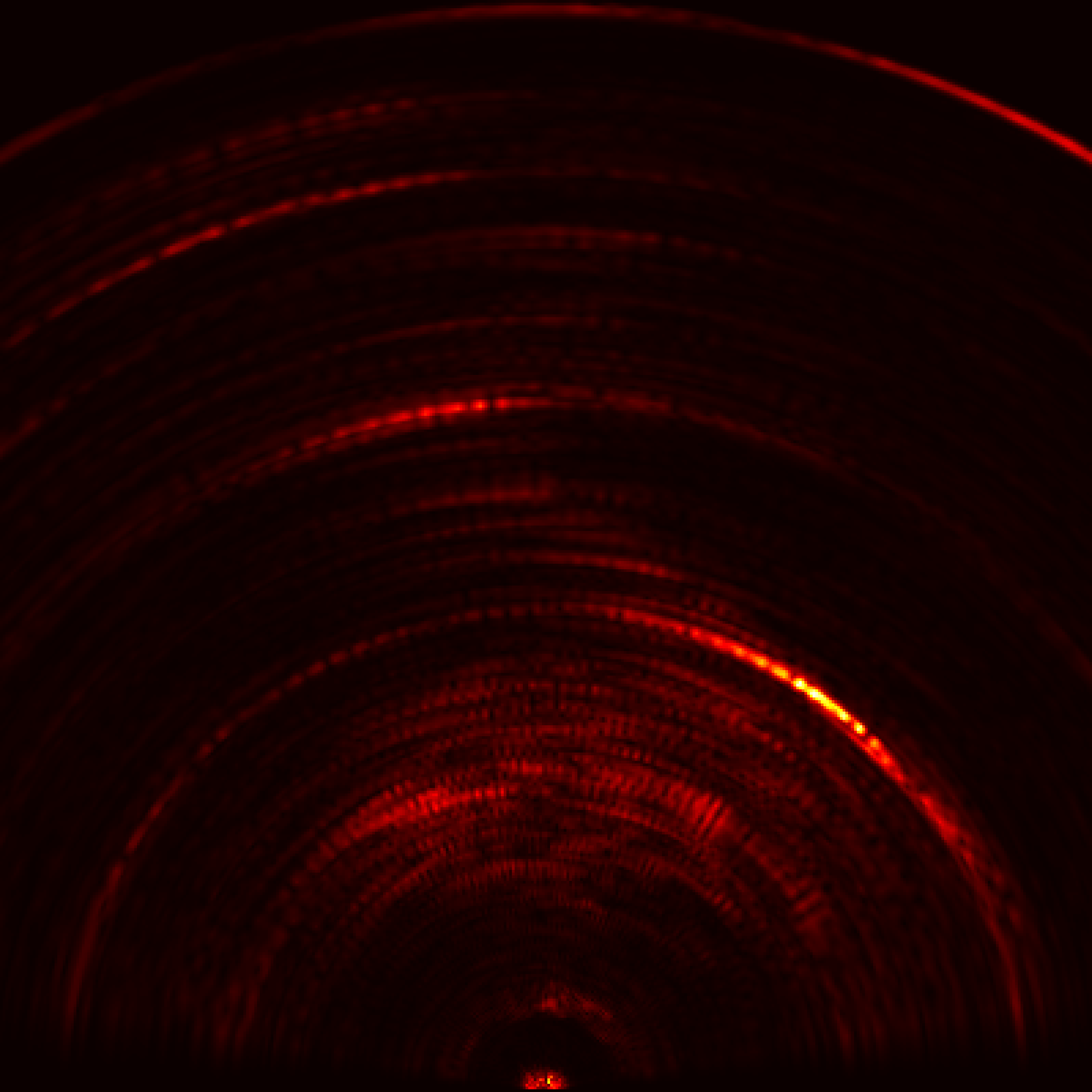} & \includegraphics[width=0.165\linewidth]{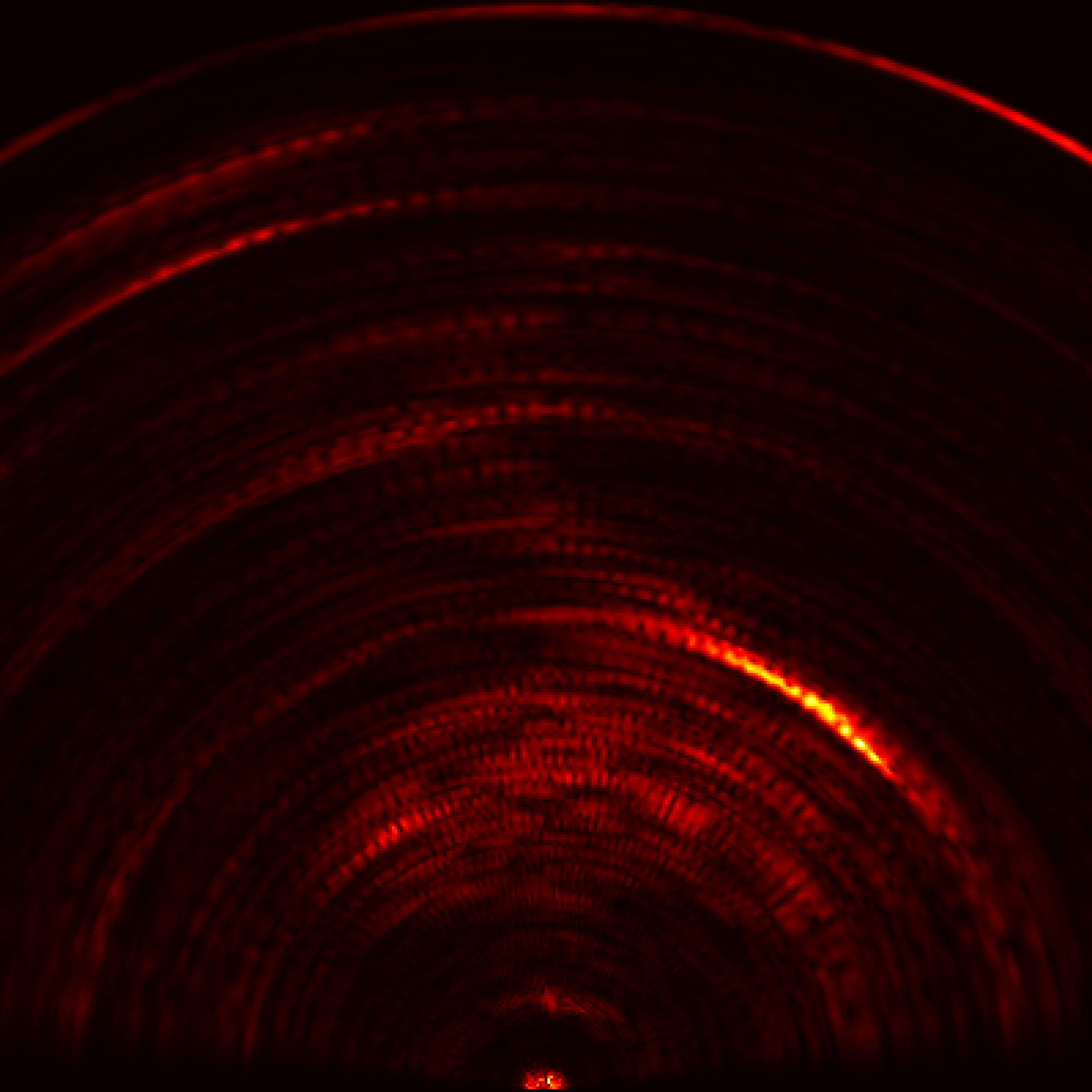} & \includegraphics[width=0.165\linewidth]{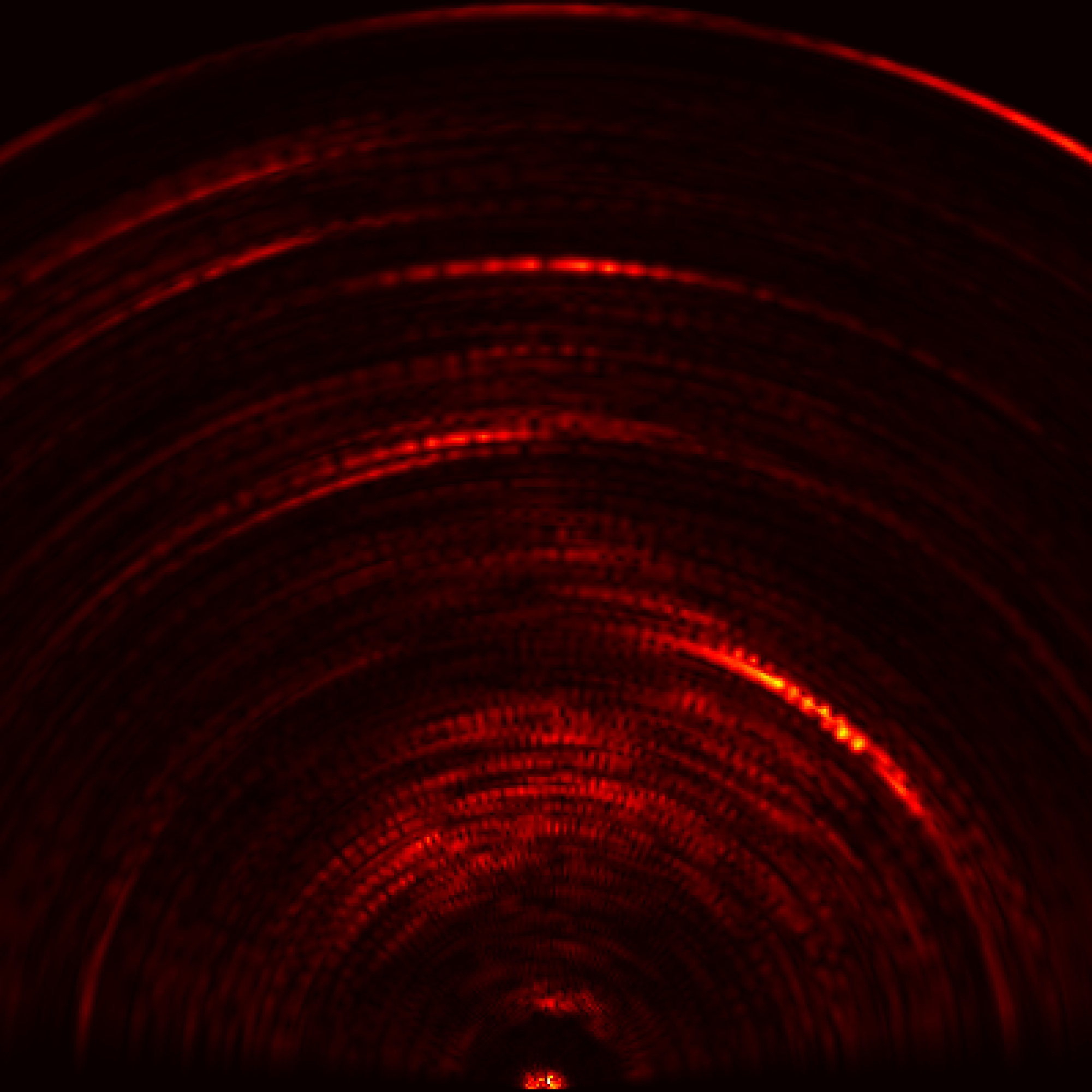} & \includegraphics[width=0.165\linewidth]{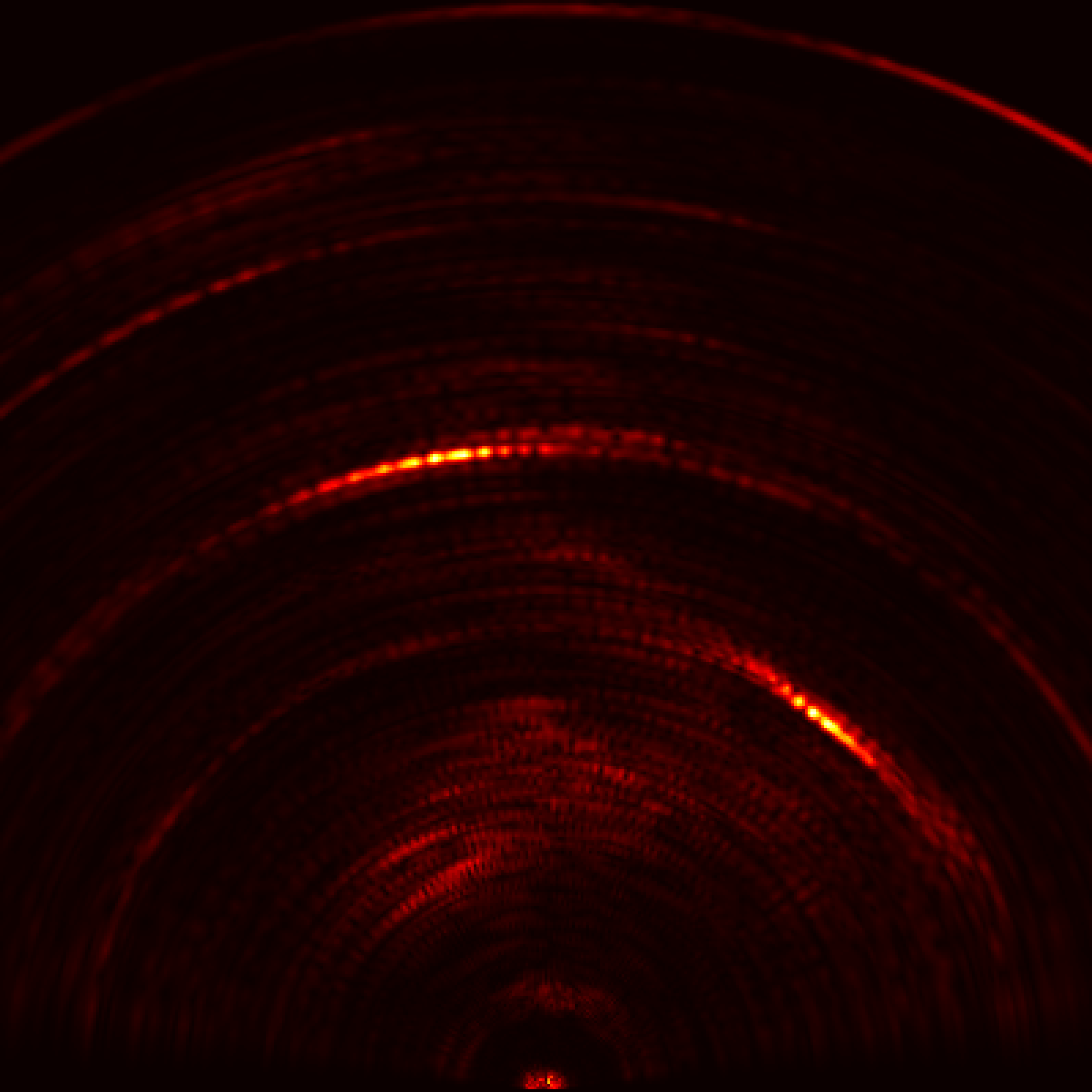} & \includegraphics[width=0.165\linewidth]{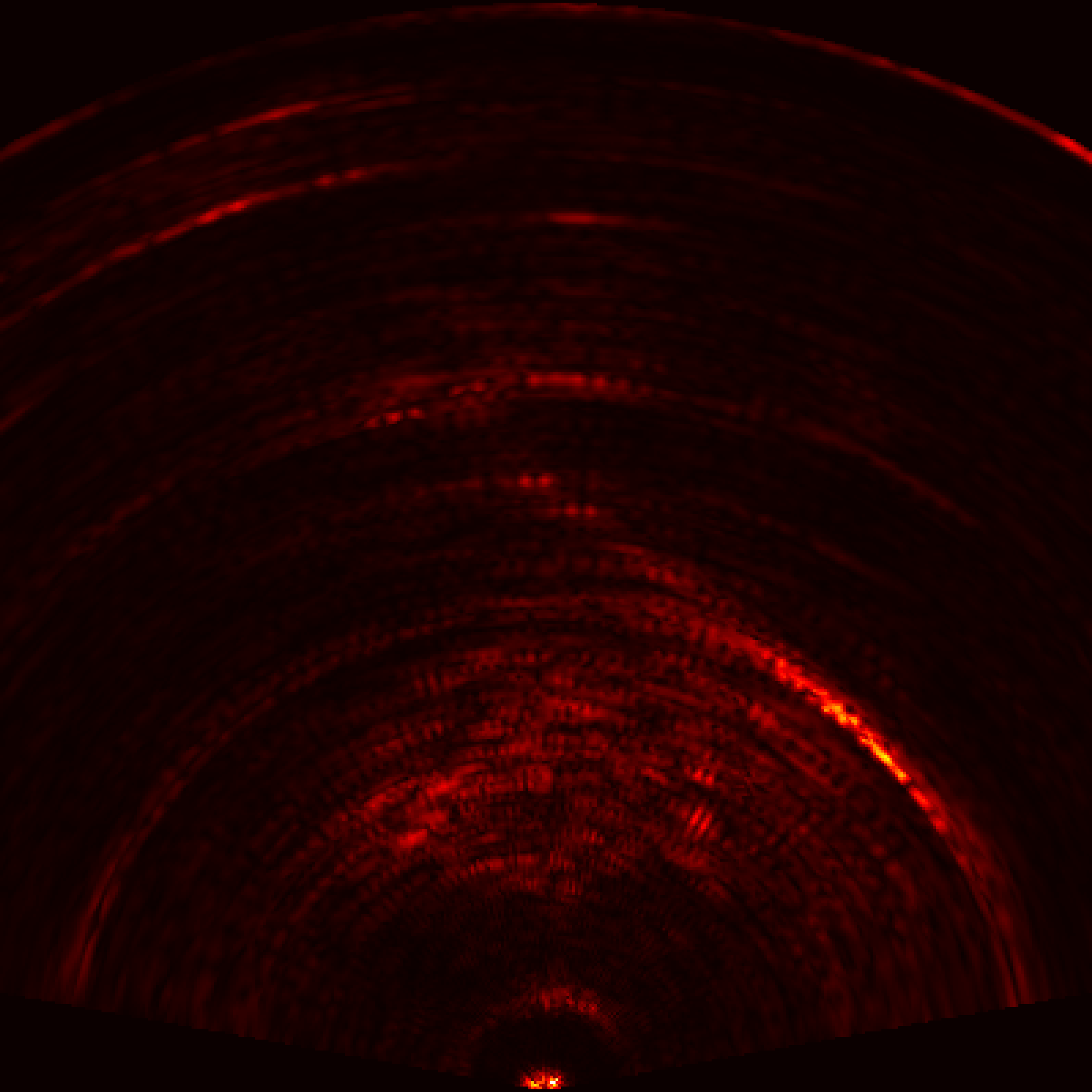} \\
\raisebox{2.0em}{\rotatebox{90}{\scriptsize\textbf{3DPS (Ours)}}} & \includegraphics[width=0.165\linewidth]{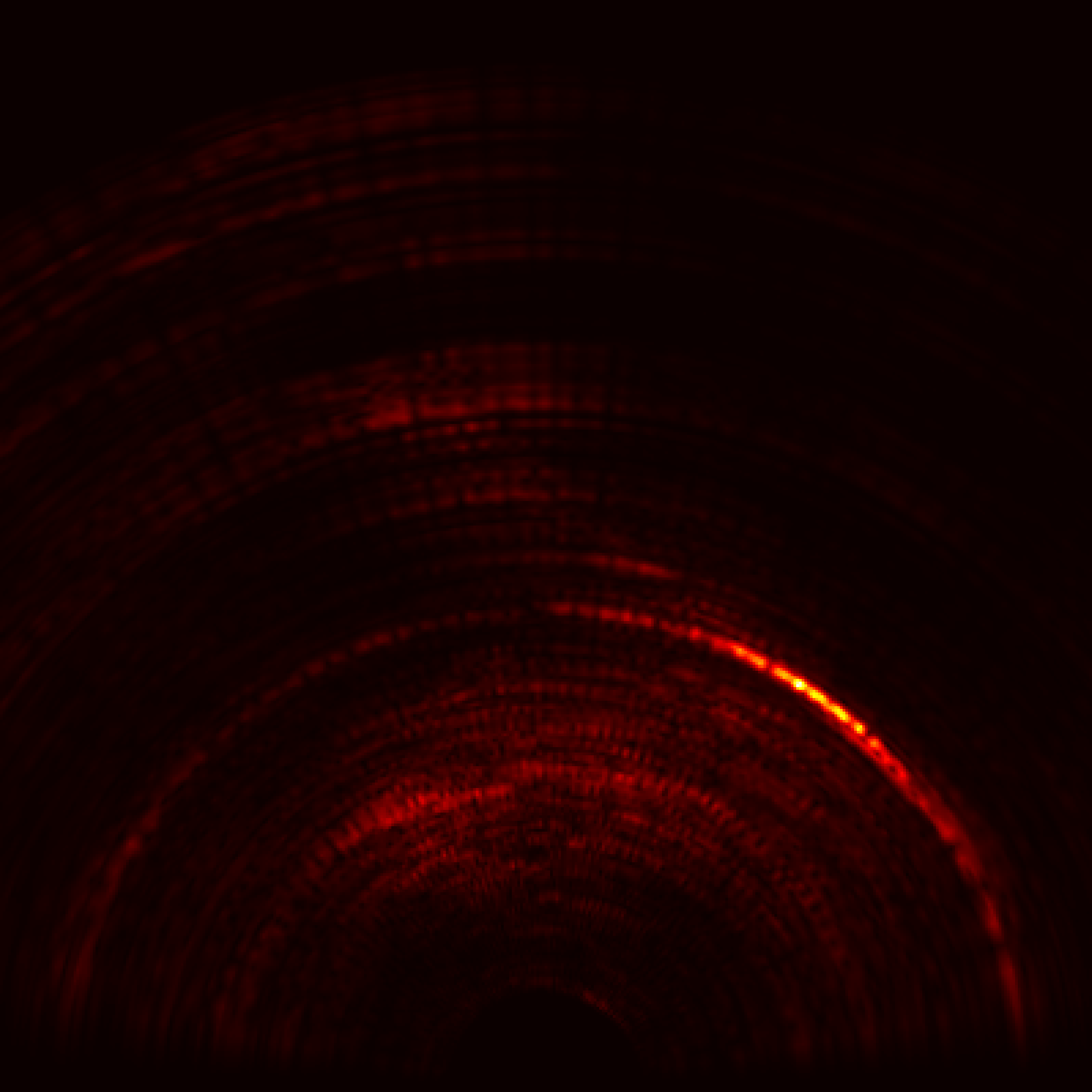} & \includegraphics[width=0.165\linewidth]{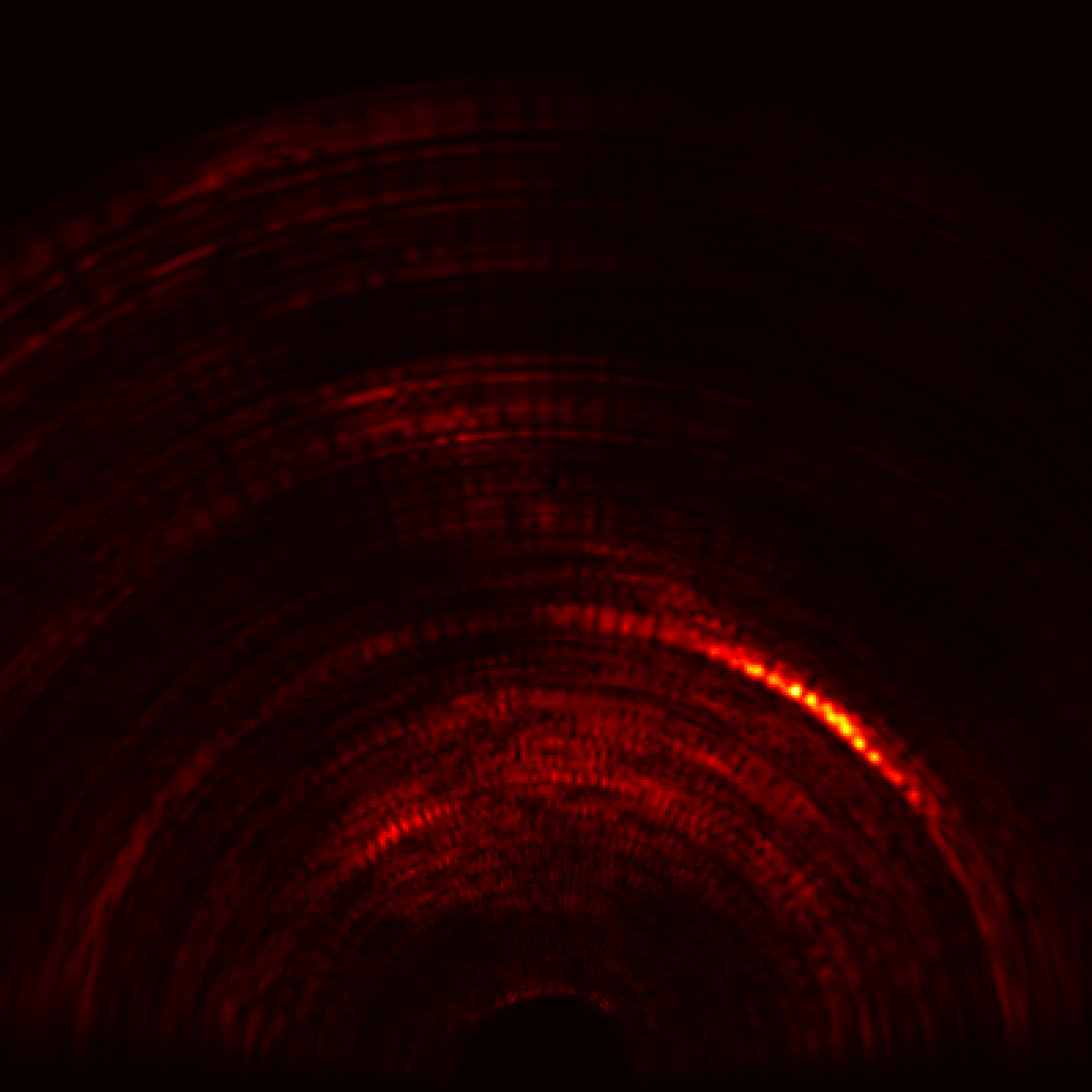} & \includegraphics[width=0.165\linewidth]{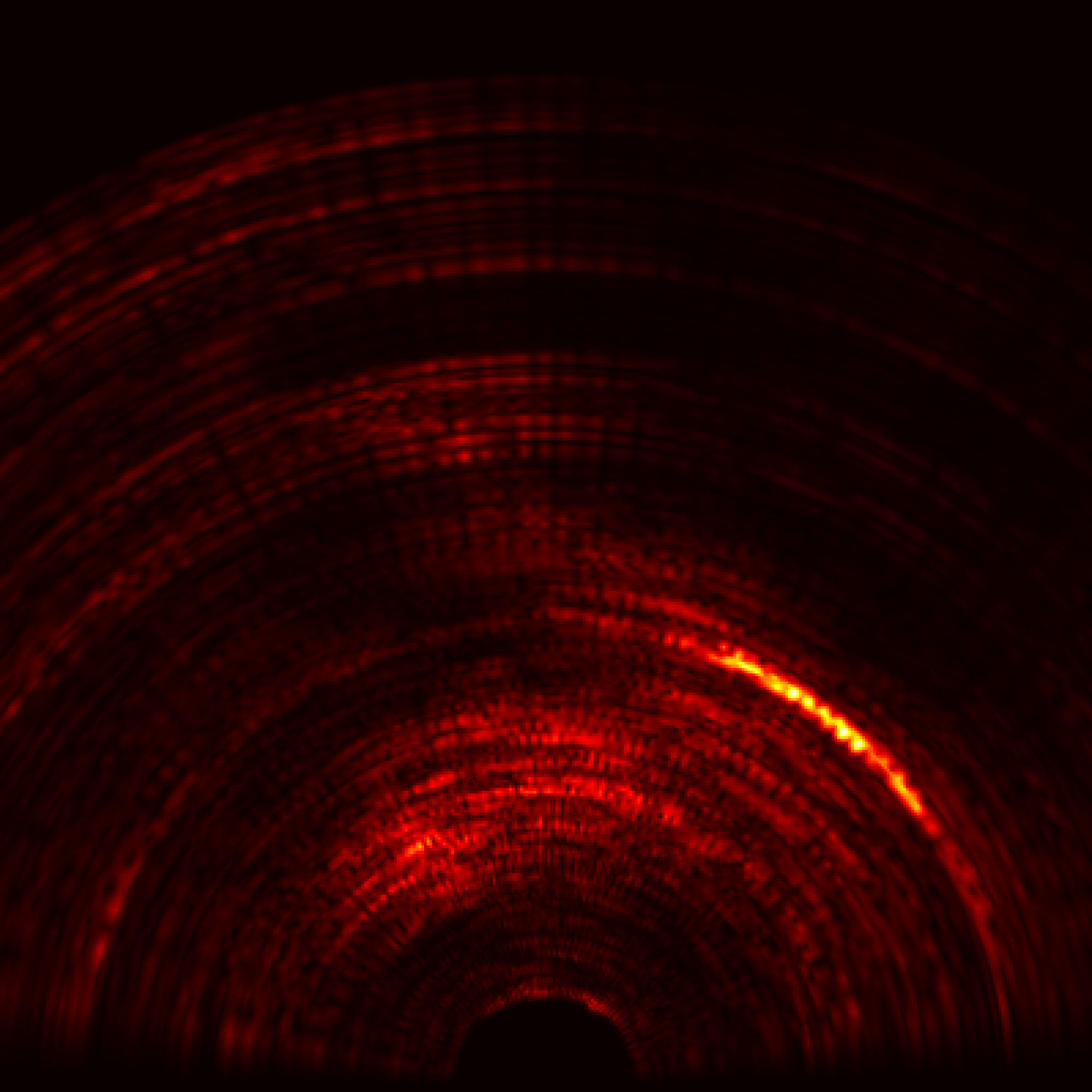} & \includegraphics[width=0.165\linewidth]{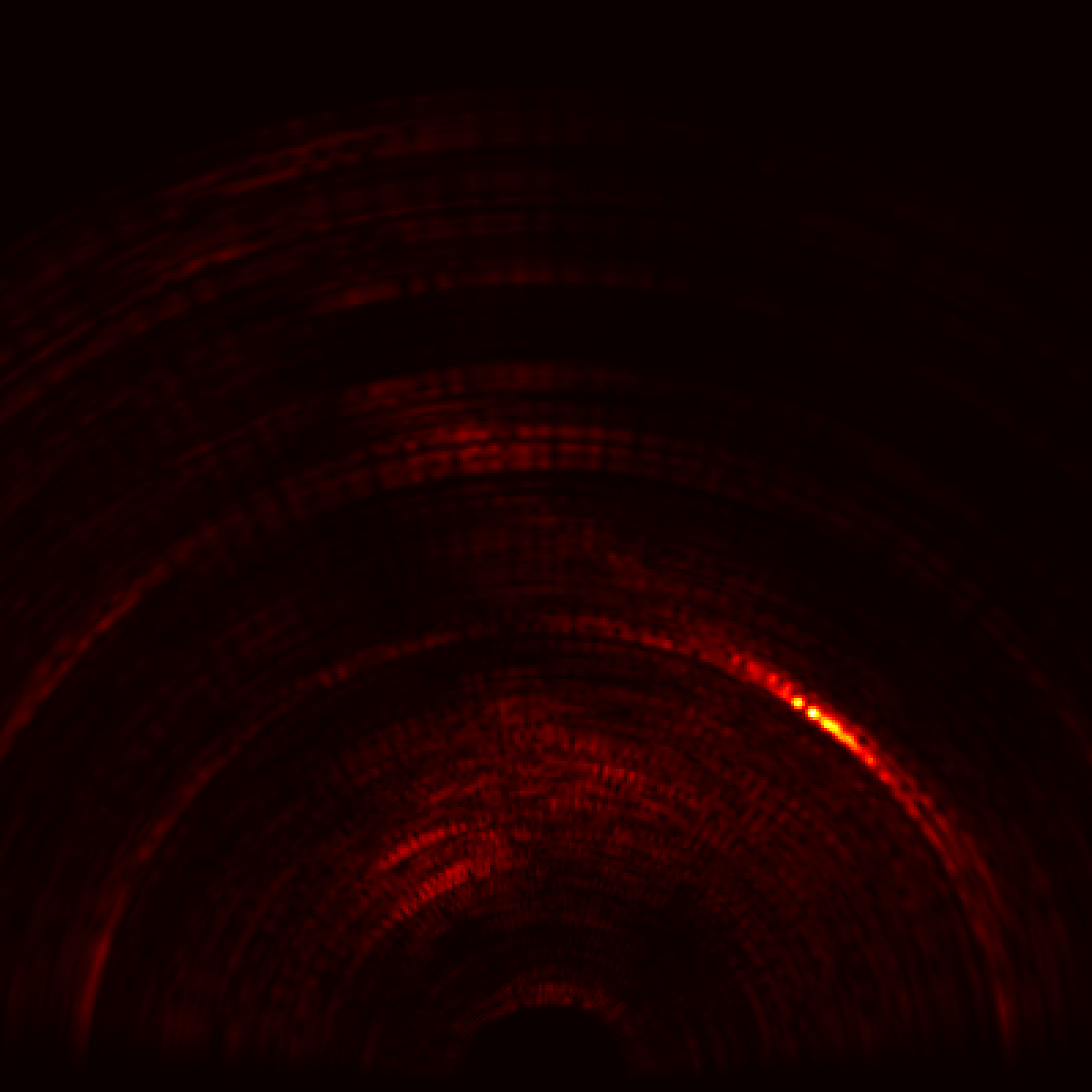} & \includegraphics[width=0.165\linewidth]{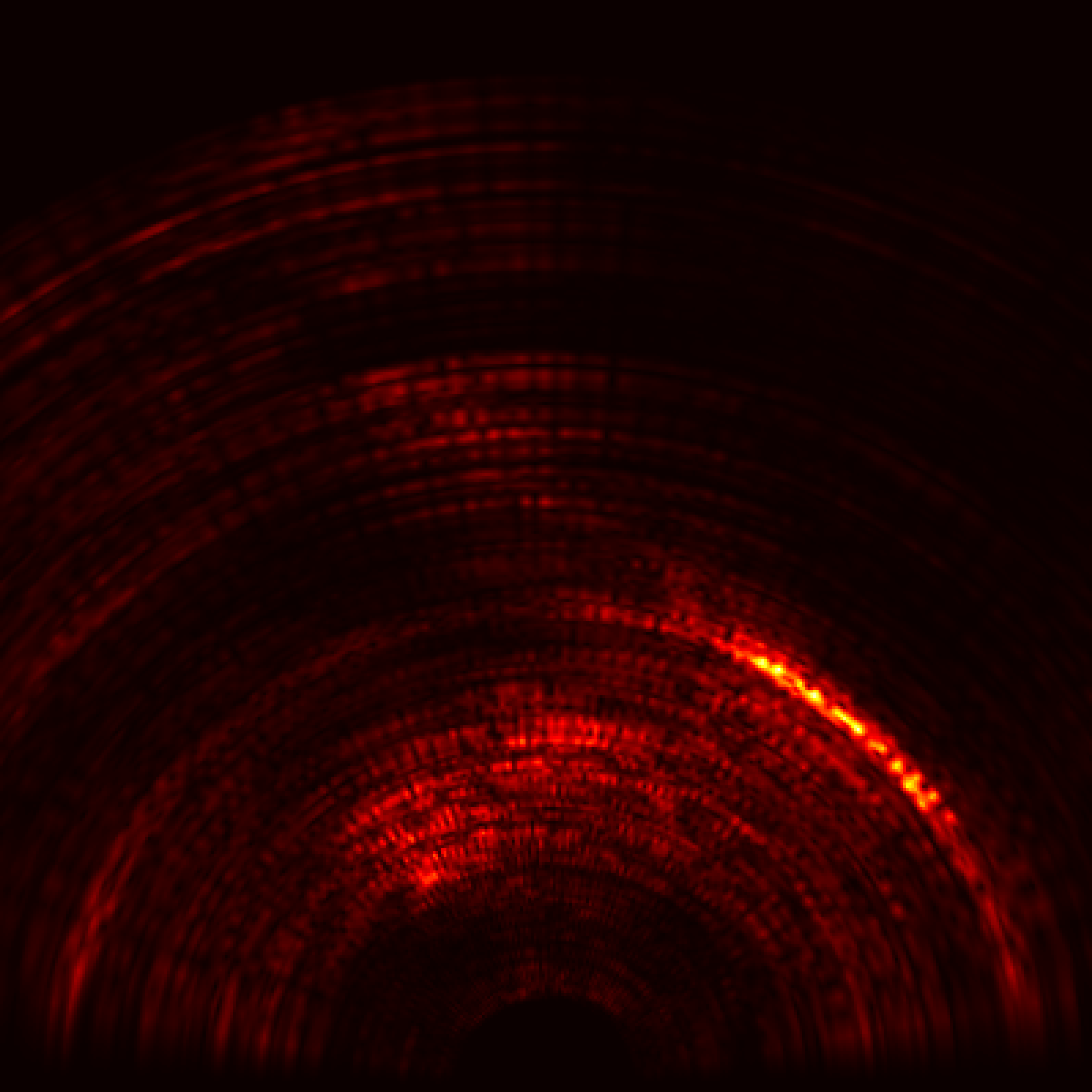} \\
\raisebox{2.0em}{\rotatebox{90}{\scriptsize\textbf{mmIR}}} & \includegraphics[width=0.165\linewidth]{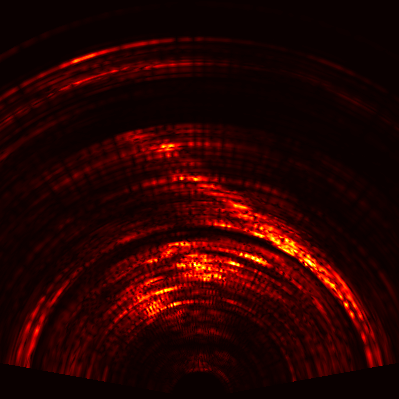} & \includegraphics[width=0.165\linewidth]{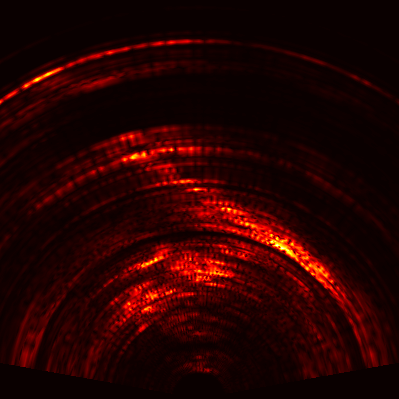} & \includegraphics[width=0.165\linewidth]{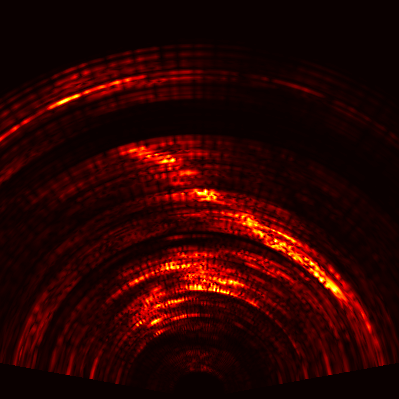} & \includegraphics[width=0.165\linewidth]{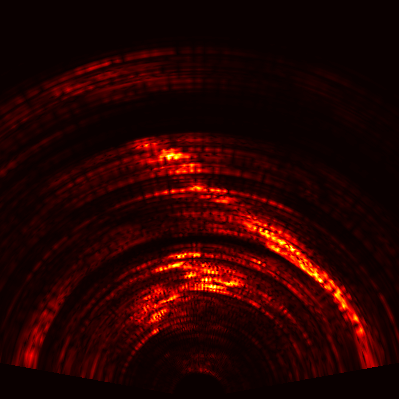} & \includegraphics[width=0.165\linewidth]{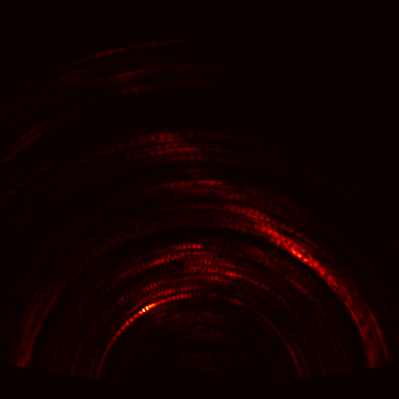} \\
\raisebox{2.0em}{\rotatebox{90}{\scriptsize\textbf{RadarSplat}}} & \includegraphics[width=0.165\linewidth]{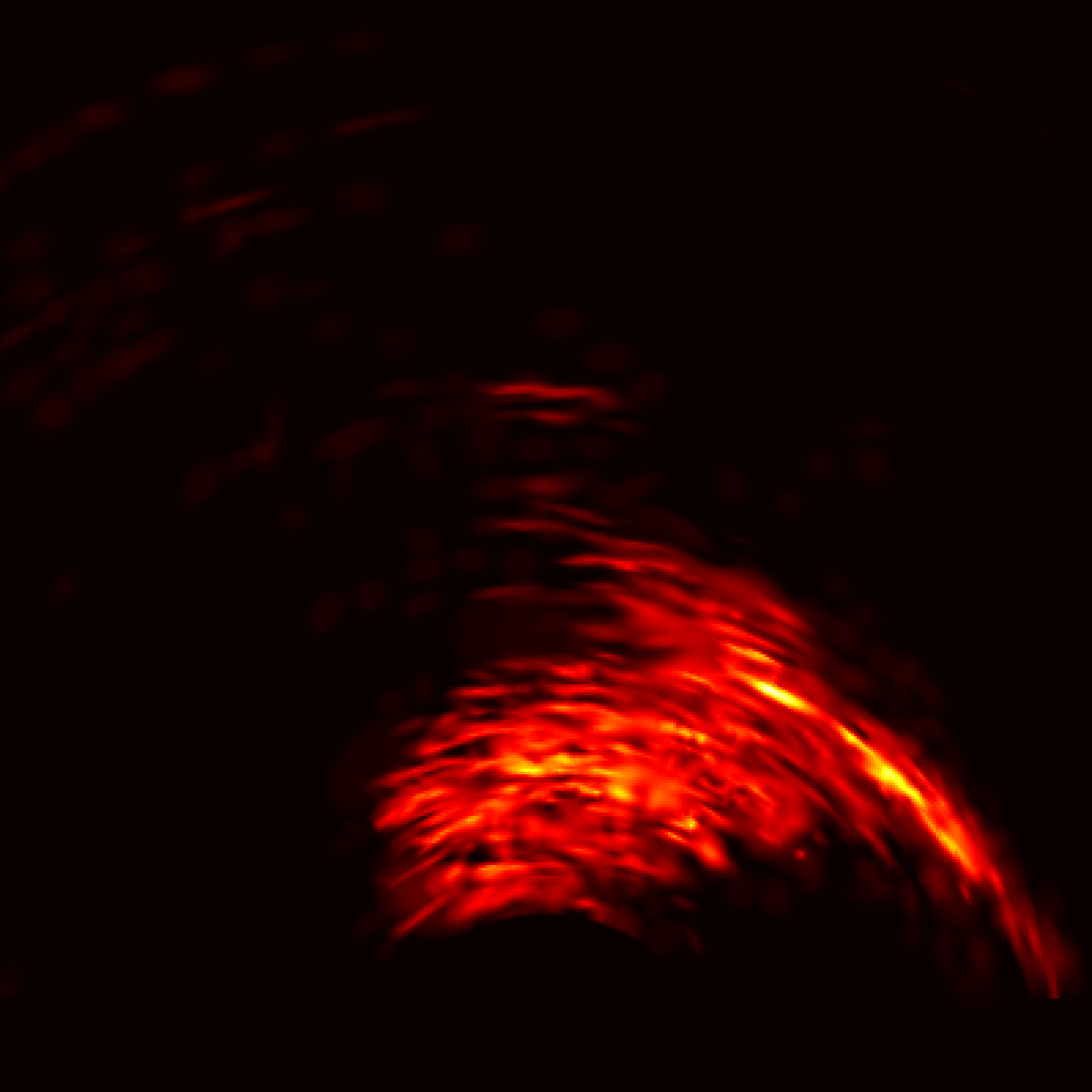} & \includegraphics[width=0.165\linewidth]{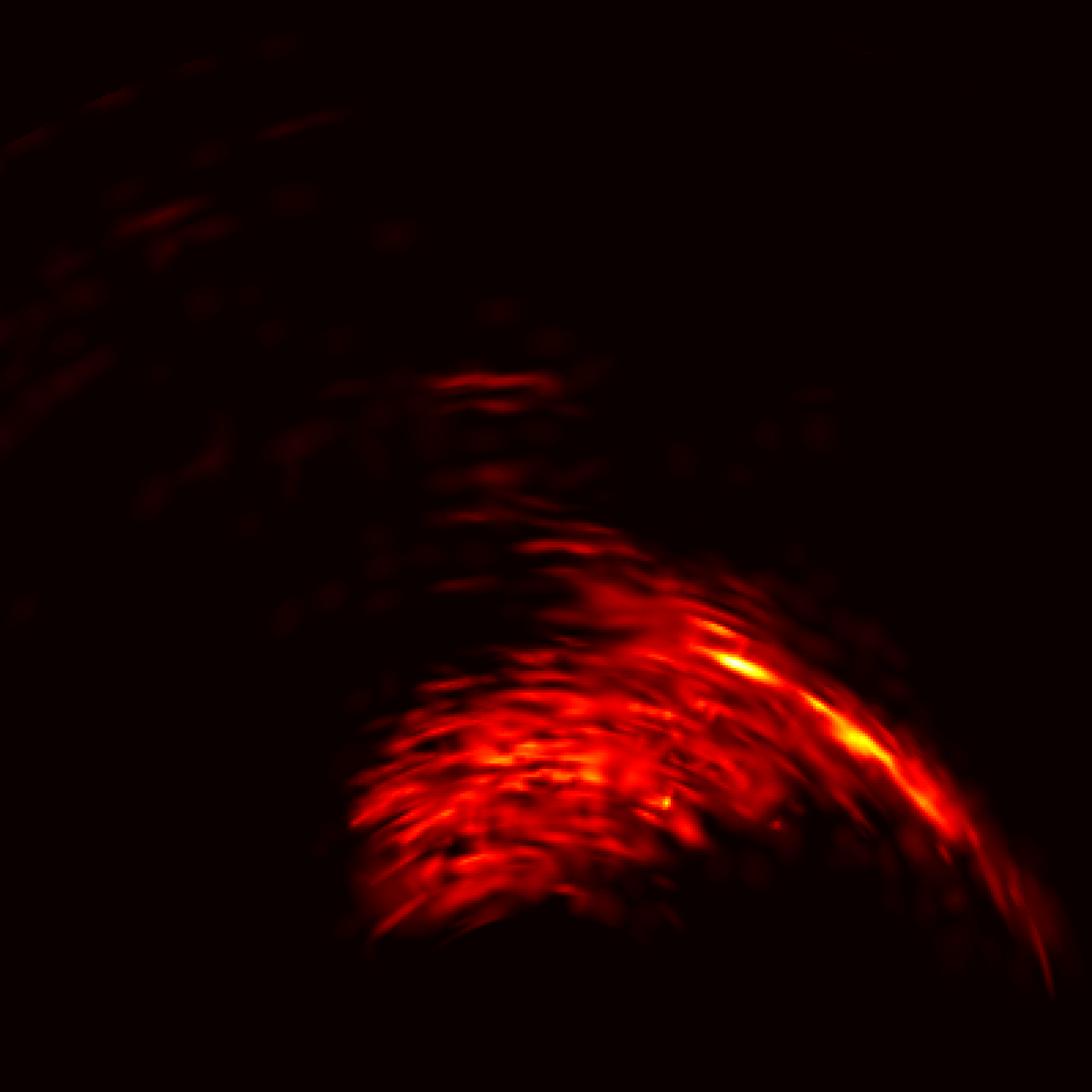} & \includegraphics[width=0.165\linewidth]{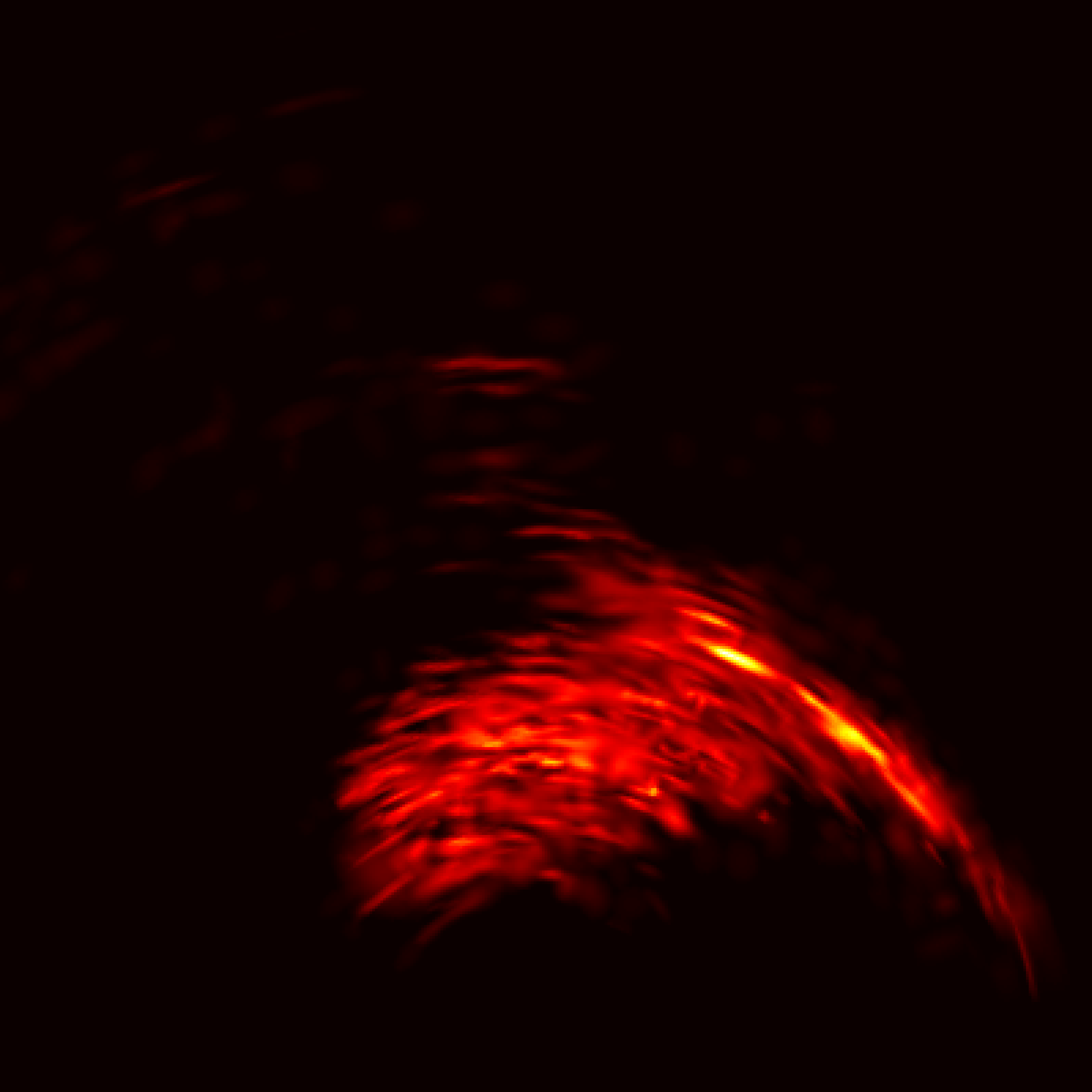} & \includegraphics[width=0.165\linewidth]{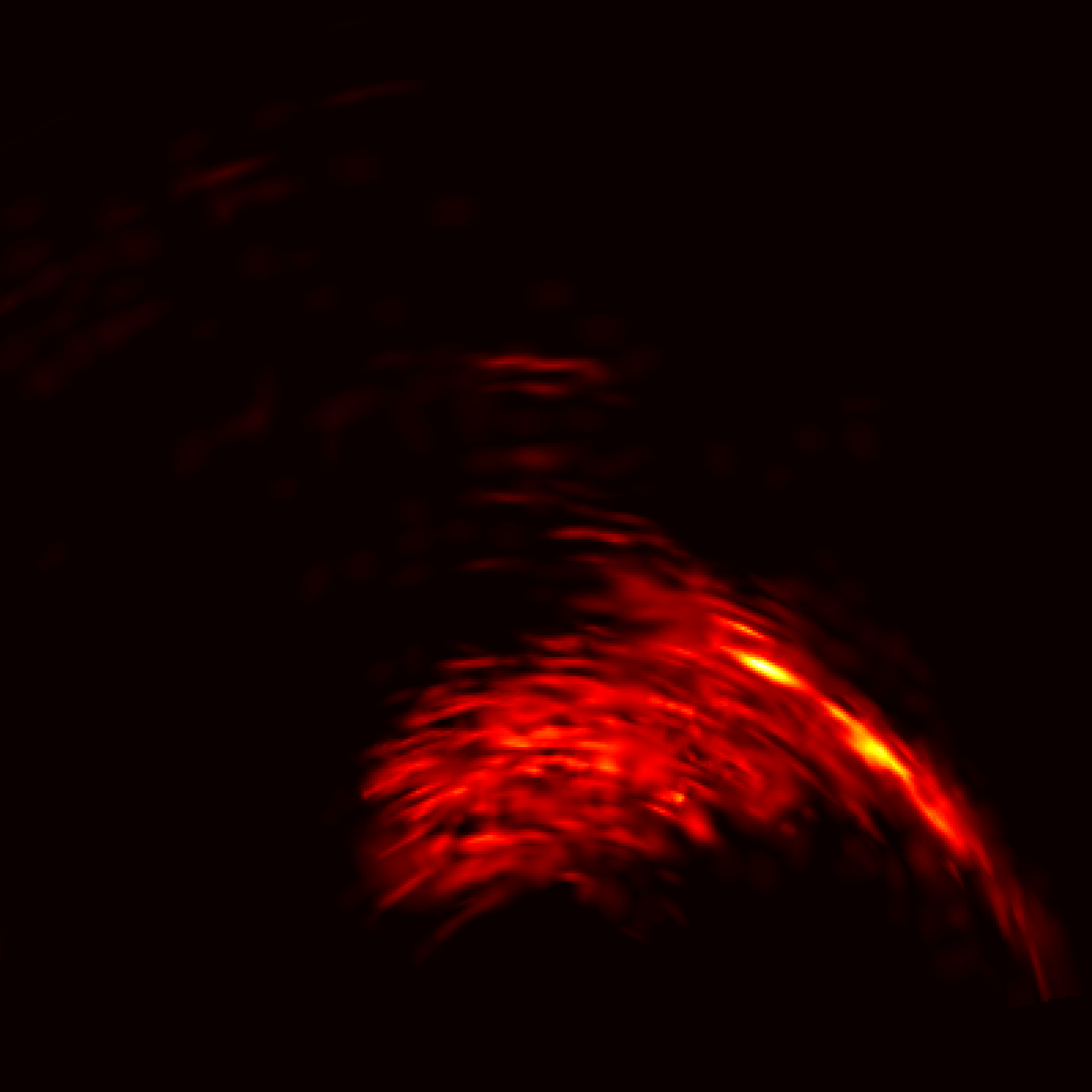} & \includegraphics[width=0.165\linewidth]{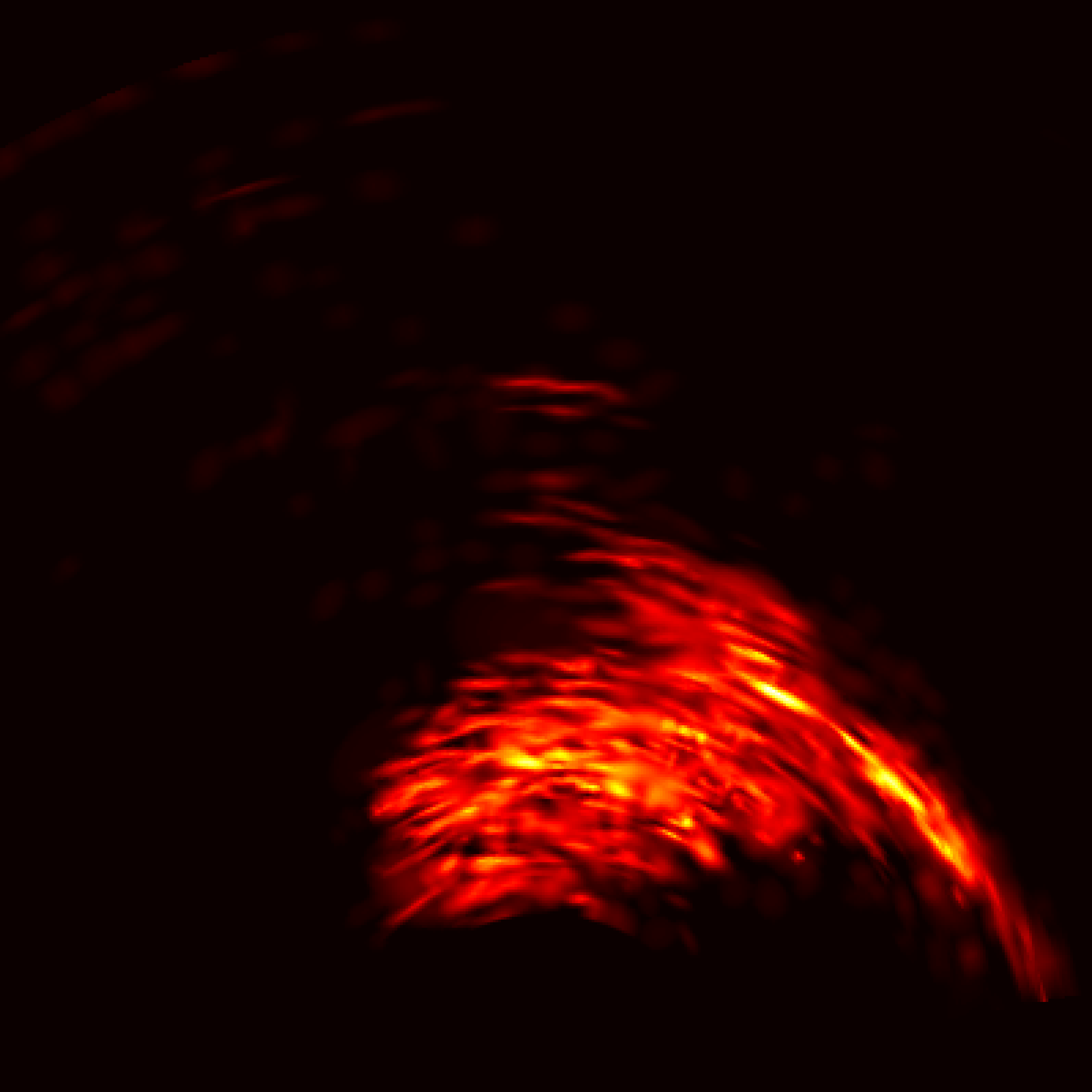} \\
\raisebox{2.0em}{\rotatebox{90}{\scriptsize\textbf{Radar Fields}}} & \includegraphics[width=0.165\linewidth]{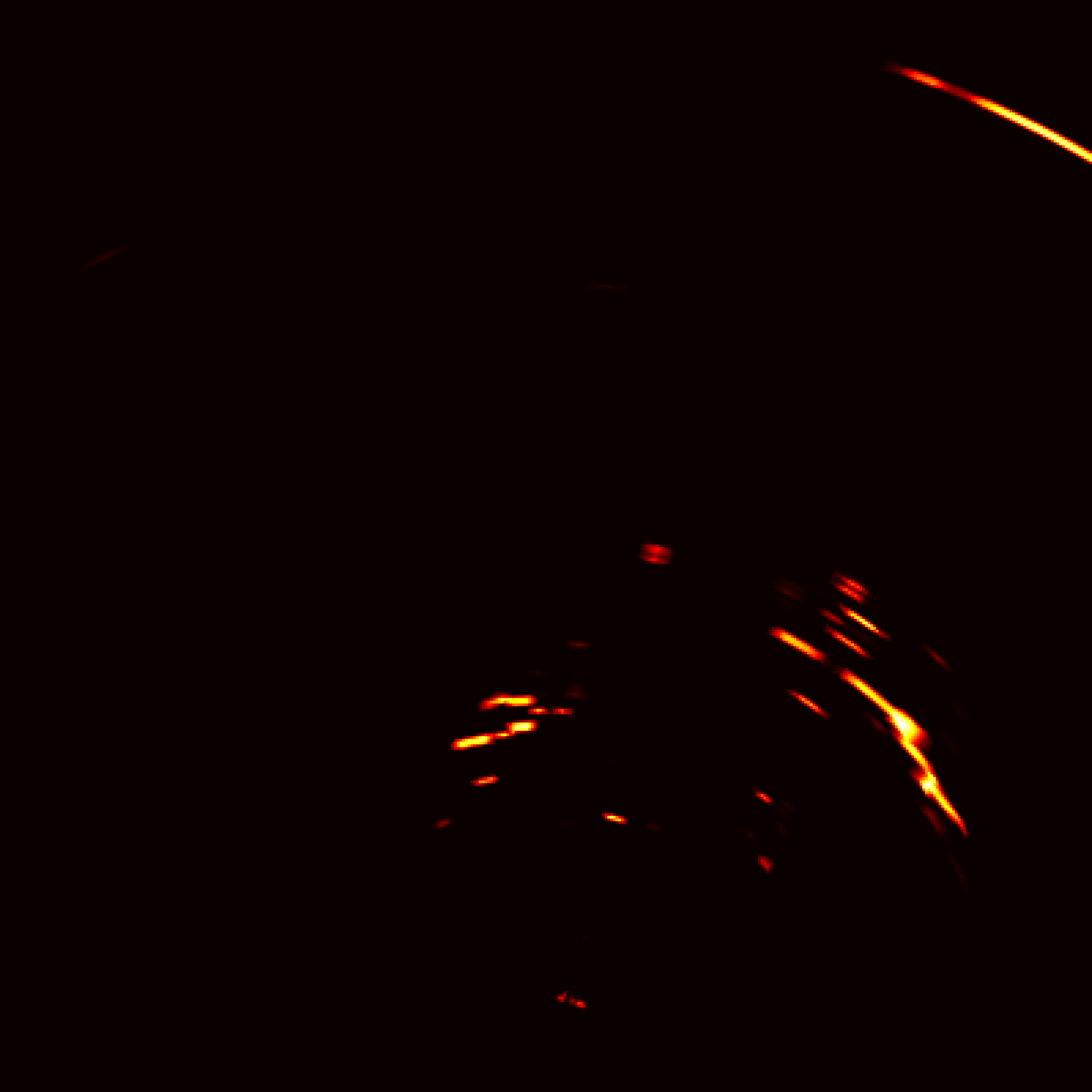} & \includegraphics[width=0.165\linewidth]{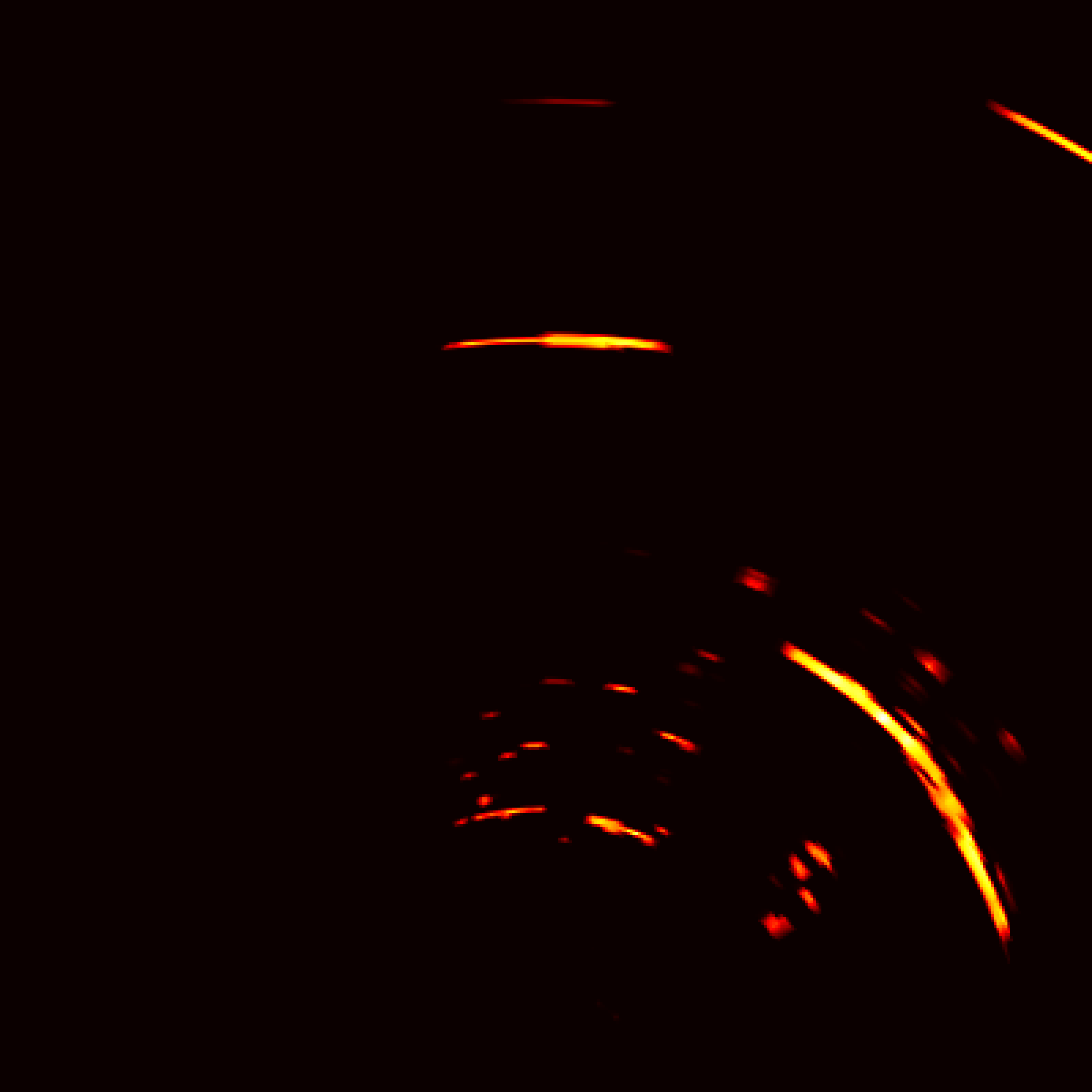} & \includegraphics[width=0.165\linewidth]{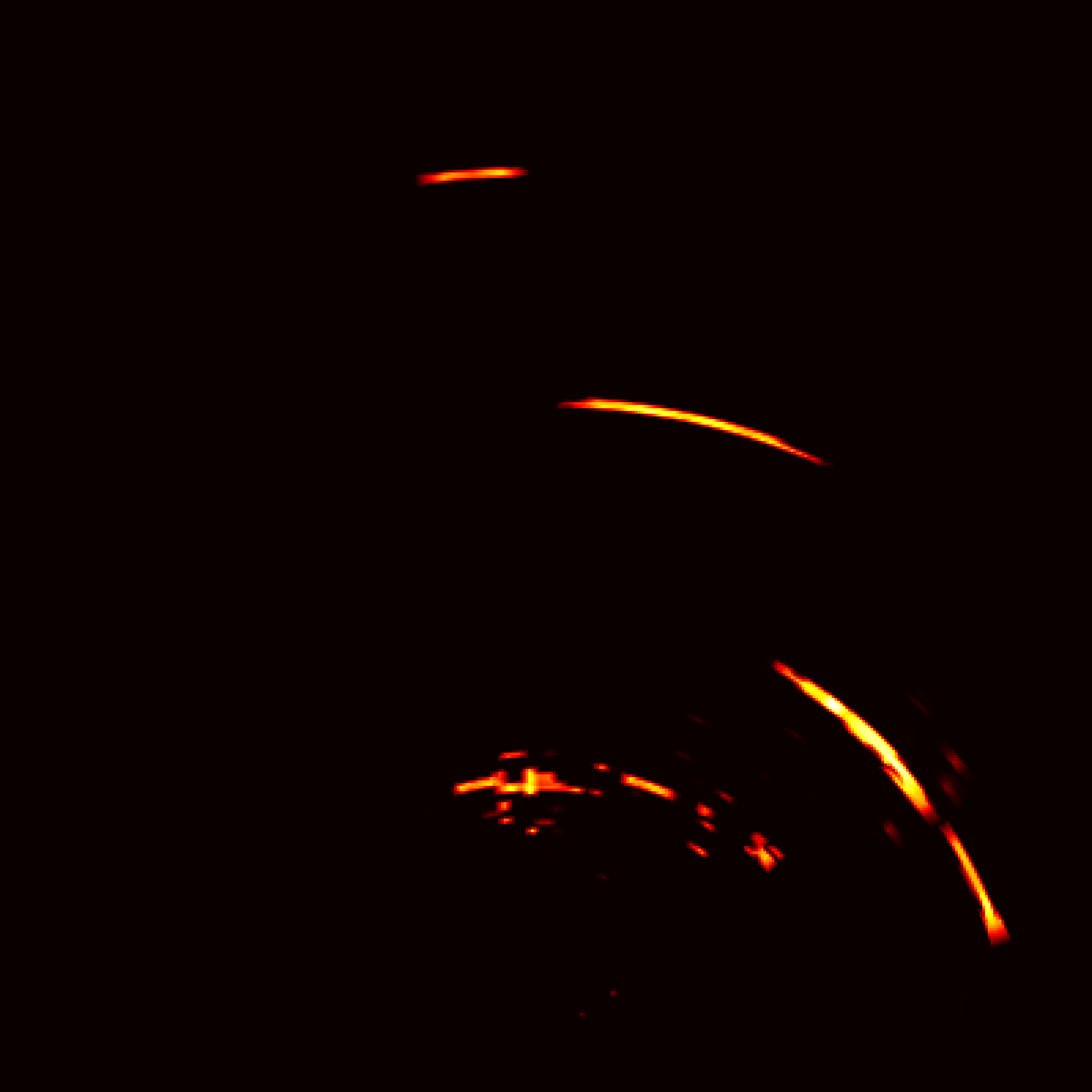} & \includegraphics[width=0.165\linewidth]{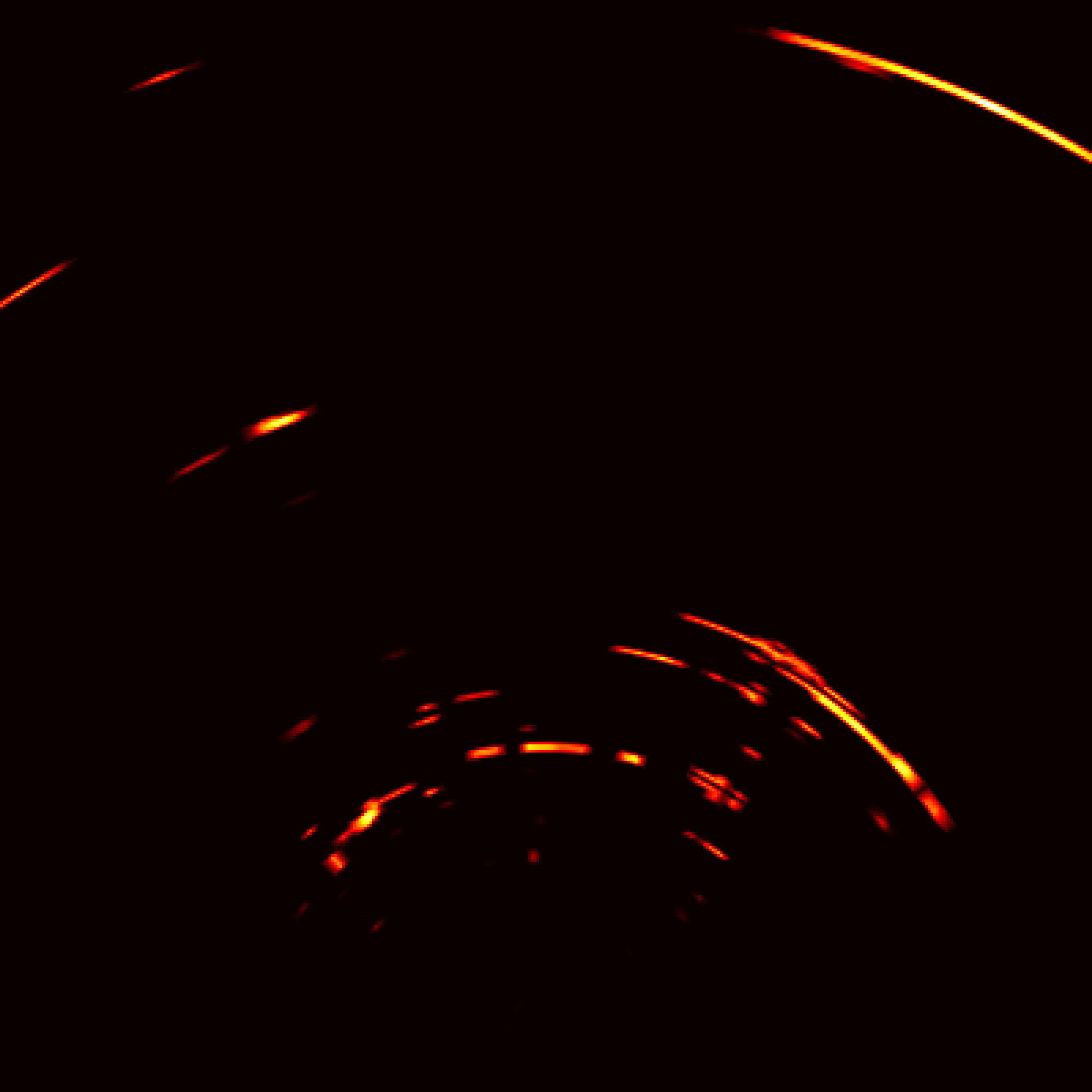} & \includegraphics[width=0.165\linewidth]{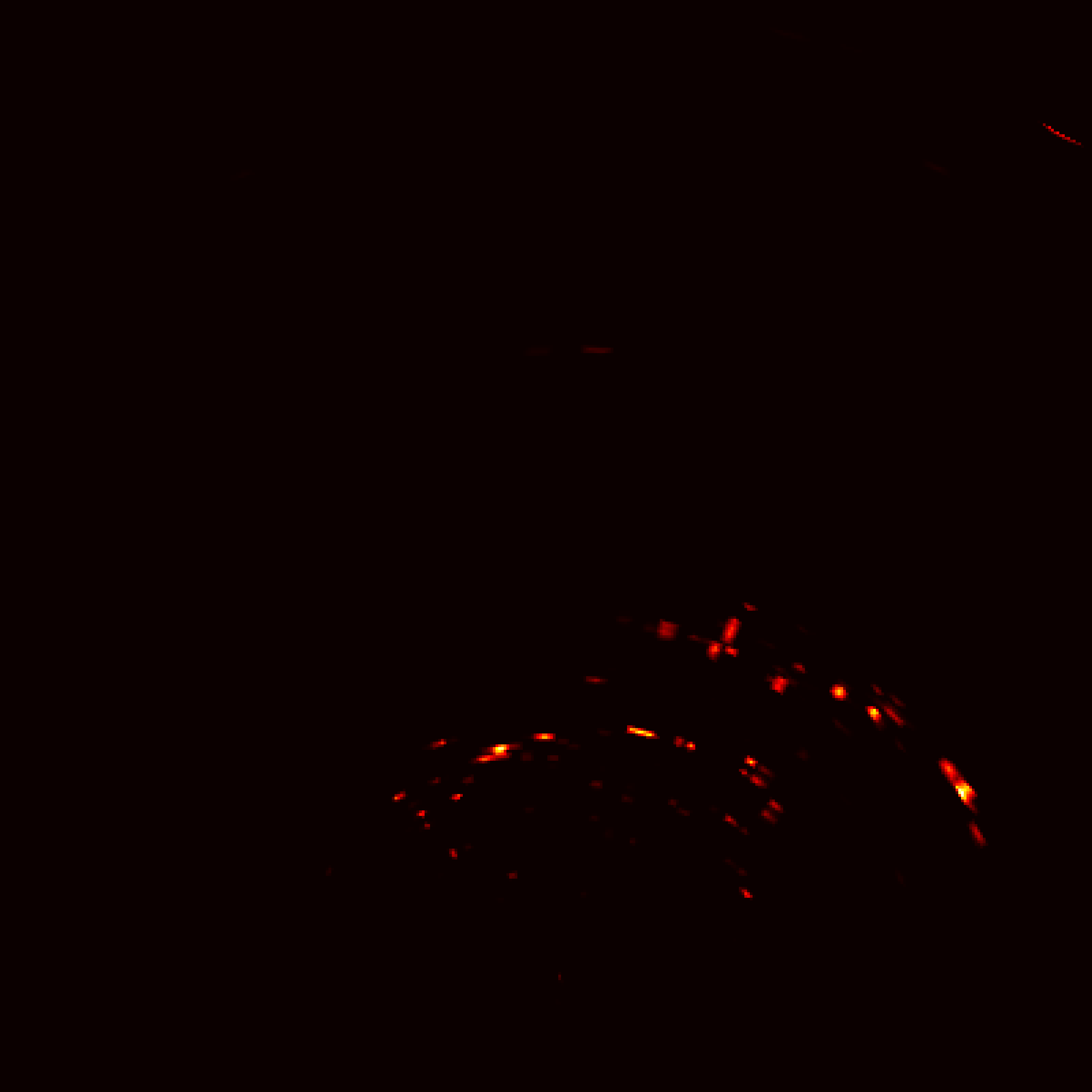} \\
\raisebox{2.0em}{\rotatebox{90}{\scriptsize\textbf{DART}}} & \includegraphics[width=0.165\linewidth]{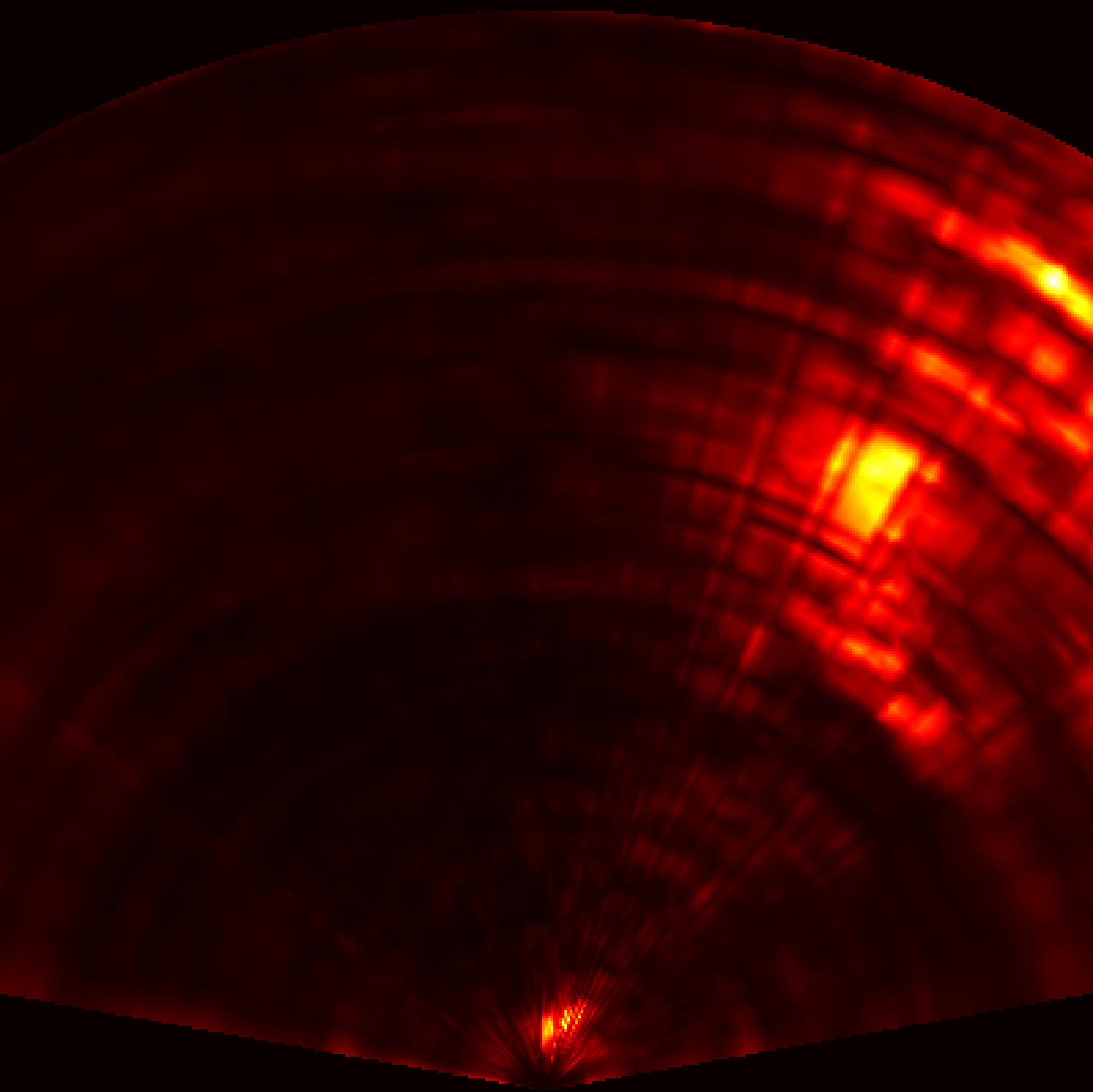} & \includegraphics[width=0.165\linewidth]{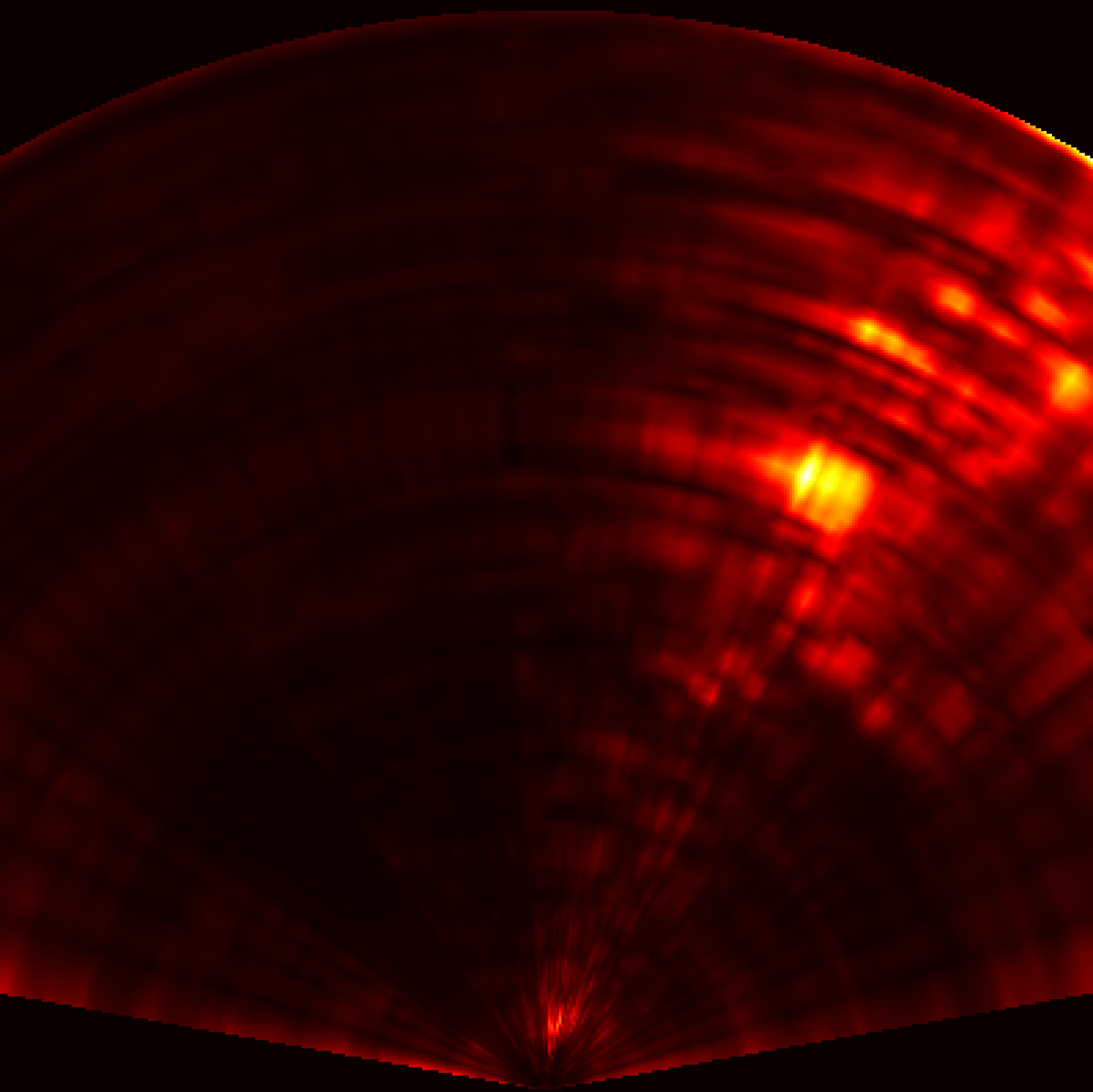} & \includegraphics[width=0.165\linewidth]{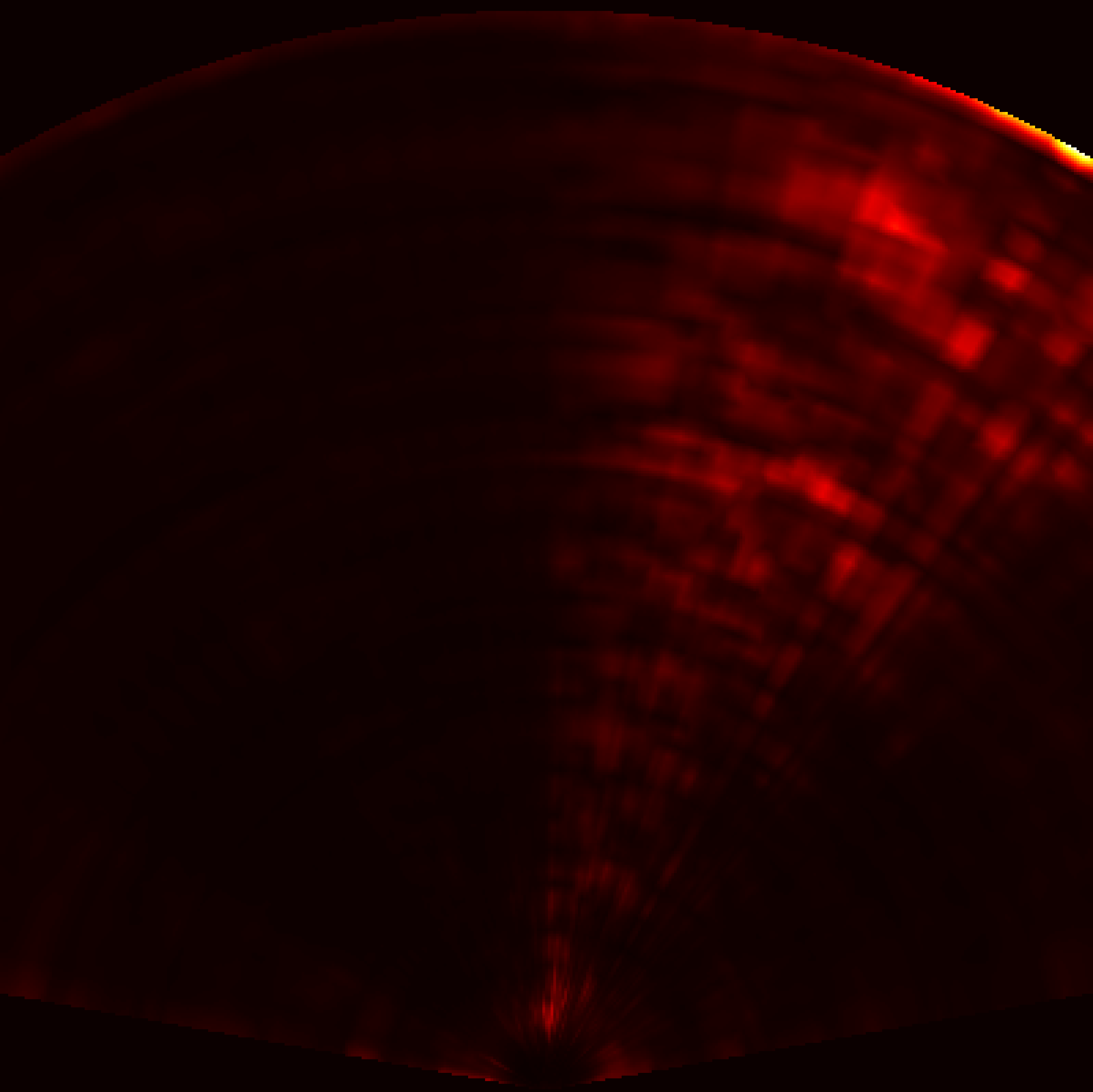} & \includegraphics[width=0.165\linewidth]{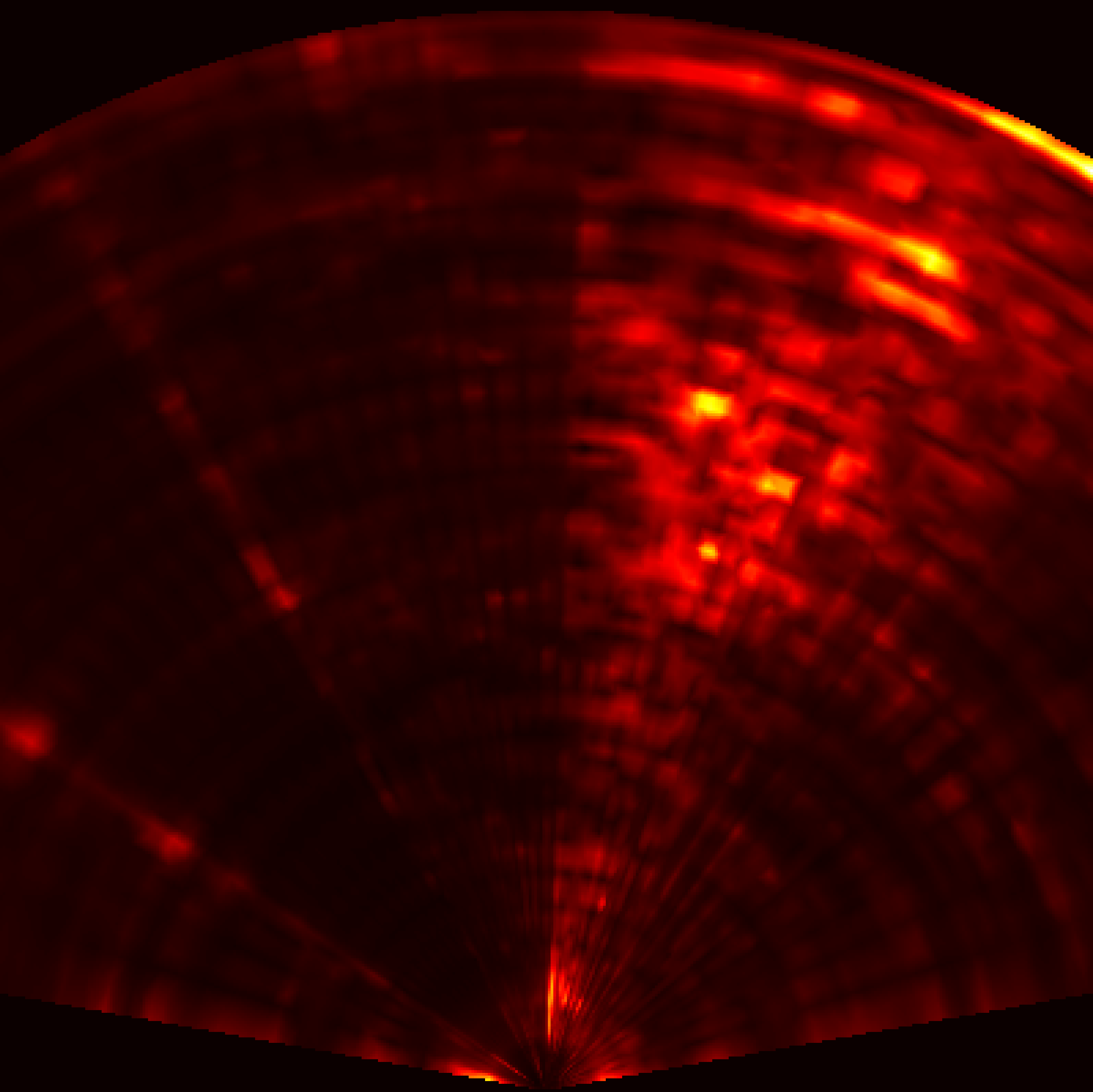} & \includegraphics[width=0.165\linewidth]{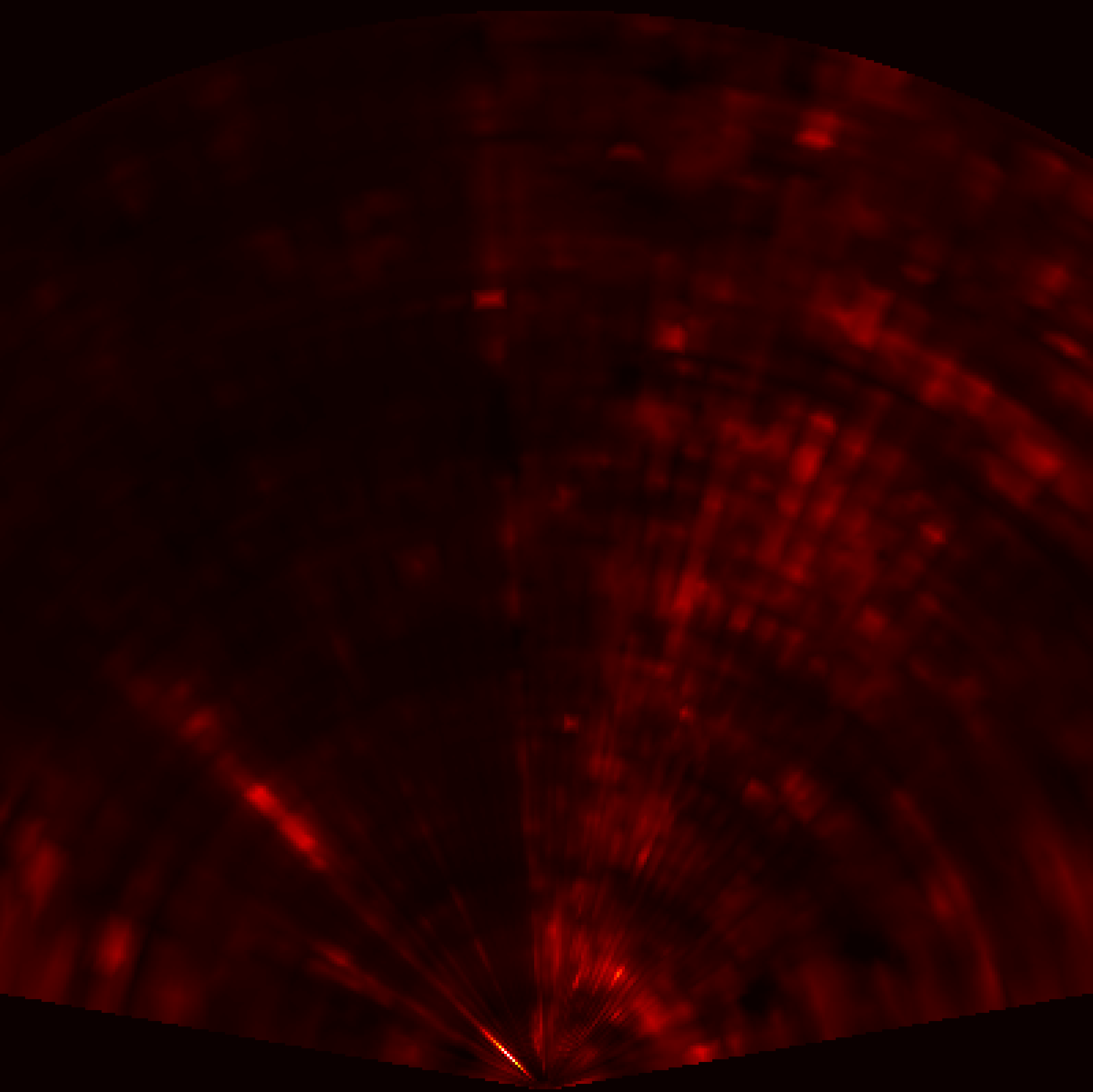} \\
\end{tabular}
\caption{\textbf{Per-frame $|\mathrm{RA}|$ comparison, scene S2\,F160, training frames $F\!+\!1$ to $F\!+\!4$ and held-out test frame $F$ (red header).} Pearson correlation per column, left to right. 3DPS: 0.85, 0.83, 0.76, 0.73, 0.66; mmIR: 0.59, 0.59, 0.53, 0.47, 0.56; RadarSplat: 0.57, 0.68, 0.59, 0.44, 0.56; Radar Fields: 0.14, 0.12, 0.14, 0.19, 0.10; DART: -0.08, -0.05, 0.00, -0.03, 0.08.}
\label{fig:supp_s2_f160_b}
\end{figure}

\begin{figure}[p]
\centering
\setlength{\tabcolsep}{1pt}\renewcommand{\arraystretch}{0.6}
\begin{tabular}{ccccc}
 & {\footnotesize $F\!-\!4$} & {\footnotesize $F\!-\!3$} & {\footnotesize $F\!-\!2$} & {\footnotesize $F\!-\!1$} \\
\raisebox{2.0em}{\rotatebox{90}{\scriptsize\textbf{GT}}} & \includegraphics[width=0.165\linewidth]{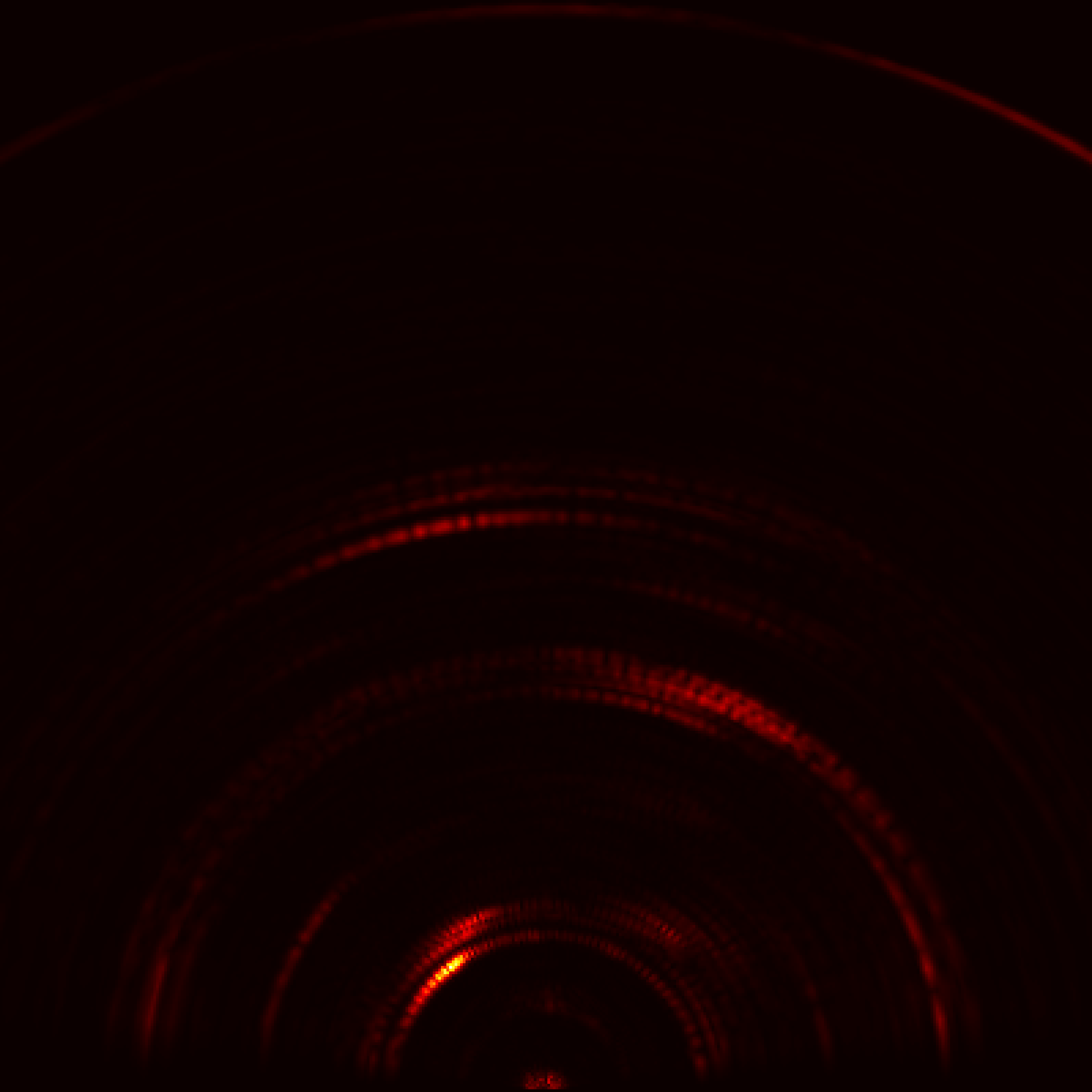} & \includegraphics[width=0.165\linewidth]{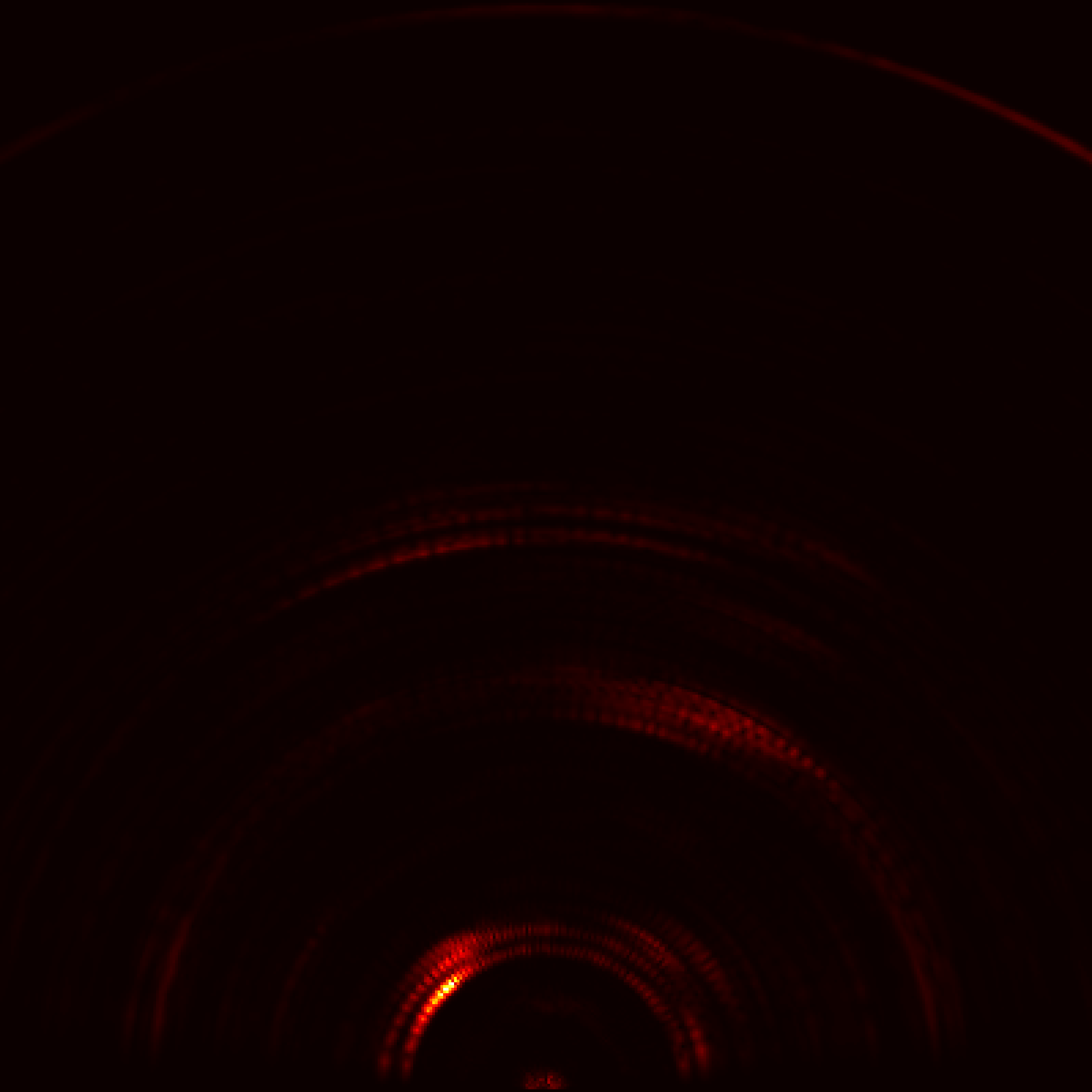} & \includegraphics[width=0.165\linewidth]{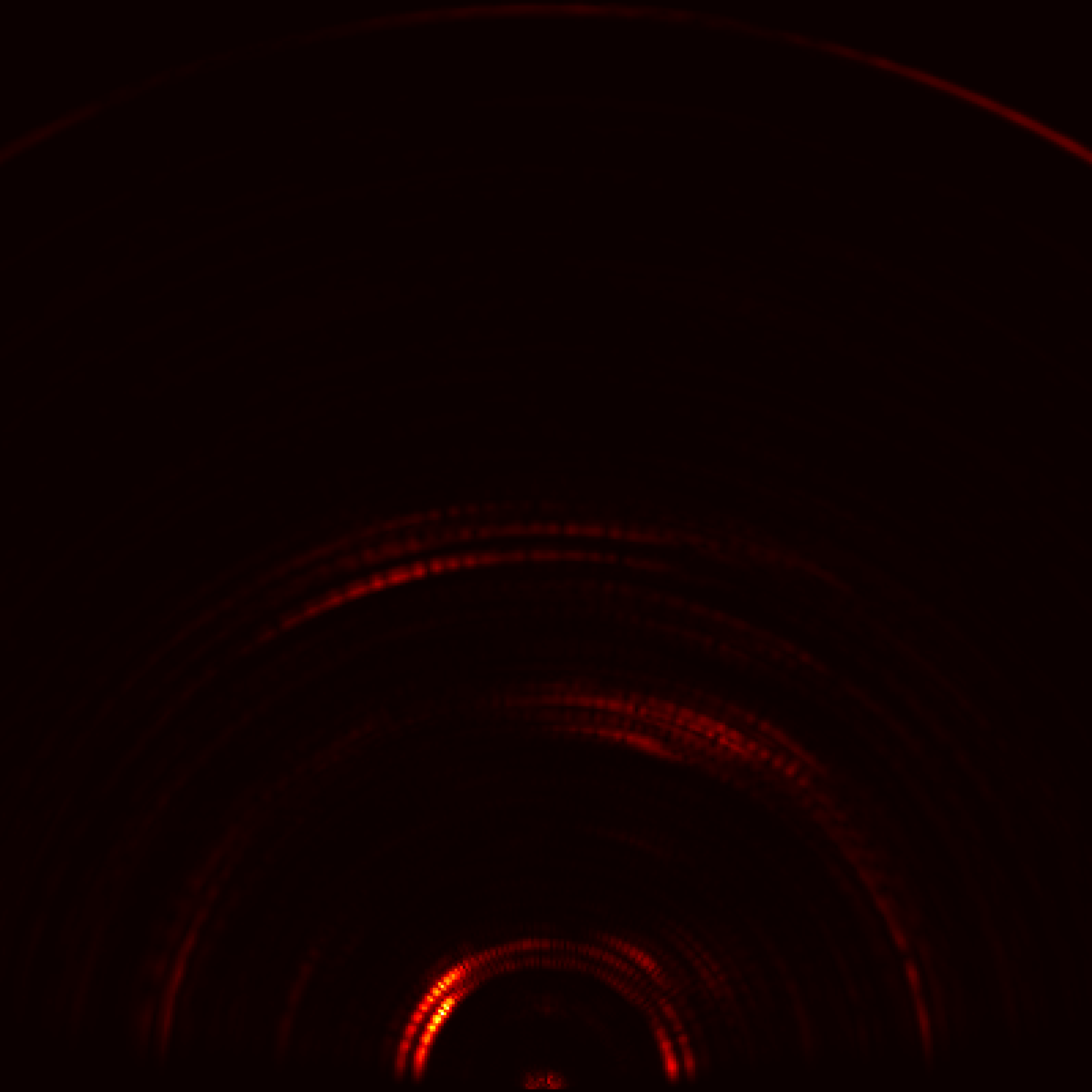} & \includegraphics[width=0.165\linewidth]{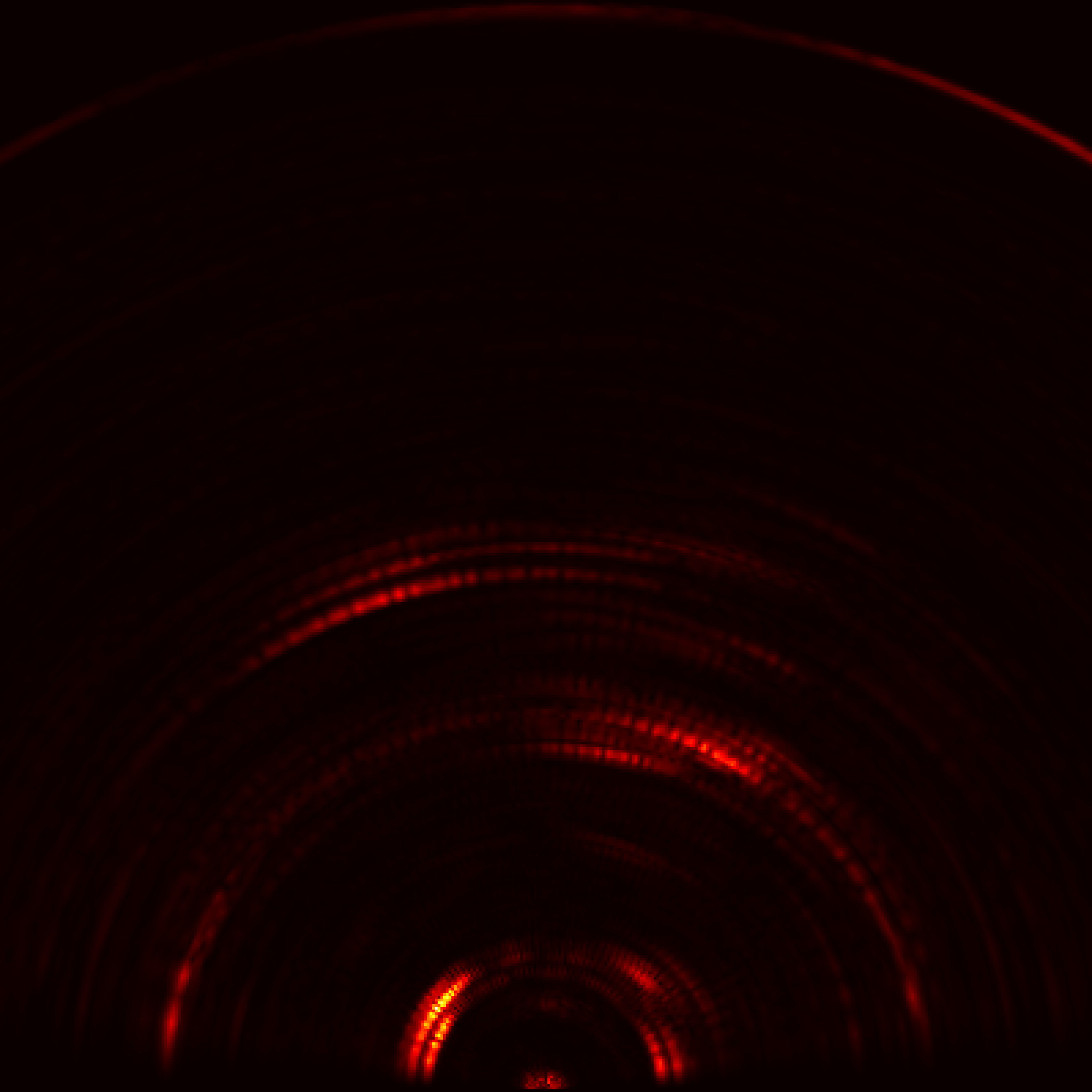} \\
\raisebox{2.0em}{\rotatebox{90}{\scriptsize\textbf{3DPS (Ours)}}} & \includegraphics[width=0.165\linewidth]{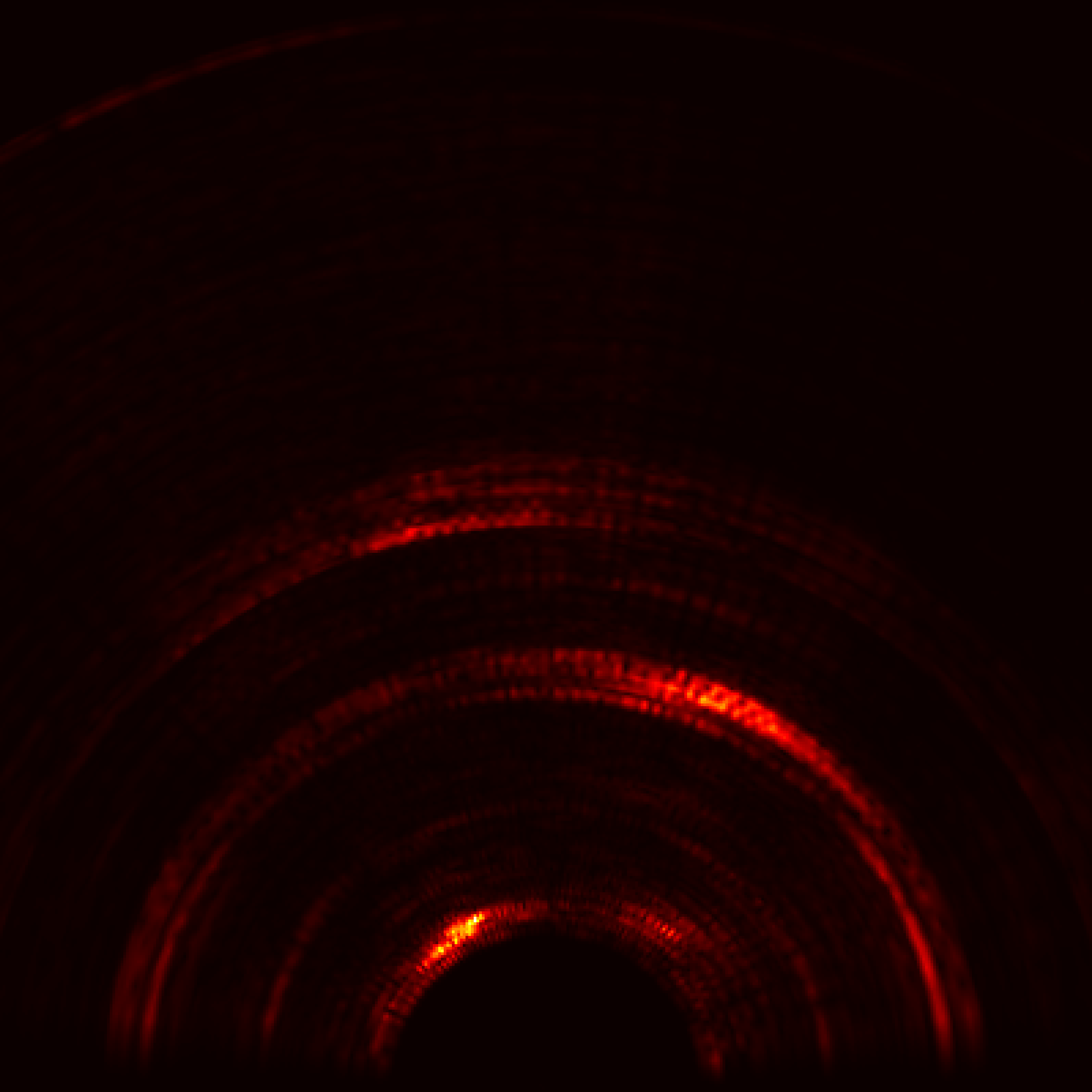} & \includegraphics[width=0.165\linewidth]{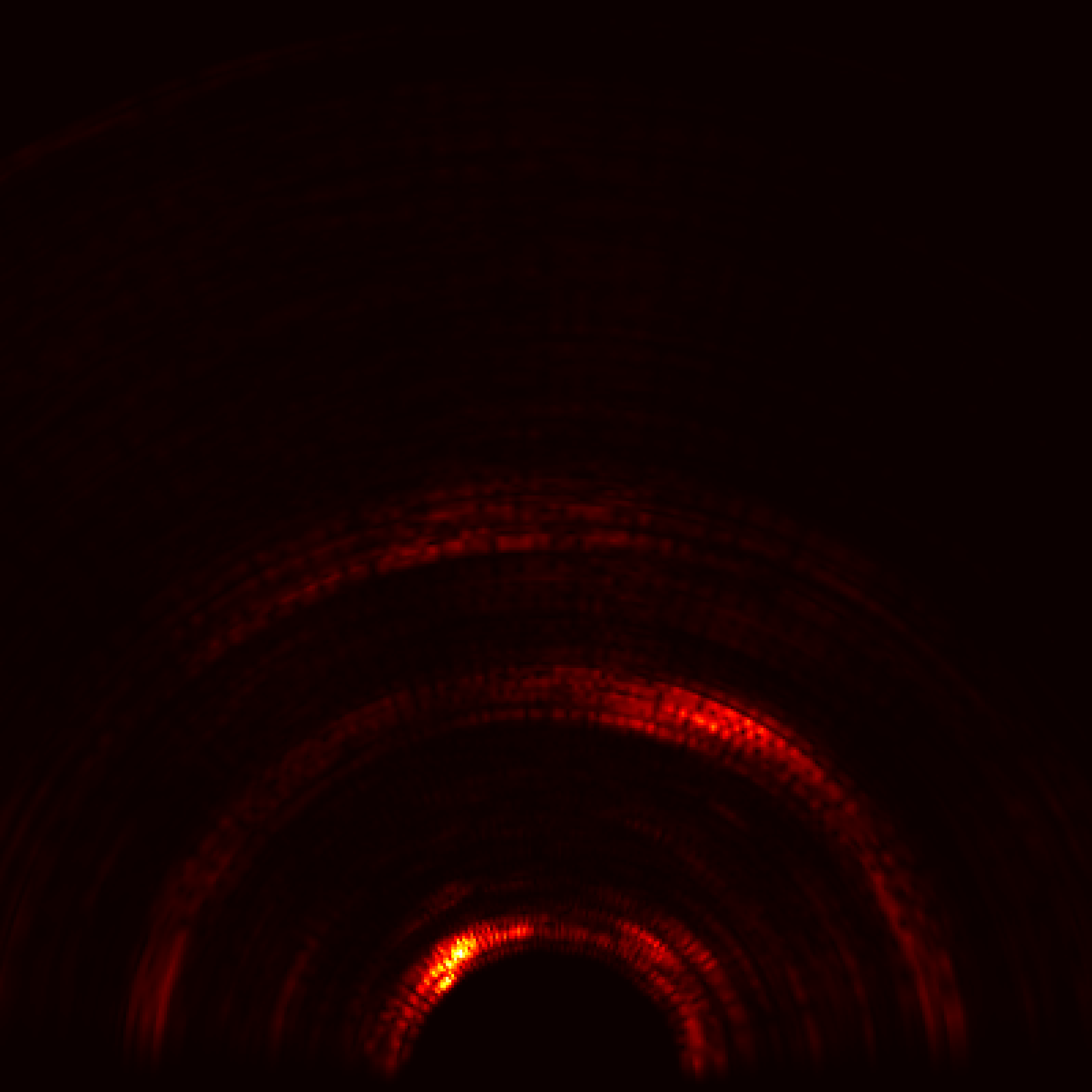} & \includegraphics[width=0.165\linewidth]{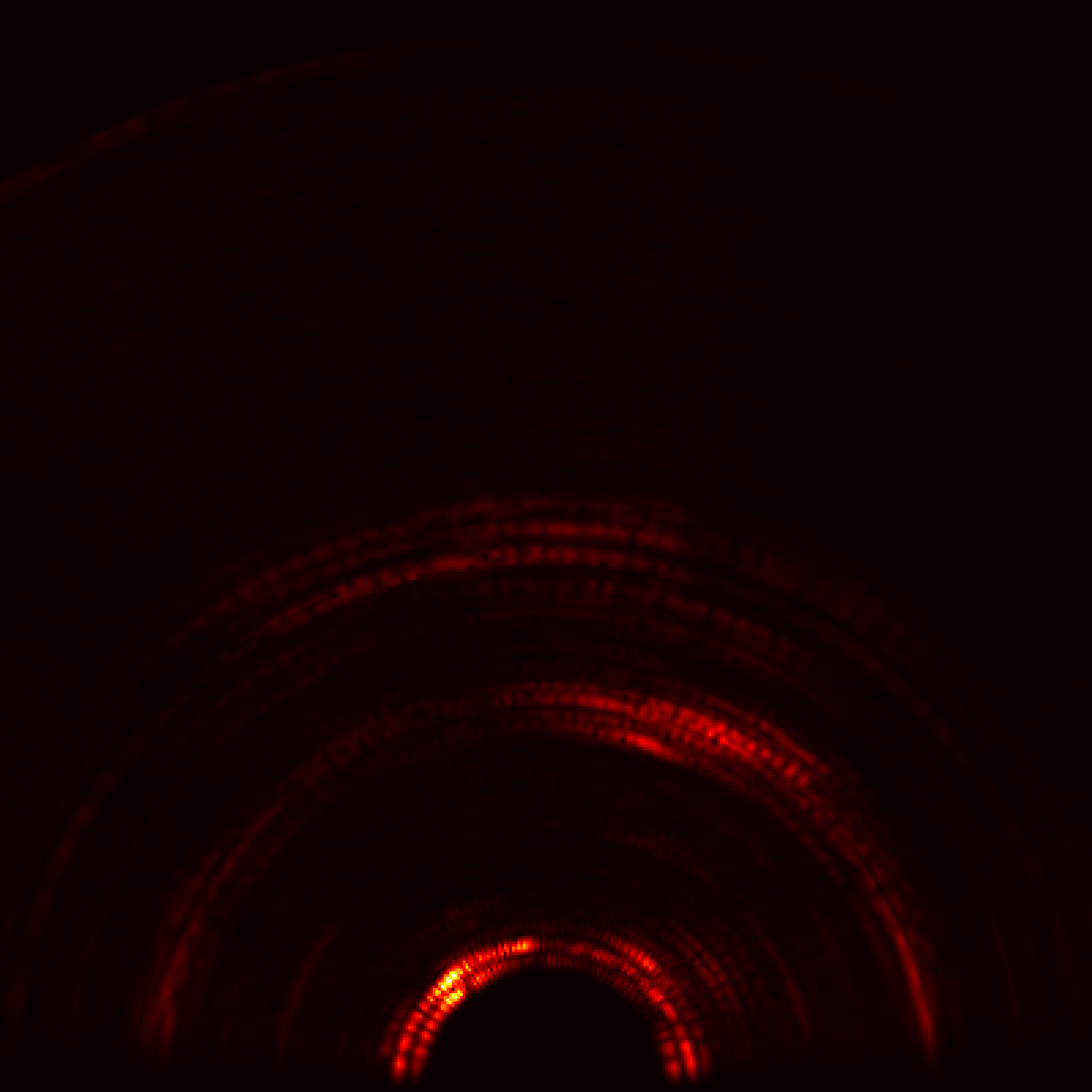} & \includegraphics[width=0.165\linewidth]{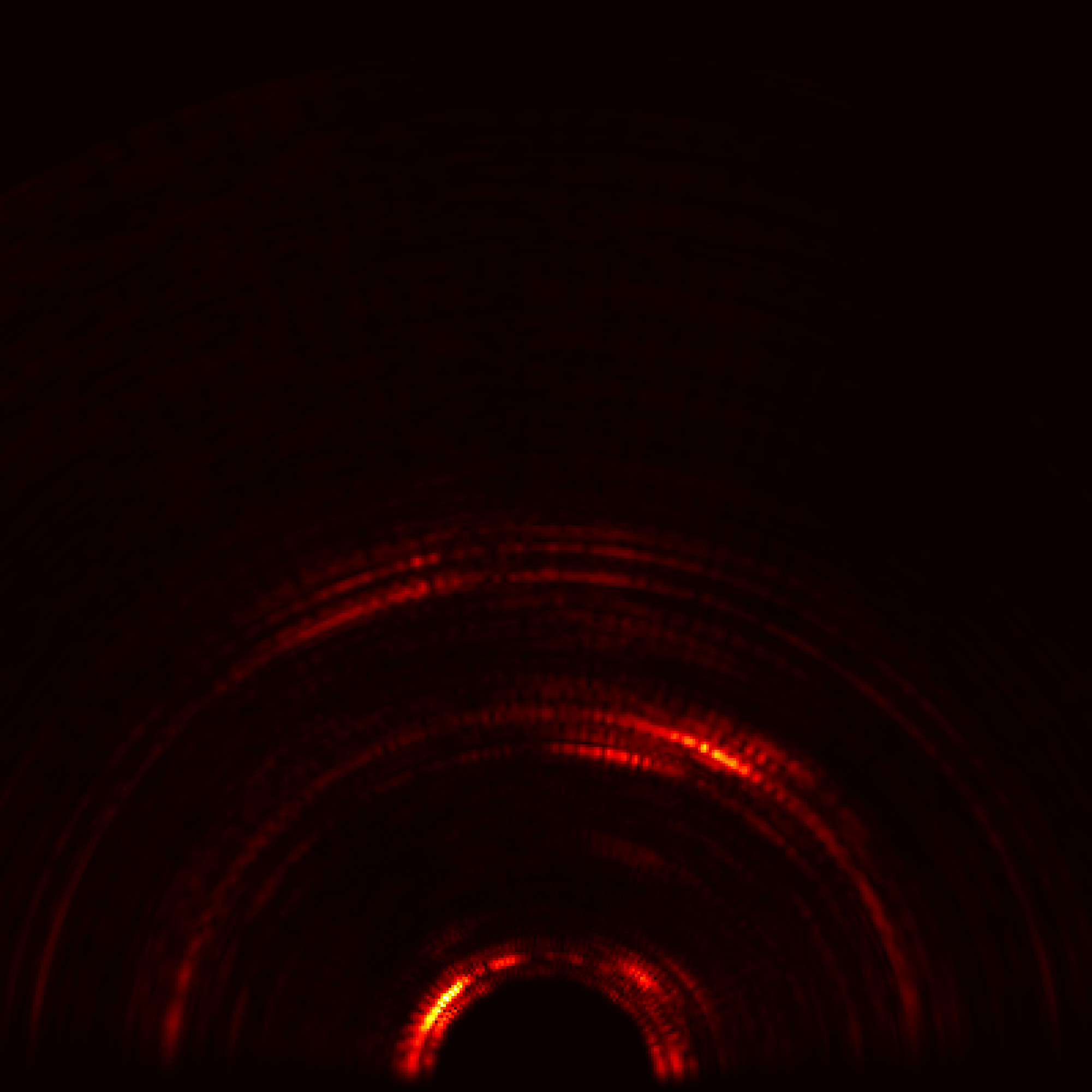} \\
\raisebox{2.0em}{\rotatebox{90}{\scriptsize\textbf{mmIR}}} & \includegraphics[width=0.165\linewidth]{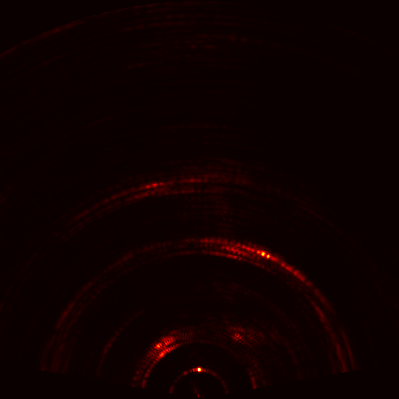} & \includegraphics[width=0.165\linewidth]{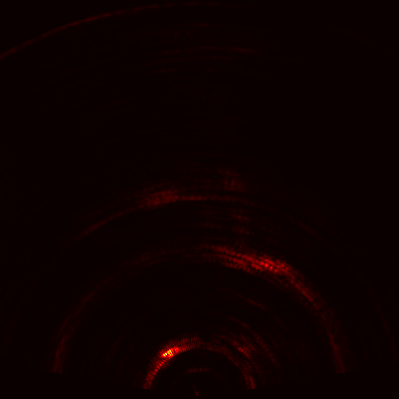} & \includegraphics[width=0.165\linewidth]{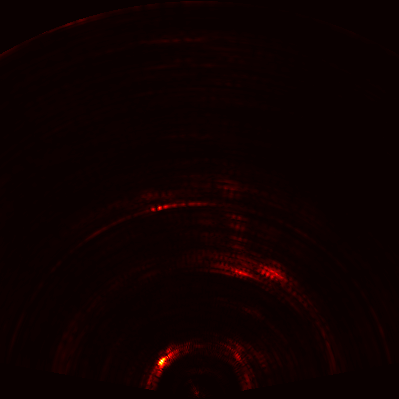} & \includegraphics[width=0.165\linewidth]{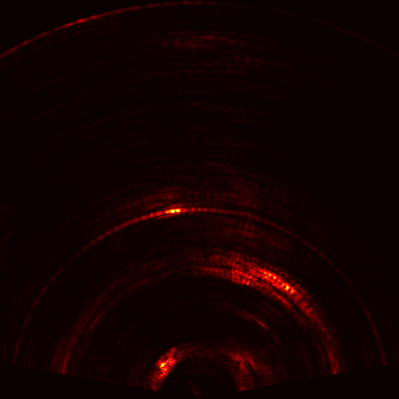} \\
\raisebox{2.0em}{\rotatebox{90}{\scriptsize\textbf{RadarSplat}}} & \includegraphics[width=0.165\linewidth]{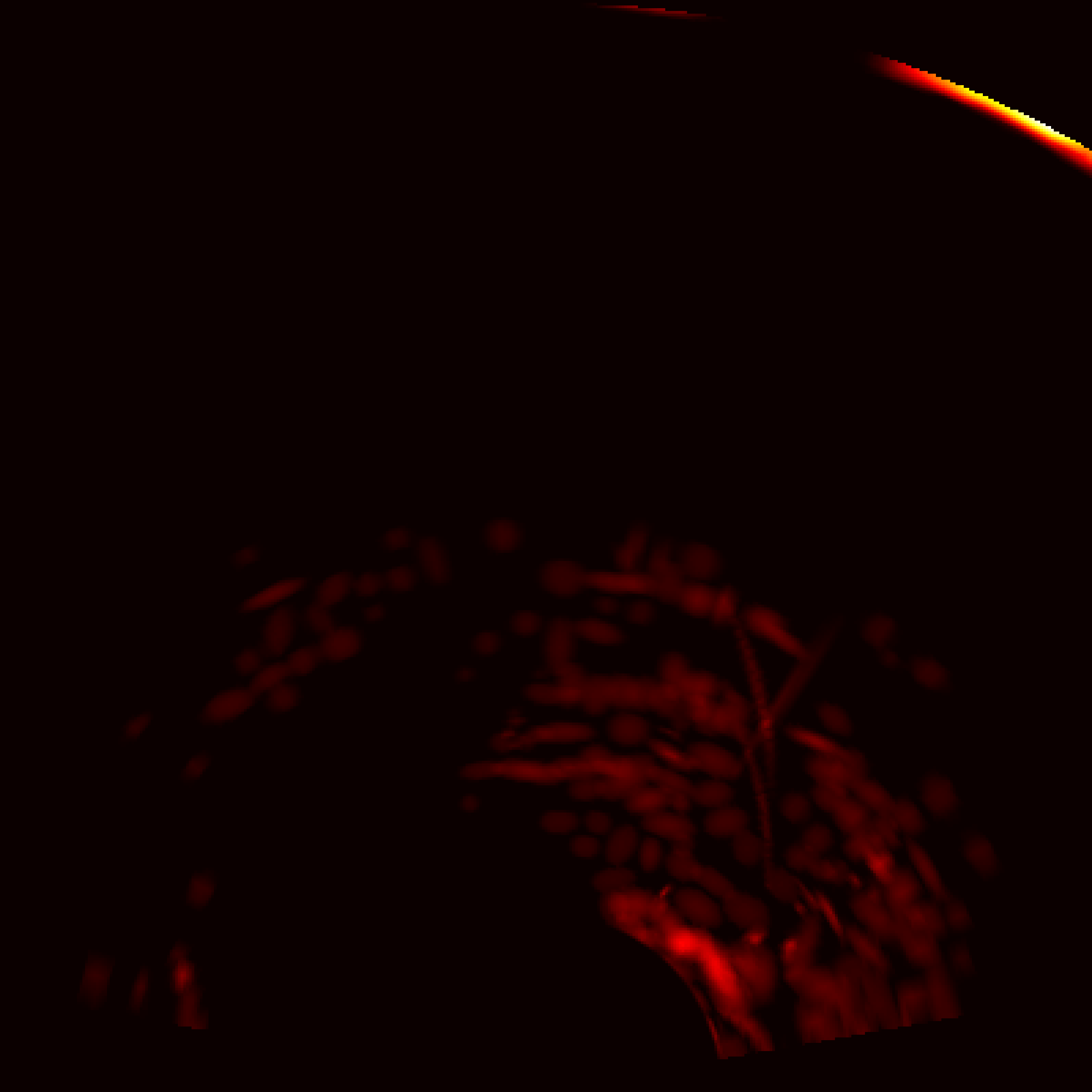} & \includegraphics[width=0.165\linewidth]{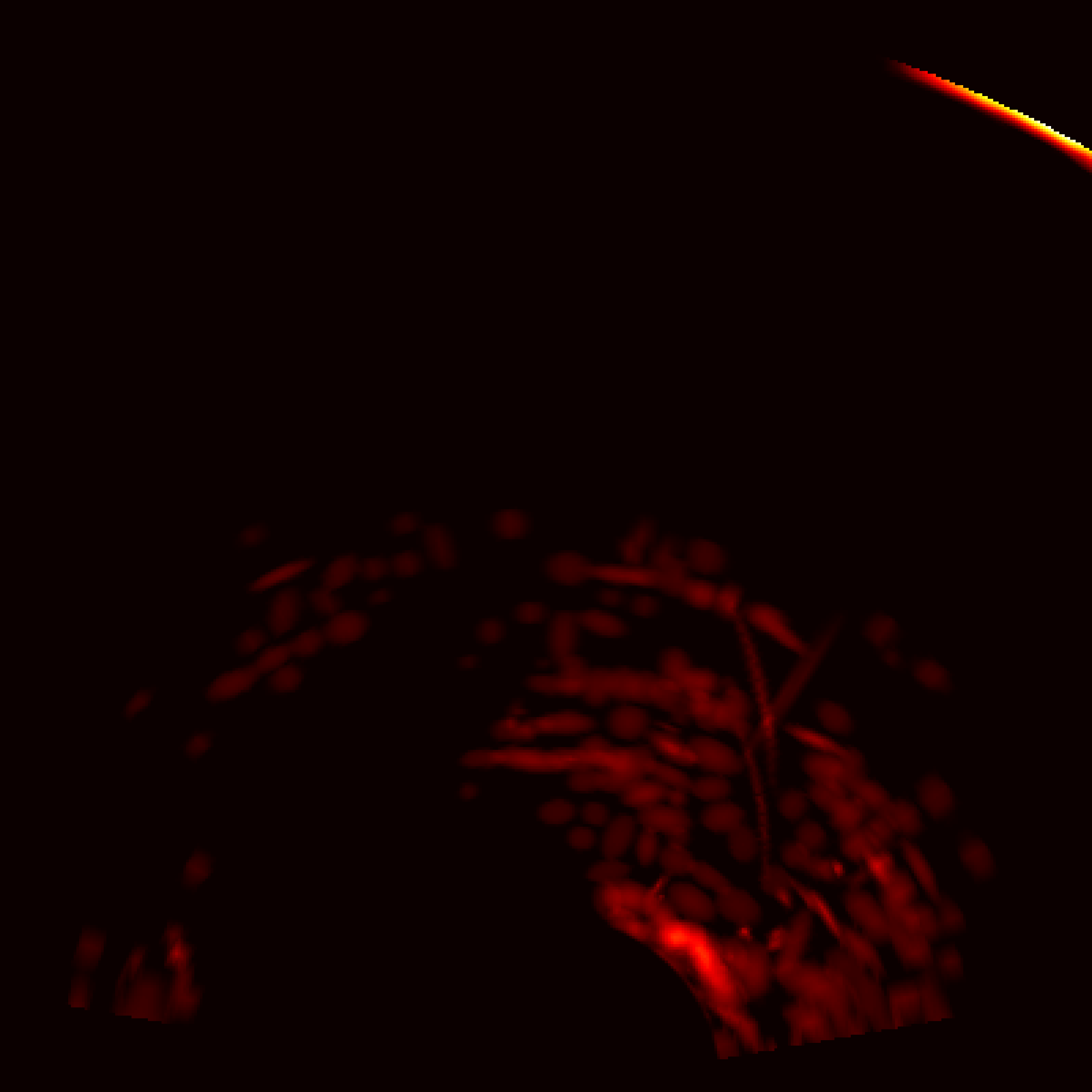} & \includegraphics[width=0.165\linewidth]{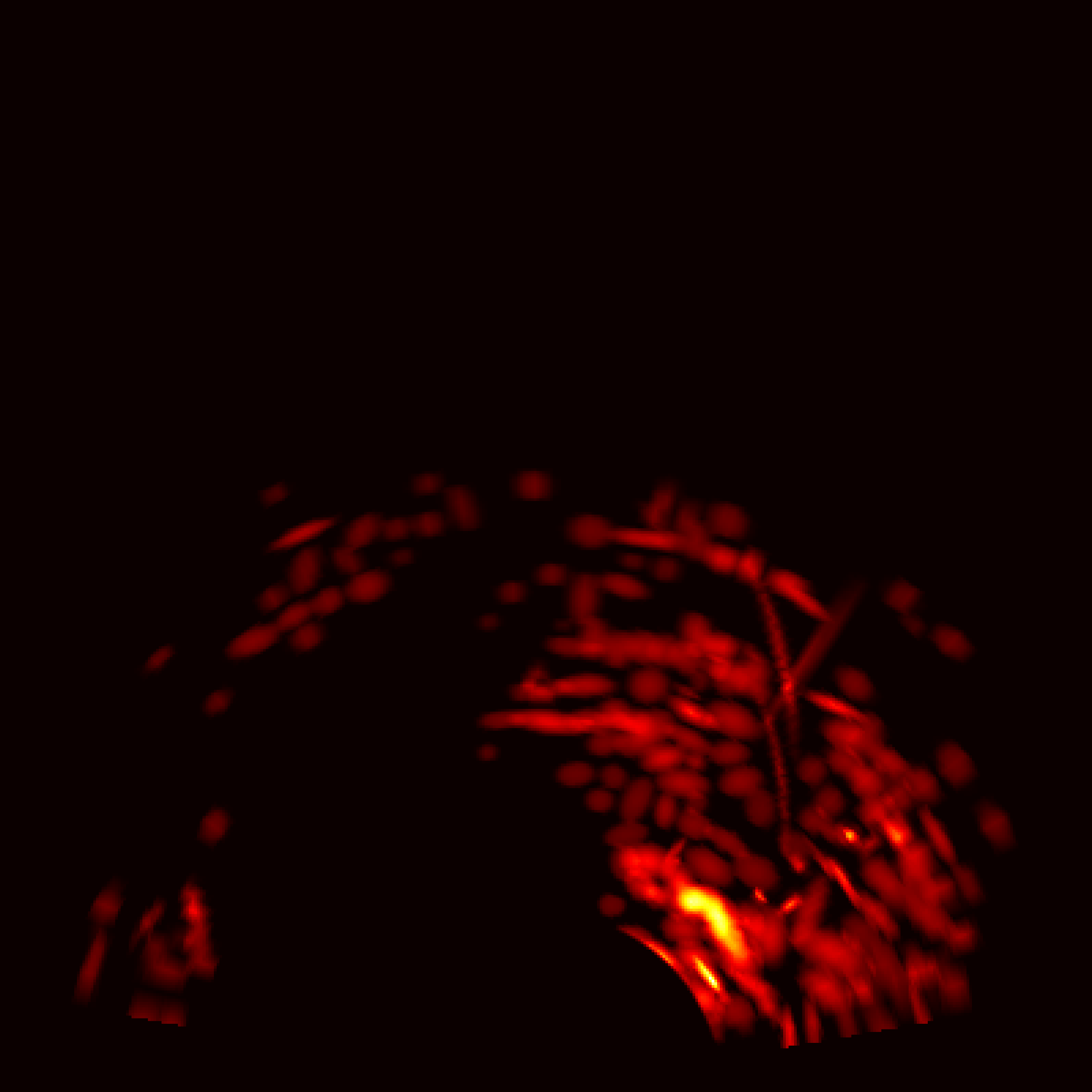} & \includegraphics[width=0.165\linewidth]{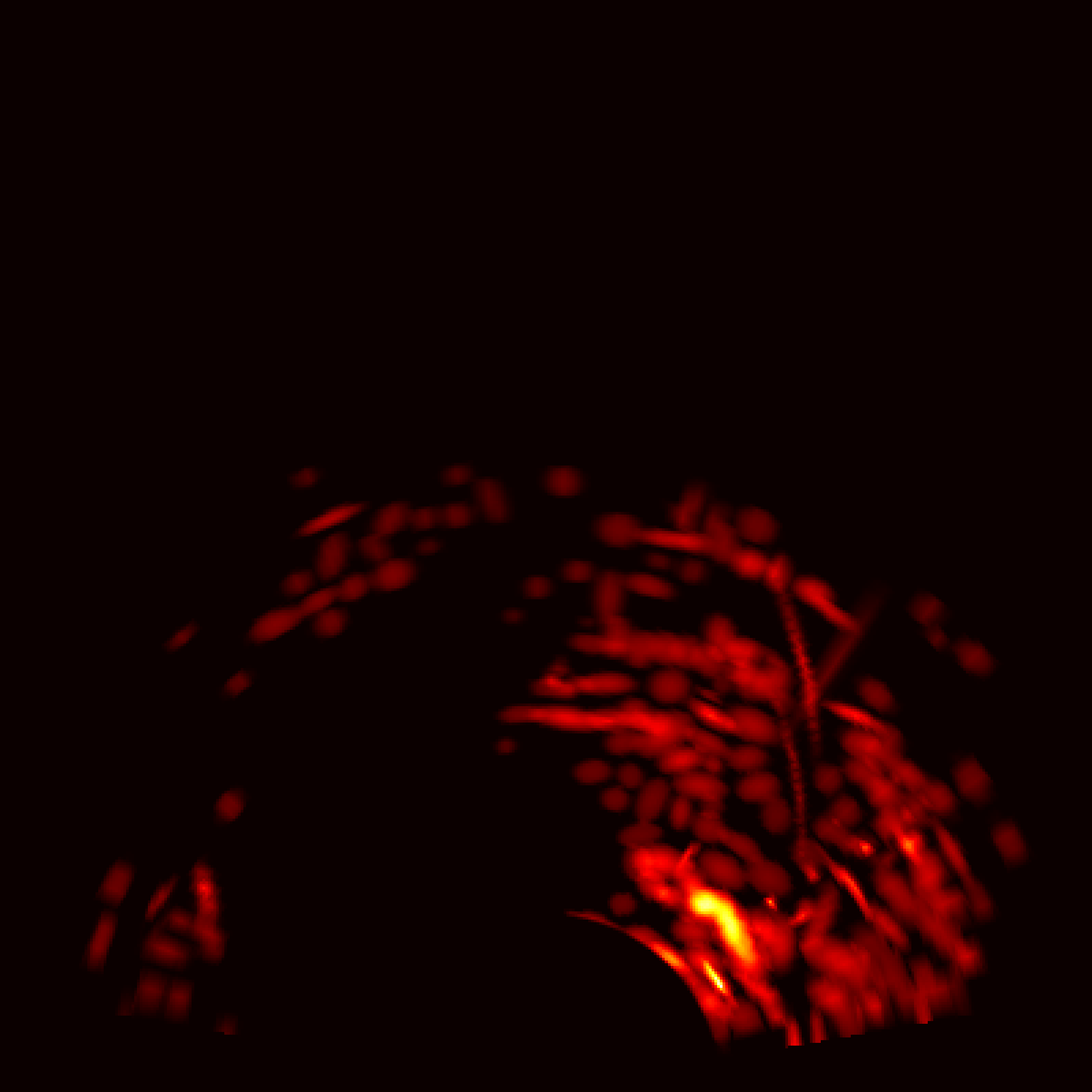} \\
\raisebox{2.0em}{\rotatebox{90}{\scriptsize\textbf{Radar Fields}}} & \includegraphics[width=0.165\linewidth]{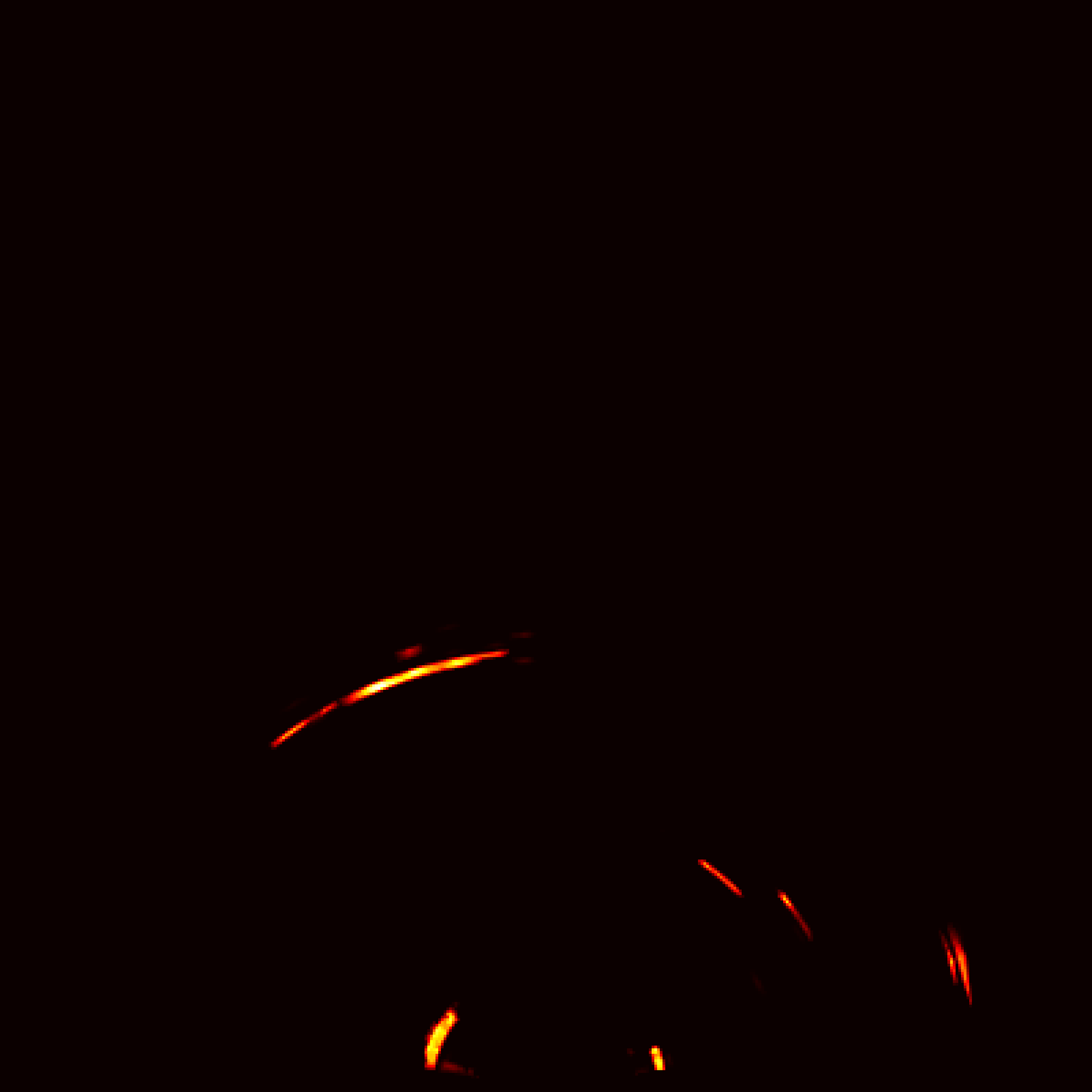} & \includegraphics[width=0.165\linewidth]{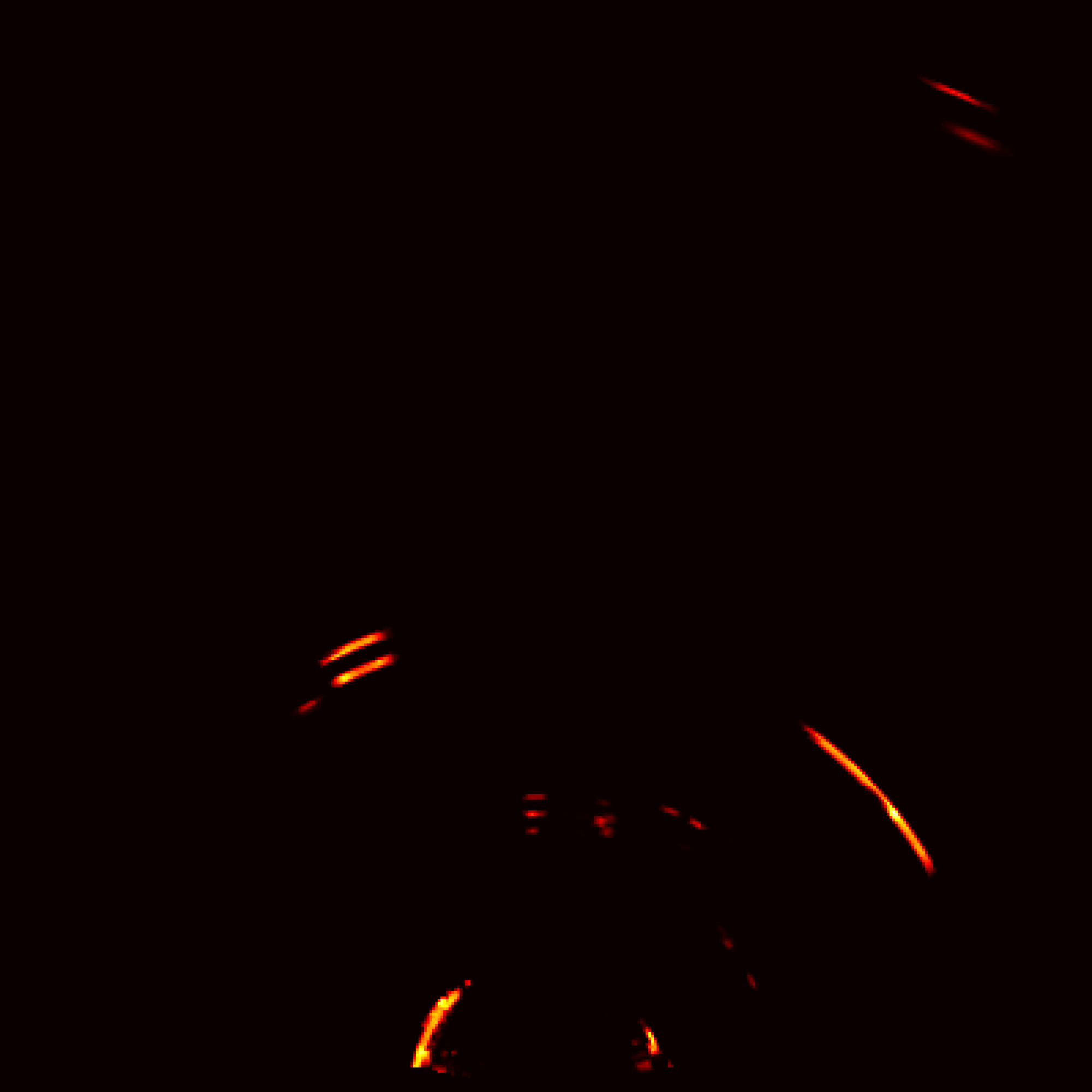} & \includegraphics[width=0.165\linewidth]{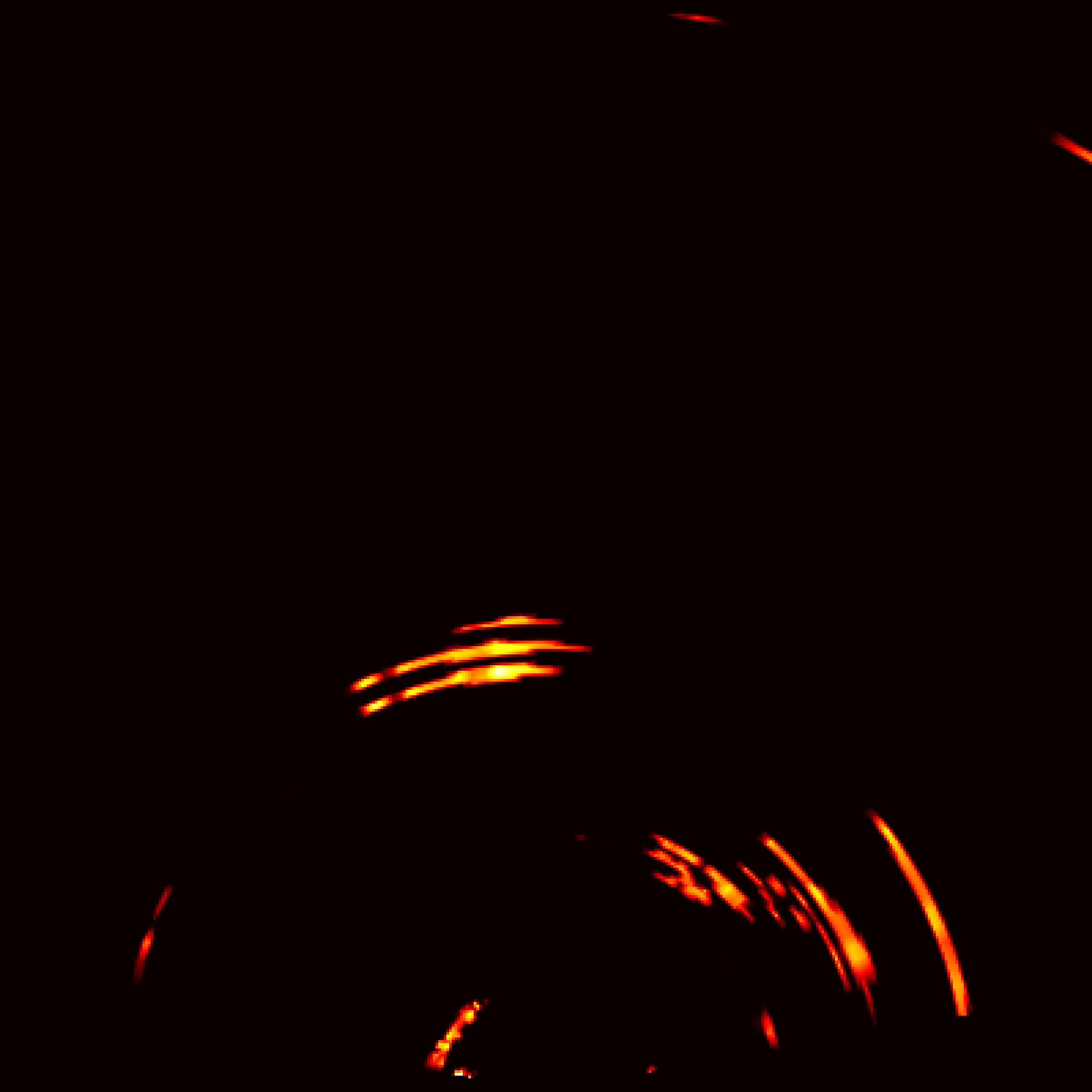} & \includegraphics[width=0.165\linewidth]{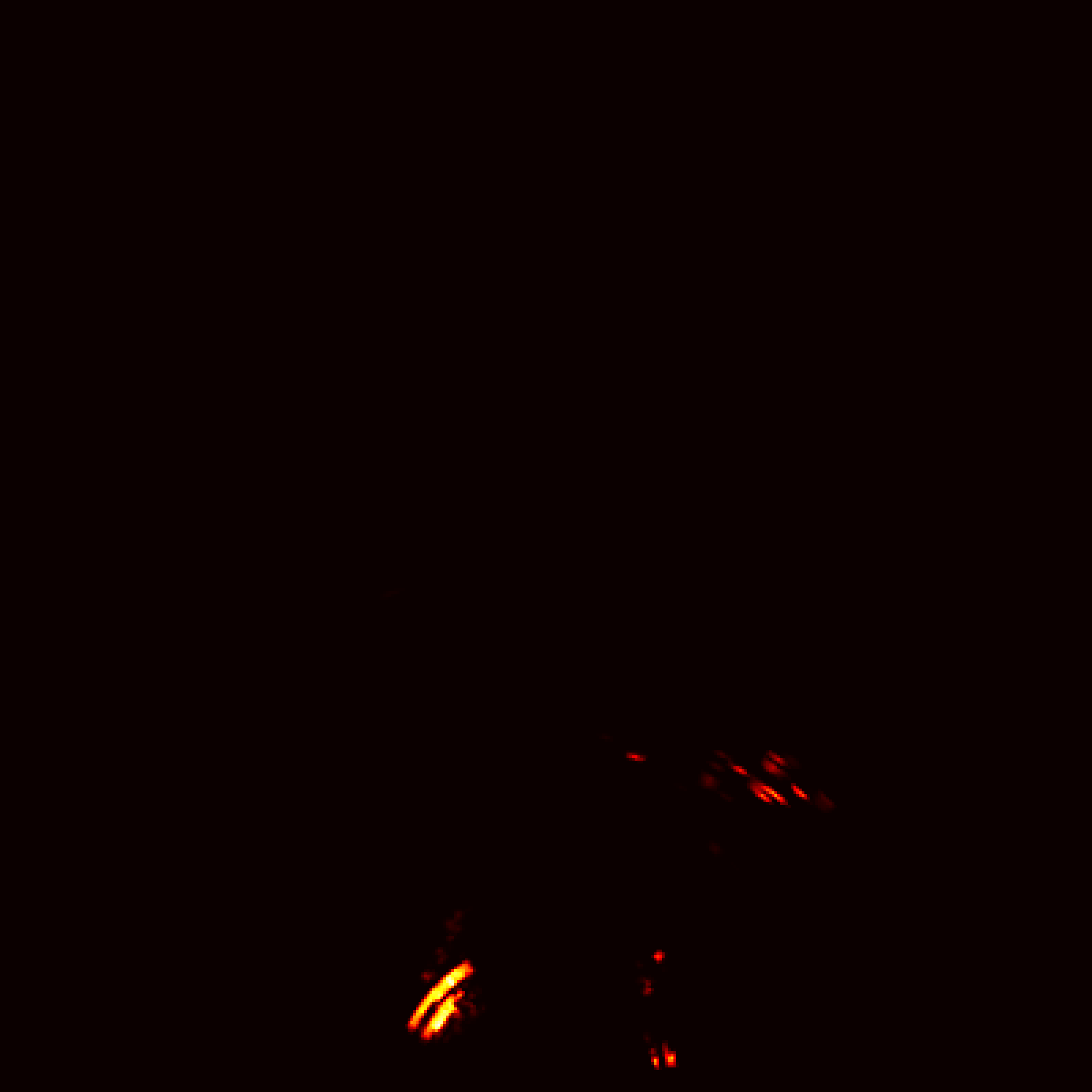} \\
\raisebox{2.0em}{\rotatebox{90}{\scriptsize\textbf{DART}}} & \includegraphics[width=0.165\linewidth]{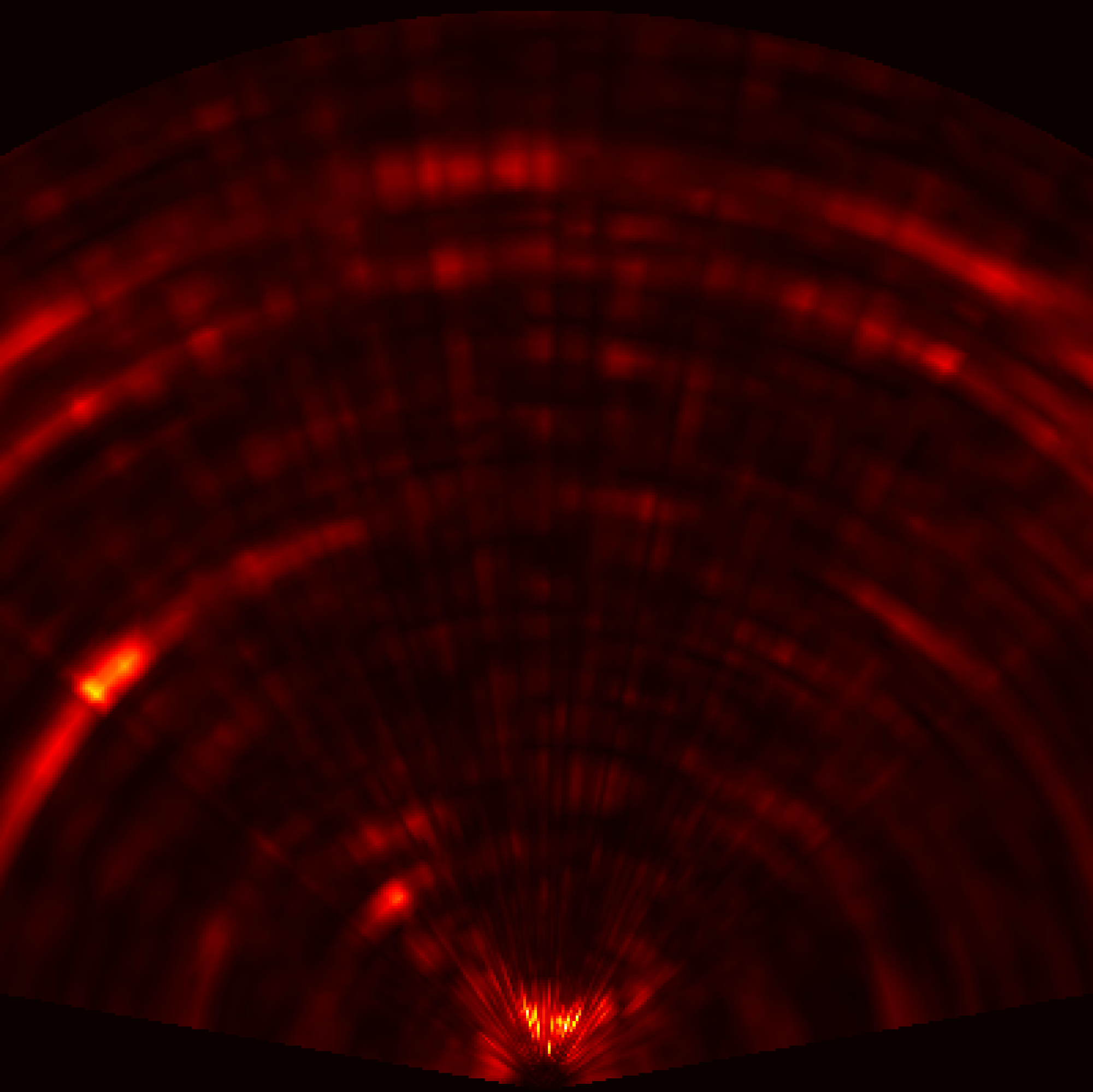} & \includegraphics[width=0.165\linewidth]{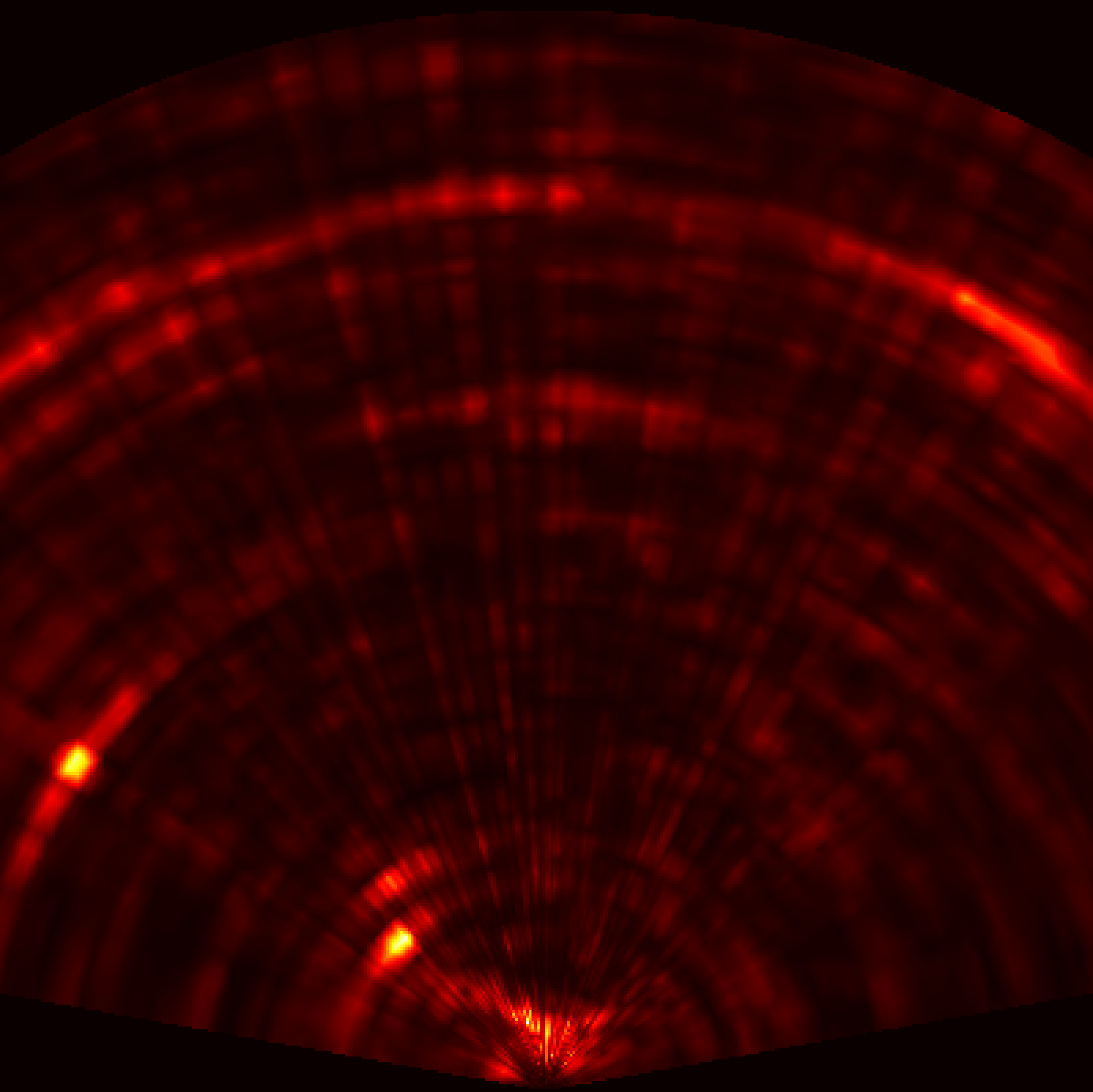} & \includegraphics[width=0.165\linewidth]{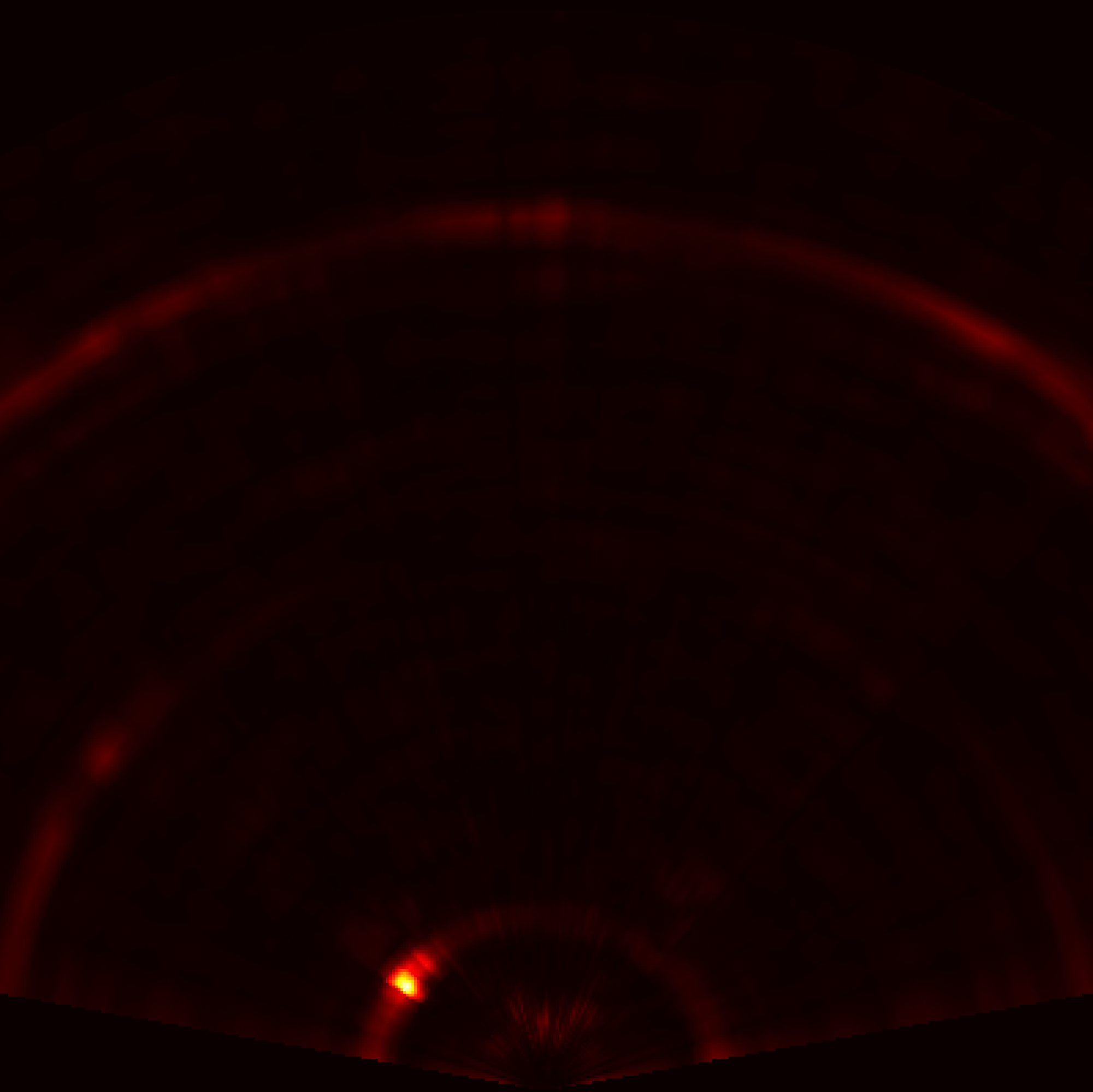} & \includegraphics[width=0.165\linewidth]{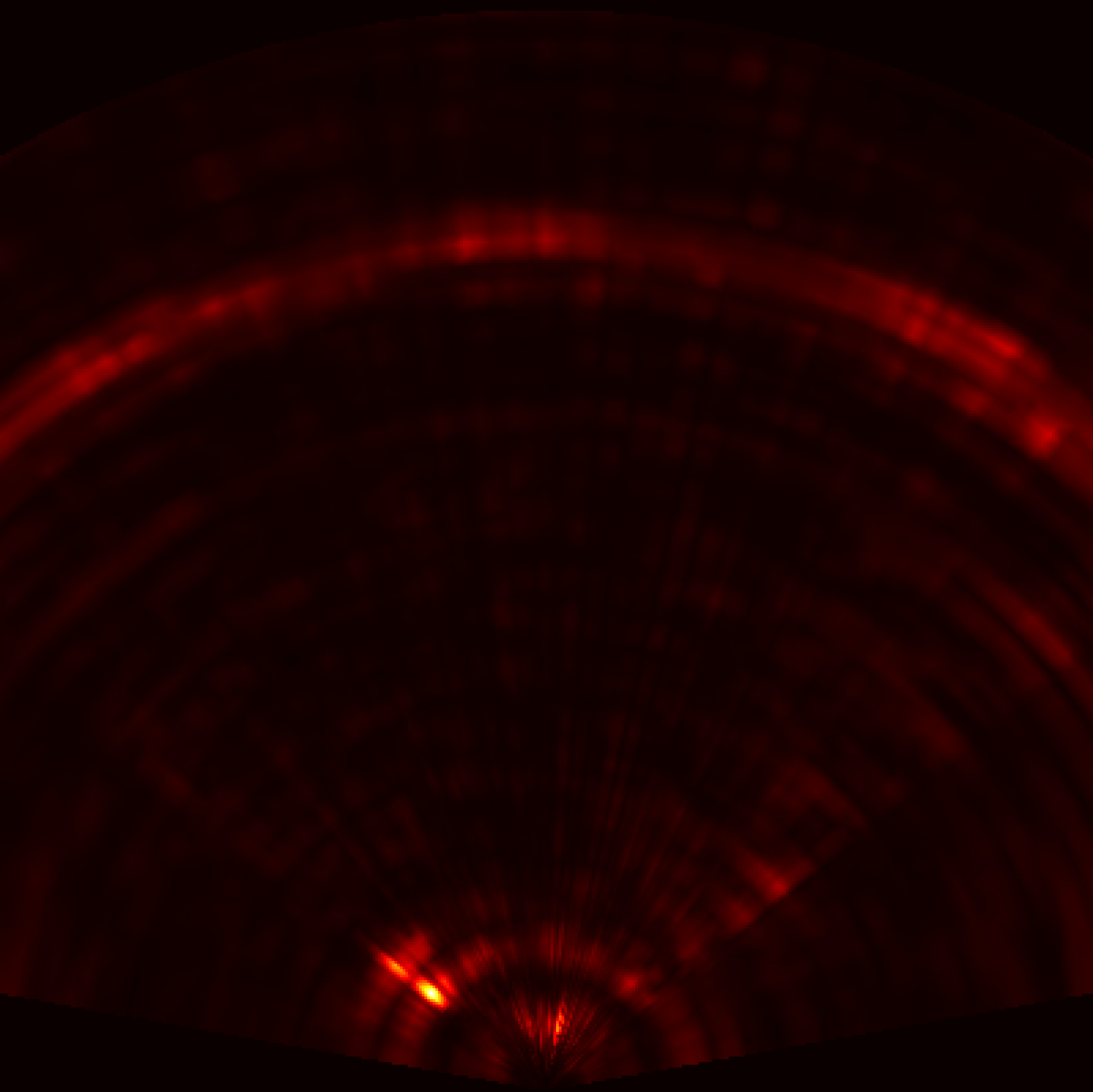} \\
\end{tabular}
\caption{\textbf{Per-frame $|\mathrm{RA}|$ comparison, scene S2\,F300, training frames $F\!-\!4$ to $F\!-\!1$.} Pearson correlation per column, left to right. 3DPS: 0.70, 0.72, 0.79, 0.86; mmIR: 0.65, 0.66, 0.67, 0.62; RadarSplat: 0.26, 0.20, 0.13, 0.17; Radar Fields: 0.02, 0.06, 0.00, 0.39; DART: -0.02, 0.04, 0.02, 0.13.}
\label{fig:supp_s2_f300_a}
\end{figure}

\begin{figure}[p]
\centering
\setlength{\tabcolsep}{1pt}\renewcommand{\arraystretch}{0.6}
\begin{tabular}{cccccc}
 & {\footnotesize $F\!+\!1$} & {\footnotesize $F\!+\!2$} & {\footnotesize $F\!+\!3$} & {\footnotesize $F\!+\!4$} & \textcolor{red}{\footnotesize\textbf{$F$}} \\
\raisebox{2.0em}{\rotatebox{90}{\scriptsize\textbf{GT}}} & \includegraphics[width=0.165\linewidth]{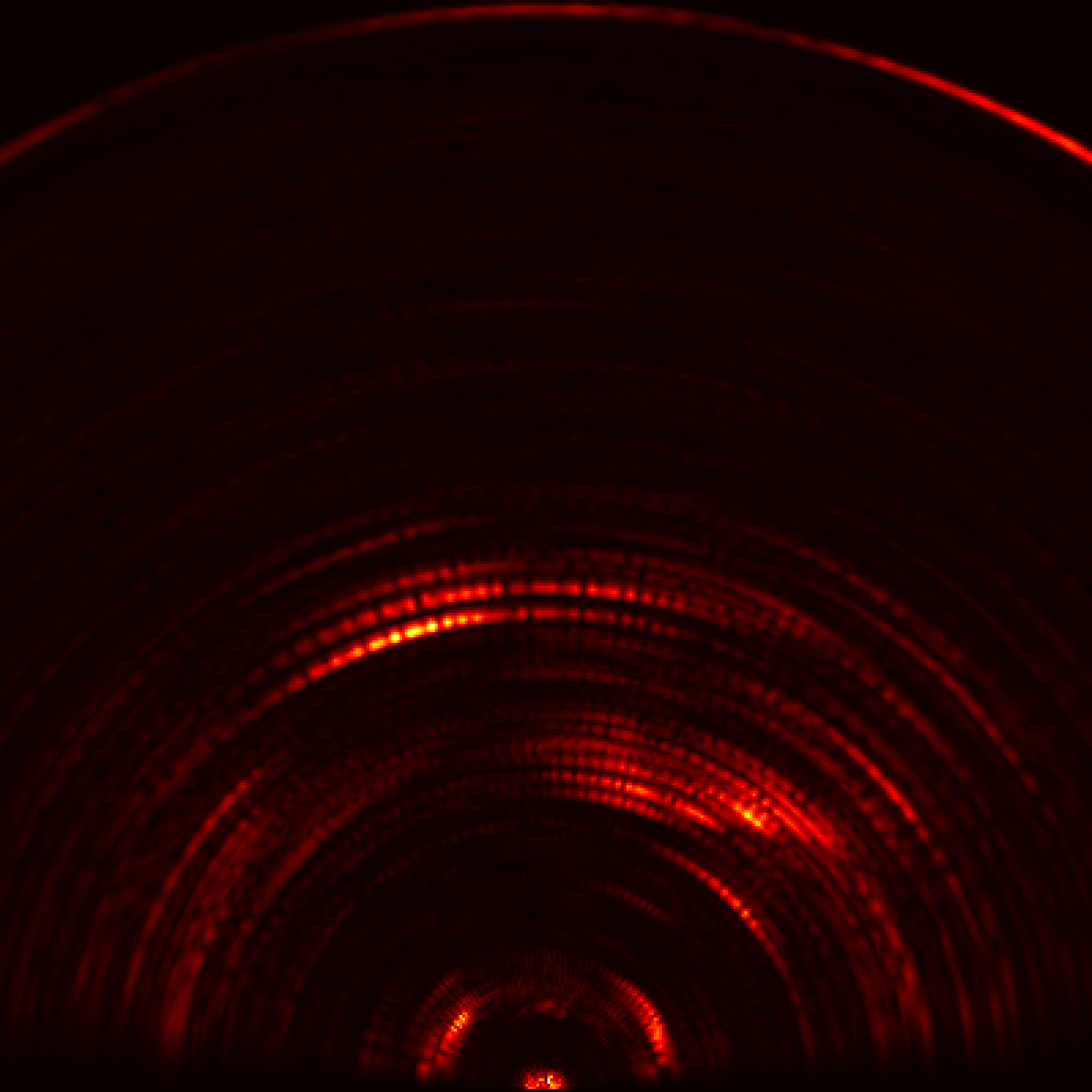} & \includegraphics[width=0.165\linewidth]{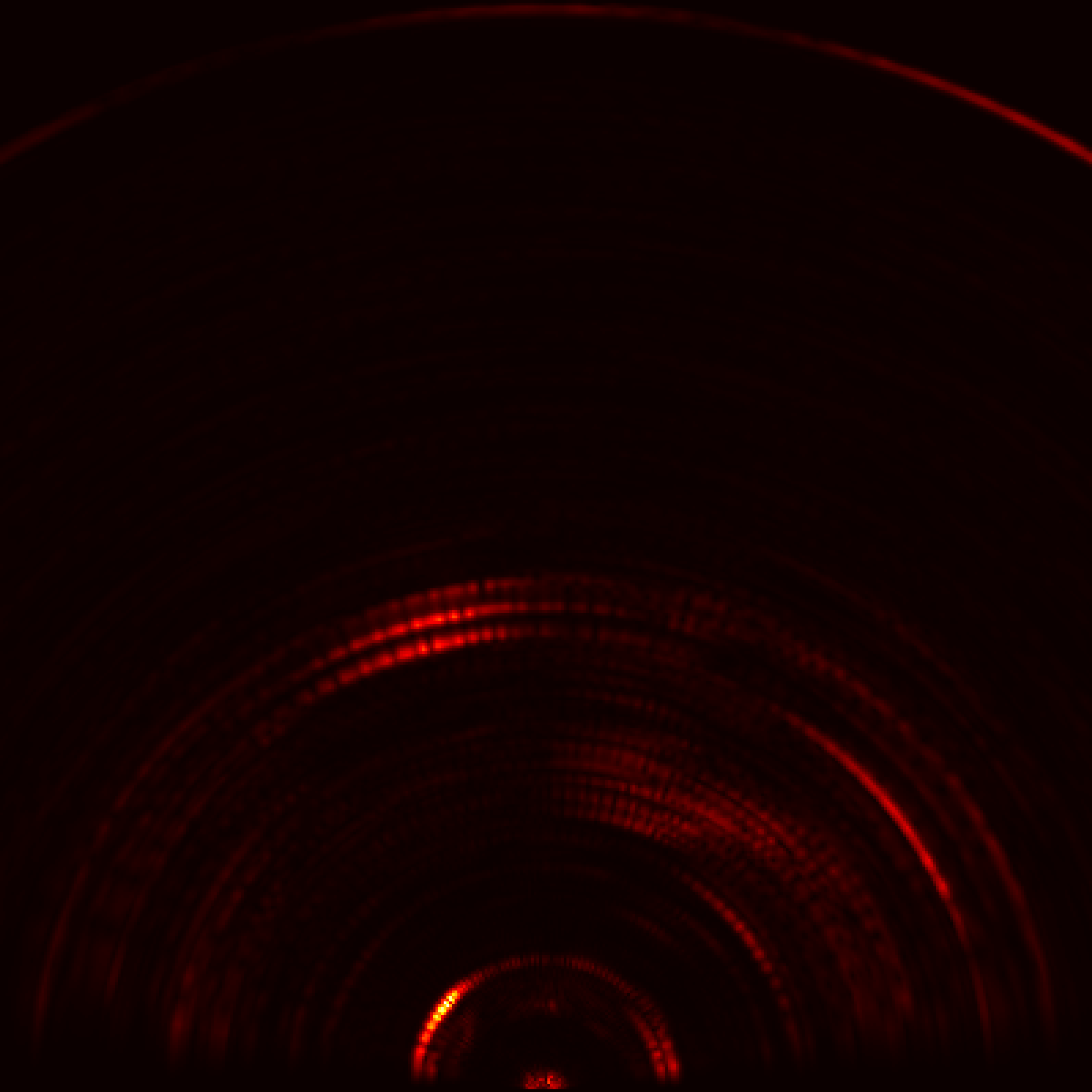} & \includegraphics[width=0.165\linewidth]{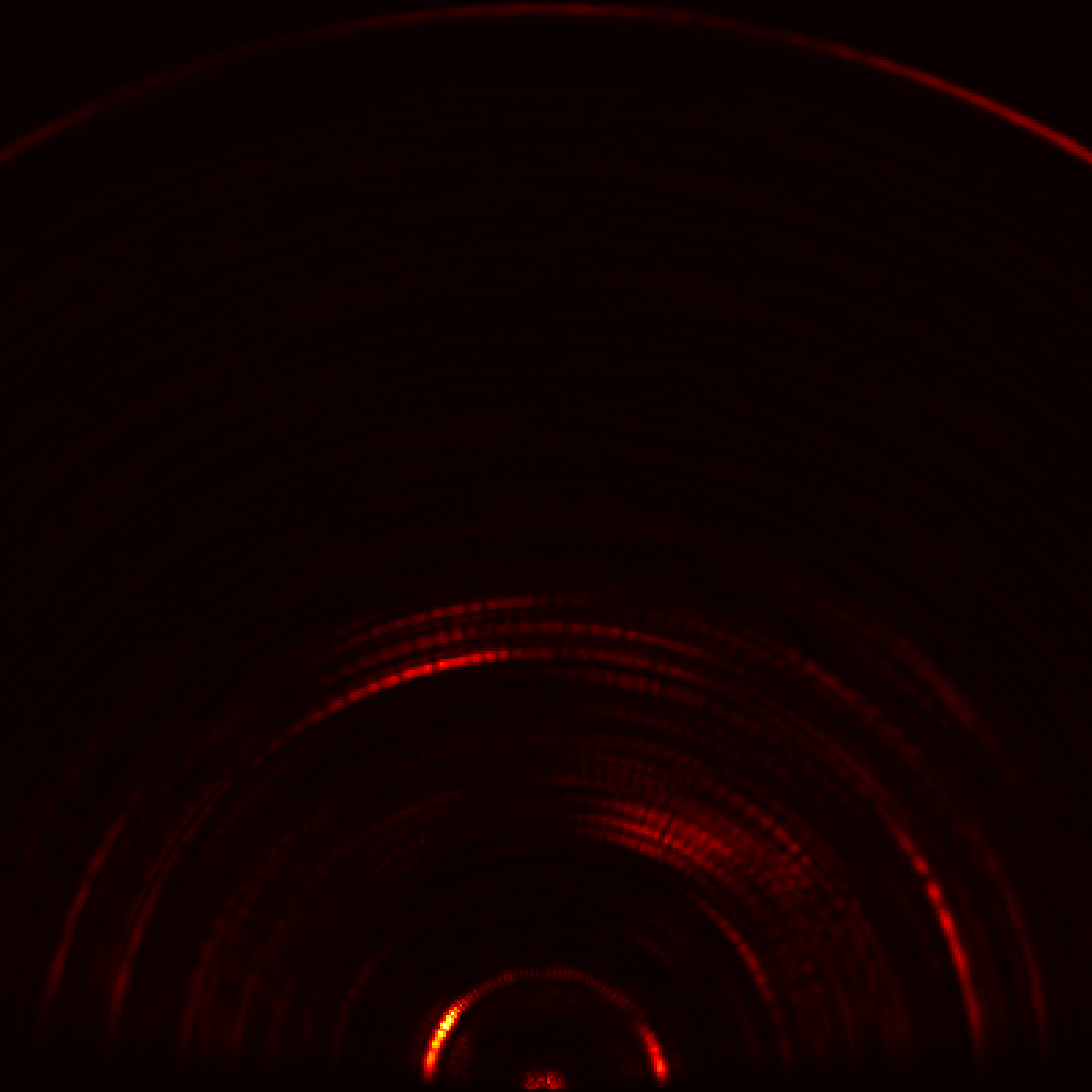} & \includegraphics[width=0.165\linewidth]{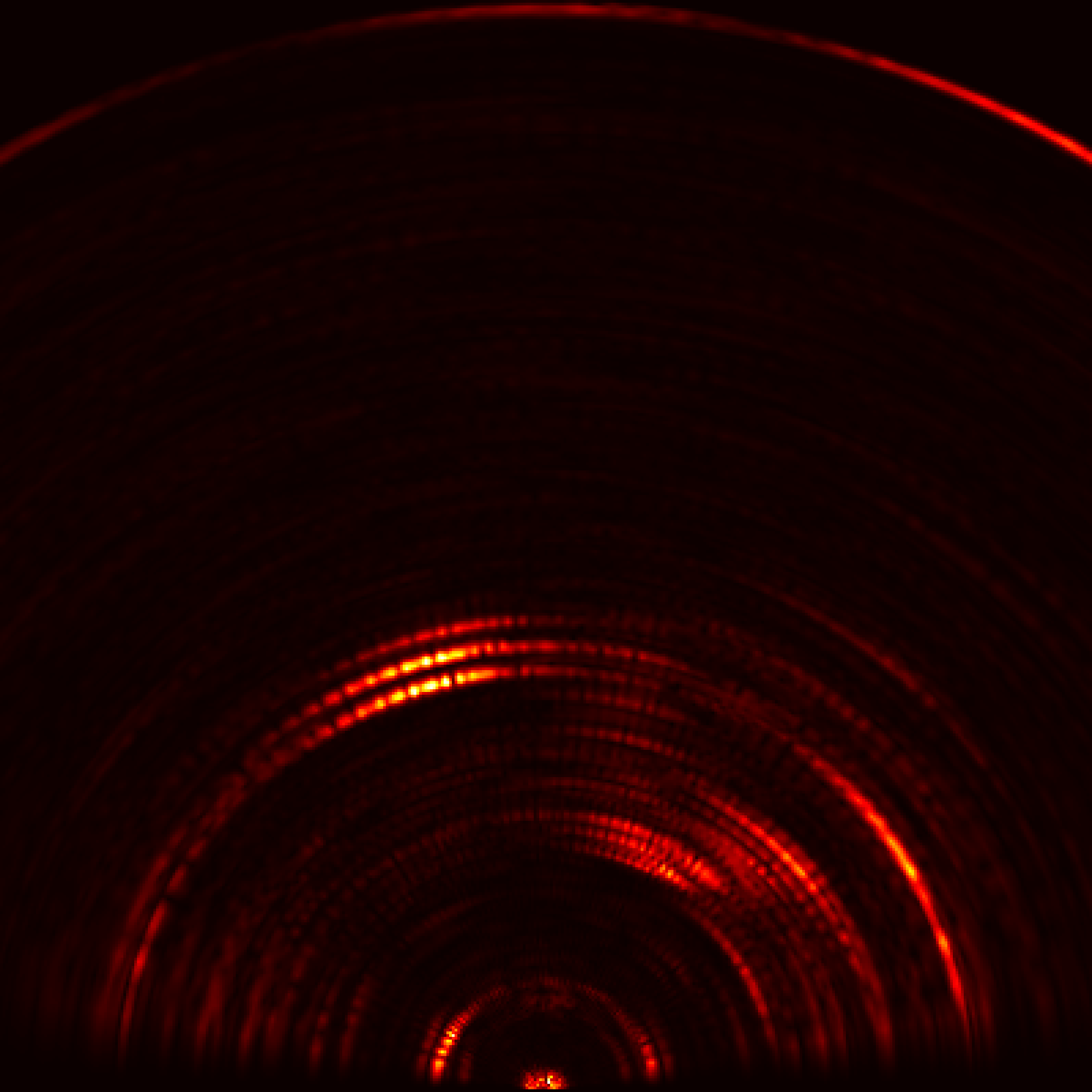} & \includegraphics[width=0.165\linewidth]{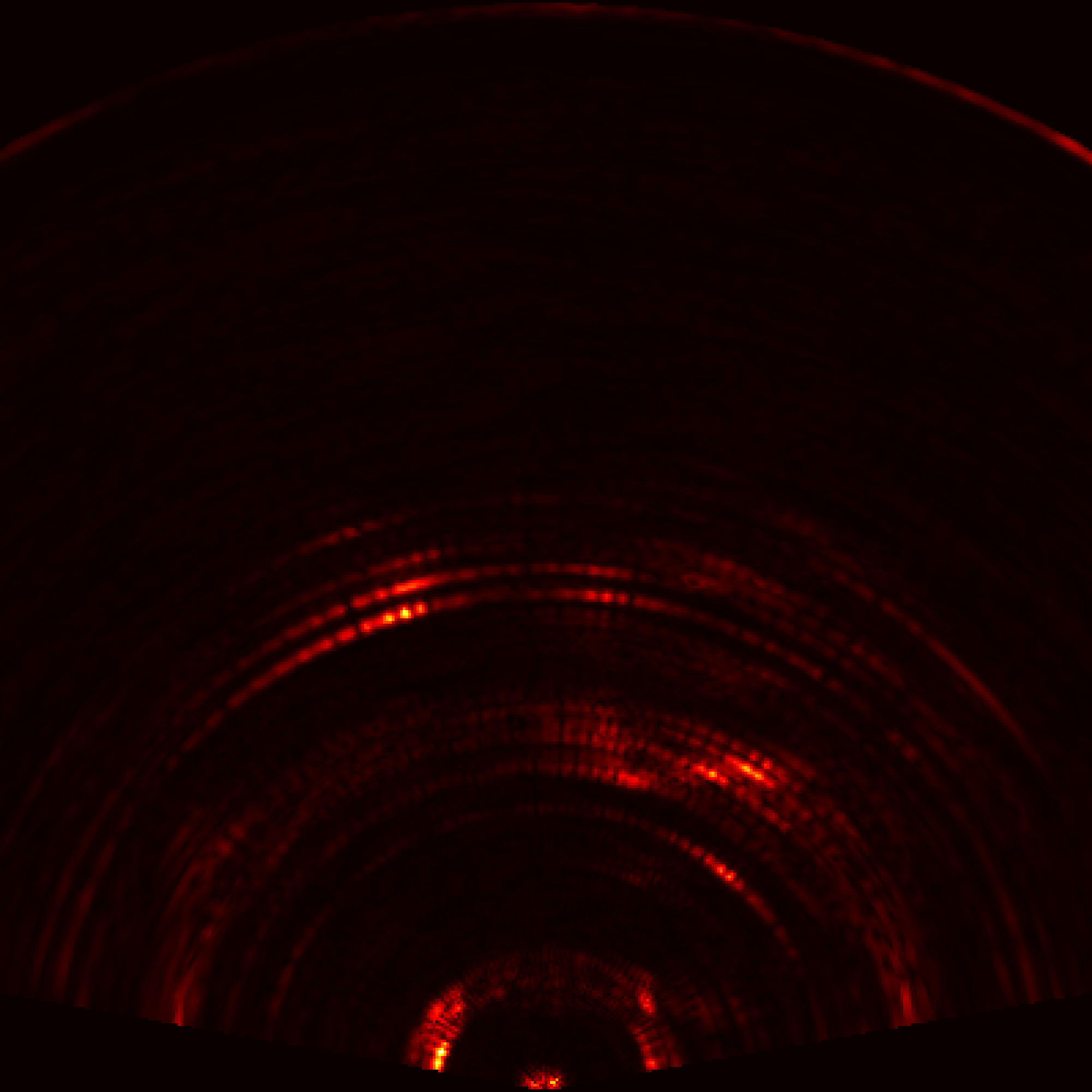} \\
\raisebox{2.0em}{\rotatebox{90}{\scriptsize\textbf{3DPS (Ours)}}} & \includegraphics[width=0.165\linewidth]{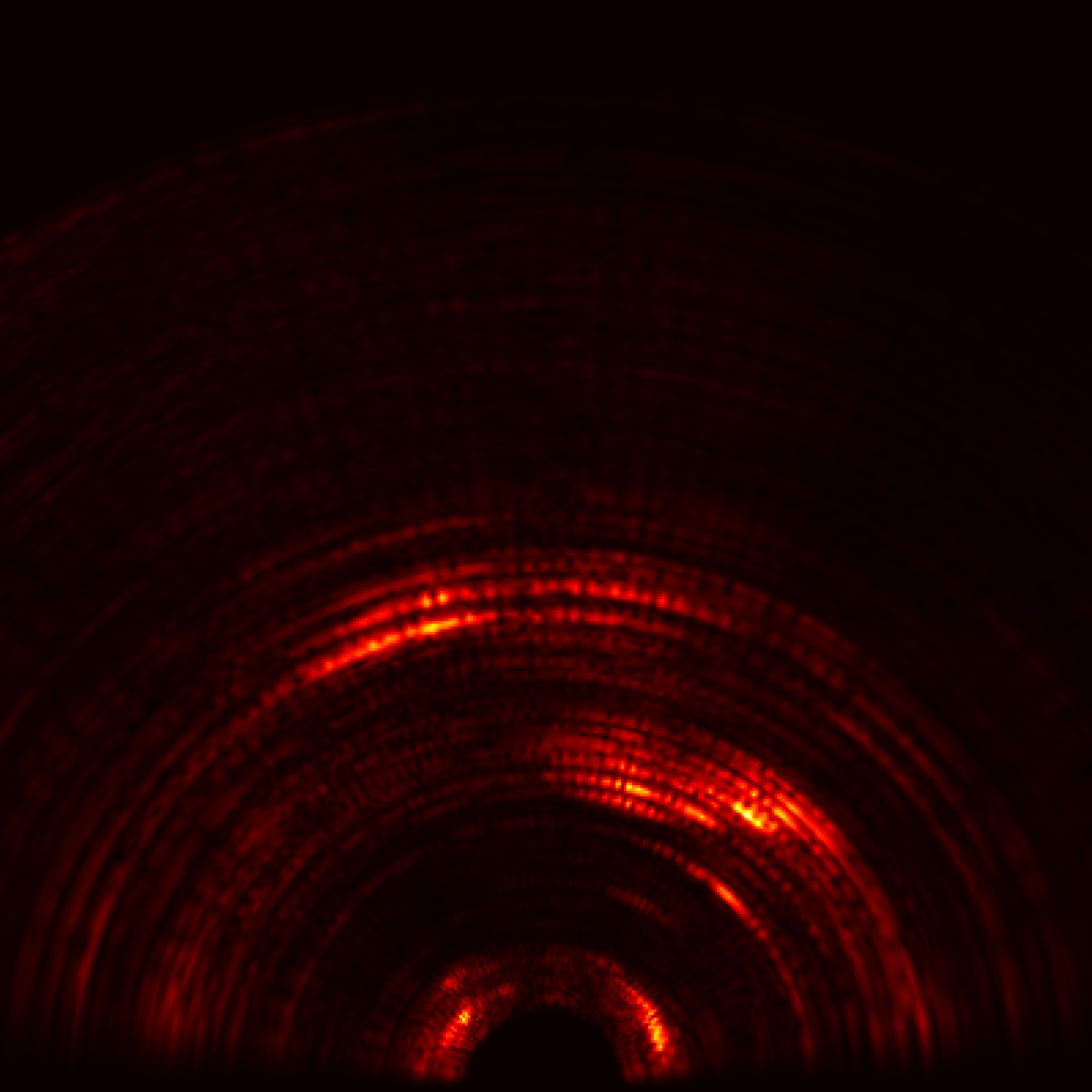} & \includegraphics[width=0.165\linewidth]{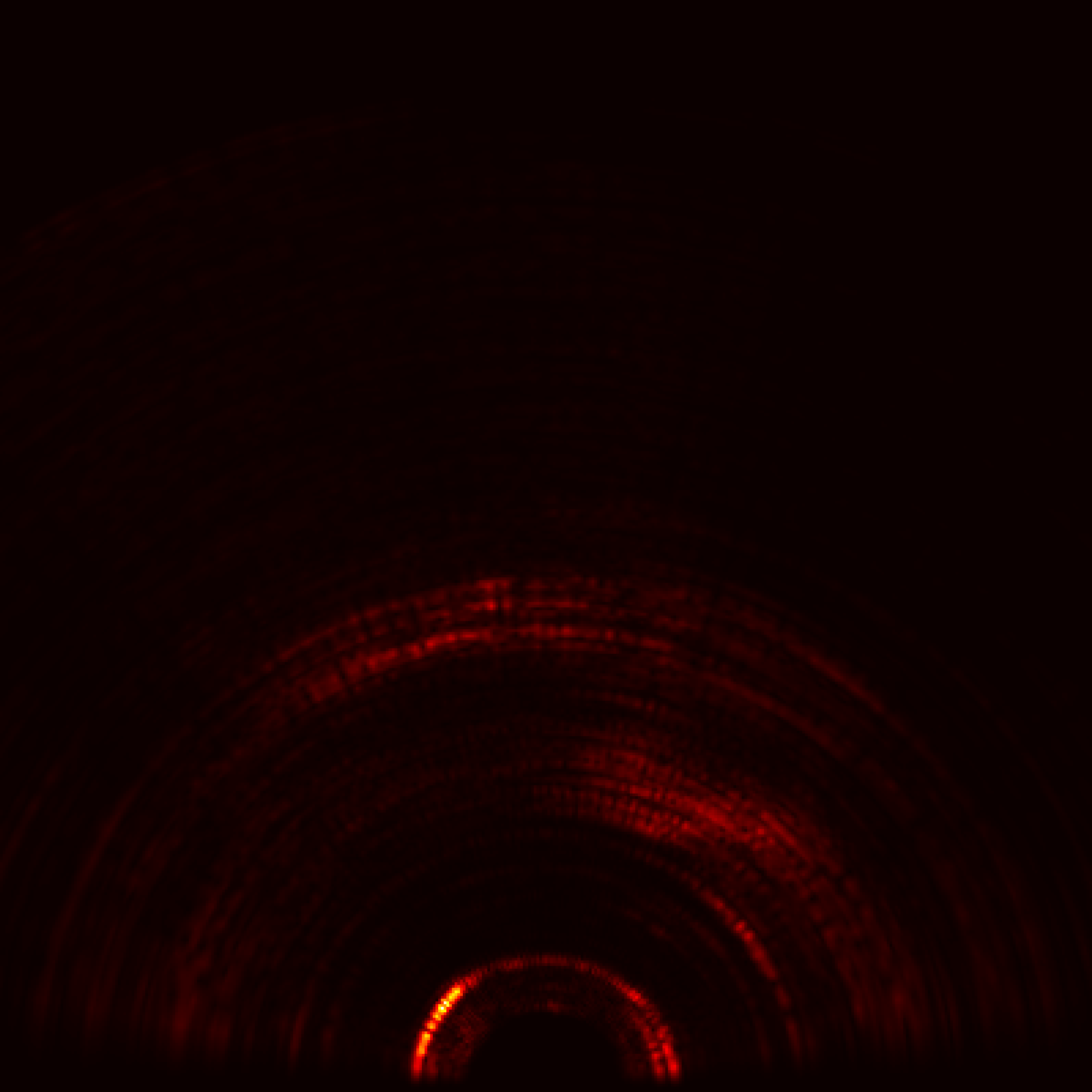} & \includegraphics[width=0.165\linewidth]{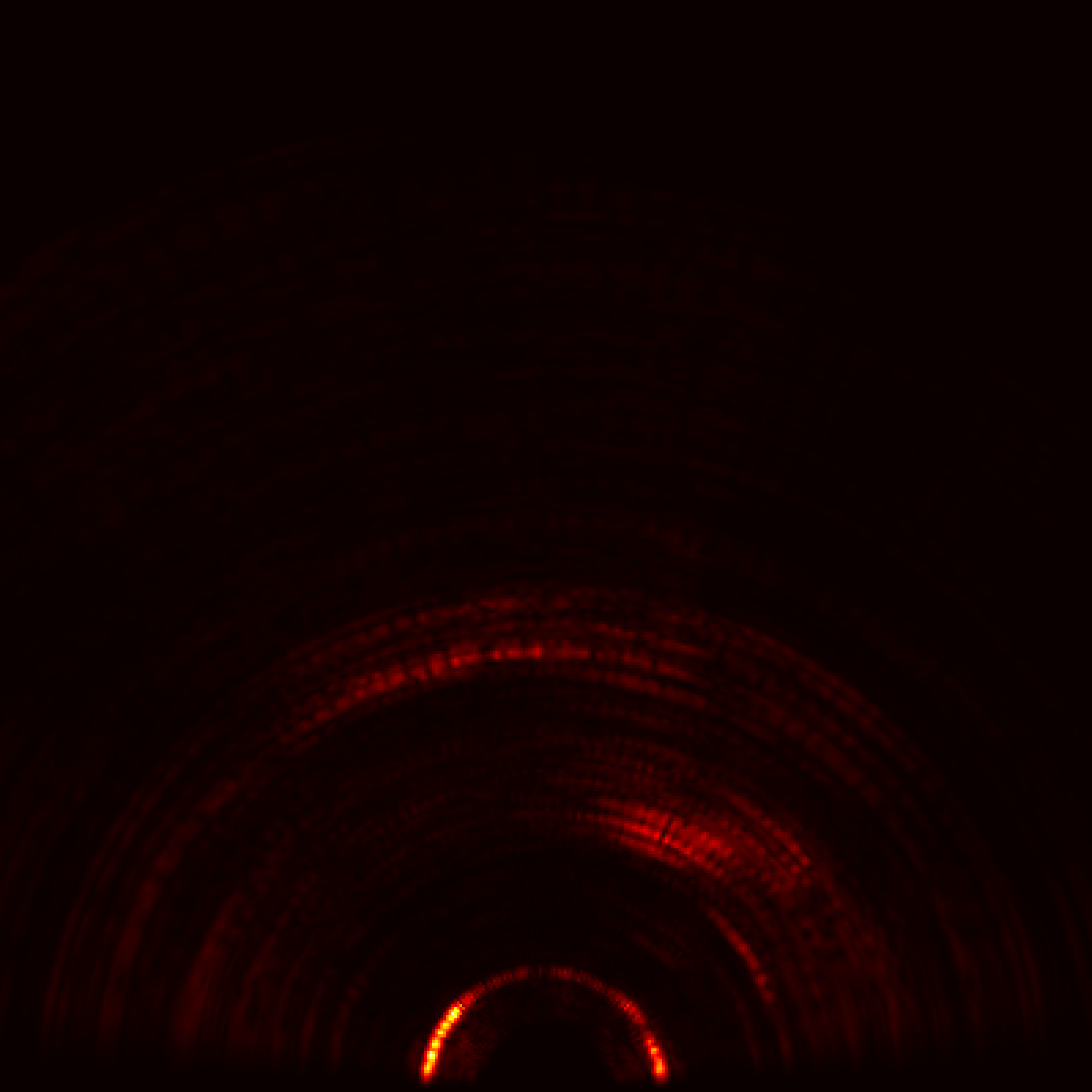} & \includegraphics[width=0.165\linewidth]{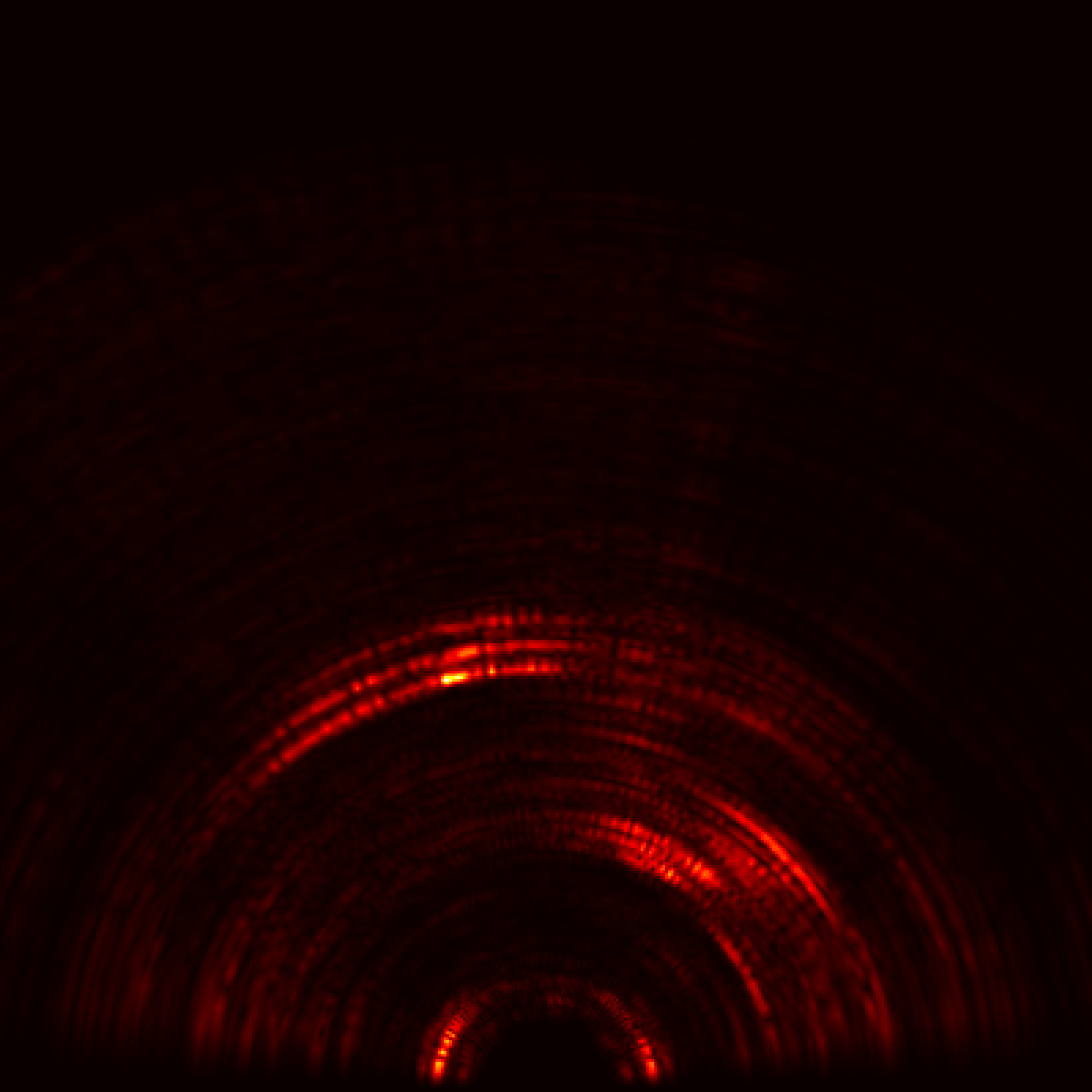} & \includegraphics[width=0.165\linewidth]{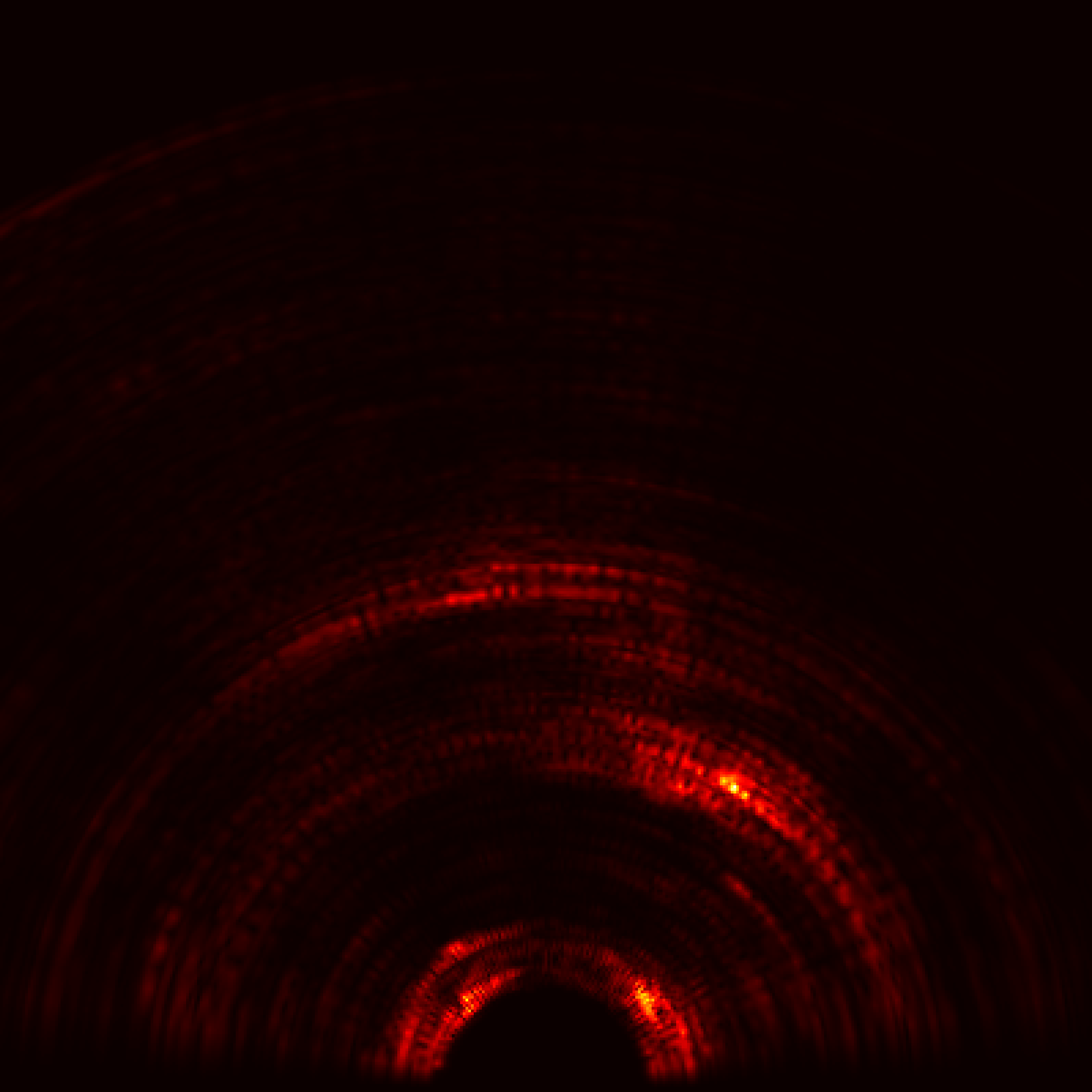} \\
\raisebox{2.0em}{\rotatebox{90}{\scriptsize\textbf{mmIR}}} & \includegraphics[width=0.165\linewidth]{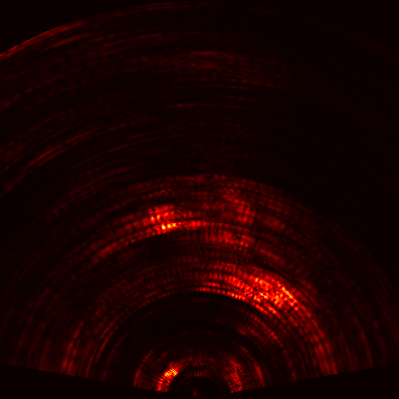} & \includegraphics[width=0.165\linewidth]{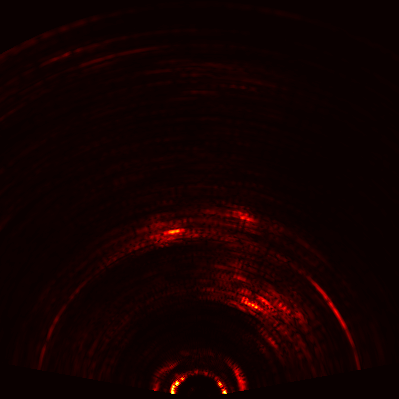} & \includegraphics[width=0.165\linewidth]{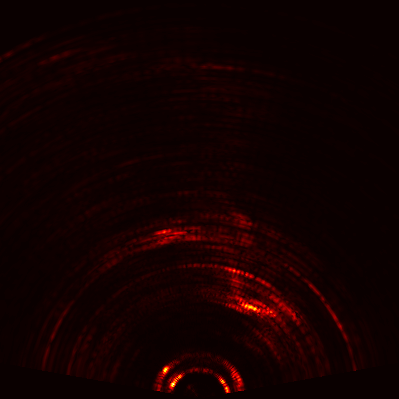} & \includegraphics[width=0.165\linewidth]{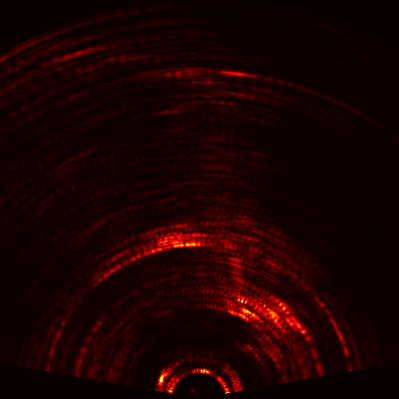} & \includegraphics[width=0.165\linewidth]{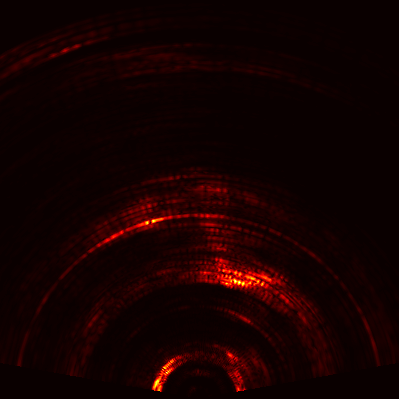} \\
\raisebox{2.0em}{\rotatebox{90}{\scriptsize\textbf{RadarSplat}}} & \includegraphics[width=0.165\linewidth]{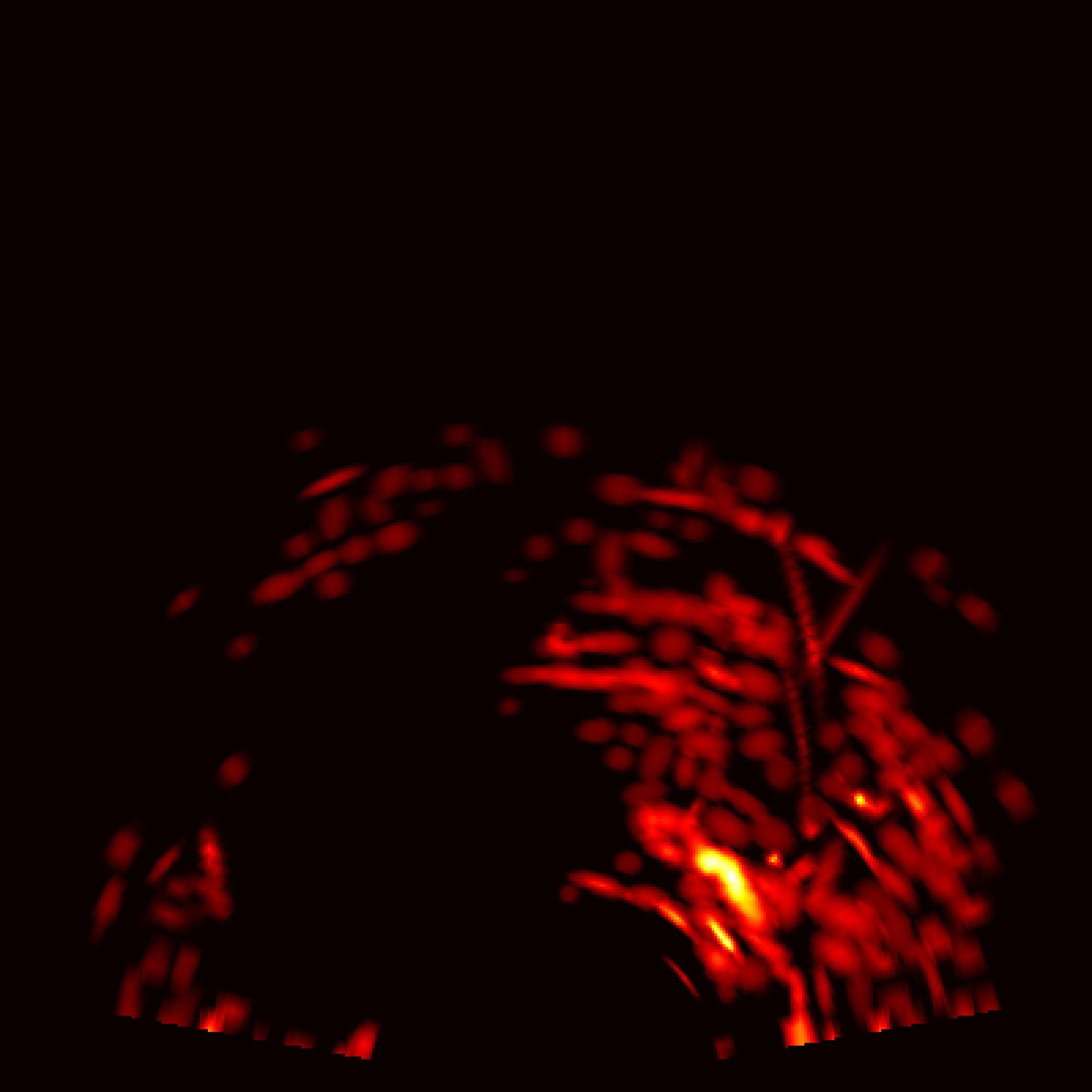} & \includegraphics[width=0.165\linewidth]{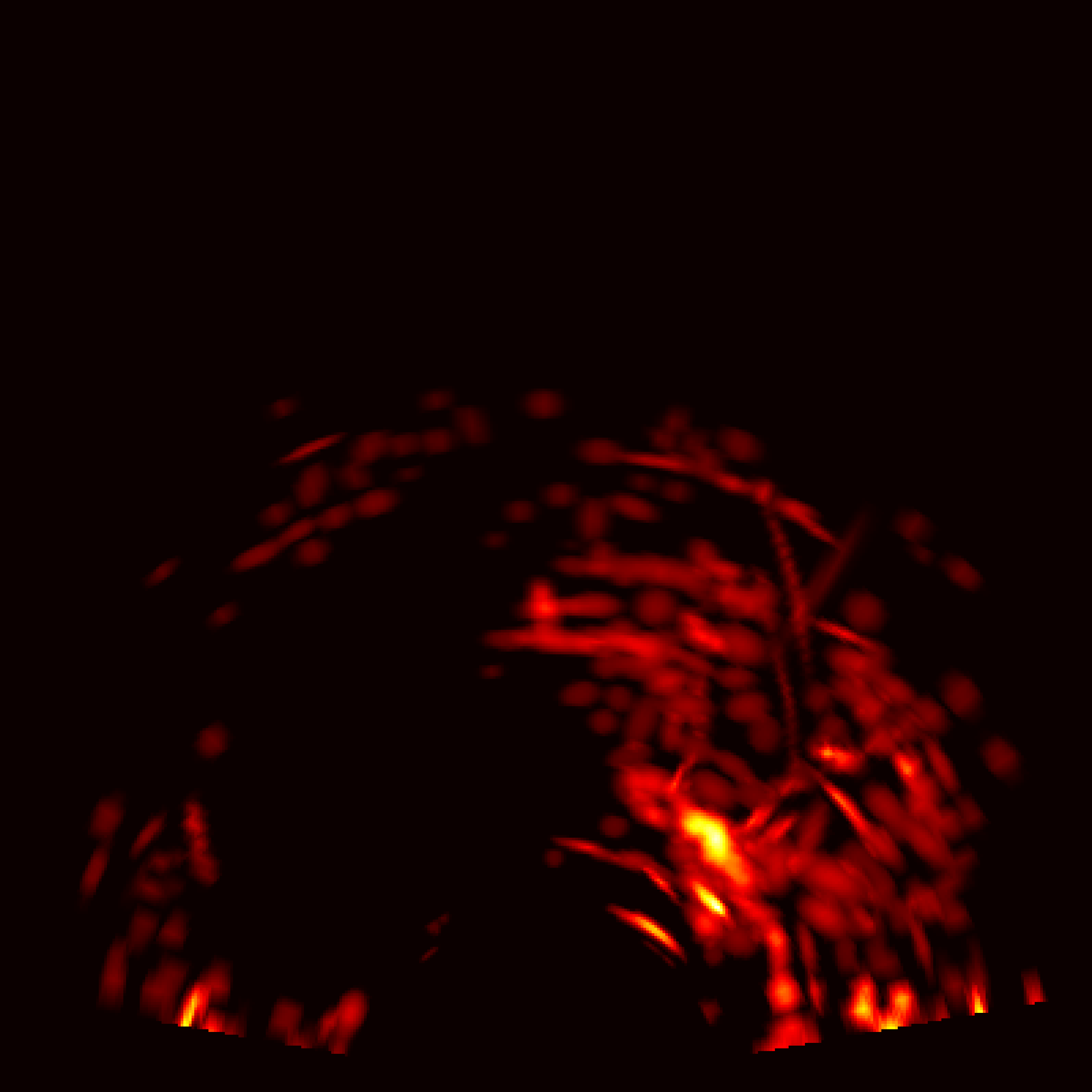} & \includegraphics[width=0.165\linewidth]{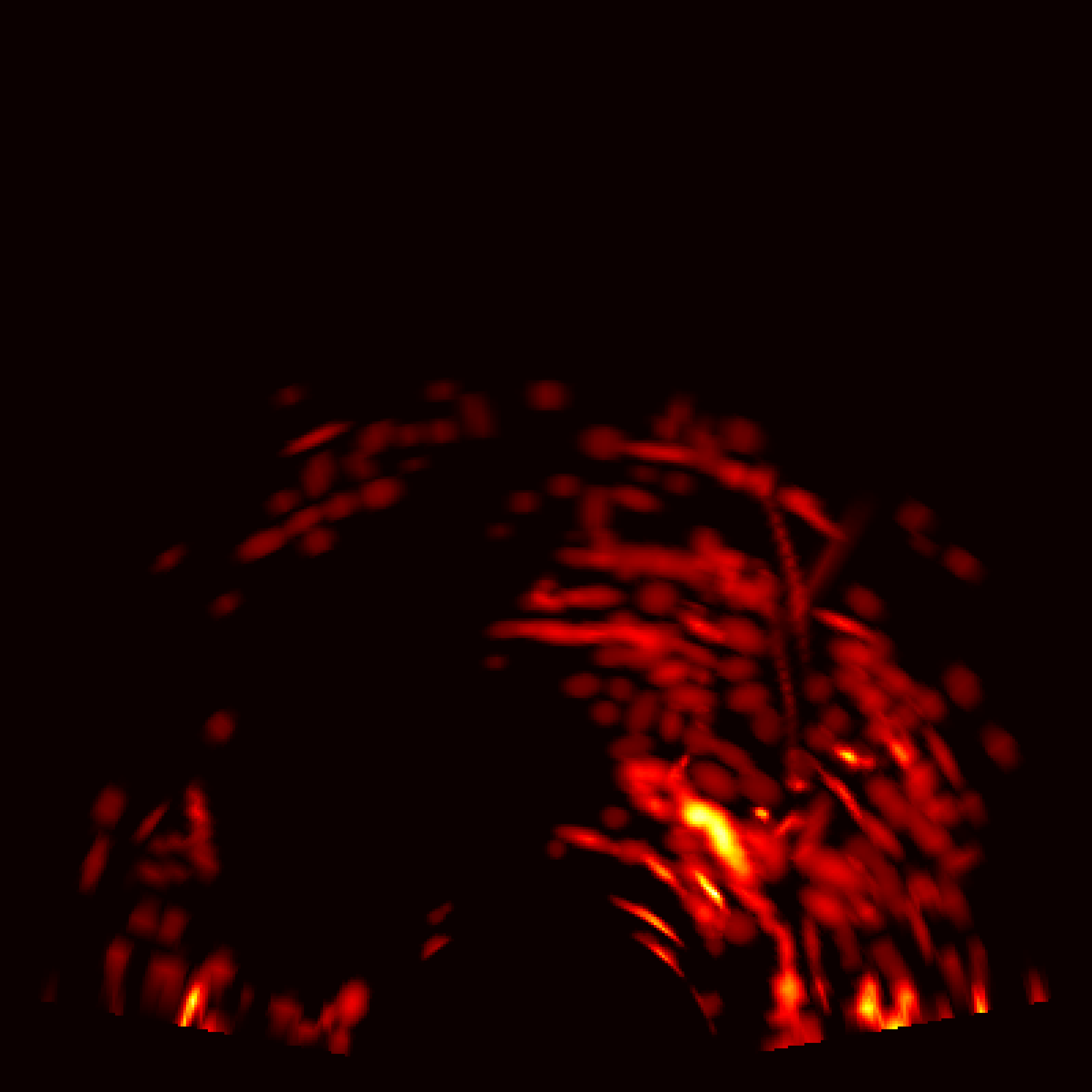} & \includegraphics[width=0.165\linewidth]{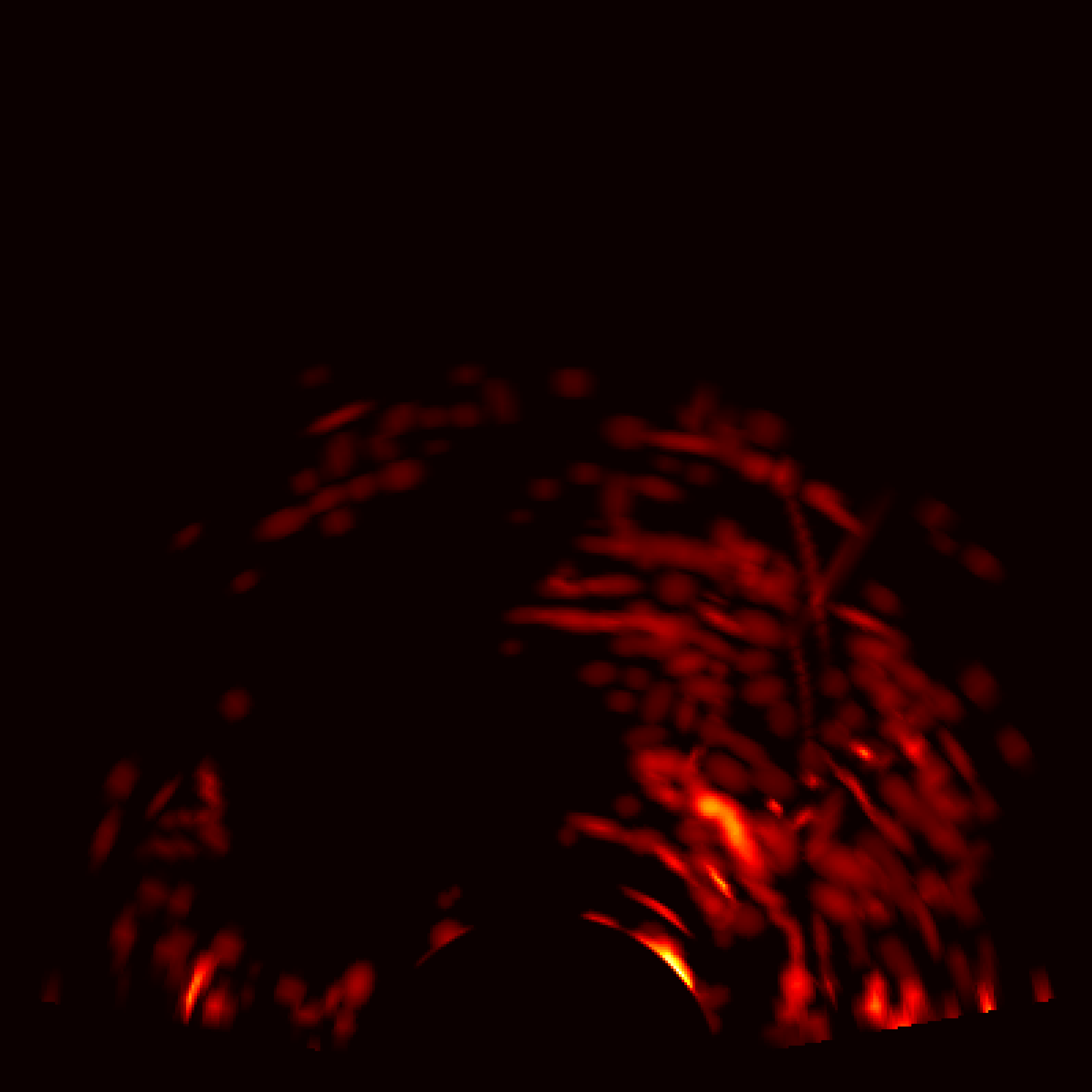} & \includegraphics[width=0.165\linewidth]{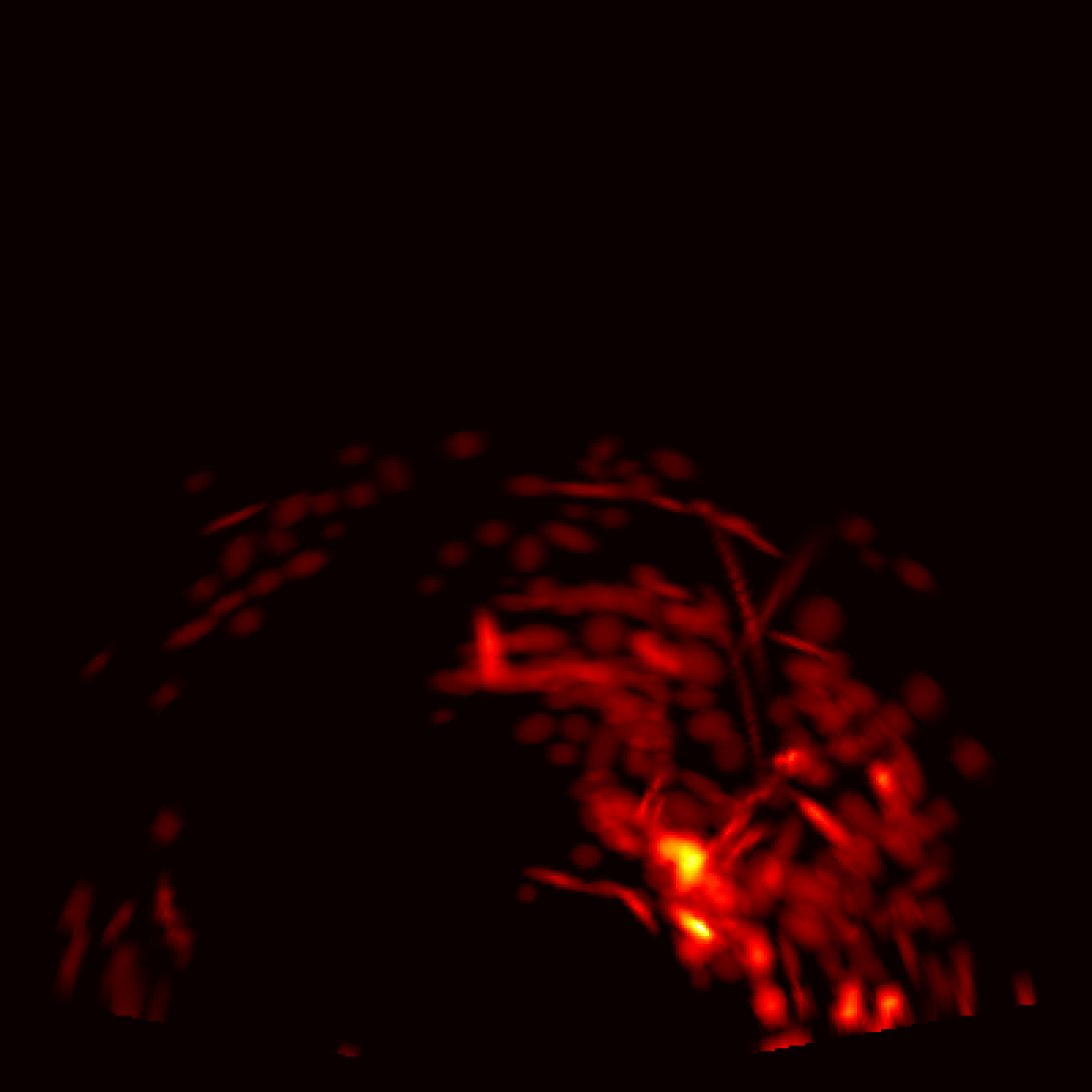} \\
\raisebox{2.0em}{\rotatebox{90}{\scriptsize\textbf{Radar Fields}}} & \includegraphics[width=0.165\linewidth]{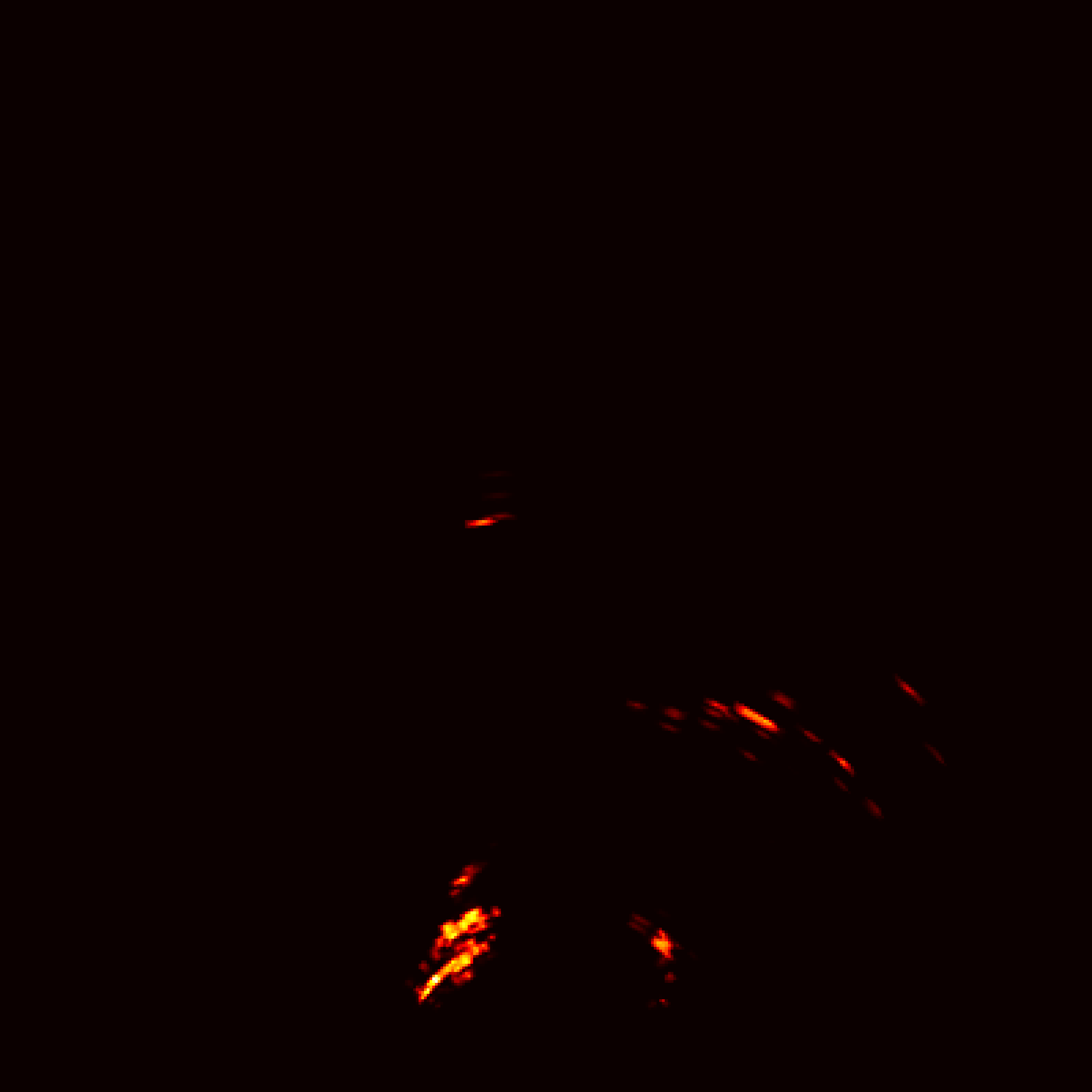} & \includegraphics[width=0.165\linewidth]{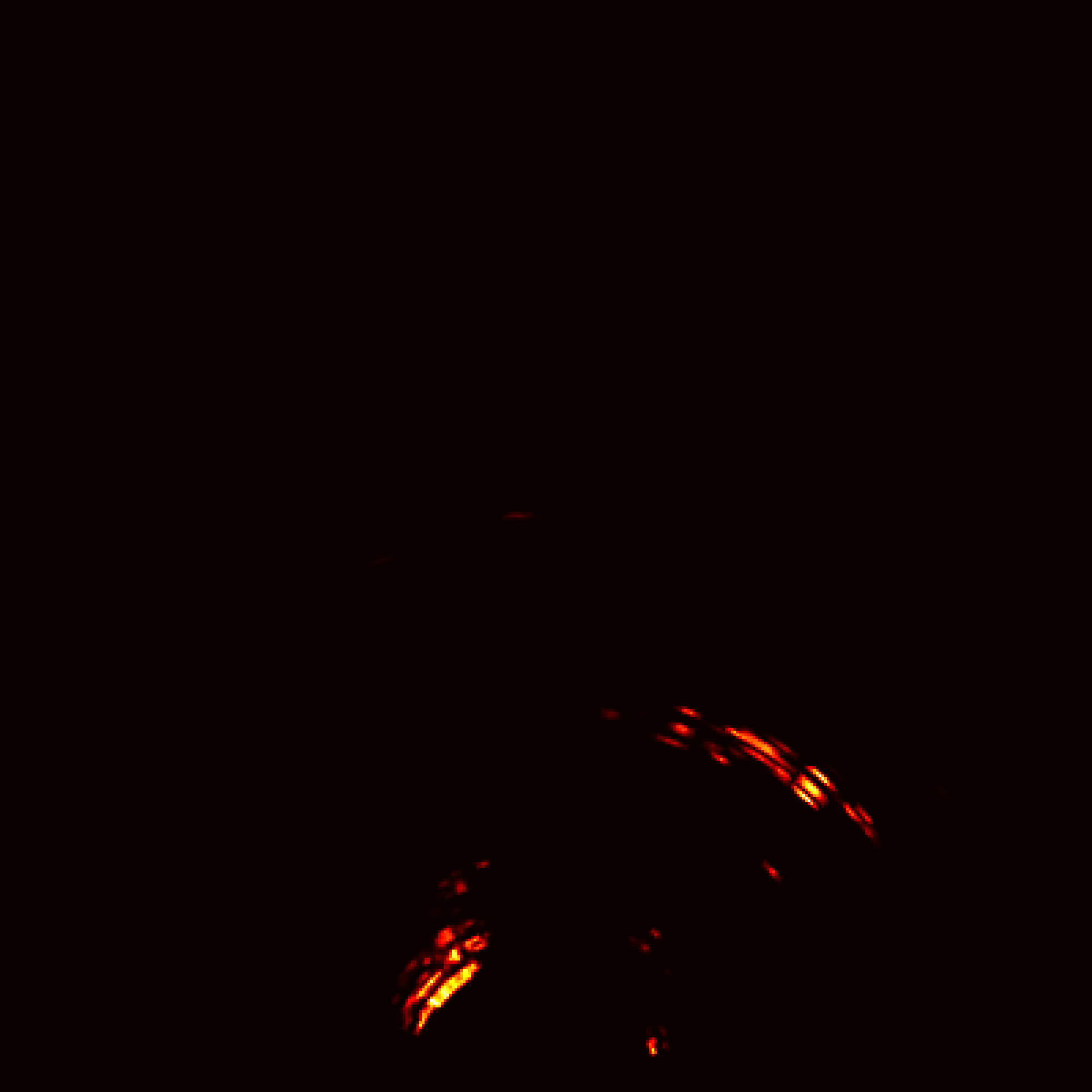} & \includegraphics[width=0.165\linewidth]{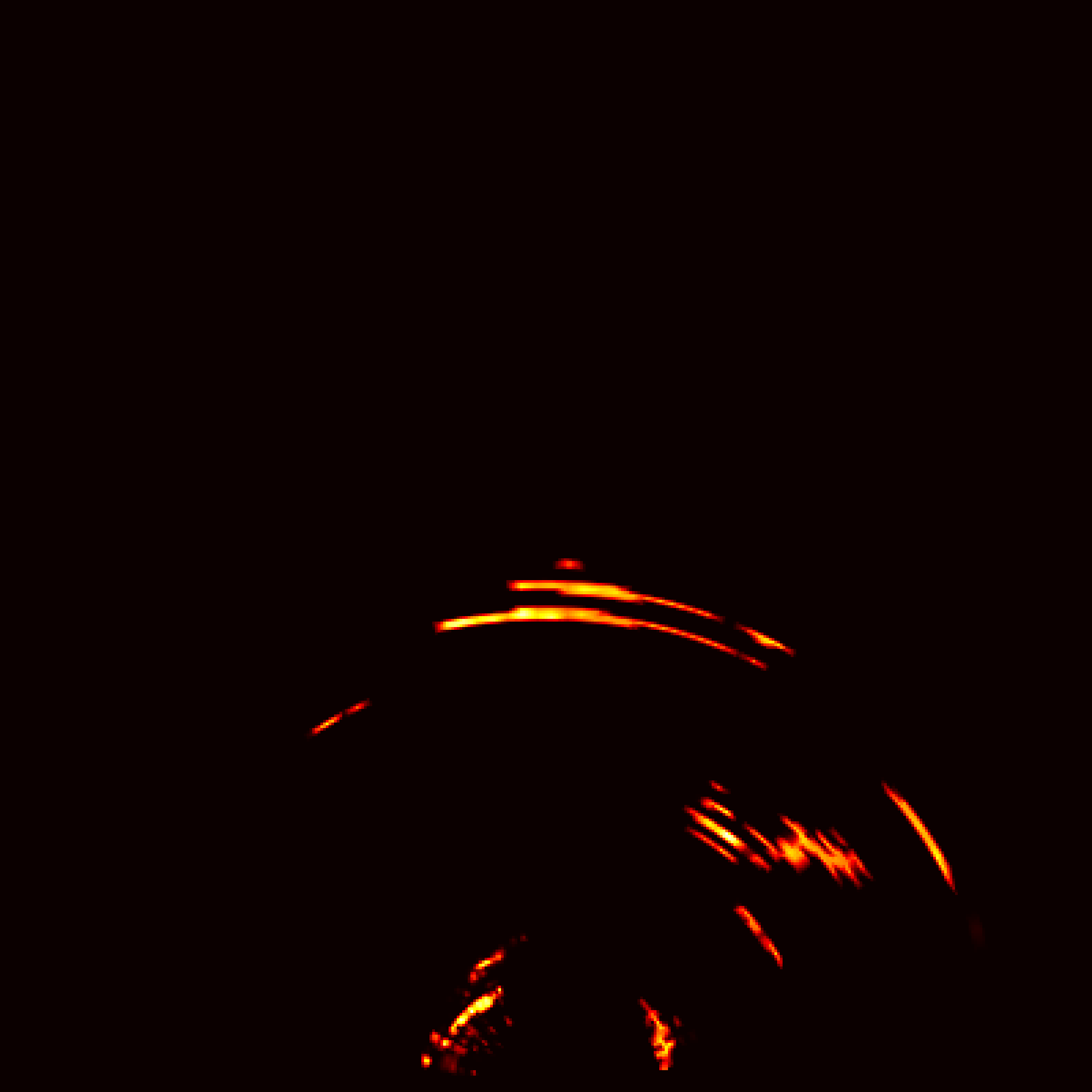} & \includegraphics[width=0.165\linewidth]{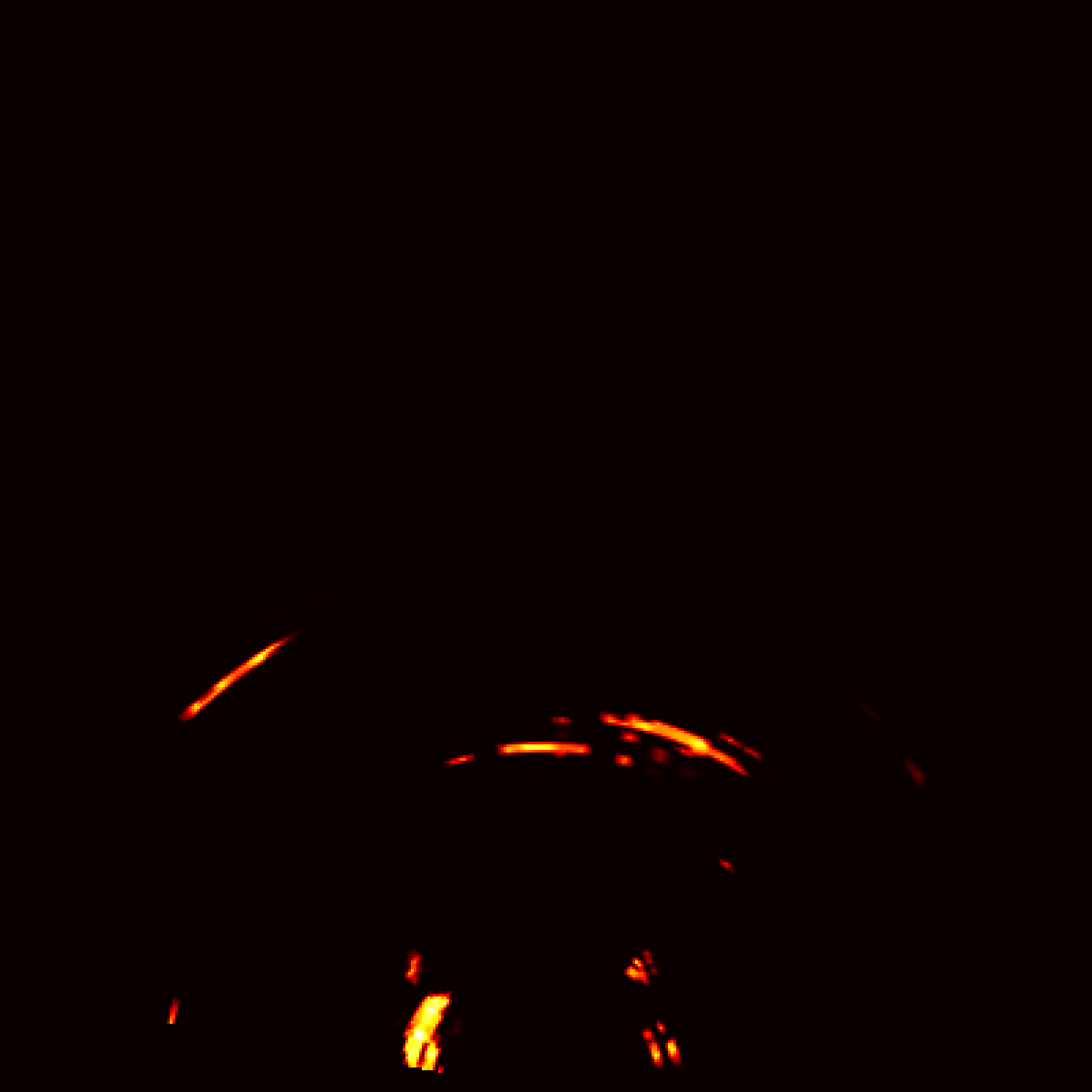} & \includegraphics[width=0.165\linewidth]{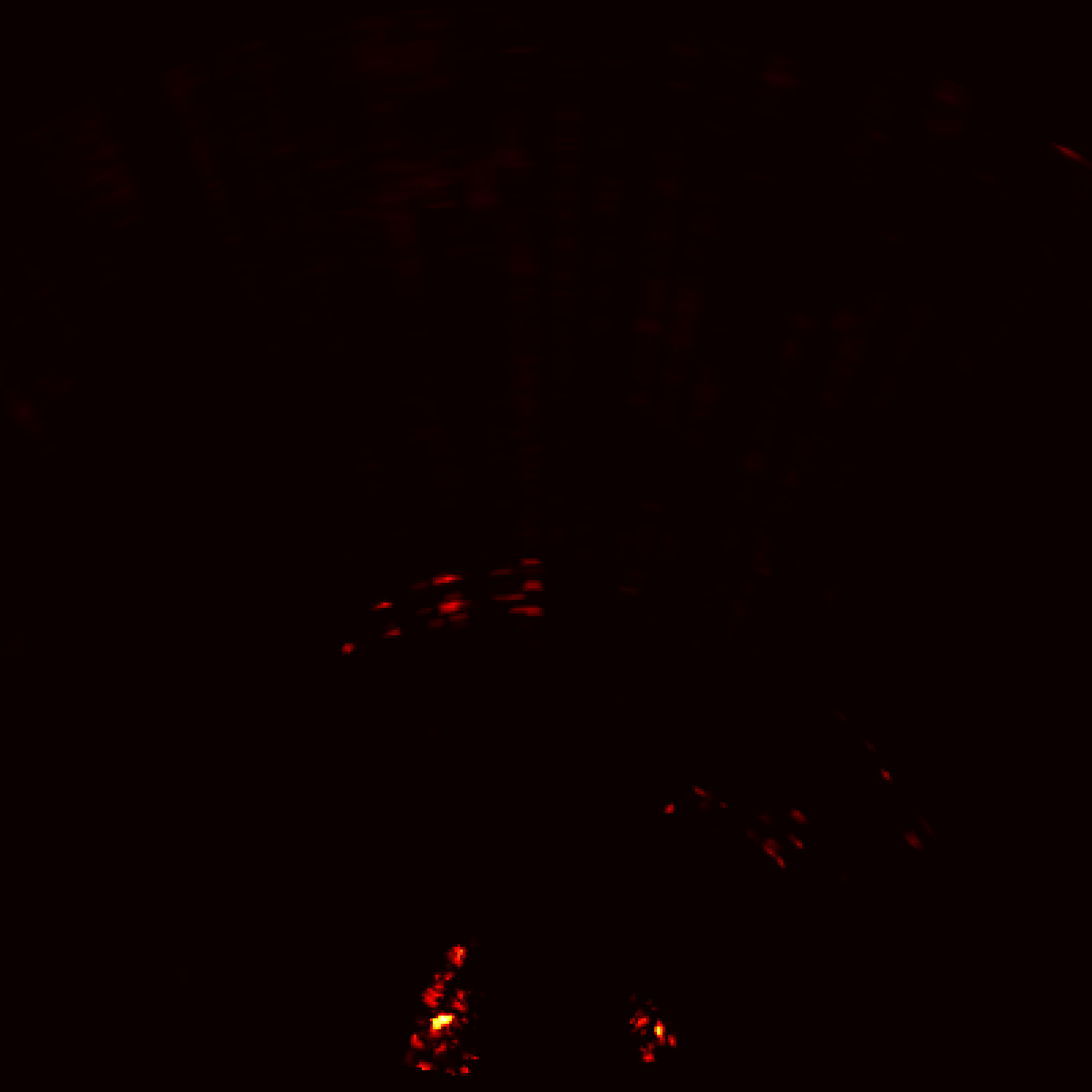} \\
\raisebox{2.0em}{\rotatebox{90}{\scriptsize\textbf{DART}}} & \includegraphics[width=0.165\linewidth]{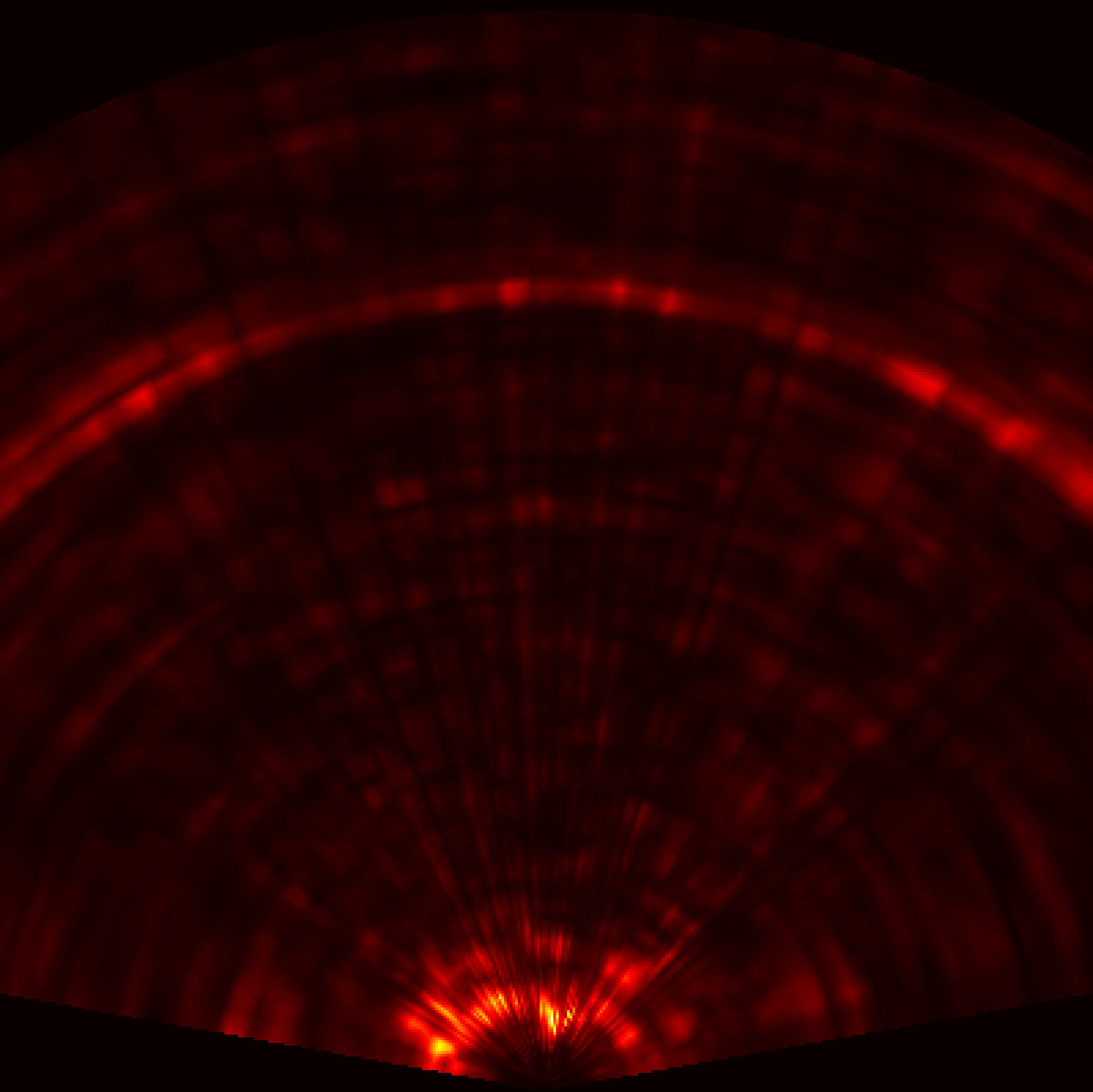} & \includegraphics[width=0.165\linewidth]{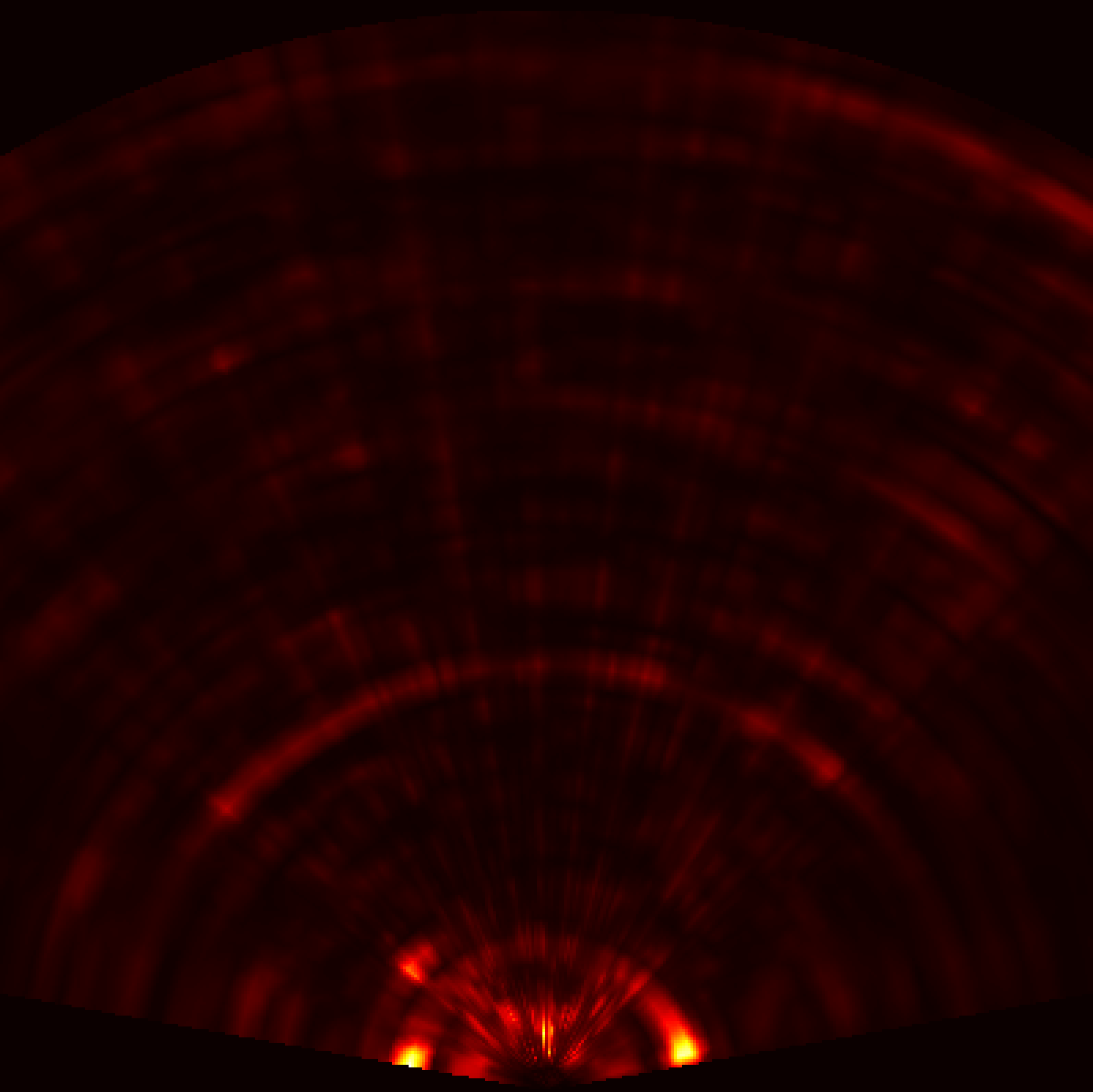} & \includegraphics[width=0.165\linewidth]{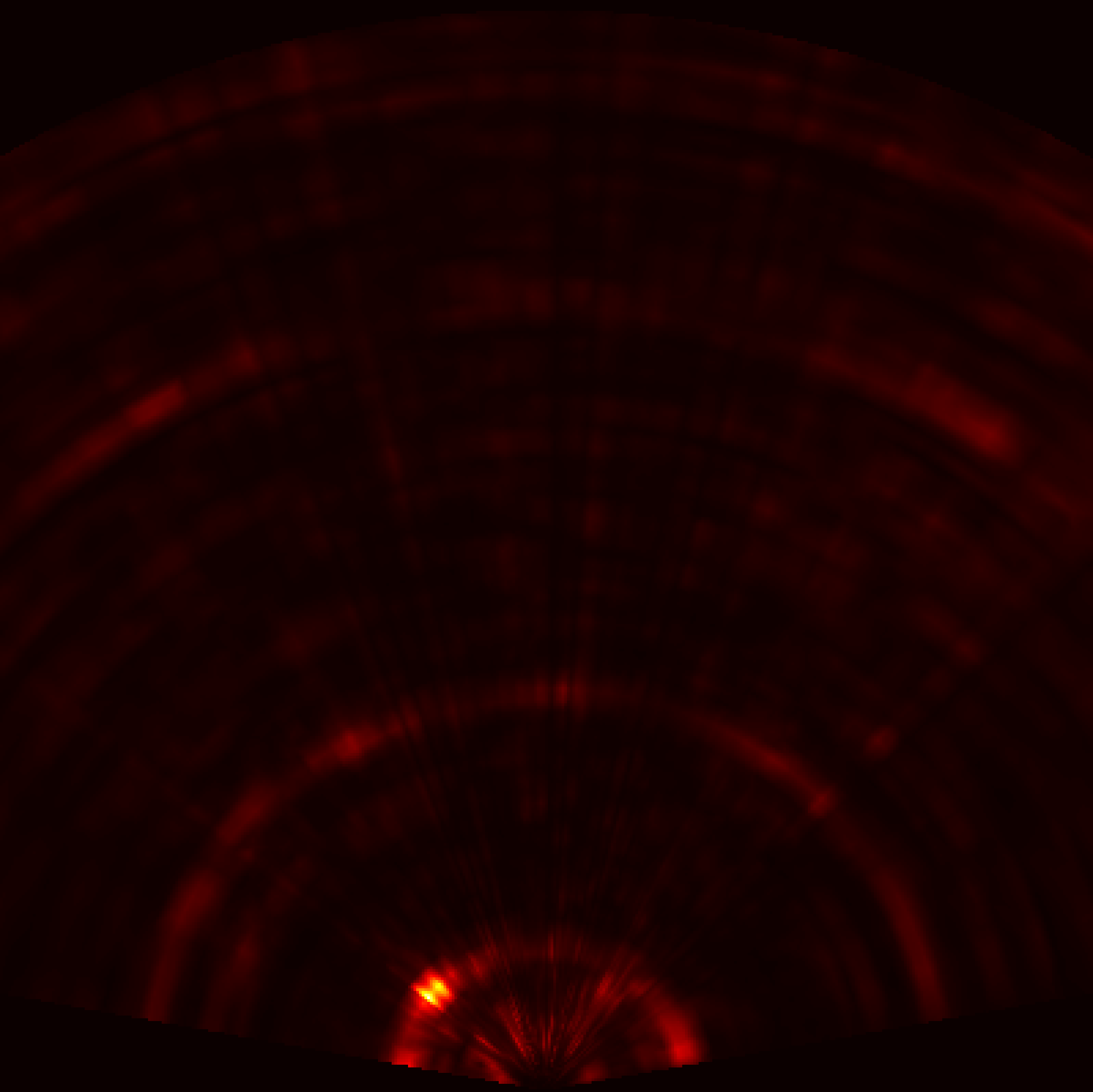} & \includegraphics[width=0.165\linewidth]{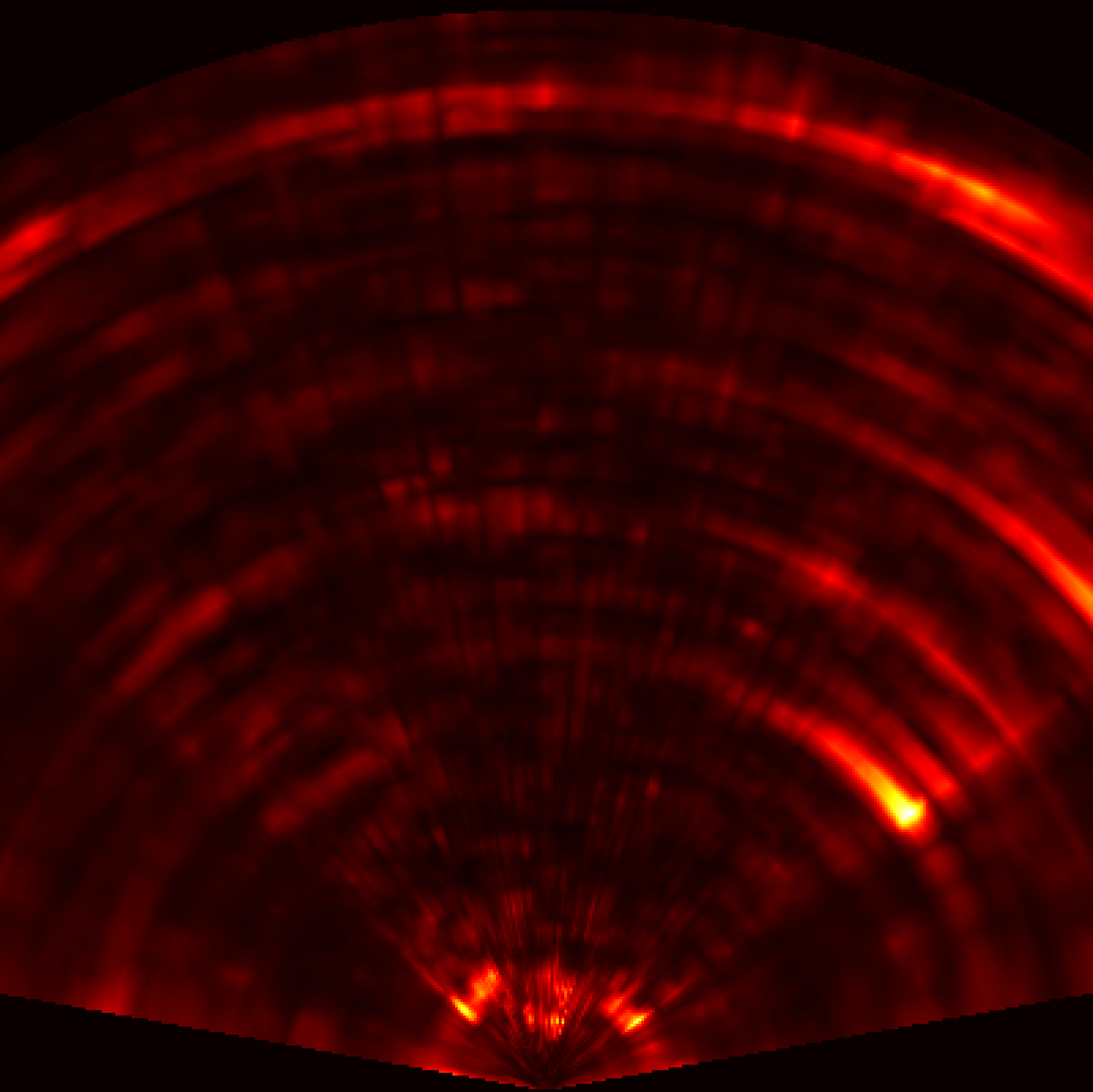} & \includegraphics[width=0.165\linewidth]{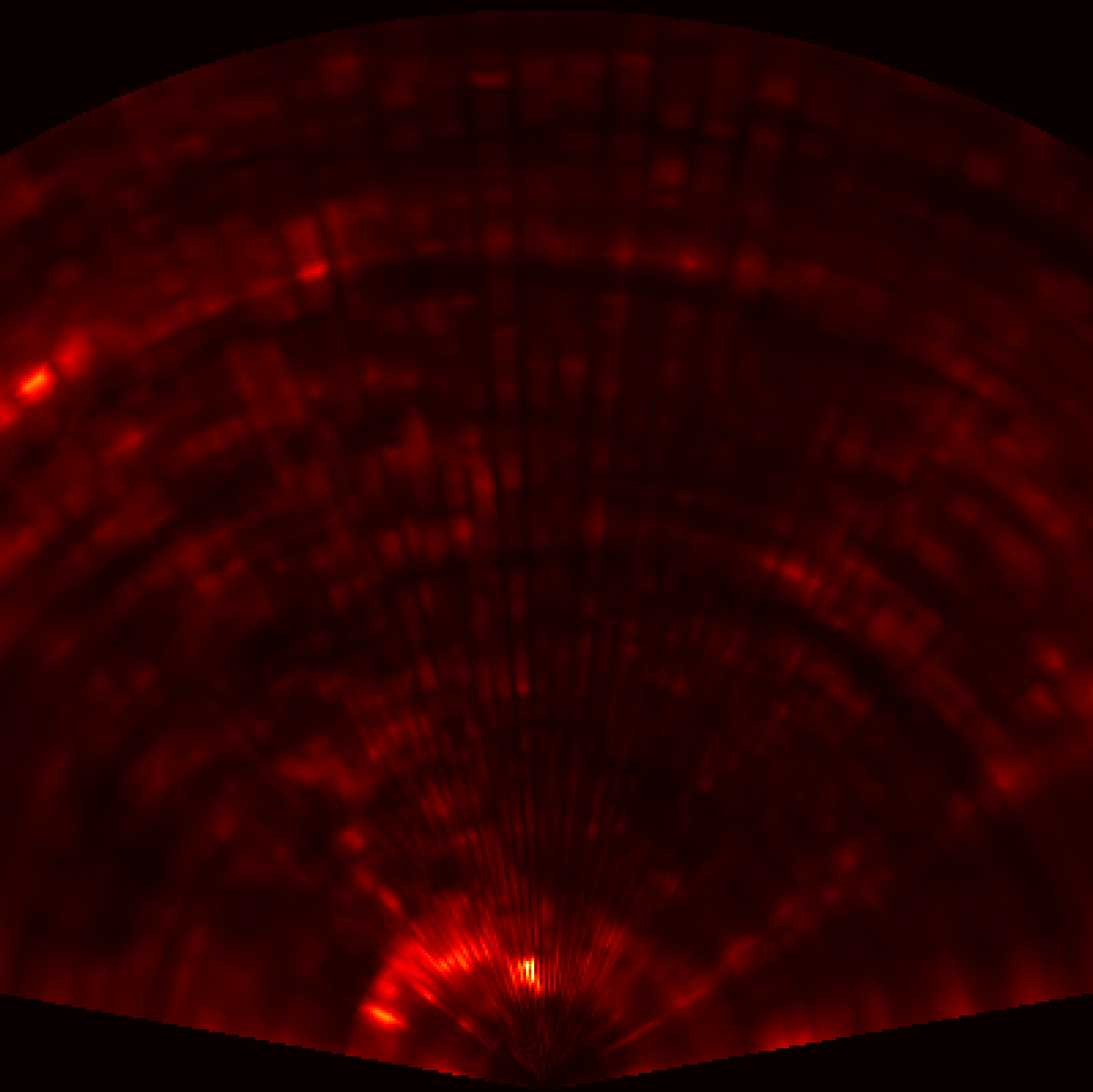} \\
\end{tabular}
\caption{\textbf{Per-frame $|\mathrm{RA}|$ comparison, scene S2\,F300, training frames $F\!+\!1$ to $F\!+\!4$ and held-out test frame $F$ (red header).} Pearson correlation per column, left to right. 3DPS: 0.81, 0.75, 0.82, 0.74, 0.51; mmIR: 0.63, 0.48, 0.53, 0.56, 0.59; RadarSplat: 0.23, 0.27, 0.24, 0.24, 0.15; Radar Fields: -0.00, 0.06, 0.13, 0.03, 0.17; DART: 0.09, 0.14, 0.05, 0.03, 0.03.}
\label{fig:supp_s2_f300}
\end{figure}

Figures~\ref{fig:supp_s0_f135}--\ref{fig:supp_s2_f300} show per-scene side-by-side $|\mathrm{RA}|$ renders for 3DPS, mmIR, RadarSplat, Radar Fields and DART (cascaded) on every training frame and on the held-out test frame $F$ (red header). Each scene spans two figures, with per-frame correlations in the captions. These figures complement the per-scene tables in Sec.~\ref{app:per_scene_ra} and confirm visually that 3DPS preserves dominant scene structure across all eight training frames as well as the held-out test frame, while the three optical-NVS baselines exhibit the failure modes discussed in Sec.~\ref{sec:experiments:results}.

\subsection{CRP and ADC heatmaps and I/Q traces}
\label{app:rdadc:figures}

\begin{figure}[h]
\centering
\includegraphics[width=\linewidth]{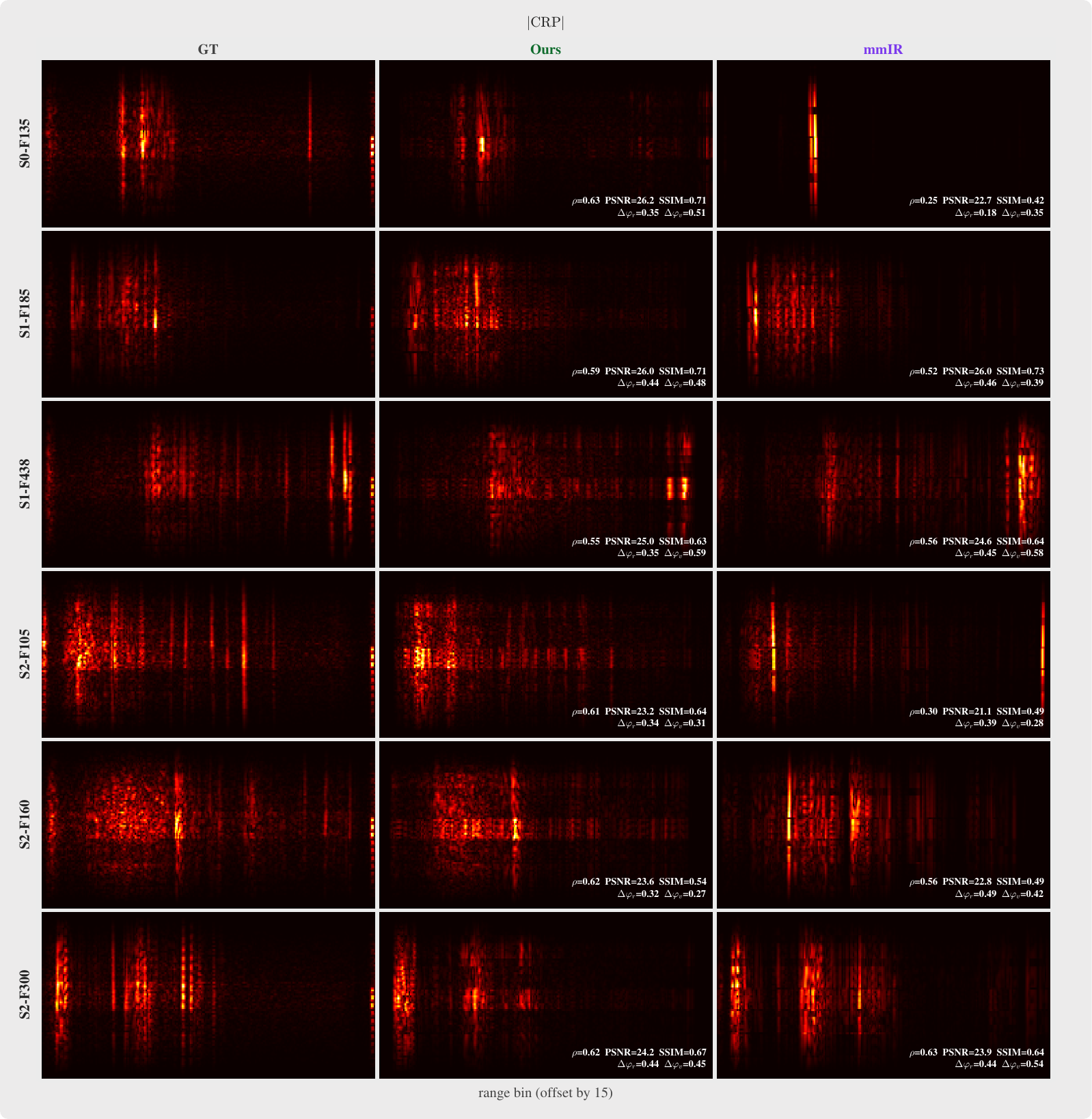}
\caption{\textbf{$|\mathrm{CRP}|$ heatmaps on held-out test frames.} Rows: six ColoRadar scenes. Columns: GT, 3DPS, mmIR. Range bins $[15{:}256]$. Panels are min-max normalized to $[0,1]$. Per-cell overlay: $\rho$ / PSNR / SSIM (magnitude) and $\Delta\varphi_r$ / $\Delta\varphi_v$ (magnitude-weighted phase coherence $\langle\cos(\Delta\varphi)\rangle$ in $[-1, 1]$, random reference 0).}
\label{fig:crp_heatmaps}
\end{figure}

\begin{figure}[h]
\centering
\includegraphics[width=\linewidth]{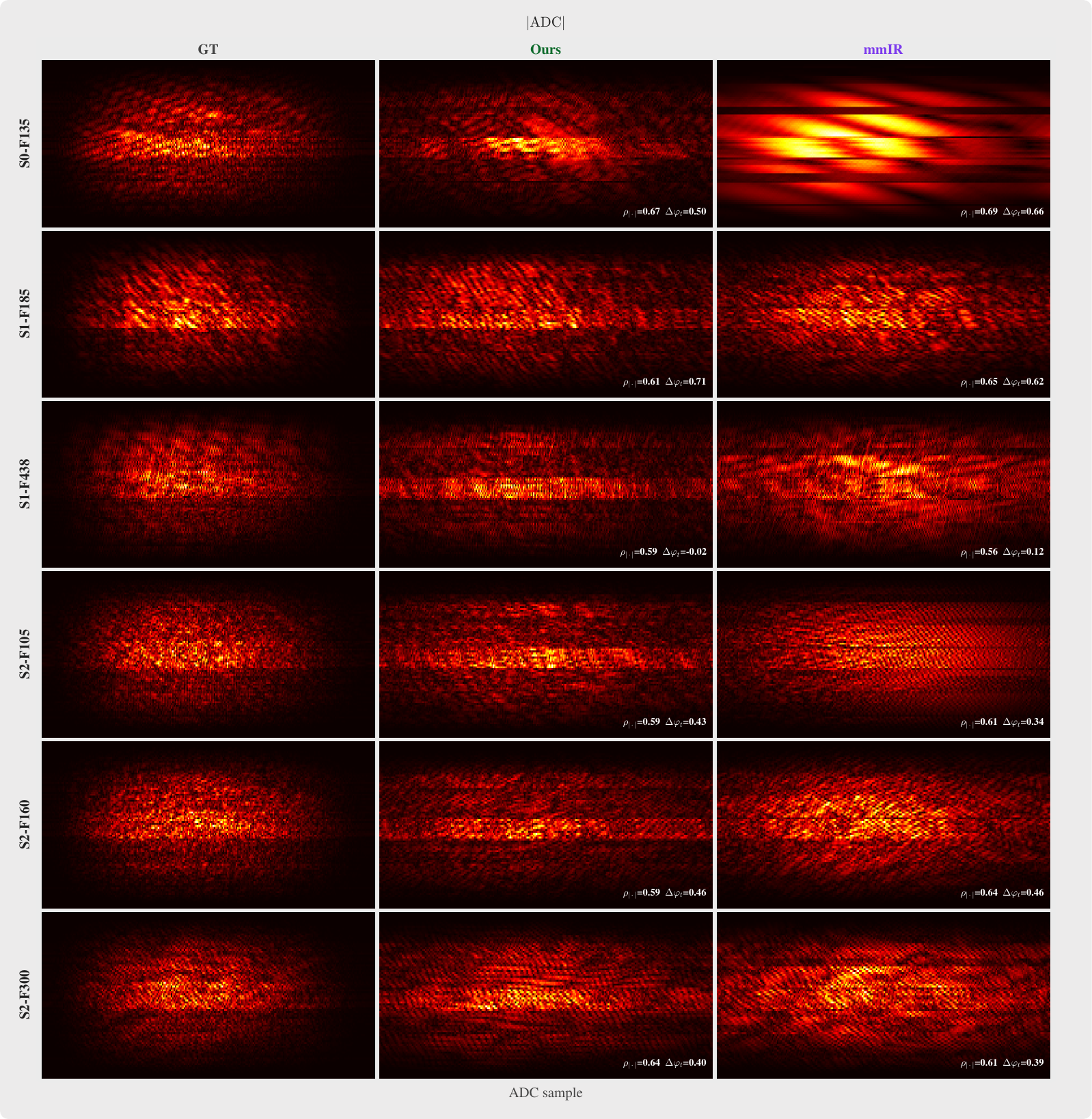}
\caption{\textbf{$|\mathrm{ADC}|$ heatmaps on held-out test frames.} Rows: six ColoRadar scenes. Columns: GT, 3DPS, mmIR. $|\mathrm{ADC}|$ is the inverse range FFT of the near-field-masked CRP. Panels are min-max normalized to $[0,1]$. Per-cell overlay: envelope $\rho_{|\cdot|}$ and $\Delta\varphi_t$.}
\label{fig:adc_heatmaps}
\end{figure}

\begin{figure}[h]
\centering
\includegraphics[width=\linewidth]{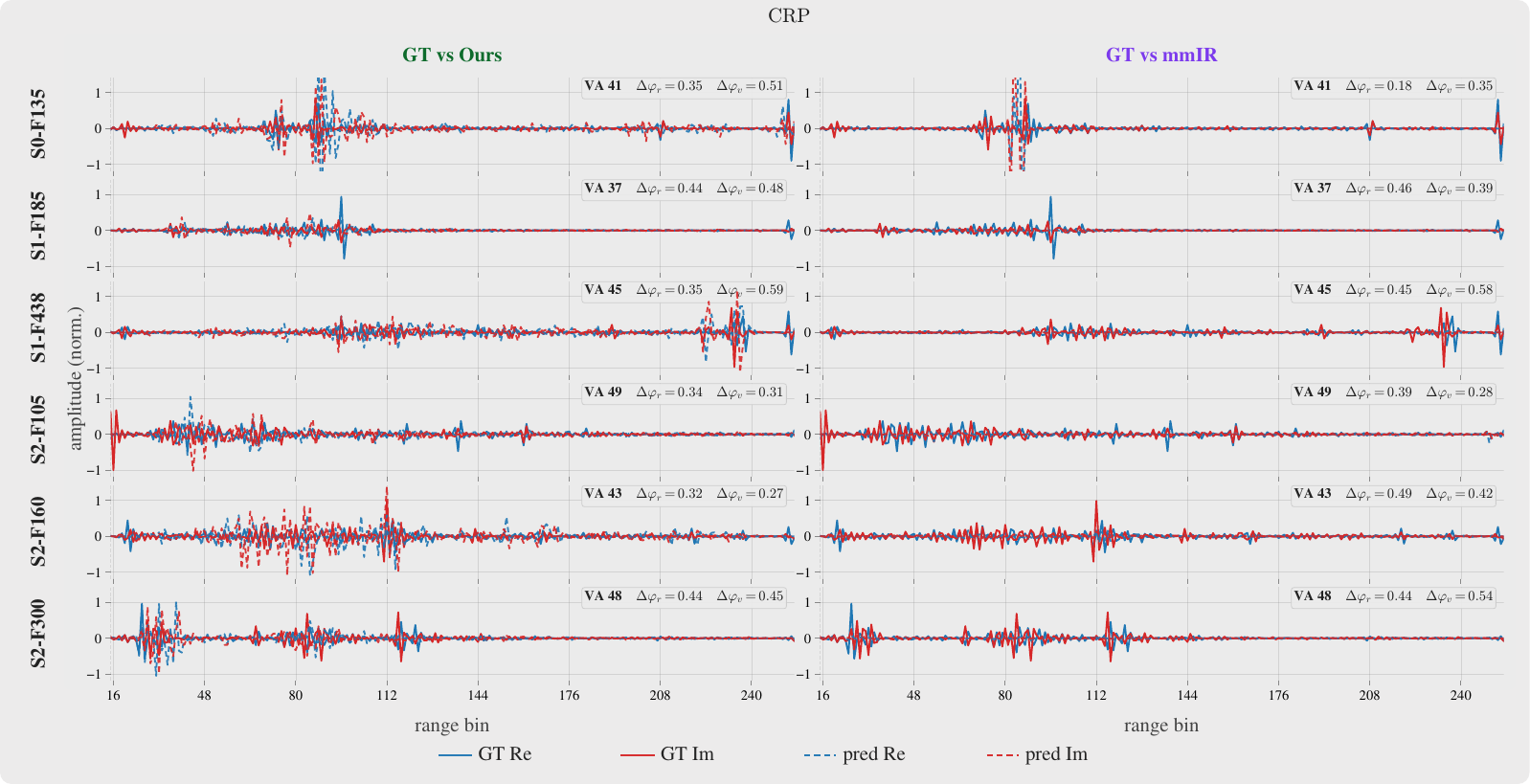}
\caption{\textbf{CRP I/Q traces on held-out test frames.} Rows: six ColoRadar scenes, at the highest-energy virtual antenna per scene, range bins $[15{:}256]$. Left: GT vs 3DPS. Right: GT vs mmIR. Solid lines are GT real (blue) and imaginary (red); dashed lines are predictions. Per-panel: $\Delta\varphi_r$ / $\Delta\varphi_v$ (relative-phase coherence $\langle\cos(\Delta\varphi)\rangle$). The traces show that 3DPS recovers the dominant amplitude envelope while the per-bin absolute phase varies (per-(VA, range) absolute phase is unconstrained by the magnitude-only training loss; only the relative-phase content is captured).}
\label{fig:crp_iq_traces}
\end{figure}

\begin{figure}[h]
\centering
\includegraphics[width=\linewidth]{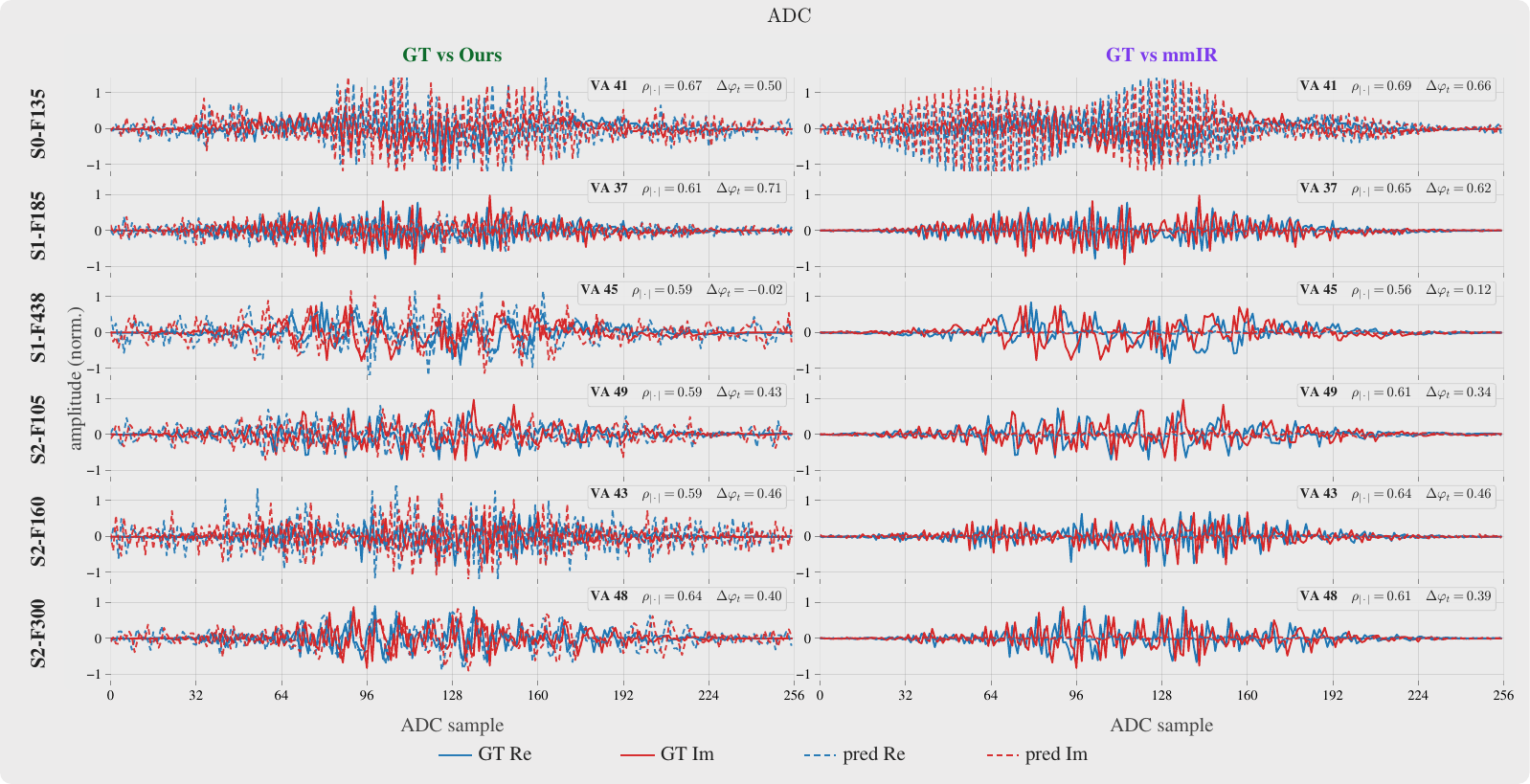}
\caption{\textbf{ADC I/Q traces on held-out test frames.} Rows: six ColoRadar scenes, at the same virtual antenna as Fig.~\ref{fig:crp_iq_traces}, across 256 fast-time samples. Left: GT vs 3DPS. Right: GT vs mmIR. Solid lines are GT real (blue) and imaginary (red); dashed lines are predictions. Per-panel: envelope $\rho_{|\cdot|}$ and $\Delta\varphi_t$.}
\label{fig:adc_iq_traces}
\end{figure}

Figures~\ref{fig:crp_heatmaps}--\ref{fig:adc_iq_traces} show the qualitative comparison of $|\mathrm{CRP}|$ and $|\mathrm{ADC}|$ between GT, 3DPS and mmIR across all six scenes (held-out test frames). Sionna RT also emits complex outputs, and its scores are in Tables~\ref{tab:crp_adc_train} and~\ref{tab:crp_adc_test}. The three optical-NVS baselines emit magnitude-only RA. Each panel is normalized identically to its $|\mathrm{RA}|$ counterpart in the main paper, so the visual structure is directly comparable.

\FloatBarrier
\section{Wall-clock and peak GPU memory}
\label{app:runtime}

Table~\ref{tab:runtime} reports per-scene wall-clock and peak GPU memory measured on a single NVIDIA RTX~4090 (24\,GB) for 3DPS and the baselines, on the same 6-scene ColoRadar benchmark and the same 9-frame split (8 training viewpoints, 1 held-out test frame). All measurements include forward+backward+optimizer+I/O. mmIR and Sionna~RT~\cite{sionna} run the same 8-viewpoint, 500-iteration fit.

\begin{table}[h]
\centering
\caption{\textbf{Wall-clock and peak GPU memory per scene} on the 6-scene benchmark, single RTX~4090. Wall-clock is reported as mean across scenes with per-scene min--max range for measured methods. }
\label{tab:runtime}
\small
\setlength{\tabcolsep}{4pt}
\resizebox{\linewidth}{!}{%
\begin{tabular}{lcccc}
\toprule
\textbf{Method} & \textbf{Mean wall-clock (min)} & \textbf{Per-scene range (min)} & \textbf{Per view (ms)} & \textbf{Peak GPU mem (GB)} \\
\midrule
DART (cascaded)~\cite{huang2024dart}            & 0.43   & 0.4--0.5    & --   & 4.4 \\
Radar Fields~\cite{10.1145/3641519.3657510}     & 1.53   & 1.52--1.54    & --   & 2.4 \\
RadarSplat~\cite{kung2025radarsplat}            & 19.74  & 14.3--22.8  & 591  & 14.2 \\
mmIR~\cite{mmIR2026}                            & 77.58  & 75.3--81.8  & 1164 & 4.4 \\
Sionna~RT~\cite{sionna}                         & 12.20  & 11.5--13.1  & 183  & 21.8 \\
\midrule
\textbf{3DPS (Ours)}                            & \textbf{3.02} & \textbf{2.3--3.7} & \textbf{45} & \textbf{0.35} \\
\bottomrule
\end{tabular}%
}
\end{table}

Per training view per iteration, 3DPS takes 45 ms against 183 ms for Sionna RT, 591 ms for RadarSplat, and 1164 ms for mmIR. DART and Radar Fields sample rays rather than render whole views, so this measure does not apply to them. DART's peak memory is the process peak, which occurs in its render step.

3DPS peak memory grows linearly with $N$: 45, 96, 181, 355 and 863 MiB at $N$ = 2k, 5k, 10k, 20k and 50k on S0\,F135, about 17 MiB per 1k points. The range-profile buffer is fixed, so a larger scene at the same $N$ costs no more memory.

\FloatBarrier
\section{mmIR and Sionna RT baselines}
\label{app:mmir_baseline}

mmIR~\cite{mmIR2026} was published as a single-viewpoint fit, so we extend it to joint optimization over the same eight training views and held-out frame, using the same aligned poses. It runs 500 iterations with its Jones BSDF, up to three bounces, learned per-vertex materials and normals, and frozen geometry. Antenna patterns stay frozen and error-map initialization stays off, because both assume one viewpoint. Main-paper results use a log-compressed $|\mathrm{RA}|$ loss, which reaches 0.570 on training views and 0.454 held-out. The linear loss of the mmIR paper reaches 0.822 and 0.370. Sionna RT~\cite{sionna} runs on the LiDAR mesh with per-vertex ITU parameters, three bounces, the measured antenna patterns, and our loss, split and scoring. It reaches 0.430 on training views and 0.380 held-out.

\FloatBarrier
\section{RadarSplat hyperparameter selection}
\label{app:radarsplat_baseline}

We benchmark RadarSplat~\cite{kung2025radarsplat} using its public implementation at its demo settings: 20{,}000 randomly initialized Gaussians and 2{,}000 iterations, refining every 100 iterations from iteration 500. We adapt it to the cascade with a 9-frame sequence, a $90^\circ$ pose rotation, a loss crop to the populated azimuth wedge, and resampling to uniform azimuth. Its multipath and occupancy terms get zero weight, because their inputs do not exist for these captures. We did not tune RadarSplat per scene. Per-scene wall-clock and test correlation are in Table~\ref{tab:runtime} and Sec.~\ref{app:per_scene_ra}.

\FloatBarrier
\section{Limitations and future work}
\label{app:limitations}

The conclusion (Sec.~\ref{sec:conclusion}) summarizes the main limitations briefly. This section expands on each and identifies the corresponding open directions.

\textbf{Single-bounce path tracing.} The renderer evaluates a single TX$\to$scatterer$\to$RX path per oriented point. For static outdoor scenes dominated by direct surface returns (the regime of our ColoRadar benchmark) this reaches 0.587 mean test $|\mathrm{RA}|$ Pearson. In indoor multipath-dominated environments, where second- and third-order reflections carry comparable energy to direct returns~\cite{lu2020see}, it reaches 0.541 (Sec.~\ref{app:per_scene_ra}). Integrating the closed-form BSDF + PSF splatting recipe into a multi-bounce framework is open.

\textbf{Magnitude-only loss leaves per-bin absolute phase unconstrained.} 3DPS supervises on $|\mathrm{RA}|$ for fair comparison with the three magnitude-only baselines, leaving $\sim\!22{,}000$ per-(VA, range) absolute phase DOFs unconstrained by the loss --- specifically, free up to per-VA RF calibration drift and per-range ADC sample-zero offsets. We side-step that nuisance phase at evaluation time by reporting differential-phase metrics ($\Delta\varphi_r$, $\Delta\varphi_v$, $\Delta\varphi_t$ in Table~\ref{tab:results}) that are absolute-phase-invariant by construction; a tighter \emph{absolute} phase fit would require adding a complex (Re/Im) loss term during training. A natural future extension is a two-stage training recipe (first stage: magnitude loss to fit geometry and material; second stage: bounded position refinement under a complex loss to recover absolute phase) that locks in the magnitude performance and adds phase fidelity on top.

\textbf{LiDAR-dependent initialization.} The point cloud that seeds 3DPS comes from a co-registered LiDAR scan. This is a strong prior that simplifies geometric initialization and accelerates convergence to a reasonable scene fit. Pure radar-driven NVS --- removing the LiDAR dependence --- is an important direction for radar-only deployments and an open challenge for our framework.

\textbf{Dynamic scenes and downstream perception.} Extending range-Doppler-coherent rendering to scenes with moving targets, and validating NVS-augmented training data on radar-trained perception tasks (object detection, semantic segmentation), are natural next steps that the product-agnostic complex-output property of 3DPS enables but does not directly address.

\FloatBarrier
\section{Ablations}
\label{app:ablations}

We ablate every design choice of 3DPS that does not follow directly from the 3D Gaussian Splatting literature: point count $N$, the multi-frame training-view budget, our adaptive density-control schedule, each of the four LiDAR-initialisation stages (azimuth-cone cull / occlusion ray-cast / cosine-weighted resample / farthest-point sampling), the quasi-static MIMO factorisation that fuses the per-(TX,\,RX) BSDF + range-splat into a single CUDA kernel, the Hann-PSF kernel length $L$, the carrier-phase detach used by the position gradient, and the L2 anchor strength $\lambda_{\mathrm{pos}}$ that ties the optimised positions to the LiDAR seed. All ablations train for the same 500 Adam iterations on the same 6 ColoRadar scenes used in the main paper, with all other hyper-parameters held at the 3DPS defaults (Table~\ref{tab:ablations}, top row, bolded).

\begin{table}[h]
\centering
\caption{\textbf{Ablation sweep on the 6-scene benchmark.} Mean held-out test $|\mathrm{RA}|$ metrics, test $|\mathrm{CRP}|$ correlation, ADC envelope correlation, mean training $|\mathrm{RA}|$ correlation, and mean wall-clock per scene on a single RTX~4090. The \textbf{3DPS default} (gray reference row) is $N{=}20$k points, 8 training views, adaptive density on, full 4-stage LiDAR initialization, MIMO-factored CUDA kernels, $L{=}15$, carrier-phase detach on, $\lambda_{\mathrm{pos}}{=}100$. Each subsequent row flips one design choice. \textbf{Bold}: best per column (ties jointly). \underline{Underline}: second-best. Time is informational, not ranked.}
\label{tab:ablations}
\small
\setlength{\tabcolsep}{4pt}
\resizebox{\linewidth}{!}{
\begin{tabular}{l|cccc|cc|c|c}
\toprule
\textbf{Configuration} & \multicolumn{4}{c|}{\textbf{Test $|\mathrm{RA}|$}} & \multicolumn{2}{c|}{\textbf{Complex / Env.}} & \textbf{Train} & \textbf{Time} \\
\cmidrule(lr){2-5}\cmidrule(lr){6-7}
& Corr $\uparrow$ & PSNR $\uparrow$ & SSIM $\uparrow$ & RMSE $\downarrow$ & $|$CRP$|$ Corr $\uparrow$ & ADC env. $\uparrow$ & Corr $\uparrow$ & (min) \\
\midrule
\multicolumn{9}{l}{\textit{Reference}} \\
\rowcolor{lightergray}\textbf{3DPS default} & \underline{0.587} & \textbf{28.8} & \textbf{0.734} & \textbf{0.0369} & 0.603 & 0.613 & 0.812 & 3.0 \\
\midrule
\multicolumn{9}{l}{\textit{Point count $N$}} \\
$N{=}\phantom{0}2{,}000$ & 0.578 & \underline{28.4} & 0.696 & \underline{0.0389} & 0.586 & 0.571 & 0.673 & 4.0 \\
$N{=}\phantom{0}5{,}000$ & 0.571 & 28.0 & 0.691 & 0.0421 & 0.574 & 0.571 & 0.731 & 4.0 \\
$N{=}10{,}000$ & 0.586 & 28.2 & 0.704 & 0.0394 & 0.591 & 0.615 & 0.777 & 2.7 \\
$N{=}50{,}000$ & 0.538 & 27.3 & 0.696 & 0.0439 & 0.565 & 0.602 & 0.858 & 5.1 \\
\midrule
\multicolumn{9}{l}{\textit{Training views}} \\
2 train views & 0.573 & 27.0 & 0.674 & 0.0448 & 0.591 & 0.591 & \underline{0.888} & 0.9 \\
4 train views & 0.575 & \underline{28.4} & 0.717 & 0.0393 & 0.581 & 0.599 & 0.859 & 1.4 \\
6 train views & 0.586 & 28.2 & 0.704 & 0.0393 & 0.584 & 0.605 & 0.838 & 2.6 \\
\midrule
\multicolumn{9}{l}{\textit{Adaptive density}} \\
no adaptive density & 0.553 & 27.0 & 0.638 & 0.0459 & 0.587 & 0.586 & 0.760 & 3.3 \\
\midrule
\multicolumn{9}{l}{\textit{LiDAR initialization stages}} \\
no azimuth-cone cull & 0.556 & 27.0 & 0.665 & 0.0458 & 0.581 & 0.586 & 0.812 & 2.5 \\
no occlusion ray-cast & \textbf{0.600} & 27.7 & 0.697 & 0.0411 & \textbf{0.612} & 0.587 & 0.825 & 3.6 \\
no cosine resample & 0.571 & 27.7 & 0.692 & 0.0417 & 0.585 & 0.590 & 0.812 & 3.5 \\
no FPS (random sub-sample) & 0.542 & 28.1 & 0.717 & 0.0403 & 0.538 & \textbf{0.626} & 0.808 & 3.3 \\
\midrule
\multicolumn{9}{l}{\textit{MIMO factorization}} \\
no MIMO factorization & 0.555 & 27.4 & 0.685 & 0.0432 & 0.576 & 0.587 & 0.813 & 5.3 \\
\midrule
\multicolumn{9}{l}{\textit{PSF kernel $L$}} \\
PSF $L{=}5$ & 0.569 & 28.0 & 0.716 & 0.0402 & 0.590 & 0.610 & 0.812 & 4.8 \\
PSF $L{=}9$ & 0.572 & 28.2 & 0.708 & 0.0399 & 0.590 & 0.577 & 0.814 & 3.4 \\
PSF $L{=}21$ & 0.578 & 27.8 & 0.691 & 0.0413 & 0.589 & 0.595 & 0.813 & 3.6 \\
PSF $L{=}25$ & 0.569 & 27.4 & 0.682 & 0.0433 & 0.605 & 0.574 & 0.814 & 3.7 \\
\midrule
\multicolumn{9}{l}{\textit{Carrier-phase detach}} \\
no carrier-phase detach & 0.537 & 27.1 & 0.676 & 0.0450 & 0.562 & 0.585 & \textbf{0.894} & 4.5 \\
\midrule
\multicolumn{9}{l}{\textit{Position anchor $\lambda_{\mathrm{pos}}$}} \\
$\lambda_{\mathrm{pos}}{=}0$ & 0.584 & 28.1 & \underline{0.725} & 0.0398 & 0.606 & 0.594 & 0.821 & 2.3 \\
$\lambda_{\mathrm{pos}}{=}1$ & 0.584 & 27.7 & 0.710 & 0.0419 & \underline{0.608} & \underline{0.616} & 0.819 & 3.7 \\
$\lambda_{\mathrm{pos}}{=}1000$ & 0.566 & 28.2 & 0.707 & 0.0394 & 0.590 & 0.584 & 0.822 & 3.3 \\
\bottomrule
\end{tabular}
}
\end{table}

\paragraph{Point count $N$.} Test $|\mathrm{RA}|$ correlation is essentially saturated by $N{=}10$k (0.586) and peaks at the $N{=}20$k default (0.587); $N{=}50$k regresses to 0.538 because the adaptive split/prune schedule begins to displace already-converged points. The implication is that 20k is near-optimal rather than merely past the knee, and that 5k--10k remain viable choices for memory-constrained deployments.

\paragraph{Number of training views.} 3DPS is sample-efficient at the view axis: 2 / 4 / 6 / 8 (default) views land within 0.014 of one another on test correlation (0.573, 0.575, 0.586, 0.587). Wall-clock scales linearly (0.9, 1.4, 2.6, 3.0 minutes per scene) but quality does not. Two adjacent frames are therefore enough when the test view lies between them; wider spacing costs correlation (Sec.~\ref{app:per_scene_ra}).

\paragraph{Adaptive density control.} Disabling the split and prune schedule at iterations 100, 200, 300 and 400 drops test correlation by $0.034$ (0.553 vs.~0.587). The CRP and ADC envelope numbers change less (0.587, 0.586). 3DGS-style densification therefore carries over to radar.

\paragraph{LiDAR initialization stages.} Of the four initialization stages, three contribute: azimuth-cone cull ($-0.031$ when removed), cosine-weighted resample ($-0.016$), and farthest-point sampling ($-0.045$). Removing the Mitsuba-based occlusion ray-cast instead \emph{improves} test correlation by $+0.013$ (0.600 vs.~0.587) while saving $\sim$2\,s of initialization time per scene. The likely cause is that adaptive density with the position-gradient split criterion already prunes occluded points, while the explicit ray-cast over-rejects partially visible points. We retain occlusion ray-casting to keep the LiDAR-prior pipeline aligned with the rendering literature.

\paragraph{Quasi-static MIMO factorisation.} Replacing the fused CUDA BSDF and range-splat kernels with an unfactored implementation that materialises the full $(M, N_{TX}, N_{RX})$ BSDF and $(\text{spread}, M\,N_{TX}\,N_{RX})$ splat tensors raises per-scene wall-clock from 3.0 to 5.3 minutes ($1.8\times$) and drops test correlation from 0.587 to 0.555. The factorisation is therefore both faster \emph{and} numerically more stable in the backward pass, since the fused kernel uses an analytical gradient while the unfactored implementation relies on automatic differentiation through the expanded tensors.

\paragraph{Hann-PSF kernel length $L$.} The sweep $\{5, 9, 15, 21, 25\}$ produces a clean unimodal curve with the peak at the $L{=}15$ default (0.587). Smaller (0.569, 0.572) and larger (0.578, 0.569) all underperform by $0.009$--$0.018$. A small $L$ underspreads the per-(TX, RX) range bins and aliases nearby scatterers; a large $L$ adds little, because the tails are already small. The unimodal shape supports our derivation that the Hann window's discrete-time impulse response decays into the noise floor at $|n|{>}\sim 7$ bins, making $L{=}15$ ($\pm 7$ neighbours) the natural truncation.

\paragraph{Carrier-phase detach.} Allowing position gradients to flow through the carrier phase $\phi_{\text{carrier}} = 2\pi f_0 (d_{\text{TX}} + d_{\text{RX}})/c$ rather than detaching it drops test correlation by $0.050$ (0.537 vs.~0.587), the largest single ablation effect. The wavelength at 77\,GHz is 3.9\,mm, so a sub-mm position update produces a $\mathcal{O}(1)$ phase rotation: the loss landscape becomes locally periodic and Adam oscillates rather than converging. Detaching the carrier phase preserves only the amplitude-path gradient (geometric path length, antenna pattern, BSDF), which is smooth in position. This supports detaching the phase in the fused splat kernel.

\paragraph{L2 position anchor $\lambda_{\mathrm{pos}}$.} The sweep $\{0, 1, 100, 1000\}$ shows that the L2 anchor is a soft regulariser: $\lambda_{\mathrm{pos}} = 0$ and $\lambda_{\mathrm{pos}} = 1$ both yield 0.584, essentially indistinguishable from the $\lambda_{\mathrm{pos}}{=}100$ default (0.587). Pushing to $\lambda_{\mathrm{pos}} = 1000$ over-constrains positions toward the LiDAR seed and degrades fit to 0.566. The amplitude-path-only gradient combined with the small position learning rate ($10^{-5}$) already limits position drift; the explicit anchor is largely redundant at the default settings but provides a guard against high-$N$ regimes where individual points carry less data and can drift.

\paragraph{Seed-normal perturbation.} Rotating each seed normal by an angle drawn from $\mathcal{N}(0,\sigma)$ gives held-out correlation 0.586, 0.580, 0.592, 0.584 at $\sigma=0^\circ, 5^\circ, 10^\circ, 20^\circ$, within the pipeline's run-to-run spread (per-scene s.d.\ 0.031).

\paragraph{Summary.} Of the eleven ablated axes, five are load-bearing for the $|\mathrm{RA}|$ result ($\Delta \geq 0.03$): adaptive density, FPS-stage initialization, cull-stage initialization, MIMO factorisation, and carrier-phase detach. Three are mild contributors ($0.01 \leq \Delta < 0.03$): cosine-resample initialization, PSF length $L$, and point count below 20k. Two are essentially neutral: number of training views (provided $\geq 2$) and $\lambda_{\mathrm{pos}}$ (anywhere in $[0, 100]$). Removing the occlusion stage changes correlation by $+0.013$, so that stage is not required for this result.

\FloatBarrier
\section{Per-point material visualizations}
\label{app:materials}

3DPS optimizes a 6-vector ITU-R~P.2040 material per oriented point ($\varepsilon_r'$, $\varepsilon_r''$, $\sigma_h$, $\ell_c$, $\tau$, $d$; Sec.~\ref{app:bsdf_full}) jointly with the magnitude-only $|\mathrm{RA}|$ loss. Because the parameters are physically meaningful per point, the optimized scene is interpretable as a per-point material map rather than as opaque learned features. We visualize the material content three ways below: (i)~a single ``overall change'' heatmap per scene quantifying per-point deviation from the ITU concrete prior, (ii)~a side-by-side initial-vs-optimized comparison showing what the 8-viewpoint training actually moved, and (iii)~a per-parameter breakdown across all six ITU axes for each scene. All renders use the optimized scenes behind Table~\ref{tab:results}, the inferno colormap on per-parameter physics ranges (linear for sigmoid parameters $\varepsilon_r'$ and $\tau$; log-scale for $\varepsilon_r''$, $\sigma_h$, $\ell_c$, $d$; bounds in Sec.~\ref{app:bsdf_full}), and the LiDAR scaffold mesh as faint grey context geometry. Per-scene viewpoints follow the same scheme used for the qualitative supplement figures.

\begin{figure}[h]
\centering
\includegraphics[width=\linewidth]{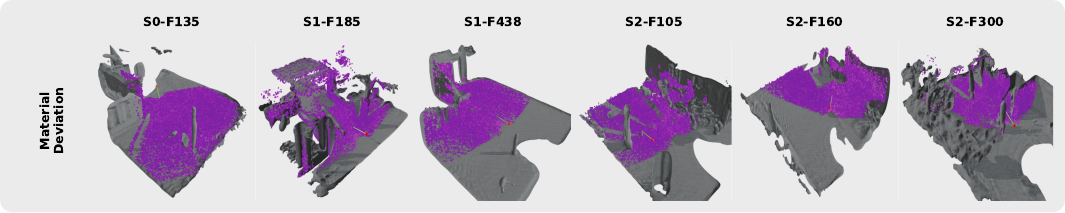}
\caption{\textbf{Per-point material deviation from ITU concrete defaults} (single row, six ColoRadar scenes). Inferno colormap on the magnitude-of-deviation in physics-space (mean across the 6 ITU parameters; log-space for $\varepsilon_r''$, $\sigma_h$, $\ell_c$, $d$ and linear-space for $\varepsilon_r'$, $\tau$). Brighter = larger deviation. Most points settle close to ITU concrete (the seed prior), with localized hot regions corresponding to strong specular returns where the optimizer pushed materials away from the prior to fit the radar response.}
\label{fig:materials_deviation}
\end{figure}

\begin{figure}[h]
\centering
\includegraphics[width=\linewidth]{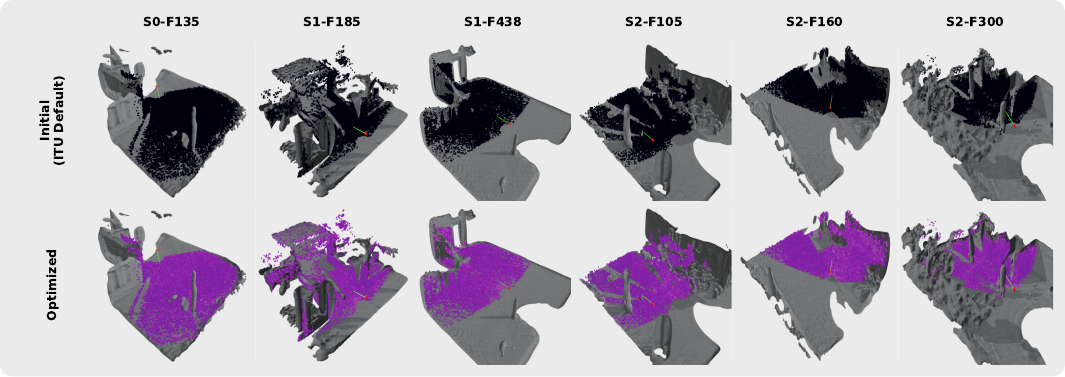}
\caption{\textbf{Initial vs.\ optimized per-point materials.} Top row: every point initialized to ITU concrete (uniform purple = zero deviation). Bottom row: per-point materials after 500 Adam iterations on the 8-viewpoint $|\mathrm{RA}|$ loss. The contrast between rows visualizes \emph{which} regions the optimizer chose to move away from the concrete prior --- predominantly the boresight strip --- consistent with the cosine-weighted resample (Sec.~\ref{app:lidar_reduction}) concentrating points where the antenna pattern weights them most.}
\label{fig:materials_init_vs_opt}
\end{figure}

\begin{figure}[h]
\centering
\includegraphics[width=\linewidth]{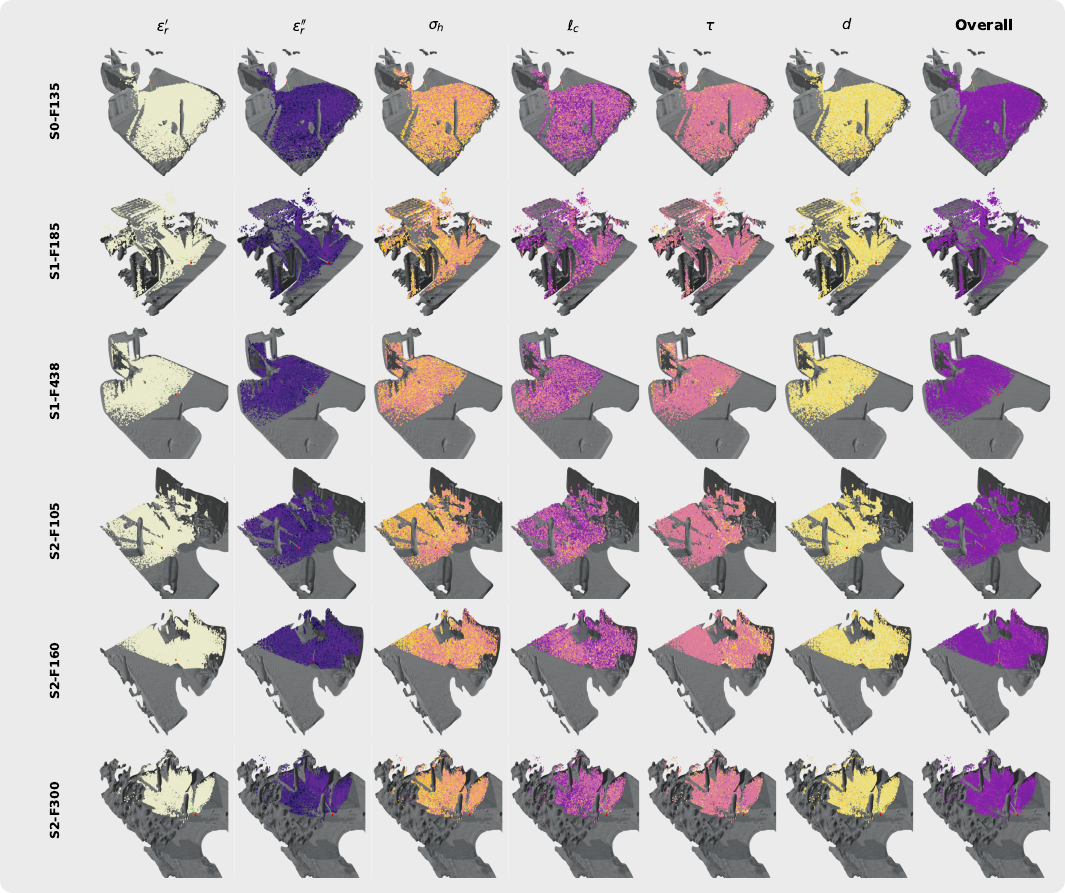}
\caption{\textbf{Per-parameter optimized materials, all six ITU-R~P.2040 axes.} Rows: six ColoRadar scenes. Columns: $\varepsilon_r'$, $\varepsilon_r''$, $\sigma_h$, $\ell_c$, $\tau$, $d$, plus an overall-deviation column on the right (matching Fig.~\ref{fig:materials_deviation}). Inferno colormap with per-parameter normalization to the ITU-R~P.2040 valid bounds (Sec.~\ref{app:bsdf_full}); linear scale for sigmoid parameters ($\varepsilon_r'$, $\tau$) and log scale for the others. Brighter = higher value. Most points stay near the prior (median $\varepsilon_r'$ 5.3; 5--95\% range 4.7--5.9).}
\label{fig:materials_optimized}
\end{figure}

\FloatBarrier
\subsection{Per-point surface normals}
\label{app:normals}

3DPS also optimizes a per-point quaternion that orients the local surface frame; the surface normal is the rotation of $+\hat{\mathbf{z}}$ by that quaternion (seeded from LiDAR). The quaternion is trainable so the optimizer can refine the LiDAR-derived normals against the magnitude-only $|\mathrm{RA}|$ loss; the figures below show what that refinement does. We use the same 6-scene benchmark and viewpoint convention as the material visualizations (Sec.~\ref{app:materials}).

\begin{figure}[h]
\centering
\includegraphics[width=\linewidth]{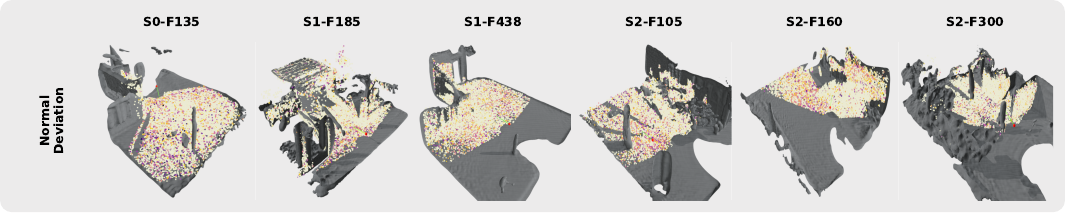}
\caption{\textbf{Per-point normal deviation from the LiDAR seed} (single row, six ColoRadar scenes). Inferno colormap on per-point angular deviation in degrees, clipped to $[0^\circ, 30^\circ]$ (dark = LiDAR-aligned, bright yellow = $\geq 30^\circ$ rotation). Initial-state normals come from nearest-vertex lookup into the LiDAR mesh at each optimized point's position. Most points stay within $\sim$10$^\circ$ of the seed (consistent with LiDAR normal noise), with localized hot regions where the radar response disagrees with the LiDAR-derived normal --- e.g., on smooth specular surfaces (planar walls, signage) where the LiDAR's local-plane fit averages over surface fine-structure that the radar return is sensitive to.}
\label{fig:normals_deviation}
\end{figure}

\begin{figure}[h]
\centering
\includegraphics[width=\linewidth]{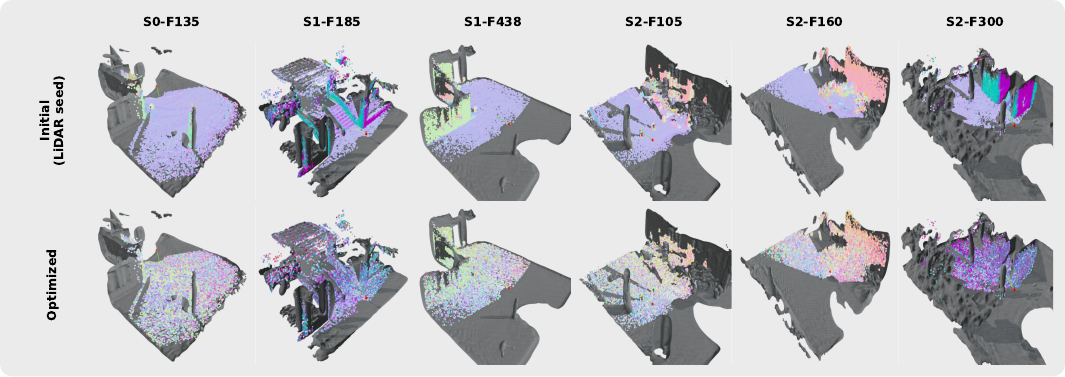}
\caption{\textbf{Initial vs.\ optimized per-point normals, RGB-encoded.} World-space normal-map encoding $((\mathbf{n}+\mathbf{1})/2 \to \mathrm{RGB})$ with back-facing normals camera-flip-corrected for visual consistency. Top row: initial state (LiDAR mesh nearest-neighbour at each optimized point's position). Bottom row: learned state (per-point quaternion $\to$ normal after 500 Adam iterations). The visual contrast between rows tracks Fig.~\ref{fig:normals_deviation}: the optimizer adds high-frequency normal variation to the locally-flat LiDAR seed, predominantly on the boresight strip where the antenna pattern weights points most heavily.}
\label{fig:normals_init_vs_opt}
\end{figure}

\begin{figure}[h]
\centering
\includegraphics[width=\linewidth]{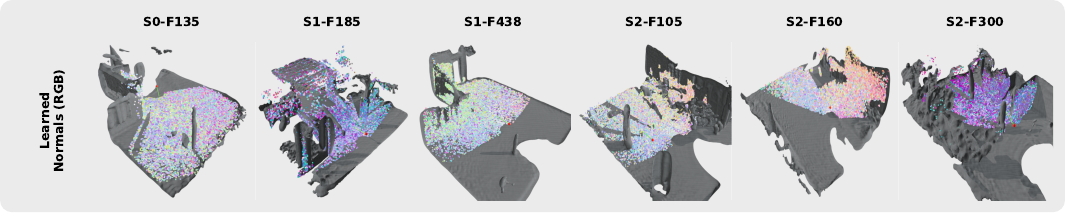}
\caption{\textbf{Learned per-point normals only}, RGB-encoded as in Fig.~\ref{fig:normals_init_vs_opt} (single row, six scenes). Provided alongside Fig.~\ref{fig:normals_init_vs_opt} for direct visual reference of the optimized normal field at full per-scene scale.}
\label{fig:normals_learned}
\end{figure}

\FloatBarrier
\section{Direction of arrival from rendered data}
\label{app:doa}

On held-out frames, we replace the azimuth FFT with MUSIC and ESPRIT on the 86-element virtual-array snapshot at the strongest GT range bins, with forward-backward smoothing over 48-element sub-apertures. Median angular error in beamwidths: 3DPS 2.84 (MUSIC), 2.39 (ESPRIT); the same render with its phase set to zero, 9.76 and 10.03. Optical-NVS baselines cannot run these estimators, because they lack per-antenna complex data. The render therefore carries recoverable angular information, but not deployment-grade DOA.

\FloatBarrier
\section{Broader impact and dual-use considerations}
\label{app:broader_impact}

3DPS targets sparse-view novel view synthesis for mmWave radar in static outdoor scenes. The intended applications are radar simulation, sensor-driven scene understanding, and data augmentation for downstream perception tasks such as object detection and free-space estimation in autonomous driving and robotics. By making radar data more consistent across viewpoints and reducing the data-collection cost of new deployments, 3DPS lowers the barrier to safety-critical evaluation of radar perception systems and supports sim-to-real transfer for robust autonomy.

\textbf{Dual-use considerations.} mmWave radar is also used in surveillance and military sensing systems. The methods presented here apply, in principle, to these settings, but the inputs we require (a co-registered LiDAR scan, full radar array geometry, and dense pose annotations) are not generally available in covert deployments, and our contribution does not advance any covert sensing capability beyond what is already achievable with conventional radar simulation tools. We do not provide trained models or reconstructions of any sensitive site, and the released code is intended for noncommercial research on public datasets such as ColoRadar~\cite{kramer2022coloradar}.

\textbf{Failure modes and deployment caveats.} 3DPS is not a substitute for direct radar measurement. The single-bounce assumption limits accuracy in indoor multipath-dominated environments, and the LiDAR-conditioned scaffold means failure of the scaffold (poor calibration, missing scan coverage) propagates to the renderer output. Users deploying 3DPS in safety-critical settings should treat its output as supplementary to, not replacing, real radar acquisitions. We list further limitations in Sec.~\ref{app:limitations}.

\textbf{Energy and compute footprint.} A complete training run for one scene takes approximately 3 minutes on a single NVIDIA RTX~4090 (Sec.~\ref{app:runtime}). The full set of experiments reported in this paper consumed less than 100 GPU hours of compute.

\clearpage
\section*{NeurIPS Paper Checklist}

\begin{enumerate}

\item {\bf Claims}
    \item[] Question: Do the main claims made in the abstract and introduction accurately reflect the paper's contributions and scope?
    \item[] Answer: \answerYes{}
    \item[] Justification: The abstract and Sec.~\ref{sec:intro} state our three contributions (the first differentiable point splatting renderer for radar with closed-form BSDFs, product-agnostic ADC/CRP/RA outputs from the same optimized scene, and sparse-view radar NVS at the state of the art with the highest held-out $|\mathrm{RA}|$ correlation at a $25\times$ per-scene speedup over the closest physics-based baseline, the mmIR ray tracer). The Scope paragraph in Sec.~\ref{sec:intro} explicitly bounds LiDAR initialization and single-bounce path tracing as design choices.

\item {\bf Limitations}
    \item[] Question: Does the paper discuss the limitations of the work performed by the authors?
    \item[] Answer: \answerYes{}
    \item[] Justification: Limitations are summarized in Sec.~\ref{sec:conclusion} and expanded in Sec.~\ref{app:limitations}, covering the single-bounce path-tracing assumption (insufficient for indoor multipath), the magnitude-only training loss (per-RA-bin phase under-constrained), the LiDAR-conditioned scaffold (LiDAR-free radar NVS as future work), and dynamic-scene extension.

\item {\bf Theory assumptions and proofs}
    \item[] Question: For each theoretical result, does the paper provide the full set of assumptions and a complete (and correct) proof?
    \item[] Answer: \answerNA{}
    \item[] Justification: The paper presents a forward-model derivation (Sec.~\ref{sec:methods:fwd}, full surface-to-hemisphere derivation in Sec.~\ref{app:fwd_derivation}) rather than new theoretical results requiring formal proofs. All equations follow from textbook radar signal processing and rendering literature, with explicit citations.

\item {\bf Experimental result reproducibility}
    \item[] Question: Does the paper fully disclose all the information needed to reproduce the main experimental results of the paper to the extent that it affects the main claims and/or conclusions of the paper (regardless of whether the code and data are provided or not)?
    \item[] Answer: \answerYes{}
    \item[] Justification: Sec.~\ref{sec:methods:impl} specifies the full optimization recipe (Adam parameters, per-parameter learning rates, density-control schedule, fused CUDA kernel structure, carrier-phase detach, L2 anchor strength). Sec.~\ref{app:lidar_reduction} details LiDAR reduction parameters, Sec.~\ref{app:bsdf_full} provides the full BSDF closed form, and Sec.~\ref{app:radarsplat_baseline} documents RadarSplat hyperparameter selection. Code and configuration files are available at \url{https://3d-point-splatting.github.io} (Sec.~\ref{sec:conclusion}).

\item {\bf Open access to data and code}
    \item[] Question: Does the paper provide open access to the data and code, with sufficient instructions to faithfully reproduce the main experimental results, as described in supplemental material?
    \item[] Answer: \answerYes{}
    \item[] Justification: The ColoRadar dataset~\cite{kramer2022coloradar} is publicly available at \url{https://arpg.github.io/coloradar/}. Code, configurations, and all analysis scripts that generated every table and figure in the paper are available at \url{https://3d-point-splatting.github.io}.

\item {\bf Experimental setting/details}
    \item[] Question: Does the paper specify all the training and test details (e.g., data splits, hyperparameters, how they were chosen, type of optimizer) necessary to understand the results?
    \item[] Answer: \answerYes{}
    \item[] Justification: Sec.~\ref{sec:experiments:setup} specifies the scene splits (held-out middle frame plus eight bracketing training frames at 5\,Hz, NVS-inferred test pose), baseline grouping by family, metric definitions, and 3DPS implementation. Sec.~\ref{app:radarsplat_baseline} documents RadarSplat hyperparameter selection.

\item {\bf Experiment statistical significance}
    \item[] Question: Does the paper report error bars suitably and correctly defined or other appropriate information about the statistical significance of the experiments?
    \item[] Answer: \answerYes{}
    \item[] Justification: Per-method per-metric standard deviations across the six ColoRadar scenes are reported in supplement Table~\ref{tab:test_ra} (held-out test, Std row) and Table~\ref{tab:train_ra} (training views, Std row), and aggregate standard deviations for 3DPS in Tables~\ref{tab:crp_adc_train} and~\ref{tab:crp_adc_test} (CRP/ADC, Std row). Std is computed as the sample standard deviation (Bessel's correction) over the six per-scene values, capturing scene-to-scene variability.

\item {\bf Experiments compute resources}
    \item[] Question: For each experiment, does the paper provide sufficient information on the computer resources (type of compute workers, memory, time of execution) needed to reproduce the experiments?
    \item[] Answer: \answerYes{}
    \item[] Justification: Sec.~\ref{sec:experiments:setup} reports per-scene wall-clock for 3DPS and the baselines on a single NVIDIA RTX~4090. Sec.~\ref{app:runtime} (Table~\ref{tab:runtime}) reports per-scene min--max wall-clock and peak GPU memory for every method. Sec.~\ref{app:broader_impact} reports the total project compute budget (less than 100 GPU-hours).

\item {\bf Code of ethics}
    \item[] Question: Does the research conducted in the paper conform, in every respect, with the NeurIPS Code of Ethics \url{https://neurips.cc/public/EthicsGuidelines}?
    \item[] Answer: \answerYes{}
    \item[] Justification: We have reviewed and adhere to the NeurIPS Code of Ethics. The paper uses only the publicly-available ColoRadar dataset, does not involve human subjects or scraped data, and discloses dual-use considerations in Sec.~\ref{app:broader_impact}.

\item {\bf Broader impacts}
    \item[] Question: Does the paper discuss both potential positive societal impacts and negative societal impacts of the work performed?
    \item[] Answer: \answerYes{}
    \item[] Justification: Sec.~\ref{app:broader_impact} discusses positive impacts (sim-to-real transfer for radar perception, lower data-collection cost for safety-critical autonomy, support for radar-trained detectors), dual-use considerations (radar surveillance and military sensing), failure modes and deployment caveats (single-bounce limit, LiDAR-scaffold dependence), and the energy and compute footprint of the project.

\item {\bf Safeguards}
    \item[] Question: Does the paper describe safeguards that have been put in place for responsible release of data or models that have a high risk for misuse (e.g., pre-trained language models, image generators, or scraped datasets)?
    \item[] Answer: \answerNA{}
    \item[] Justification: The paper introduces a differentiable forward renderer trained on a public radar dataset. It does not release pre-trained generative models, large language models, image generators, or scraped datasets that pose high misuse risk. Released artifacts are code and configurations for public ColoRadar scenes only (Sec.~\ref{app:broader_impact}).

\item {\bf Licenses for existing assets}
    \item[] Question: Are the creators or original owners of assets (e.g., code, data, models), used in the paper, properly credited and are the license and terms of use explicitly mentioned and properly respected?
    \item[] Answer: \answerYes{}
    \item[] Justification: All assets are cited at first use. The ColoRadar dataset~\cite{kramer2022coloradar} is distributed under the BSD~3-Clause License at \url{https://arpg.github.io/coloradar/}. Mitsuba~3~\cite{jakob2022dr} is distributed under the BSD~3-Clause License. The ITU-R~P.2040 material model~\cite{itu_p2040} is a public ITU-R Recommendation. Baselines (DART~\cite{huang2024dart}, Radar Fields~\cite{10.1145/3641519.3657510}, RadarSplat~\cite{kung2025radarsplat}, mmIR~\cite{mmIR2026}, Sionna RT~\cite{sionna}) and 3D Gaussian Splatting~\cite{3dgs} are used per their respective public licenses, all of which permit academic research use; configuration details are in Sec.~\ref{app:radarsplat_baseline}.

\item {\bf New assets}
    \item[] Question: Are new assets introduced in the paper well documented and is the documentation provided alongside the assets?
    \item[] Answer: \answerYes{}
    \item[] Justification: Code, configuration files, and analysis scripts that generated every table and figure in the paper are available at \url{https://3d-point-splatting.github.io}, with documentation for running the renderer and reproducing each table and figure.

\item {\bf Crowdsourcing and research with human subjects}
    \item[] Question: For crowdsourcing experiments and research with human subjects, does the paper include the full text of instructions given to participants and screenshots, if applicable, as well as details about compensation (if any)?
    \item[] Answer: \answerNA{}
    \item[] Justification: The paper does not involve crowdsourcing or research with human subjects.

\item {\bf Institutional review board (IRB) approvals or equivalent for research with human subjects}
    \item[] Question: Does the paper describe potential risks incurred by study participants, whether such risks were disclosed to the subjects, and whether Institutional Review Board (IRB) approvals (or an equivalent approval/review based on the requirements of your country or institution) were obtained?
    \item[] Answer: \answerNA{}
    \item[] Justification: The paper does not involve research with human subjects.

\item {\bf Declaration of LLM usage}
    \item[] Question: Does the paper describe the usage of LLMs if it is an important, original, or non-standard component of the core methods in this research? Note that if the LLM is used only for writing, editing, or formatting purposes and does \emph{not} impact the core methodology, scientific rigor, or originality of the research, declaration is not required.
    \item[] Answer: \answerNA{}
    \item[] Justification: LLMs were used only for writing, editing, and formatting assistance (manuscript polishing, caption tightening, sentence revision). They did not impact the core methodology, mathematical derivations, algorithm design, or experimental results, so per the NeurIPS LLM policy, declaration is not required.

\end{enumerate}

\end{document}